\documentclass[11pt]{article}

\usepackage[final]{acl}

\usepackage{times}
\usepackage{latexsym}

\usepackage[T1]{fontenc}

\usepackage[utf8]{inputenc}

\usepackage{microtype}

\usepackage{inconsolata}

\usepackage{graphicx}

\usepackage{amsmath}
\usepackage{amsfonts}
\usepackage{amssymb} 
\usepackage{multirow}

\usepackage{subcaption}
\usepackage{booktabs} 
\usepackage{tikz}  
\usetikzlibrary{arrows.meta, positioning, calc}

\usepackage{algorithm}
\usepackage{algpseudocode}

\newcommand{\pnext}{p_\theta}
\newcommand{\anchorTok}{A}
\newcommand{\plusoneTok}{\texttt{+1}}

\usepackage{xcolor}
\usepackage{array}
\usepackage{adjustbox} 

\newlength{\FigGridLabelW}        
\newlength{\FigGridSubLabelW}     
\newlength{\FigGridPlotW}         
\newlength{\FigGridCellH}         

\newlength{\FigGridColSep}        
\newlength{\FigGridLabelSep}      
\newlength{\FigGridSubLabelSep}   

\newlength{\FigGridBoxSep}
\newlength{\FigGridBoxRule}

\colorlet{FigGridBorderColor}{gray!55}
\colorlet{FigGridFillColor}{white}

\newcommand{\FigHdr}{\small\bfseries}
\newcommand{\FigLabelFont}{\small\bfseries}

\newcommand{\FigGridSetup}[1]{%
  \setlength{\tabcolsep}{\FigGridColSep}%
  \setlength{\FigGridPlotW}{%
    \dimexpr(\textwidth-\FigGridLabelW-\FigGridLabelSep-\numexpr2*(#1-1)\relax\tabcolsep)/#1\relax%
  }%
}
\newcommand{\FigGridSetupTwo}[1]{%
  \setlength{\tabcolsep}{\FigGridColSep}%
  \setlength{\FigGridPlotW}{%
    \dimexpr(\textwidth-\FigGridLabelW-\FigGridLabelSep-\FigGridSubLabelW-\FigGridSubLabelSep-\numexpr2*(#1-1)\relax\tabcolsep)/#1\relax%
  }%
}

\newcommand{\FigGridVLabel}[1]{%
  \begin{minipage}[c][\FigGridCellH][c]{\FigGridLabelW}%
    \centering
    \rotatebox[origin=c]{90}{\FigLabelFont\strut #1}%
  \end{minipage}%
}

\newcommand{\FigGridVLabelSpanTwo}[1]{%
  \begin{minipage}[c][\dimexpr2\FigGridCellH\relax][c]{\FigGridLabelW}%
    \centering
    \rotatebox[origin=c]{90}{\FigLabelFont\strut #1}%
  \end{minipage}%
}

\newcommand{\FigGridVSubLabel}[1]{%
  \begin{minipage}[c][\FigGridCellH][c]{\FigGridSubLabelW}%
    \centering
    \rotatebox[origin=c]{90}{\FigLabelFont\strut #1}%
  \end{minipage}%
}

\newcommand{\FigGridCell}[1]{%
  \begingroup
  \setlength{\fboxsep}{\FigGridBoxSep}%
  \setlength{\fboxrule}{\FigGridBoxRule}%
  \fcolorbox{FigGridBorderColor}{FigGridFillColor}{%
    \begin{minipage}[c][\dimexpr\FigGridCellH-2\FigGridBoxSep-2\FigGridBoxRule\relax][c]{%
      \dimexpr\FigGridPlotW-2\FigGridBoxSep-2\FigGridBoxRule\relax}%
      \centering
      \includegraphics[
        width=\linewidth,
        height=\dimexpr\FigGridCellH-2\FigGridBoxSep-2\FigGridBoxRule\relax,
        keepaspectratio
      ]{#1}%
    \end{minipage}%
  }%
  \endgroup
}

\title{Beyond Semantics: How Temporal Biases Shape Retrieval in Transformer and State-Space Models}

\author{
  Anooshka Bajaj\textsuperscript{*} \quad
  Deven Mahesh Mistry\textsuperscript{*} \quad 
  Sahaj Singh Maini \\
  \textbf{Yash Aggarwal \quad Zoran Tiganj} \\
  Department of Computer Science \\
  Indiana University Bloomington \\
  \texttt{\{anobajaj,demistry,sahmaini,yaggarw,ztiganj\}@iu.edu} 
}

\begin{document}
\maketitle

\def\thefootnote{*}\footnotetext{These authors contributed equally to this work.}\def\thefootnote{\arabic{footnote}}

\begin{abstract}
In-context learning depends not only on \emph{what} appears in the prompt but also on \emph{when} it appears. To isolate this temporal component from semantic confounds, we construct prompts with repeated anchor tokens and average the model's predictions over hundreds of random permutations of the intervening context. This approach ensures that any observed position-dependent effects are driven purely by temporal structure rather than token identity or local semantics.
Across four transformer LLMs and three state-space/recurrent models, we observe a robust serial recall signature: models allocate disproportionate probability mass to the tokens that previously followed the anchor, but the strength of this signal is modulated by serial position, yielding model-specific primacy/recency profiles. We then introduce an overlapping-episode probe in which only a short cue from one episode is re-presented; retrieval is reliably weakest for episodes embedded in the middle of the prompt, consistent with ``lost-in-the-middle'' behavior.
Mechanistically, ablating high-induction-score attention heads in transformers reduces serial recall and episodic separation. For state-space models, ablating a small fraction of high-attribution channels produces analogous degradations, suggesting a sparse subspace supporting induction-style copying.
Together, these results clarify how temporal biases shape retrieval across architectures and provide controlled probes for studying long-context behavior.
\end{abstract}

\section{Introduction}

The remarkable ability of Large Language Models (LLMs) for in-context learning (ICL) allows them to adapt to new tasks using only the information provided within the input prompt, without explicit parameter updates \citep{brown2020language}. However, while much research has focused on their semantic processing and reasoning capabilities, the mechanisms governing how LLMs retrieve and utilize contextual information, particularly concerning its temporal structure, remain less understood. It is increasingly recognized that the temporal position of information within the context significantly influences retrieval \citep{liu2024lost}. In particular, models often exhibit better recall for information presented at the very beginning or end of the input context, a phenomenon termed the ``lost in the middle'' effect \citep{liu2024lost}, mirroring the well-documented primacy and recency effects observed in human memory studies \citep{murdock1962serial, ebbinghaus1913memory}.

\begin{figure*}[h!]
\centering
\includegraphics[width=1\textwidth]{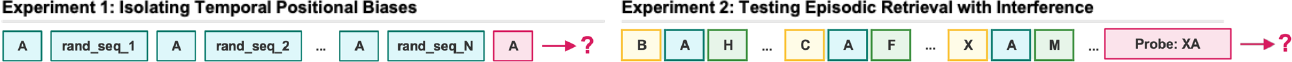}
\caption{
\textbf{Schematic of the experiments.}
\textbf{Left:} Experiment~1 (serial recall). Prompts contain repeated occurrences of an anchor token $\anchorTok$ separated by random gap tokens.
Across many permutations of the gap tokens, we measure the probability assigned to each successor (\plusoneTok) token to isolate position-dependent retrieval biases (primacy/recency) from semantic confounds.
\textbf{Right:} Experiment~2 (episodic retrieval with overlap). Five episodes $(C_i,\anchorTok,T_i)$ are embedded in long random fillers; the prompt ends with a probe pair $(C_k,\anchorTok)$.
Successful retrieval corresponds to assigning high probability to the matching target token $T_k$ while suppressing targets from the non-probed episodes.
}

\label{fig:schematic} 
\end{figure*}

This parallel extends to the principles of human episodic memory, where the temporal organization of experiences is fundamental for segregating and retrieving specific past events, even when they share semantic content. 
This ability relies on encoding not just \emph{what} happened, but \emph{when} it happened relative to other events, enabling the formation and recall of distinct episodes \citep{howard2002contextual, kahana1996associative}. Recent work has begun exploring analogous processes in LLMs, investigating how they might implement memory-like functions to track and utilize temporal context \citep{ji2024linking, mistry2025emergence, pink2024assessing}.

Mechanistically, specific components within transformer architectures, known as ``induction heads'', have been identified as crucial for ICL \citep{olsson2022context, elhage2021mathematical}. These heads operate by finding previous occurrences of a current token and attending to the token that followed it, effectively learning and reproducing sequences based on temporal association \citep{olsson2022context, singh2024transient}. Ablation studies confirm their essential role, demonstrating that removing induction heads significantly impairs ICL capabilities and temporal dependency processing, such as the tendency for serial recall (predicting the token immediately following a repeated token) \citep{mistry2025emergence, crosbie2024induction}. Furthermore, the behavior of these induction heads exhibits characteristics reminiscent of human episodic memory recall, including temporal contiguity (recalling items presented close together in time) and forward asymmetry (preferential recall in the forward direction) \citep{ji2024linking, mistry2025emergence, kahana1996associative}. This connection has been formalized by linking induction head mechanisms to computational models of human memory like the Context Maintenance and Retrieval (CMR) model \citep{ji2024linking, polyn2009context, howard2002contextual}.

While semantic relationships in LLM retrieval are critical, temporal separation is essential for effective context use because semantics alone cannot disambiguate repeated content. Recent work has begun to map these dynamics, drawing connections between LLM behavior and serial-position effects like primacy and recency in both transformers and state-space models \citep{janik2023aspects,morita2025primacy_ssm,airlangga2025primacy_recency_mamba}. However, differentiating pure temporal bias from semantic confounds remains a challenge. Our contribution addresses this by introducing a permutation-based probe that controls the marginal token distribution at each position, ensuring that observed modulations cannot be attributed to token identity or local semantic regularity. We further connect these biases to causal mechanisms via ablations, grounding broader phenomena of position bias in specific architectural ingredients \citep{wu2025position_bias}.
We apply this methodology to a diverse range of architectures, including transformers and state-space models (SSMs), prompting them with sequences containing multiple, temporally distinct presentations of the same tokens. By fixing the positions of these repeated tokens and analyzing next-token predictions across permutations of intervening tokens, we isolate the model's ability to use pure temporal cues for retrieval. Our findings reveal consistent temporal biases across all tested models, characterized by a distinct preference for information near the beginning or end of the context. In transformers, we mechanistically link this behavior to induction heads; remarkably, SSMs exhibit comparable biases despite their architectural differences, suggesting these are fundamental properties of sequential processing. These results expand our understanding of ICL dynamics, illustrating how temporal biases—analogous to principles of human memory—enable episodic separation within LLMs.

\section{Experiments}
To characterize temporal biases and the ability to retrieve temporal context, we conducted two experiments evaluating seven open-weight LLMs using distinct prompt structures. The selected models represent both the transformer architecture \citep{vaswani2017attention} and the SSM  architecture \citep{gu2023mamba}.

Models based on the transformer architecture include Llama-3.1-8B-Instruct \citep{dubey2024llama}, Gemma-2-9b-it \citep{team2024gemma}, Mistral-7B-Instruct-v0.1 \citep{jiang2023mistral}, and Qwen2.5-7B-Instruct \citep{yang2024qwen2}. The SSM-based models include mamba-130m-hf \citep{gu2023mamba}, Falcon3-Mamba-7B-Instruct \citep{zuo2024falcon}, and Recurrent-Gemma-9b-it \citep{botev2024recurrentgemma}.


\subsection{Permutation-controlled prompt family and scoring}
\label{sec:permutation_control}

\paragraph{Why permutations}
Token identity and frequency priors do influence next-token probabilities, even without context.
Our goal is \emph{not} to eliminate these priors; rather, we design prompts so that such priors are \emph{constant across positions}
in expectation. Under this construction, systematic differences between early and late retrieval signals are attributable to temporal
structure (e.g., position bias, recency/primacy) rather than particular tokens occurring at particular positions.

\paragraph{Sampling scheme}
For each prompt instance, we sample a single anchor token $\anchorTok$ uniformly from the vocabulary. All non-anchor tokens are sampled uniformly from the remaining pool, excluding $\anchorTok$.
Within each gap (the tokens between anchor occurrences), we sample without replacement so the marginal distribution of the token at any fixed within-gap position is identical across positions.

\paragraph{Position-indexed probability profile}
Given a prompt $x=(x_1,\ldots,x_T)$ and the model next-token distribution $\pnext(\cdot\mid x)$, we define a \emph{position-indexed probability profile} by mapping each position $t$ to the probability assigned to the token that actually appears
at that position:
\[
\pi_t(x) \;=\; \pnext(x_t \mid x).
\]
Our figures plot $\pi_t(x)$ against token position $t$. Throughout, vertical gray lines indicate anchor occurrences.

\paragraph{\plusoneTok\ tokens}
If anchor occurrences are at positions $\{a_1,\ldots,a_N\}$, we call the successor tokens $x_{a_i+1}$ the \plusoneTok\ tokens.
We summarize serial recall with
\[
p_i^{(+1)} \;=\; \pnext(x_{a_i+1} \mid x),
\]
and report means and standard errors across $M$ independently sampled prompts (typically $M=5000$).

\begin{algorithm}[t]
\caption{Experiment 1: Serial recall probe}
\label{alg:perm_serial}
\small
\begin{algorithmic}[1]
\Require model $\pnext$, repeats $N$, gap length $L$, permutations $M$
\For{$m=1$ to $M$}
  \State sample anchor $\anchorTok$ uniformly from allowed tokens
  \For{$i=1$ to $N$}
    \State sample gap tokens $r_{i,1:L}$ uniformly without replacement from allowed tokens excluding $\anchorTok$
  \EndFor
  \State $x \gets (\anchorTok, r_{1,1:L}, \anchorTok, r_{2,1:L}, \ldots, \anchorTok, r_{N,1:L}, \anchorTok)$ 
  \State compute $\pnext(\cdot\mid x)$ and record $p_i^{(+1)}=\pnext(r_{i,1}\mid x)$
\EndFor
\State \Return $\{p_i^{(+1)}\}_{i=1}^N$ statistics across permutations
\end{algorithmic}
\end{algorithm}

\begin{figure*}[h!]
\centering
\begingroup
\setlength{\FigGridLabelW}{0.65cm}
\setlength{\FigGridCellH}{2.35cm} 
\setlength{\FigGridLabelSep}{1pt}
\setlength{\FigGridColSep}{0.5pt}

\FigGridSetup{5}
\renewcommand{\arraystretch}{1.0}

\begin{tabular}{@{}>{\centering\arraybackslash}m{\FigGridLabelW}@{\hspace{\FigGridLabelSep}}*{5}{>{\centering\arraybackslash}m{\FigGridPlotW}}@{}}
\toprule
& \multicolumn{5}{c}{\FigHdr \# Repeats} \\
\cmidrule(lr){2-6}
& \FigHdr 3 & \FigHdr 5 & \FigHdr 10 & \FigHdr 20 & \FigHdr 40 \\
\midrule

\FigGridVLabel{Llama} &
\FigGridCell{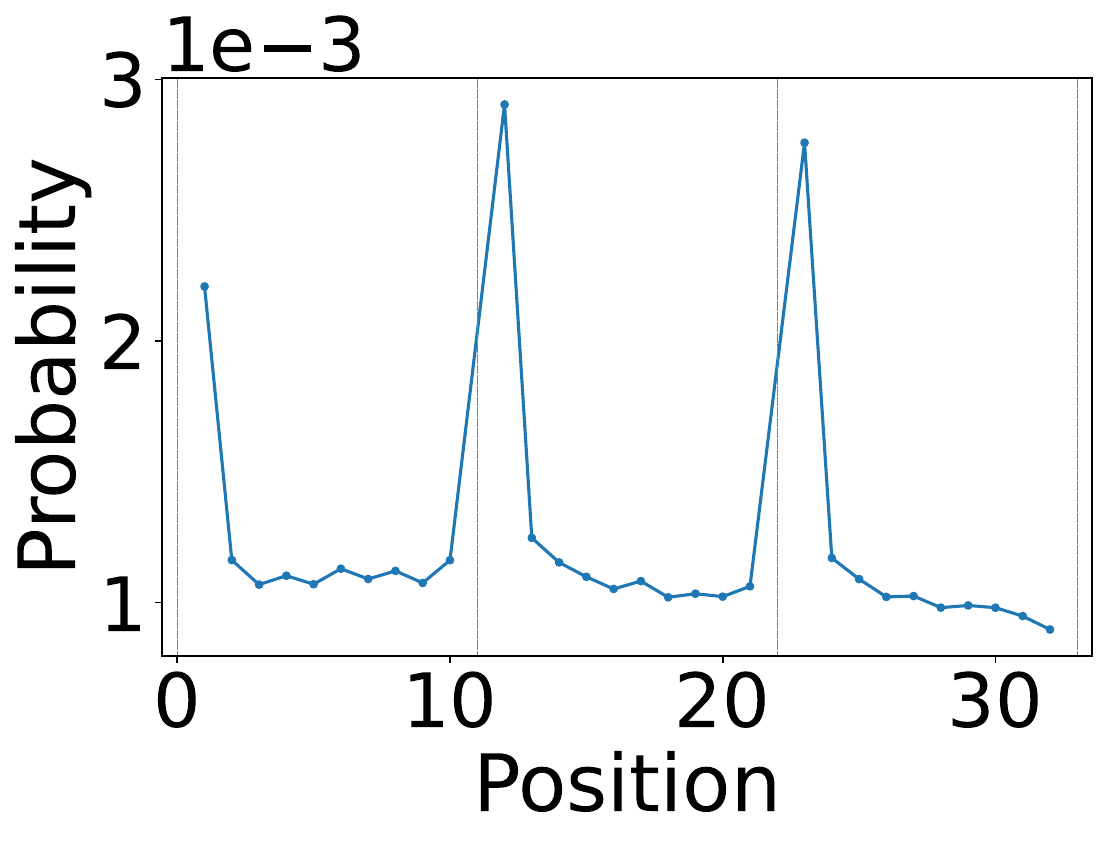} &
\FigGridCell{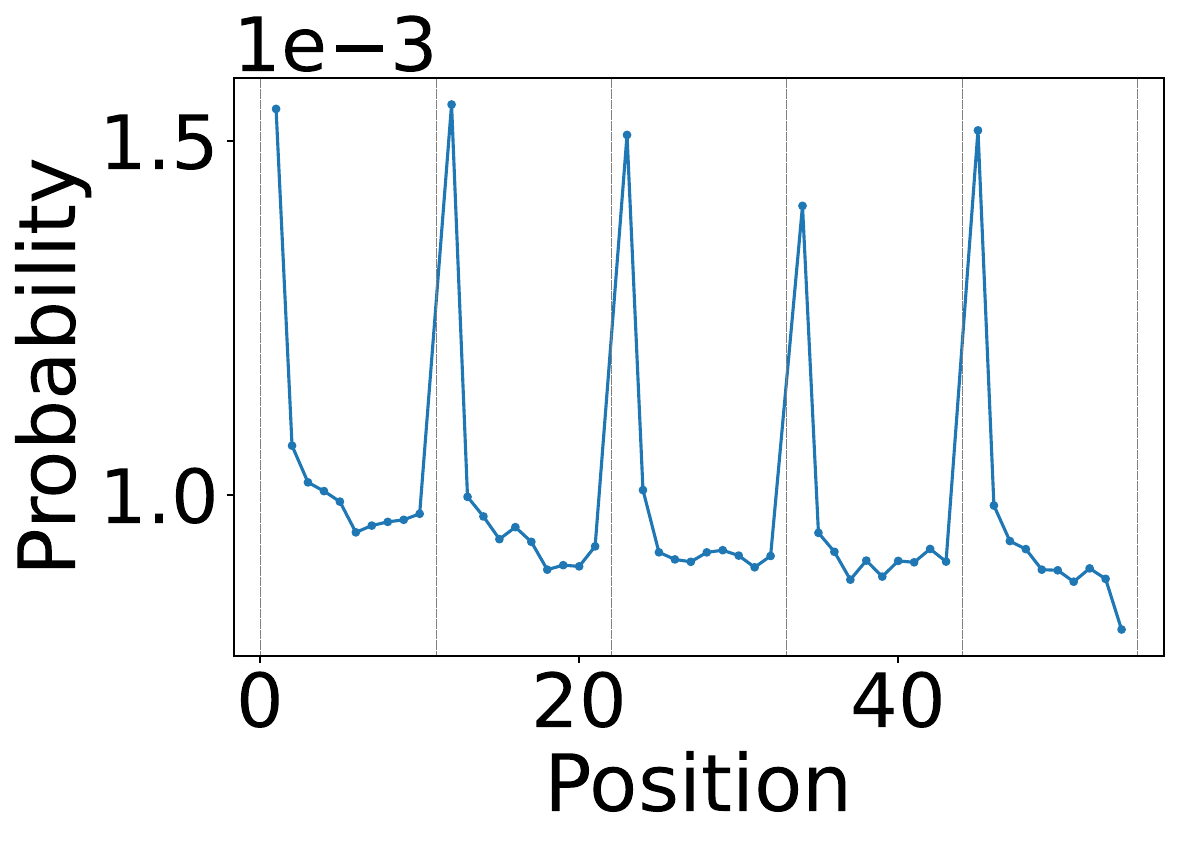} &
\FigGridCell{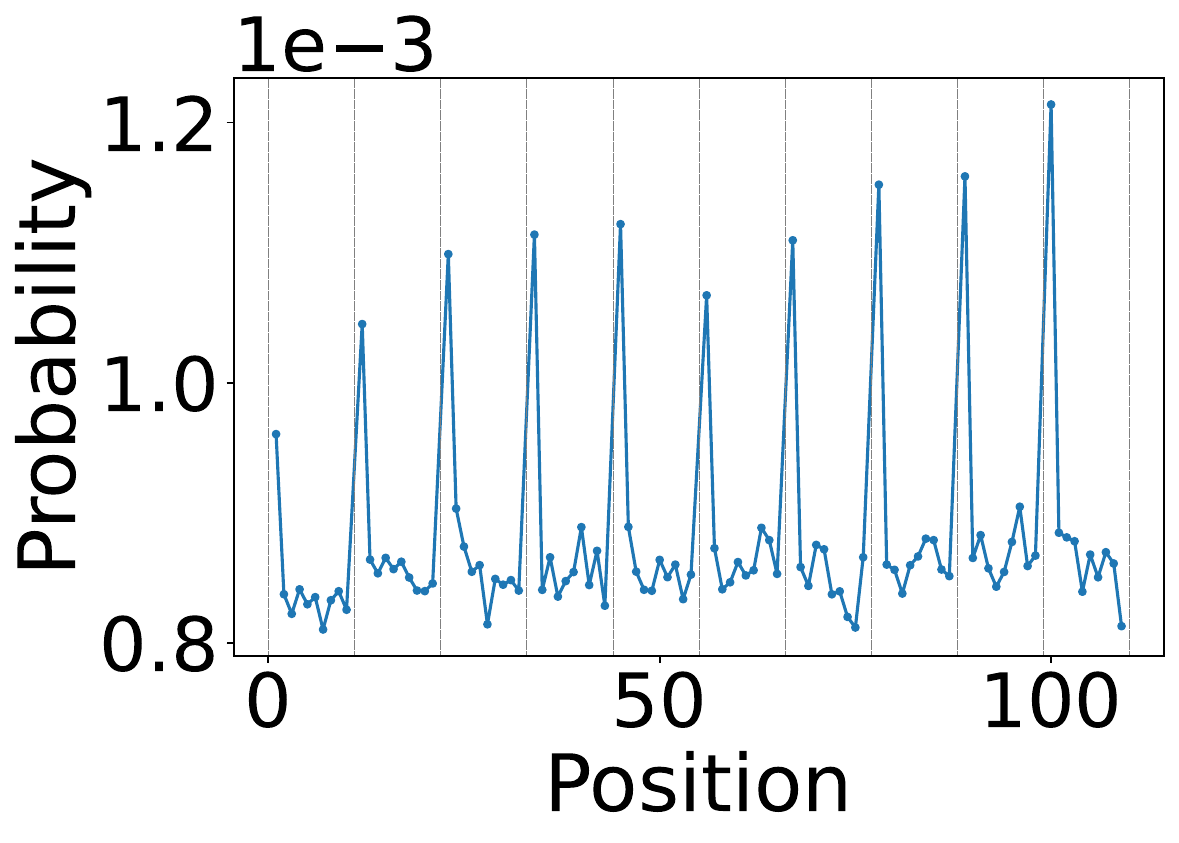} &
\FigGridCell{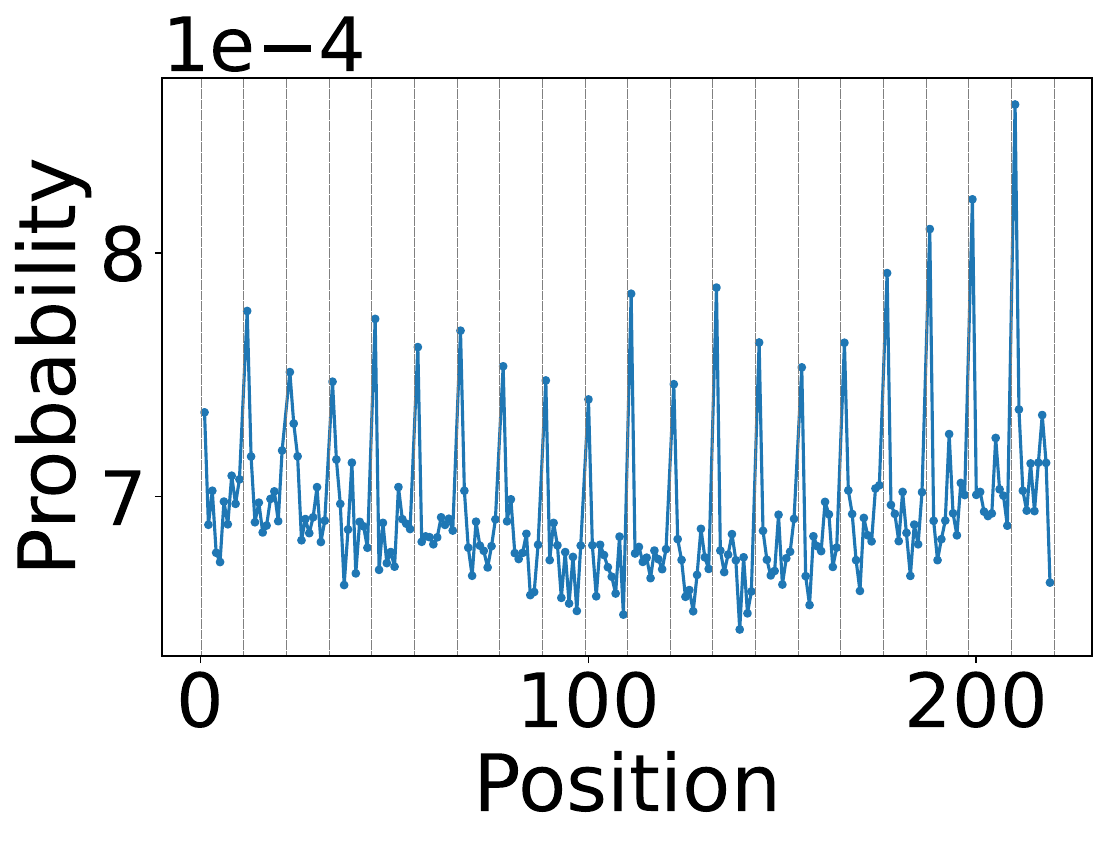} &
\FigGridCell{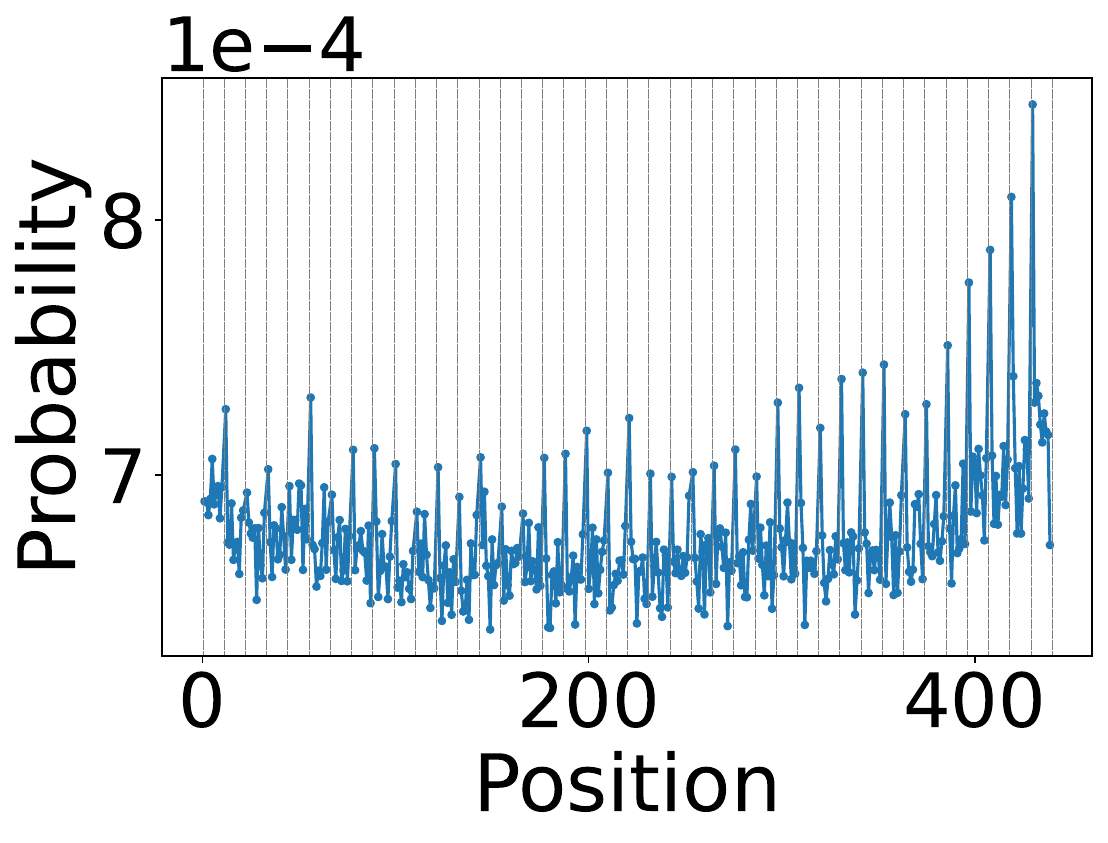} \\

\FigGridVLabel{Mistral} &
\FigGridCell{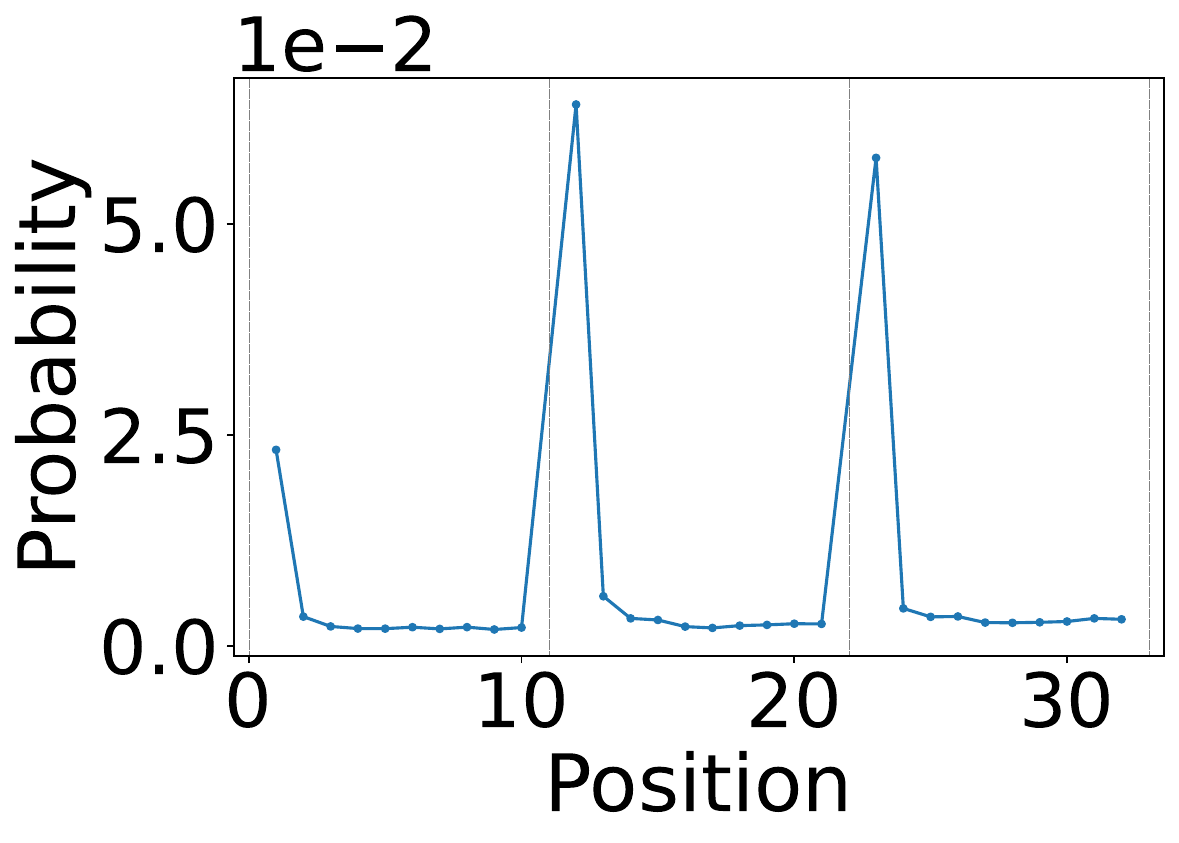} &
\FigGridCell{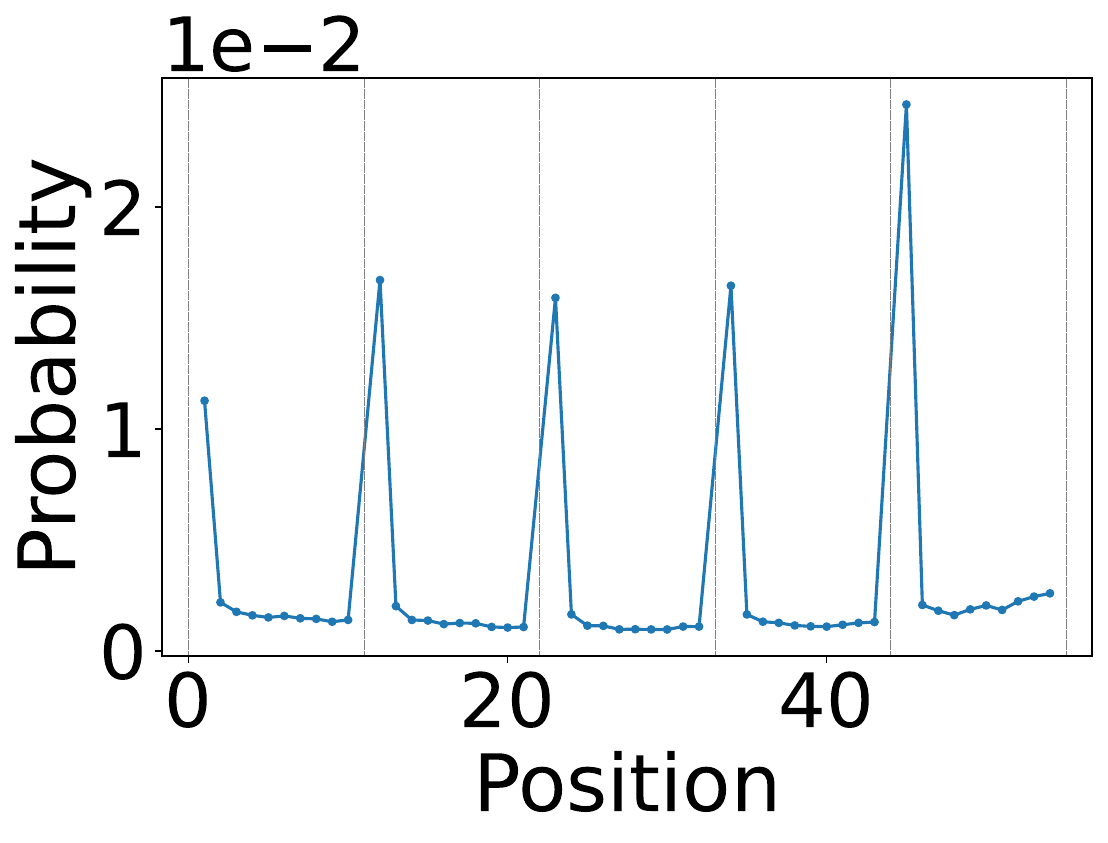} &
\FigGridCell{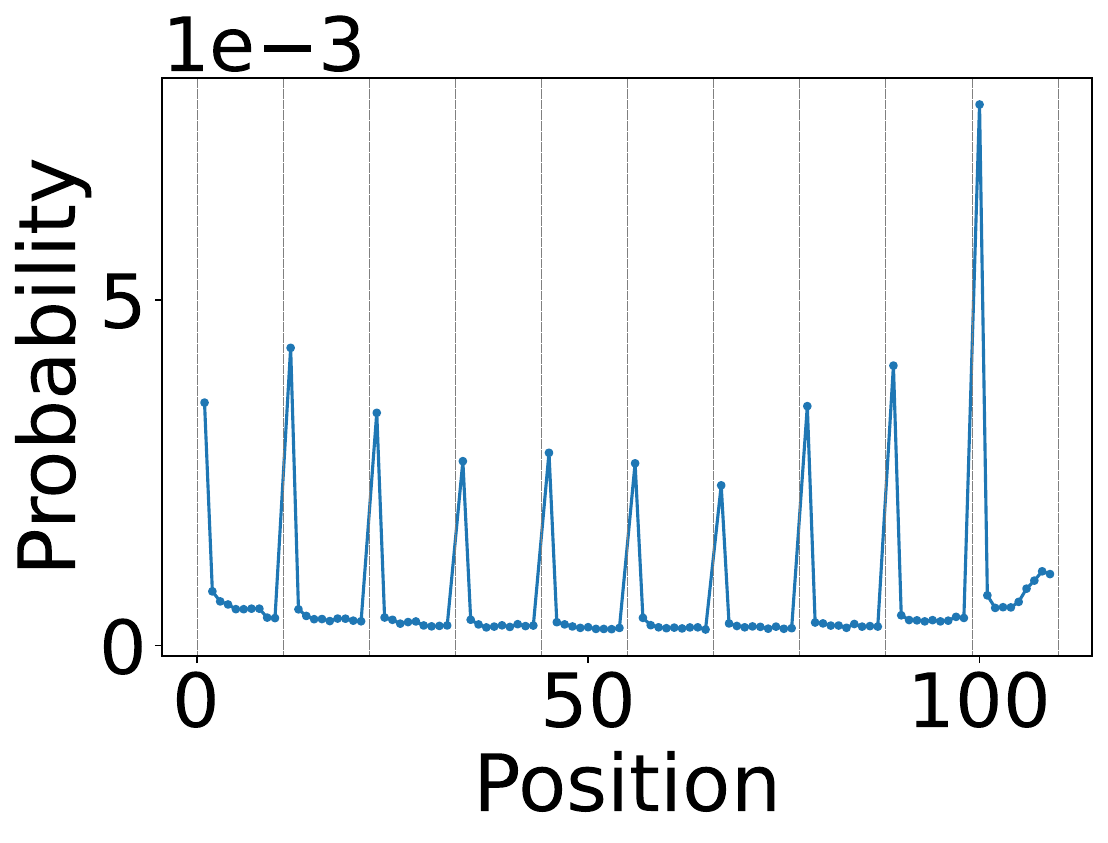} &
\FigGridCell{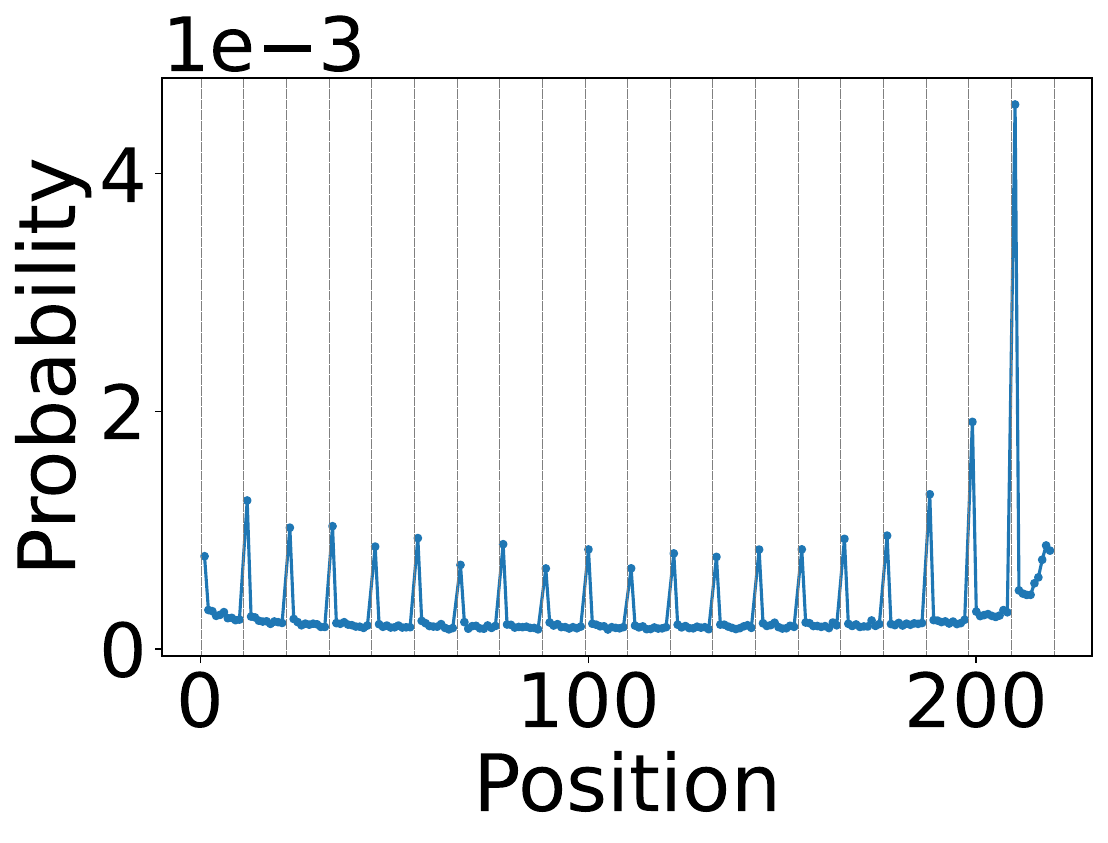} &
\FigGridCell{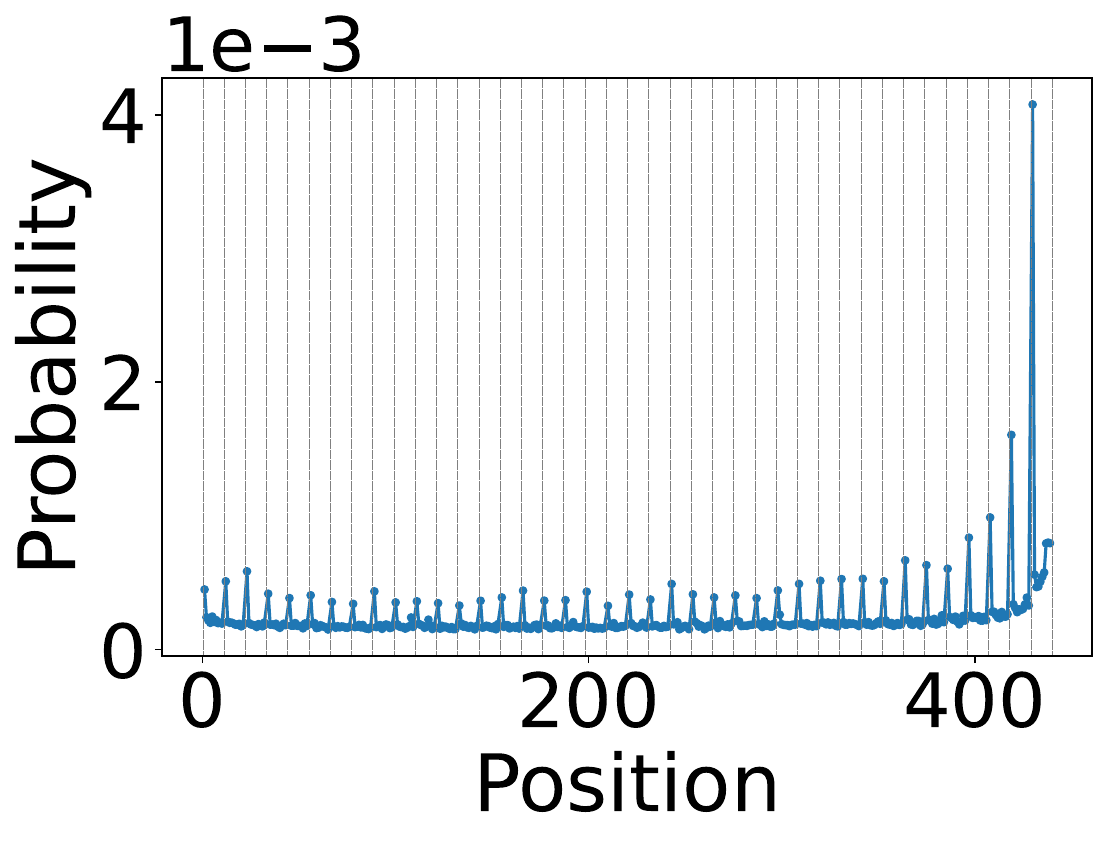} \\

\FigGridVLabel{Qwen} &
\FigGridCell{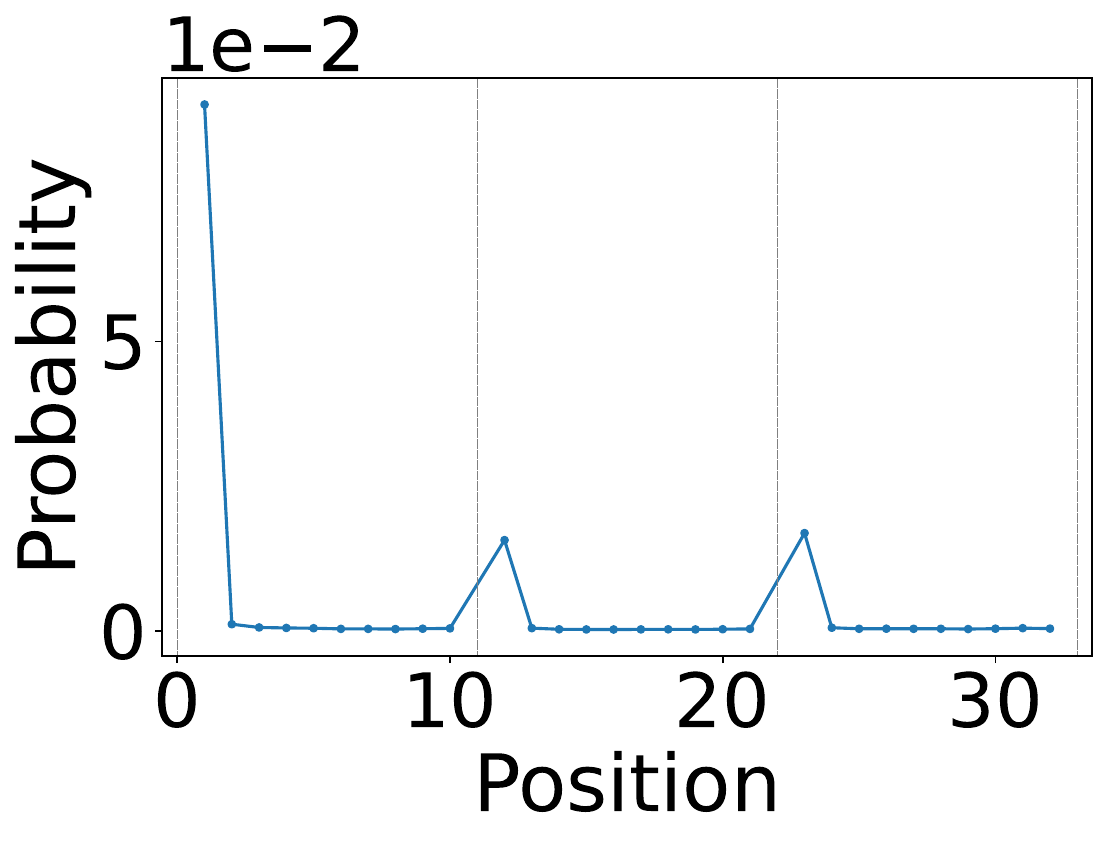} &
\FigGridCell{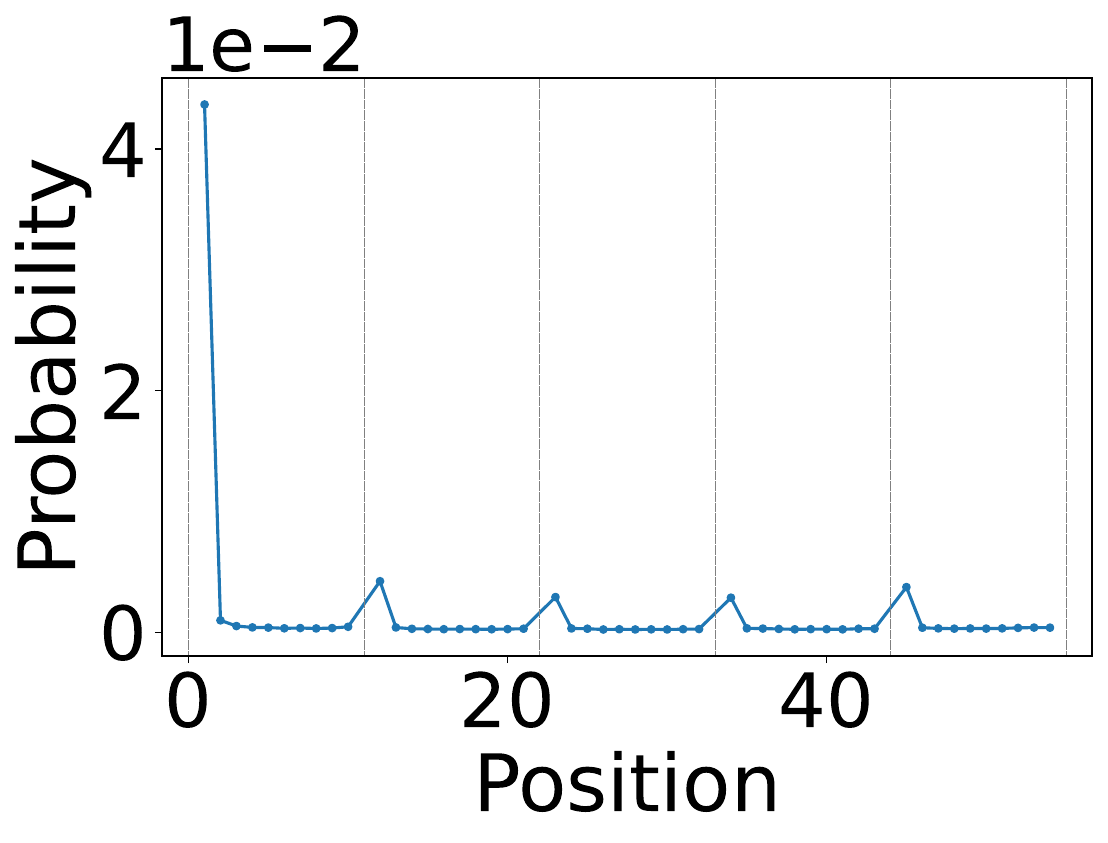} &
\FigGridCell{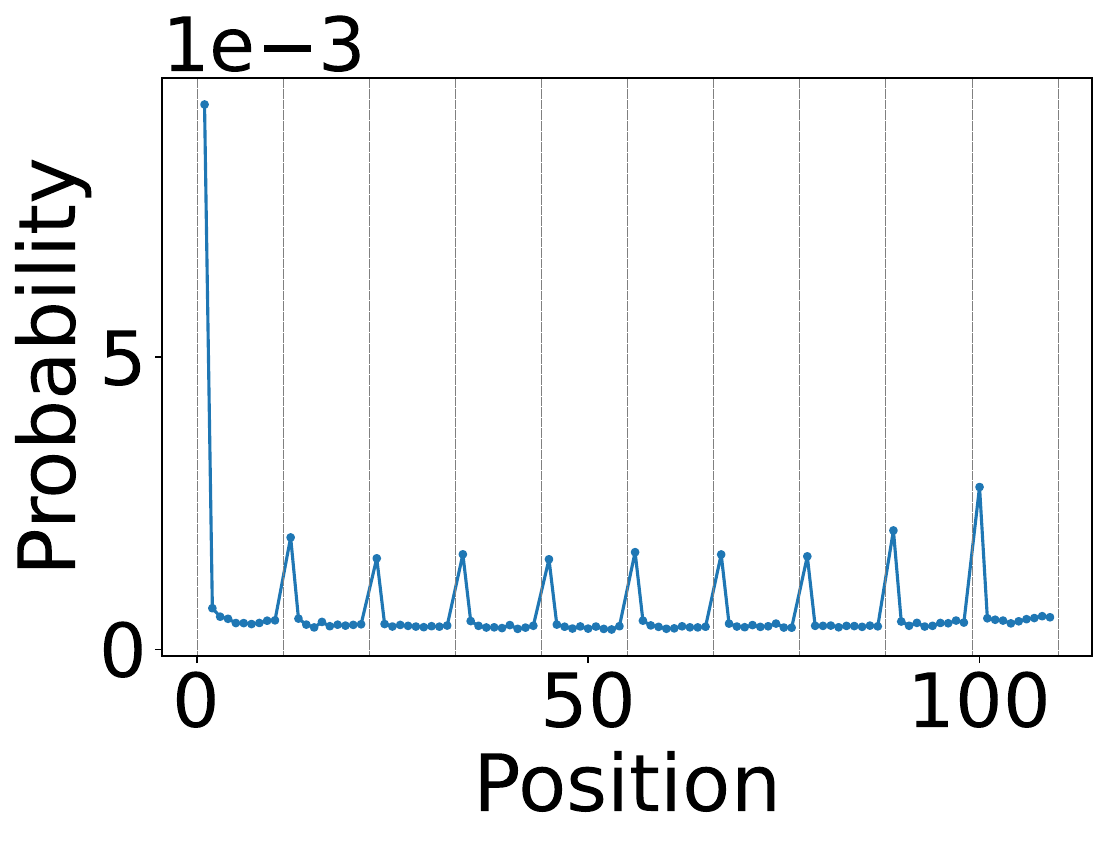} &
\FigGridCell{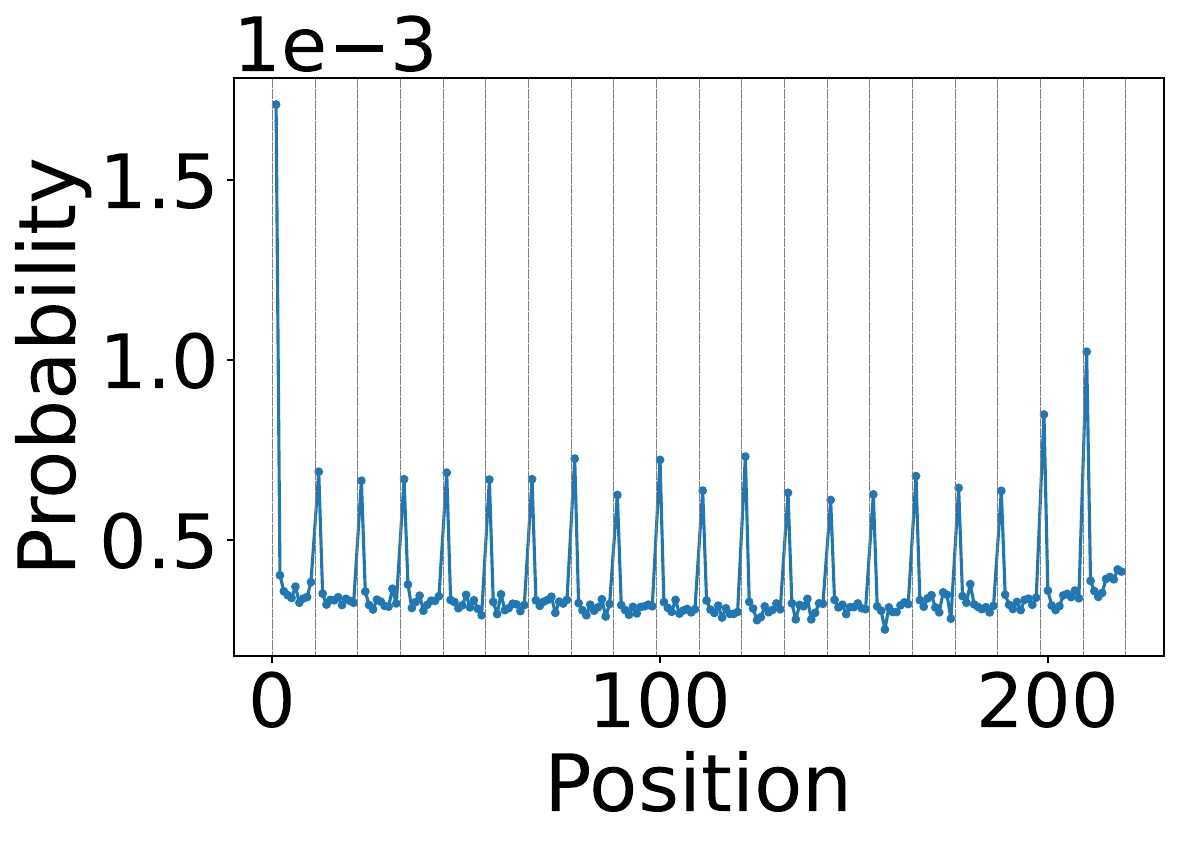} &
\FigGridCell{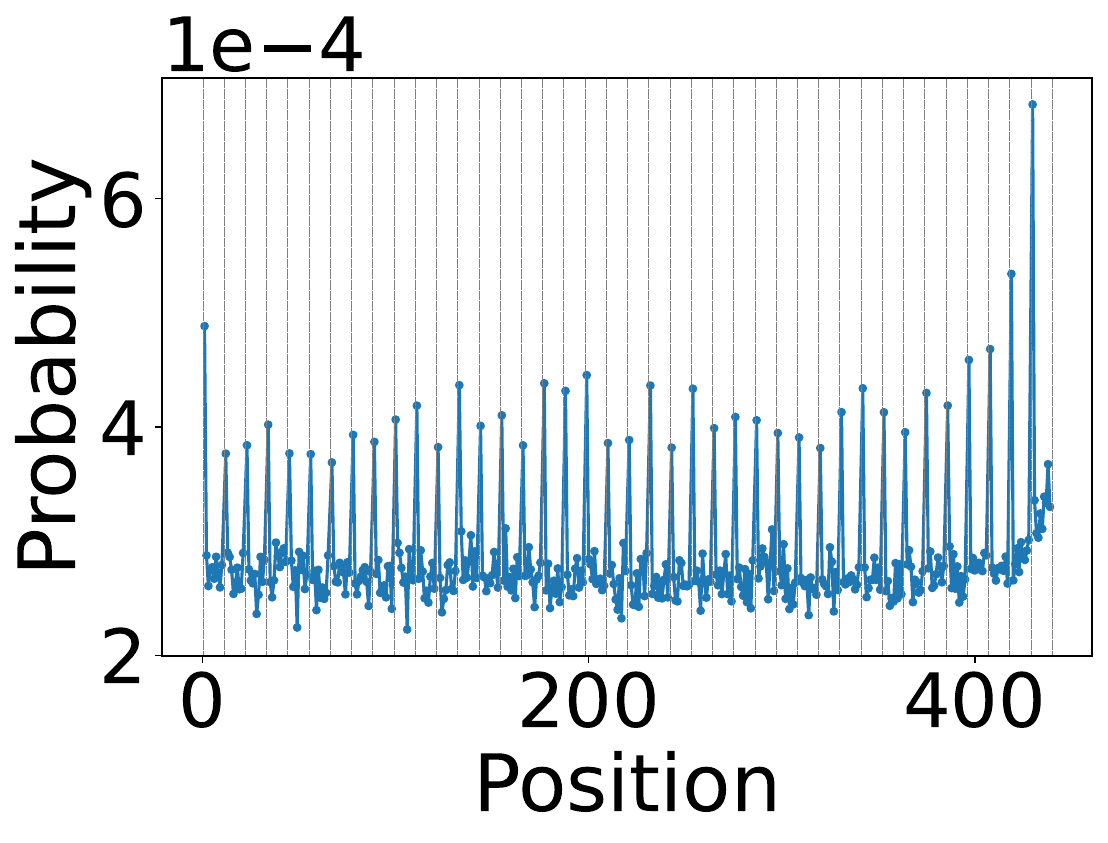} \\

\FigGridVLabel{Gemma} &
\FigGridCell{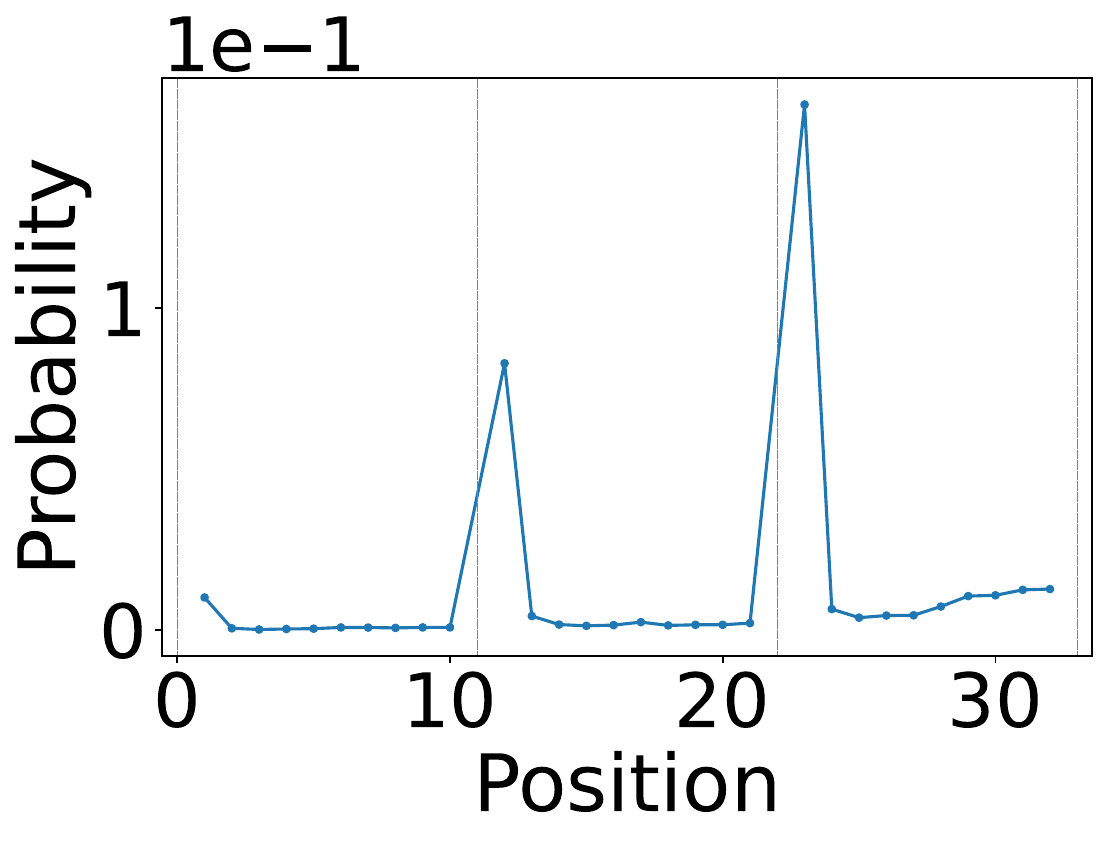} &
\FigGridCell{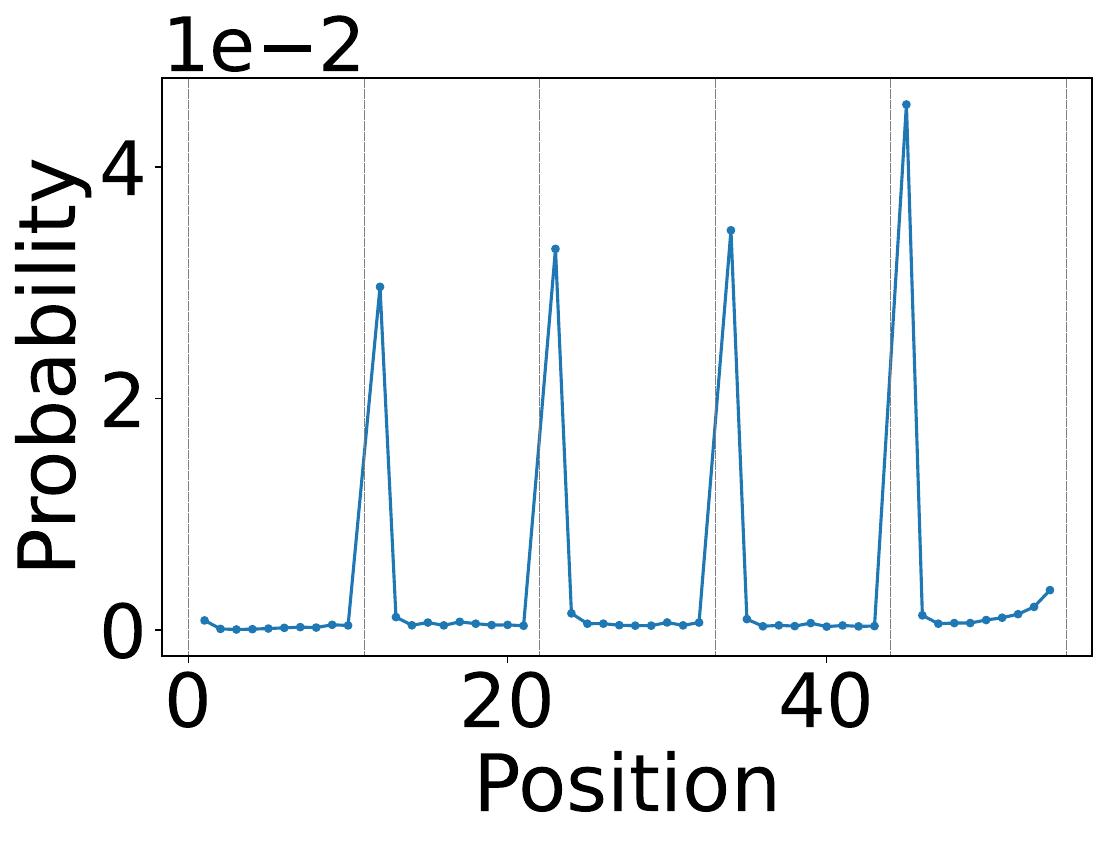} &
\FigGridCell{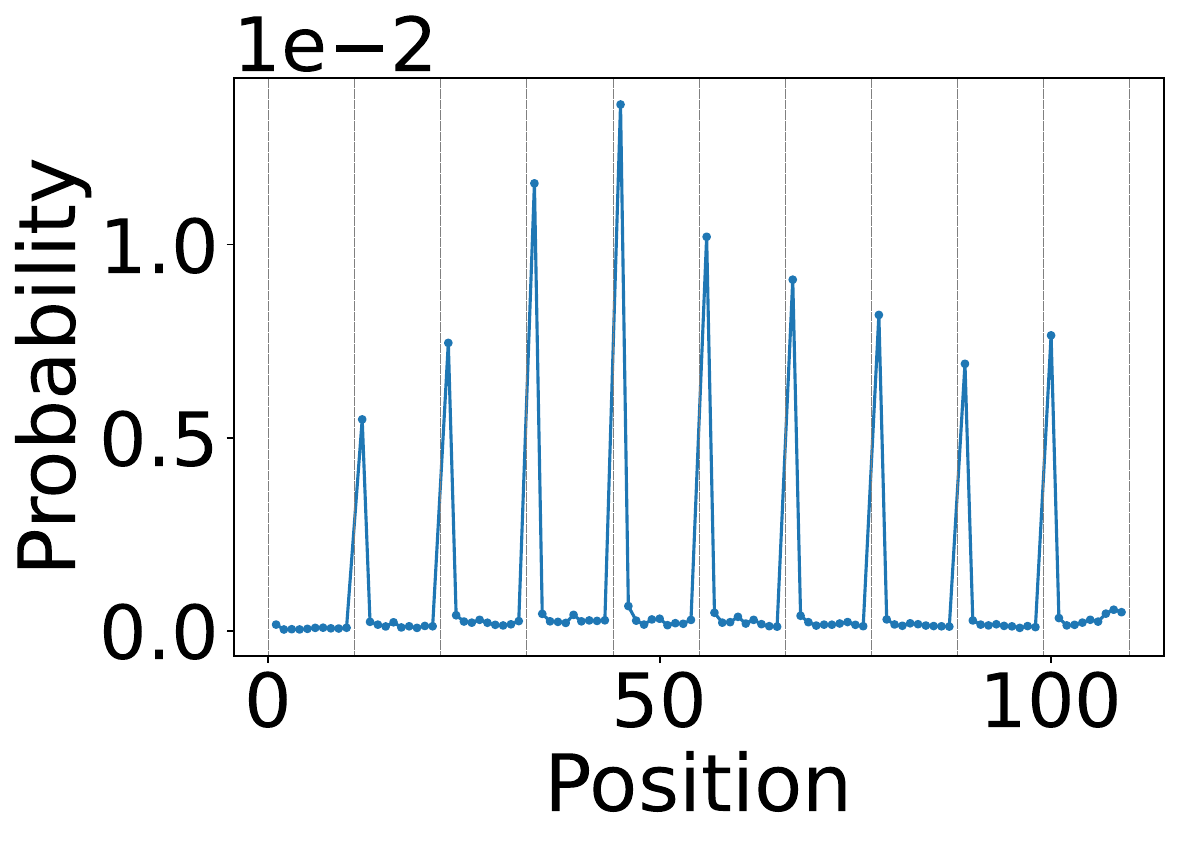} &
\FigGridCell{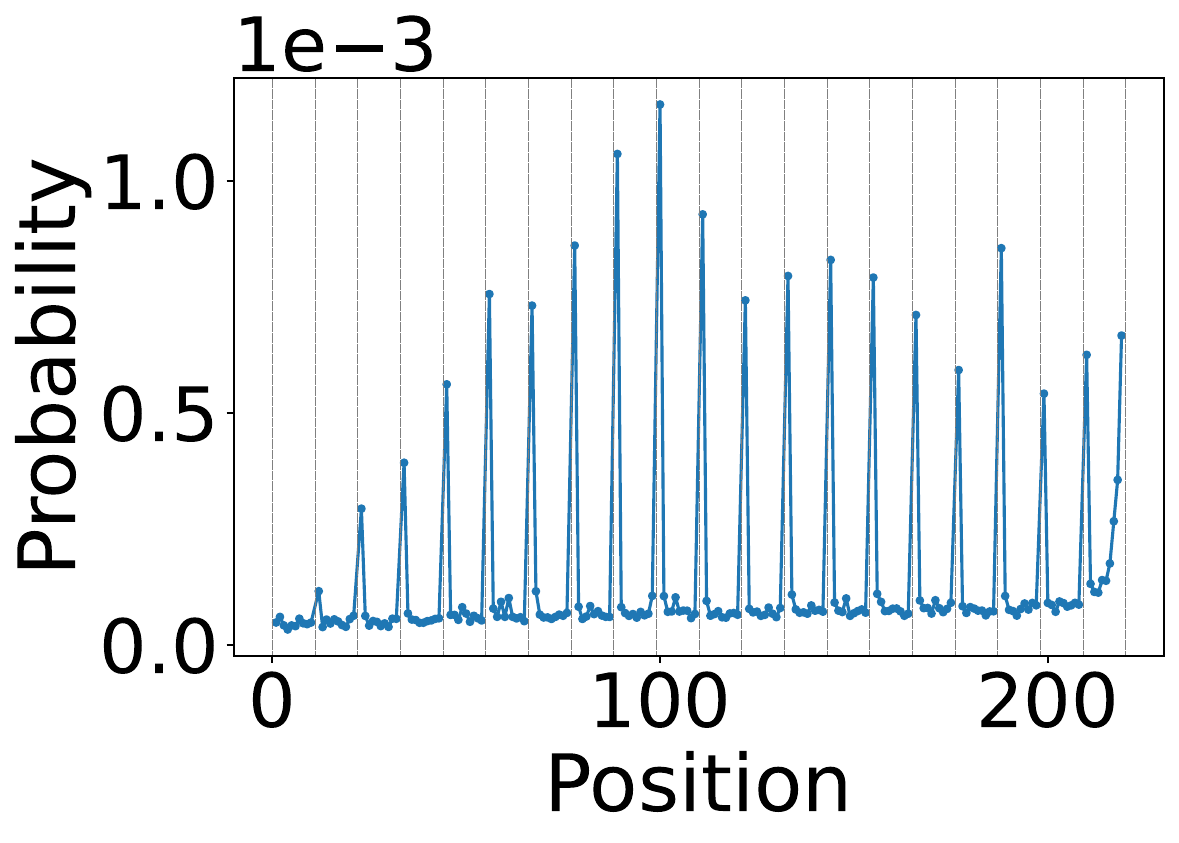} &
\FigGridCell{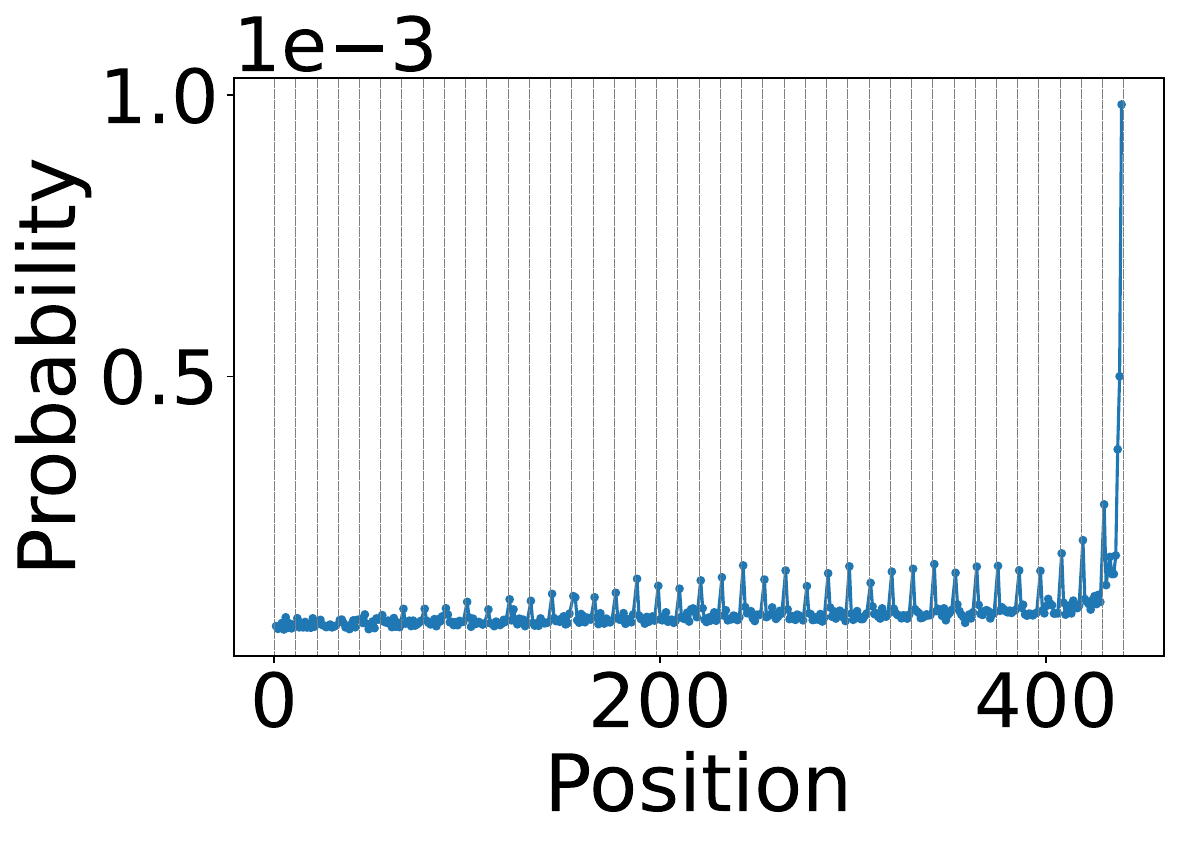} \\

\addlinespace[1pt]
\midrule
\addlinespace[1pt]

\FigGridVLabel{Mamba} &
\FigGridCell{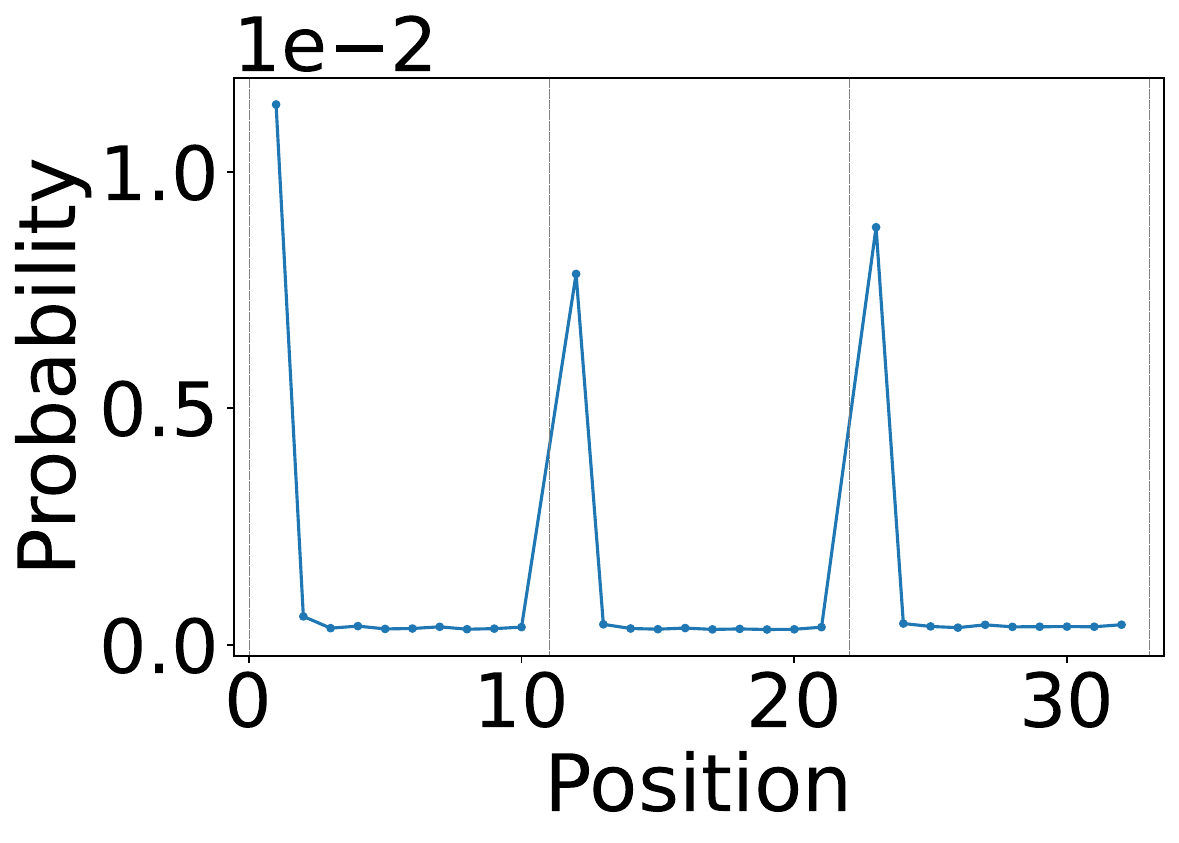} &
\FigGridCell{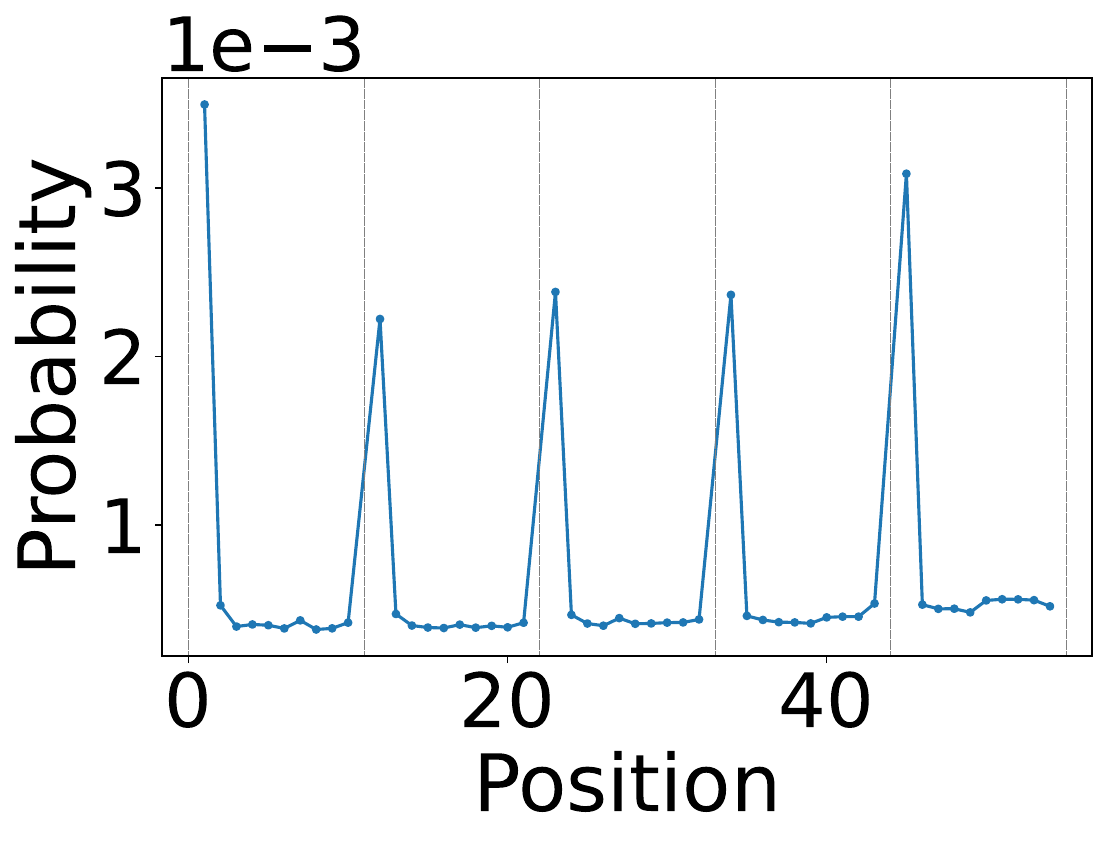} &
\FigGridCell{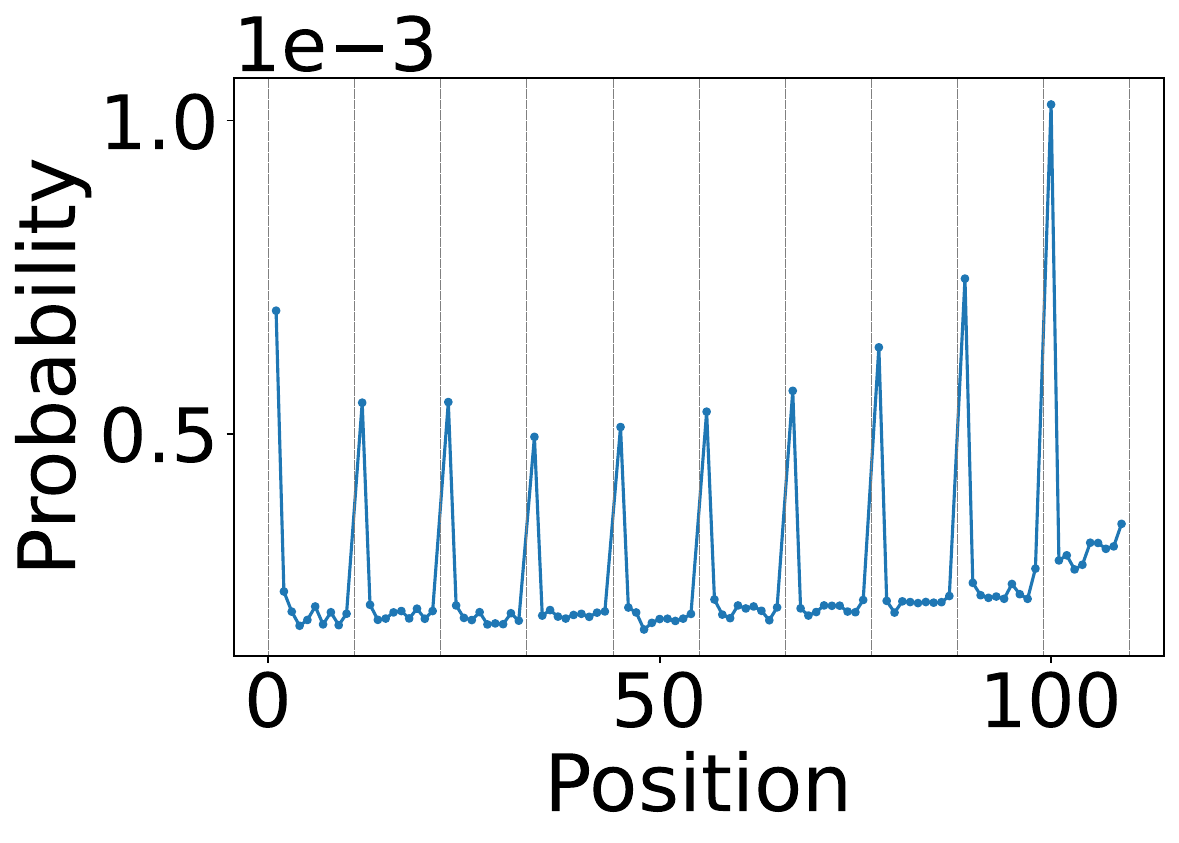} &
\FigGridCell{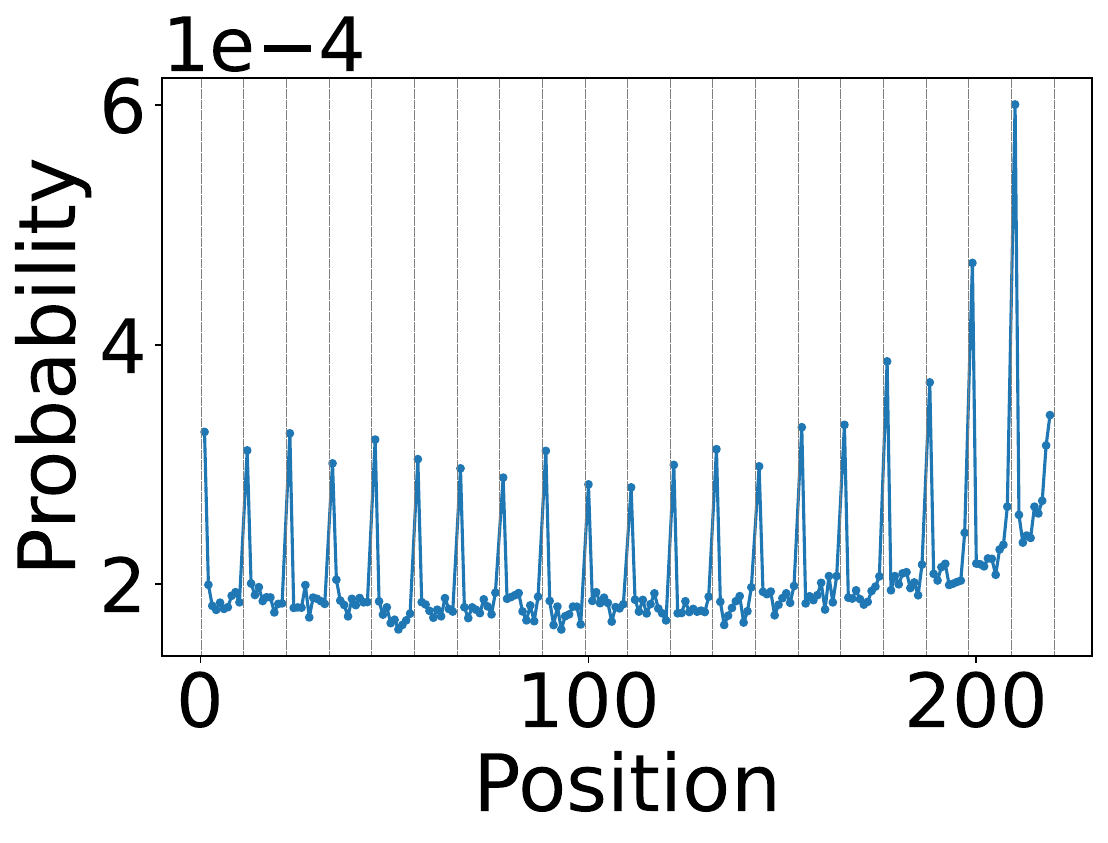} &
\FigGridCell{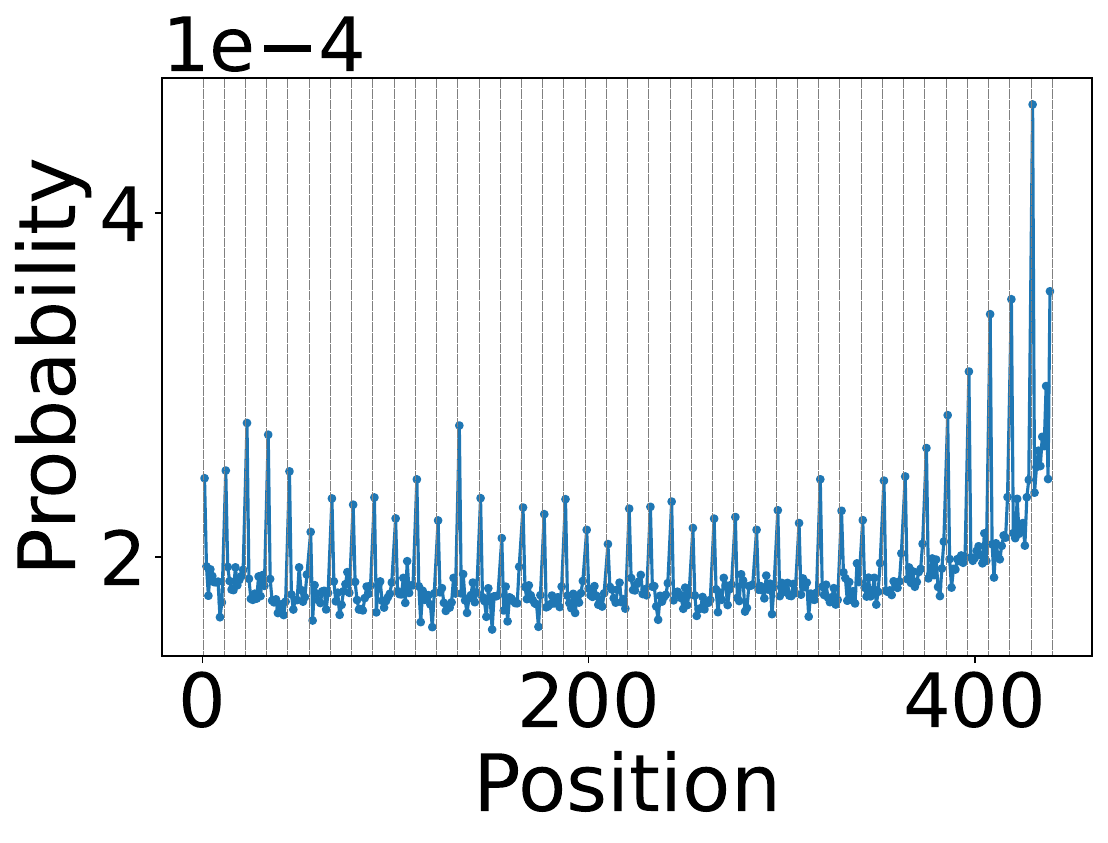} \\

\FigGridVLabel{Falcon-M} &
\FigGridCell{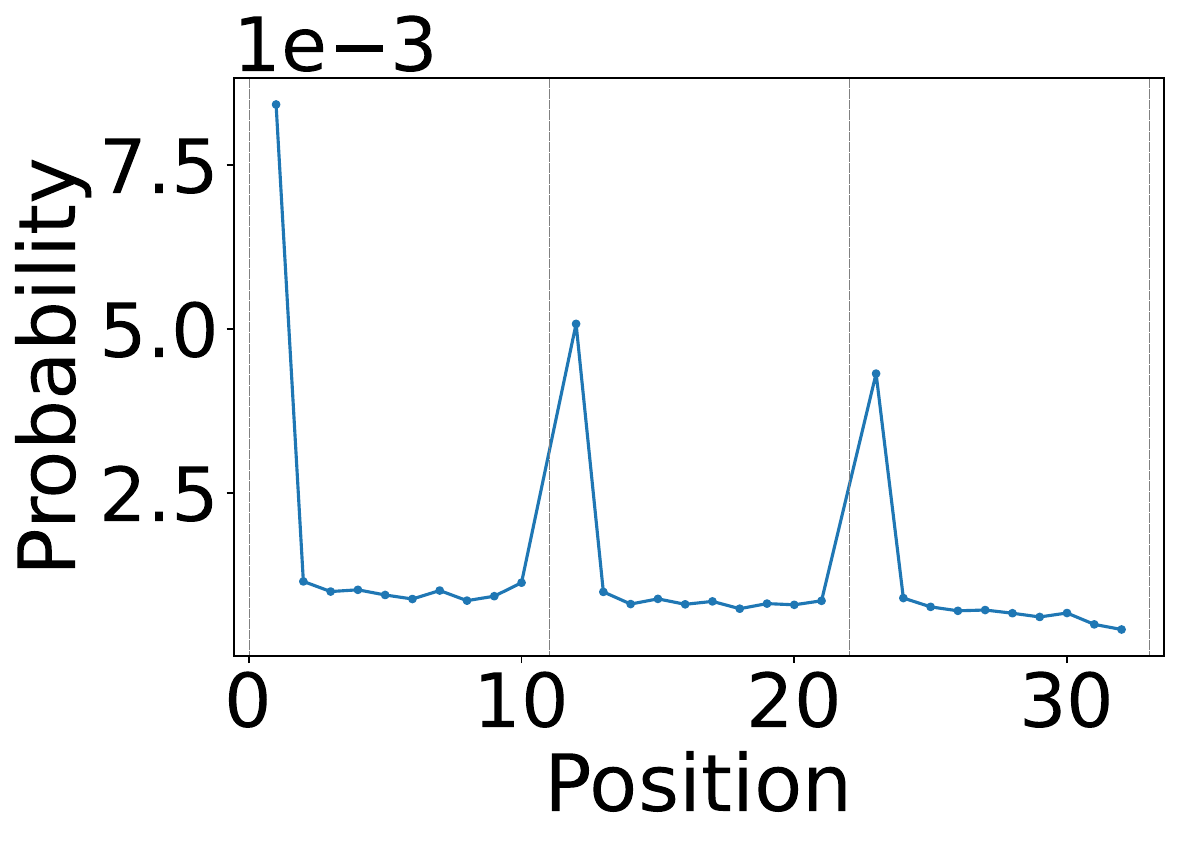} &
\FigGridCell{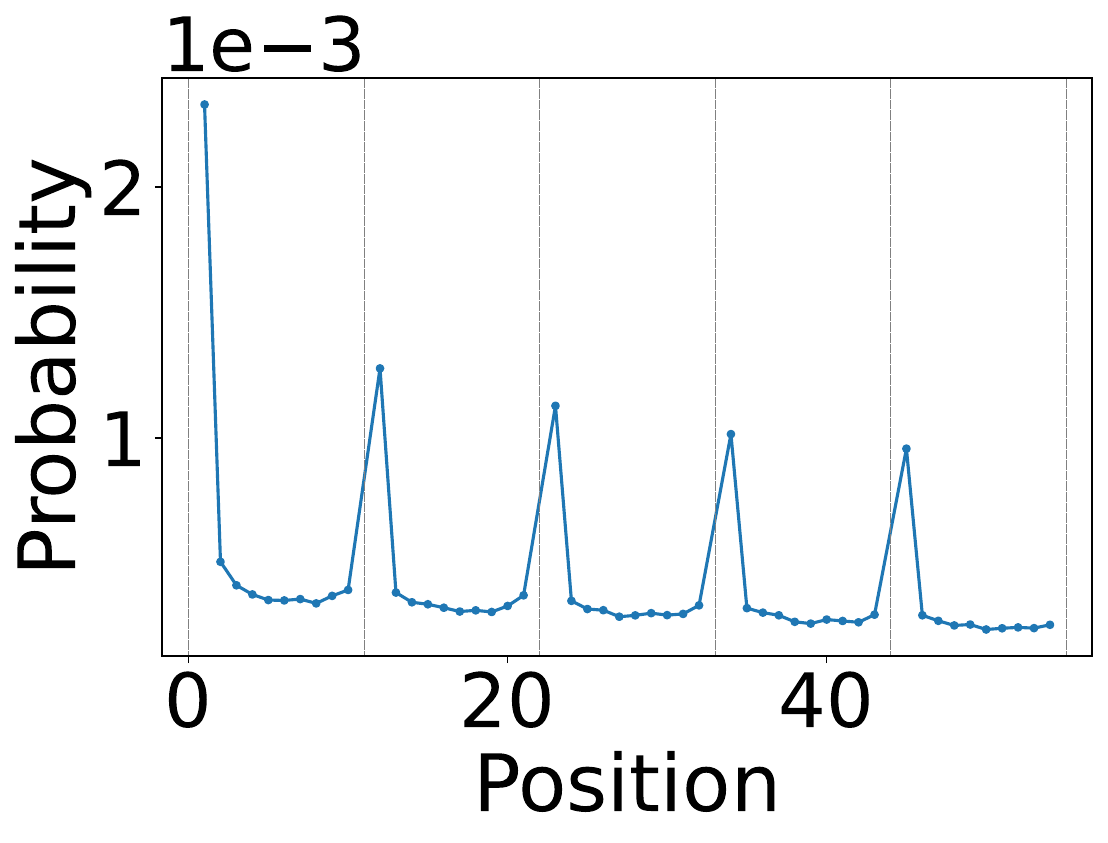} &
\FigGridCell{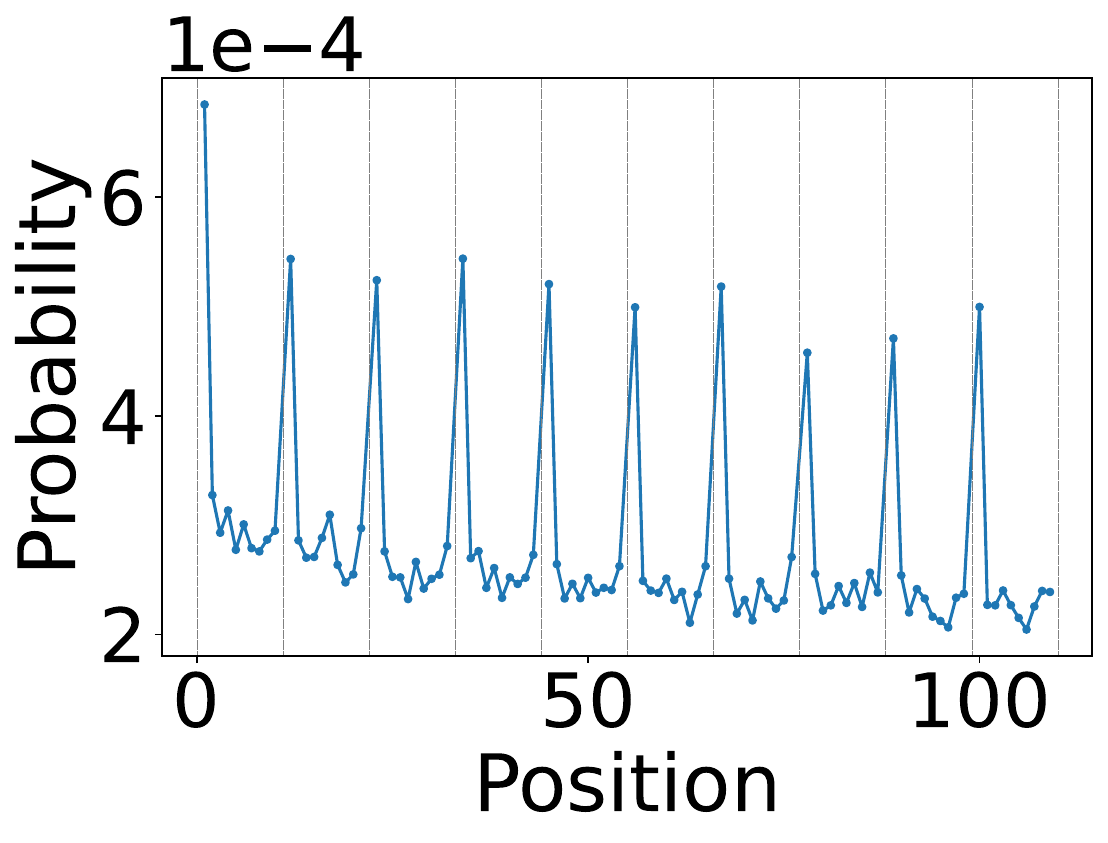} &
\FigGridCell{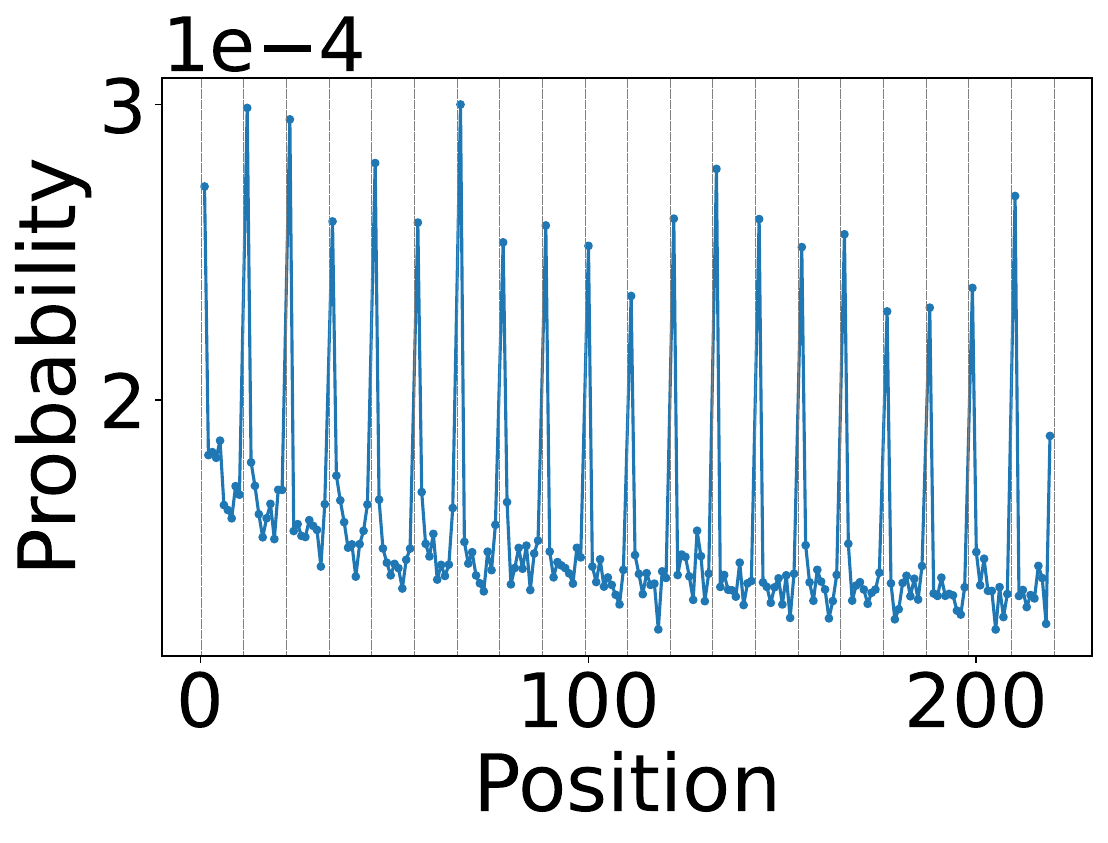} &
\FigGridCell{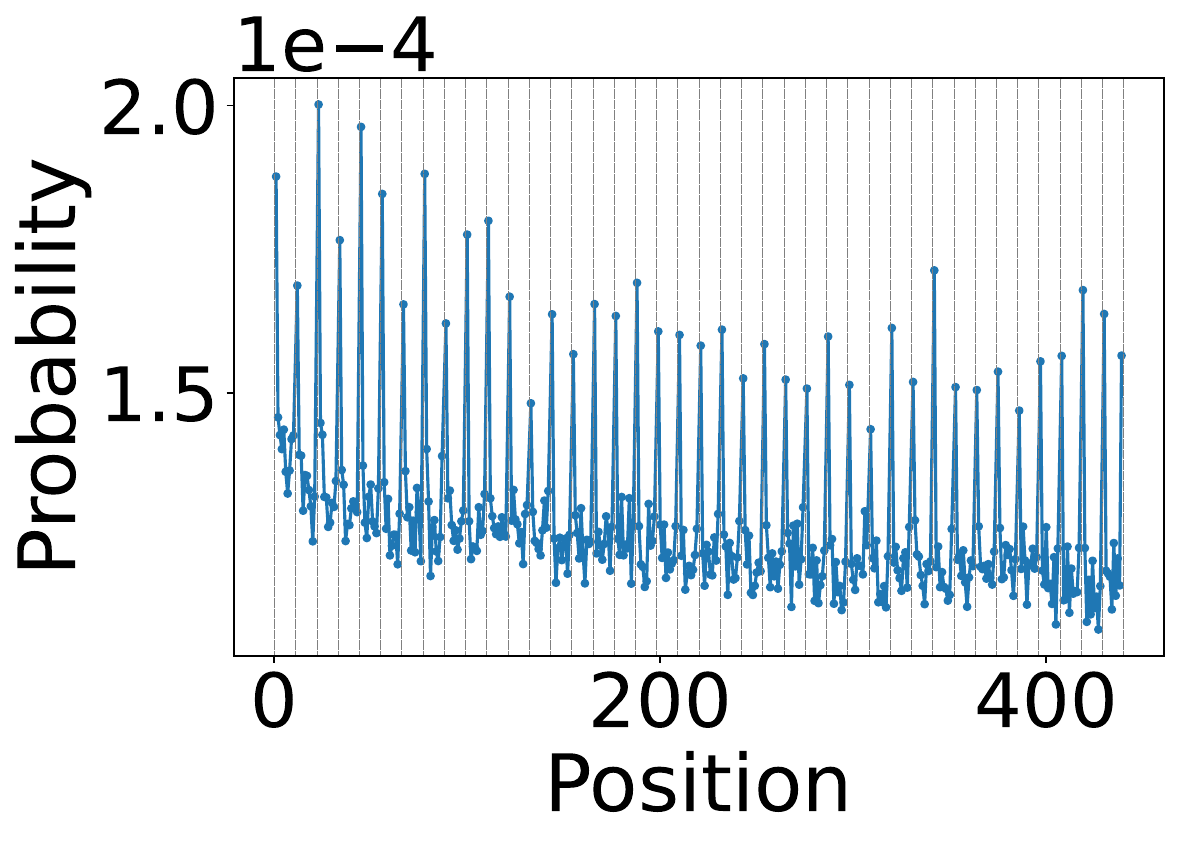} \\

\FigGridVLabel{R-Gemma} &
\FigGridCell{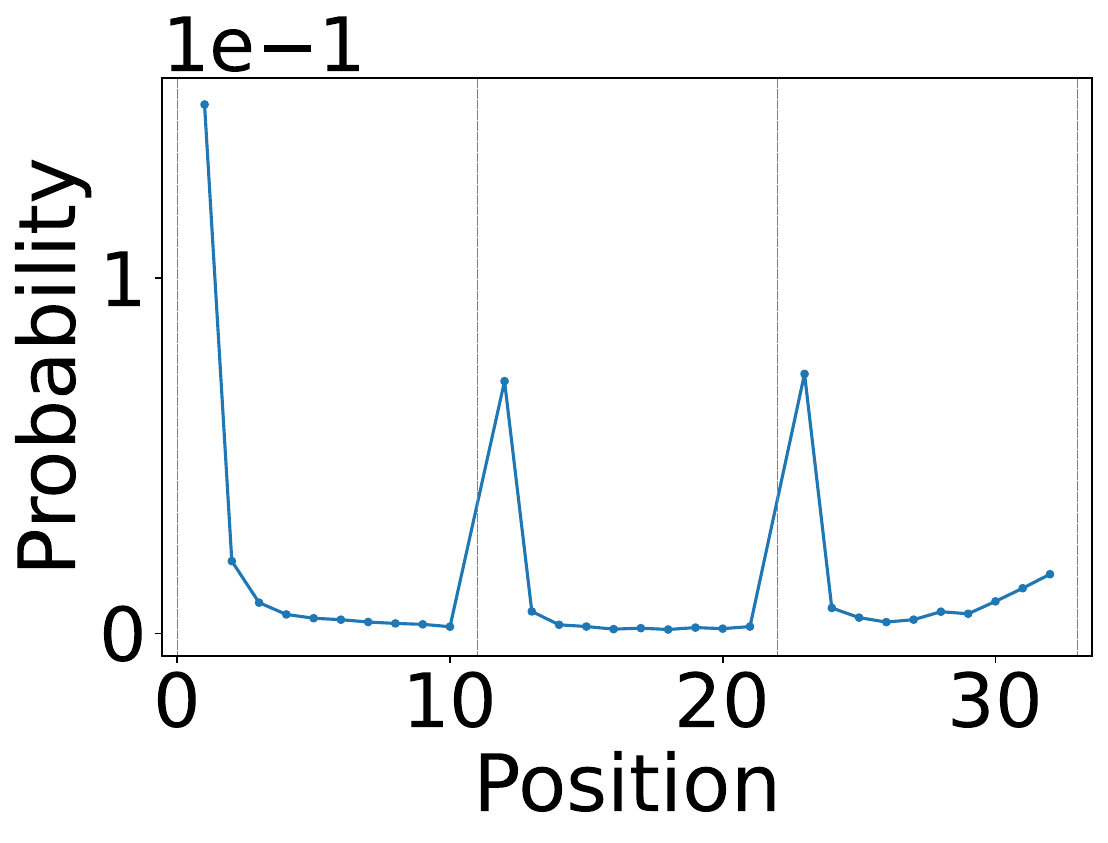} &
\FigGridCell{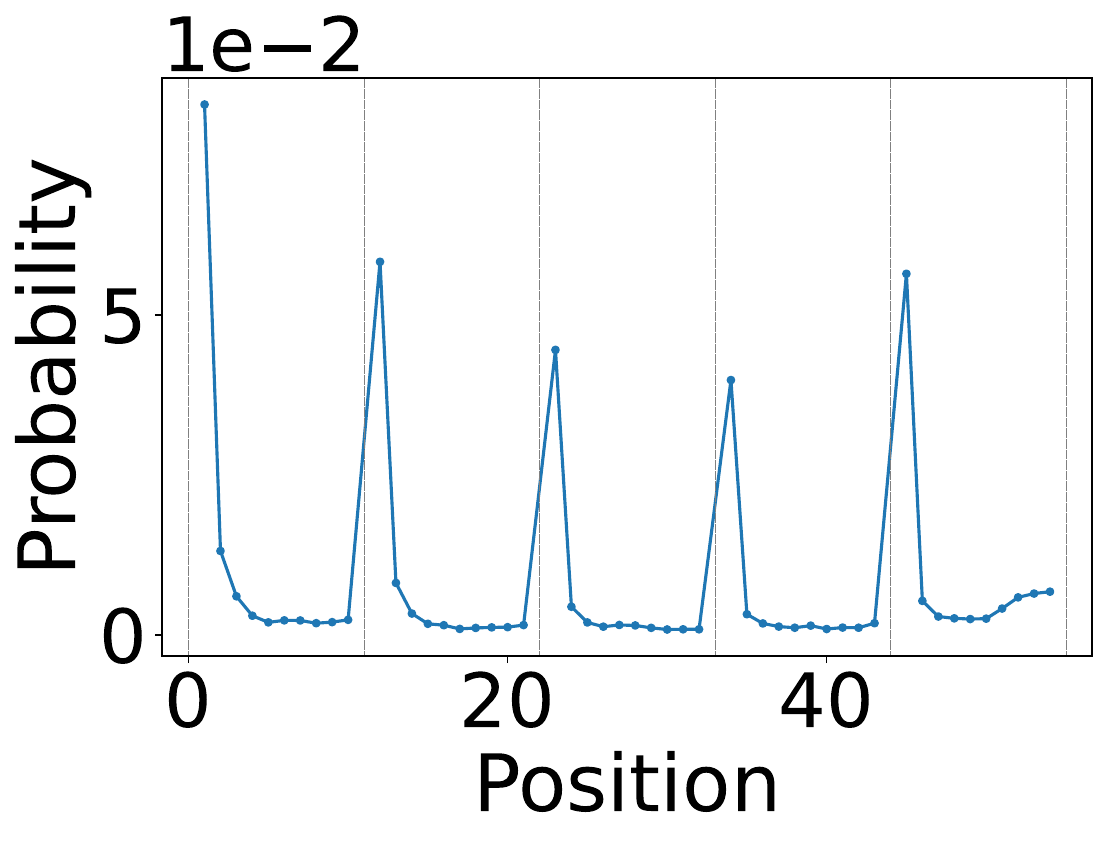} &
\FigGridCell{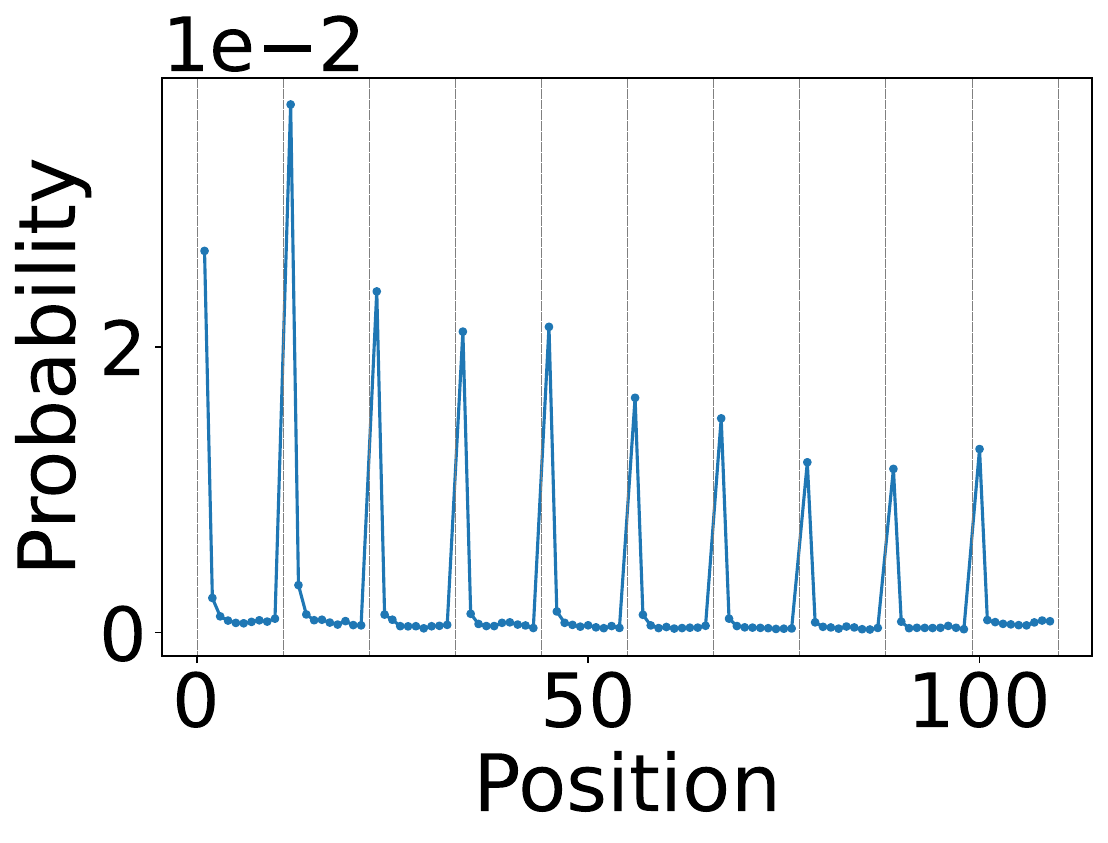} &
\FigGridCell{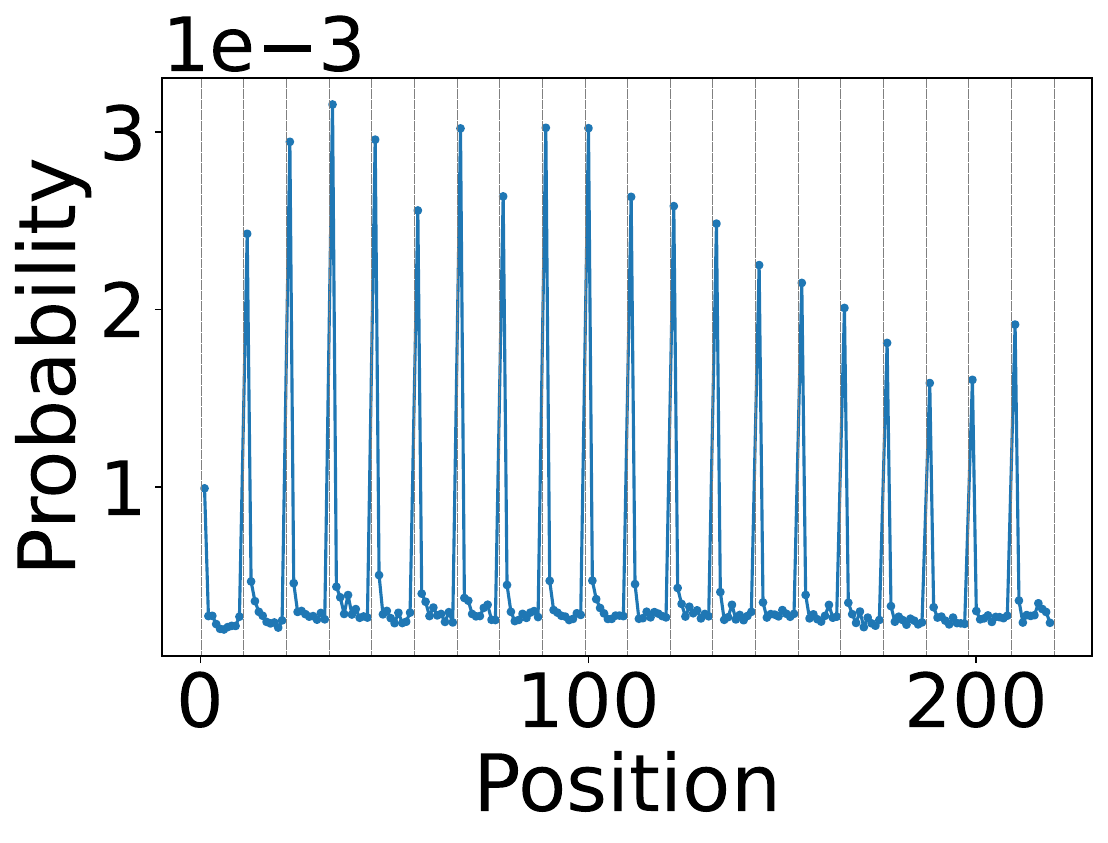} &
\FigGridCell{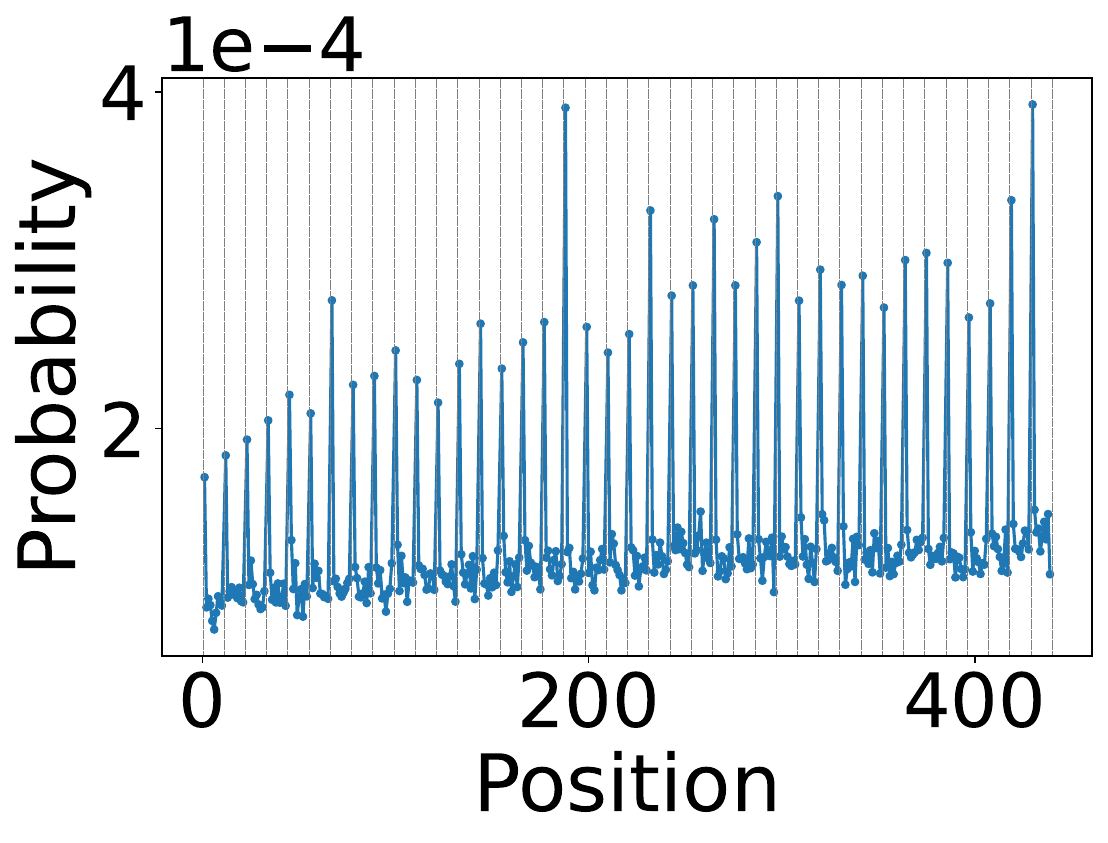} \\

\bottomrule
\end{tabular}
\endgroup

\caption{
Experiment 1: Next-token probability vs.\ position for varying number of fixed-token repetitions (columns) across models (rows).
Vertical gray lines mark the anchor token.
Peaks in the blue lines confirm the presence of \plusoneTok\ recall preference and positional biases.
}
\label{fig:single_target_all_repetitions_all_tokens}
\end{figure*}

\subsection{Experiment 1: Examining Temporal Positional Preferences}

This experiment investigated inherent temporal biases in LLM retrieval, independent of semantic content. We designed prompts containing repetitions of a specific token (`anchor token') interspersed with sequences of random tokens, and quantified how the predicted probability of the next token is influenced by the temporal position of these repetitions (Figure~\ref{fig:schematic}, left).

\paragraph{Prompt construction}
We construct prompts containing $N$ occurrences of a single anchor token $\anchorTok$, each followed by a length-$L$ gap of random tokens,
and a final anchor $\anchorTok$ at the end (Algorithm~\ref{alg:perm_serial}).
The token immediately following the $i$-th anchor occurrence is the \plusoneTok\ token for that occurrence.

\paragraph{Quantification}
For each prompt instance $x$, we compute the next-token distribution $\pnext(\cdot\mid x)$ and form the position-indexed profile
$\pi_t(x)=\pnext(x_t\mid x)$ (Section~\ref{sec:permutation_control}).
Serial recall is summarized by $p_i^{(+1)}=\pnext(x_{a_i+1}\mid x)$, the probability assigned to the \plusoneTok\ token from the $i$-th anchor.
We report means and SEM across $M=5000$ independently sampled prompts per condition.

\textbf{Results} Figure~\ref{fig:single_target_all_repetitions_all_tokens} shows the resulting probabilities (averaged across permutations) as a function of token position within the prompt (position 0 is the start), varying the number of repetitions (columns) while keeping the spacing fixed at 10 tokens. Figure~\ref{fig:single_target_all_spacings_all_tokens} shows results varying the spacing (columns) with a fixed number of 10 repetitions. Probabilities for the repeated fixed token itself were excluded from visualization due to their artificially high values resulting from repetition frequency.

Across all seven models, probabilities exhibit distinct peaks corresponding to the token immediately following each instance of the fixed token (the \plusoneTok\ token). This preference for the \plusoneTok\ token demonstrates a tendency towards serial recall (repeating sequences in the order presented), consistent with previous findings in transformer models \citep{mistry2025emergence}.

Our results reveal several novel insights: 1) \emph{SSMs exhibit serial recall:} SSMs demonstrate a similar \plusoneTok\ token preference, indicating a comparable tendency for serial recall despite architectural differences from transformers. 2) \emph{Positional modulation of peaks:} The magnitude of the \plusoneTok\ probability peaks varies depending on the repetition's position within the prompt, indicating temporal biases (primacy/recency). This is highlighted in Figure~\ref{fig:single_target_all_repetitions_plusone_tokens}, which plots only the \plusoneTok\ token probabilities against the repetition number (depth in prompt). 3) \emph{Model-specific biases:} These temporal preferences differ across models. Mistral shows a recency bias (higher probabilities towards the end). Falcon-Mamba exhibits a primacy bias (higher probabilities near the beginning). Gemma's preference shifts: with fewer repetitions, the peak is mid-prompt, but with more repetitions, it shifts towards the end. These shifts are also observable when varying spacing (Figure~\ref{fig:single_target_all_spacings_all_tokens}). No globally distinct temporal patterns consistently separated transformer and state-space architectures.

\subsection{Experiment 2: Characterizing Episodic Retrieval as a Function of Temporal Distance and Context Overlap}

Experiment 1 probes serial recall for a repeated anchor in a largely context-free setting. Here we test a more episodic setting with partially overlapping structure, in which retrieval requires selecting the correct continuation given a short cue.

\begin{figure*}[h!]
\centering
\begingroup
\setlength{\FigGridLabelW}{0.65cm}
\setlength{\FigGridCellH}{2.35cm} 
\setlength{\FigGridLabelSep}{1pt}
\setlength{\FigGridColSep}{0.5pt}

\FigGridSetup{5}
\renewcommand{\arraystretch}{1.0}

\begin{tabular}{@{}>{\centering\arraybackslash}m{\FigGridLabelW}@{\hspace{\FigGridLabelSep}}*{5}{>{\centering\arraybackslash}m{\FigGridPlotW}}@{}}
\toprule
& \multicolumn{5}{c}{\FigHdr Probe episode} \\
\cmidrule(lr){2-6}
& \FigHdr 1 & \FigHdr 2 & \FigHdr 3 & \FigHdr 4 & \FigHdr 5 \\
\midrule
\FigGridVLabel{Llama} &
\FigGridCell{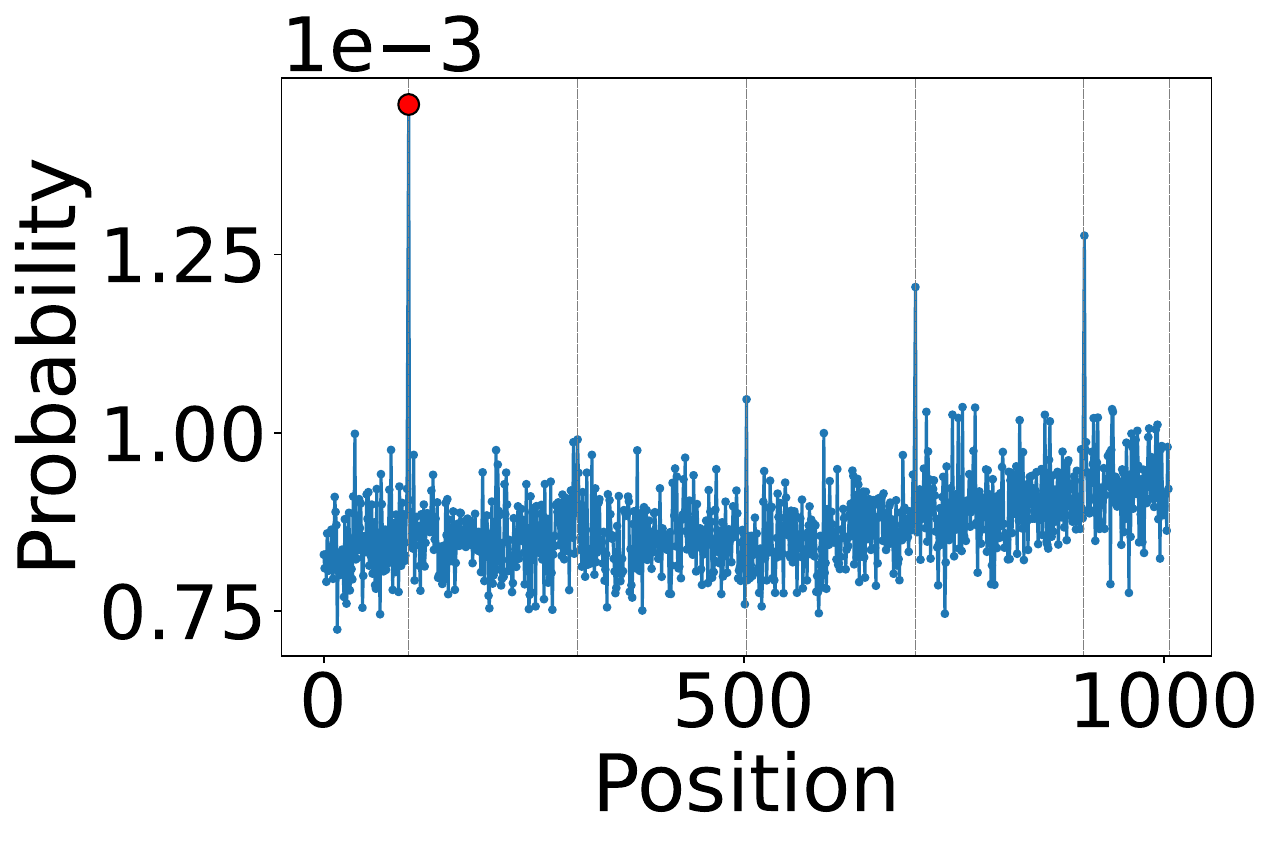} &
\FigGridCell{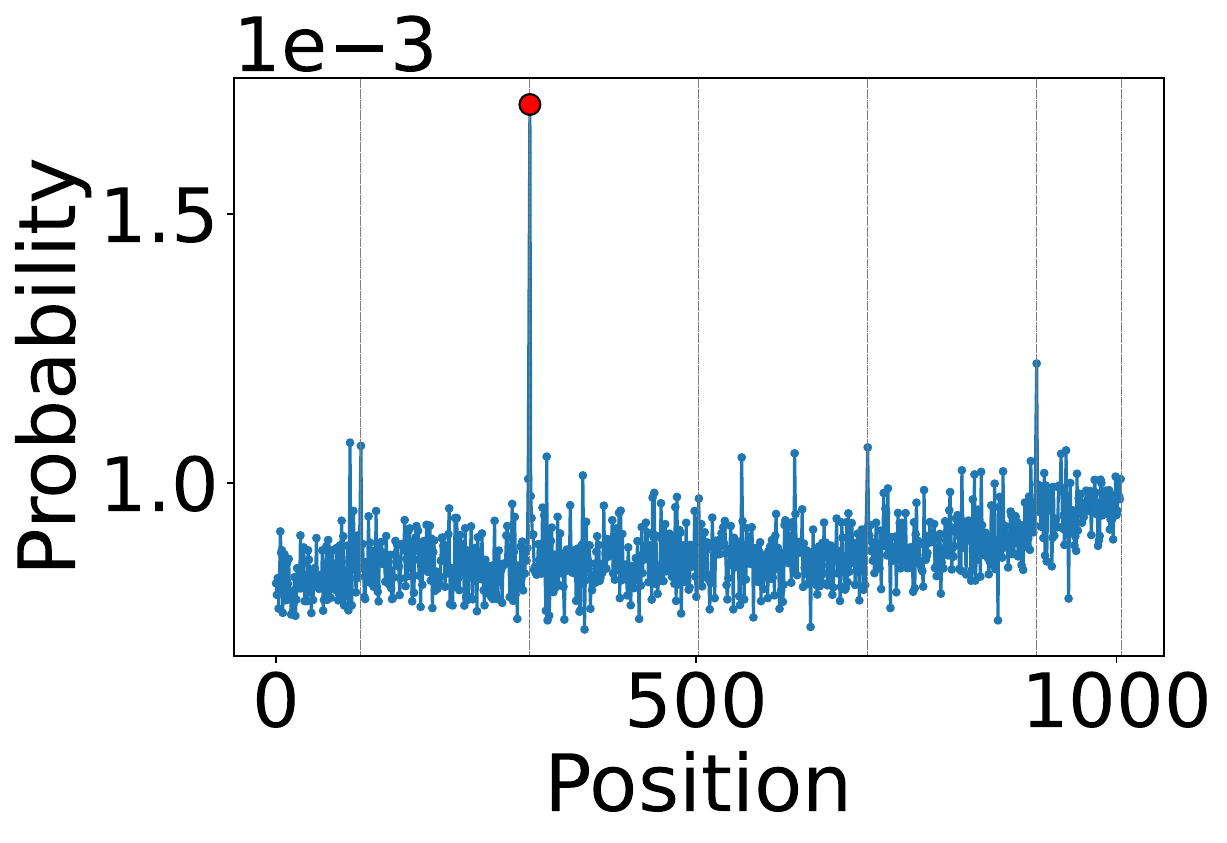} &
\FigGridCell{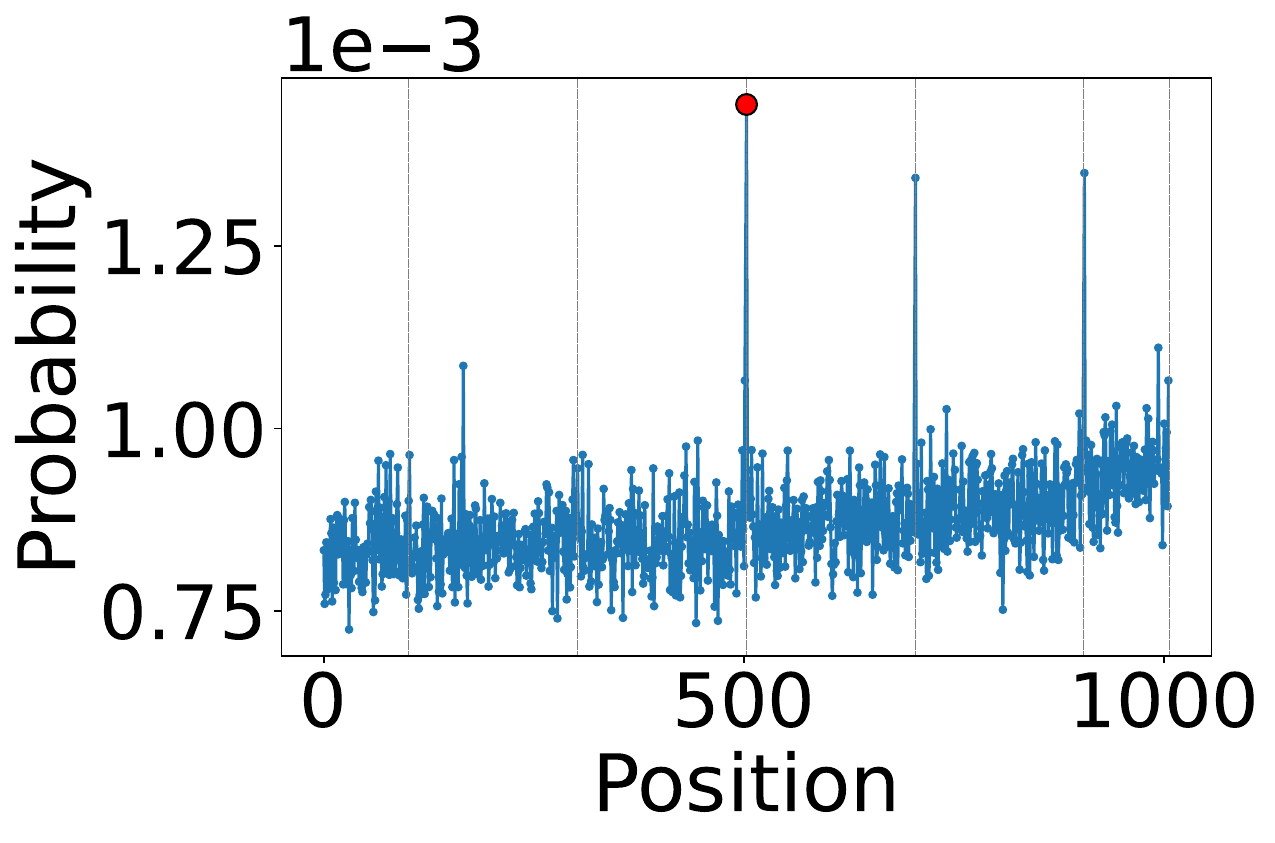} &
\FigGridCell{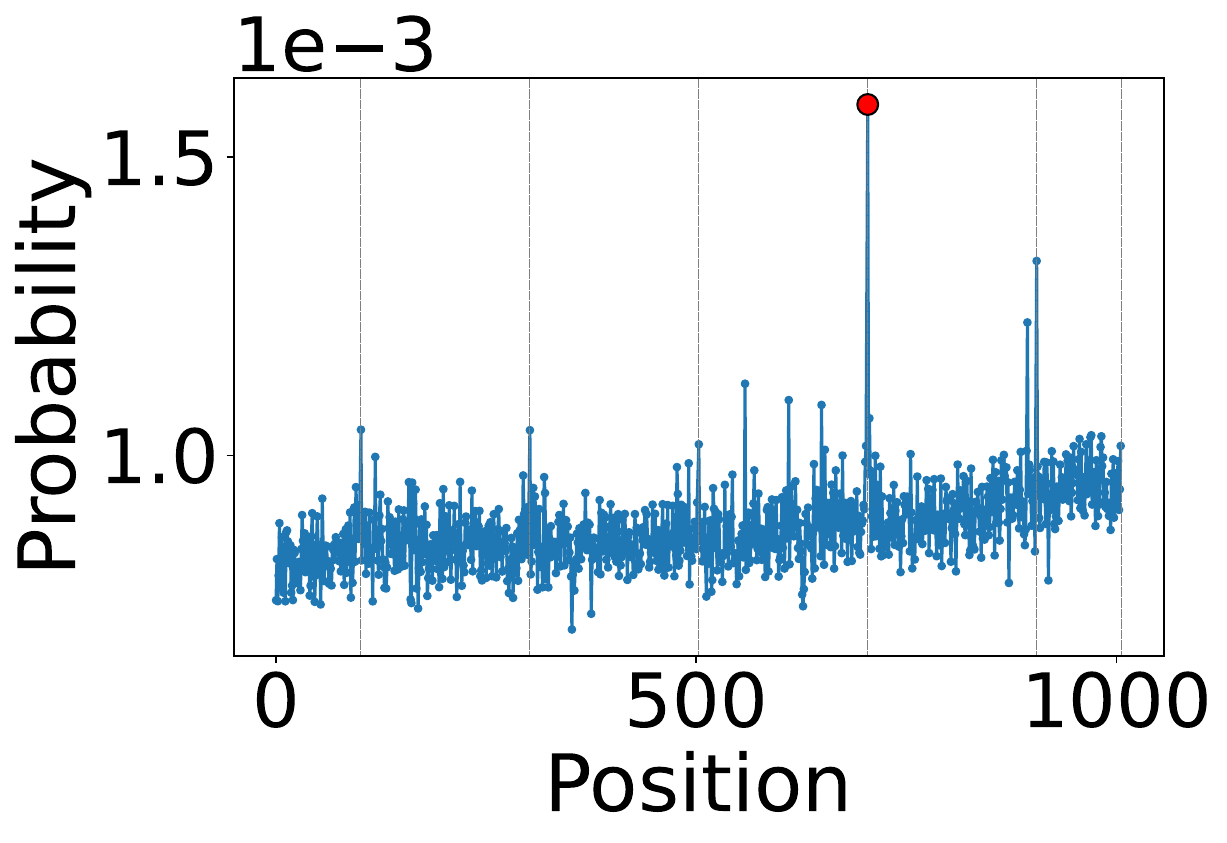} &
\FigGridCell{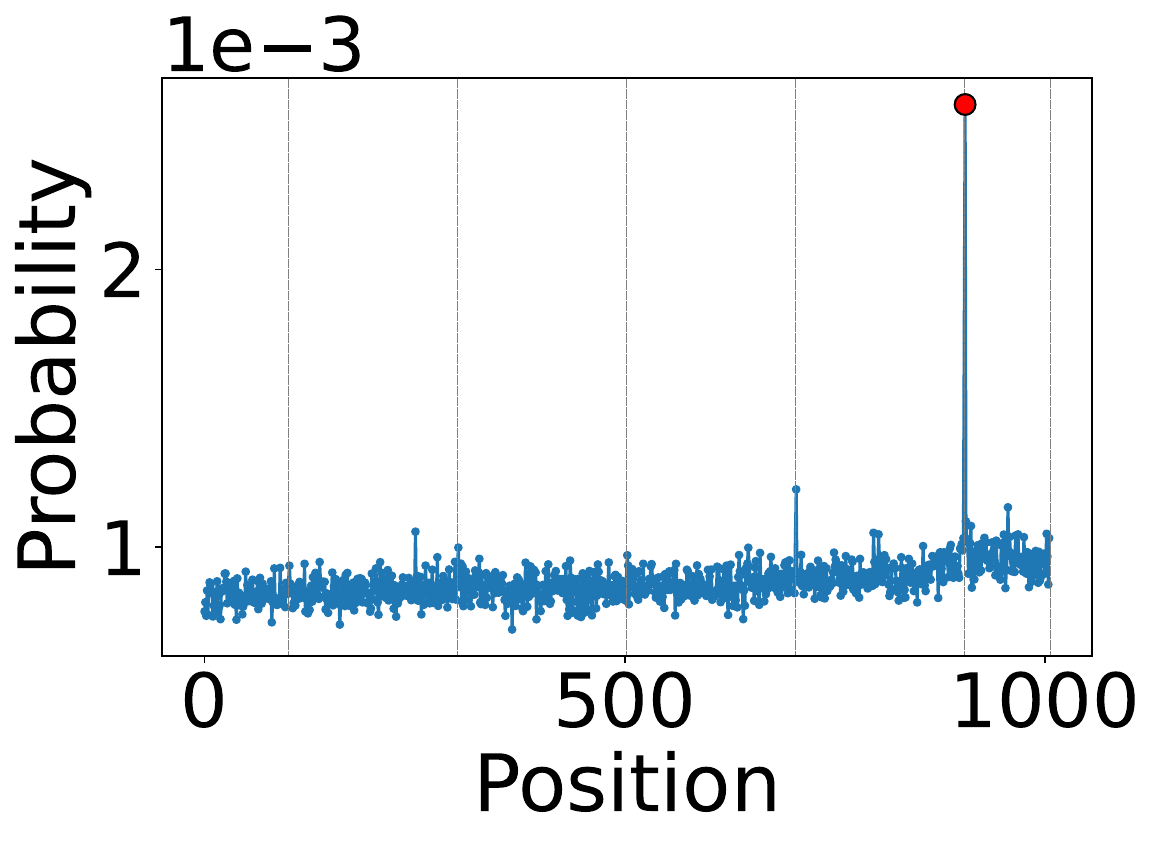} \\

\FigGridVLabel{Mistral} &
\FigGridCell{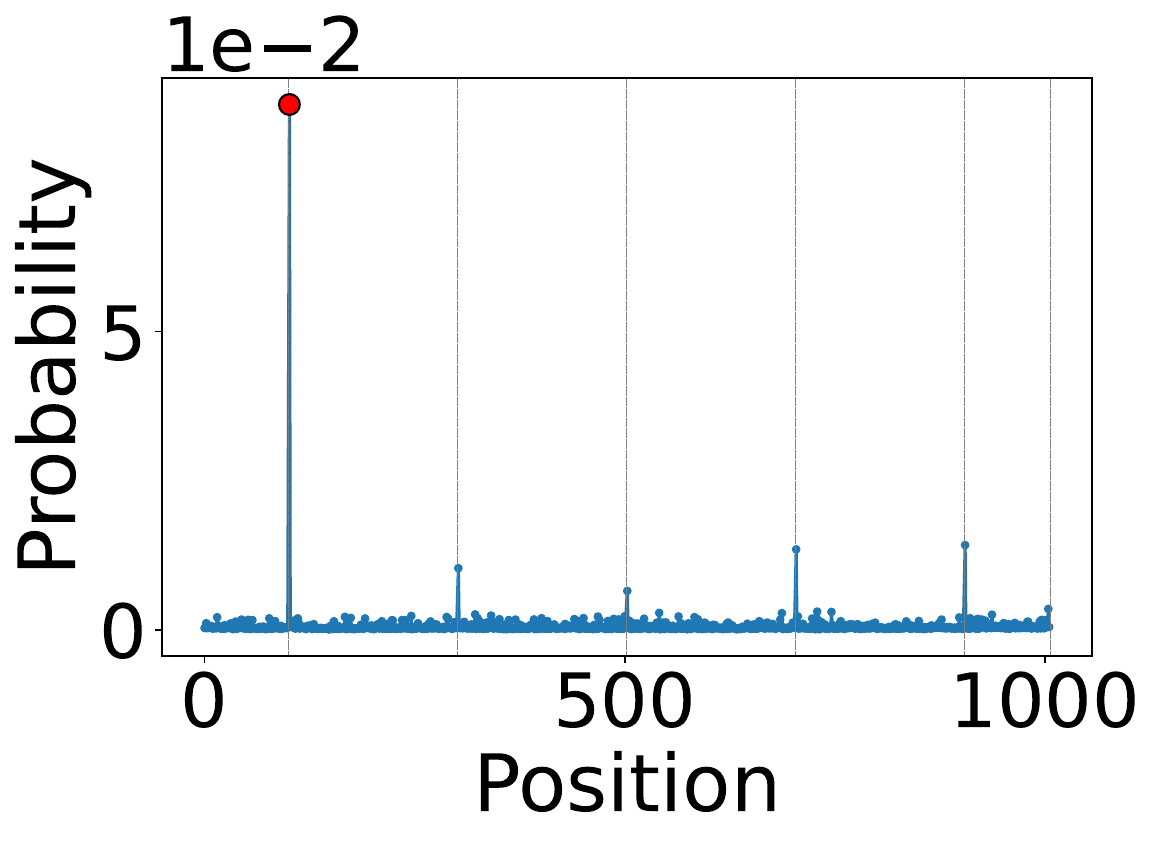} &
\FigGridCell{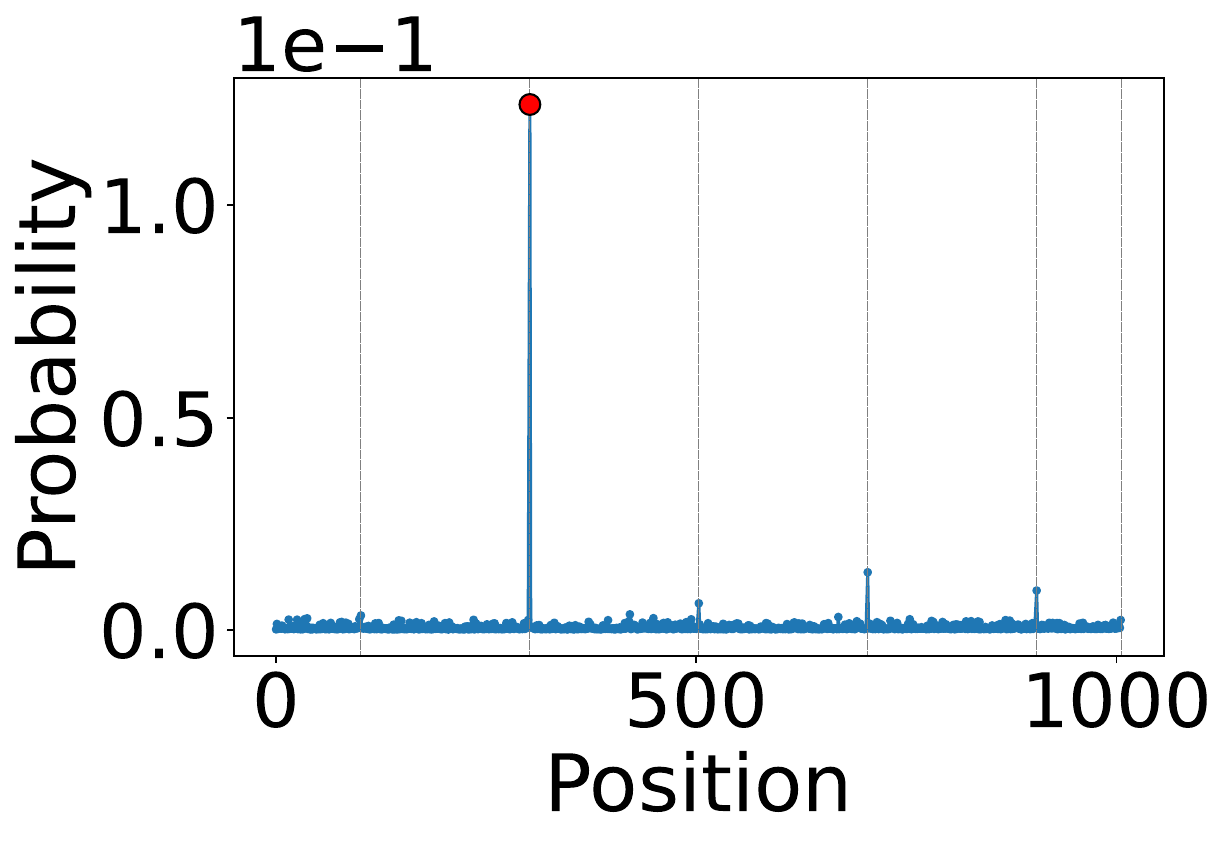} &
\FigGridCell{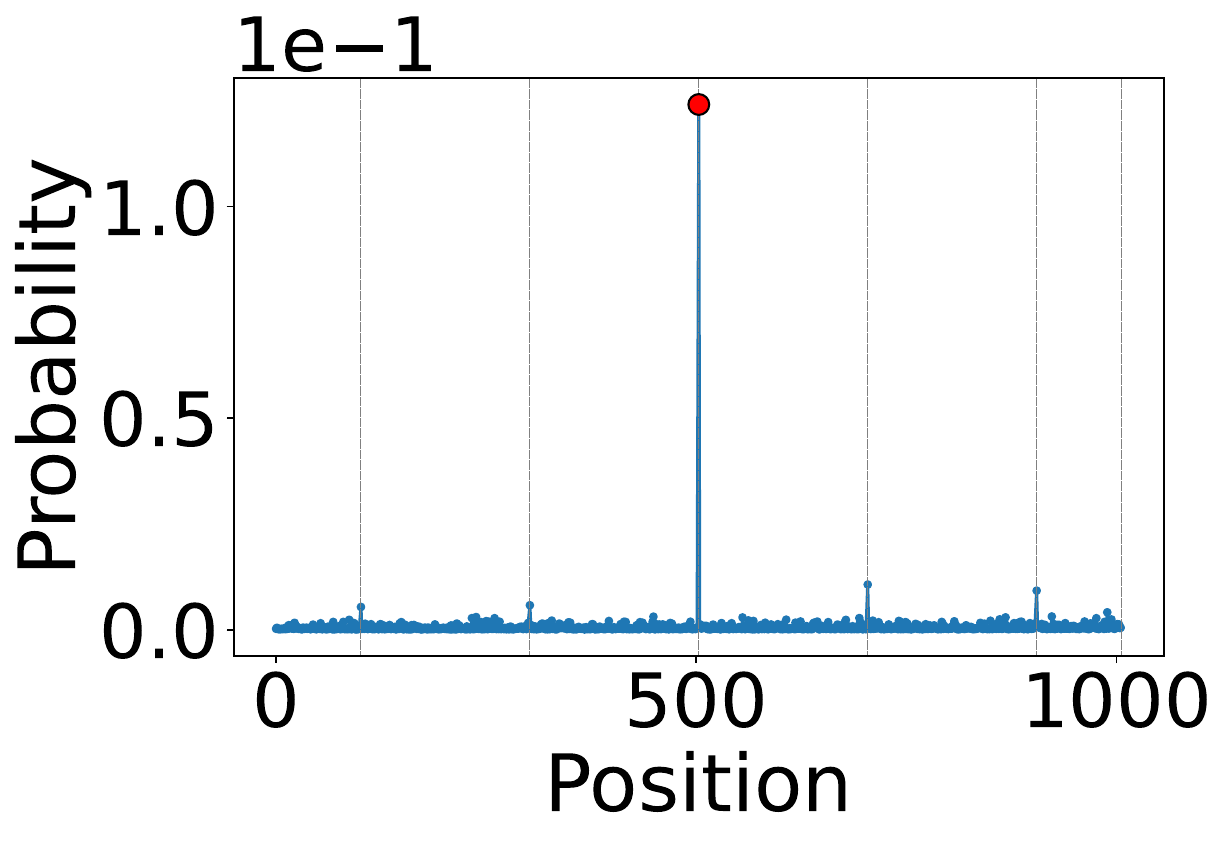} &
\FigGridCell{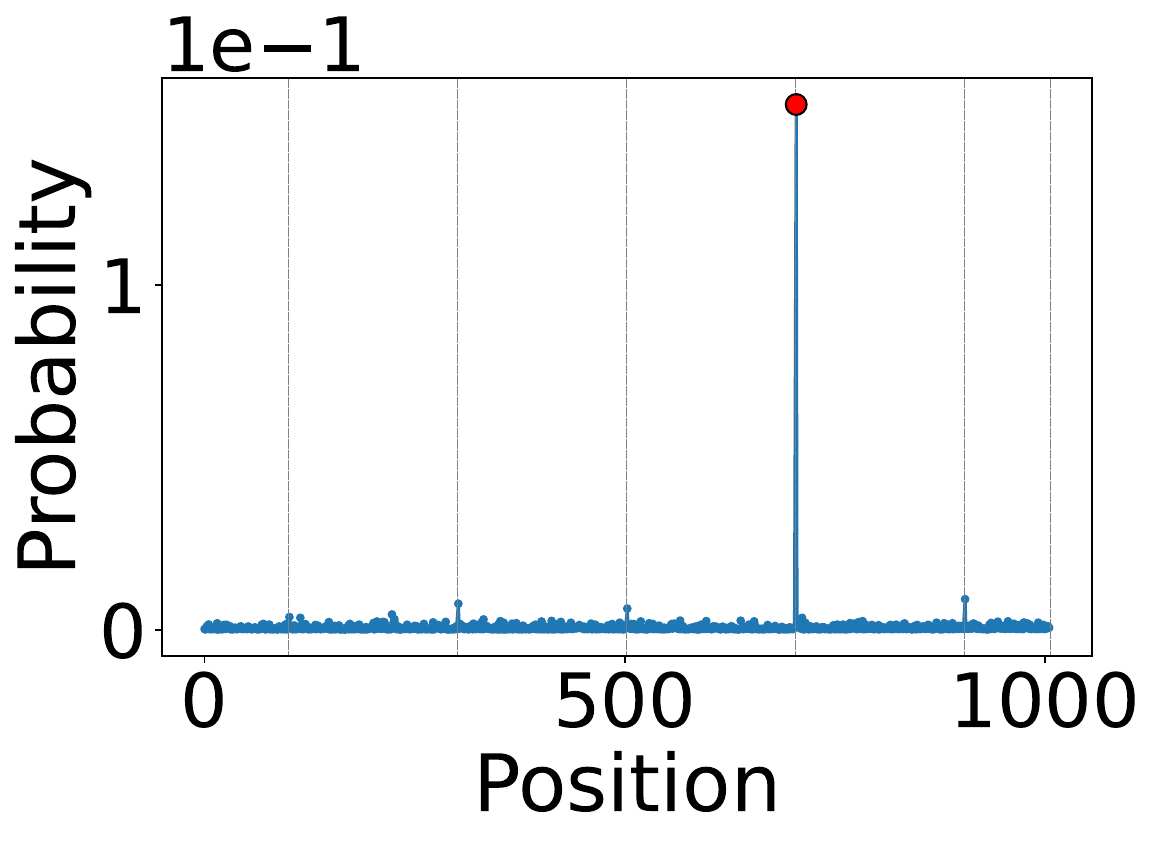} &
\FigGridCell{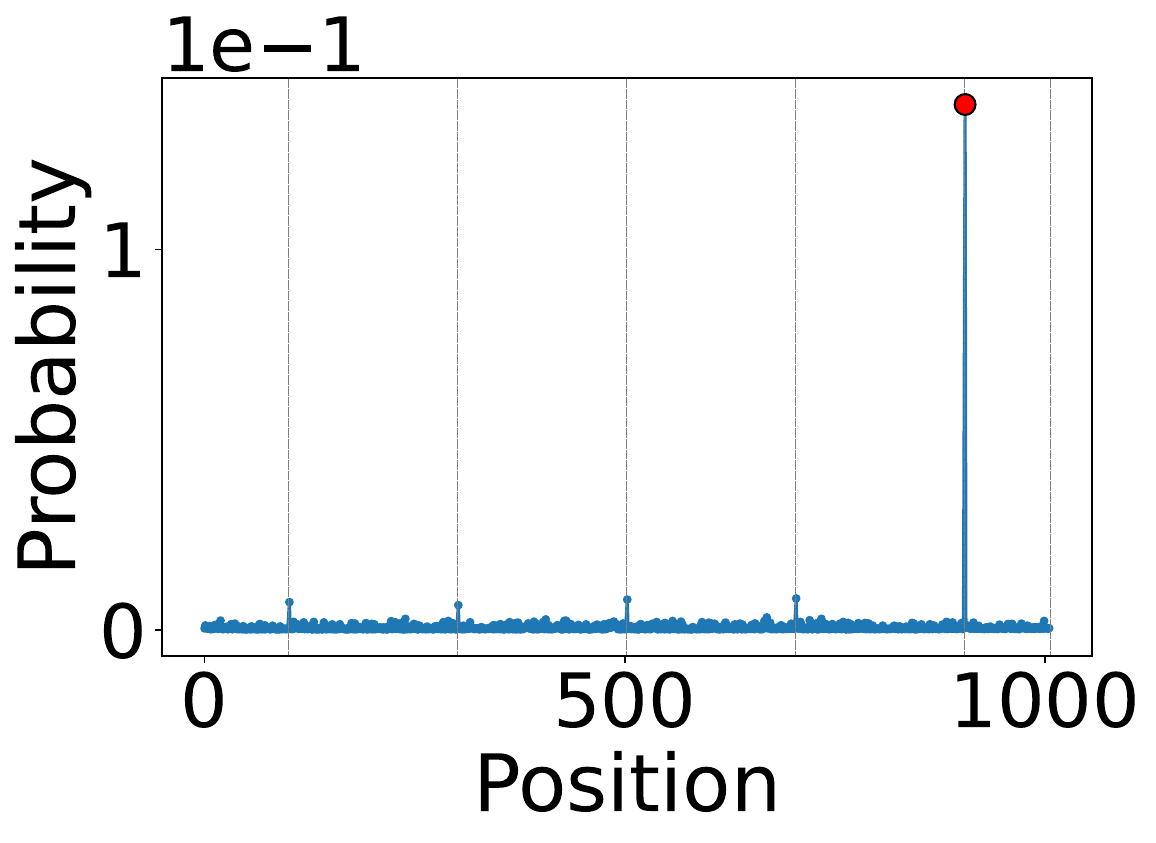} \\

\FigGridVLabel{Qwen} &
\FigGridCell{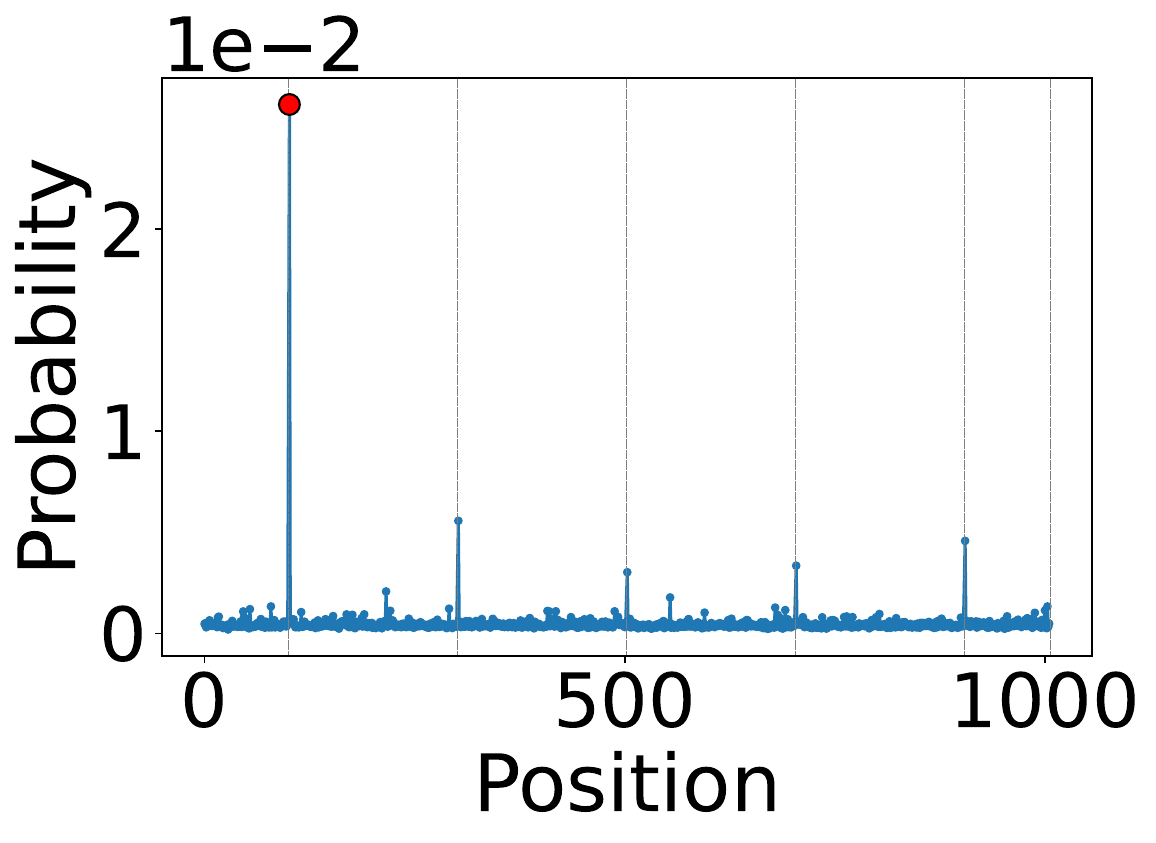} &
\FigGridCell{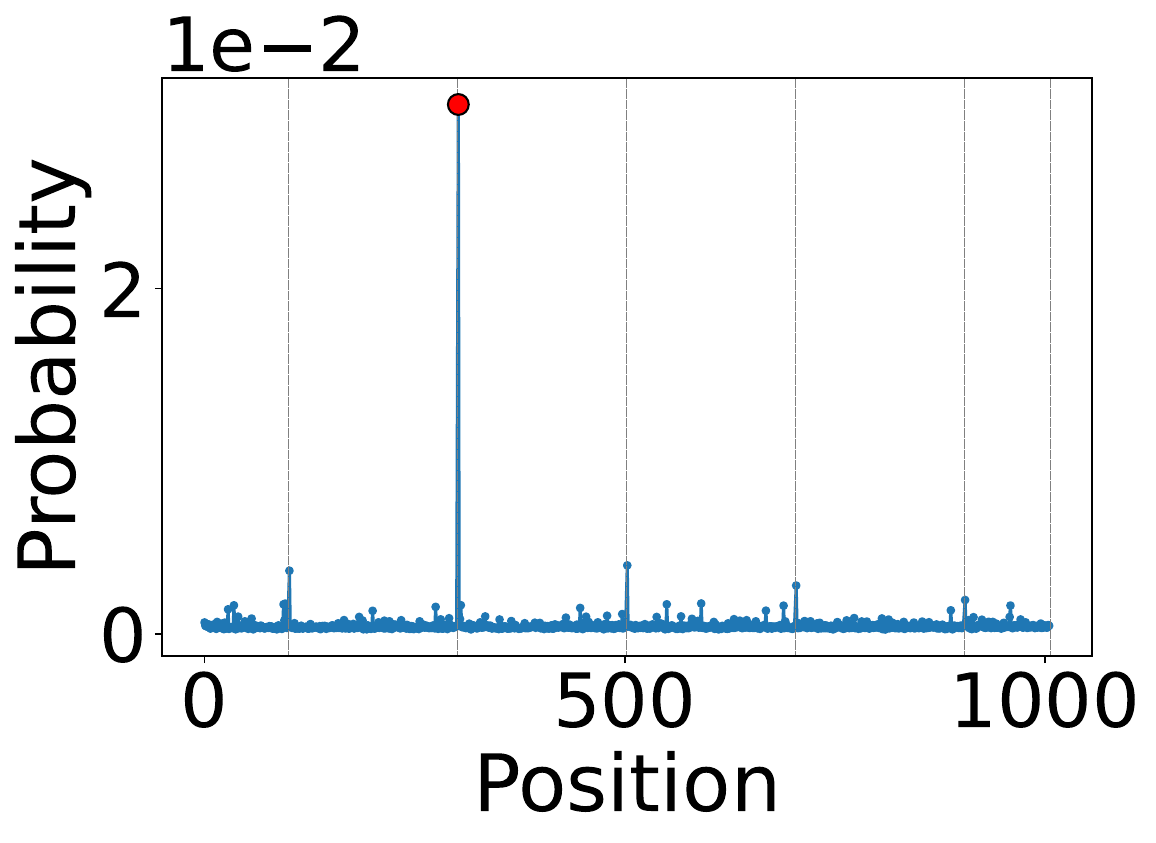} &
\FigGridCell{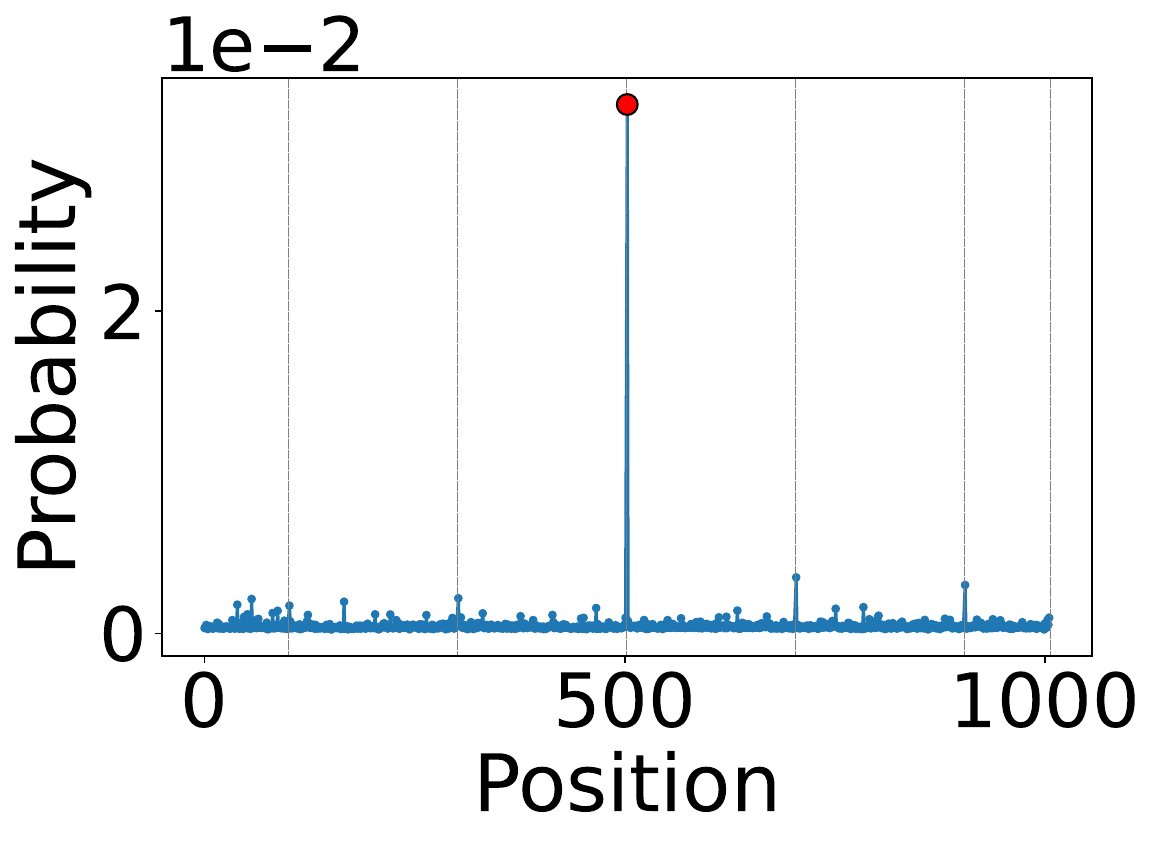} &
\FigGridCell{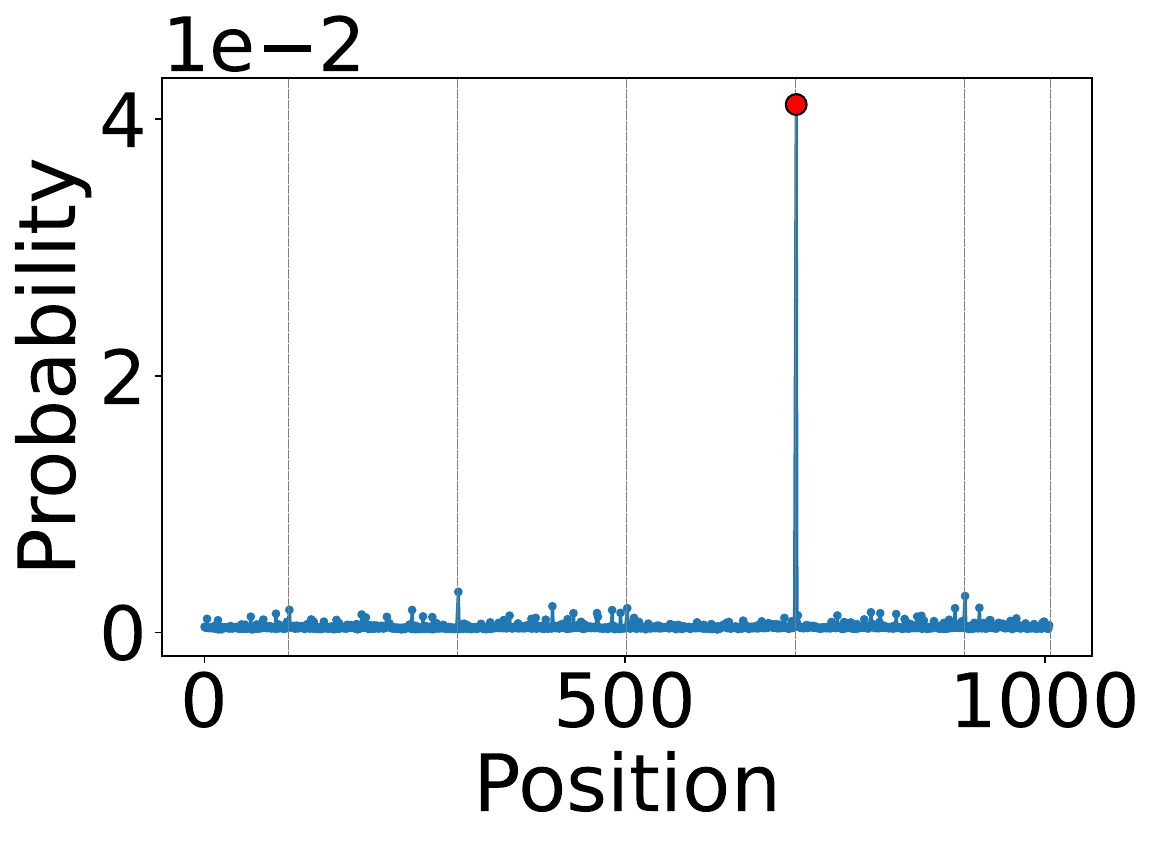} &
\FigGridCell{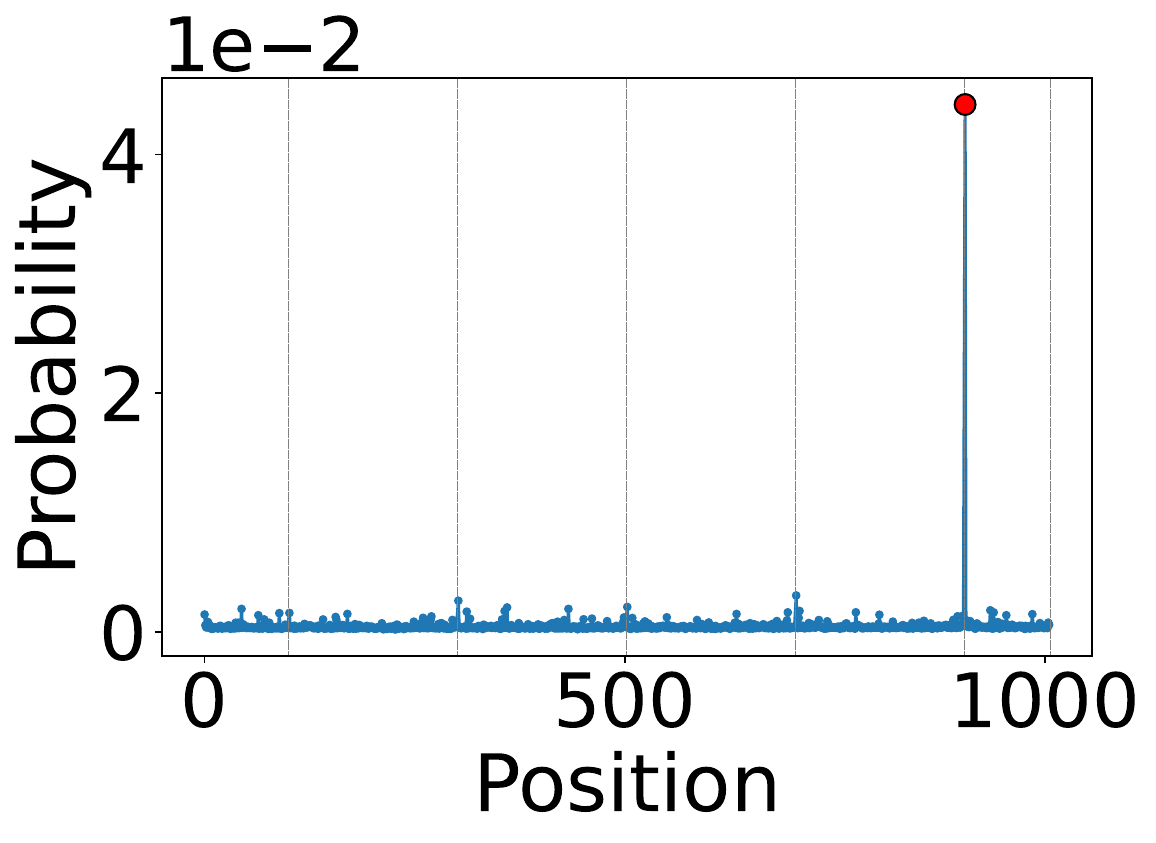} \\

\FigGridVLabel{Gemma} &
\FigGridCell{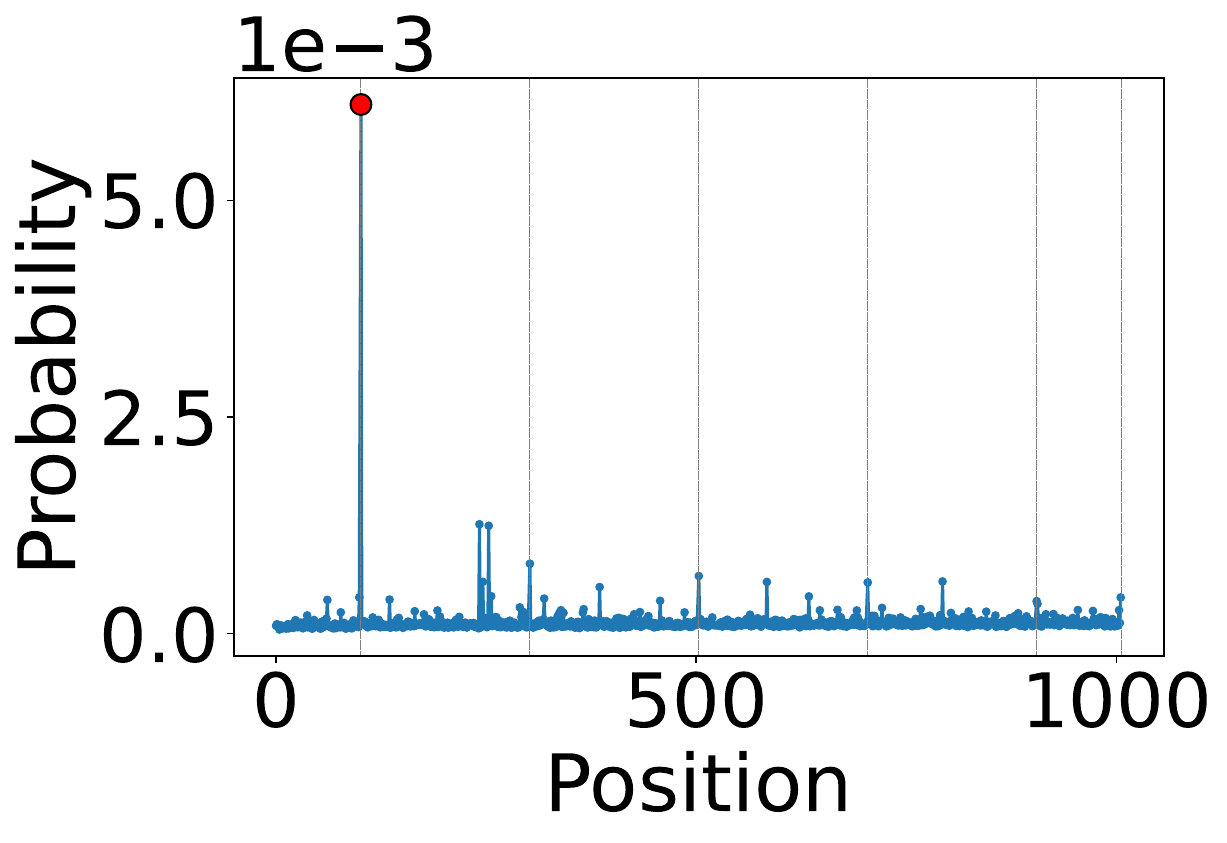} &
\FigGridCell{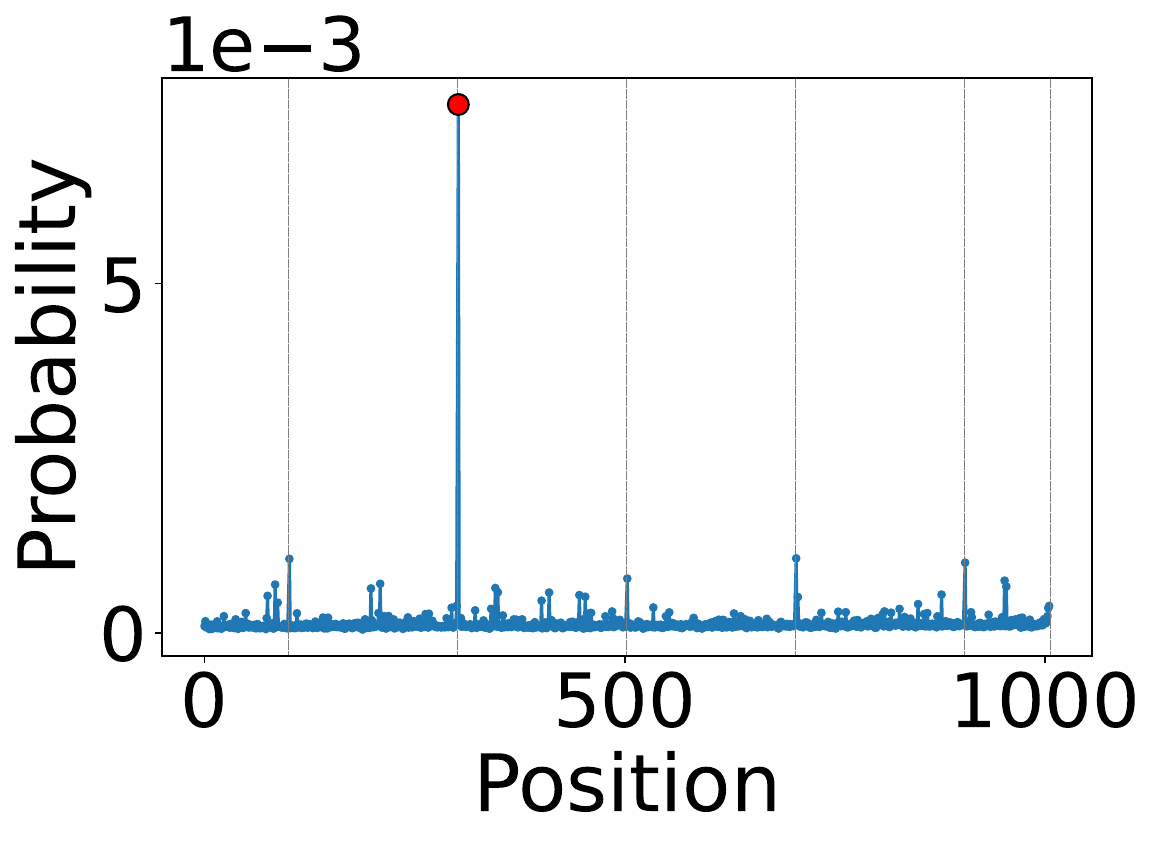} &
\FigGridCell{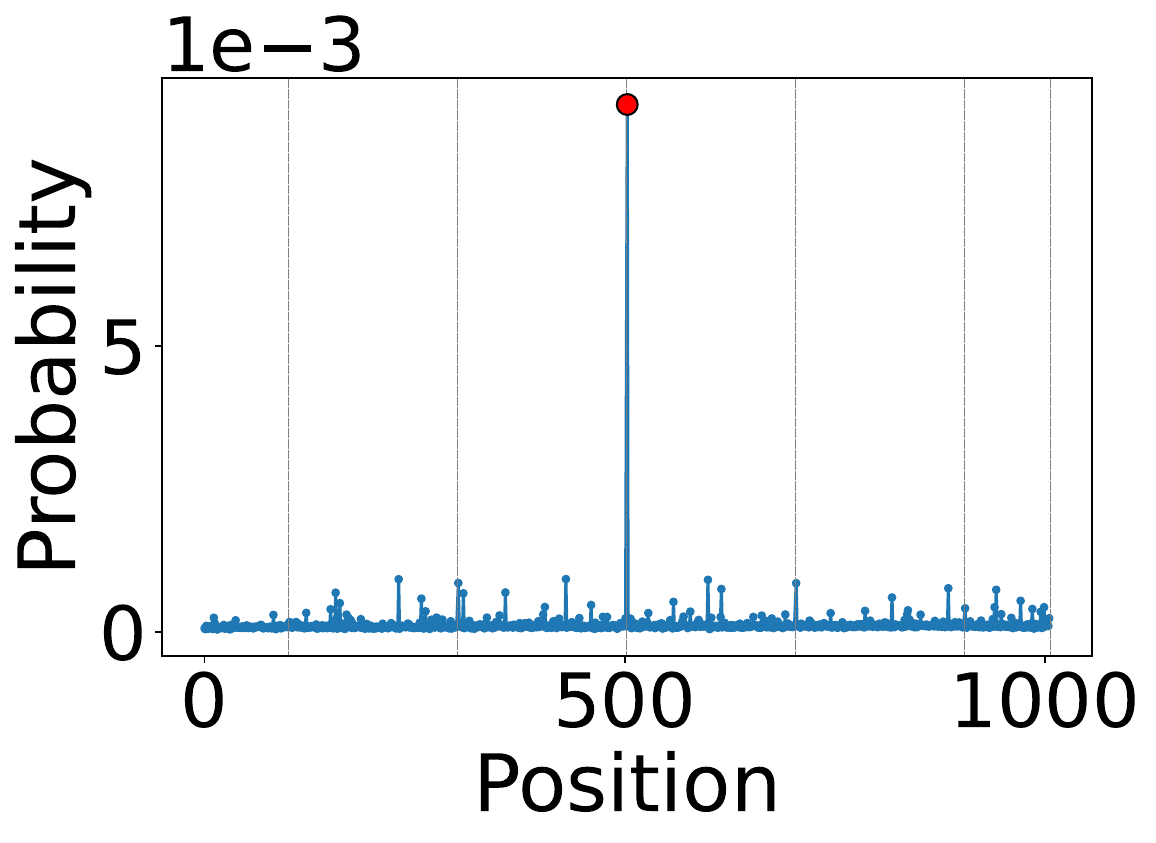} &
\FigGridCell{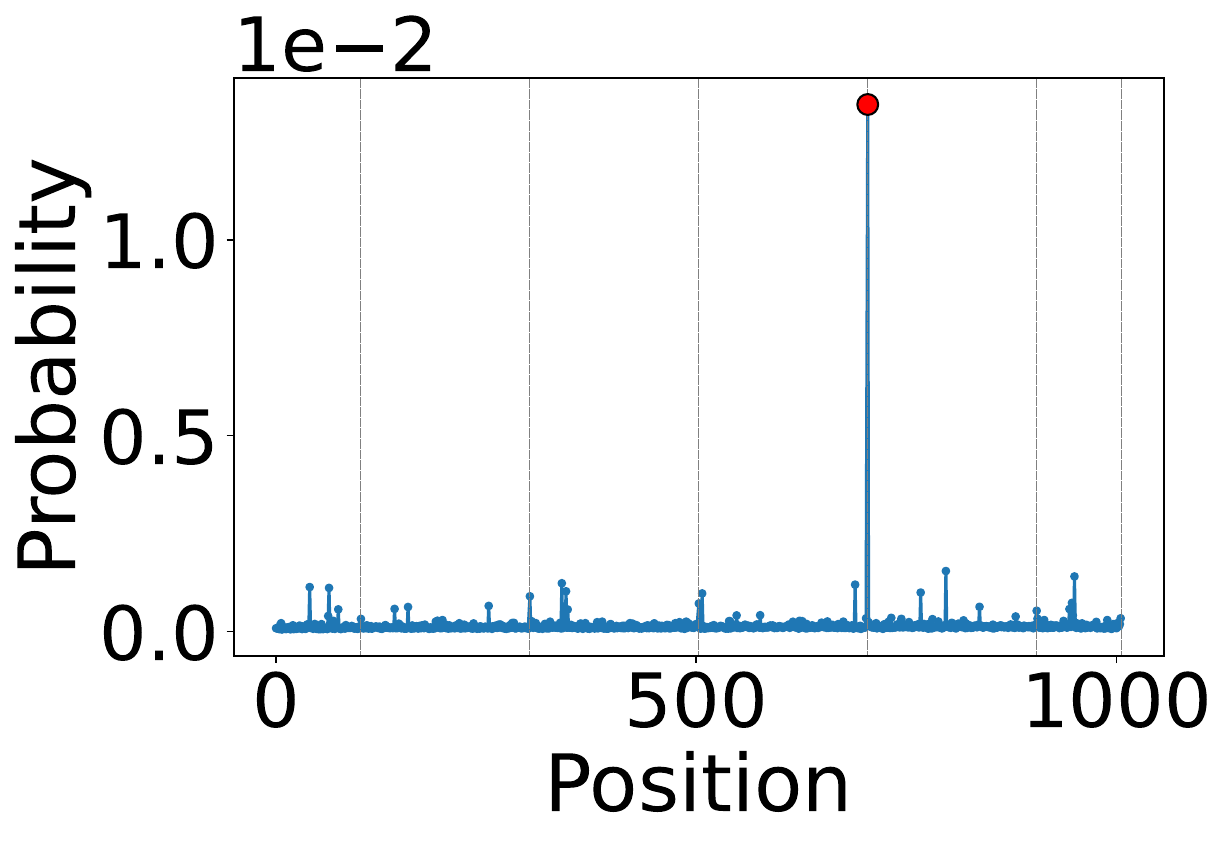} &
\FigGridCell{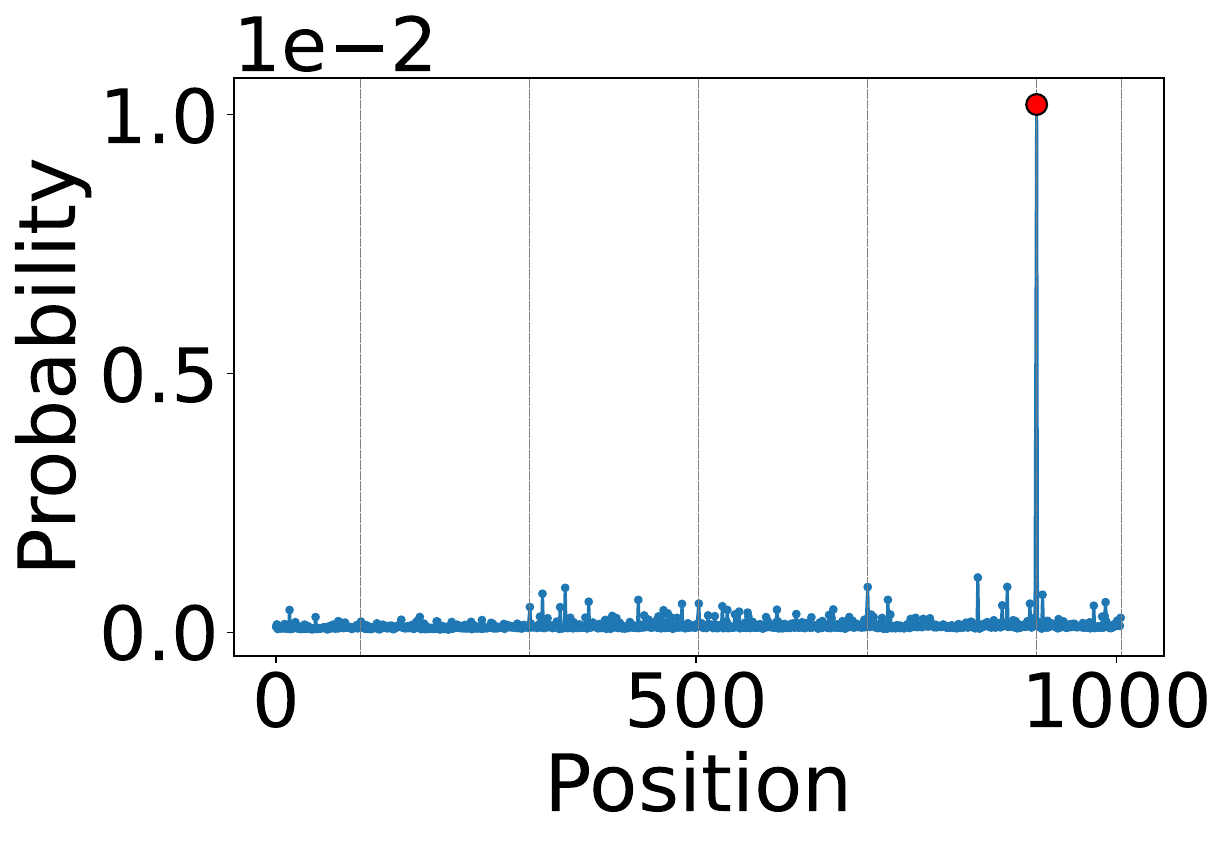} \\

\addlinespace[1pt]
\midrule
\addlinespace[1pt]

\FigGridVLabel{Mamba} &
\FigGridCell{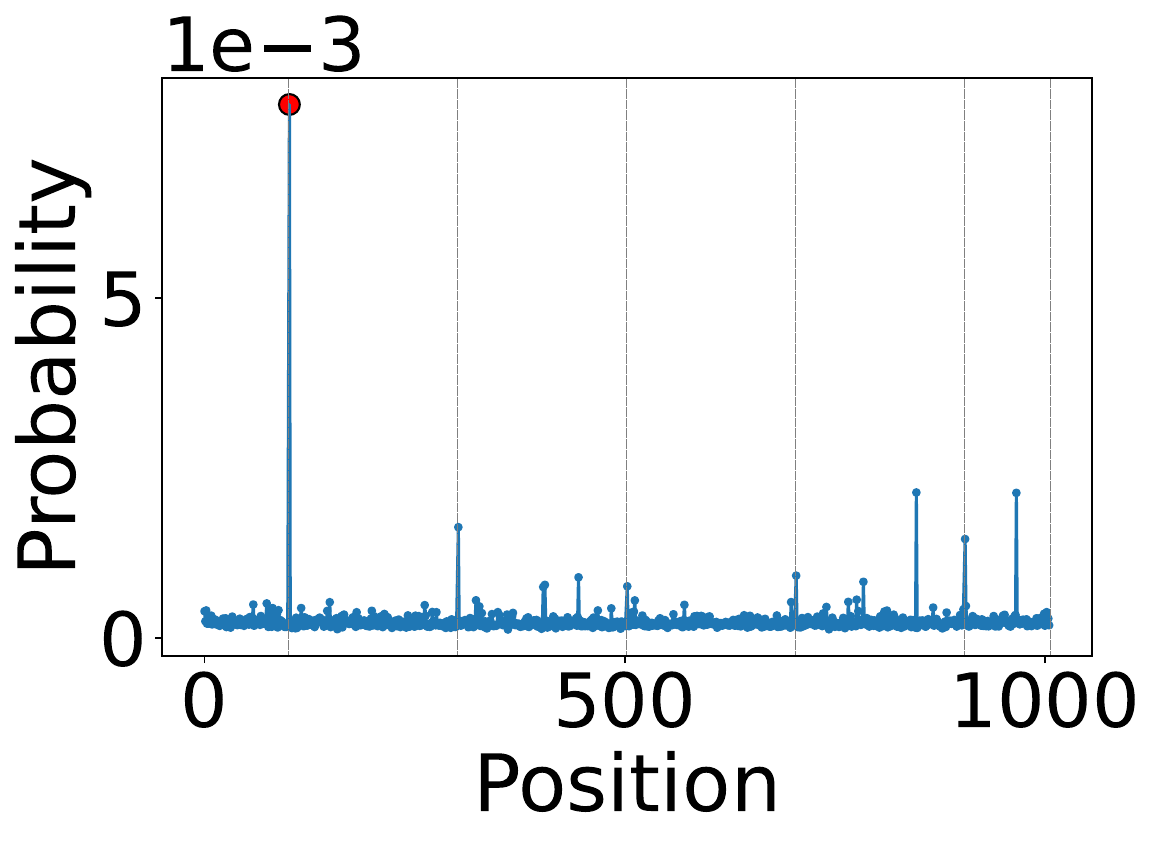} &
\FigGridCell{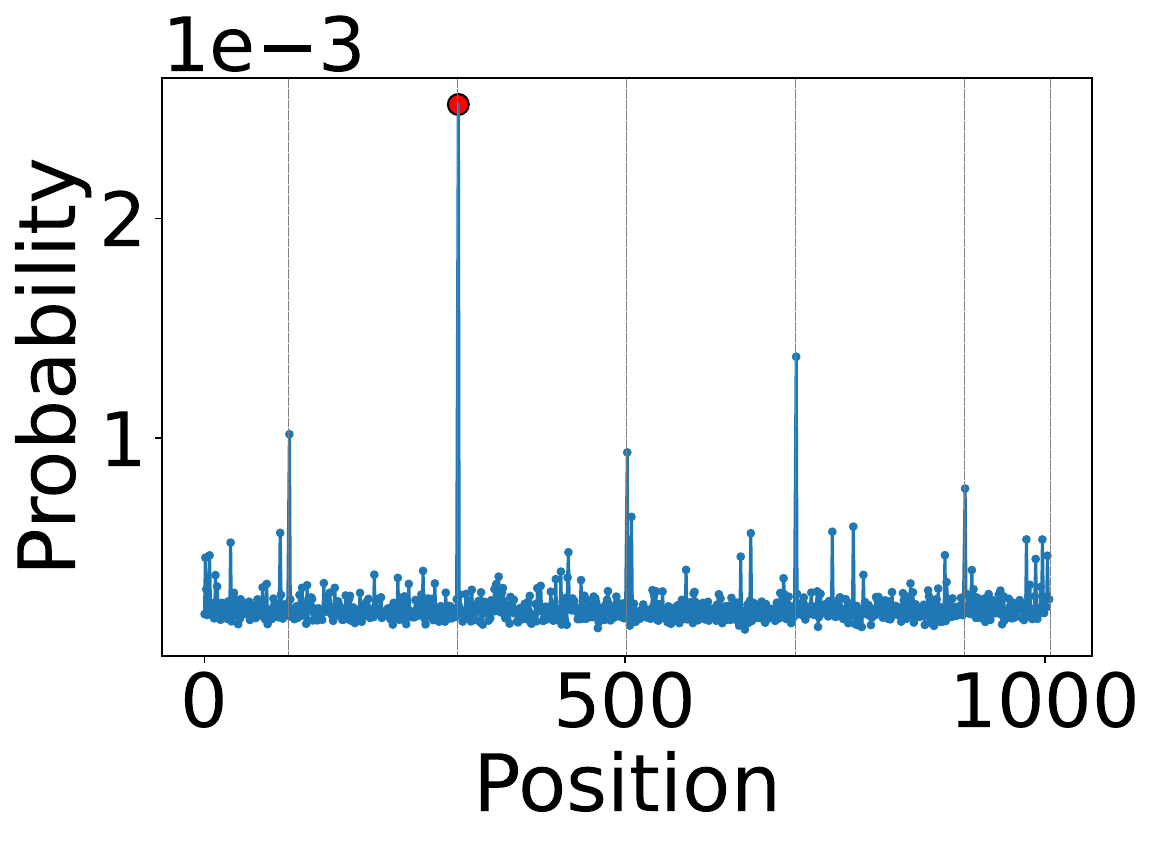} &
\FigGridCell{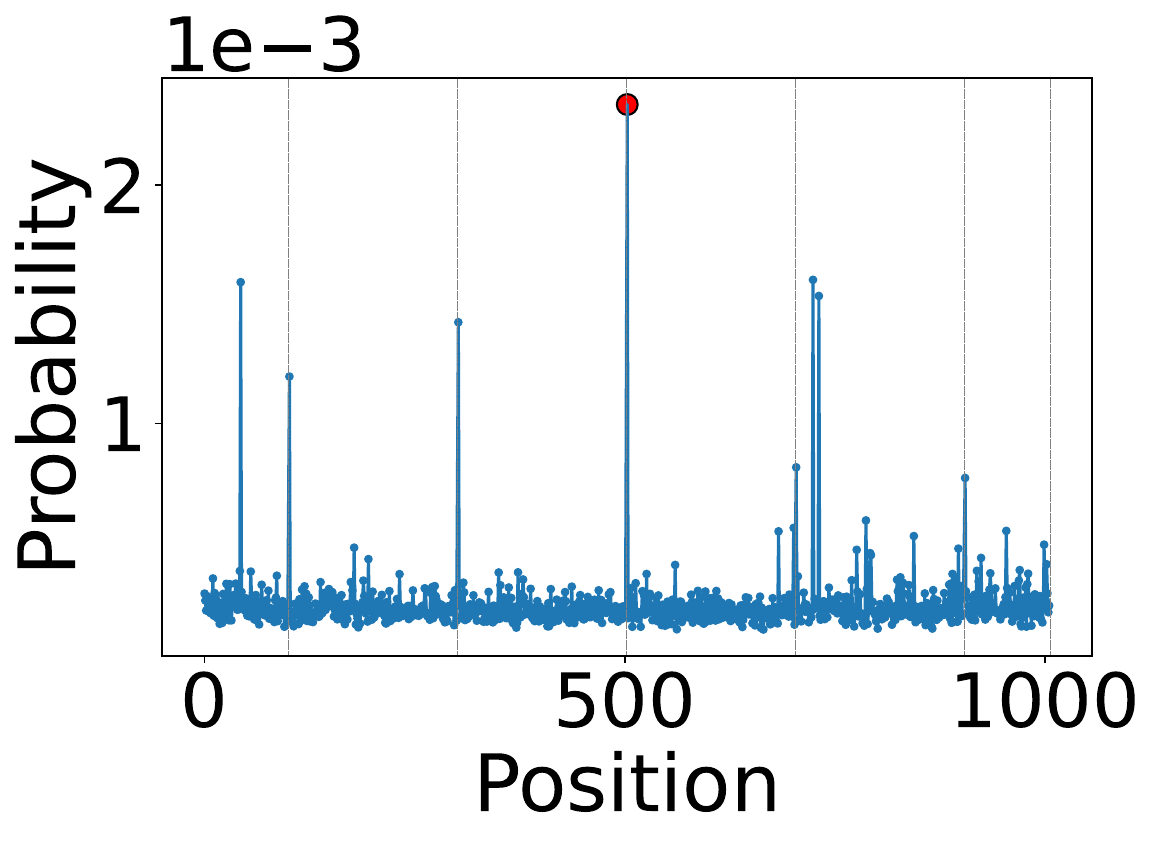} &
\FigGridCell{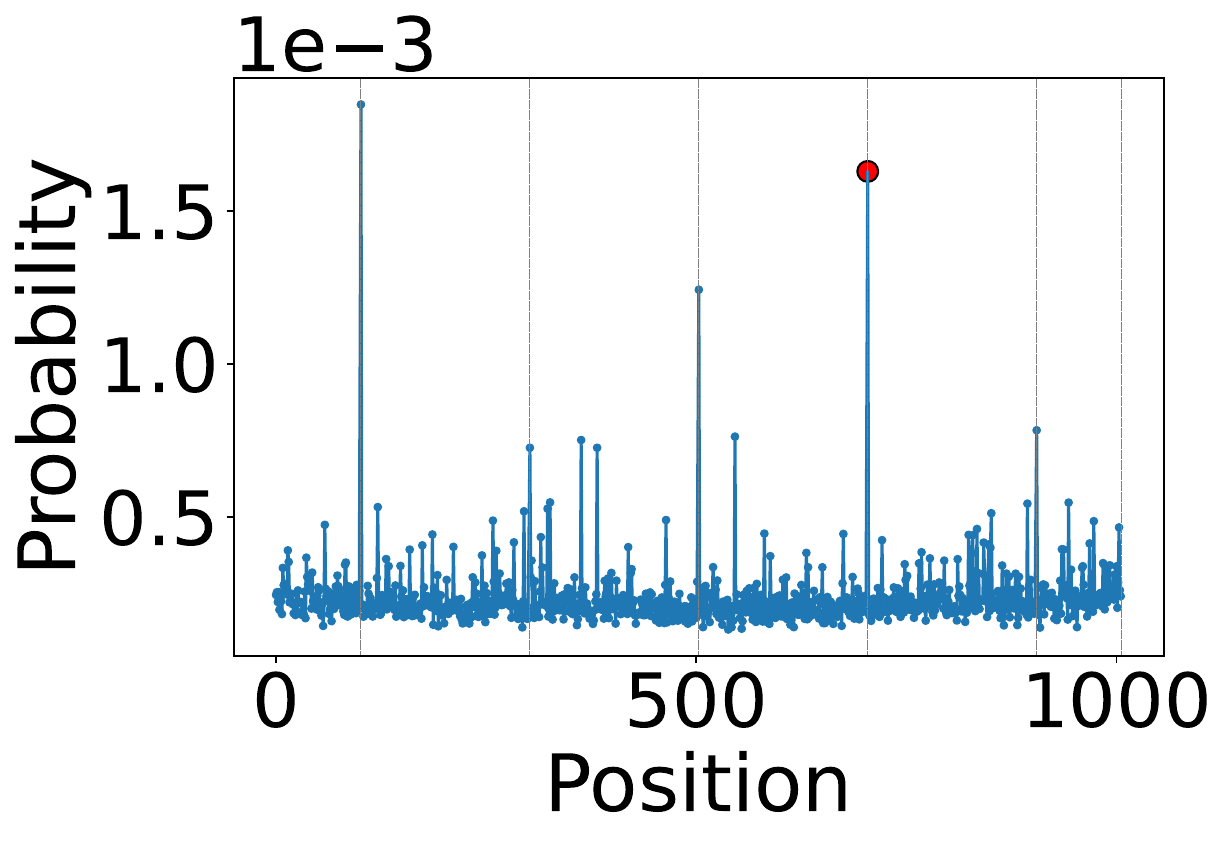} &
\FigGridCell{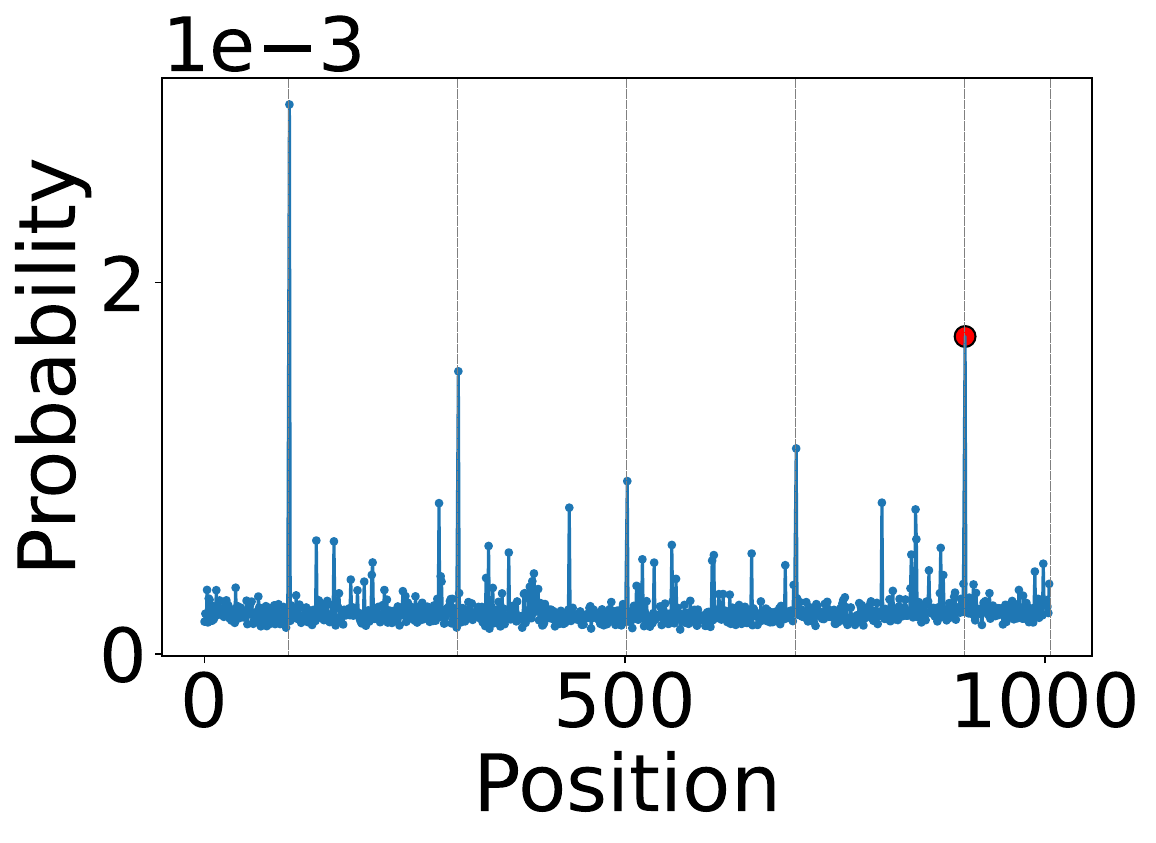} \\

\FigGridVLabel{Falcon-M} &
\FigGridCell{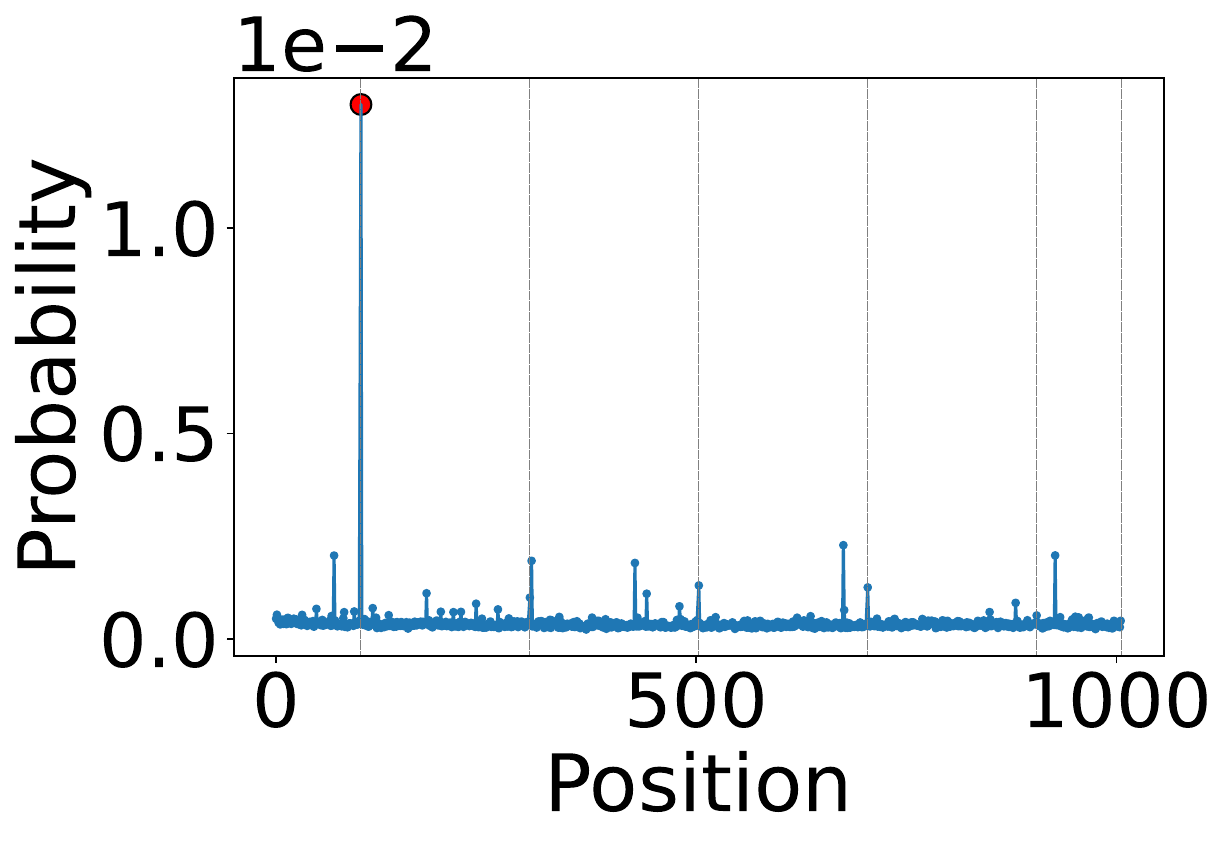} &
\FigGridCell{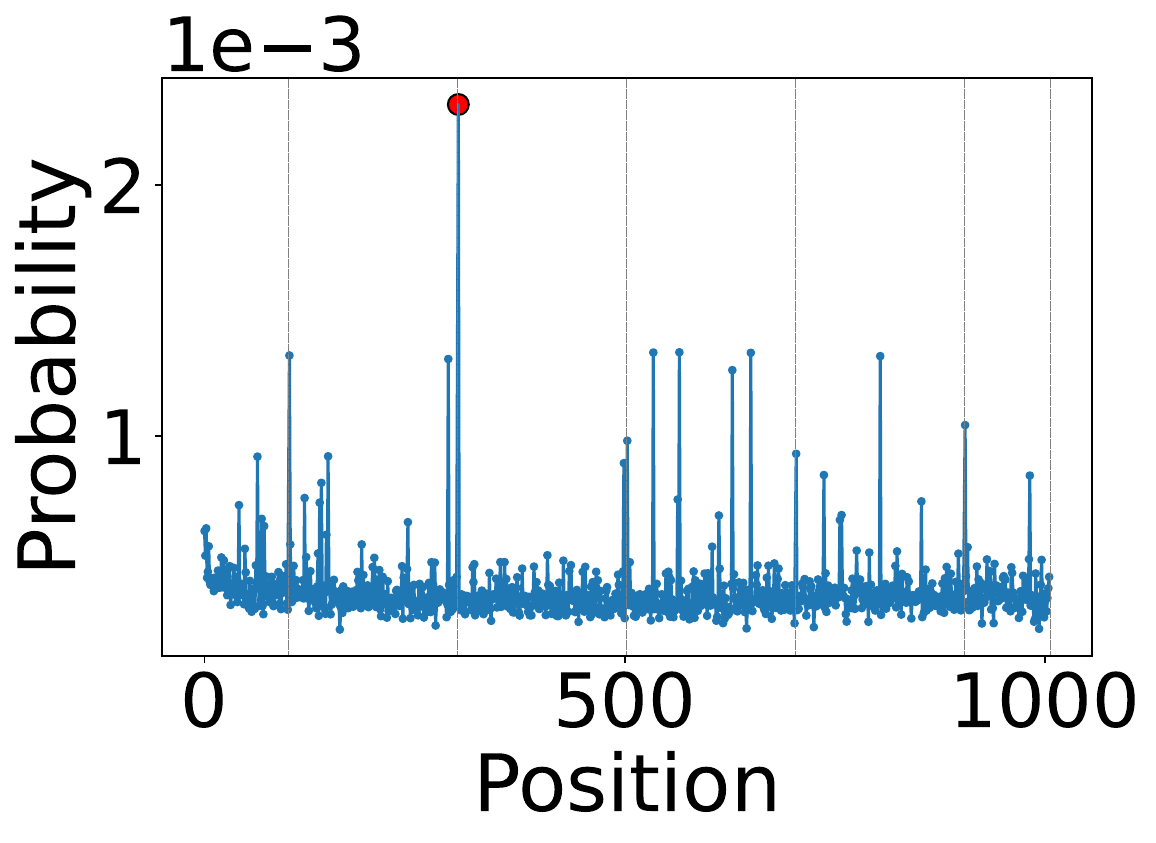} &
\FigGridCell{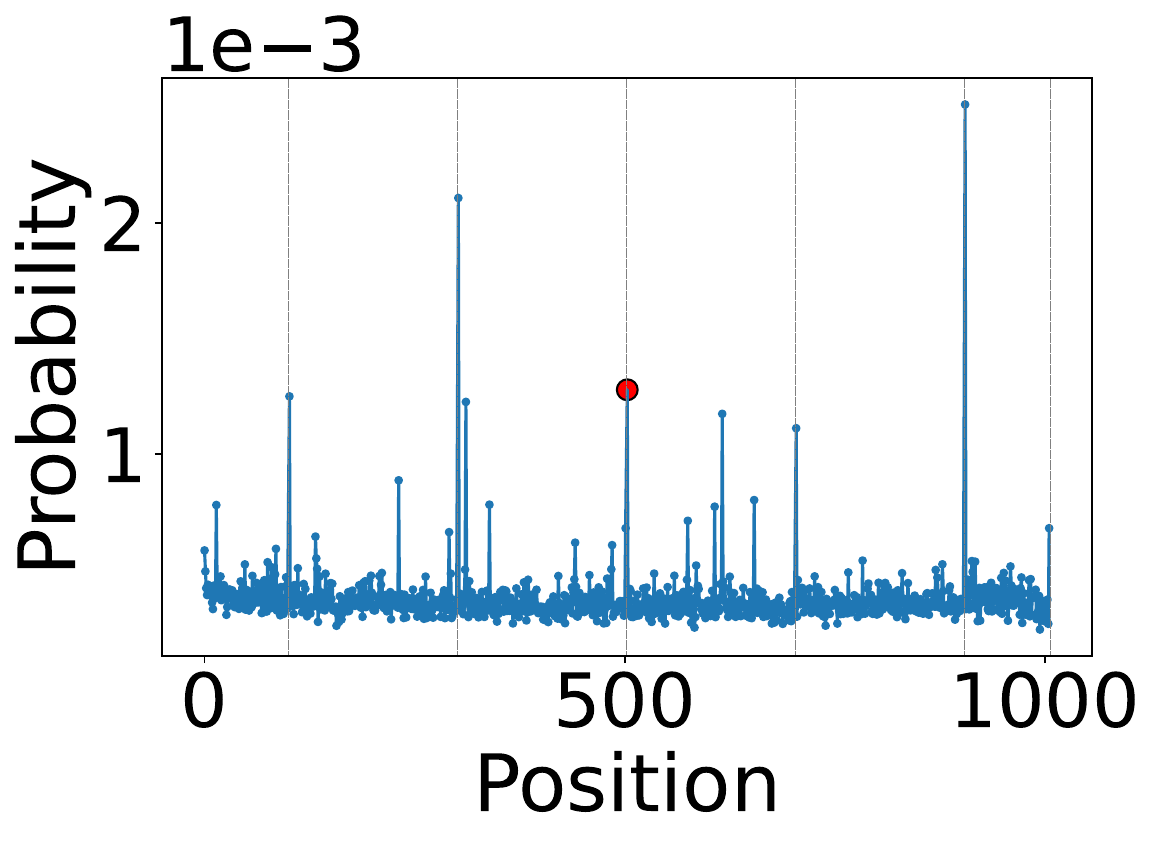} &
\FigGridCell{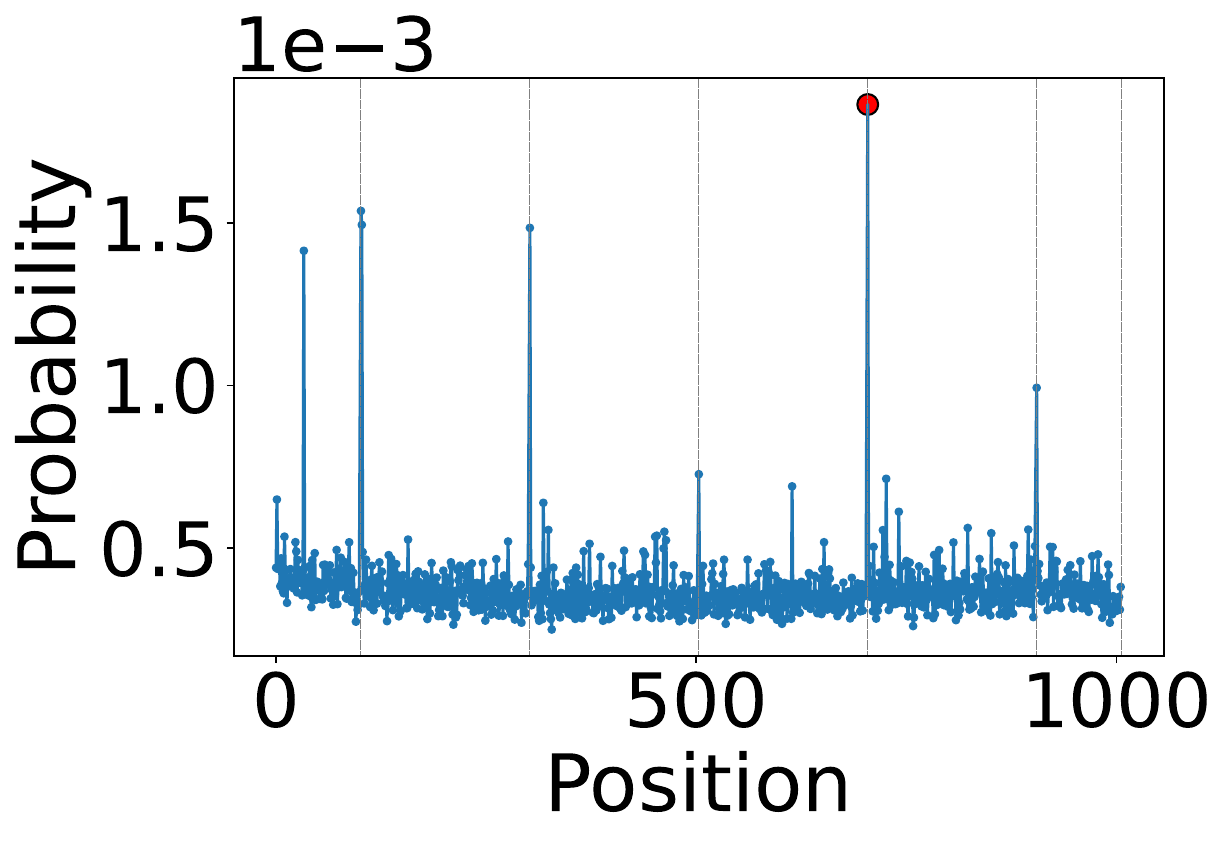} &
\FigGridCell{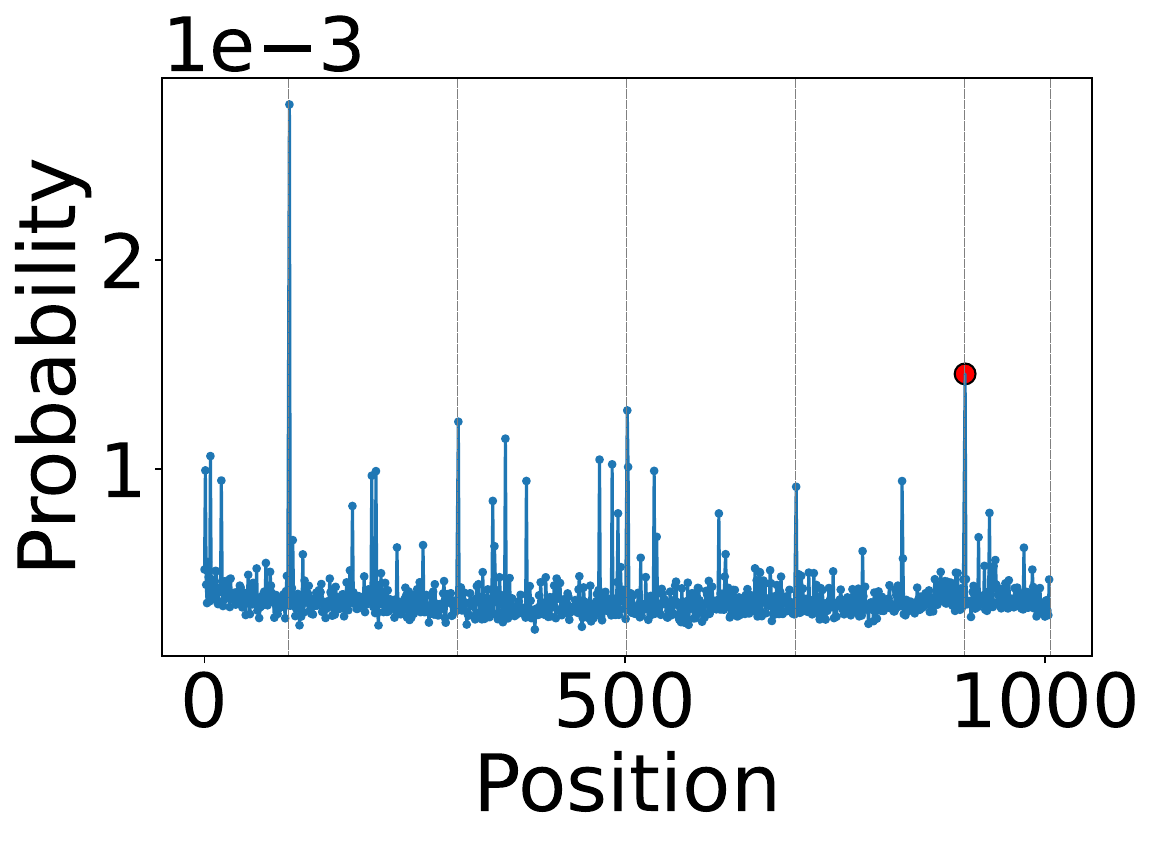} \\

\FigGridVLabel{R-Gemma} &
\FigGridCell{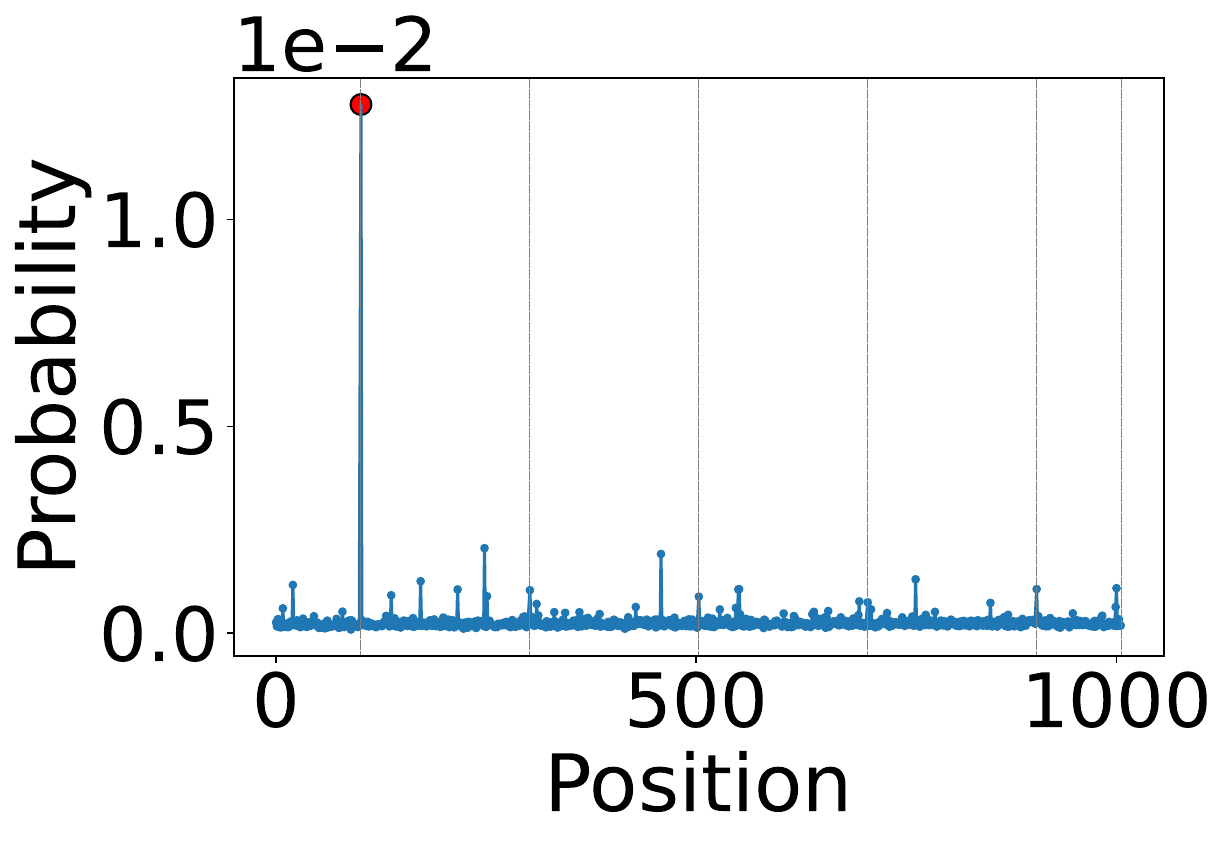} &
\FigGridCell{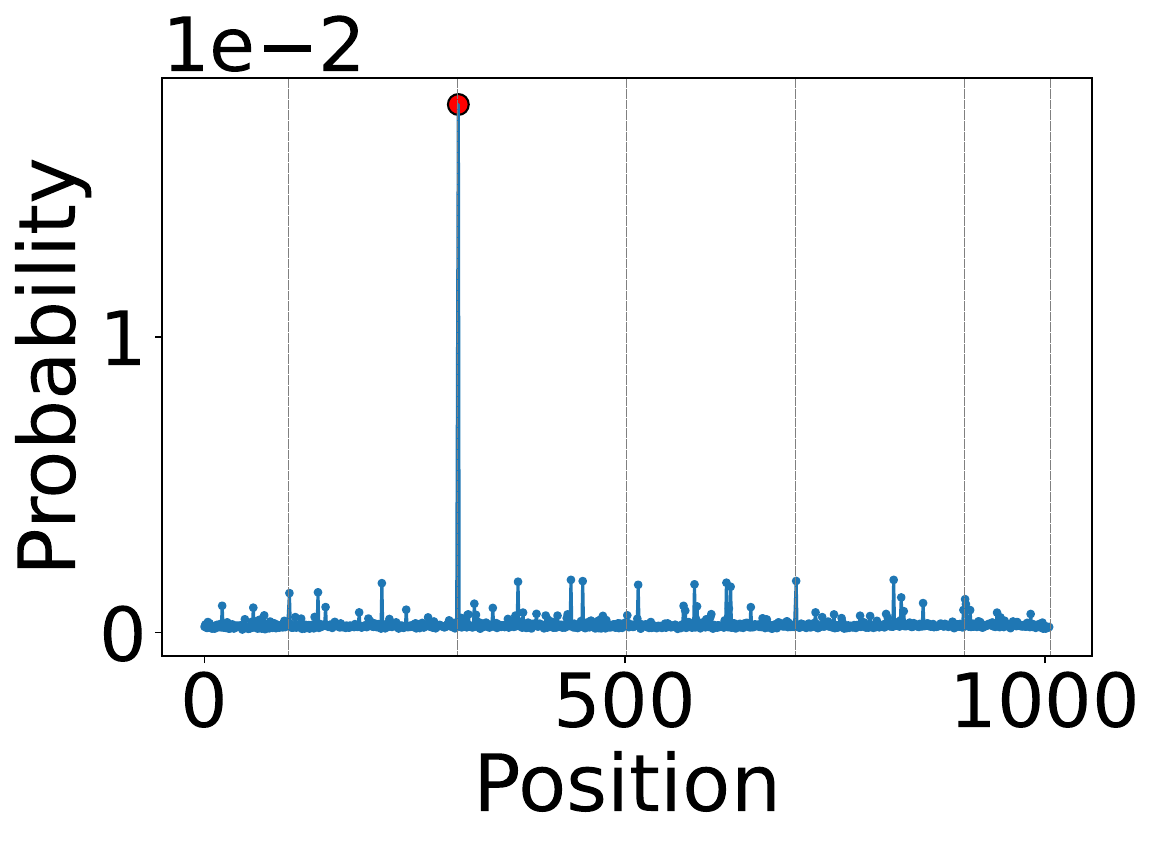} &
\FigGridCell{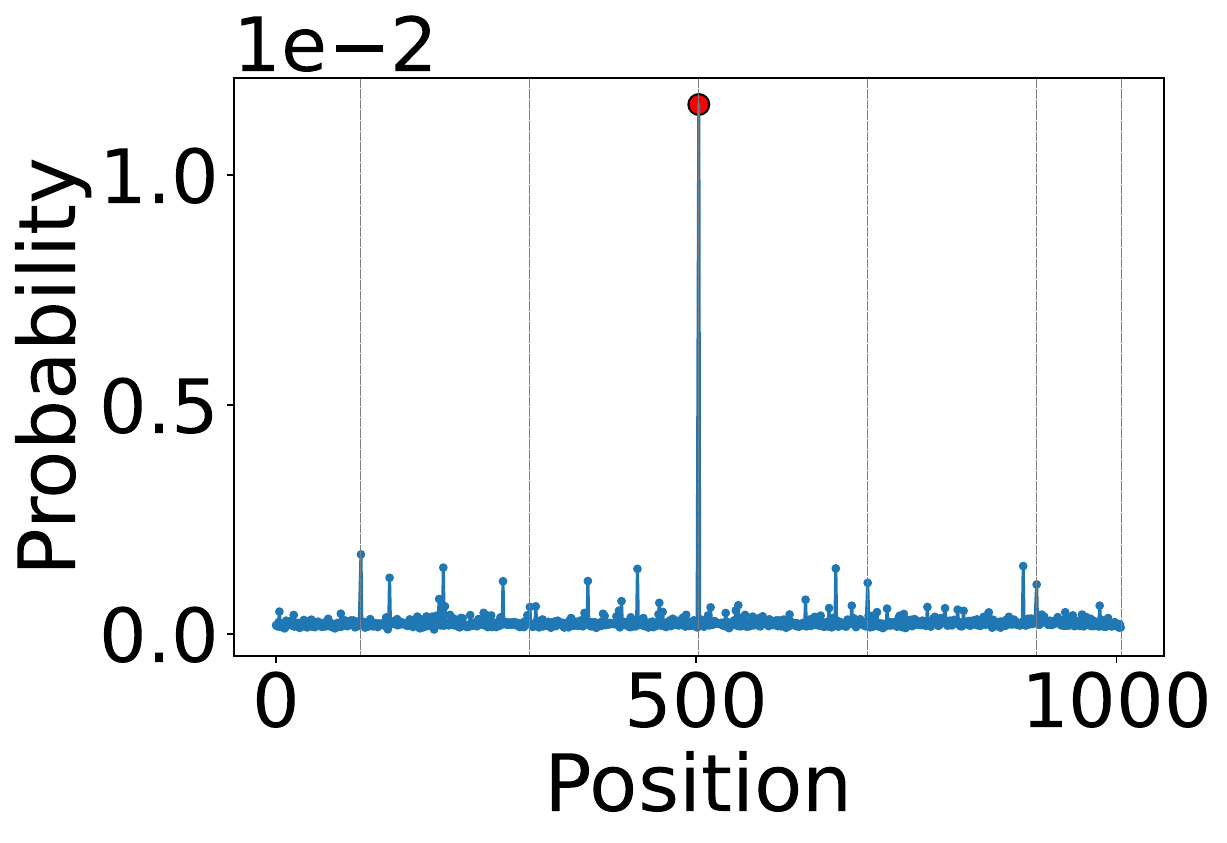} &
\FigGridCell{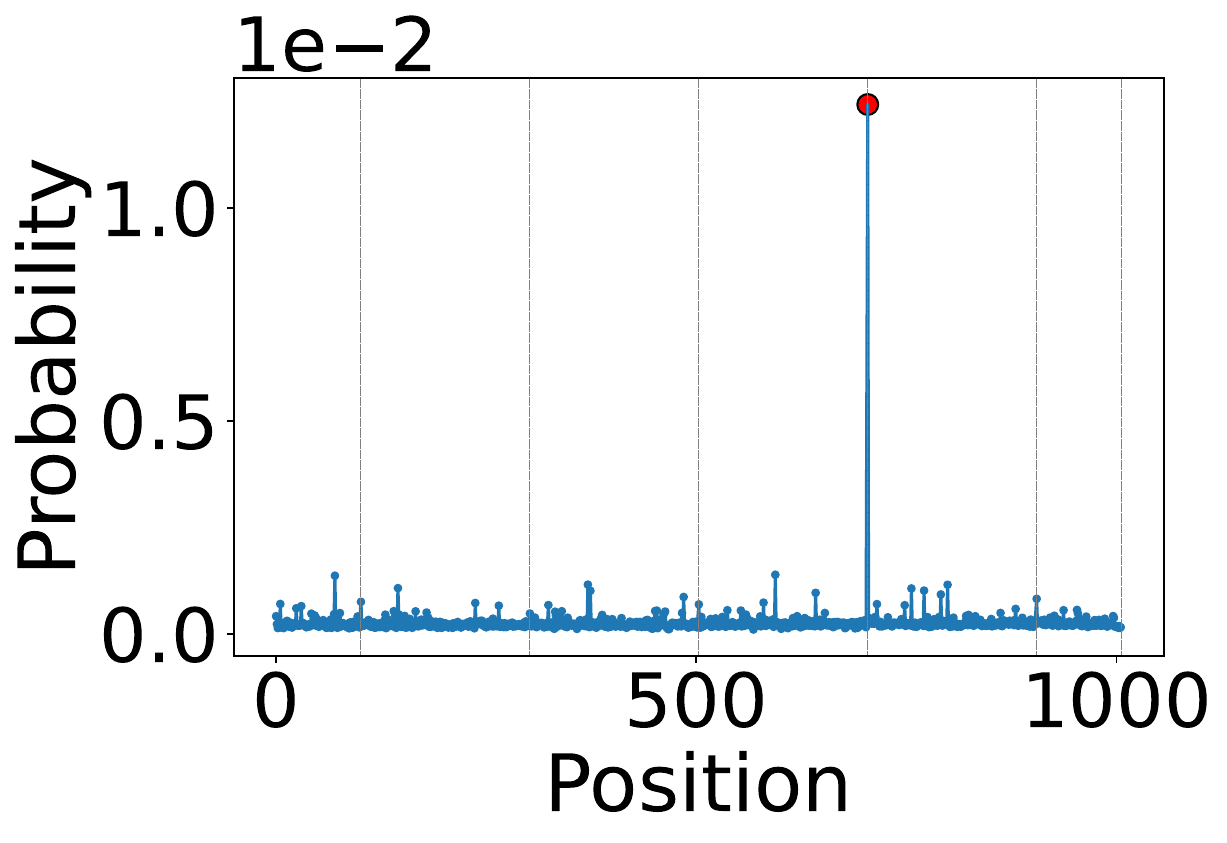} &
\FigGridCell{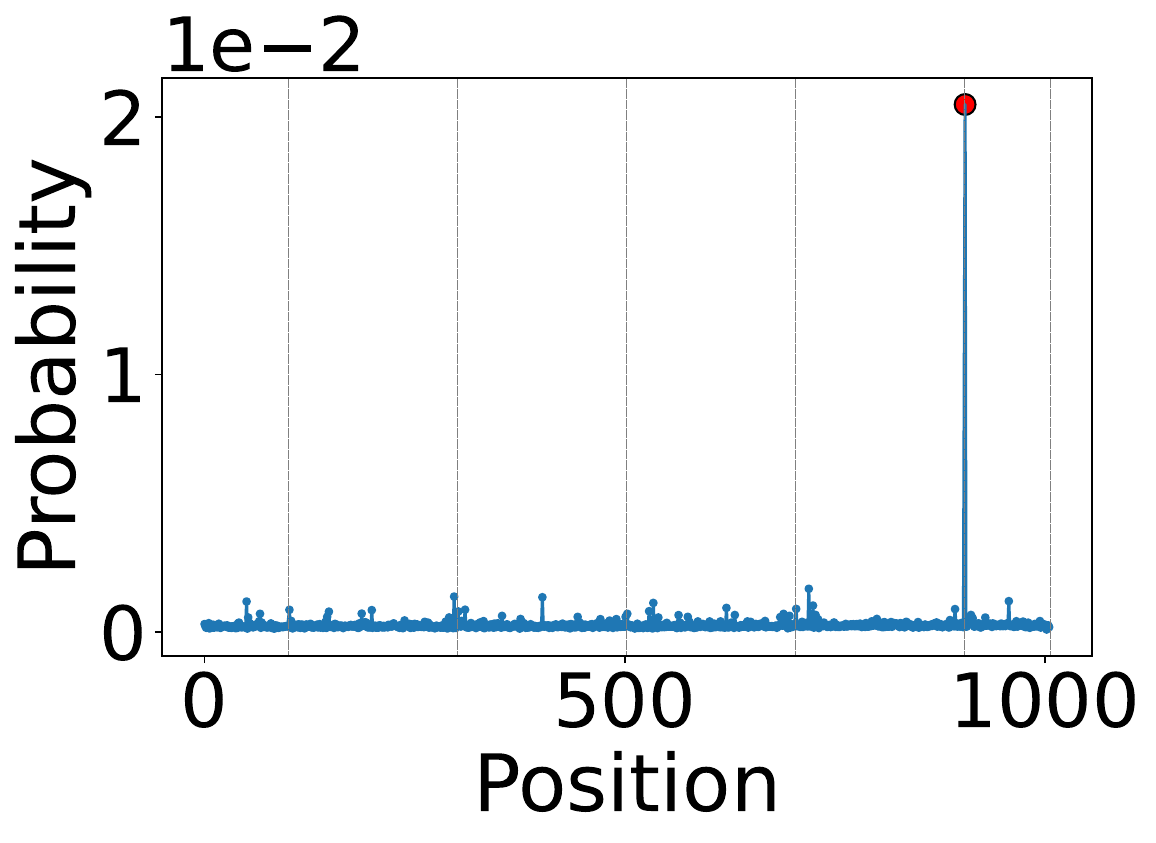} \\
\bottomrule
\end{tabular}
\endgroup
\caption{
Experiment 2: Episodic retrieval results. Probability of potential target tokens after probing episodes at different temporal positions (columns 1-5). Vertical gray lines mark the anchor token. Red dots mark the target token corresponding to the probed episode.  A high probability at the red dot indicates successful retrieval of the episode’s target token.
}
\label{fig:exp2_all_repetitions}
\end{figure*}

\textbf{Prompt Construction} We created prompts containing five distinct episodes. Each episode consisted of a unique `context' token, followed by the `anchor' token (same `A' across all episodes), followed by a unique `target' token (e.g., `BAH', `CAF', `XAM', `GAD', `RAP'). These five episodes were embedded within longer sequences of random tokens, separated by 200 random tokens each. The final probe consisted of the context token and fixed token from one specific target episode (e.g., ending the prompt with `...XA'), Figure~\ref{fig:schematic}, right.

\textbf{Quantification} The models were evaluated on their ability to predict the token following the probe pair (e.g., predict `M' given `XA'). Successful retrieval requires distinguishing the target episode (`XAM') from the others sharing the fixed token `A'. We systematically varied which of the five episodes was used for the probe (testing retrieval from positions 1 through 5). Probabilities were averaged over 500 permutations with shuffled intervening random tokens to isolate temporal effects.

\textbf{Results} Figure~\ref{fig:exp2_all_repetitions} displays the next-token probabilities following the probe, plotted against the five possible target episode positions (columns 1-5). Each plot shows the probabilities assigned to the five potential target tokens (`H', `F', `M', `D', `P' in the example) and other tokens at the relevant positions (the fixed token `A' probability is omitted).

When probing for the first episode (Column 1, episode near the start), most models correctly assign the highest probability to the target token corresponding to that episode (e.g., `H' for probe `BA'), indicating successful retrieval. This pattern holds when probing for subsequent episodes (Columns 2-5). Llama, Mistral, Qwen, Gemma, and RecurrentGemma consistently assign the highest probability to the correct target token across most positions. However, Mamba and Falcon-Mamba show less robust retrieval, particularly when the target episode is closer to the end of the prompt (e.g., column 5).

Beyond the highest peak for the target episode, smaller peaks corresponding to the target tokens of non-probed episodes are often visible, indicating interference or similarity matching based on the shared fixed token `A'. The magnitude of the correct target peak often varies with position, generally being strongest for episodes nearer the end of the prompt (recency bias) and weaker for earlier episodes in models like Llama, Mistral, Qwen, and RecurrentGemma.

\subsection{Ablation Study}

\subsubsection{Transformers: ablating high-induction-score heads}

To investigate the mechanisms underlying these temporal effects, we performed an ablation study focusing on induction heads, known contributors to ICL and temporal processing \citep{olsson2022context, elhage2021mathematical, ji2024linking}. As defined by \citet{olsson2022context}, induction heads are attention heads exhibiting pattern completion or copying behavior; they identify previous occurrences of the current token and attend to the subsequent token.

\begin{figure*}[h!]
\centering
\begingroup
\setlength{\FigGridLabelW}{0.75cm}
\setlength{\FigGridSubLabelW}{0.55cm}
\setlength{\FigGridCellH}{1.85cm} 
\setlength{\FigGridLabelSep}{1pt}
\setlength{\FigGridSubLabelSep}{1pt}
\setlength{\FigGridColSep}{0.5pt}

\FigGridSetupTwo{5}
\renewcommand{\arraystretch}{1.0}

\begin{tabular}{@{}>{\centering\arraybackslash}m{\FigGridLabelW}@{\hspace{\FigGridLabelSep}}
                >{\centering\arraybackslash}m{\FigGridSubLabelW}@{\hspace{\FigGridSubLabelSep}}
                *{5}{>{\centering\arraybackslash}m{\FigGridPlotW}}@{}}
\toprule
& & \multicolumn{5}{c}{\FigHdr Ablations} \\
\cmidrule(lr){3-7}
& & \FigHdr 0 & \FigHdr 1 & \FigHdr 10 & \FigHdr 50 & \FigHdr 100 \\
\midrule
\multirow{2}{*}{\FigGridVLabelSpanTwo{Llama}} &
\FigGridVSubLabel{Ind} &
\FigGridCell{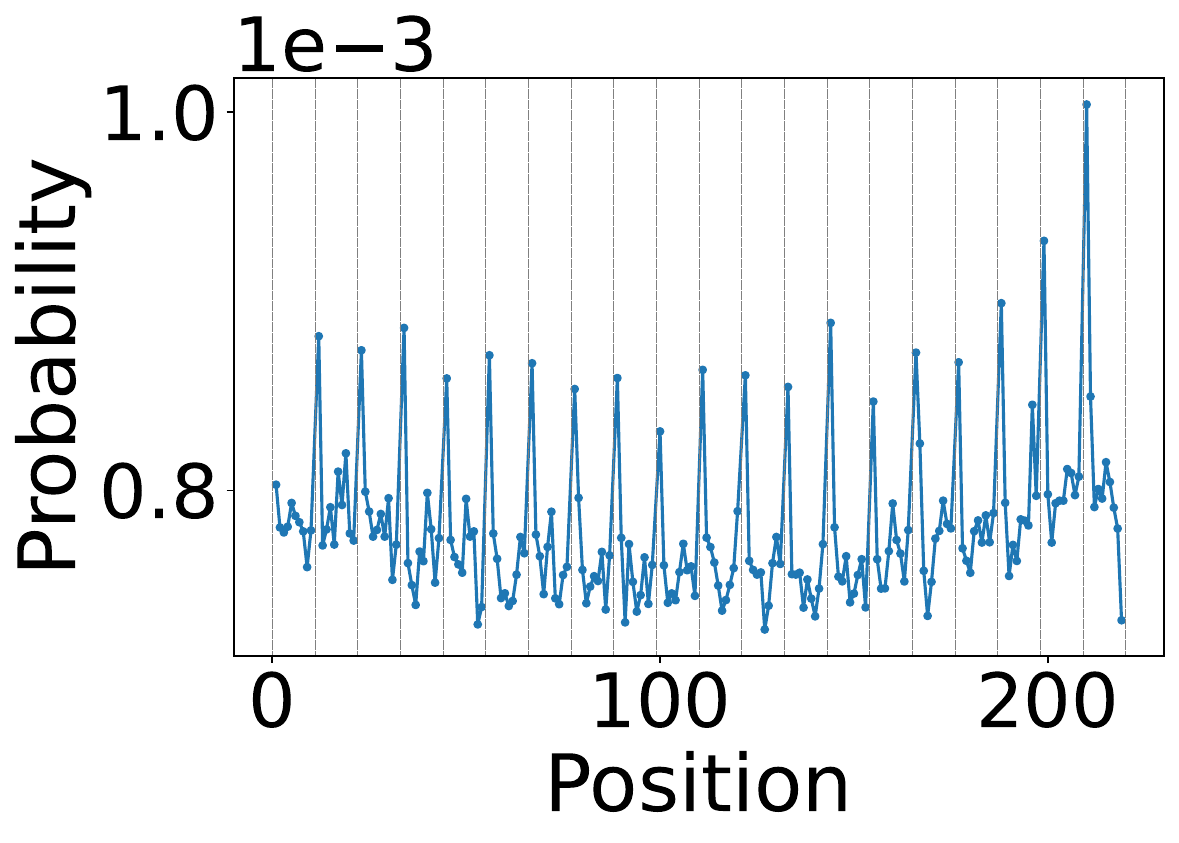} &
\FigGridCell{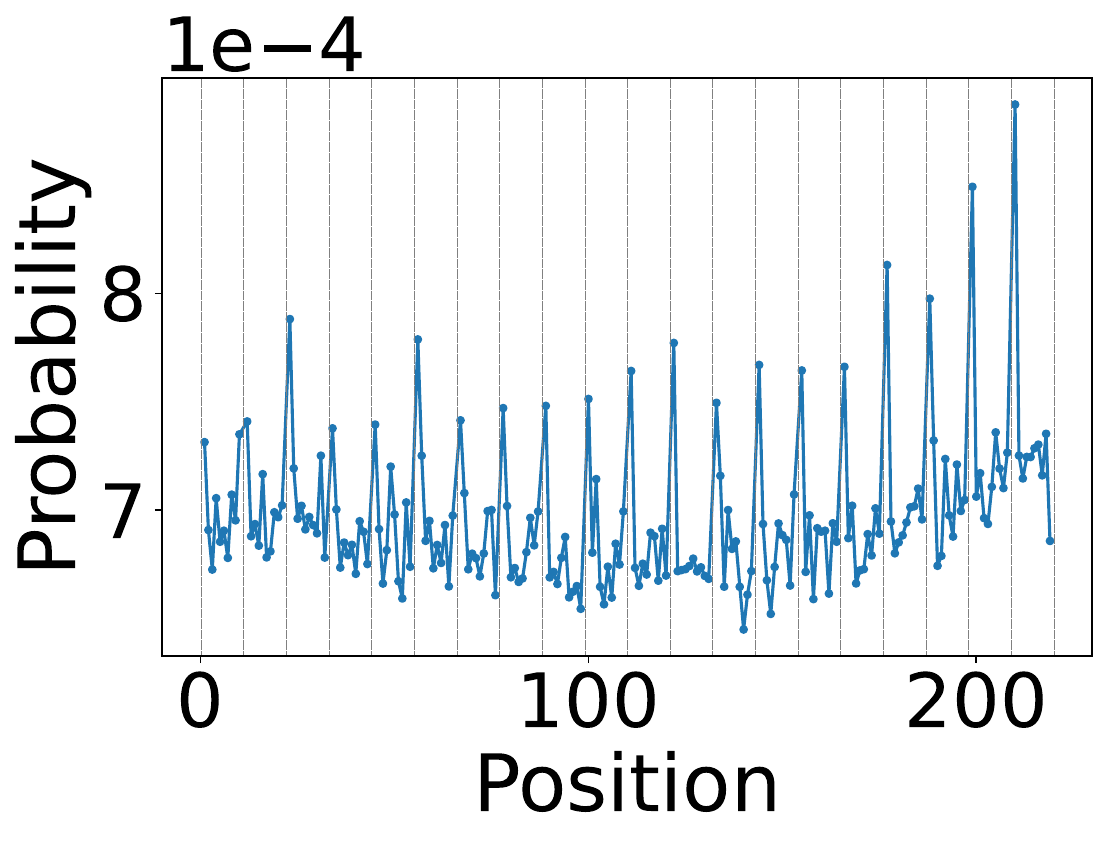} &
\FigGridCell{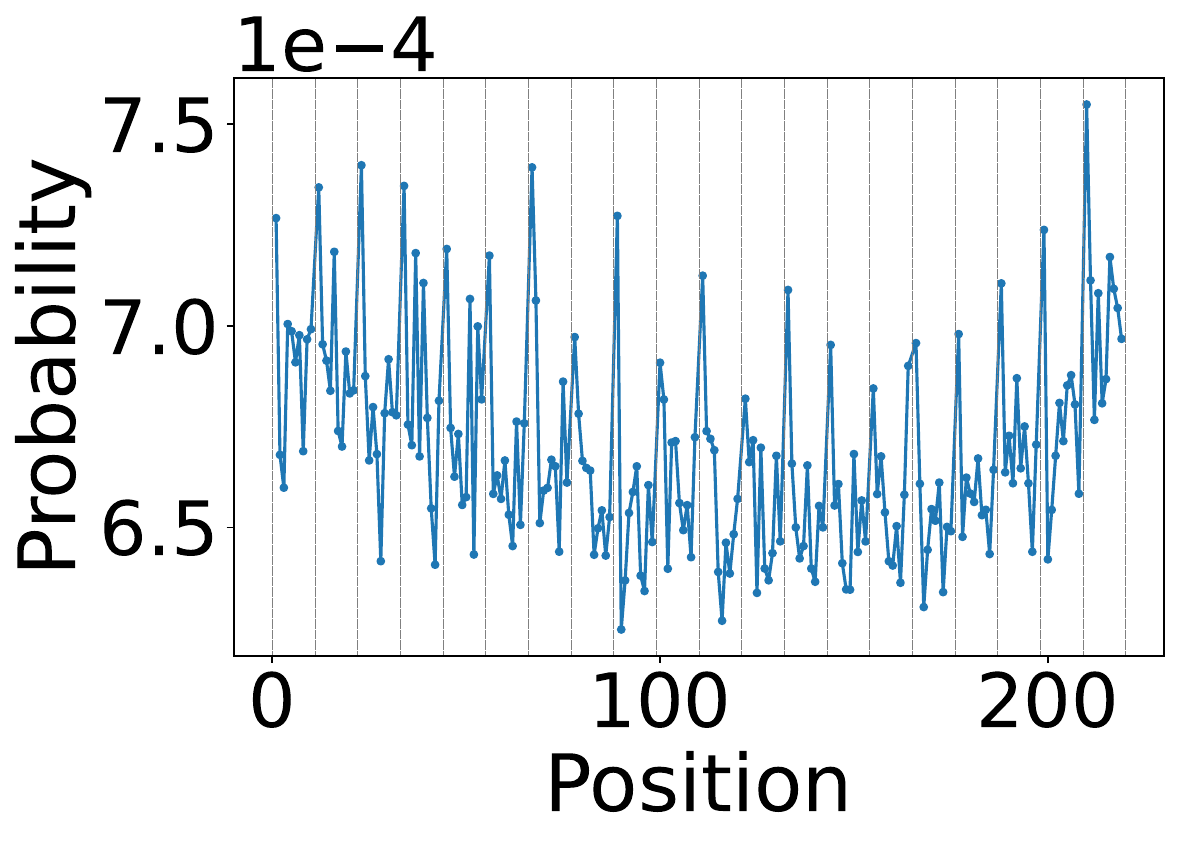} &
\FigGridCell{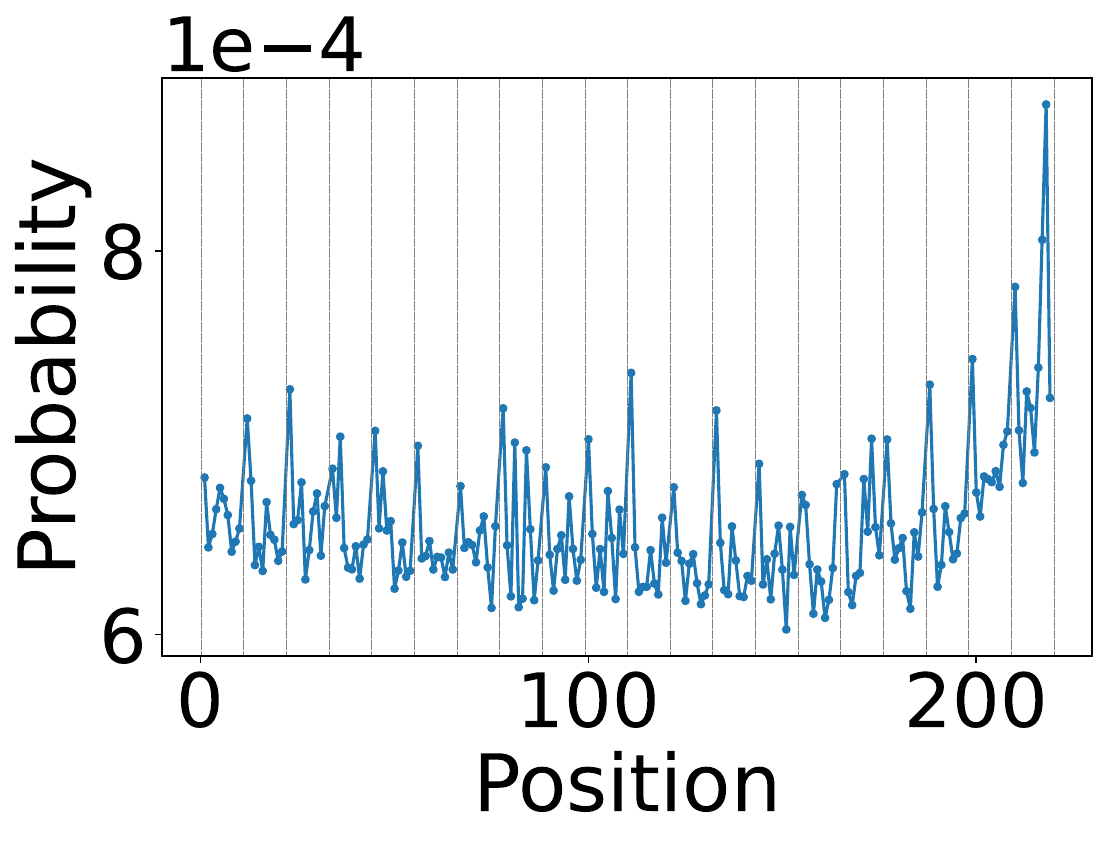} &
\FigGridCell{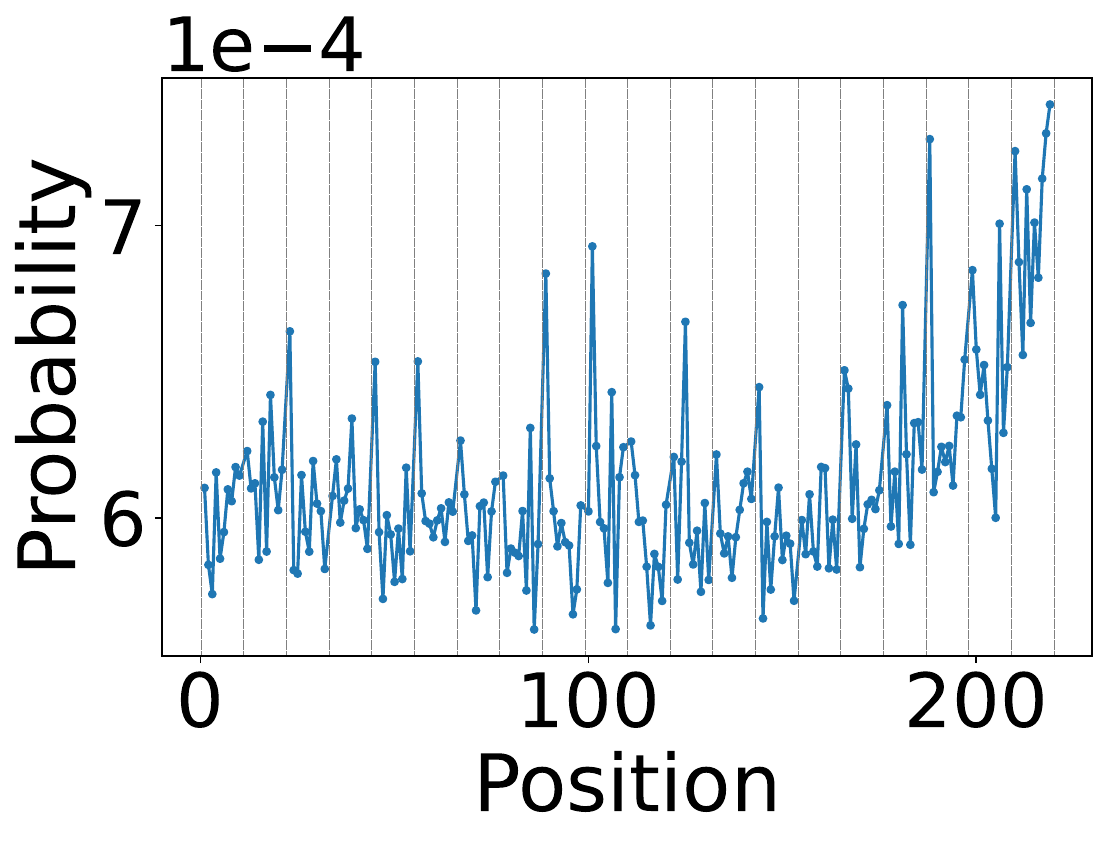} \\
& \FigGridVSubLabel{Rand} &
\FigGridCell{Figures/abl_1_prob_without_A/Llama-3.1-8B-Instruct_20_Repeats_10_Length_0_ablations_induction_5000_Permutations.pdf} &
\FigGridCell{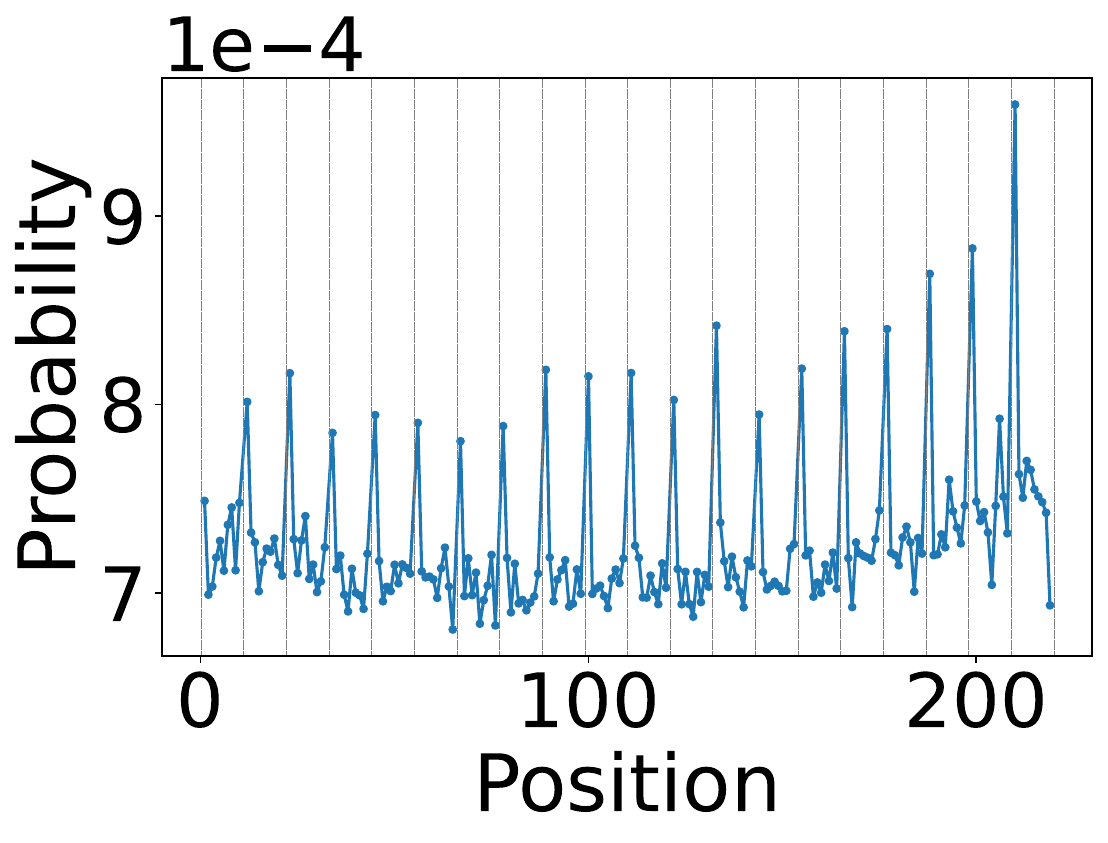} &
\FigGridCell{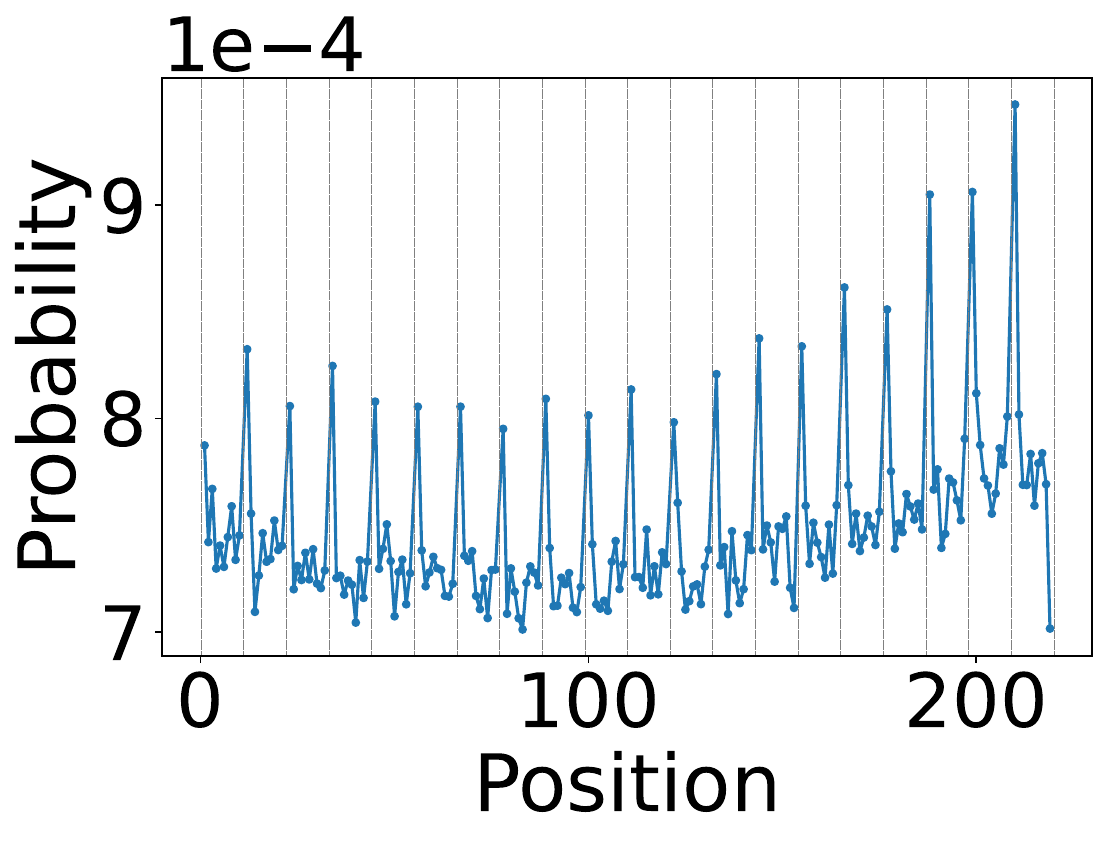} &
\FigGridCell{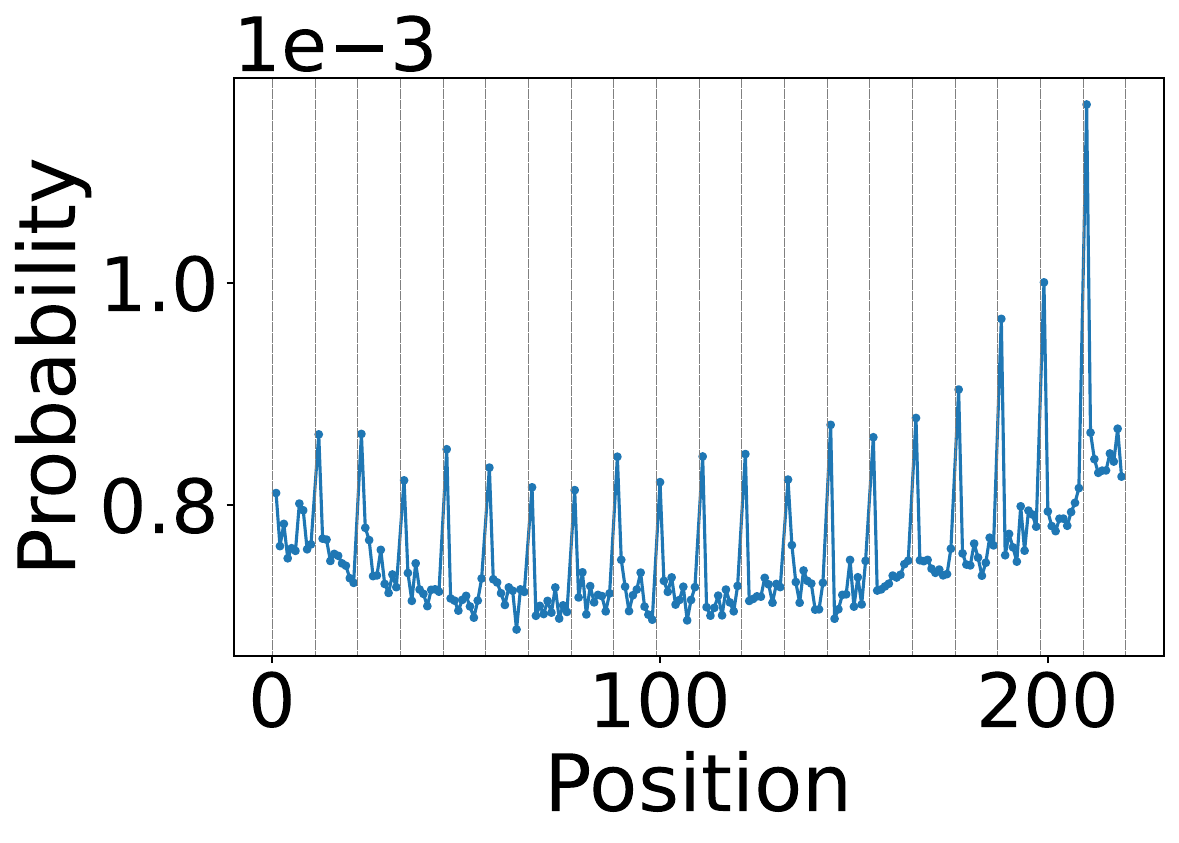} &
\FigGridCell{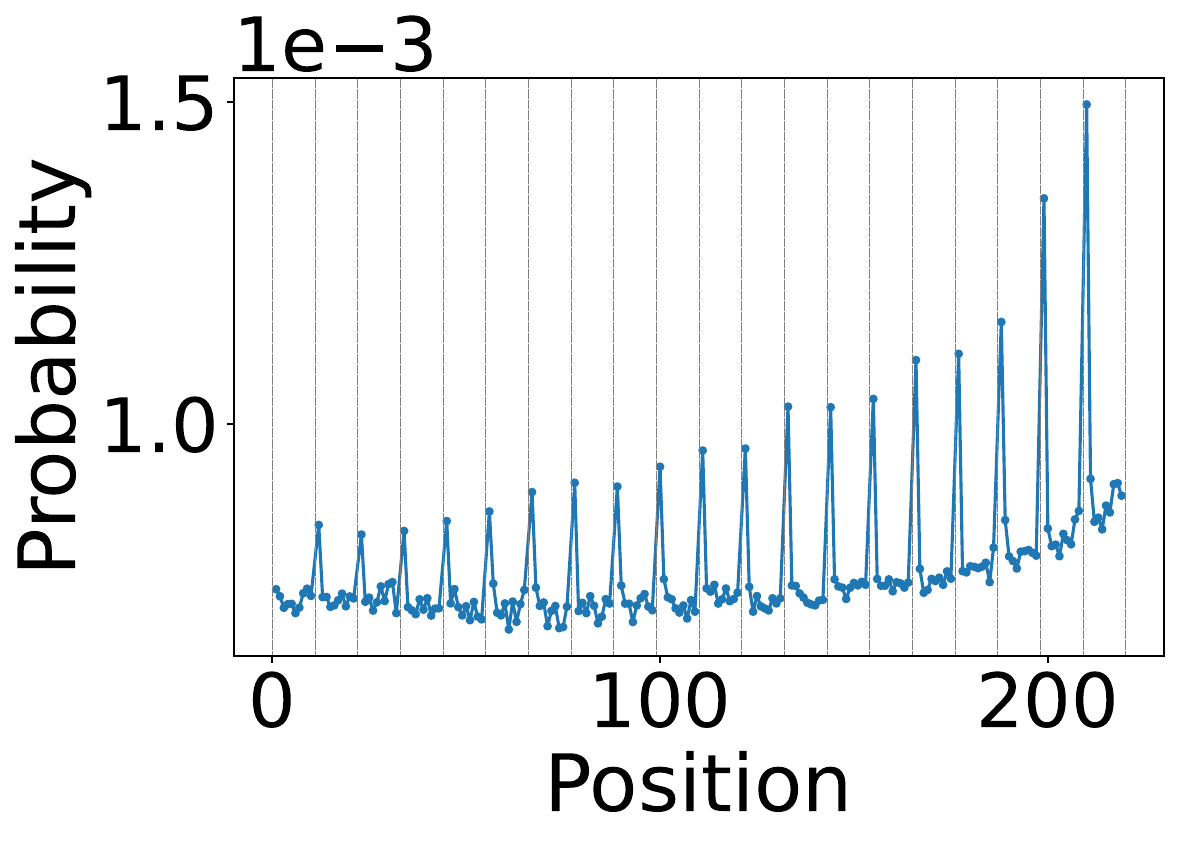} \\

\multirow{2}{*}{\FigGridVLabelSpanTwo{Mistral}} &
\FigGridVSubLabel{Ind} &
\FigGridCell{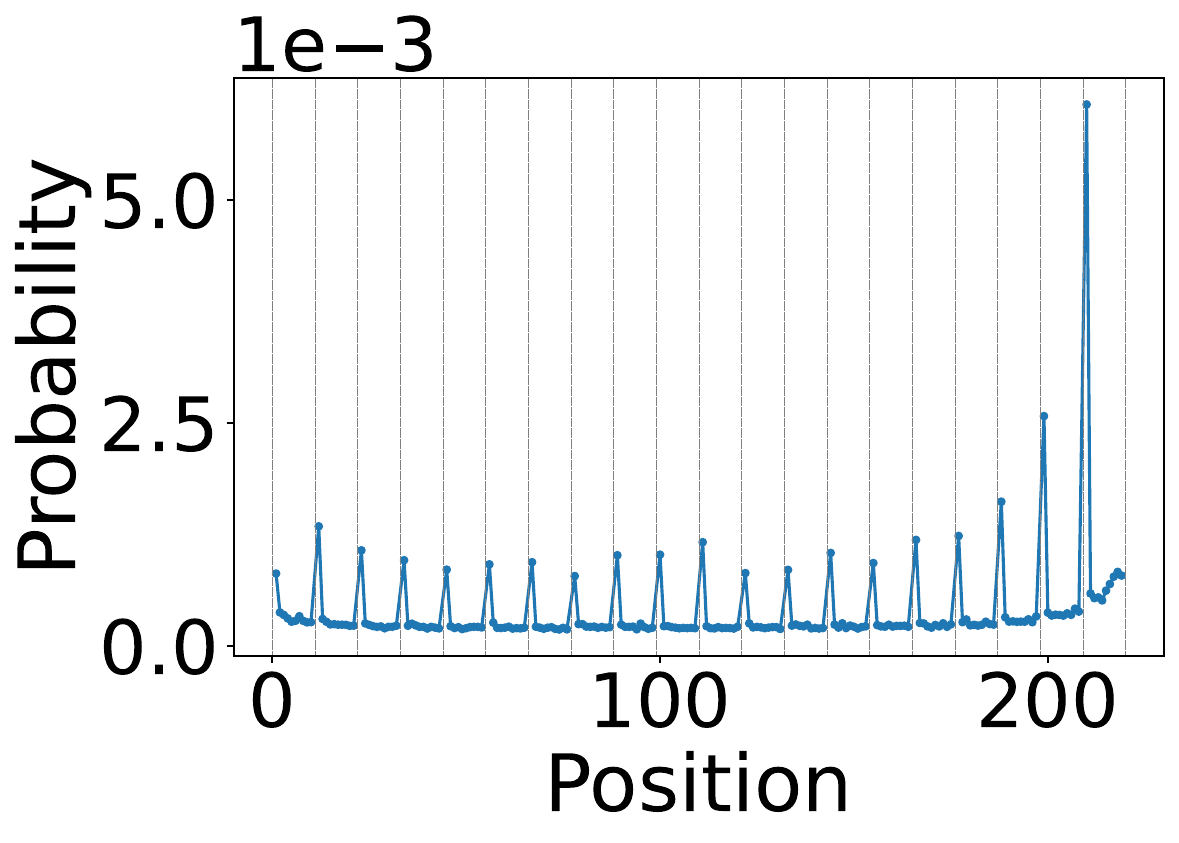} &
\FigGridCell{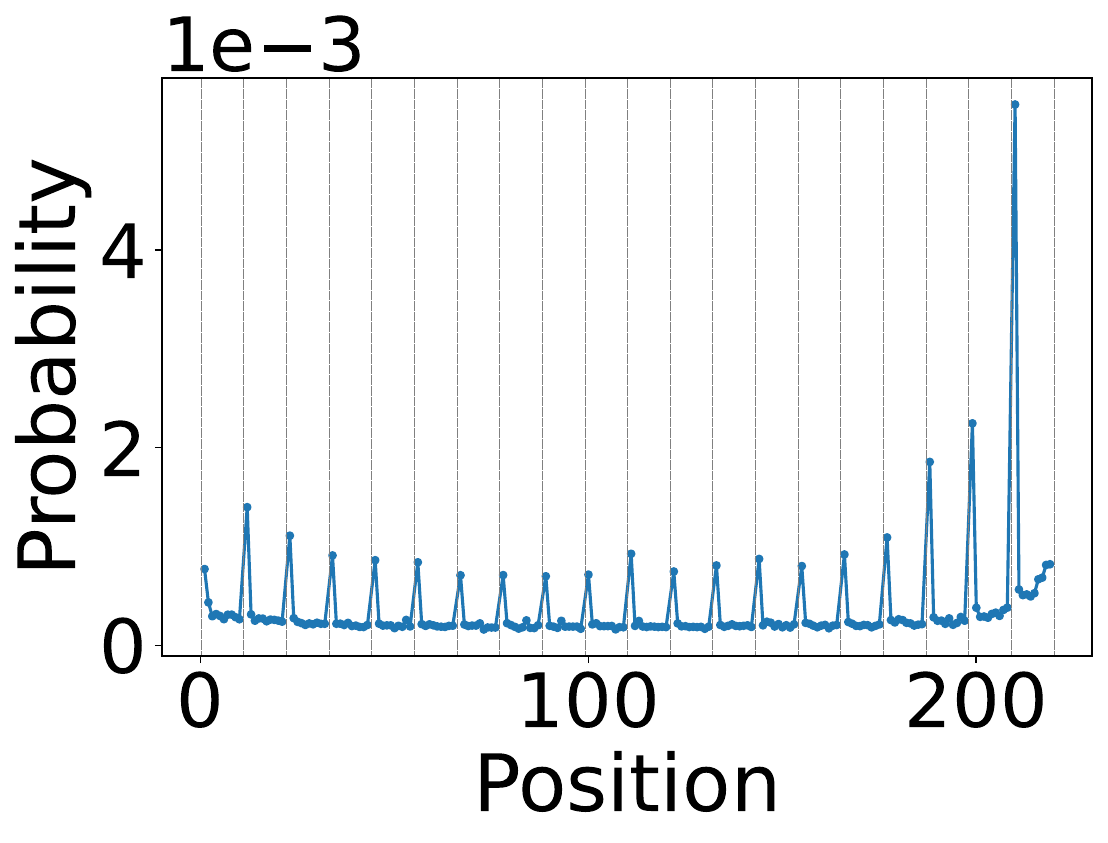} &
\FigGridCell{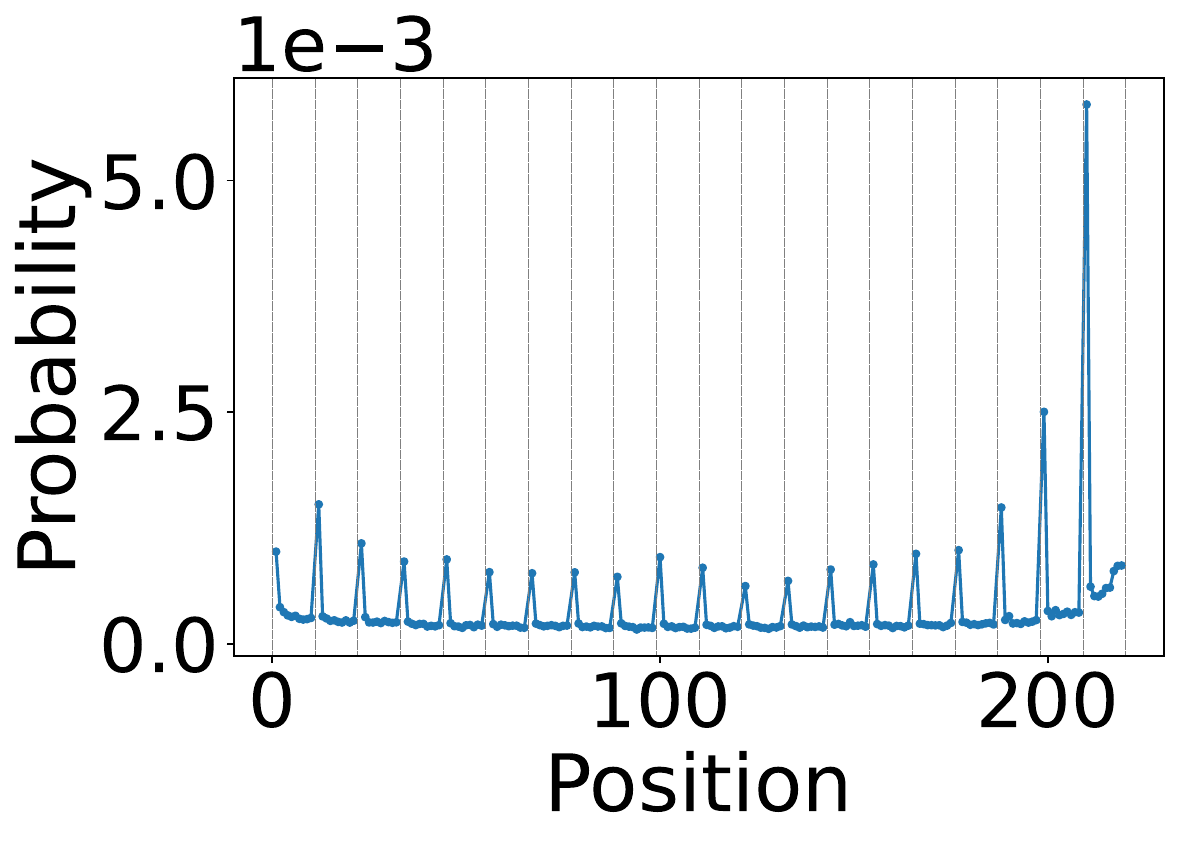} &
\FigGridCell{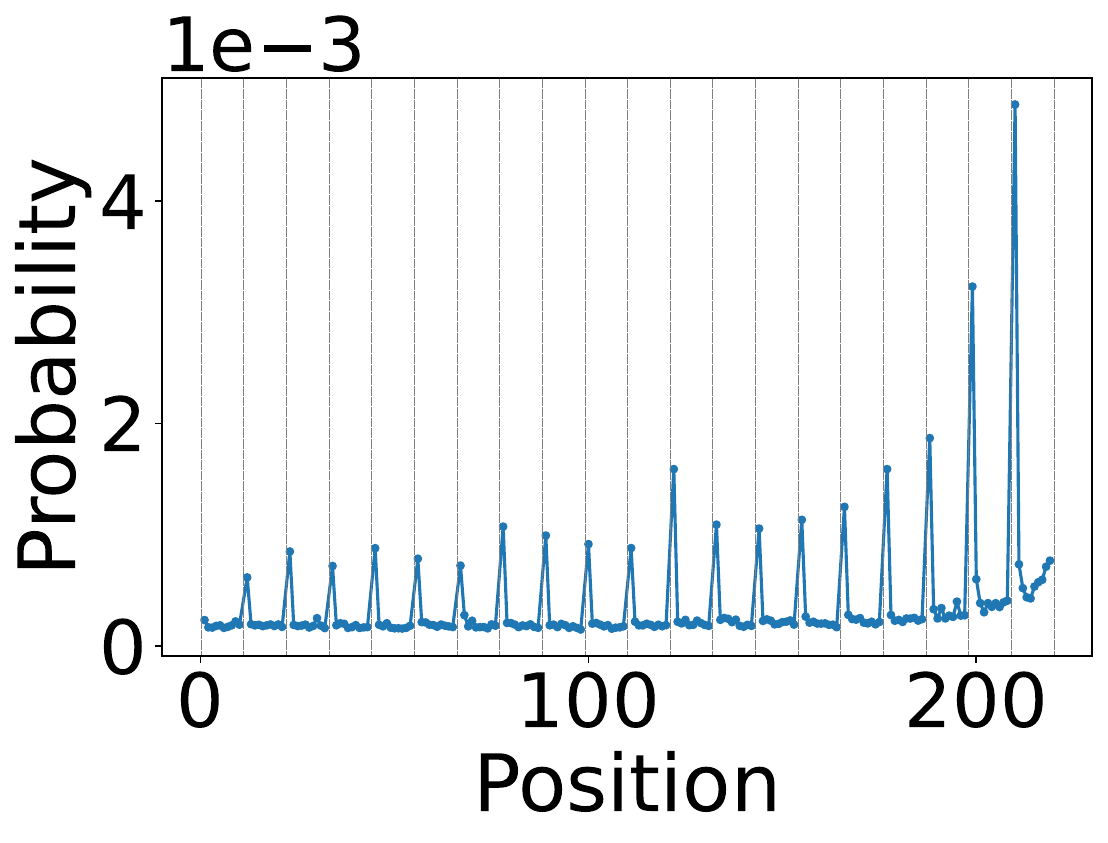} &
\FigGridCell{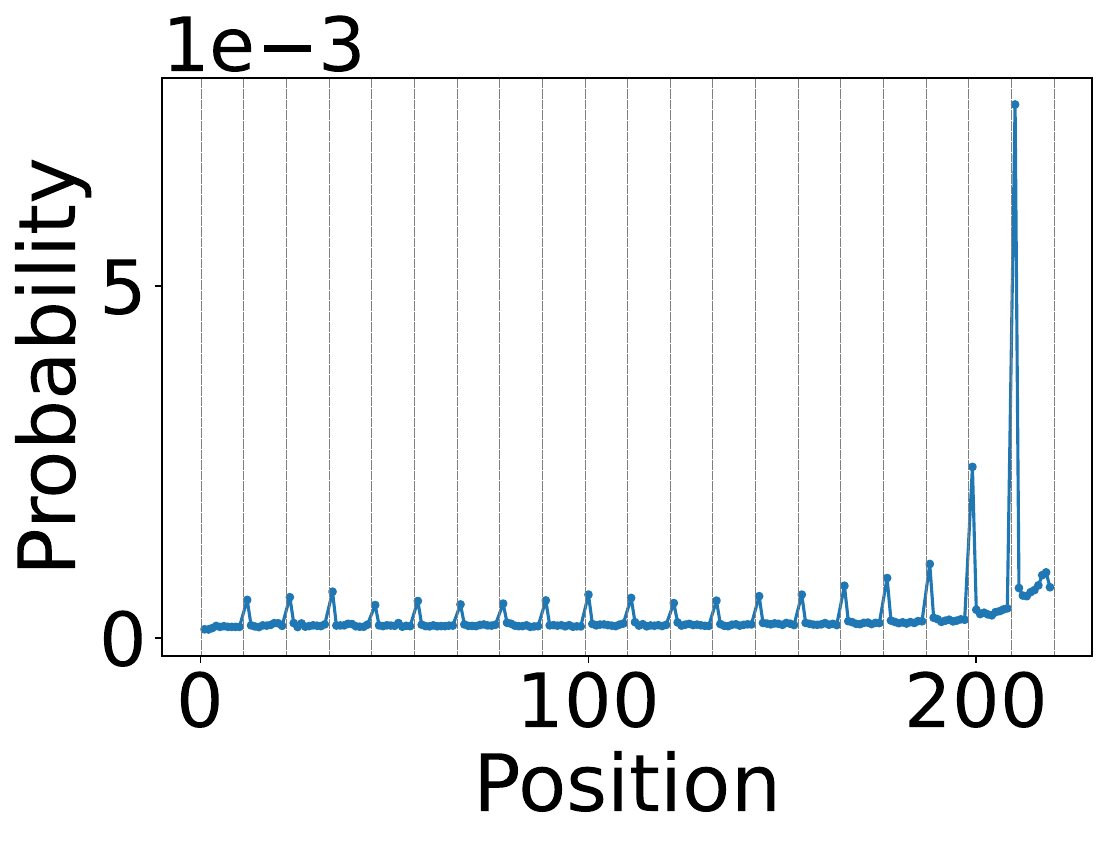} \\
& \FigGridVSubLabel{Rand} &
\FigGridCell{Figures/abl_1_prob_without_A/Mistral-7B-Instruct-v0.1_20_Repeats_10_Length_0_ablations_induction_5000_Permutations.pdf} &
\FigGridCell{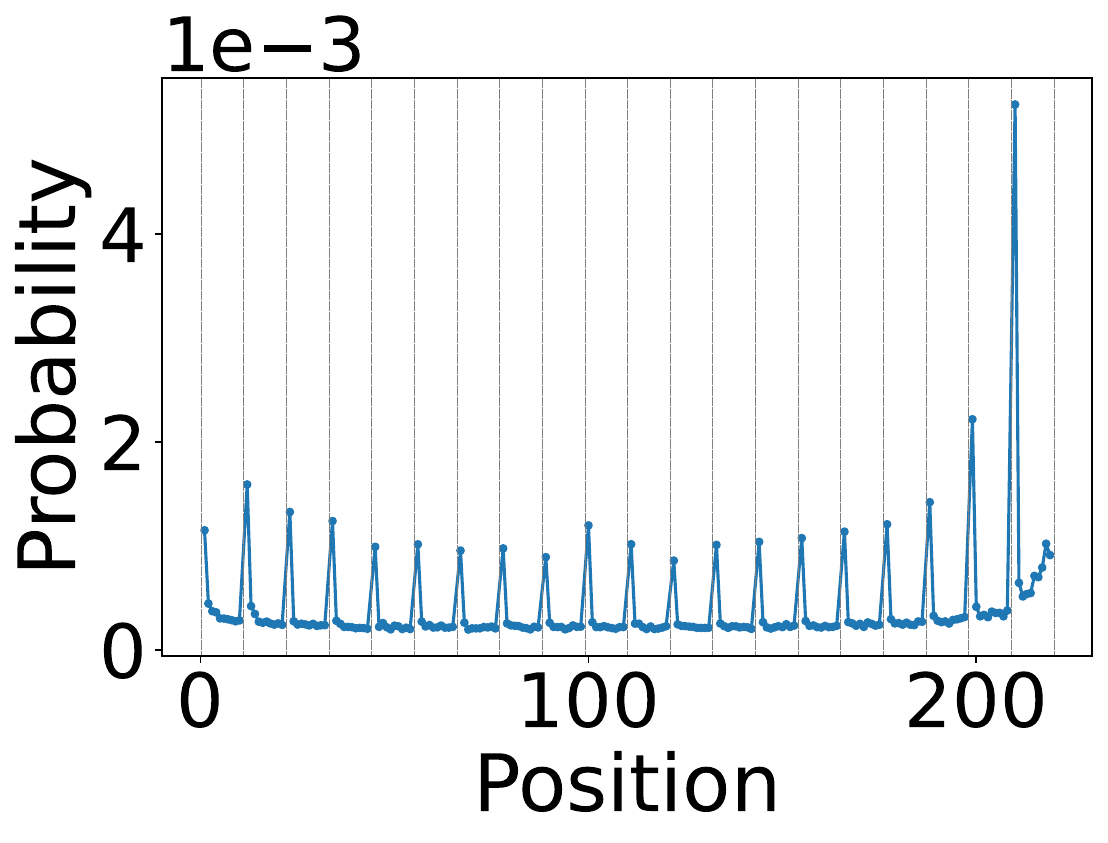} &
\FigGridCell{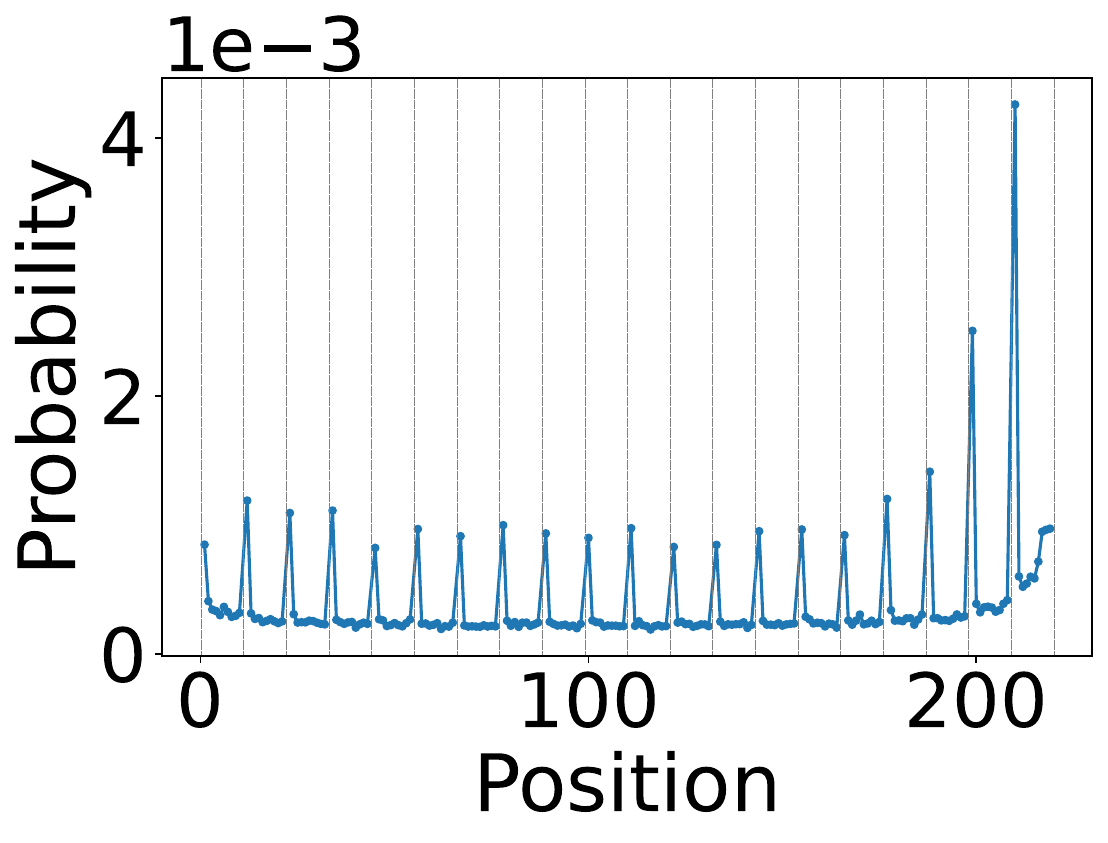} &
\FigGridCell{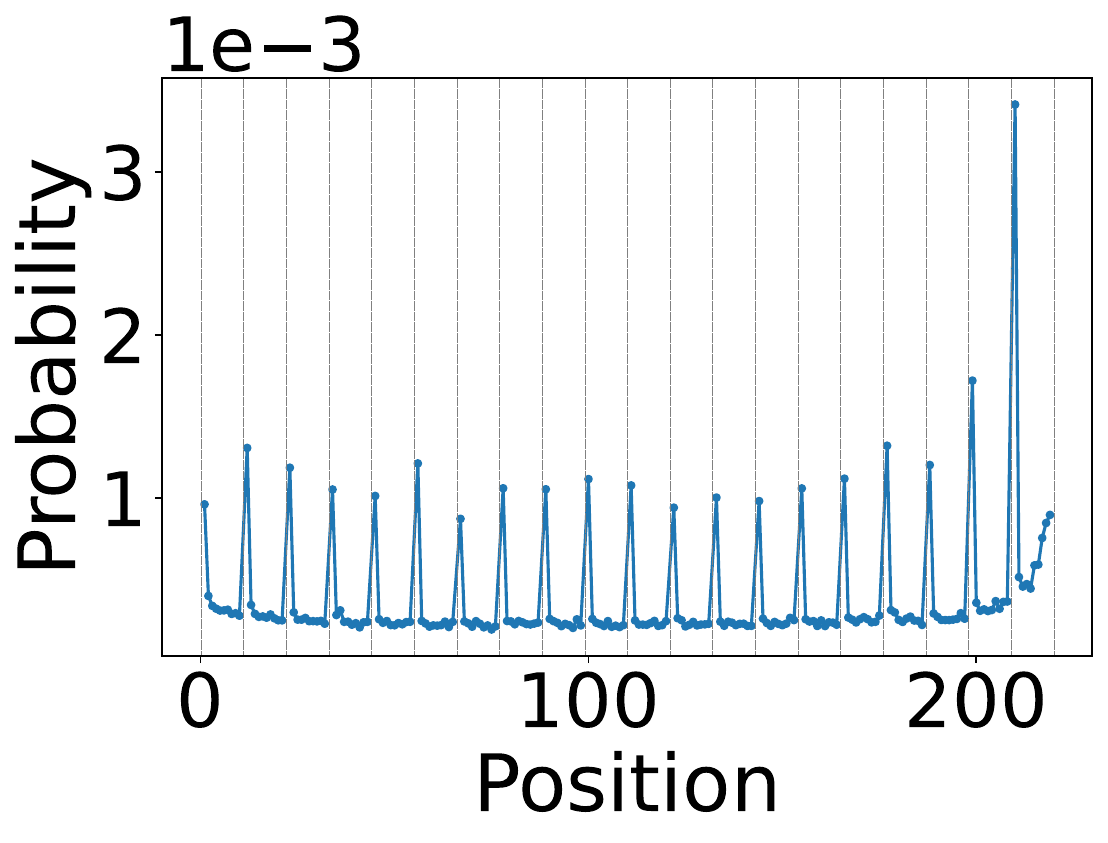} &
\FigGridCell{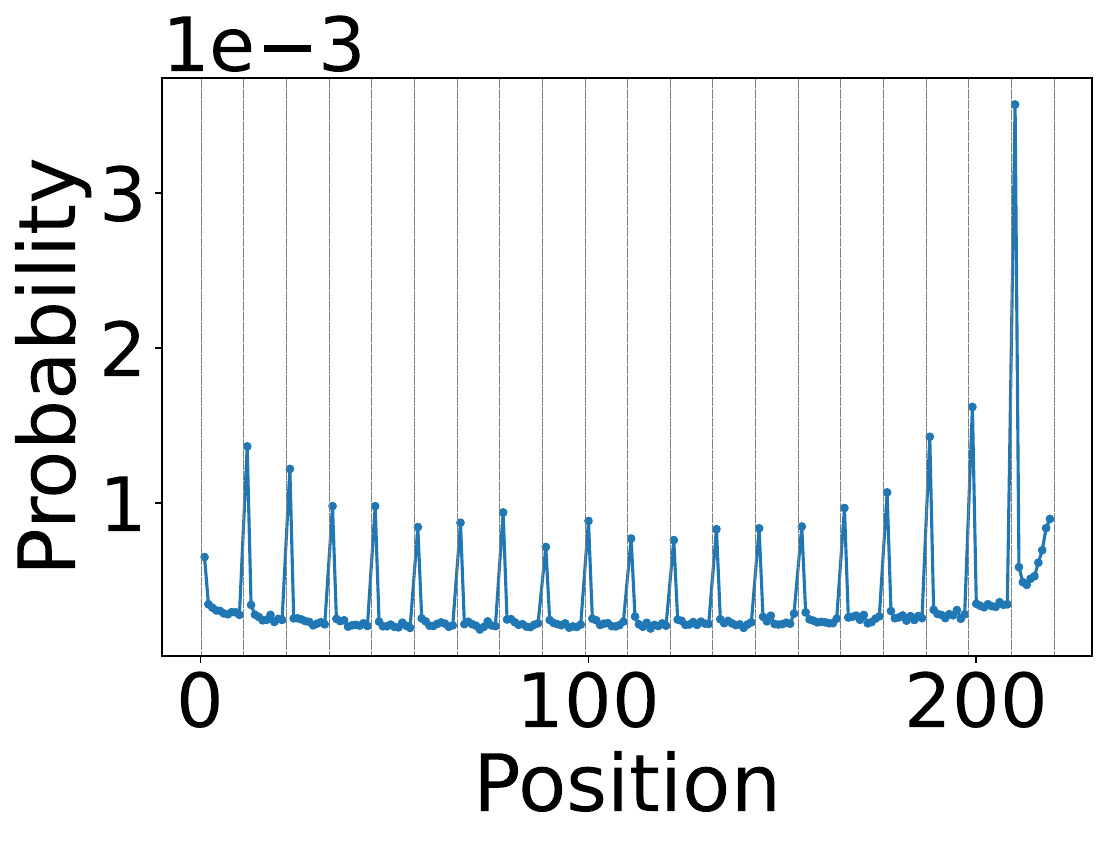} \\

\multirow{2}{*}{\FigGridVLabelSpanTwo{Qwen}} &
\FigGridVSubLabel{Ind} &
\FigGridCell{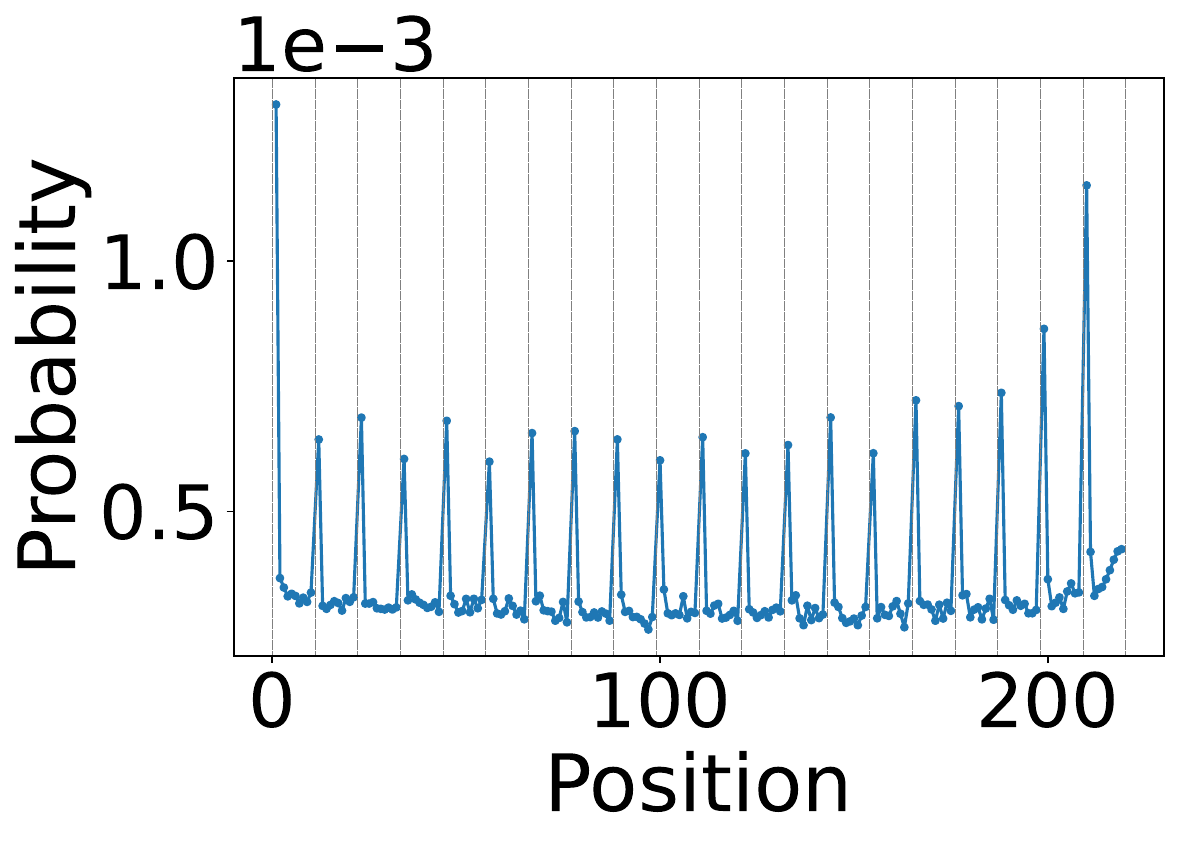} &
\FigGridCell{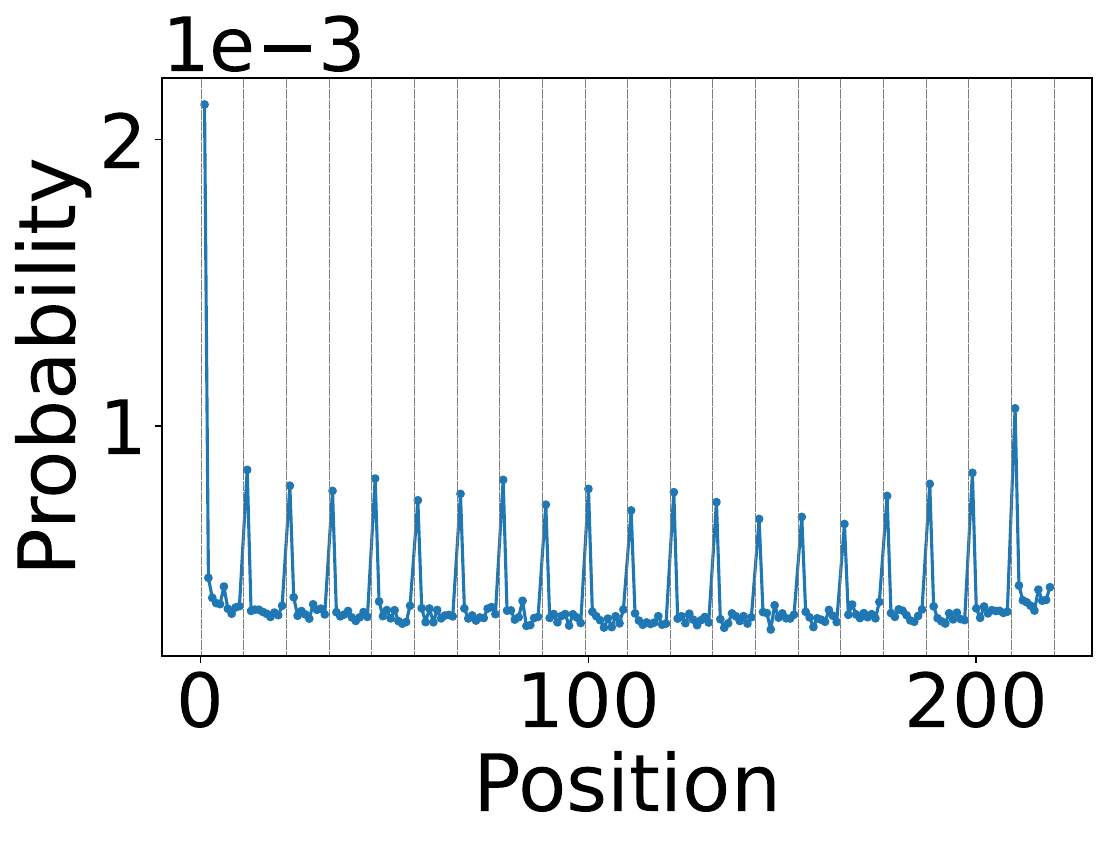} &
\FigGridCell{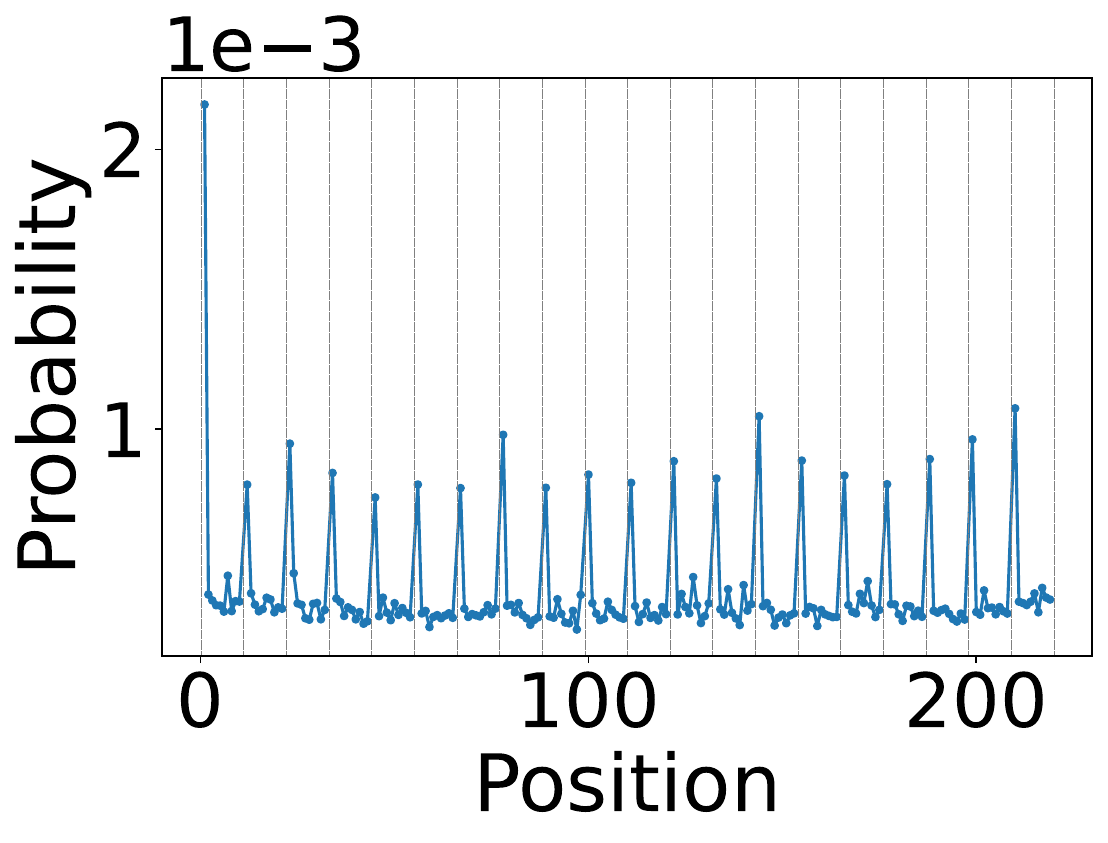} &
\FigGridCell{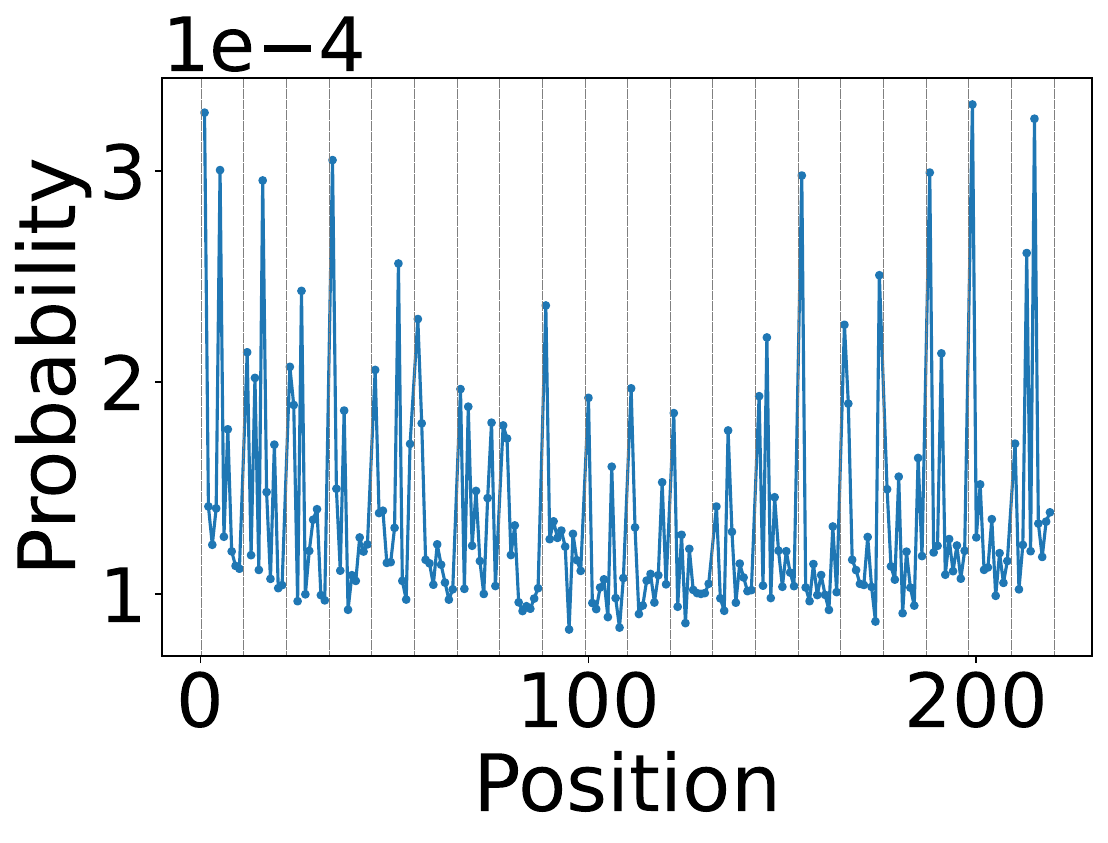} &
\FigGridCell{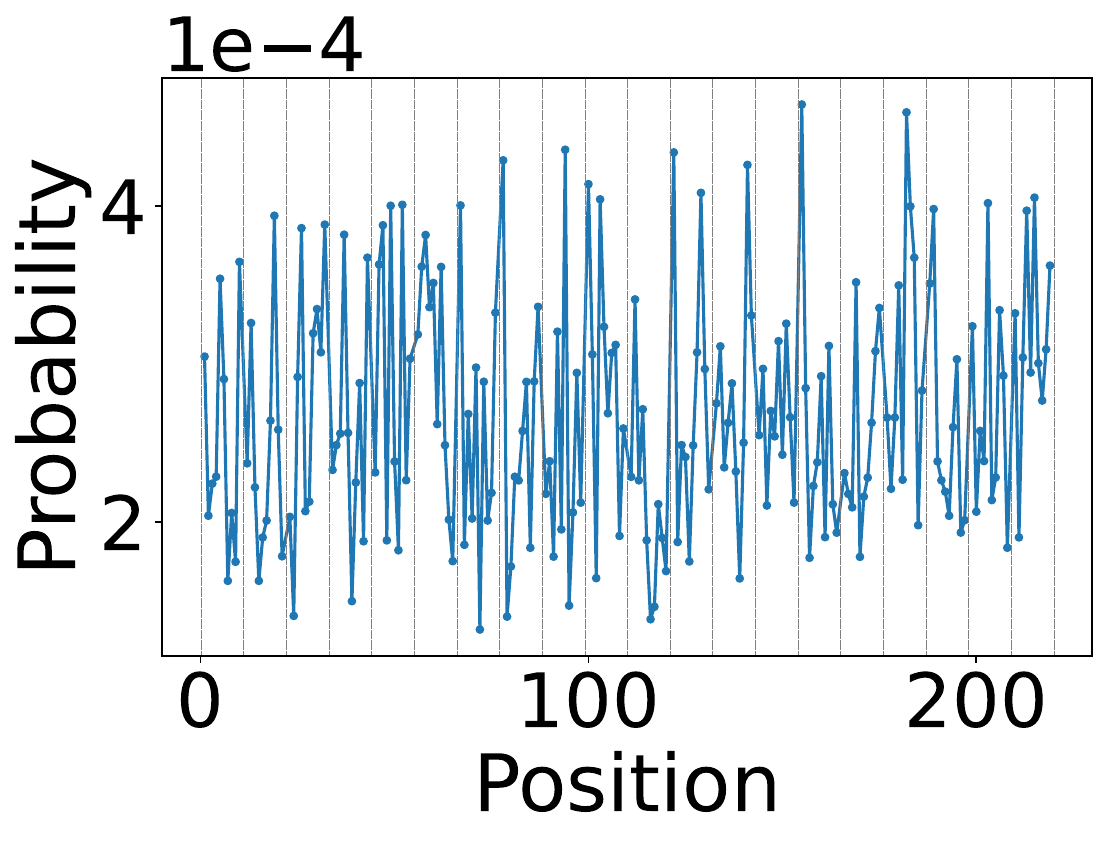} \\
& \FigGridVSubLabel{Rand} &
\FigGridCell{Figures/abl_1_prob_without_A/Qwen2.5-7B-Instruct_20_Repeats_10_Length_0_ablations_induction_5000_Permutations.pdf} &
\FigGridCell{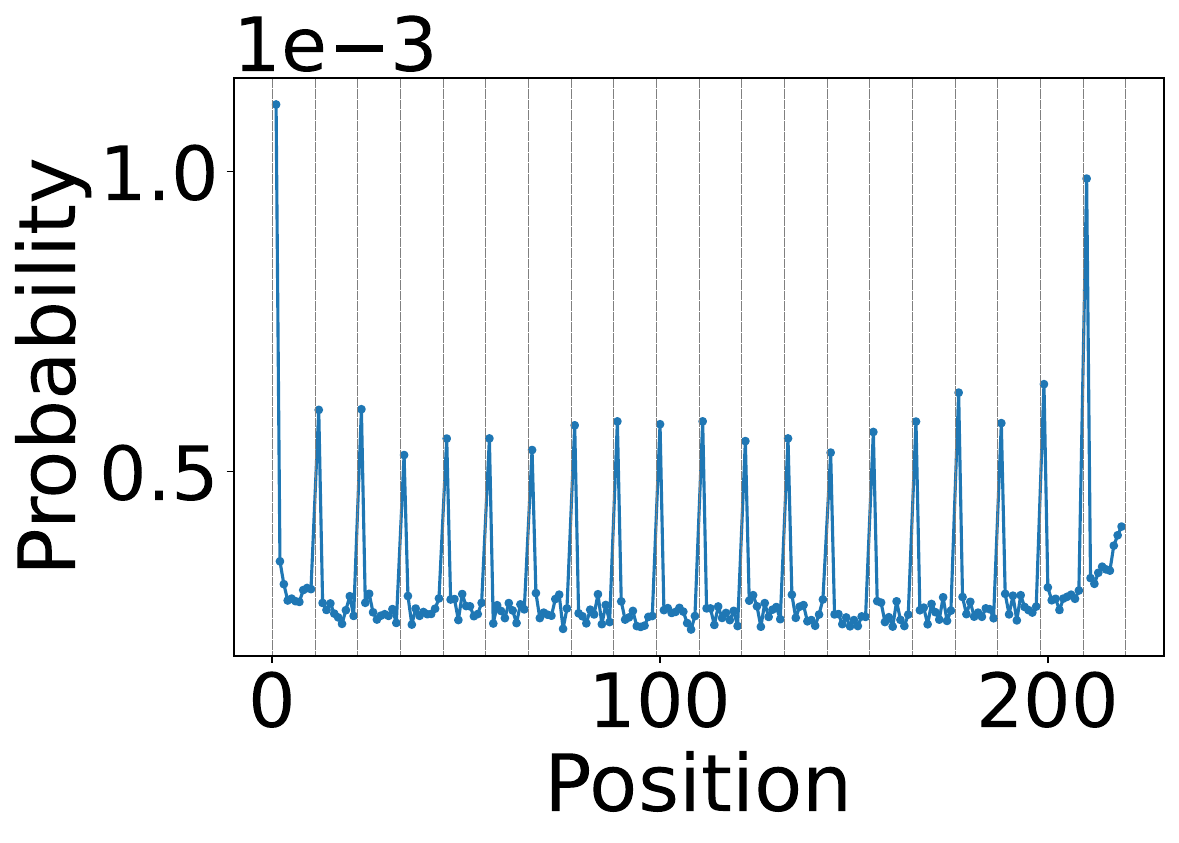} &
\FigGridCell{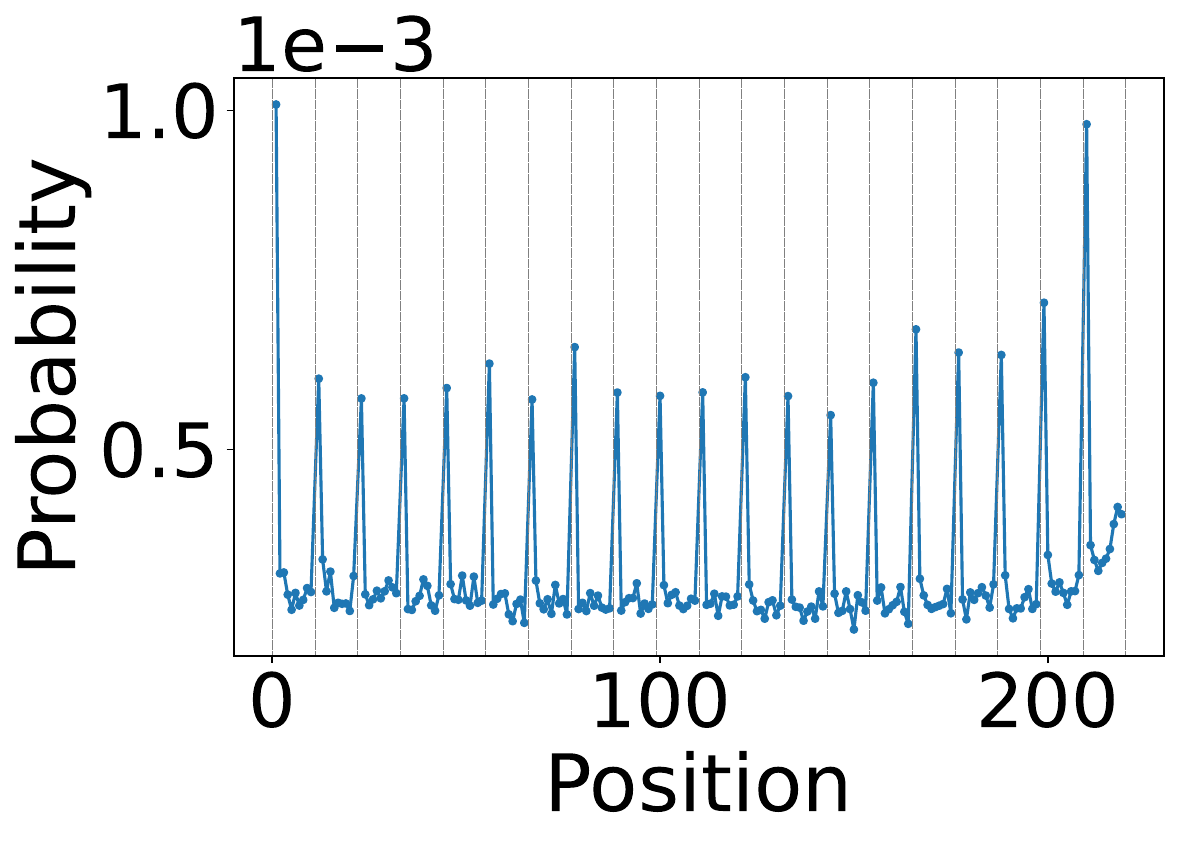} &
\FigGridCell{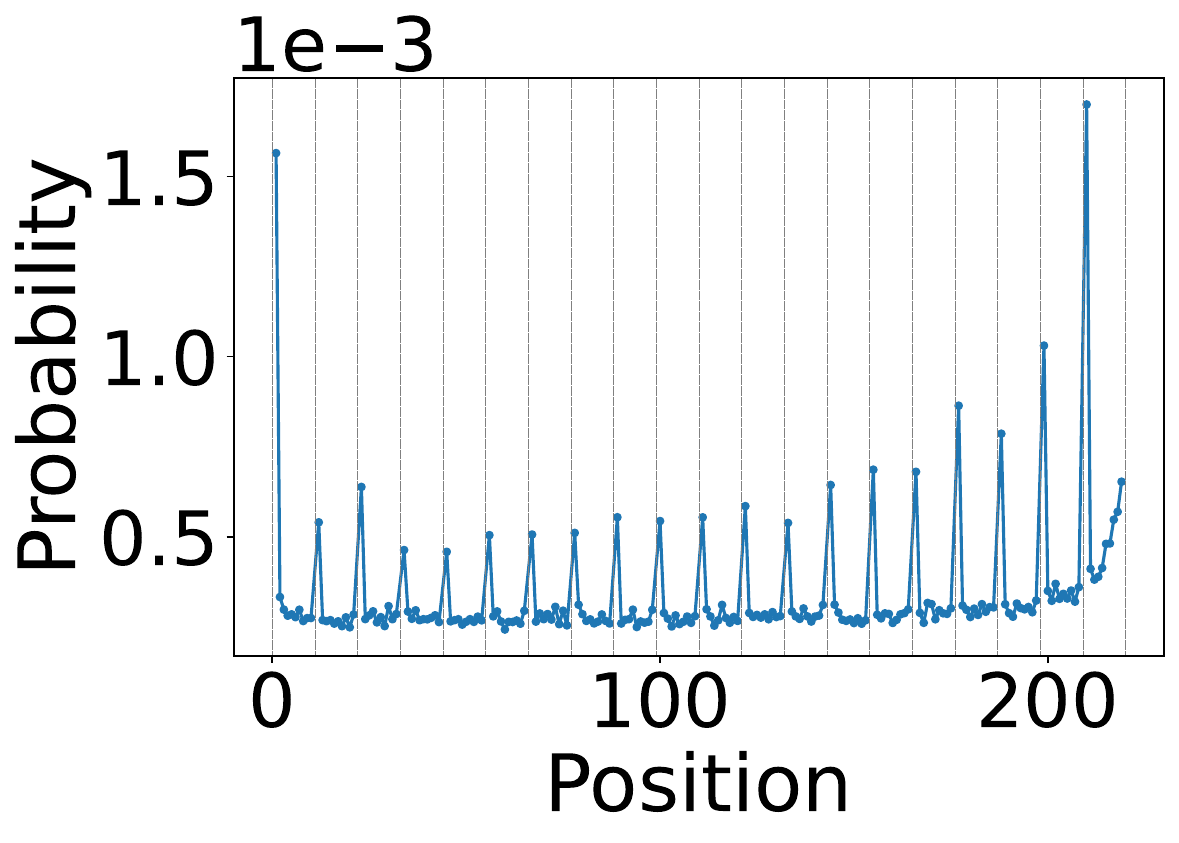} &
\FigGridCell{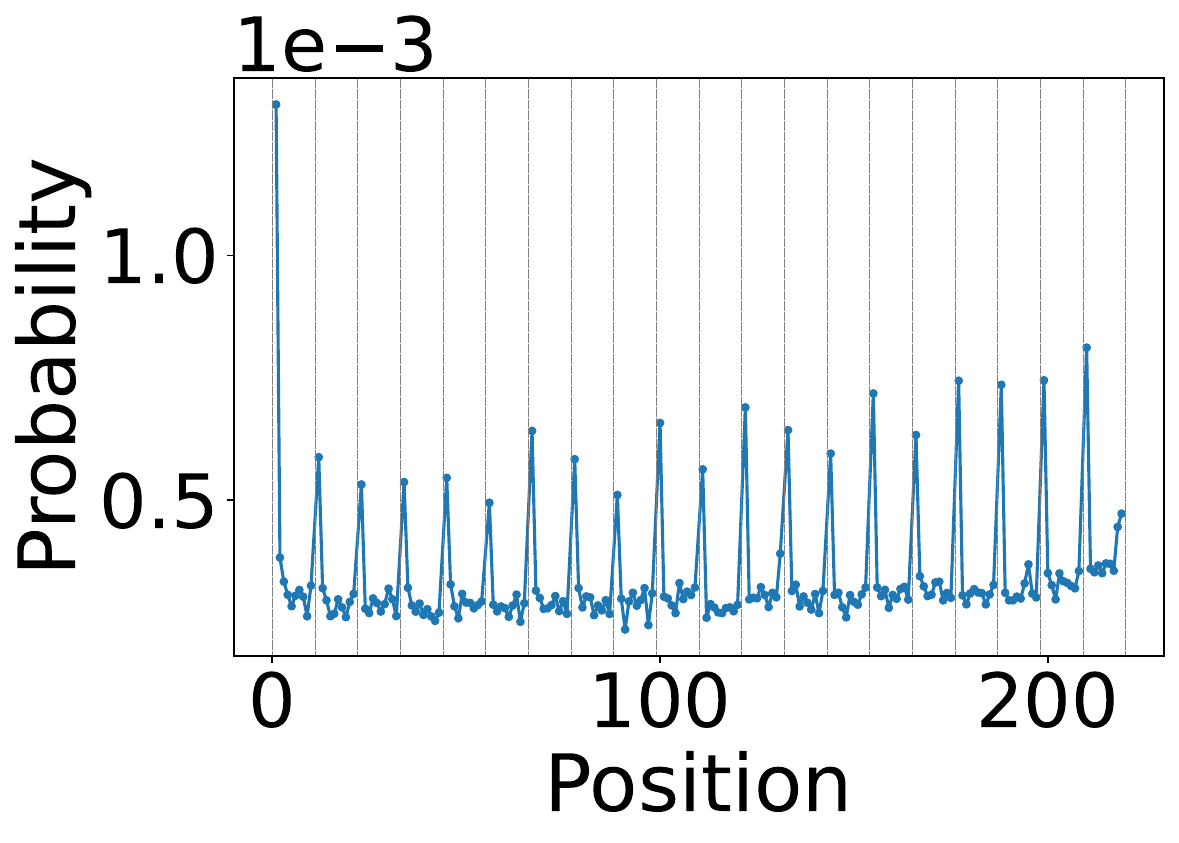} \\

\multirow{2}{*}{\FigGridVLabelSpanTwo{Gemma}} &
\FigGridVSubLabel{Ind} &
\FigGridCell{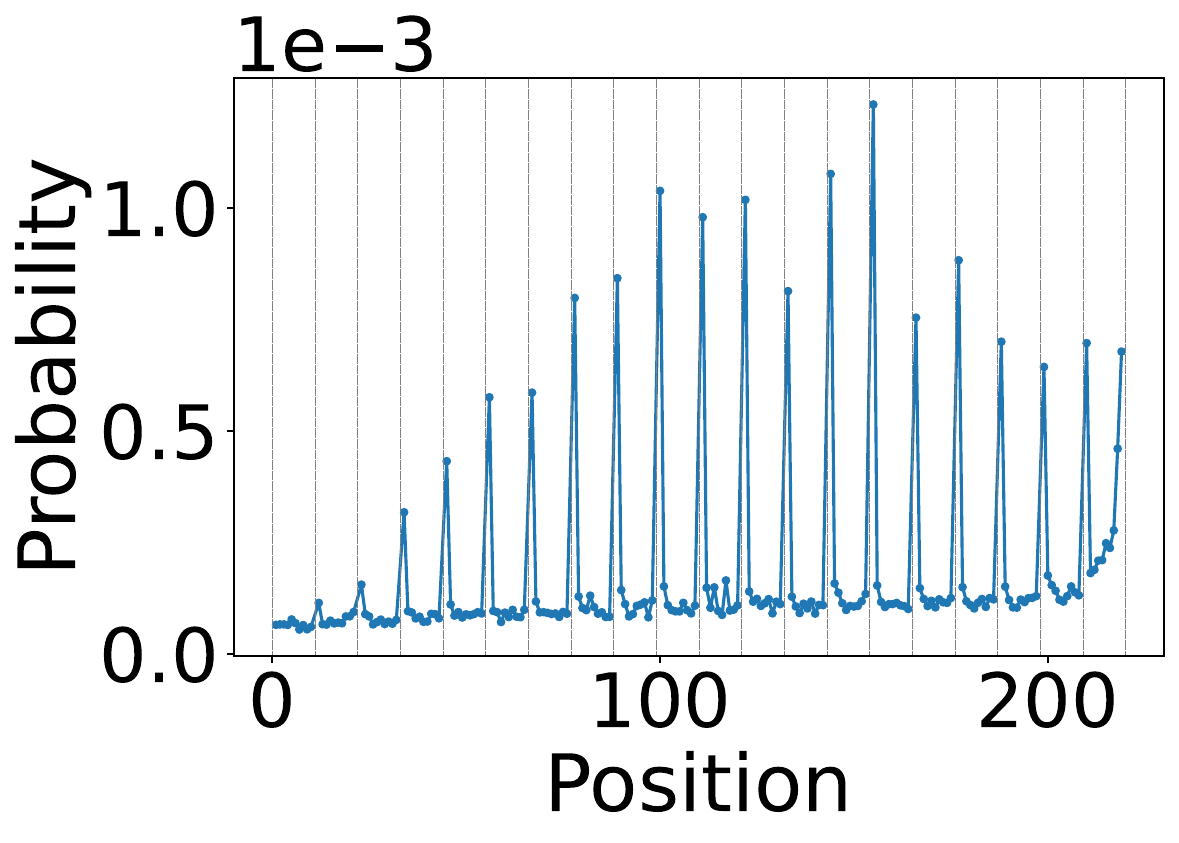} &
\FigGridCell{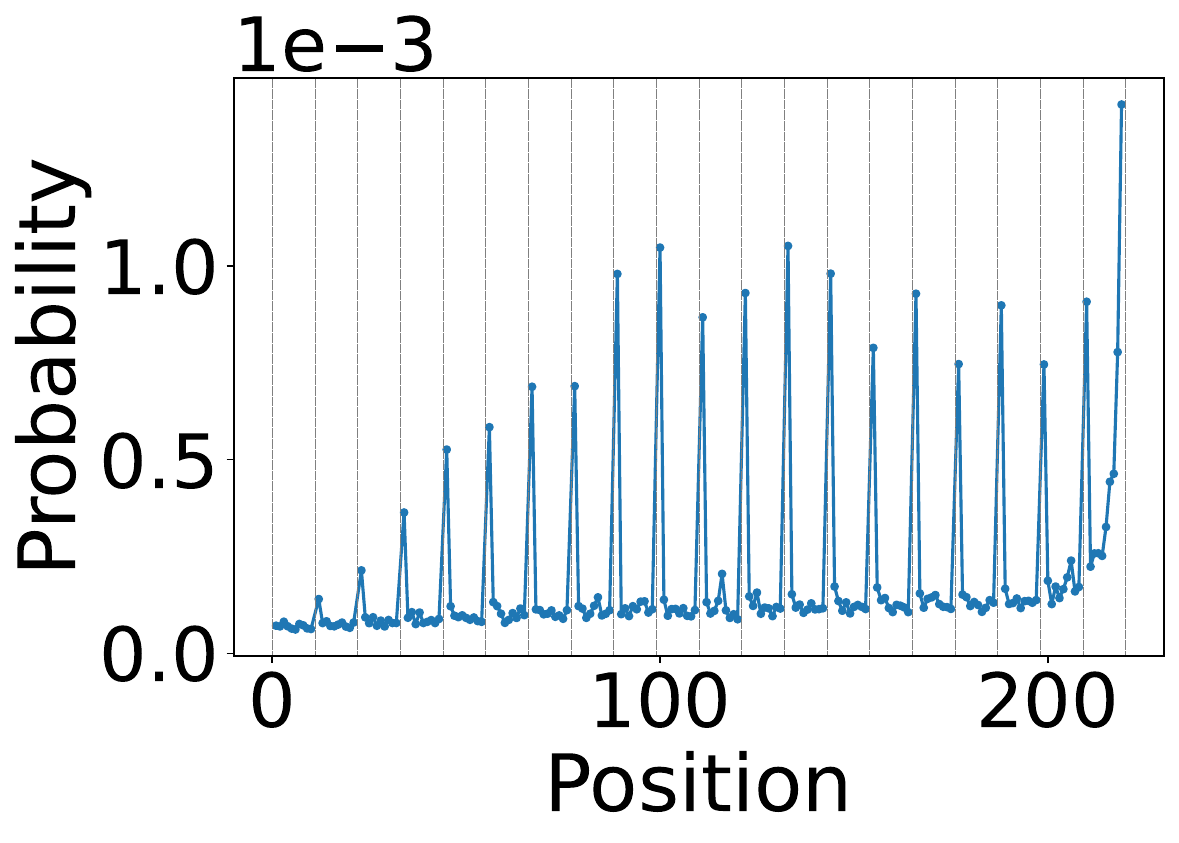} &
\FigGridCell{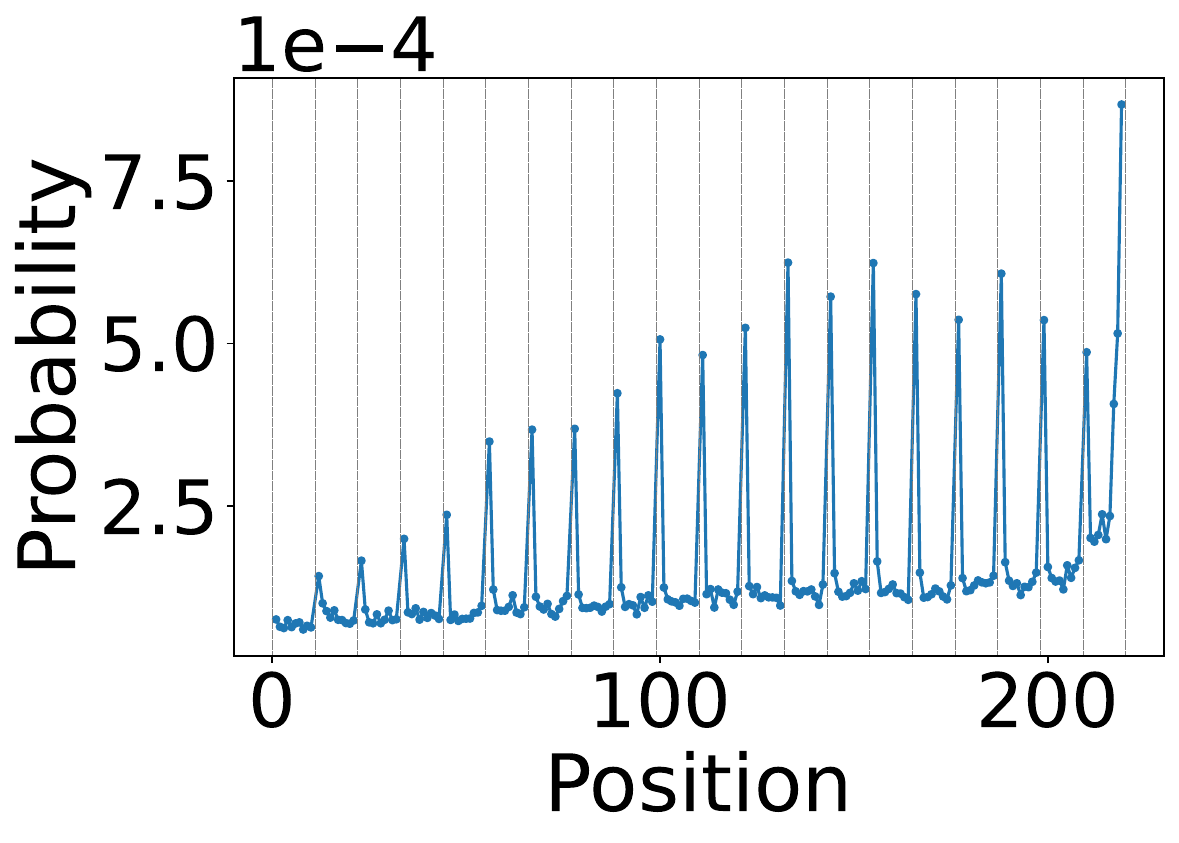} &
\FigGridCell{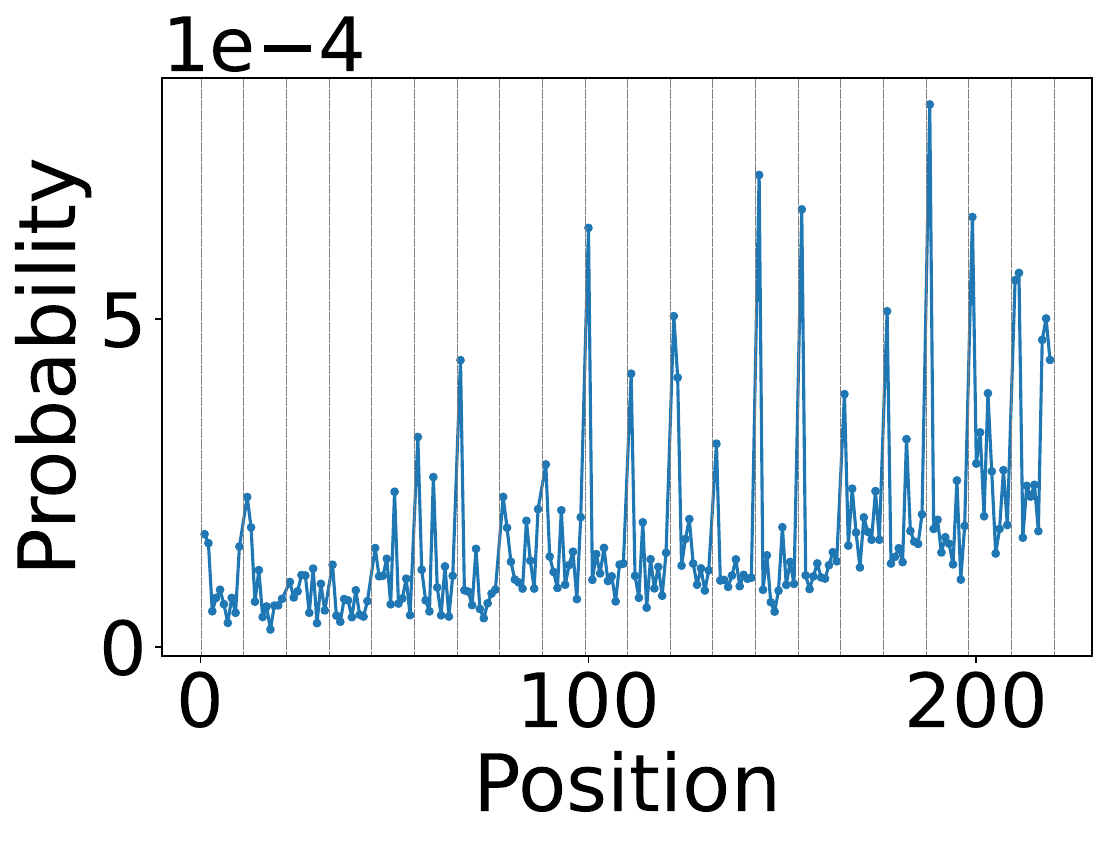} &
\FigGridCell{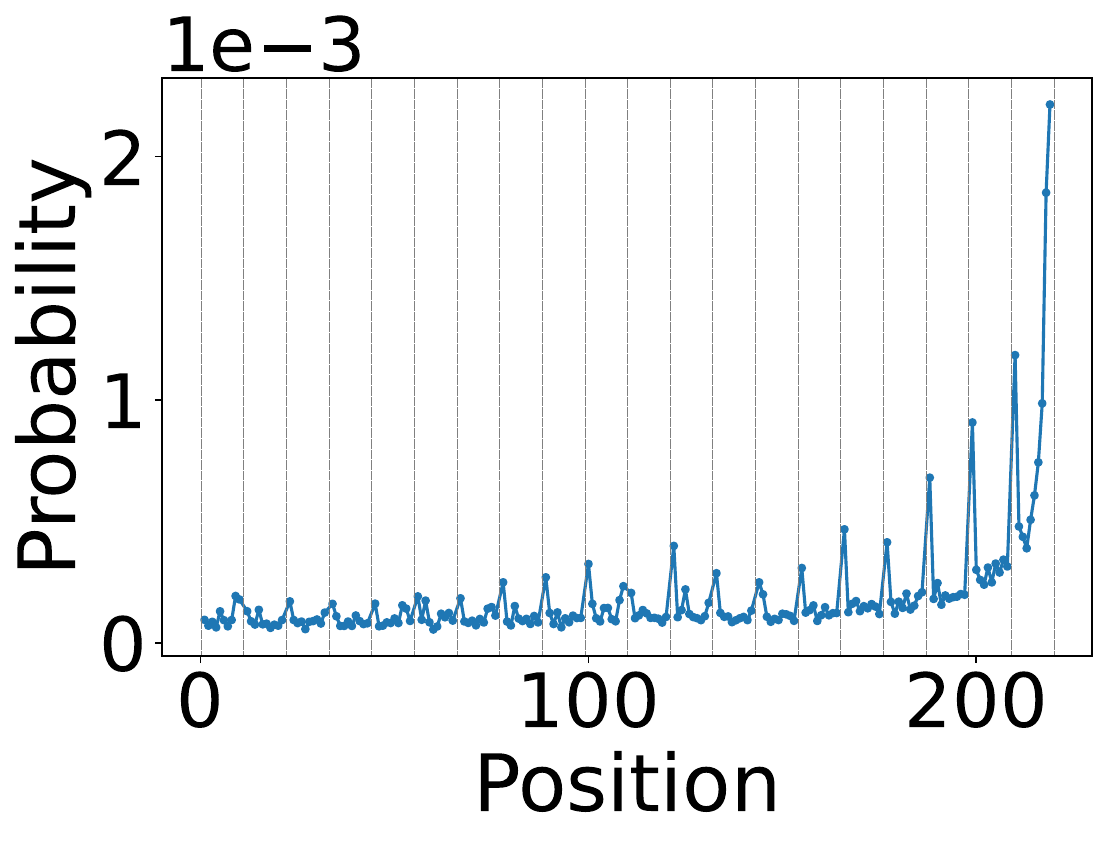} \\
& \FigGridVSubLabel{Rand} &
\FigGridCell{Figures/abl_1_prob_without_A/gemma-2-9b-it_20_Repeats_10_Length_0_ablations_induction_5000_Permutations.pdf} &
\FigGridCell{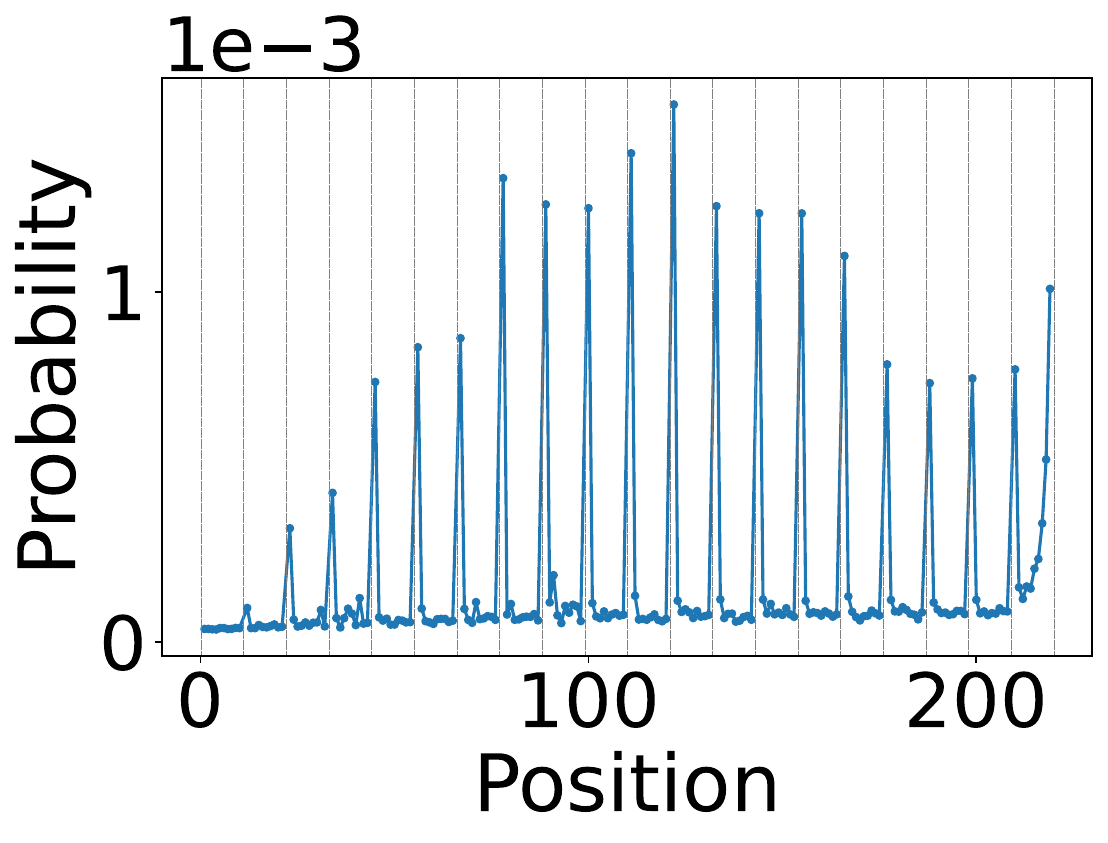} &
\FigGridCell{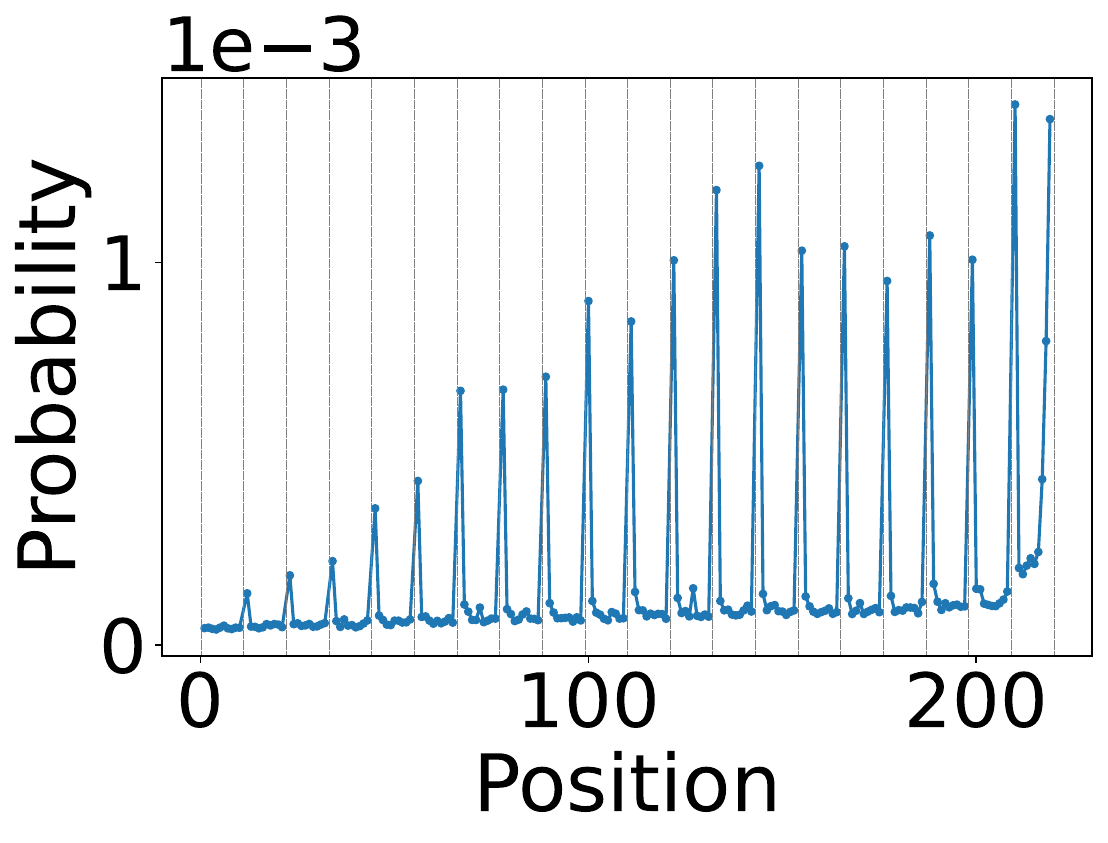} &
\FigGridCell{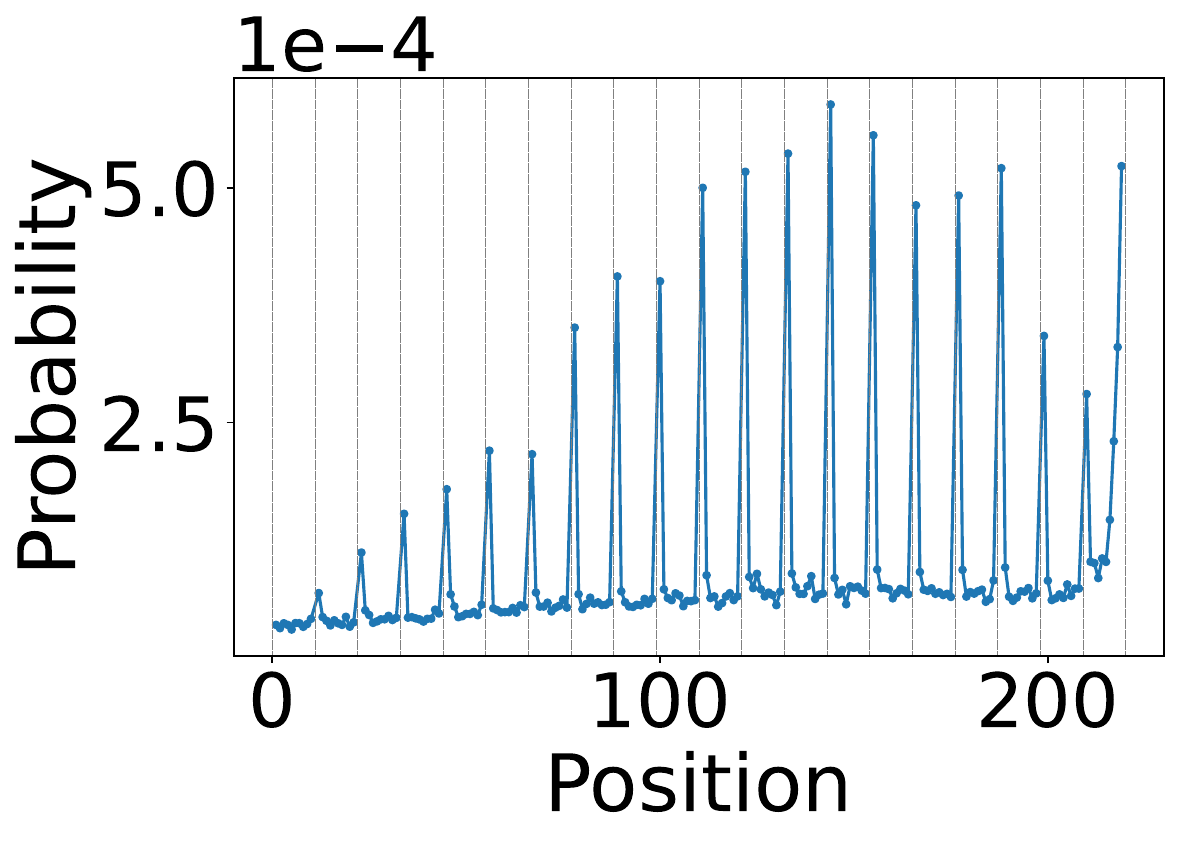} &
\FigGridCell{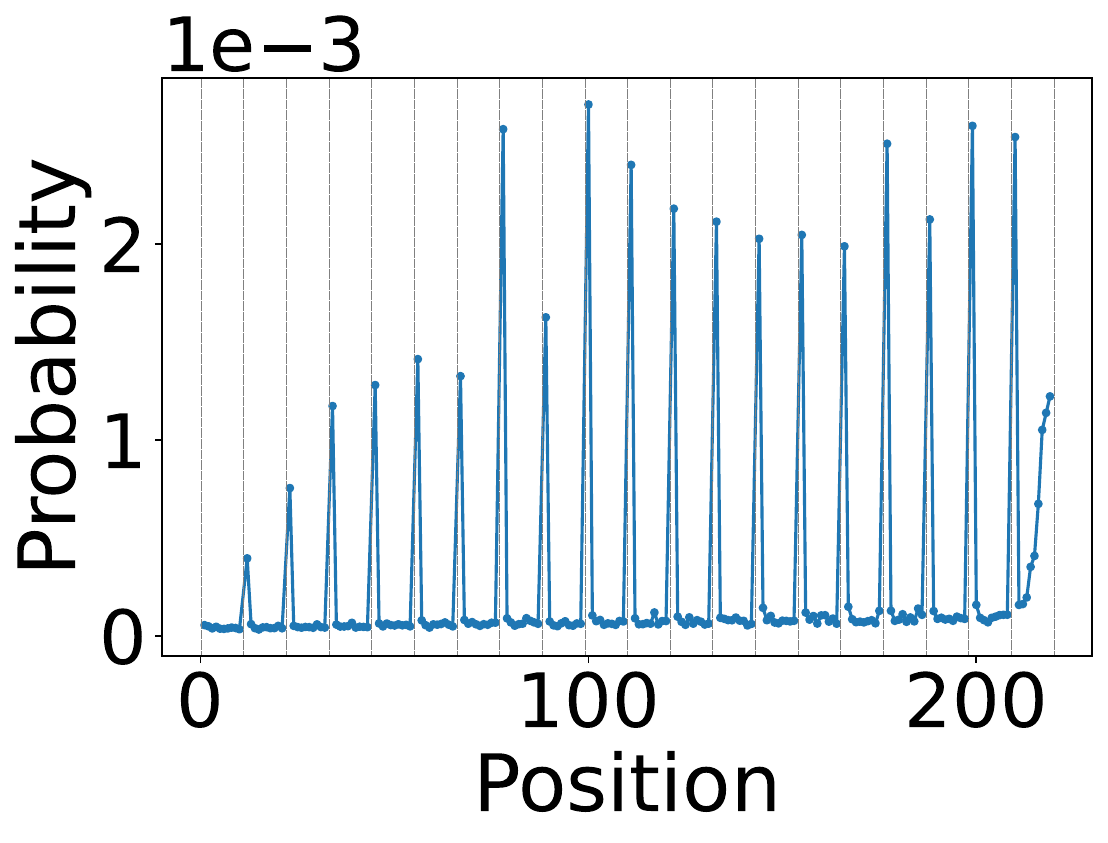} \\
\bottomrule
\end{tabular}
\endgroup

\caption{
Transformer ablation effect (Exp. 1). Probability vs. position after ablating Induction (Ind) or Random (Rand) heads (rows) for Llama, Mistral, Qwen, Gemma (model pairs per 2 rows). Columns show the number of ablated heads.
}
\label{fig:exp1_ablations}
\end{figure*}

\textbf{Procedure} For each transformer model (Llama, Mistral, Qwen, Gemma), we calculated induction scores for all attention heads (Layer $\times$ Head) using the method described in \citet{ji2024linking}. Heads were ranked by their induction scores. We then progressively ablated (set attention scores to zero, following \citet{crosbie2024induction}) the top 1, 10, 50, and 100 induction heads. As a control, we ablated the same numbers of randomly selected heads (ensuring they were not among the top 100 induction heads). 

\textbf{Results}
Figure~\ref{fig:exp1_ablations} shows the ablation results for Experiment 1. Ablating top induction heads consistently and significantly degrades the \plusoneTok\ token probability peaks, particularly with 50 or 100 heads ablated. Ablating random heads has a much less systematic effect, sometimes slightly perturbing probabilities but generally preserving the \plusoneTok\ peaks. This confirms the critical role of induction heads in the serial recall behavior. Across $4\times4$ (models$\times$ablation levels), the drop in the average \plusoneTok\ probability after ablating high–induction heads exceeds the drop after ablating the same number of random heads in $13/16$ settings (median difference $3.52\times 10^{-5}$, IQR $1.12\times 10^{-5}$–$6.09\times 10^{-5}$; maximum difference $2.26\times 10^{-4}$ for Llama-3.1-8B with 100 heads).
The few negative differences are small (at most $1.46\times 10^{-5}$).
See Table~\ref{tab:abl_summary} for complete numerical values.
These results, together with the shape changes in Figure~\ref{fig:exp1_ablations}, indicate that the serial-recall signal is disproportionately carried by high–induction heads, and the effect generally strengthens as more such heads are ablated (most clearly in Llama).

Similar effects were observed in Experiment 2 (Figures~\ref{fig:exp2_llama_ablation}-\ref{fig:exp2_gemma_ablation}). Ablating induction heads (top 5 rows in each figure, corresponding to probes 1-5) often disrupted the model's ability to selectively retrieve the single target episode. Instead of a dominant peak for the correct token, probabilities became more distributed across the potential target tokens from different episodes, indicating increased interference. Ablating random heads (bottom 5 rows) generally had a weaker impact, though ablating 50 or 100 random heads sometimes also perturbed the output distributions significantly. These results further underscore the importance of induction heads for temporal context separation and retrieval in transformers.

\subsubsection{State-space models: ablating high-attribution channels}
\label{sec:abl_ssm}

State-space models lack attention heads, so we adapt the same causal logic to channels (hidden dimensions) of the selective state-space block.

\textbf{Procedure}
For each channel $c$, we compute an ``induction-style'' salience score using a gradient-attribution proxy aggregated over prompts:
\[
s_c \;=\; \mathbb{E}\left[\,\left|u_c \cdot \frac{\partial \log \pnext(y\mid x)}{\partial u_c}\right|\,\right],
\]
where $u$ denotes the residual update produced by the selective SSM block and $y$ is the correct \plusoneTok\ token (Experiment 1) or correct target token (Experiment 2).
We ablate the top-ranked channels by zeroing their contribution to the residual update, and compare to ablating a matched number of random channels.
To compare fairly across model sizes, we ablate fixed fractions of channels (1\%, 5\%, and 10\%).

\paragraph{Results}
Figure~\ref{fig:exp1_ablations_ssm} shows impact of ablation in Experiment 1 and Figures~\ref{fig:exp2_mamba_ablation} and \ref{fig:exp2_r-gemma_ablation} in Experiment 2 (see also Table~\ref{tab:ssm_ablation} for numerical values).
In both experiments, for sufficiently large ablations (5--10\%), removing high-attribution channels reduces the recall signal more than random ablations across the tested SSM-family models, supporting the presence of a sparse subspace that is disproportionately responsible for the copy-like behavior.
Importantly, we do not claim a one-to-one equivalence between transformer induction heads and SSM channels; rather, these channels are \emph{functionally analogous} in the sense that they are causally implicated in the same induction-style behavioral signature.

These findings are consistent with recent analyses arguing that selective SSMs can be interpreted through an implicit-attention lens \citep{ali2025hidden_attention_mamba,zimerman2025implicit_attention},
and with reports of sparse memory channels that support primacy/recency effects in Mamba-like models \citep{airlangga2025primacy_recency_mamba,morita2025primacy_ssm}.

\section{Discussion}

This study investigated the role of temporal structure in shaping information retrieval within LLMs. By employing experiments designed to isolate temporal effects, we probed how both transformer and SSM architectures handle repeated information and overlapping temporal contexts. Our findings reveal significant temporal biases influencing retrieval and offer insights into the mechanisms underpinning these effects.

A key finding is a better understanding of positional biases in LLM context utilization, often described as the ``lost in the middle'' phenomenon \citep{liu2024lost}. Our results from Experiment 1 demonstrate that models exhibit strong primacy and recency effects, favoring information associated with tokens at the beginning or end of the context, even when semantic content is neutralized through permutation. This suggests the bias is deeply rooted in the sequential processing capabilities of these models, not merely an artifact of document structure or semantic coherence. The specific nature of this bias (primacy vs. recency dominance) varied across models and context parameters (Figures \ref{fig:single_target_all_repetitions_all_tokens}, \ref{fig:single_target_all_repetitions_plusone_tokens}), hinting at complex interactions between model architecture, training data, and the specific structure of the input sequence. In transformers, ``attention sinks'' can allocate baseline attention to early tokens \citep{xiao2023streamingllm, attn_sink_emerges_2025}. These sinks plausibly amplify primacy by biasing which earlier occurrence is selected by induction heads.  

Furthermore, Experiment 1 established that the tendency for serial recall, prioritizing the token immediately following a repeated instance (\plusoneTok\ token), is robust across both transformer and SSM architectures. This echoes prior work on induction heads in transformers \citep{olsson2022context, mistry2025emergence} but for the first time demonstrates its presence in SSMs like Mamba and Recurrent-Gemma as well. This shared characteristic suggests that basic sequence copying or pattern completion might be a convergent capability learned by different sequence modeling architectures.

Experiment 2 tests the models' ability to perform episodic-like retrieval by distinguishing between partially overlapping temporal contexts based on unique preceding tokens. Most models demonstrated a capacity for this temporal separation, correctly identifying the target token associated with the probed episode (Figure \ref{fig:exp2_all_repetitions}). However, this retrieval was imperfect, often showing interference from competing, temporally adjacent episodes (non-target peaks) and exhibiting strong positional effects, with episodes located near the end of the prompt generally retrieved more reliably (recency bias). This resonates with computational models of human episodic memory, which explicitly account for interference based on contextual similarity and temporal distance \citep{howard2002contextual, polyn2009context}. While LLMs showed some ability for temporal separation, their strong reliance on serial position contrasts with the more graded temporal contiguity effects seen in human recall \citep{kahana1996associative, ji2024linking}, suggesting potential differences in how temporal context is represented and utilized.

Our ablation studies in both transformer and state space models (Figures \ref{fig:exp1_ablations}, \ref{fig:exp1_ablations_ssm},  \ref{fig:exp2_llama_ablation} --\ref{fig:exp2_r-gemma_ablation}) provide mechanistic insights, confirming the crucial role of induction heads and their SSM analogues. Ablation significantly impaired both the \plusoneTok\ serial recall preference in Experiment 1 and the ability to selectively retrieve the correct episode in the presence of interference in Experiment 2. This aligns with the established function of induction heads in pattern matching and ICL \citep{olsson2022context, elhage2021mathematical, crosbie2024induction} and supports their proposed link to episodic memory functions \citep{ji2024linking}.

Despite lacking explicit attention mechanisms, SSMs showed serial‑position structure in retrieval. For example, Recurrent‑Gemma displayed U‑shaped positional preference curves in the serial‑recall probe (Figure \ref{fig:single_target_all_repetitions_plusone_tokens}) and a recency advantage in episodic retrieval under overlap (Figure \ref{fig:exp2_all_repetitions}), whereas Mamba and Falcon‑Mamba exhibited less robust episodic retrieval when the probed episode occurred near the end of the prompt (Figure \ref{fig:exp2_all_repetitions}). Together, these patterns suggest that temporal biases are not solely a consequence of attention, but can also arise from more general constraints on representing and querying long sequences. Potential contributing factors include how models encode positional information (explicitly via positional representations in transformers, or implicitly through recurrent/state dynamics in SSMs) and limitations in maintaining distinct, queryable traces over long temporal distances. SSMs, for instance, compress context history into a fixed‑size state \citep{gu2023mamba, jelassi2024repeat}; the evolution, saturation, and selective‑forgetting dynamics of this state could systematically privilege certain parts of the sequence, yielding primacy/recency‑like effects. Consistent with this, our ablations in state‑space models show that removing a small fraction of high‑attribution channels produces degradations analogous to ablating high‑induction heads in transformers, supporting the presence of a sparse subspace that implements induction‑style copying and episodic separation. While some studies suggest transformers excel at exact copying tasks \citep{jelassi2024repeat}, our interference‑based probes indicate that when retrieval depends on relative temporal position and resisting overlap, SSMs can exhibit comparable position‑dependent limitations and biases.

These findings have several implications. For LLM development, they underscore that addressing the ``lost in the middle'' problem requires tackling fundamental temporal processing limitations, potentially related to positional information or state management, which may persist even in non-attention architectures. For cognitive science, our methodology provides a controlled paradigm for comparing how different computational architectures handle temporal context and interference, offering insights into the functional constraints shaping memory-like phenomena in artificial systems. 

\section*{Limitations}
We primarily study token-random prompts rather than natural text and analyze next-token probabilities. While methodologically necessary to isolate temporal effects from semantic confounds, random-token sequences represent a simplified information environment. Future work can build upon this understanding of pure temporal biases to explore the more complex interplay between \emph{when} something was said and \emph{what} was said in rich, semantic contexts.

Our SSM ``high-attribution channel'' identification uses gradient-based salience as a heuristic.
Although targeted ablations provide causal evidence that a sparse subset of channels supports the induction-style signal, this is not a one-to-one analog of transformer induction heads, and direct alignment with other proposed SSM memory decompositions remains future work.

\bibliographystyle{acl_natbib}
\bibliography{references}

\begin{thebibliography}{31}
\providecommand{\natexlab}[1]{#1}

\bibitem[{Airlangga et~al.(2025)Airlangga, AlQuabeh, Nwadike, and
  Inui}]{airlangga2025primacy_recency_mamba}
Muhammad~Cendekia Airlangga, Hilal AlQuabeh, Munachiso~S Nwadike, and Kentaro
  Inui. 2025.
\newblock \href {https://arxiv.org/abs/2506.15156} {Emergence of primacy and
  recency effect in mamba: A mechanistic point of view}.
\newblock \emph{arXiv preprint arXiv:2506.15156}.

\bibitem[{Ali et~al.(2025)Ali, Zimerman, and
  Wolf}]{ali2025hidden_attention_mamba}
Ameen Ali, Itamar Zimerman, and Lior Wolf. 2025.
\newblock \href {https://aclanthology.org/2025.acl-long.76/} {The hidden
  attention of mamba models}.
\newblock In \emph{Proceedings of the 63rd Annual Meeting of the Association
  for Computational Linguistics (Volume 1: Long Papers)}.

\bibitem[{Botev et~al.(2024)Botev, De, Smith, Fernando, Muraru, Haroun,
  Berrada, Pascanu, Sessa, Dadashi et~al.}]{botev2024recurrentgemma}
Aleksandar Botev, Soham De, Samuel~L Smith, Anushan Fernando, George-Cristian
  Muraru, Ruba Haroun, Leonard Berrada, Razvan Pascanu, Pier~Giuseppe Sessa,
  Robert Dadashi, and 1 others. 2024.
\newblock Recurrentgemma: Moving past transformers for efficient open language
  models.
\newblock \emph{arXiv preprint arXiv:2404.07839}.

\bibitem[{Brown et~al.(2020)Brown, Mann, Ryder, Subbiah, Kaplan, Dhariwal, ...,
  and Amodei}]{brown2020language}
Tom~B. Brown, Benjamin Mann, Nick Ryder, Melanie Subbiah, Jared Kaplan,
  Prafulla Dhariwal, ..., and Dario Amodei. 2020.
\newblock Language models are few-shot learners.
\newblock In \emph{Advances in Neural Information Processing Systems}.

\bibitem[{Crosbie and Shutova(2024)}]{crosbie2024induction}
Joy Crosbie and Ekaterina Shutova. 2024.
\newblock Induction heads as an essential mechanism for pattern matching in
  in-context learning.
\newblock \emph{arXiv preprint arXiv:2407.07011}.

\bibitem[{Dubey et~al.(2024)Dubey, Jauhri, Pandey, Kadian, Al-Dahle, Letman,
  Mathur, Schelten, Yang, Fan et~al.}]{dubey2024llama}
Abhimanyu Dubey, Abhinav Jauhri, Abhinav Pandey, Abhishek Kadian, Ahmad
  Al-Dahle, Aiesha Letman, Akhil Mathur, Alan Schelten, Amy Yang, Angela Fan,
  and 1 others. 2024.
\newblock The llama 3 herd of models.
\newblock \emph{arXiv preprint arXiv:2407.21783}.

\bibitem[{Ebbinghaus(1913)}]{ebbinghaus1913memory}
Hermann Ebbinghaus. 1913.
\newblock \emph{Memory: A contribution to experimental psychology}.
\newblock 3. Teachers college, Columbia university.

\bibitem[{Elhage et~al.(2021)Elhage, Nanda, Olsson, Henighan, Joseph, Mann,
  Askell, Bai, Chen, Conerly et~al.}]{elhage2021mathematical}
Nelson Elhage, Neel Nanda, Catherine Olsson, Tom Henighan, Nicholas Joseph, Ben
  Mann, Amanda Askell, Yuntao Bai, Anna Chen, Tom Conerly, and 1 others. 2021.
\newblock A mathematical framework for transformer circuits.
\newblock \emph{Transformer Circuits Thread}, 1(1):12.

\bibitem[{Gu and Dao(2023)}]{gu2023mamba}
Albert Gu and Tri Dao. 2023.
\newblock Mamba: Linear-time sequence modeling with selective state spaces.
\newblock \emph{arXiv preprint arXiv:2312.00752}.

\bibitem[{Gu et~al.(2024)Gu, Pang, Du, Liu, Zhang, Du, Wang, and
  Lin}]{attn_sink_emerges_2025}
Xiangming Gu, Tianyu Pang, Chao Du, Qian Liu, Fengzhuo Zhang, Cunxiao Du,
  Ye~Wang, and Min Lin. 2024.
\newblock When attention sink emerges in language models: An empirical view.
\newblock \emph{arXiv preprint arXiv:2410.10781}.

\bibitem[{Howard and Kahana(1999)}]{howard2002contextual}
Marc~W Howard and Michael~J Kahana. 1999.
\newblock Contextual variability and serial position effects in free recall.
\newblock \emph{Journal of Experimental Psychology: Learning, Memory, and
  Cognition}, 25(4):923.

\bibitem[{Janik(2023)}]{janik2023aspects}
Romuald~A Janik. 2023.
\newblock Aspects of human memory and large language models.
\newblock \emph{arXiv preprint arXiv:2311.03839}.

\bibitem[{Jelassi et~al.(2024)Jelassi, Brandfonbrener, Kakade, and
  Malach}]{jelassi2024repeat}
Samy Jelassi, David Brandfonbrener, Sham~M Kakade, and Eran Malach. 2024.
\newblock Repeat after me: Transformers are better than state space models at
  copying.
\newblock \emph{arXiv preprint arXiv:2402.01032}.

\bibitem[{Ji-An et~al.(2024)Ji-An, Zhou, Benna, and Mattar}]{ji2024linking}
Li~Ji-An, Corey~Y Zhou, Marcus~K Benna, and Marcelo~G Mattar. 2024.
\newblock Linking in-context learning in transformers to human episodic memory.
\newblock \emph{arXiv preprint arXiv:2405.14992}.

\bibitem[{Jiang et~al.(2023)Jiang, Sablayrolles, Mensch, Bamford, Chaplot,
  Casas, Bressand, Lengyel, Lample, Saulnier et~al.}]{jiang2023mistral}
Albert~Q Jiang, Alexandre Sablayrolles, Arthur Mensch, Chris Bamford,
  Devendra~Singh Chaplot, Diego de~las Casas, Florian Bressand, Gianna Lengyel,
  Guillaume Lample, Lucile Saulnier, and 1 others. 2023.
\newblock Mistral 7b.
\newblock \emph{arXiv preprint arXiv:2310.06825}.

\bibitem[{Kahana(1996)}]{kahana1996associative}
Michael~J Kahana. 1996.
\newblock Associative retrieval processes in free recall.
\newblock \emph{Memory \& Cognition}, 24(1):103--109.

\bibitem[{Liu et~al.(2024)Liu, Lin, Hewitt, Paranjape, Bevilacqua, Petroni, and
  Liang}]{liu2024lost}
Nelson~F Liu, Kevin Lin, John Hewitt, Ashwin Paranjape, Michele Bevilacqua,
  Fabio Petroni, and Percy Liang. 2024.
\newblock Lost in the middle: How language models use long contexts.
\newblock \emph{Transactions of the Association for Computational Linguistics},
  12:157--173.

\bibitem[{Mistry et~al.(2025)Mistry, Bajaj, Aggarwal, Maini, and
  Tiganj}]{mistry2025emergence}
Deven~Mahesh Mistry, Anooshka Bajaj, Yash Aggarwal, Sahaj~Singh Maini, and
  Zoran Tiganj. 2025.
\newblock Emergence of episodic memory in transformers: Characterizing changes
  in temporal structure of attention scores during training.
\newblock \emph{arXiv preprint arXiv:2502.06902}.

\bibitem[{Morita(2025)}]{morita2025primacy_ssm}
Takashi Morita. 2025.
\newblock \href {https://arxiv.org/abs/2502.13729} {Emergence of the primacy
  effect in structured state-space models}.
\newblock \emph{arXiv preprint arXiv:2502.13729}.

\bibitem[{Murdock(1962)}]{murdock1962serial}
Bennet B.~Jr. Murdock. 1962.
\newblock The serial position effect of free recall.
\newblock \emph{Journal of Experimental Psychology}, 64(5):482--488.

\bibitem[{Olsson et~al.(2022)Olsson, Elhage, Nanda, Joseph, DasSarma, Henighan,
  Mann, Askell, Bai, Chen et~al.}]{olsson2022context}
Catherine Olsson, Nelson Elhage, Neel Nanda, Nicholas Joseph, Nova DasSarma,
  Tom Henighan, Ben Mann, Amanda Askell, Yuntao Bai, Anna Chen, and 1 others.
  2022.
\newblock In-context learning and induction heads.
\newblock \emph{arXiv preprint arXiv:2209.11895}.

\bibitem[{Pink et~al.(2024)Pink, Vo, Wu, Mu, Turek, Hasson, Norman, Michelmann,
  Huth, and Toneva}]{pink2024assessing}
Mathis Pink, Vy~A Vo, Qinyuan Wu, Jianing Mu, Javier~S Turek, Uri Hasson,
  Kenneth~A Norman, Sebastian Michelmann, Alexander Huth, and Mariya Toneva.
  2024.
\newblock Assessing episodic memory in llms with sequence order recall tasks.
\newblock \emph{arXiv preprint arXiv:2410.08133}.

\bibitem[{Polyn et~al.(2009)Polyn, Norman, and Kahana}]{polyn2009context}
Sean~M Polyn, Kenneth~A Norman, and Michael~J Kahana. 2009.
\newblock A context maintenance and retrieval model of organizational processes
  in free recall.
\newblock \emph{Psychological review}, 116(1):129.

\bibitem[{Singh et~al.(2024)Singh, Chan, Moskovitz, Grant, Saxe, and
  Hill}]{singh2024transient}
Aaditya Singh, Stephanie Chan, Ted Moskovitz, Erin Grant, Andrew Saxe, and
  Felix Hill. 2024.
\newblock The transient nature of emergent in-context learning in transformers.
\newblock \emph{Advances in Neural Information Processing Systems}, 36.

\bibitem[{Team et~al.(2024)Team, Riviere, Pathak, Sessa, Hardin, Bhupatiraju,
  Hussenot, Mesnard, Shahriari, Ram{\'e} et~al.}]{team2024gemma}
Gemma Team, Morgane Riviere, Shreya Pathak, Pier~Giuseppe Sessa, Cassidy
  Hardin, Surya Bhupatiraju, L{\'e}onard Hussenot, Thomas Mesnard, Bobak
  Shahriari, Alexandre Ram{\'e}, and 1 others. 2024.
\newblock Gemma 2: Improving open language models at a practical size.
\newblock \emph{arXiv preprint arXiv:2408.00118}.

\bibitem[{Vaswani et~al.(2017)Vaswani, Shazeer, Parmar, Uszkoreit, Jones,
  Gomez, Kaiser, and Polosukhin}]{vaswani2017attention}
Ashish Vaswani, Noam Shazeer, Niki Parmar, Jakob Uszkoreit, Llion Jones,
  Aidan~N. Gomez, {\L}ukasz Kaiser, and Illia Polosukhin. 2017.
\newblock Attention is all you need.
\newblock In \emph{Advances in Neural Information Processing Systems}.

\bibitem[{Wu et~al.(2025)Wu, Wang, Jegelka, and
  Jadbabaie}]{wu2025position_bias}
Xinyi Wu, Yifei Wang, Stefanie Jegelka, and Ali Jadbabaie. 2025.
\newblock \href {https://arxiv.org/abs/2502.01951} {On the emergence of
  position bias in transformers}.
\newblock \emph{arXiv preprint arXiv:2502.01951}.

\bibitem[{Xiao et~al.(2023)Xiao, Tian, Chen, Han, and
  Lewis}]{xiao2023streamingllm}
Guangxuan Xiao, Yuandong Tian, Beidi Chen, Song Han, and Mike Lewis. 2023.
\newblock Efficient streaming language models with attention sinks.
\newblock \emph{arXiv preprint arXiv:2309.17453}.

\bibitem[{Yang et~al.(2024)Yang, Yang, Zhang, Hui, Zheng, Yu, Li, Liu, Huang,
  Wei et~al.}]{yang2024qwen2}
An~Yang, Baosong Yang, Beichen Zhang, Binyuan Hui, Bo~Zheng, Bowen Yu,
  Chengyuan Li, Dayiheng Liu, Fei Huang, Haoran Wei, and 1 others. 2024.
\newblock Qwen2.5 technical report.
\newblock \emph{arXiv preprint arXiv:2412.15115}.

\bibitem[{Zimerman et~al.(2024)Zimerman, Ali, and
  Wolf}]{zimerman2025implicit_attention}
Itamar Zimerman, Ameen Ali, and Lior Wolf. 2024.
\newblock Explaining modern gated-linear rnns via a unified implicit attention
  formulation.
\newblock \emph{arXiv preprint arXiv:2405.16504}.

\bibitem[{Zuo et~al.(2024)Zuo, Velikanov, Rhaiem, Chahed, Belkada, Kunsch, and
  Hacid}]{zuo2024falcon}
Jingwei Zuo, Maksim Velikanov, Dhia~Eddine Rhaiem, Ilyas Chahed, Younes
  Belkada, Guillaume Kunsch, and Hakim Hacid. 2024.
\newblock Falcon mamba: The first competitive attention-free 7b language model.
\newblock \emph{arXiv preprint arXiv:2410.05355}.

\end{thebibliography}


\appendix

\renewcommand{\thetable}{A\arabic{table}}
\renewcommand{\thefigure}{A\arabic{figure}}
\setcounter{table}{0}
\setcounter{figure}{0}

\clearpage

\begin{table*}[t]
\centering
\small
\setlength{\tabcolsep}{6pt}
\begin{tabular}{l r r r r}
\hline
\textbf{Model} & \textbf{Ablation Level} & \textbf{Induction Change} & \textbf{Random Change} & \textbf{Difference in Change} $\boldsymbol{\Delta}$ \\
\hline
Llama & 1   & \(9.35\times10^{-5}\)  & \(4.96\times10^{-5}\)   & \textbf{\(4.39\times10^{-5}\)} \\
                      & 10  & \(1.34\times10^{-4}\)  & \(3.81\times10^{-5}\)   & \textbf{\(9.63\times10^{-5}\)} \\
                      & 50  & \(1.81\times10^{-4}\)  & \(1.72\times10^{-5}\)   & \textbf{\(1.64\times10^{-4}\)} \\
                      & 100 & \(2.24\times10^{-4}\)  & \(-1.91\times10^{-6}\)  & \textbf{\(2.26\times10^{-4}\)} \\
Mistral & 1   & \(1.90\times10^{-5}\)  & \(-7.51\times10^{-6}\)  & \textbf{\(2.65\times10^{-5}\)} \\
                         & 10  & \(2.09\times10^{-5}\)  & \(2.03\times10^{-6}\)   & \textbf{\(1.88\times10^{-5}\)} \\
                         & 50  & \(6.27\times10^{-5}\)  & \(-4.65\times10^{-6}\)  & \textbf{\(6.74\times10^{-5}\)} \\
                         & 100 & \(5.85\times10^{-5}\)  & \(-1.19\times10^{-7}\)  & \textbf{\(5.87\times10^{-5}\)} \\
Qwen  & 1   & \(-2.62\times10^{-6}\) & \(8.58\times10^{-6}\)   & \(-1.12\times10^{-5}\) \\
                      & 10  & \(1.57\times10^{-5}\)  & \(3.46\times10^{-6}\)   & \textbf{\(1.23\times10^{-5}\)} \\
                      & 50  & \(7.45\times10^{-5}\)  & \(2.49\times10^{-5}\)   & \textbf{\(4.95\times10^{-5}\)} \\
                      & 100 & \(7.93\times10^{-5}\)  & \(3.17\times10^{-5}\)   & \textbf{\(4.76\times10^{-5}\)} \\
Gemma         & 1   & \(-6.59\times10^{-6}\) & \(6.56\times10^{-7}\)   & \(-7.24\times10^{-6}\) \\
                      & 10  & \(-1.31\times10^{-5}\) & \(1.46\times10^{-6}\)   & \(-1.46\times10^{-5}\) \\
                      & 50  & \(1.72\times10^{-5}\)  & \(-2.80\times10^{-6}\)  & \textbf{\(2.00\times10^{-5}\)} \\
                      & 100 & \(1.53\times10^{-5}\)  & \(7.27\times10^{-6}\)   & \textbf{\(8.00\times10^{-6}\)} \\
\hline
\end{tabular}
\caption{Ablation effects on the average \plusoneTok\ probability in Exp.~1. 
\emph{Induction Change} and \emph{Random Change} are changes in \plusoneTok\ probability relative to the unablated model. 
\(\Delta =\) Induction Change $-$ Random Change; positive \(\Delta\) means ablating high-induction heads reduces the \((+1)\) signal more than ablating random heads.}
\label{tab:abl_summary}
\end{table*}

\begin{table*}[t]
\centering
\small
\begin{tabular}{llrr}
\toprule
\textbf{Experiment} & \textbf{Model / Ablation} & \textbf{Induction Change} & \textbf{Random Change} \\
\midrule
\multirow{9}{*}{Exp.\ 1} 
& Mamba, 1\%   & +67.8\%  & -91.7\% \\
& Mamba, 5\%   & +98.4\%  & +6.2\%  \\
& Mamba, 10\%  & +94.2\%  & +58.3\% \\
& R-Gemma, 1\% & -17.2\%  & +9.6\%  \\
& R-Gemma, 5\% & +68.2\%  & +43.2\% \\
& R-Gemma, 10\%& +96.9\%  & +58.9\% \\
& Falcon-M, 1\%& +11.2\%  & +35.7\% \\
& Falcon-M, 5\%& +92.6\%  & +11.0\% \\
& Falcon-M, 10\%& +90.4\% & +38.5\% \\
\midrule
\multirow{6}{*}{Exp.\ 2} 
& Mamba, 1\%   & +16.4\%  & -69.7\% \\
& Mamba, 5\%   & +64.1\%  & +27.3\% \\
& Mamba, 10\%  & +82.3\%  & +62.1\% \\
& R-Gemma, 1\% & -7.3\%   & +14.7\% \\
& R-Gemma, 5\% & +65.2\%  & +44.0\% \\
& R-Gemma, 10\% & +91.6\% & +70.1\% \\
\bottomrule
\end{tabular}
\caption{
State-space/recurrent channel ablations. ``Change'' denotes the relative change in the task signal. Experiment 1: mean \plusoneTok\ recall;
Experiment 2: target probability / separation signal after ablating high-attribution channels (Induction) vs.\ a matched number
of random channels. Falcon-M was too computationally demanding for Experiment 2 ablations.
}
\label{tab:ssm_ablation}
\end{table*}

\begin{figure*}[h!]
\centering
\begingroup
\setlength{\FigGridLabelW}{0.65cm}
\setlength{\FigGridCellH}{2.25cm} 
\setlength{\FigGridLabelSep}{1pt}
\setlength{\FigGridColSep}{0.5pt}

\FigGridSetup{5}
\renewcommand{\arraystretch}{1.0}

\begin{tabular}{@{}>{\centering\arraybackslash}m{\FigGridLabelW}@{\hspace{\FigGridLabelSep}}*{5}{>{\centering\arraybackslash}m{\FigGridPlotW}}@{}}
\toprule
& \multicolumn{5}{c}{\FigHdr \# Repeats} \\
\cmidrule(lr){2-6}
& \FigHdr 3 & \FigHdr 5 & \FigHdr 10 & \FigHdr 20 & \FigHdr 40 \\
\midrule
\FigGridVLabel{Llama} &
\FigGridCell{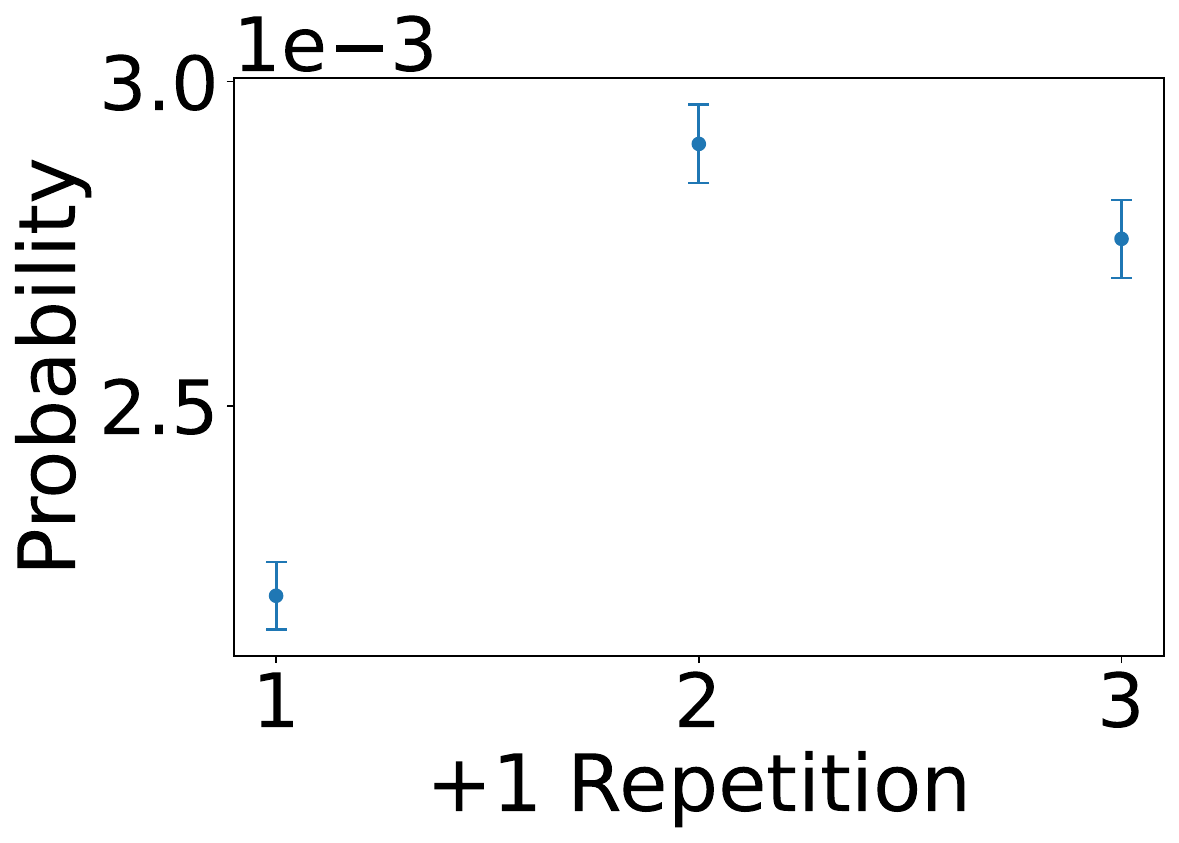} &
\FigGridCell{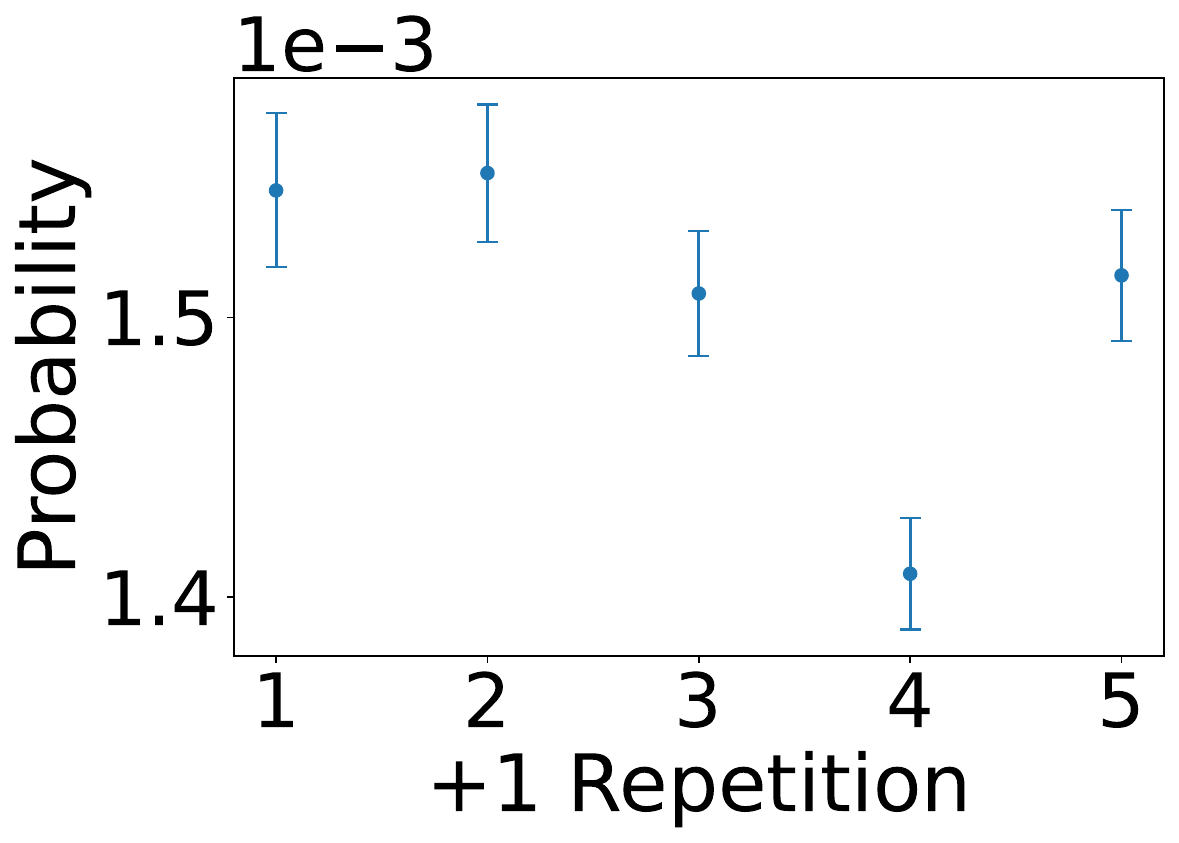} &
\FigGridCell{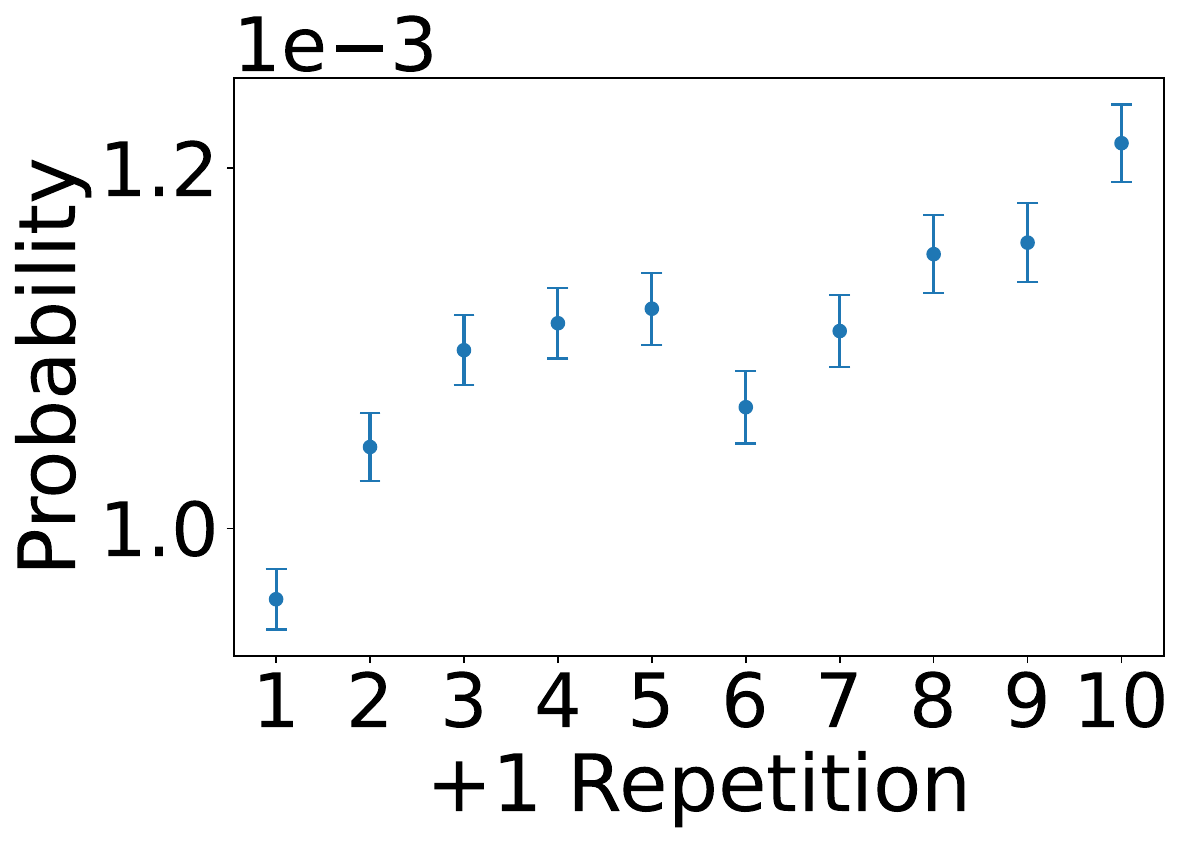} &
\FigGridCell{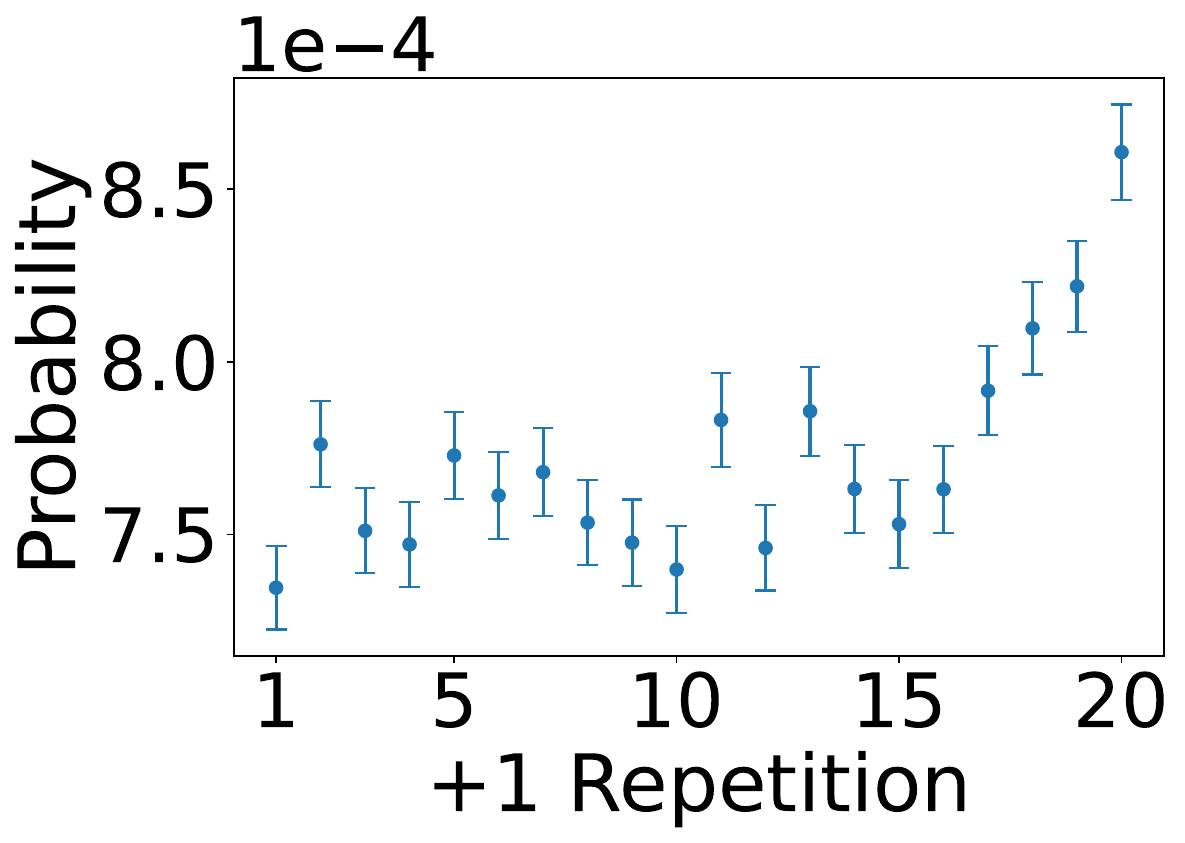} &
\FigGridCell{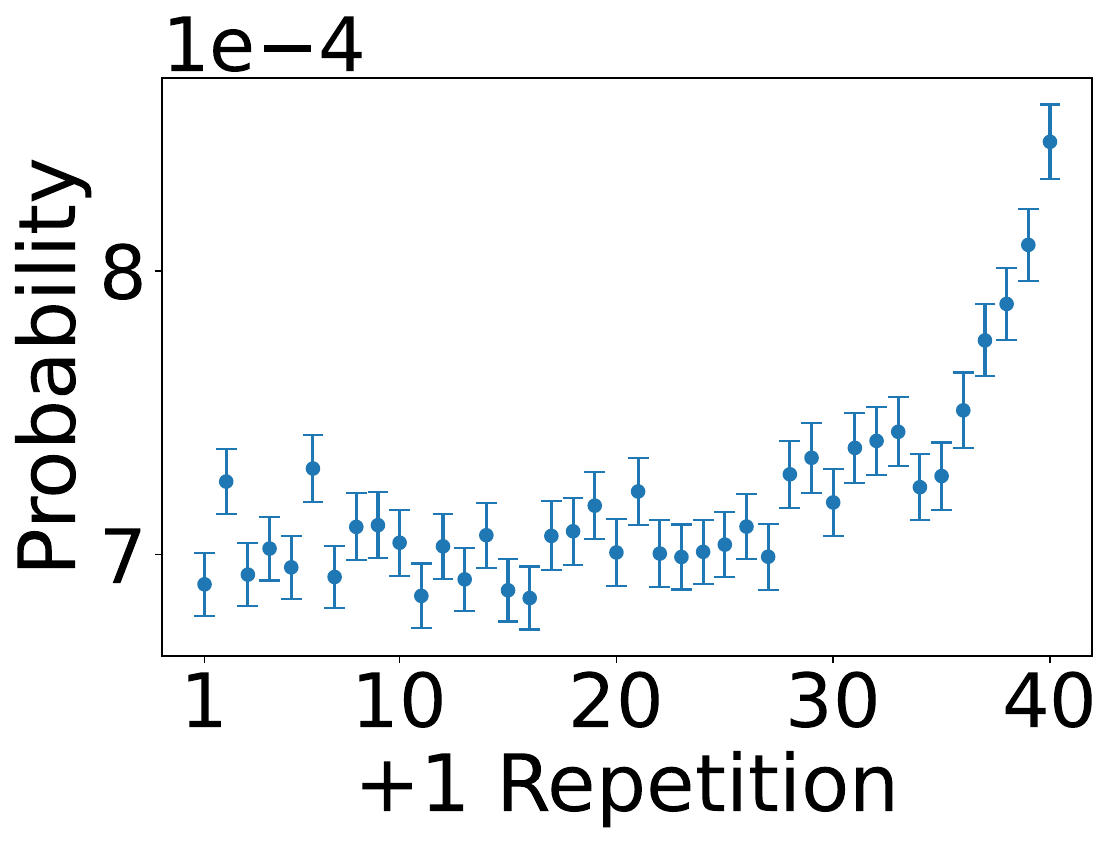} \\
\FigGridVLabel{Mistral} &
\FigGridCell{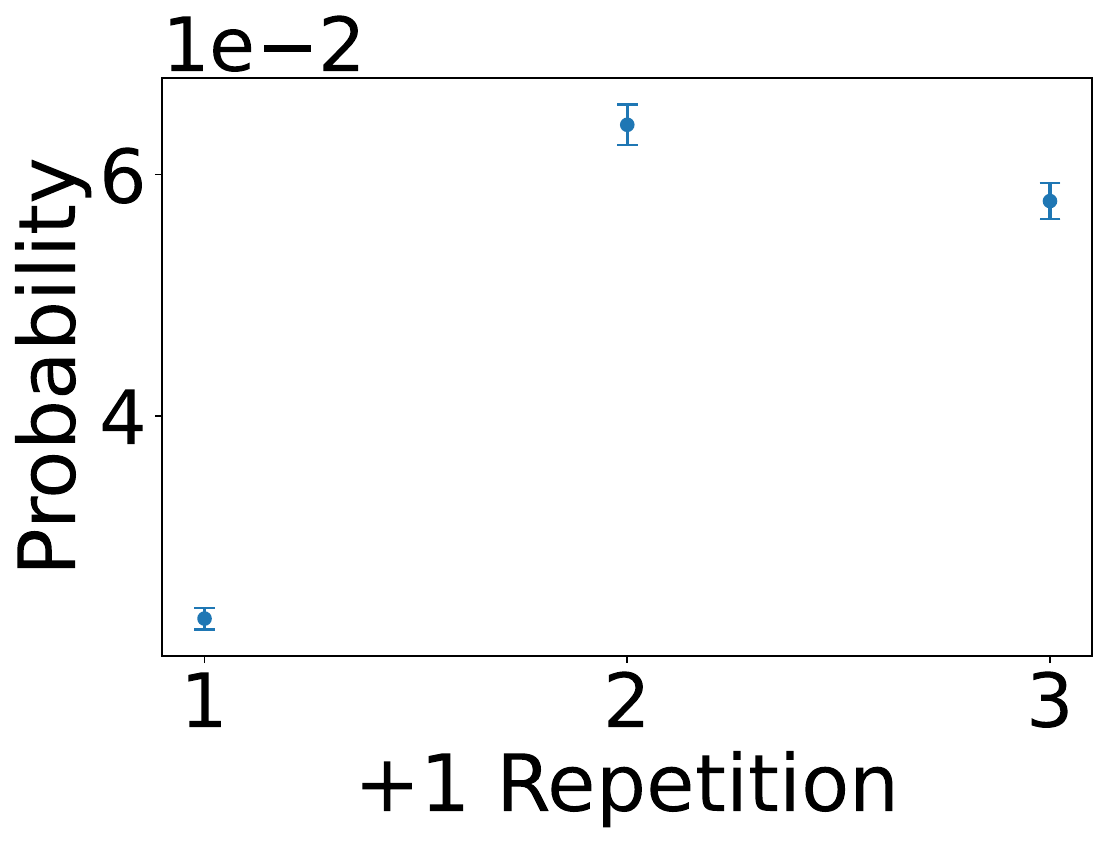} &
\FigGridCell{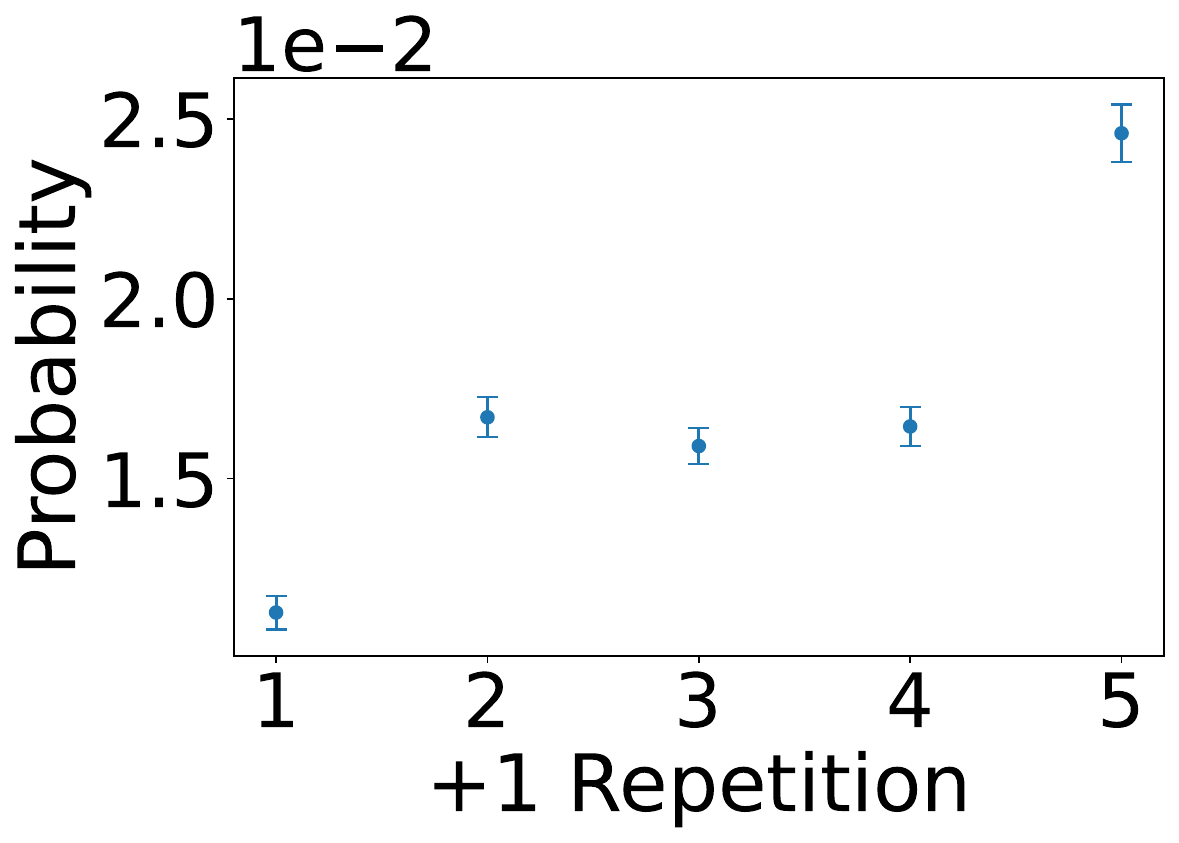} &
\FigGridCell{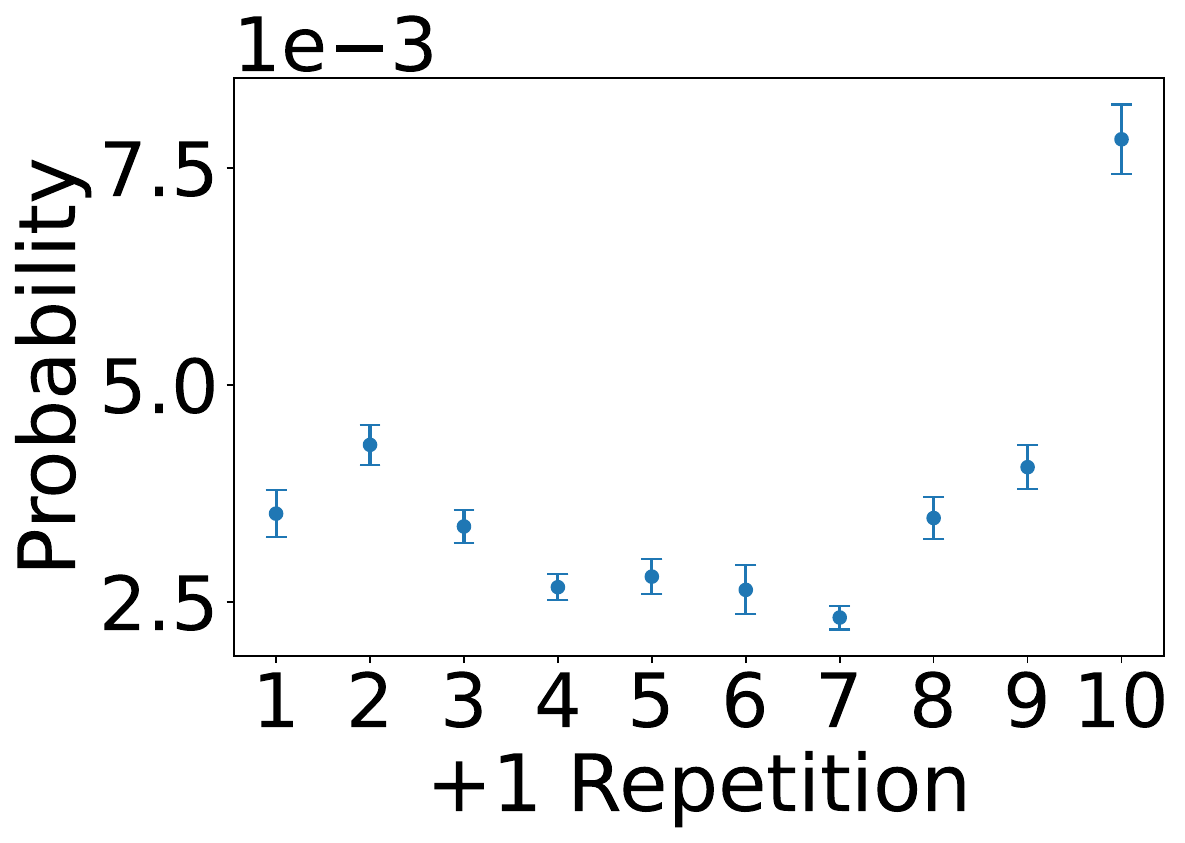} &
\FigGridCell{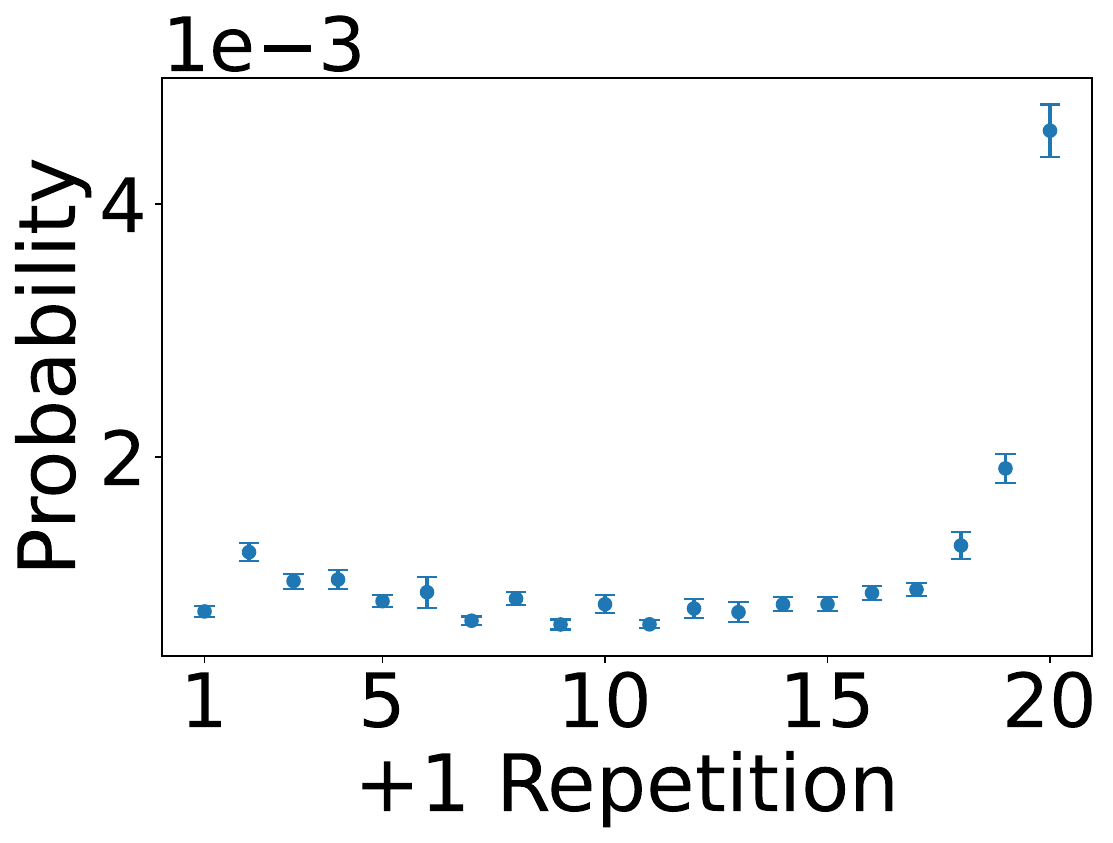} &
\FigGridCell{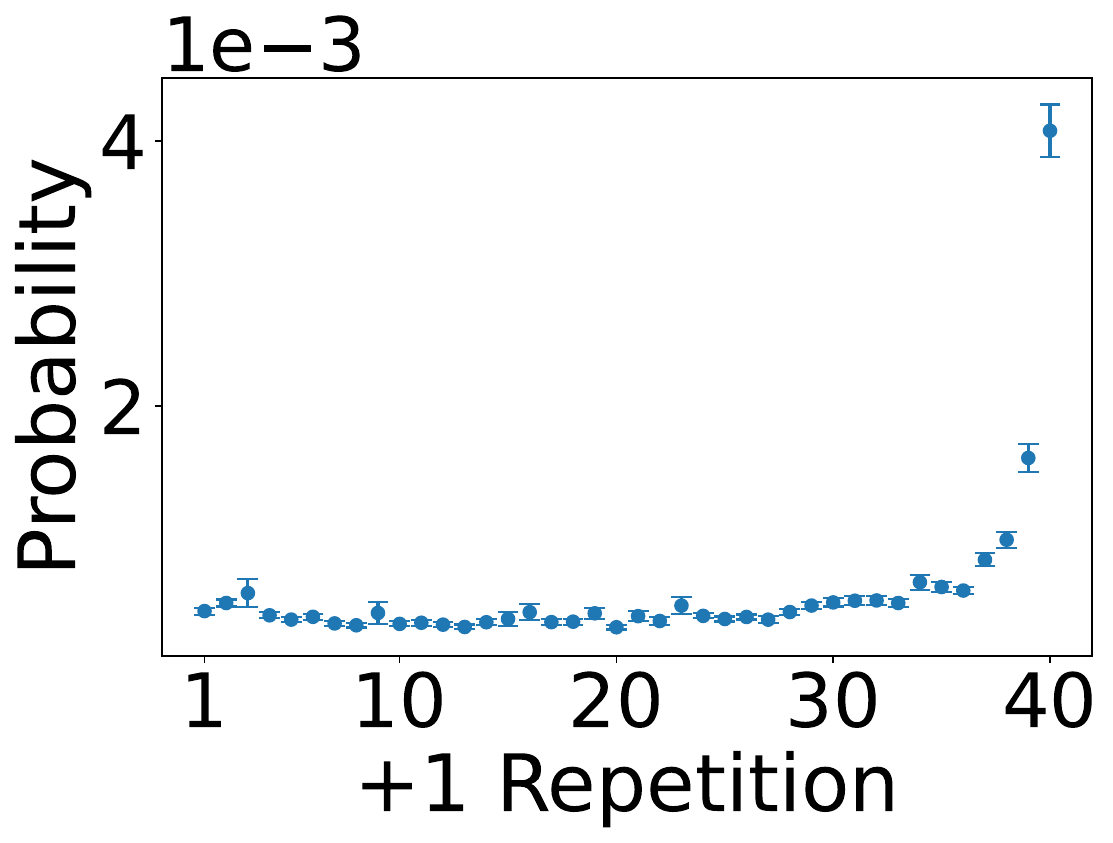} \\
\FigGridVLabel{Qwen} &
\FigGridCell{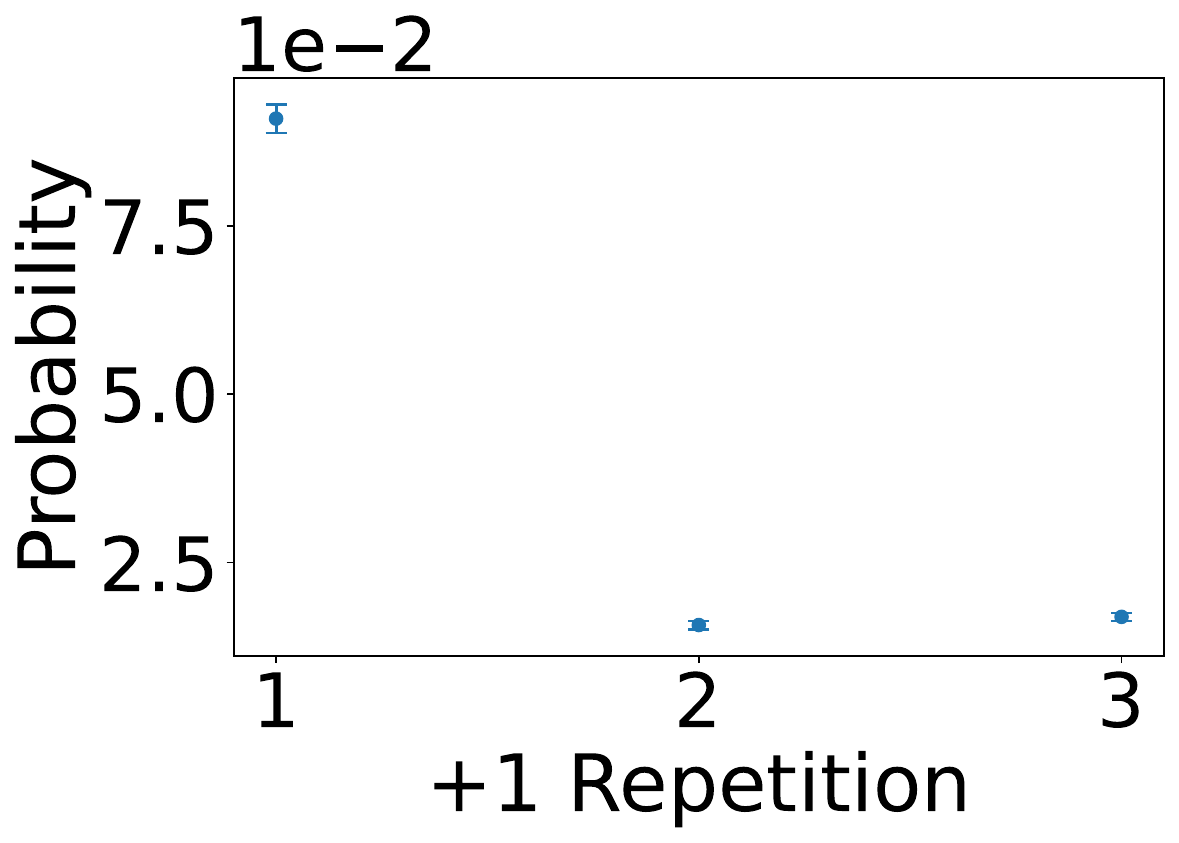} &
\FigGridCell{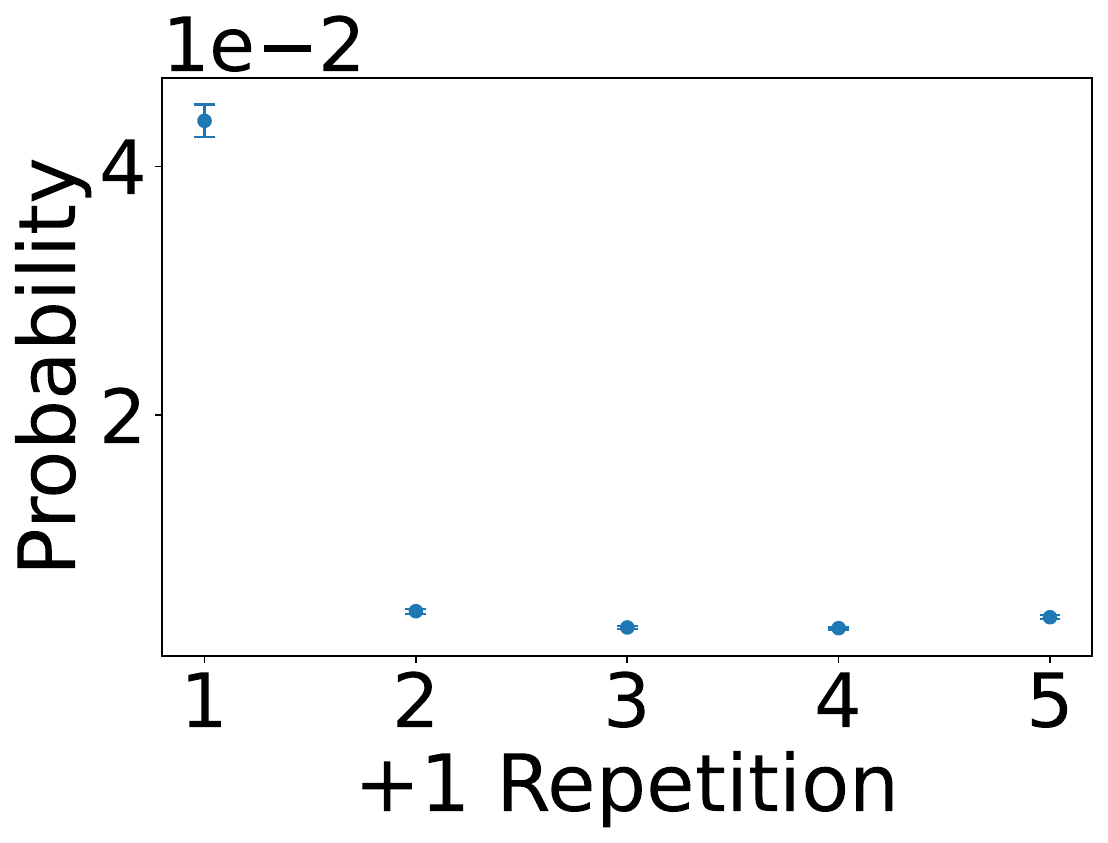} &
\FigGridCell{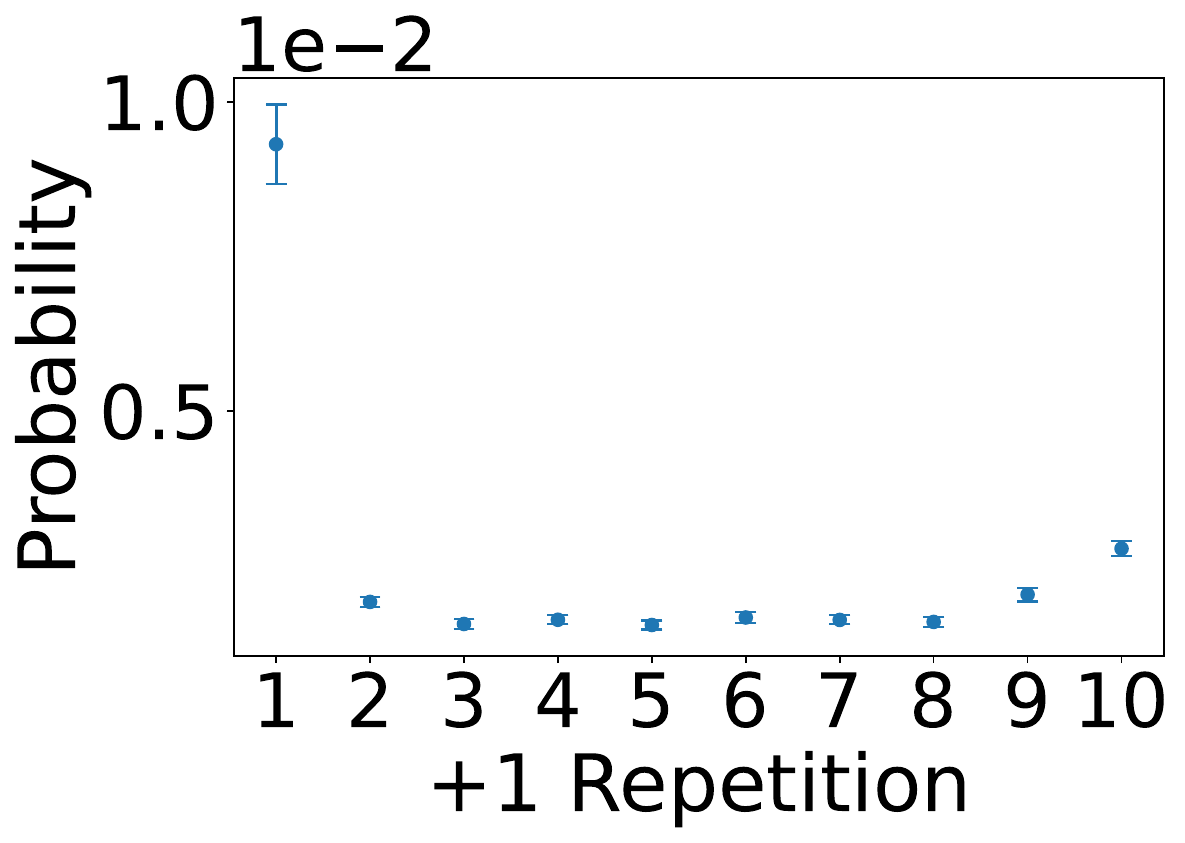} &
\FigGridCell{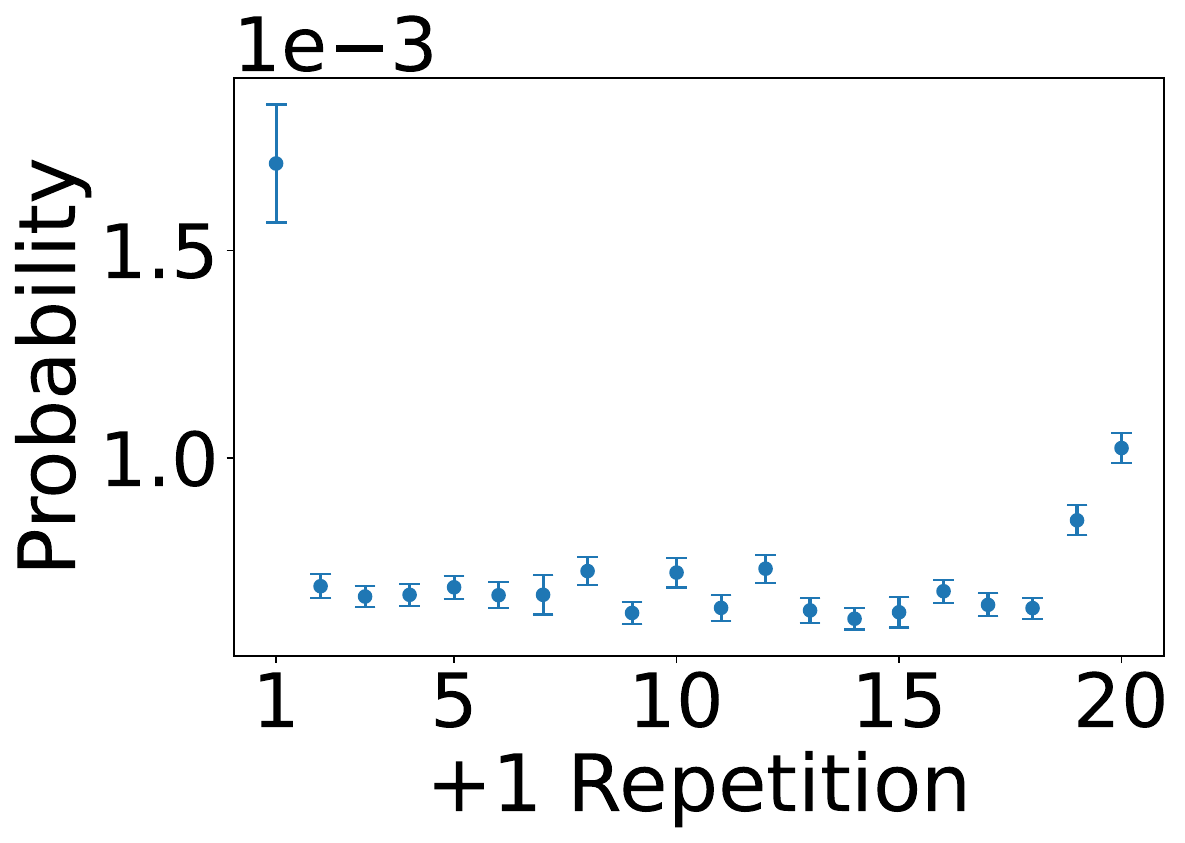} &
\FigGridCell{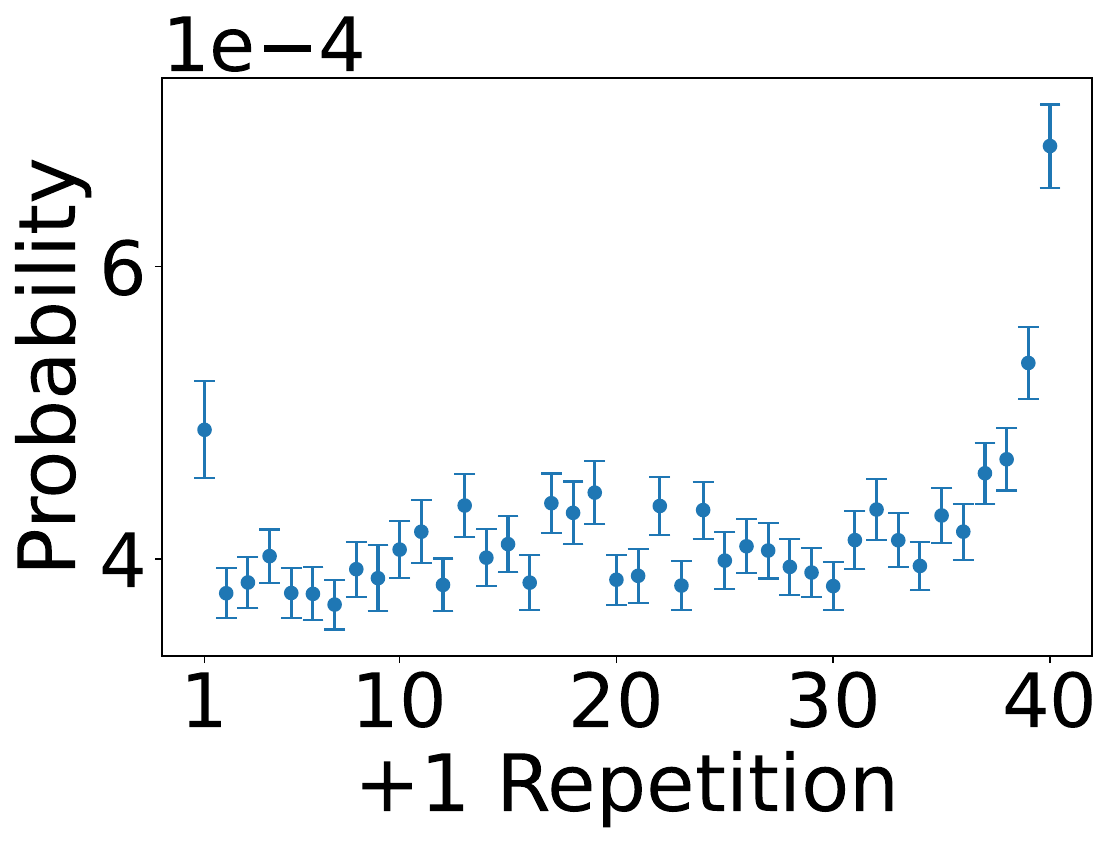} \\
\FigGridVLabel{Gemma} &
\FigGridCell{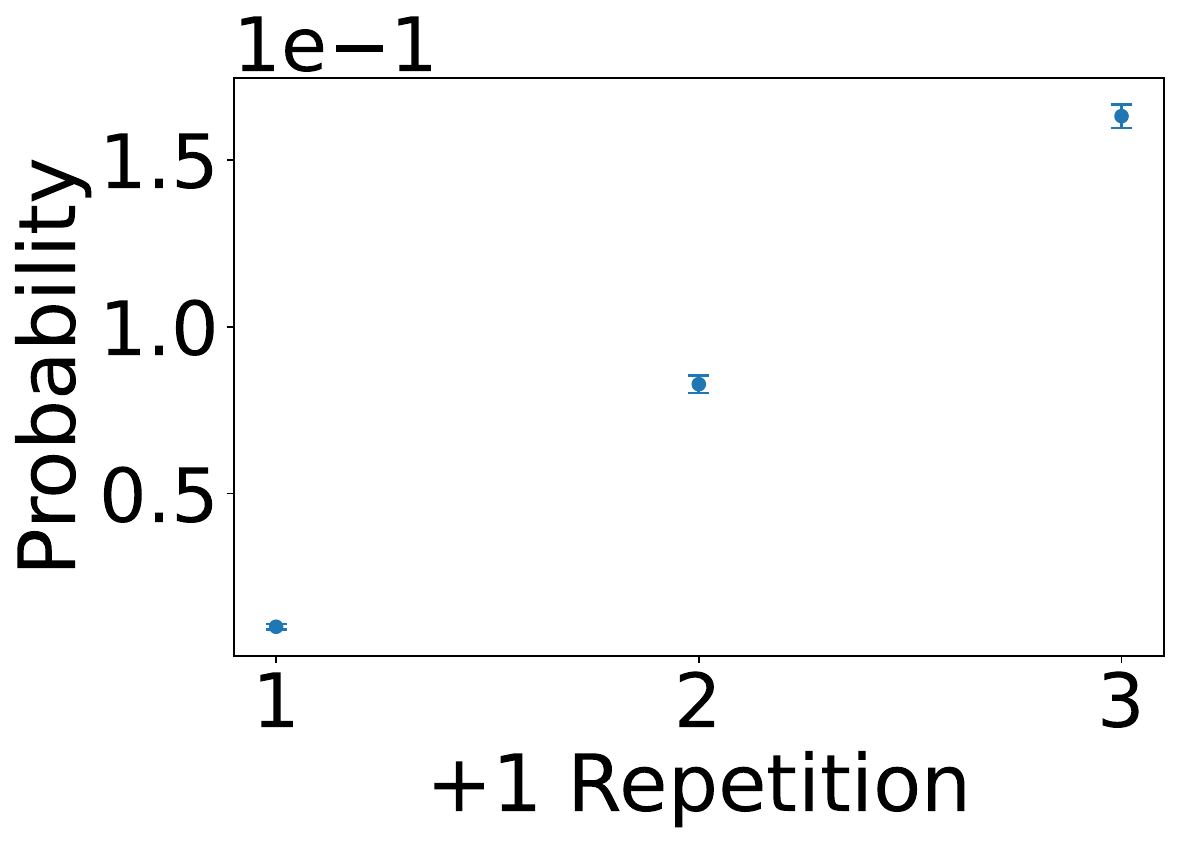} &
\FigGridCell{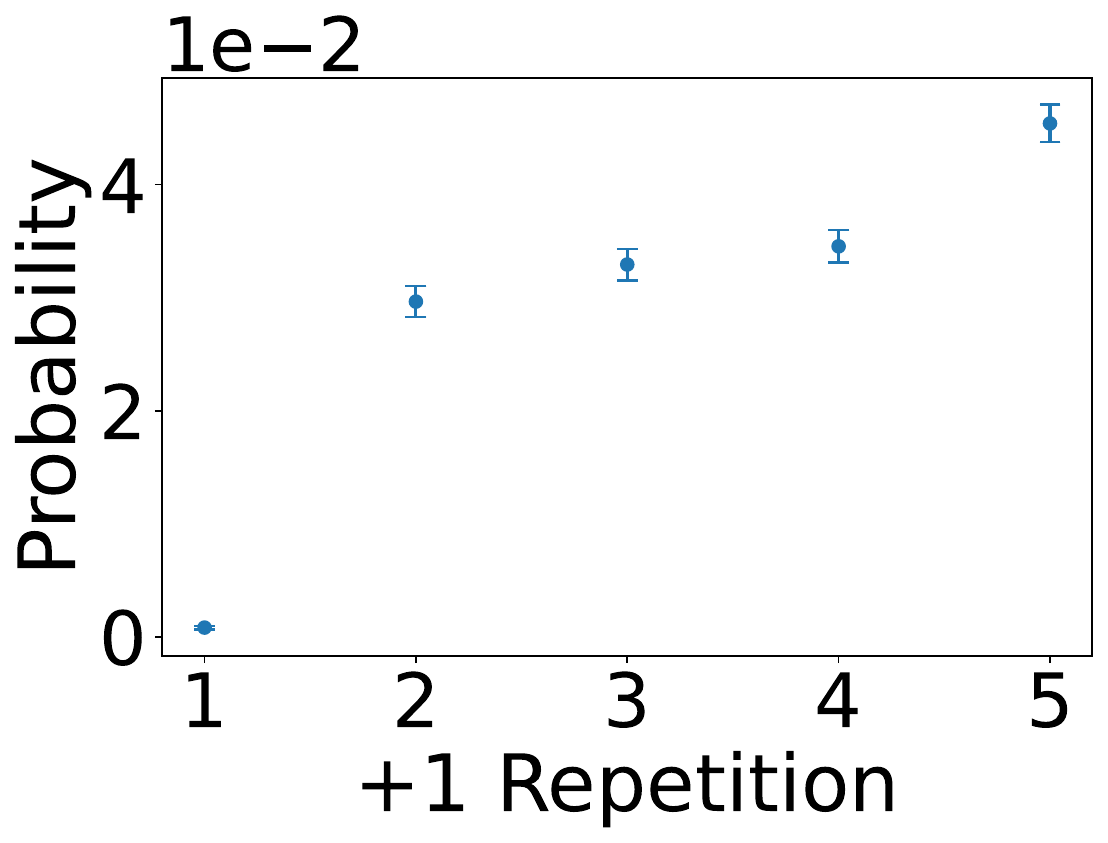} &
\FigGridCell{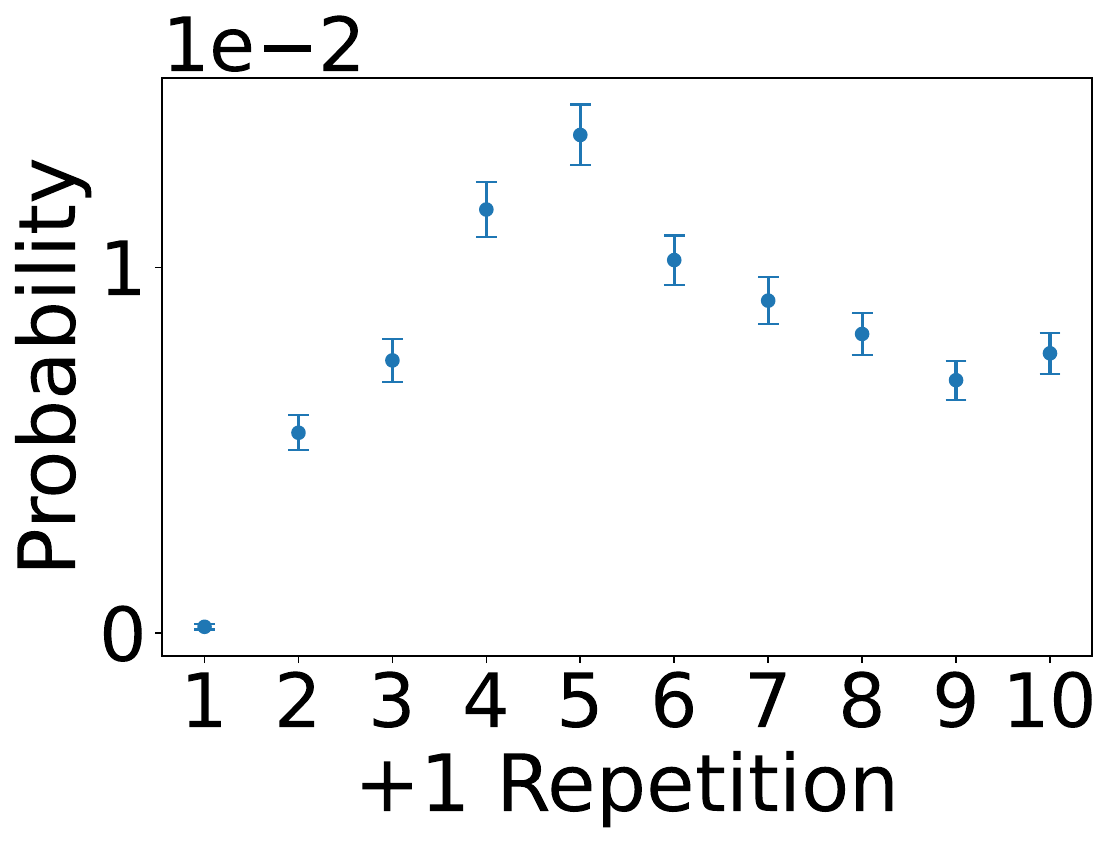} &
\FigGridCell{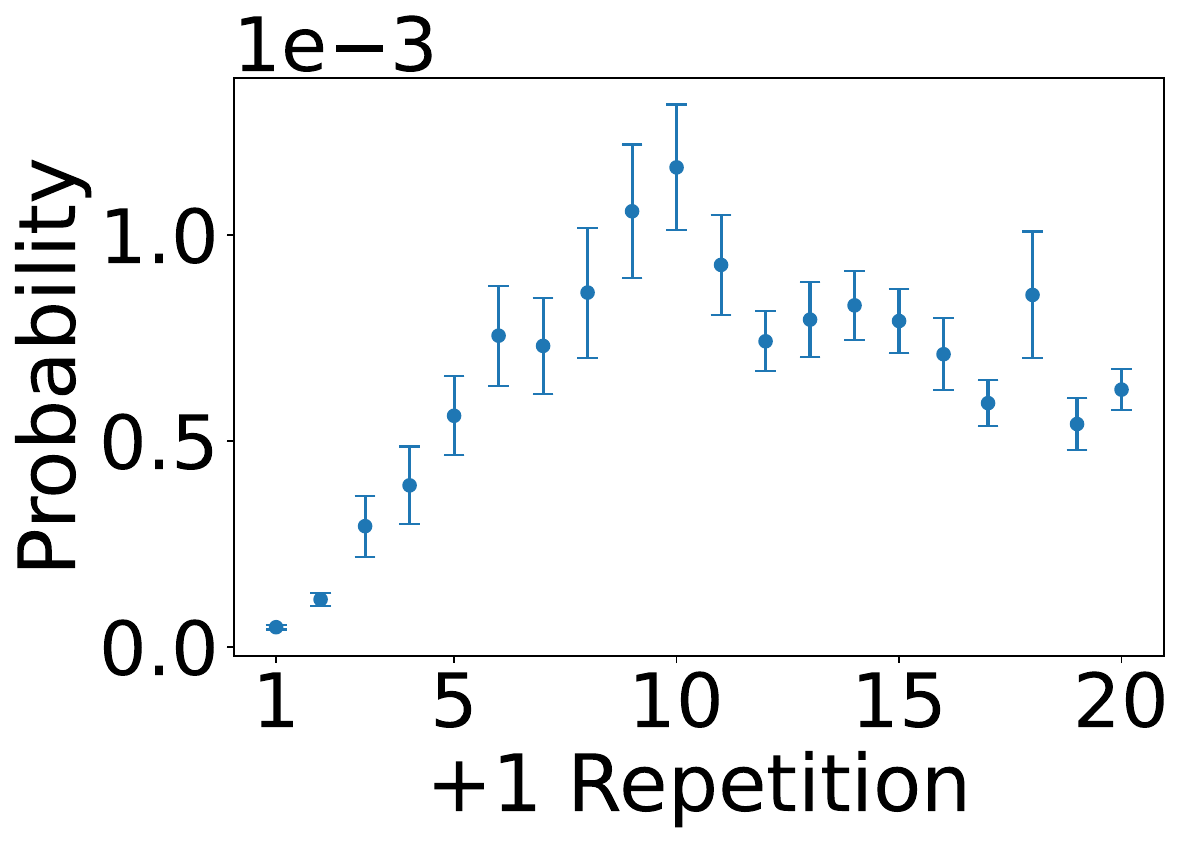} &
\FigGridCell{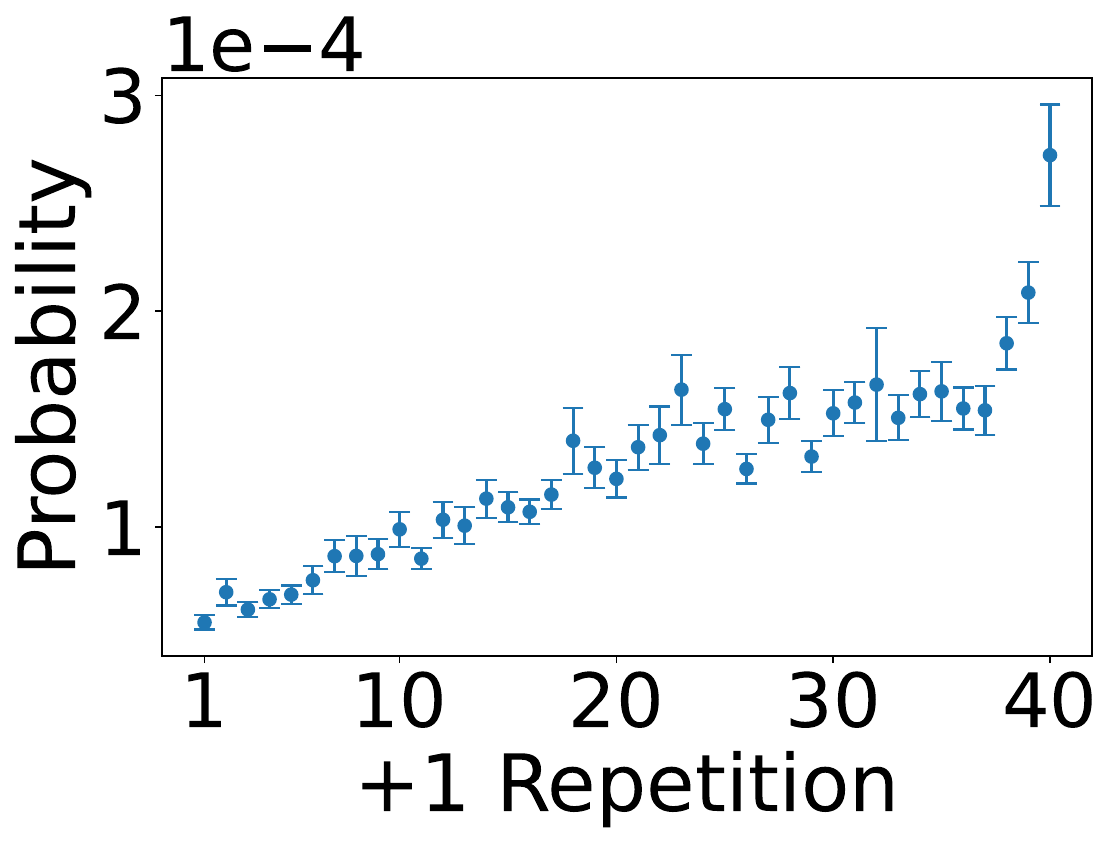} \\

\addlinespace[1pt]
\midrule
\addlinespace[1pt]

\FigGridVLabel{Mamba} &
\FigGridCell{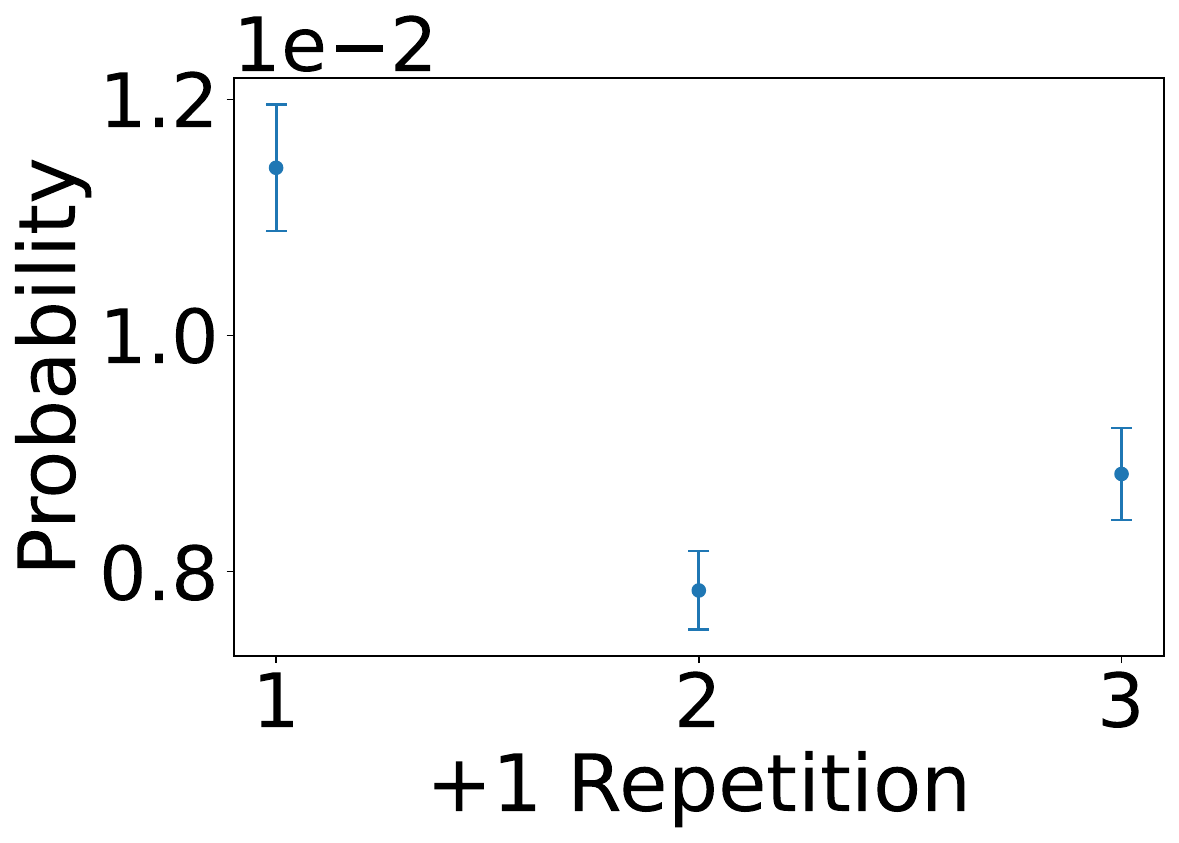} &
\FigGridCell{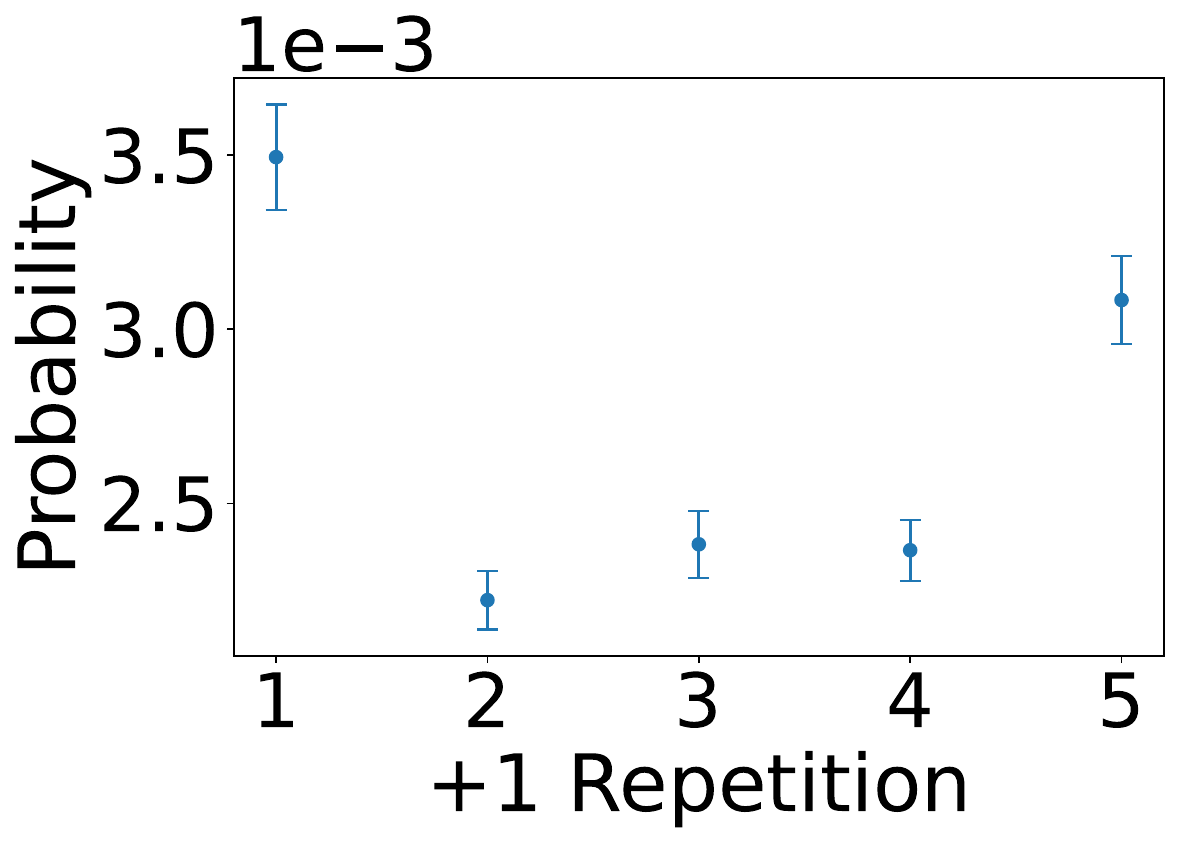} &
\FigGridCell{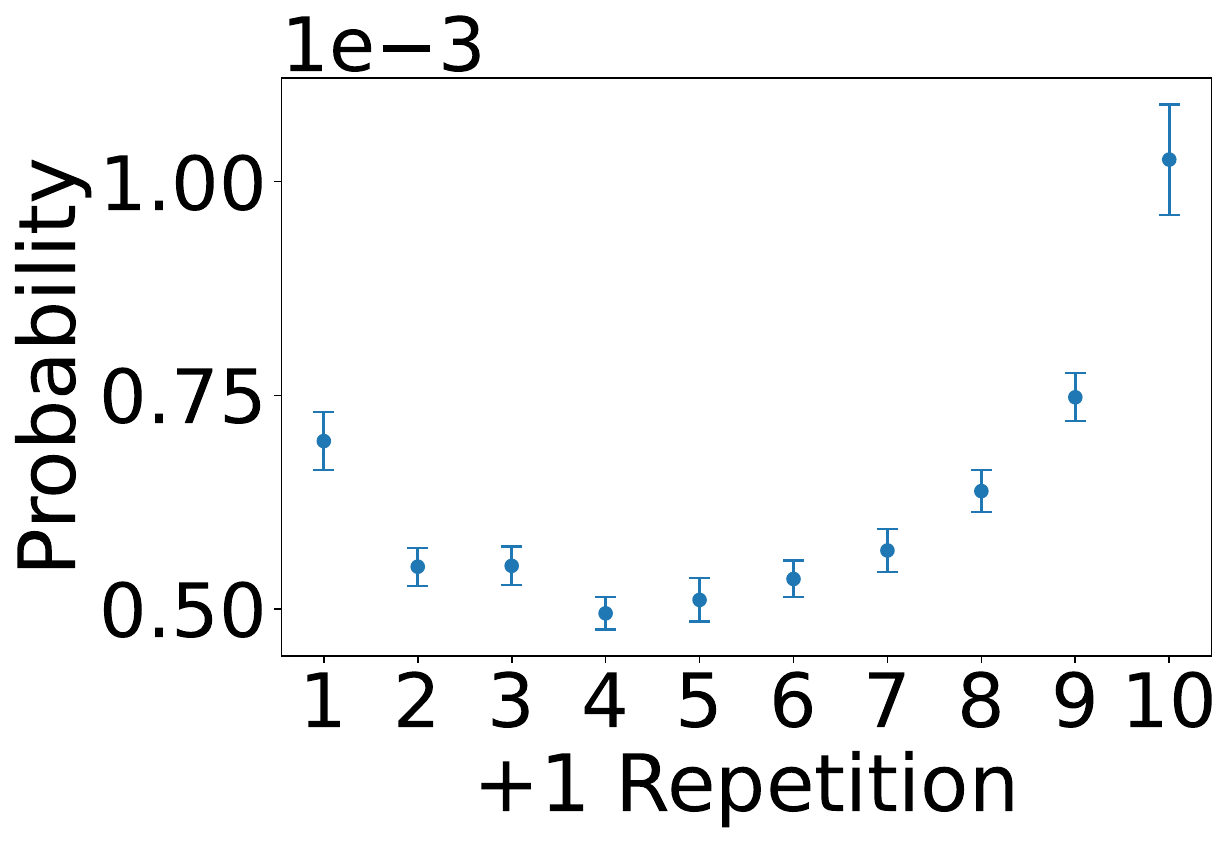} &
\FigGridCell{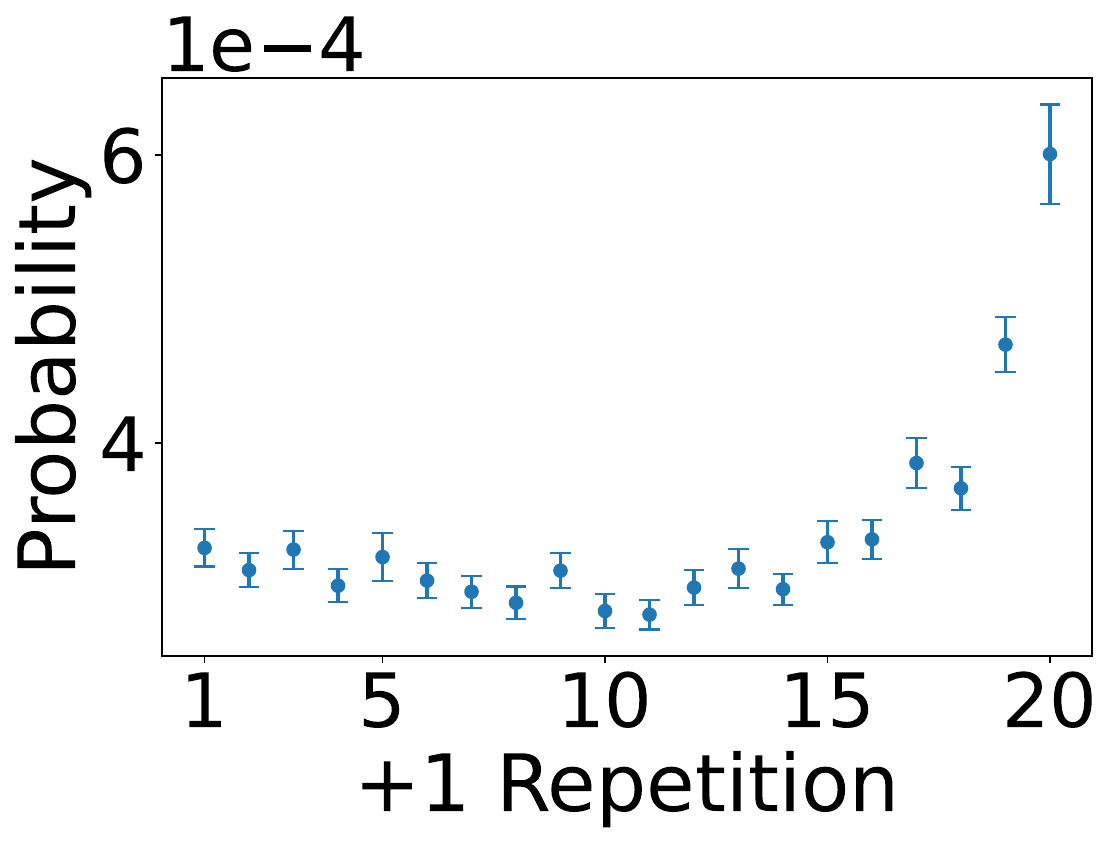} &
\FigGridCell{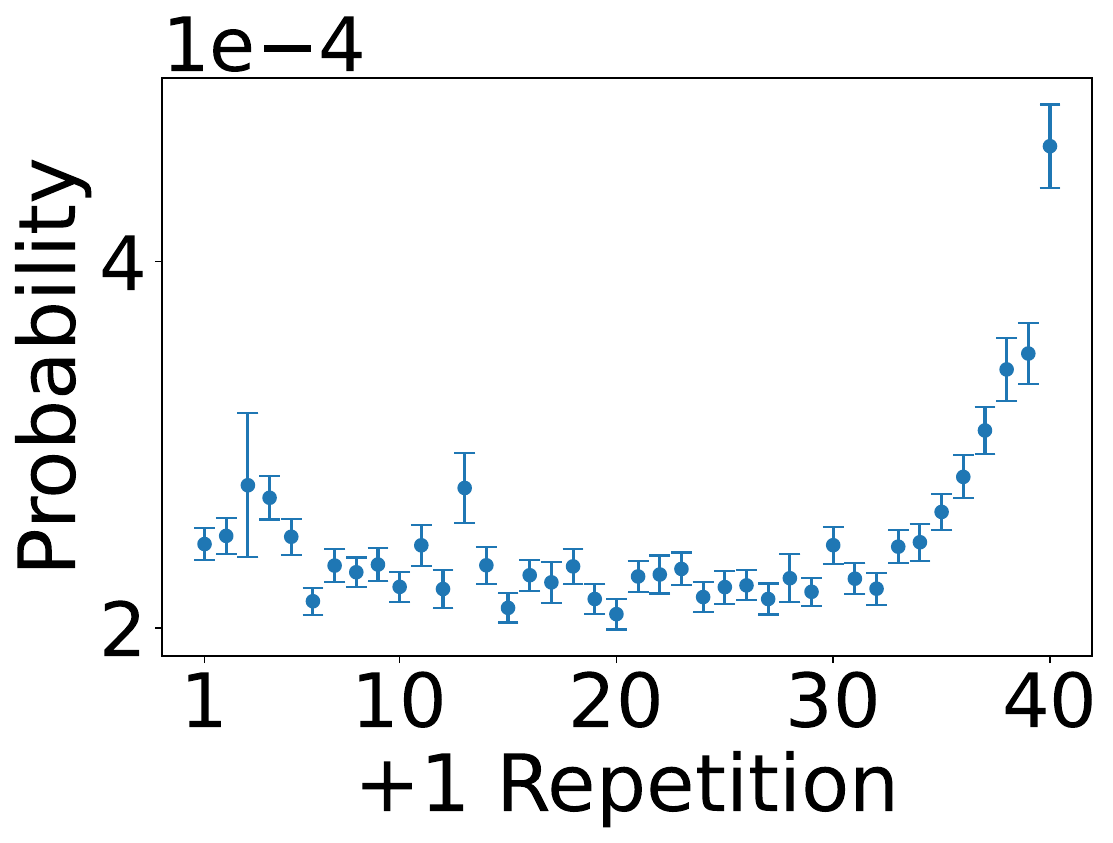} \\
\FigGridVLabel{Falcon-M} &
\FigGridCell{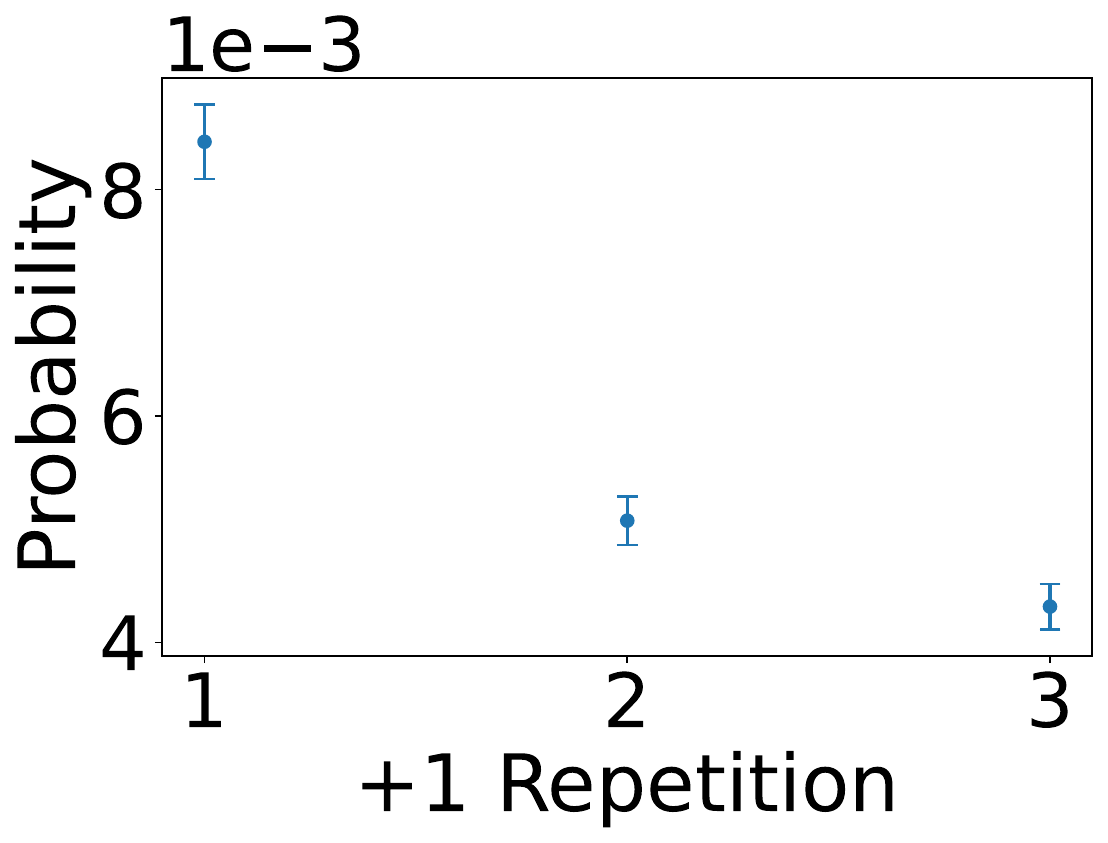} &
\FigGridCell{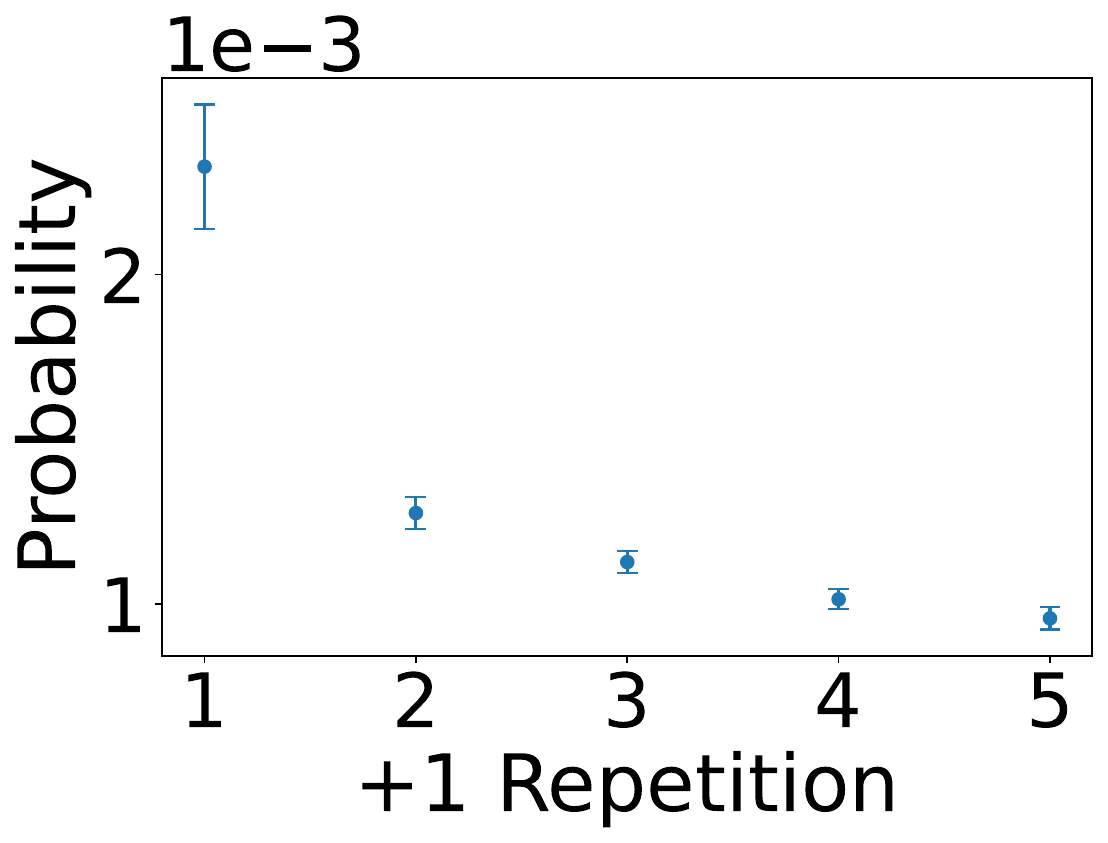} &
\FigGridCell{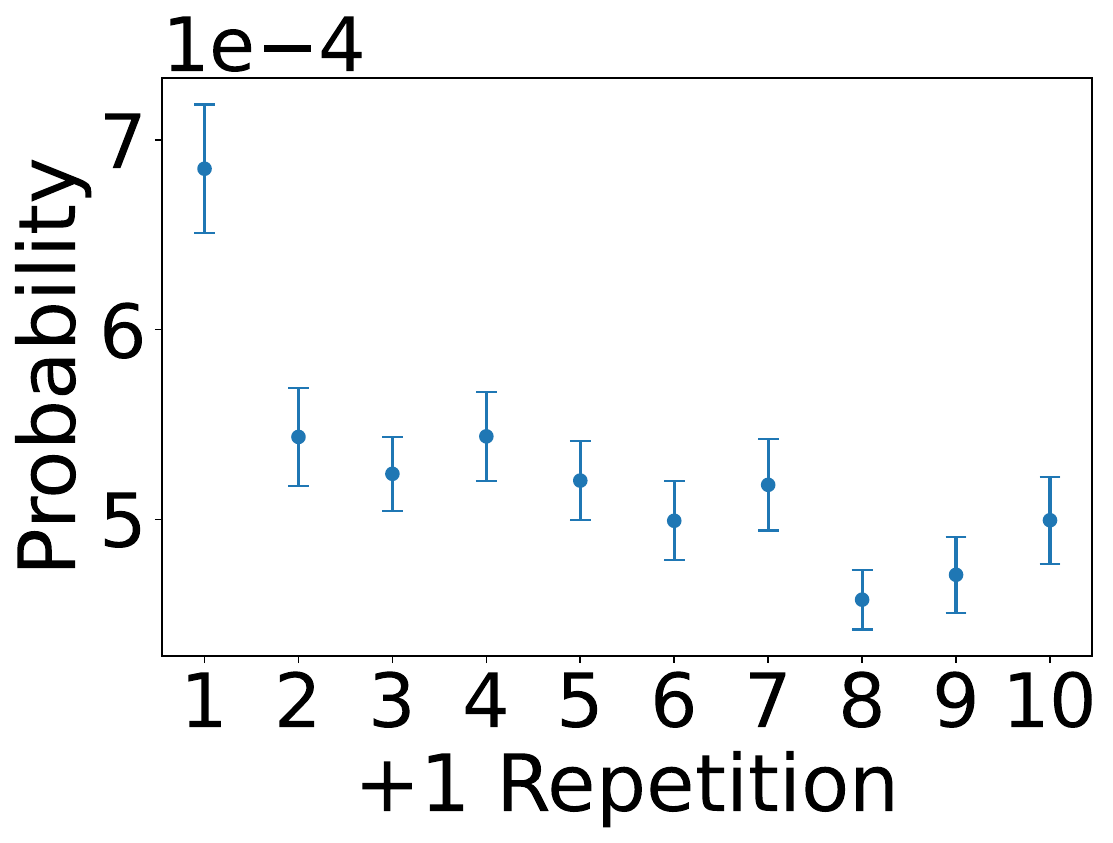} &
\FigGridCell{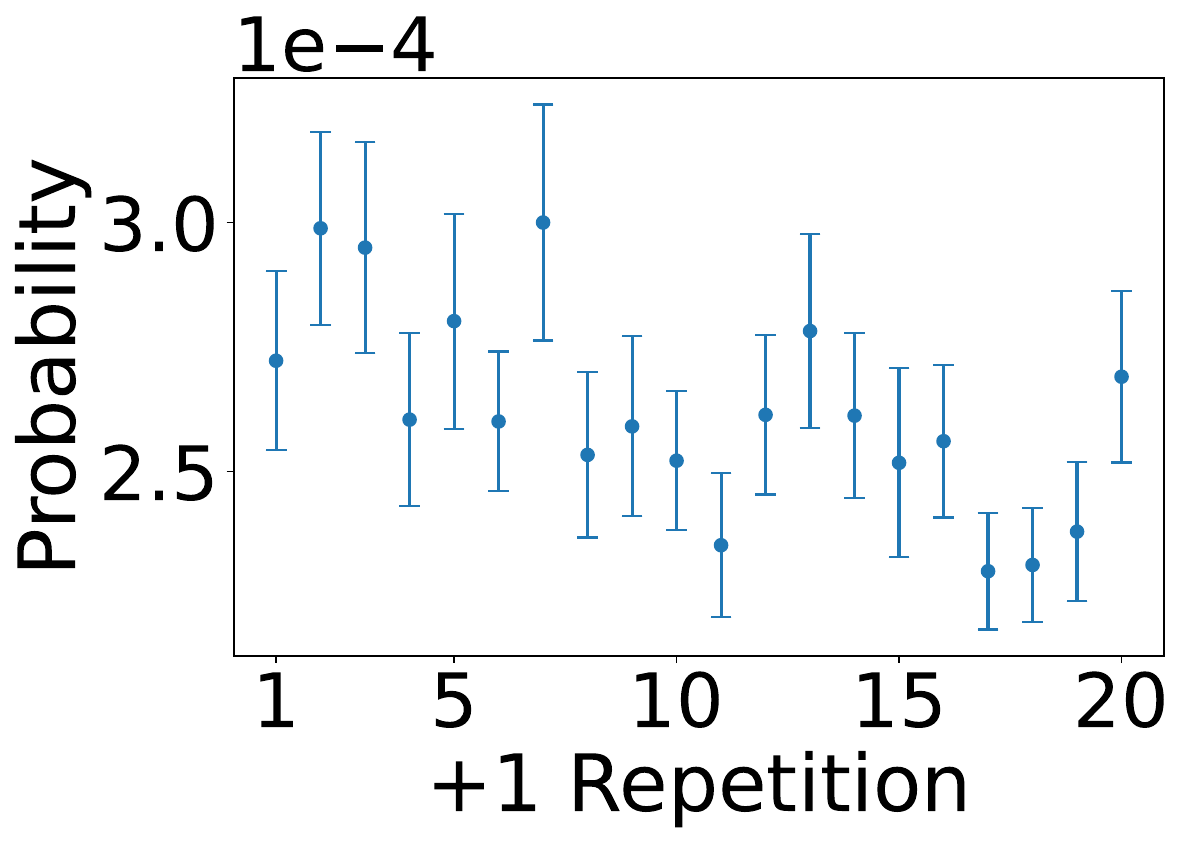} &
\FigGridCell{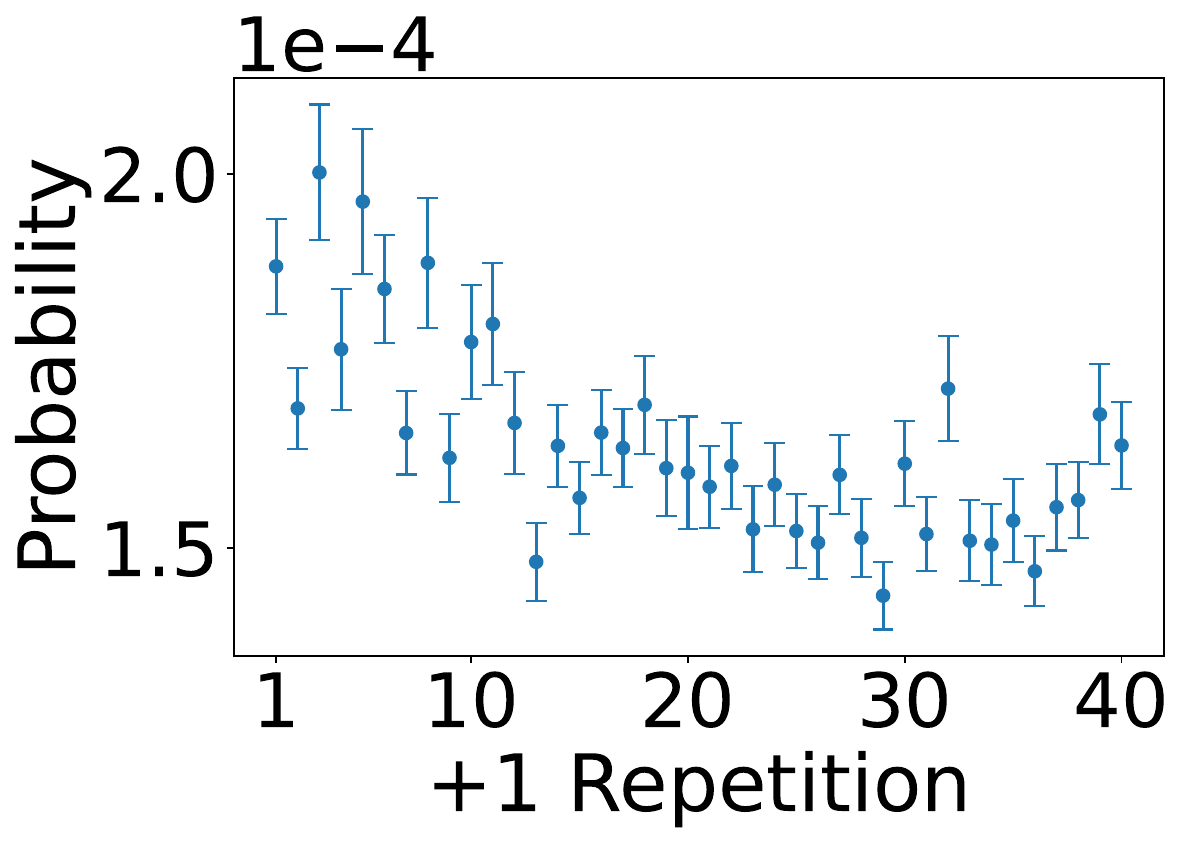} \\
\FigGridVLabel{R-Gemma} &
\FigGridCell{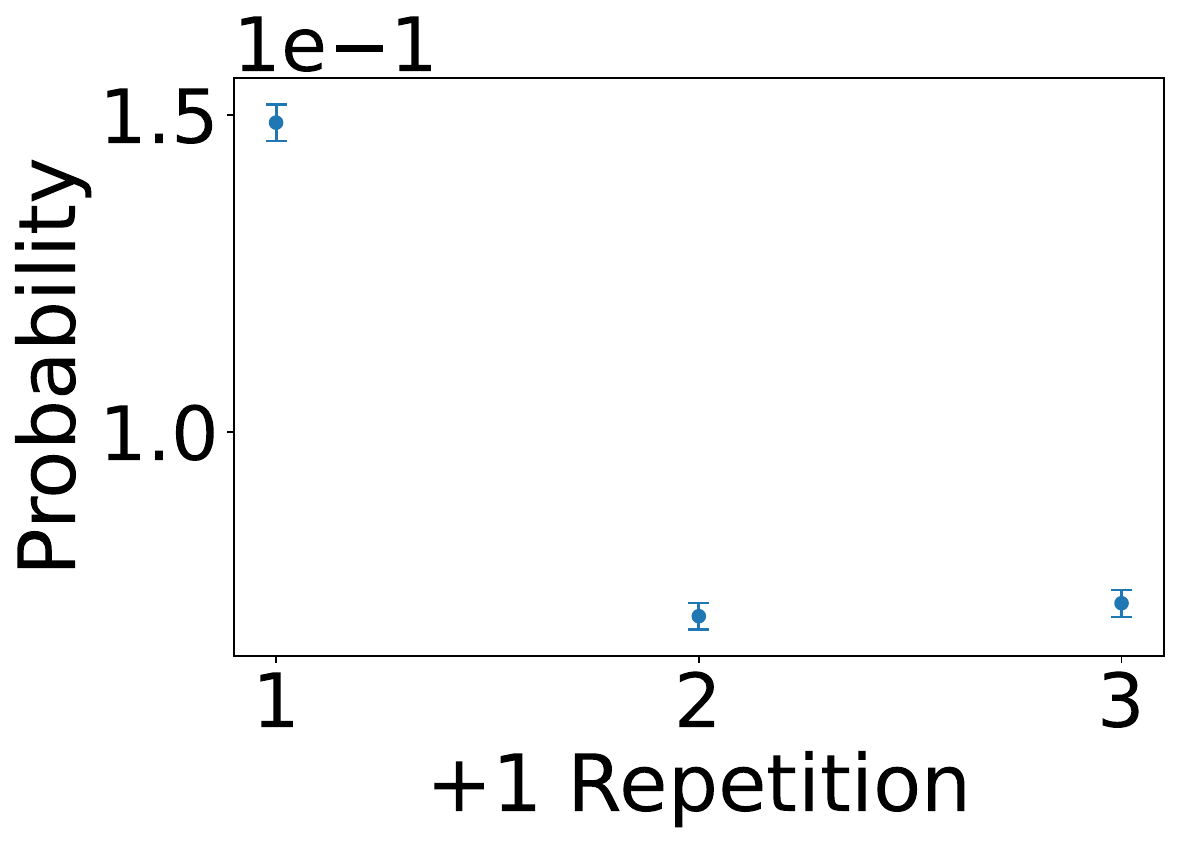} &
\FigGridCell{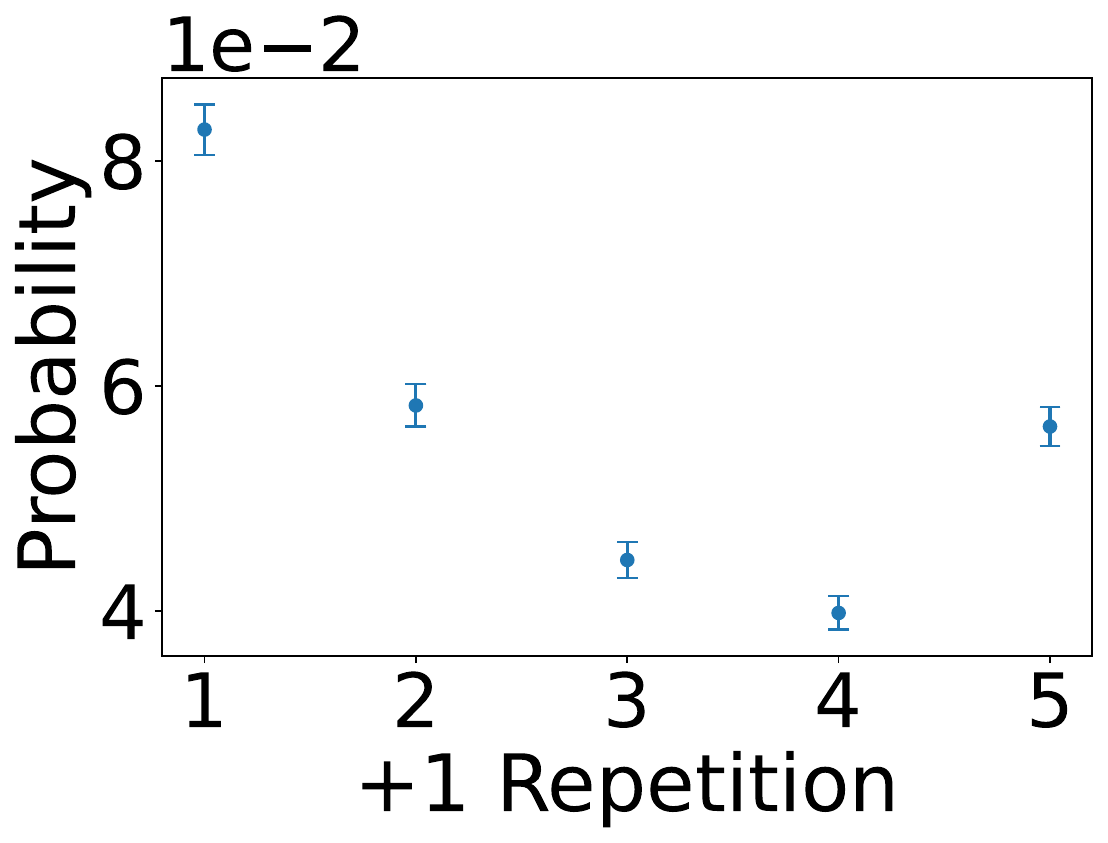} &
\FigGridCell{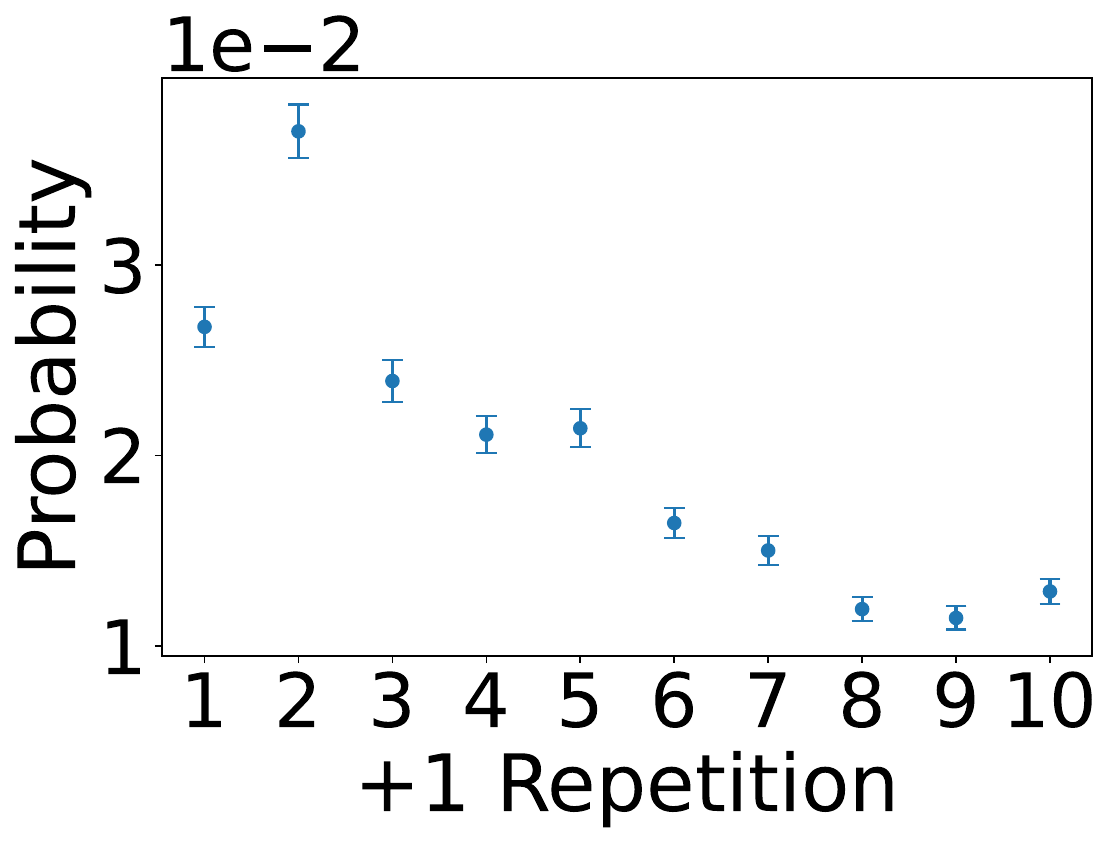} &
\FigGridCell{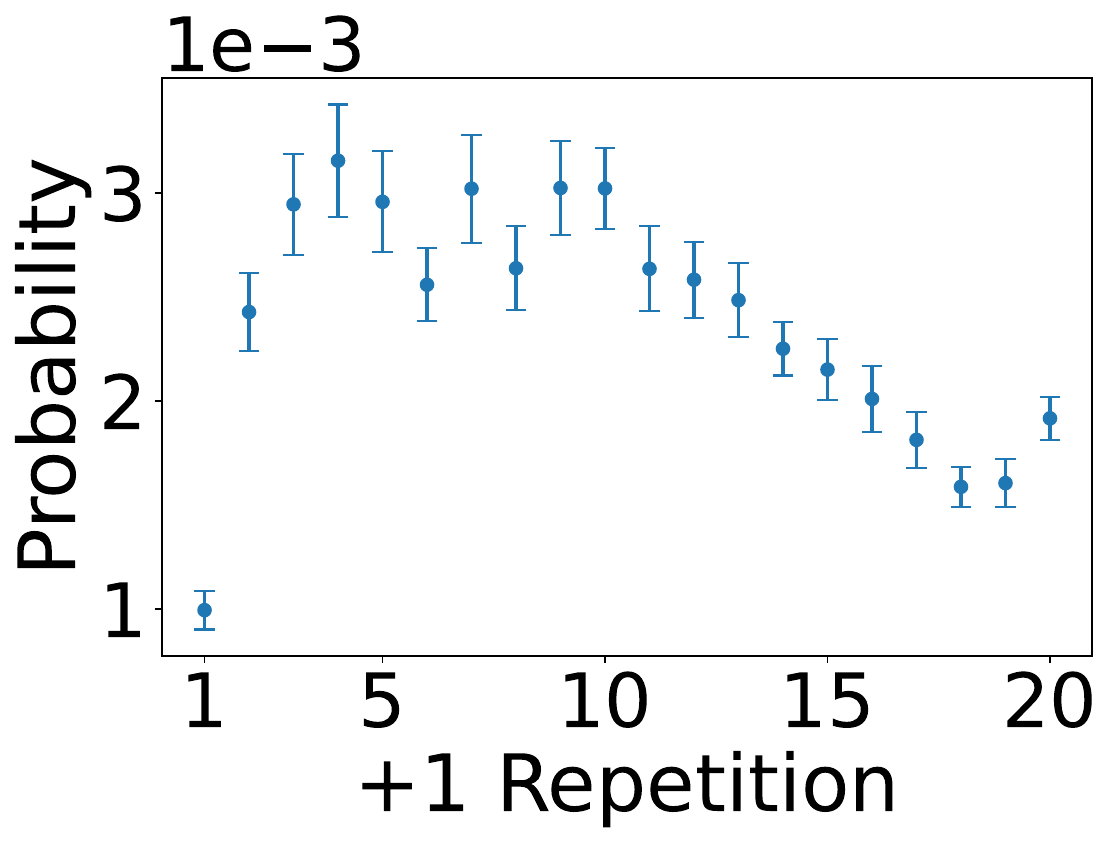} &
\FigGridCell{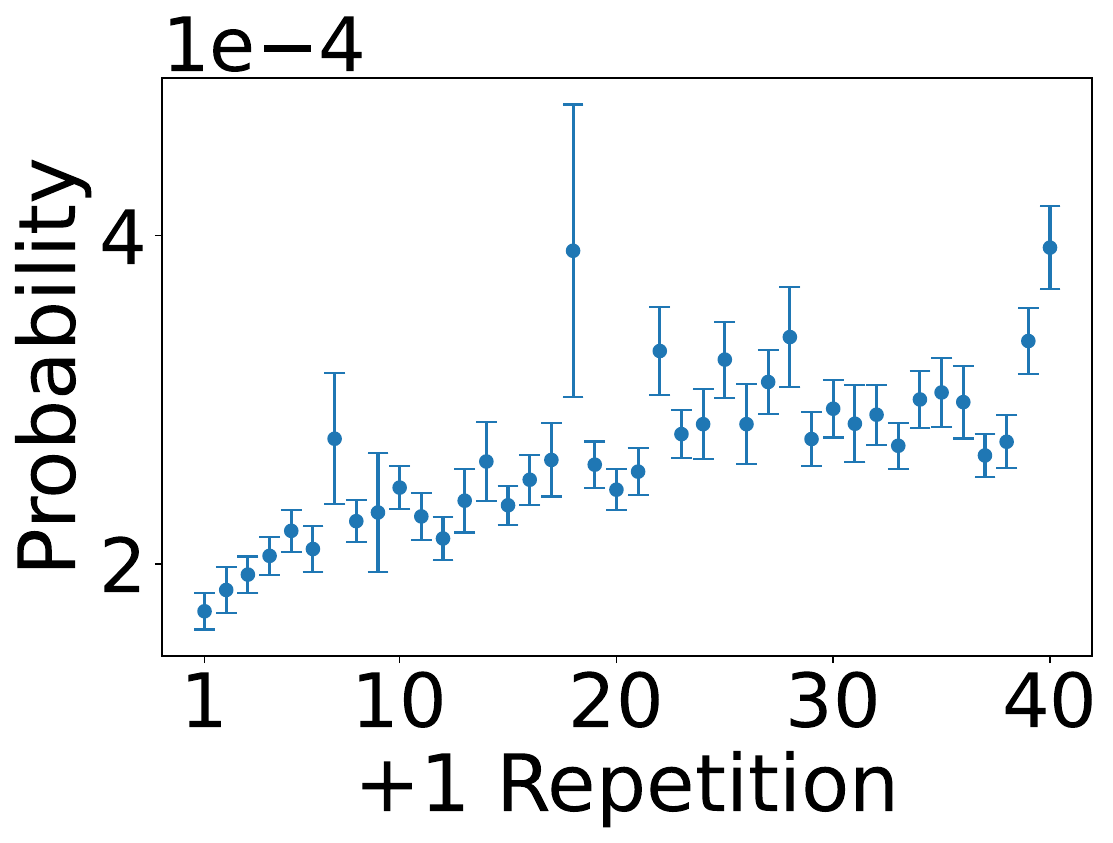} \\
\bottomrule
\end{tabular}
\endgroup

\caption{
Probability ($\pm$SEM) of \plusoneTok\ token retrieval vs. repetition position, for varying number of fixed-token repetitions (columns).
}
\label{fig:single_target_all_repetitions_plusone_tokens}
\end{figure*}

\begin{figure*}[h!]
\centering
\begingroup
\setlength{\FigGridLabelW}{0.75cm}
\setlength{\FigGridSubLabelW}{0.55cm}
\setlength{\FigGridCellH}{2.55cm} 
\setlength{\FigGridLabelSep}{1pt}
\setlength{\FigGridSubLabelSep}{1pt}
\setlength{\FigGridColSep}{0.5pt}

\FigGridSetupTwo{4}
\renewcommand{\arraystretch}{1.0}

\begin{tabular}{@{}>{\centering\arraybackslash}m{\FigGridLabelW}@{\hspace{\FigGridLabelSep}}
                >{\centering\arraybackslash}m{\FigGridSubLabelW}@{\hspace{\FigGridSubLabelSep}}
                *{4}{>{\centering\arraybackslash}m{\FigGridPlotW}}@{}}
\toprule
& & \multicolumn{4}{c}{\FigHdr Ablations (\%)} \\
\cmidrule(lr){3-6}
& & \FigHdr 0 & \FigHdr 1 & \FigHdr 5 & \FigHdr 10 \\
\midrule
\multirow{2}{*}{\FigGridVLabelSpanTwo{Mamba}} &
\FigGridVSubLabel{Ind} &
\FigGridCell{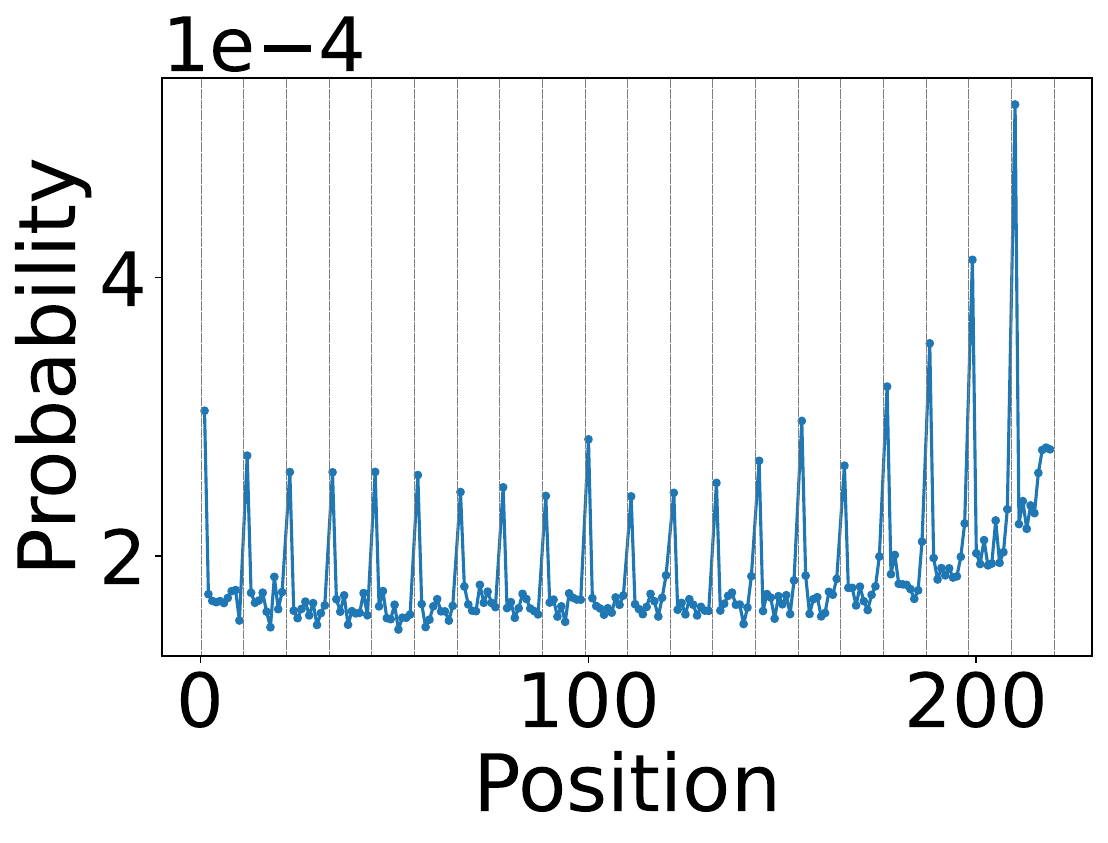} &
\FigGridCell{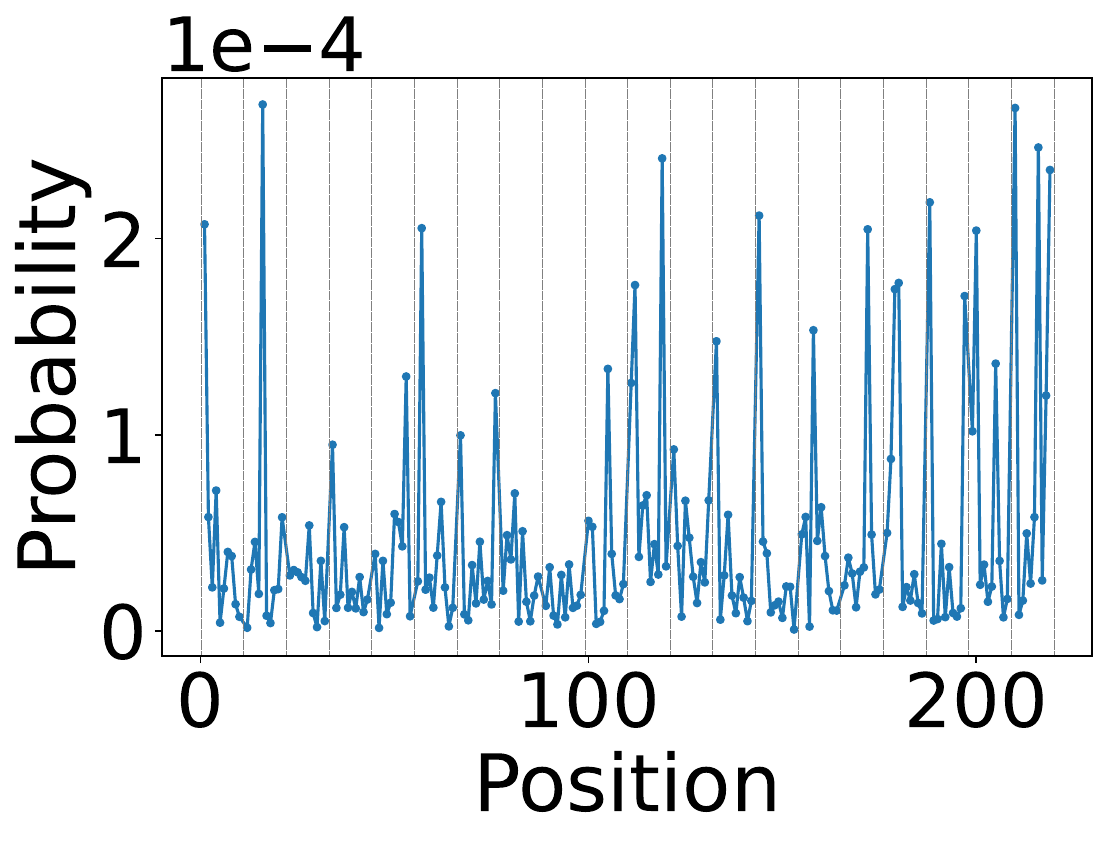} &
\FigGridCell{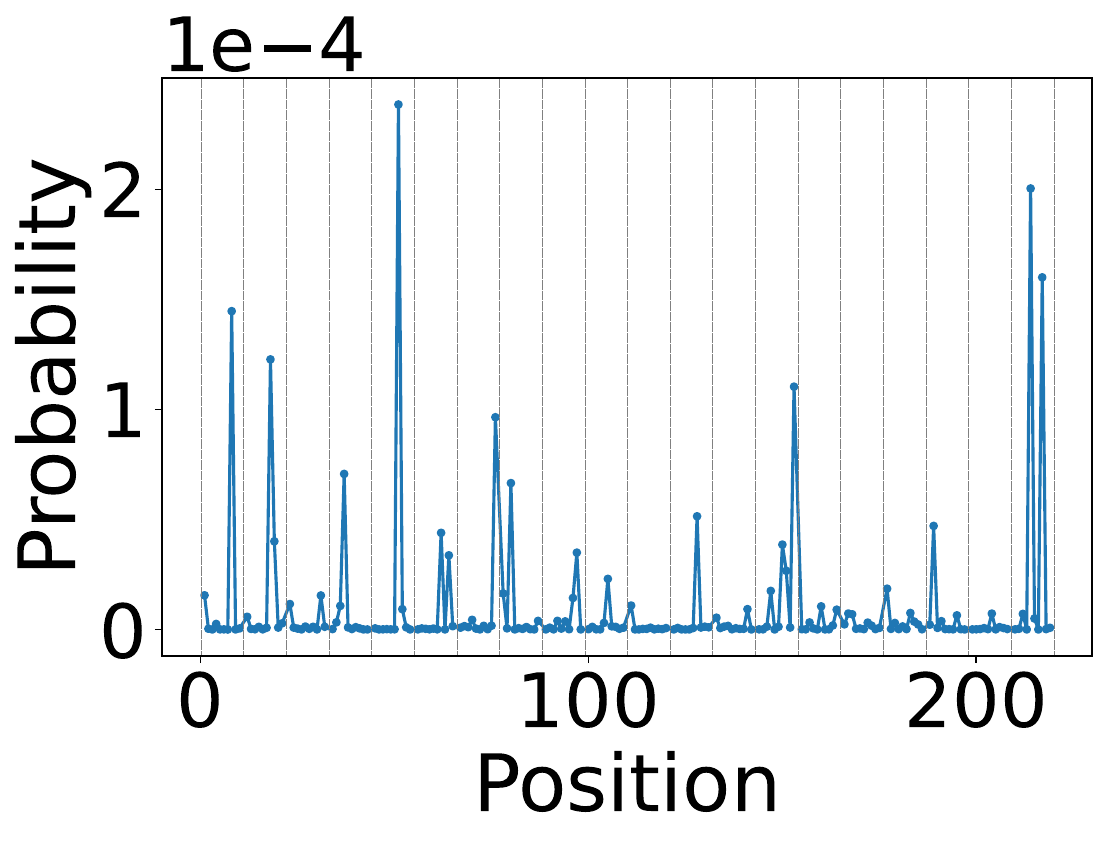} &
\FigGridCell{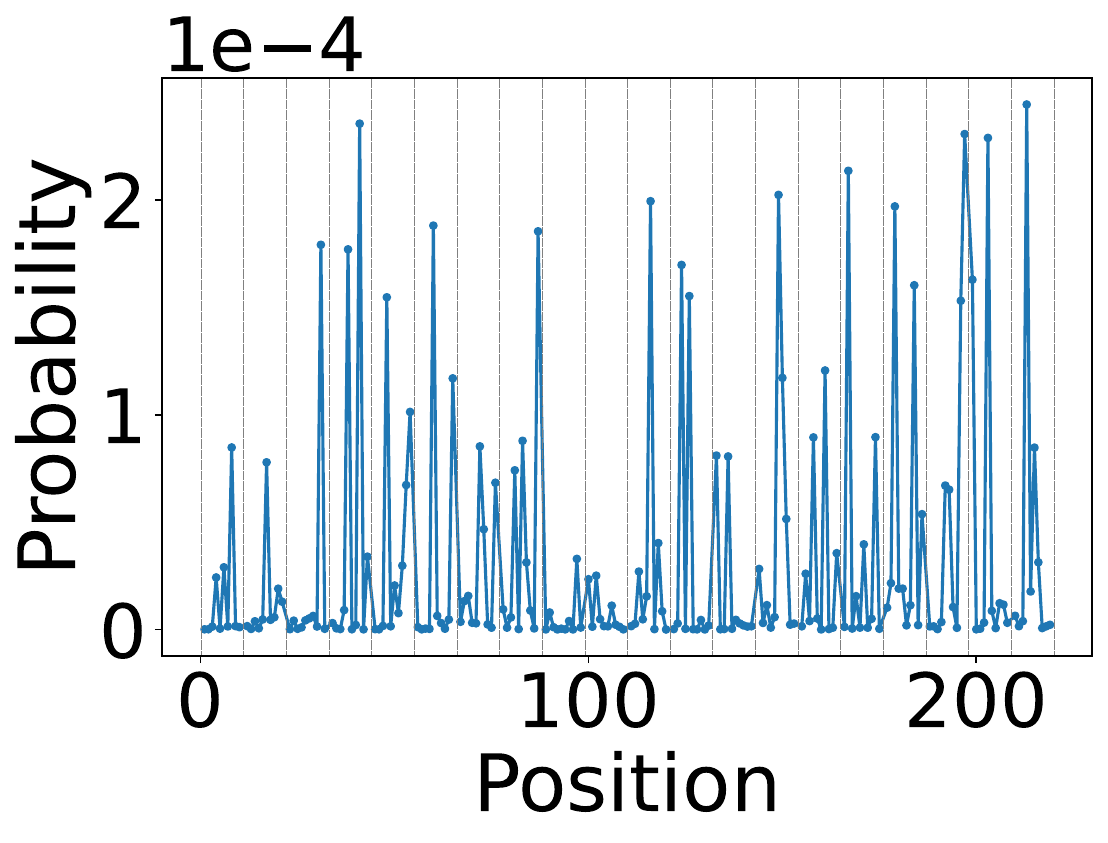} \\
& \FigGridVSubLabel{Rand} &
\FigGridCell{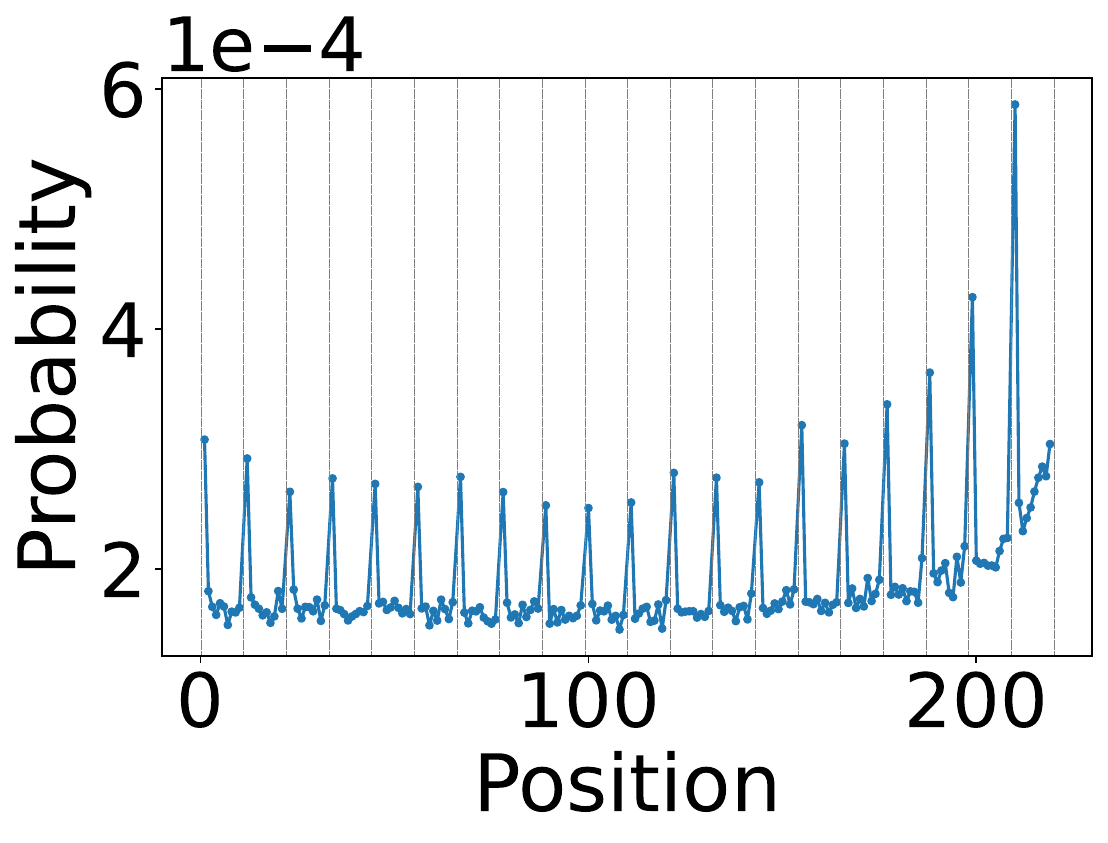} &
\FigGridCell{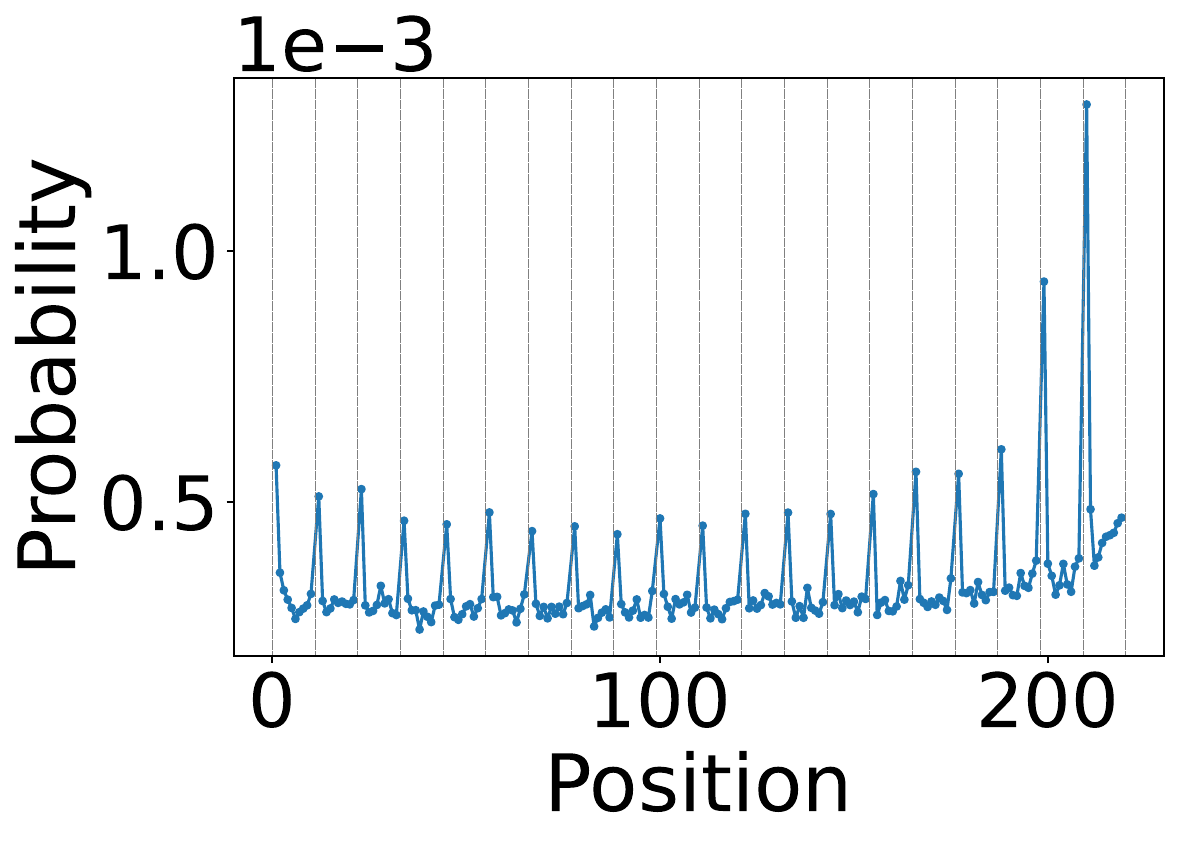} &
\FigGridCell{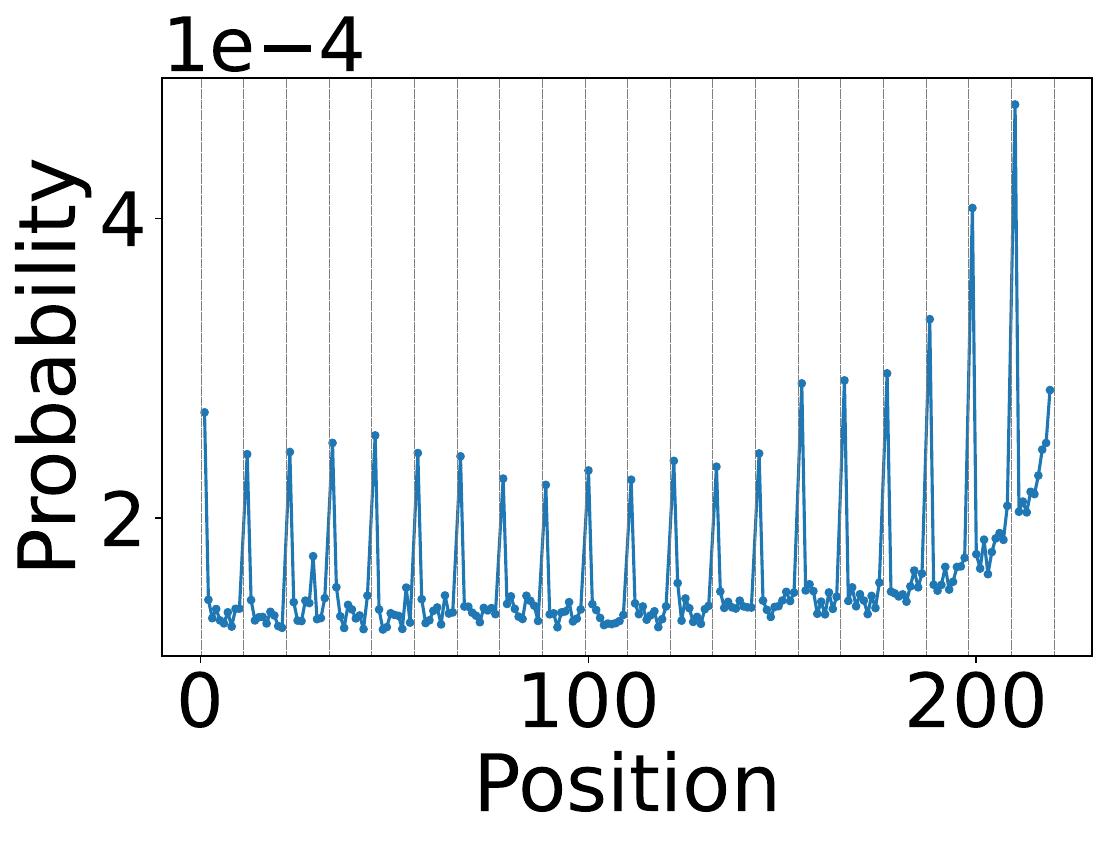} &
\FigGridCell{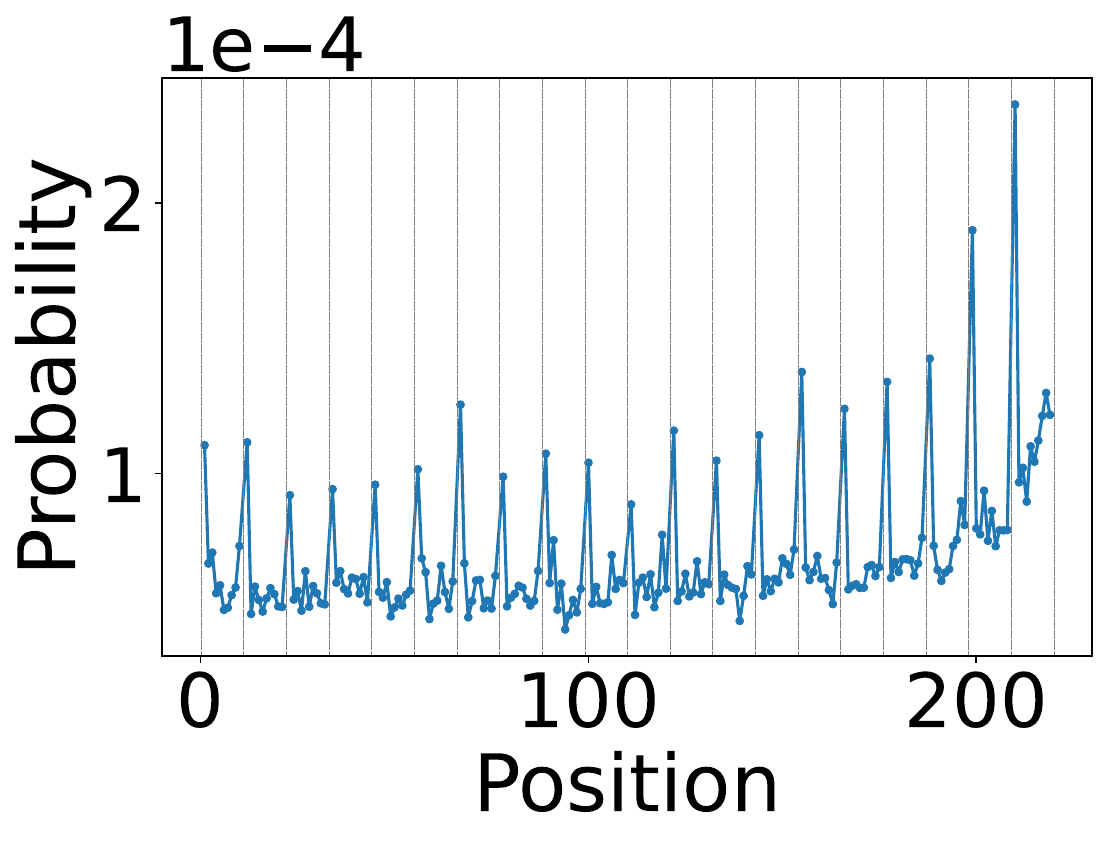} \\

\multirow{2}{*}{\FigGridVLabelSpanTwo{Falcon-M}} &
\FigGridVSubLabel{Ind} &
\FigGridCell{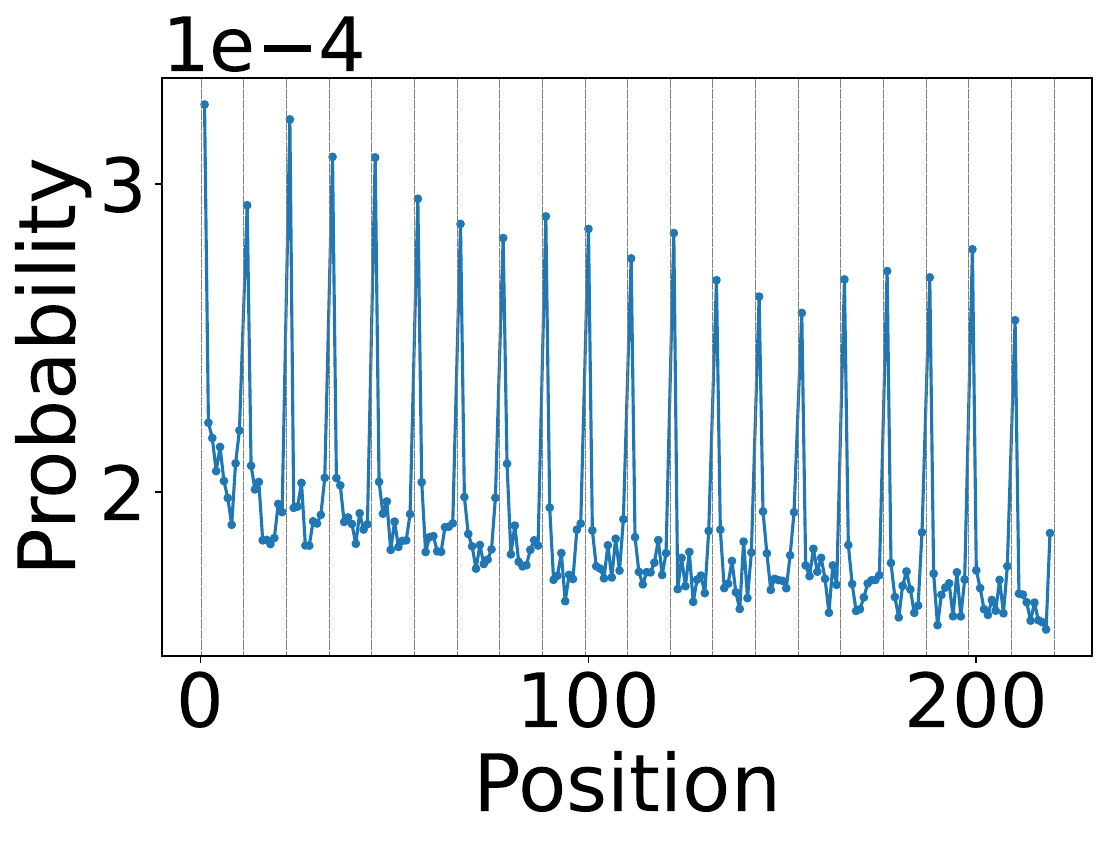} &
\FigGridCell{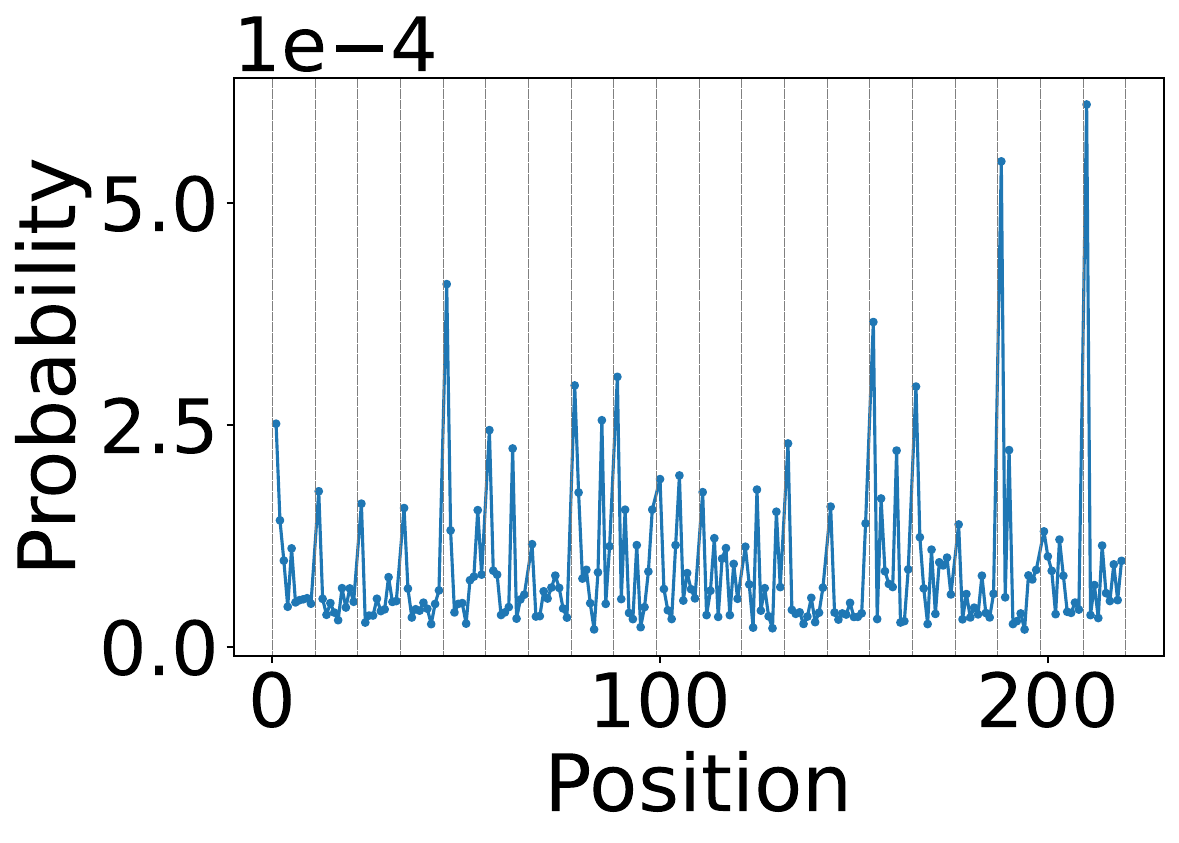} &
\FigGridCell{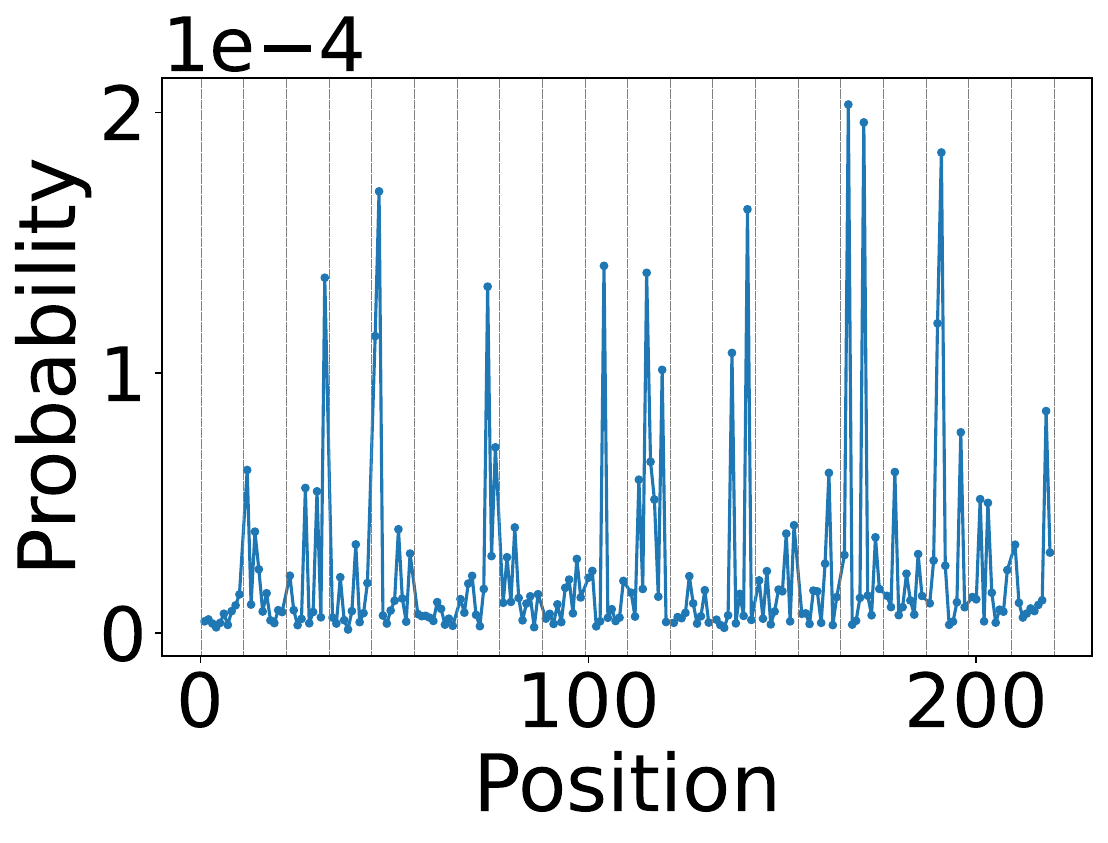} &
\FigGridCell{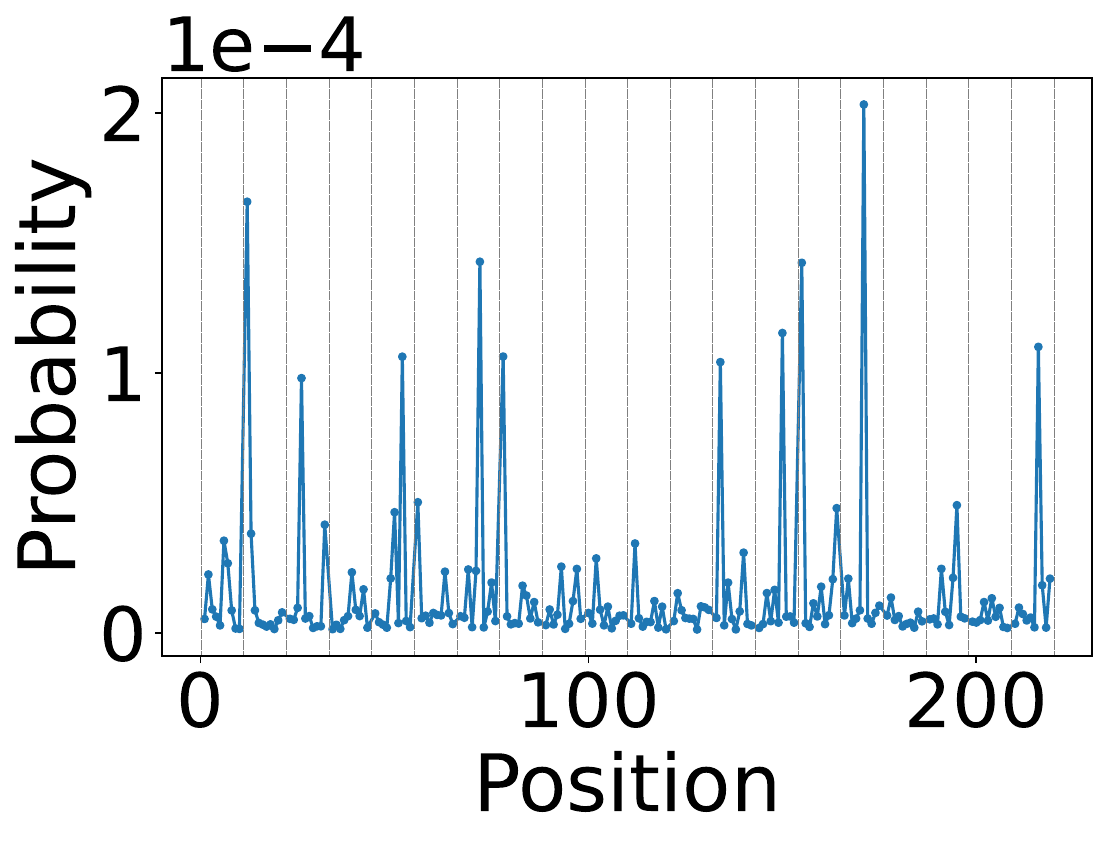} \\
& \FigGridVSubLabel{Rand} &
\FigGridCell{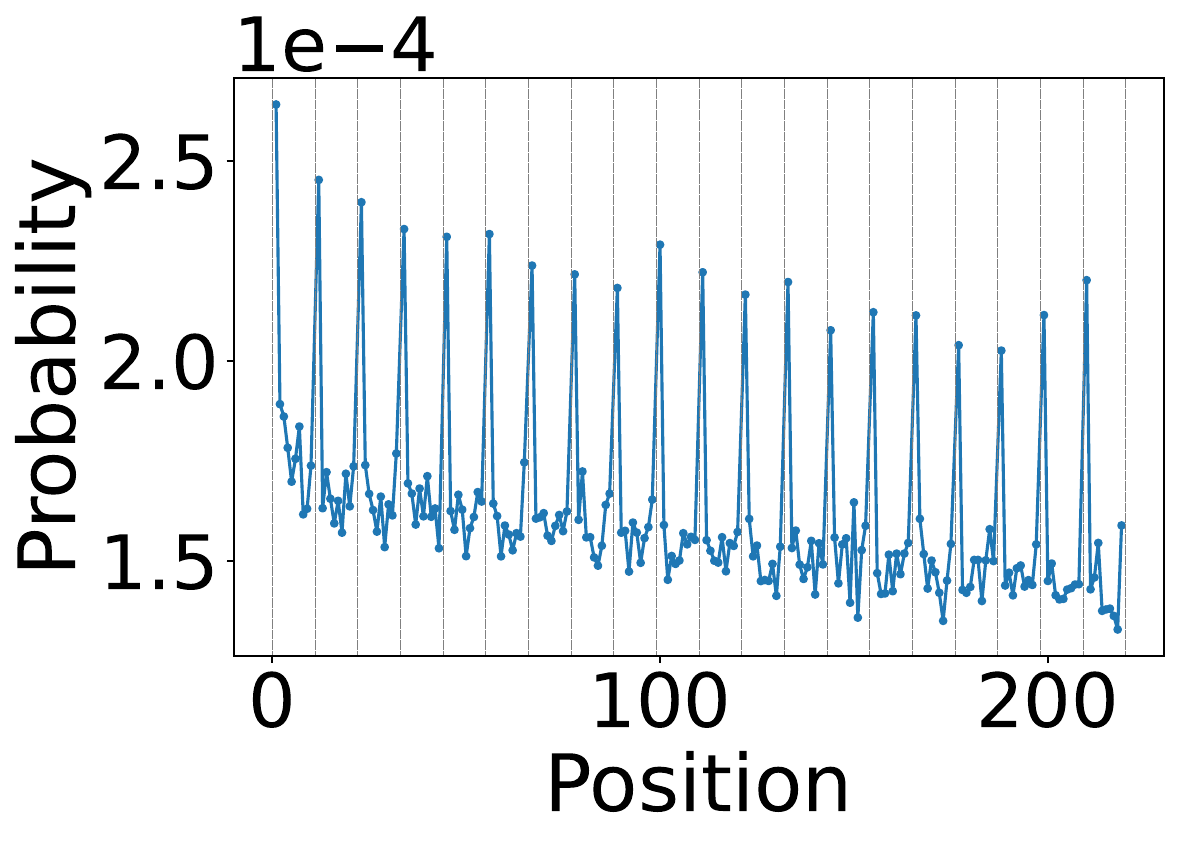} &
\FigGridCell{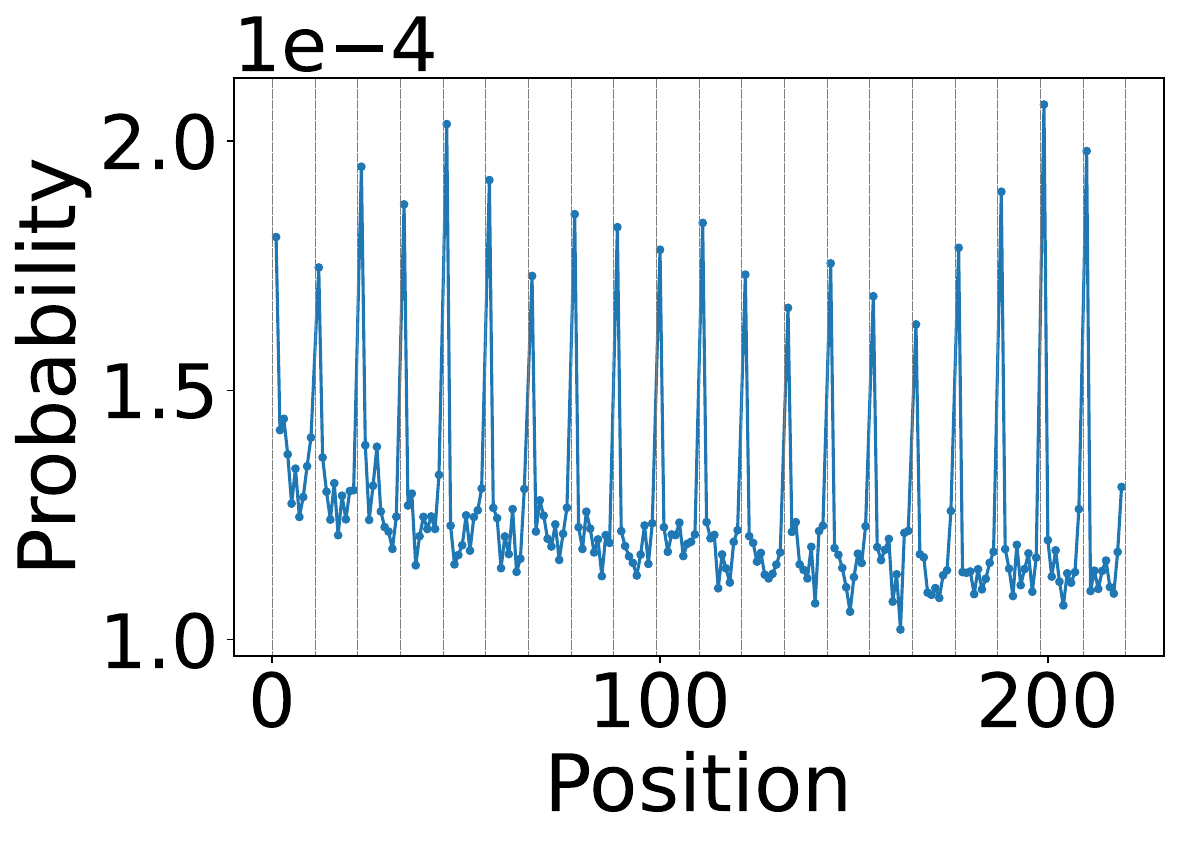} &
\FigGridCell{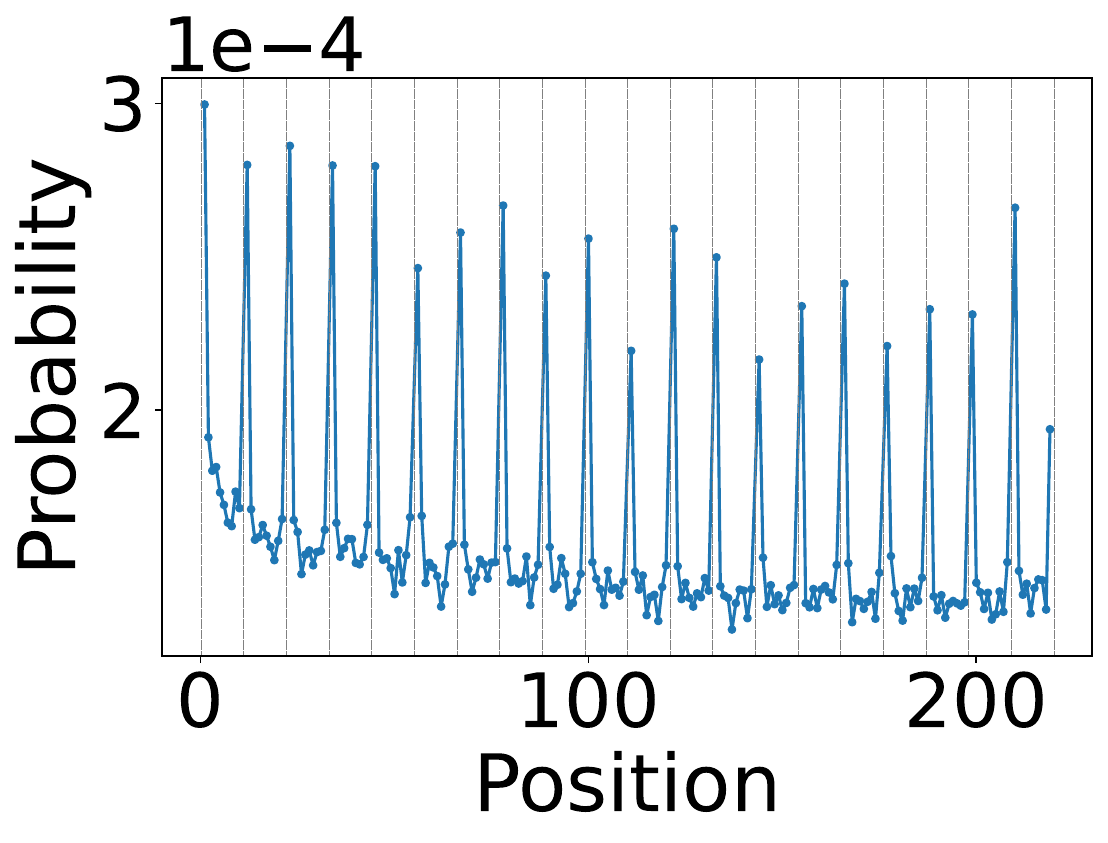} &
\FigGridCell{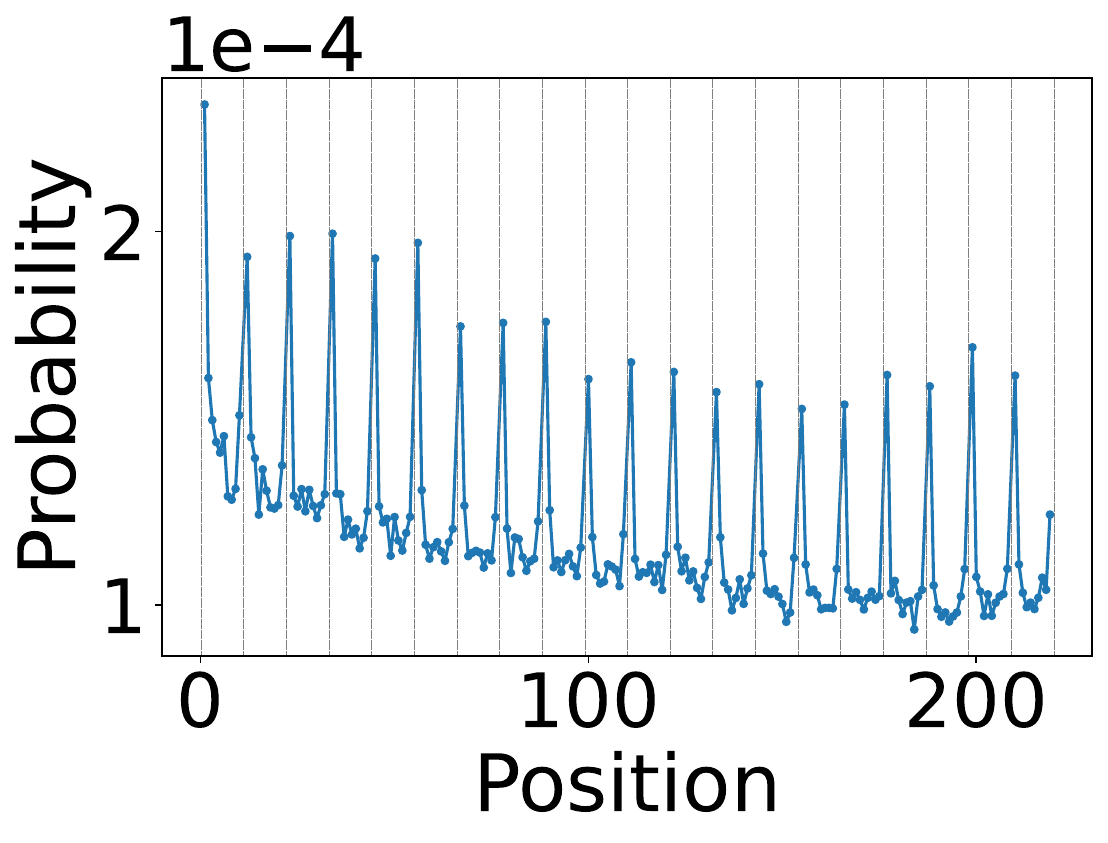} \\

\multirow{2}{*}{\FigGridVLabelSpanTwo{R-Gemma}} &
\FigGridVSubLabel{Ind} &
\FigGridCell{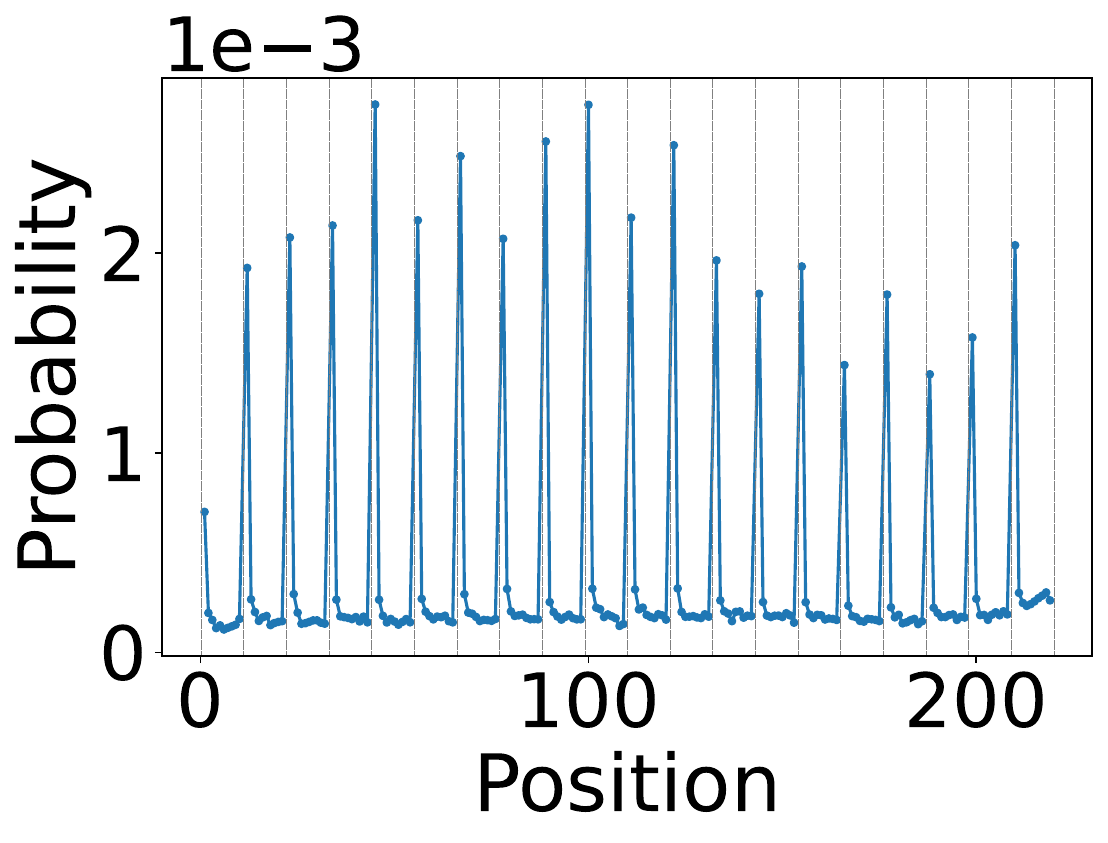} &
\FigGridCell{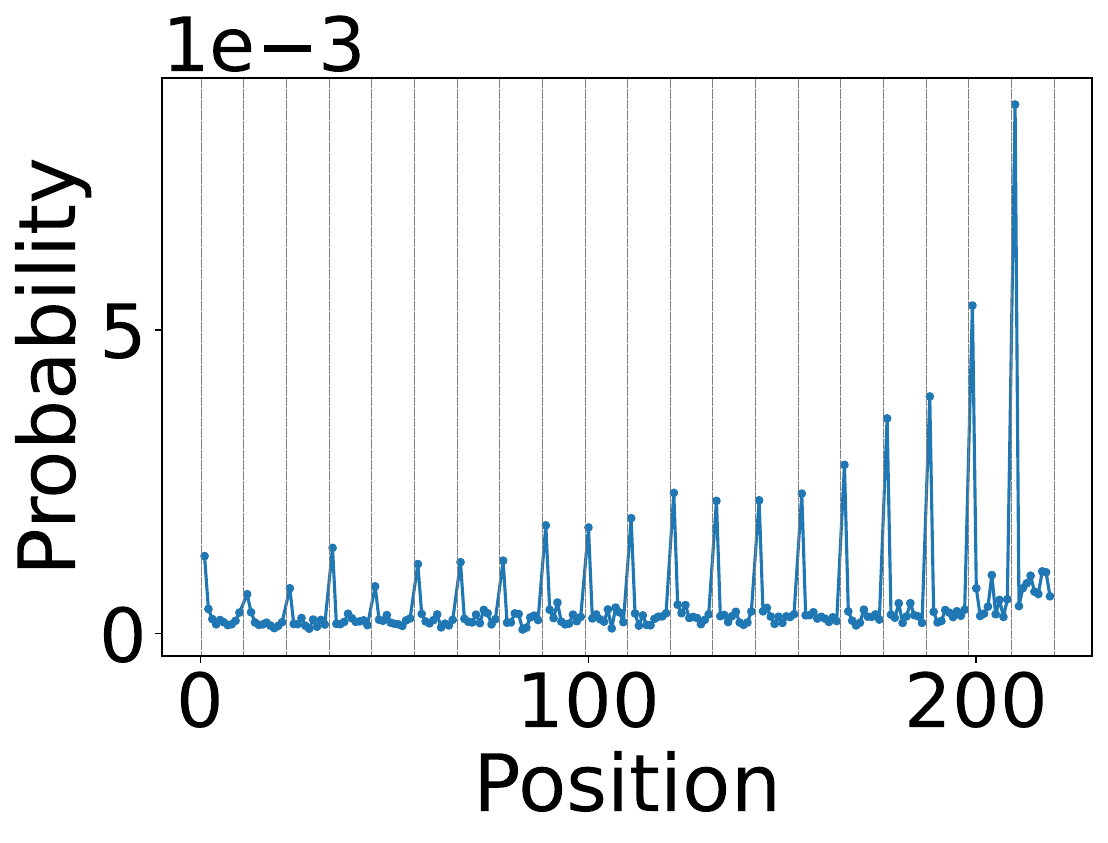} &
\FigGridCell{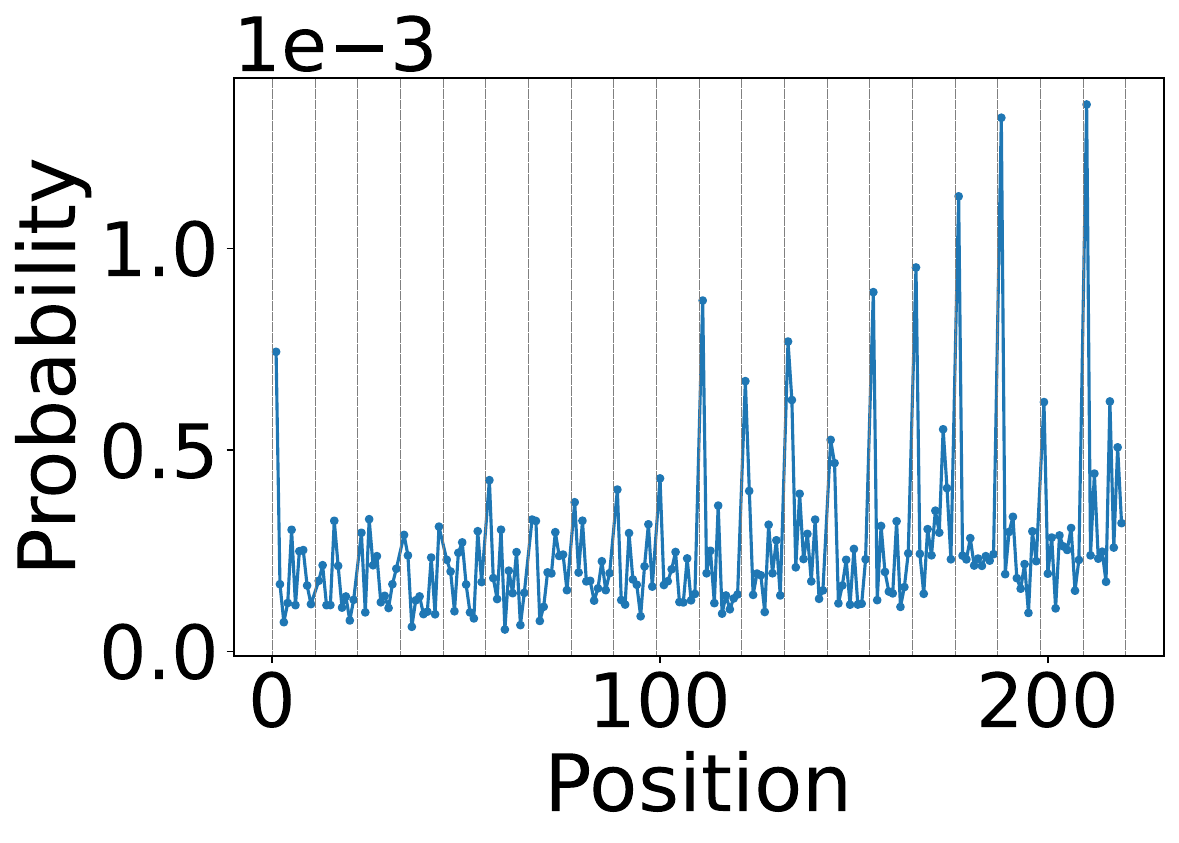} &
\FigGridCell{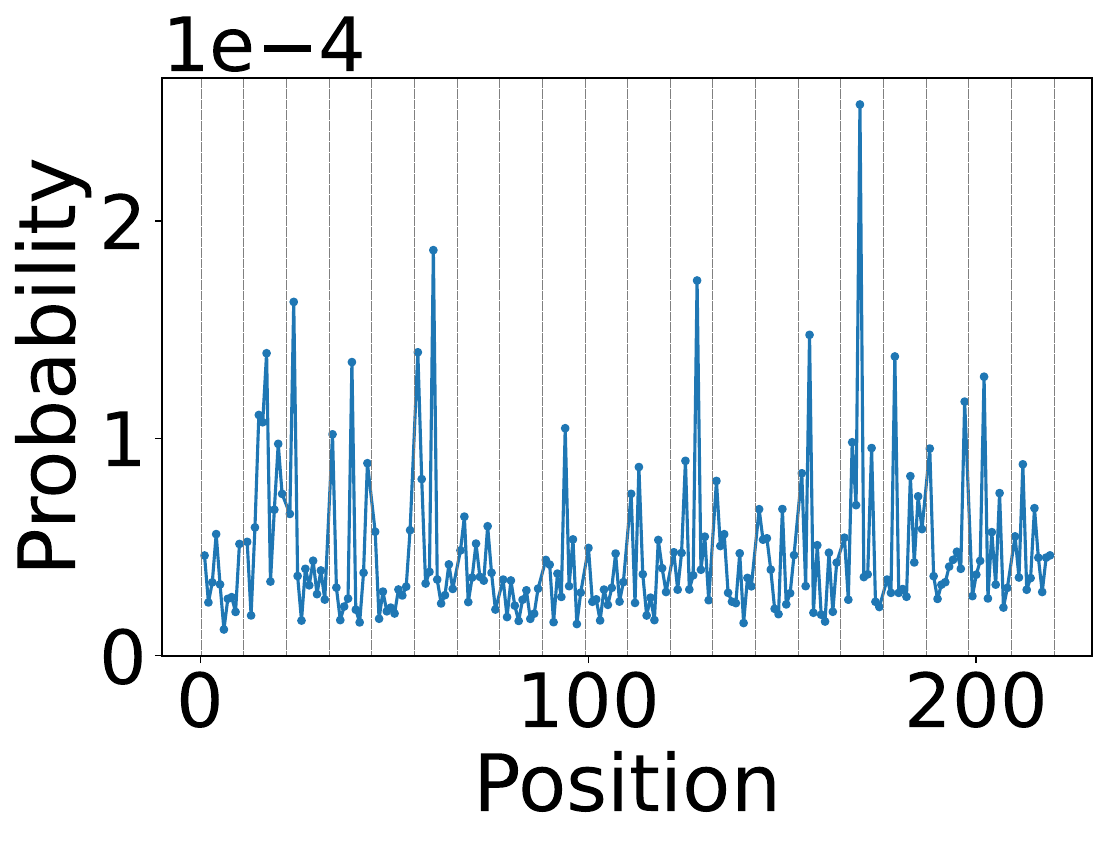} \\
& \FigGridVSubLabel{Rand} &
\FigGridCell{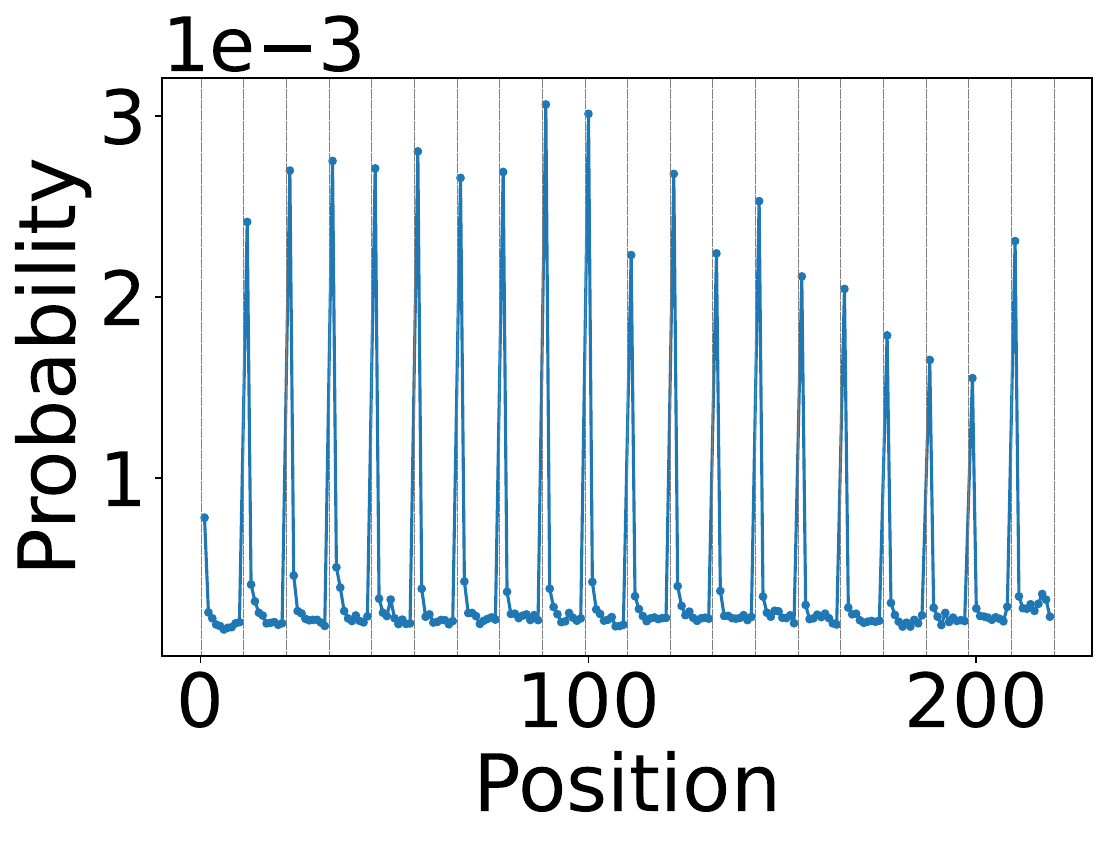} &
\FigGridCell{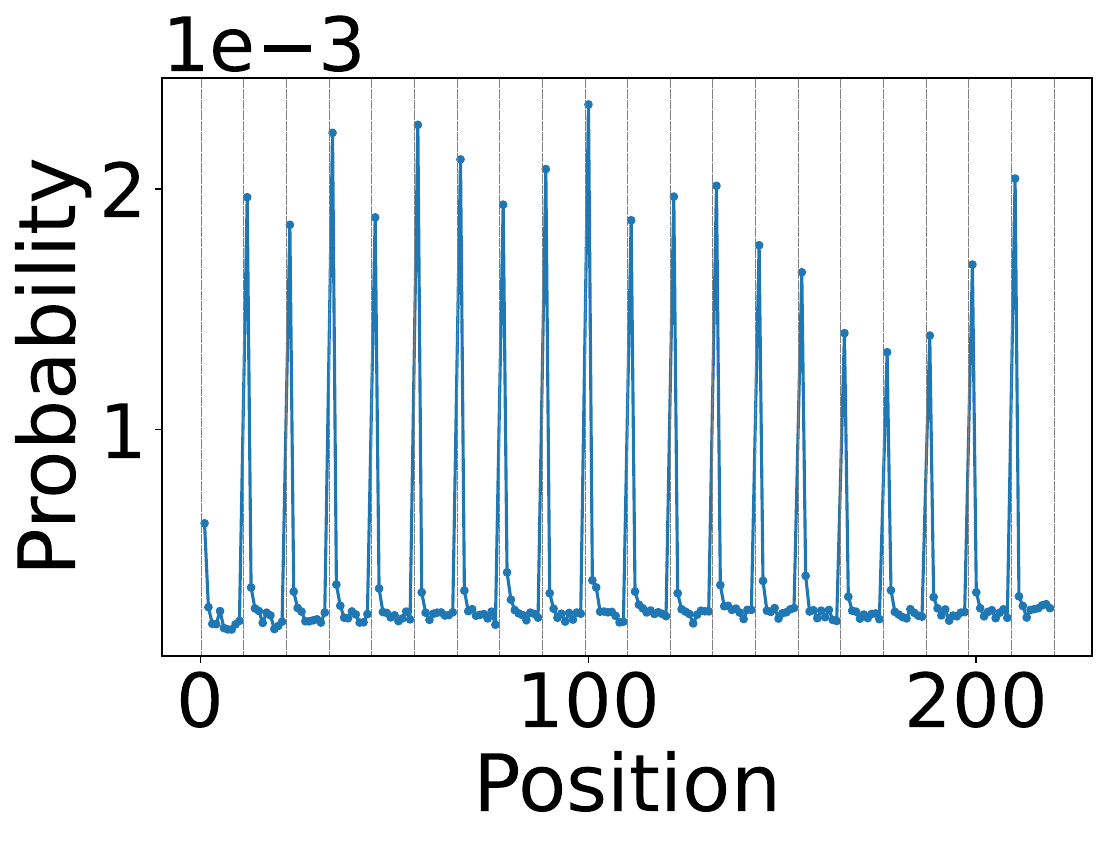} &
\FigGridCell{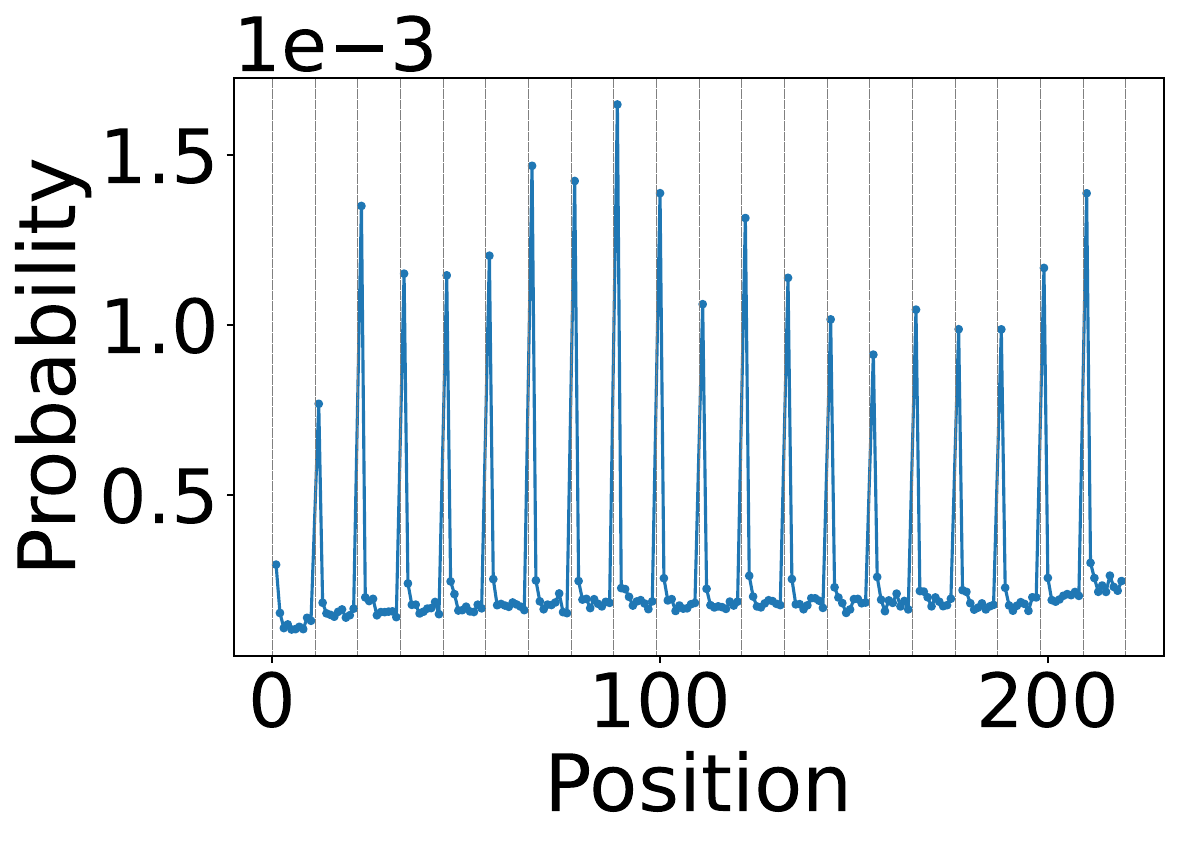} &
\FigGridCell{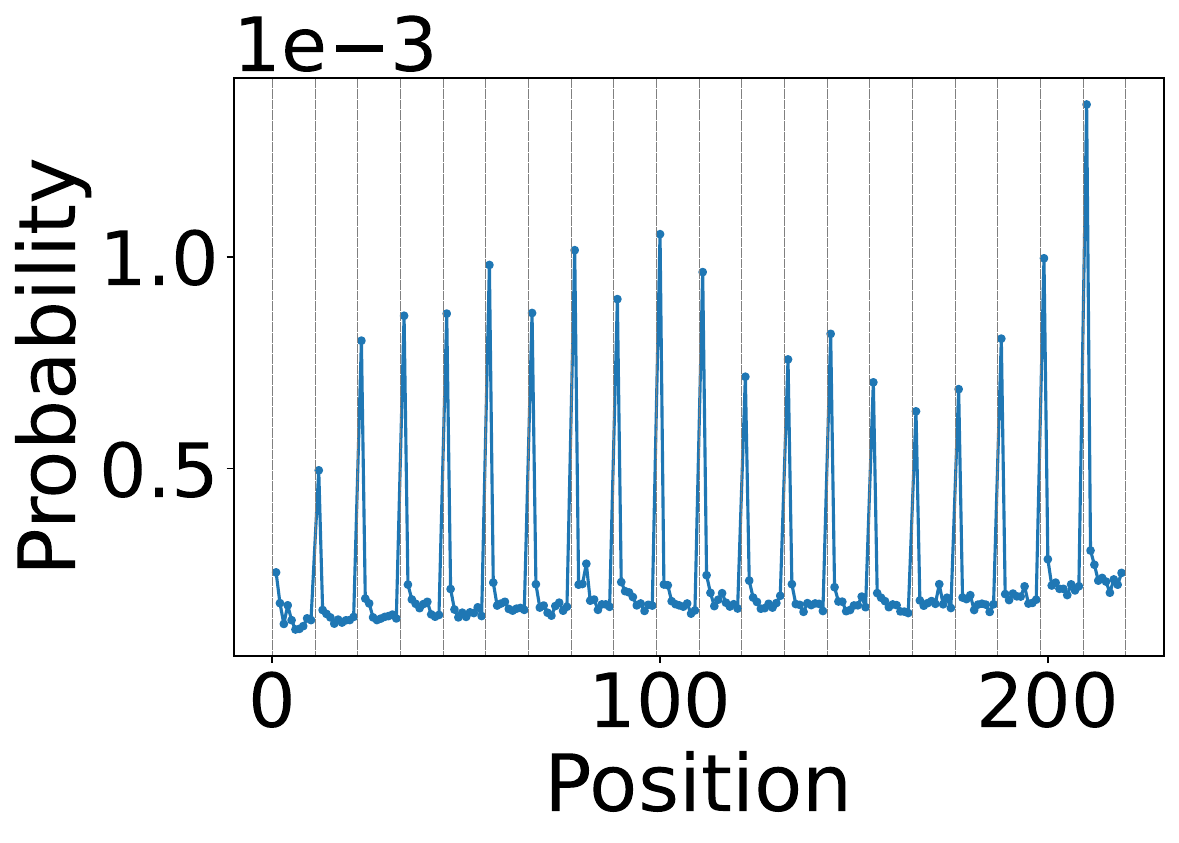} \\
\bottomrule
\end{tabular}
\endgroup
\caption{
SSM ablation effect (Exp. 1). Probability vs. position after ablating Induction (Ind) or Random (Rand) channels (rows) for Mamba, Falcon-Mamba, Recurrent-Gemma (model pairs per 2 rows). Columns show the percentage of ablated channels.
}
\label{fig:exp1_ablations_ssm}
\end{figure*}

\begin{figure*}[h!]
\centering
\begingroup
\setlength{\FigGridLabelW}{0.95cm}
\setlength{\FigGridPlotW}{0.265\textwidth}
\setlength{\FigGridCellH}{2.05cm}
\setlength{\tabcolsep}{3pt}
\renewcommand{\arraystretch}{1.0}

\begin{adjustbox}{max width=\textwidth}
\begin{tabular}{@{}>{\centering\arraybackslash}m{\FigGridLabelW}@{\hspace{4pt}}
                *{3}{>{\centering\arraybackslash}m{\FigGridPlotW}}@{}}
\toprule
& \multicolumn{3}{c}{\FigHdr Spacing} \\
\cmidrule(lr){2-4}
& \FigHdr 10 & \FigHdr 20 & \FigHdr 30 \\
\midrule
\FigGridVLabel{Llama} &
\FigGridCell{Figures/prob_without_A/llama-3.1-8B-Instruct_10_Repeats_10_Length_5000_Permutations.pdf} &
\FigGridCell{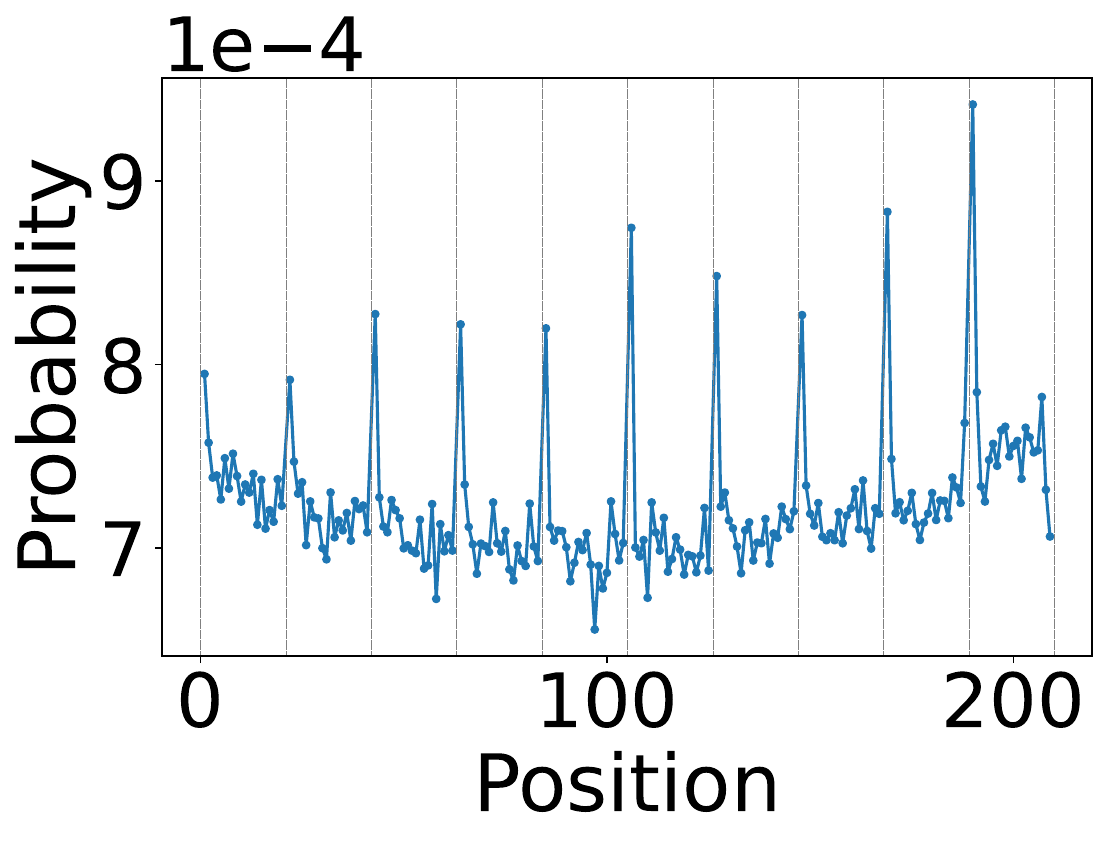} &
\FigGridCell{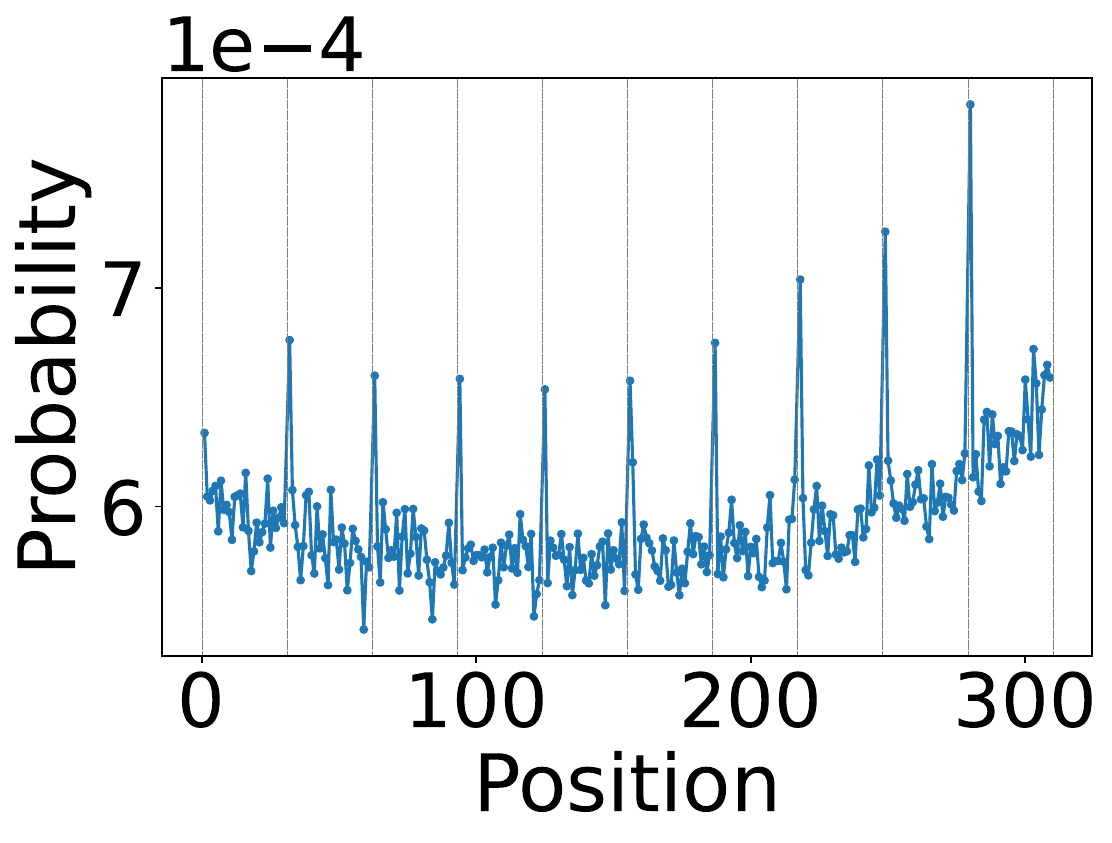} \\
\FigGridVLabel{Mistral} &
\FigGridCell{Figures/prob_without_A/Mistral-7B-Instruct-v0.1_10_Repeats_10_Length_5000_Permutations.pdf} &
\FigGridCell{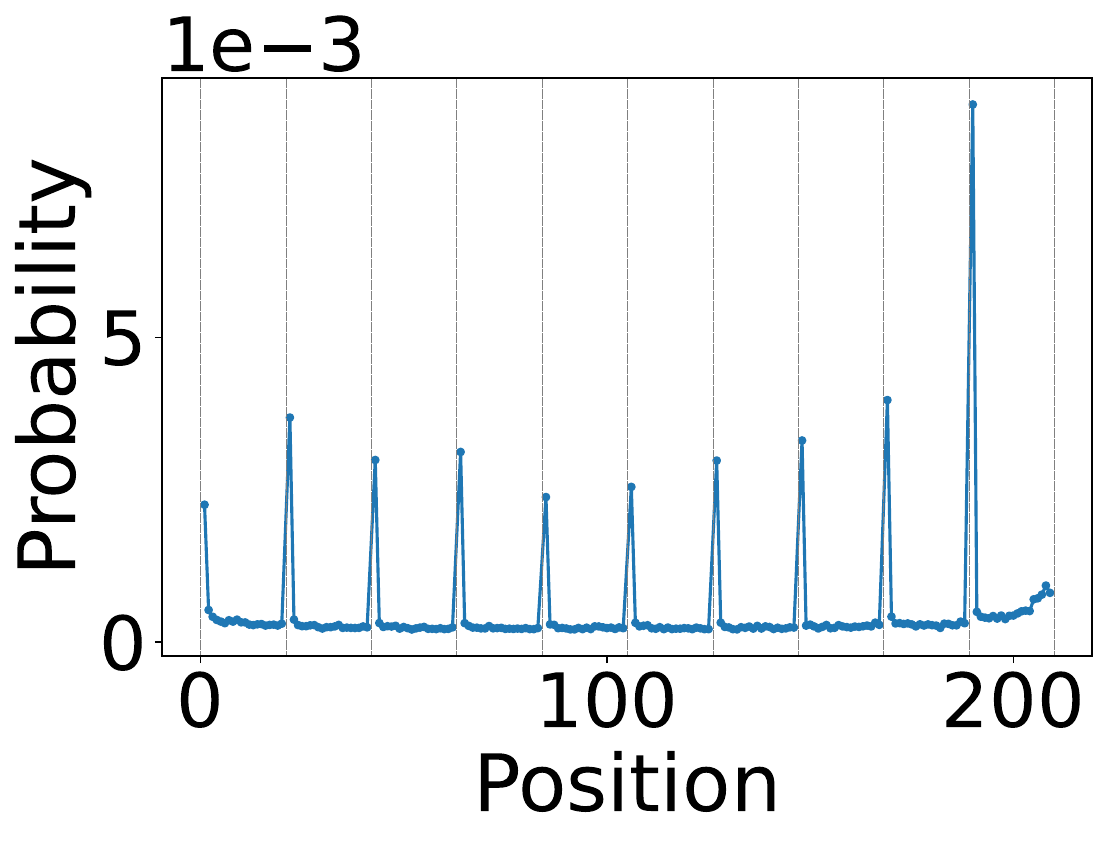} &
\FigGridCell{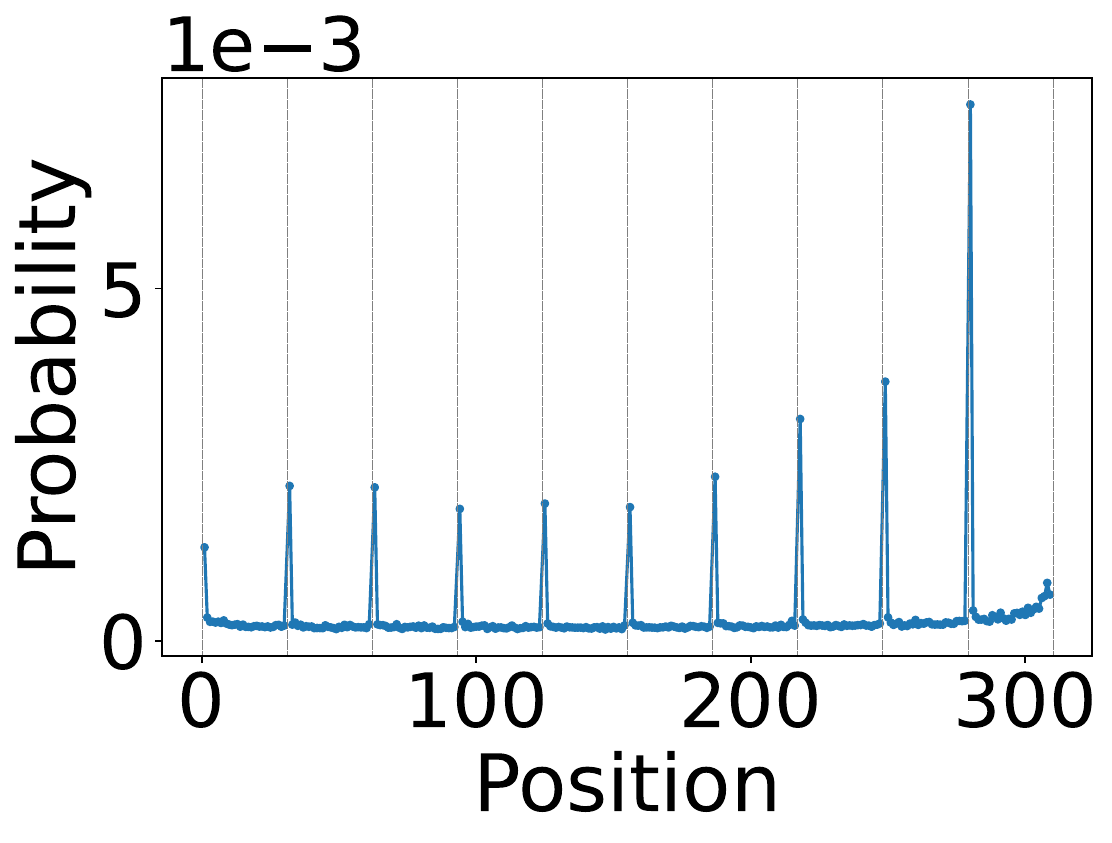} \\
\FigGridVLabel{Qwen} &
\FigGridCell{Figures/prob_without_A/Qwen2.5-7B-Instruct_10_Repeats_10_Length_5000_Permutations.pdf} &
\FigGridCell{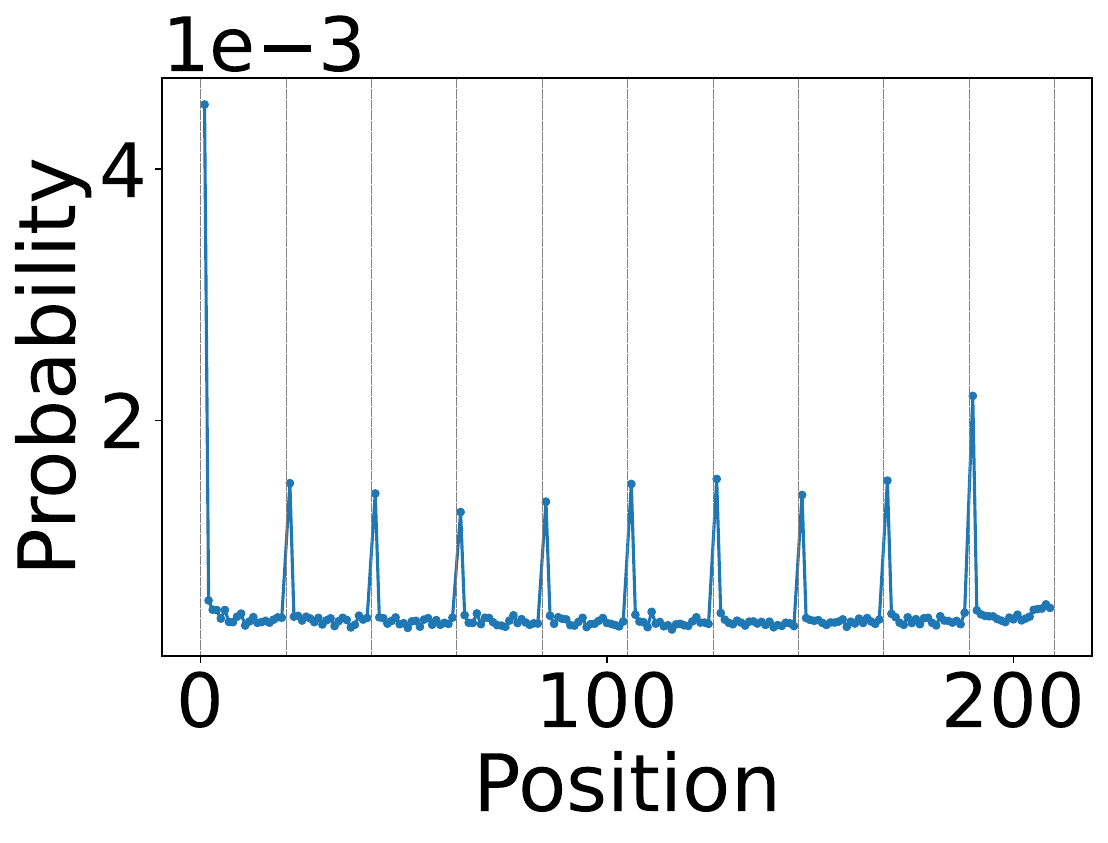} &
\FigGridCell{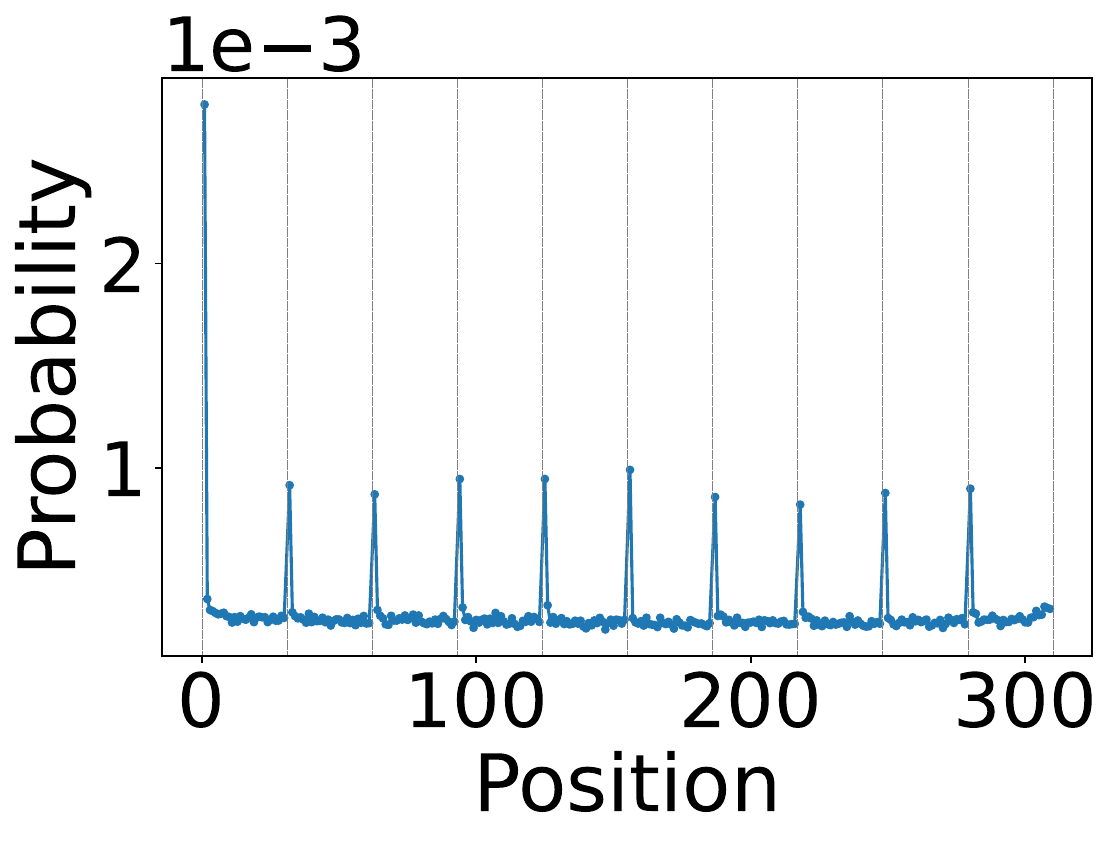} \\
\FigGridVLabel{Gemma} &
\FigGridCell{Figures/prob_without_A/gemma-2-9b-it_10_Repeats_10_Length_5000_Permutations.pdf} &
\FigGridCell{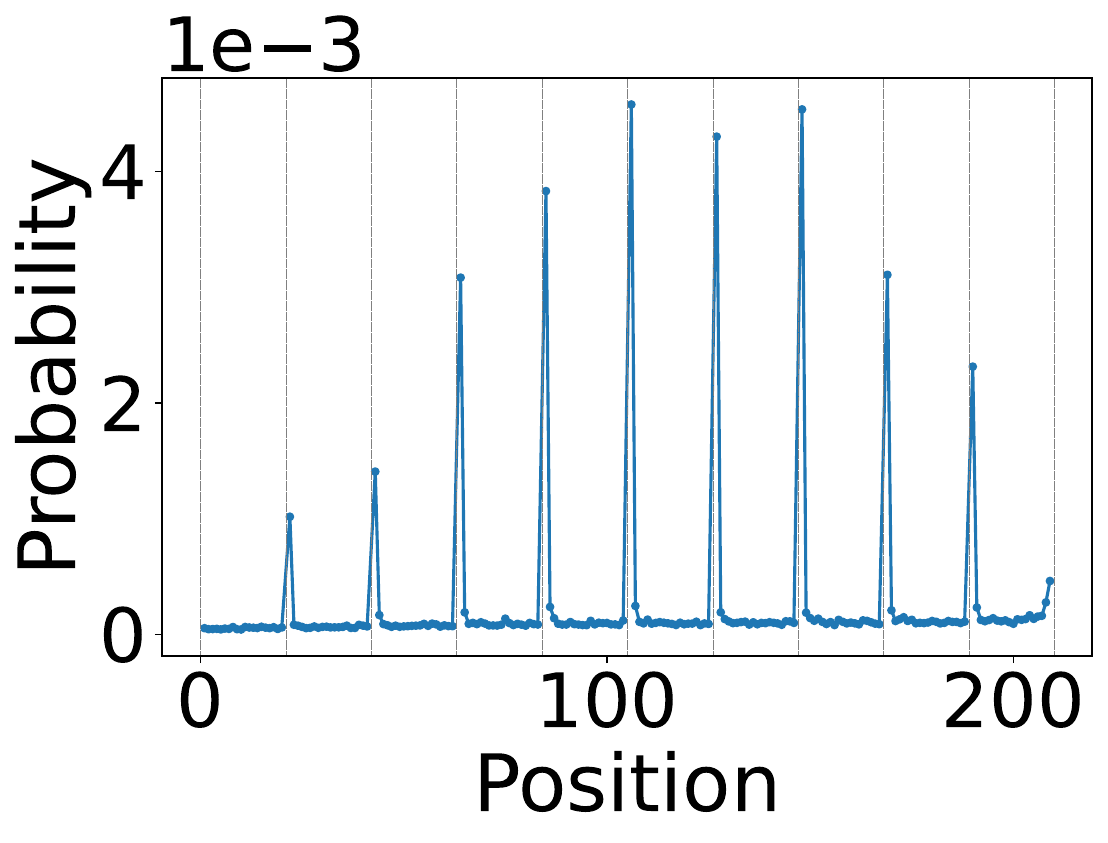} &
\FigGridCell{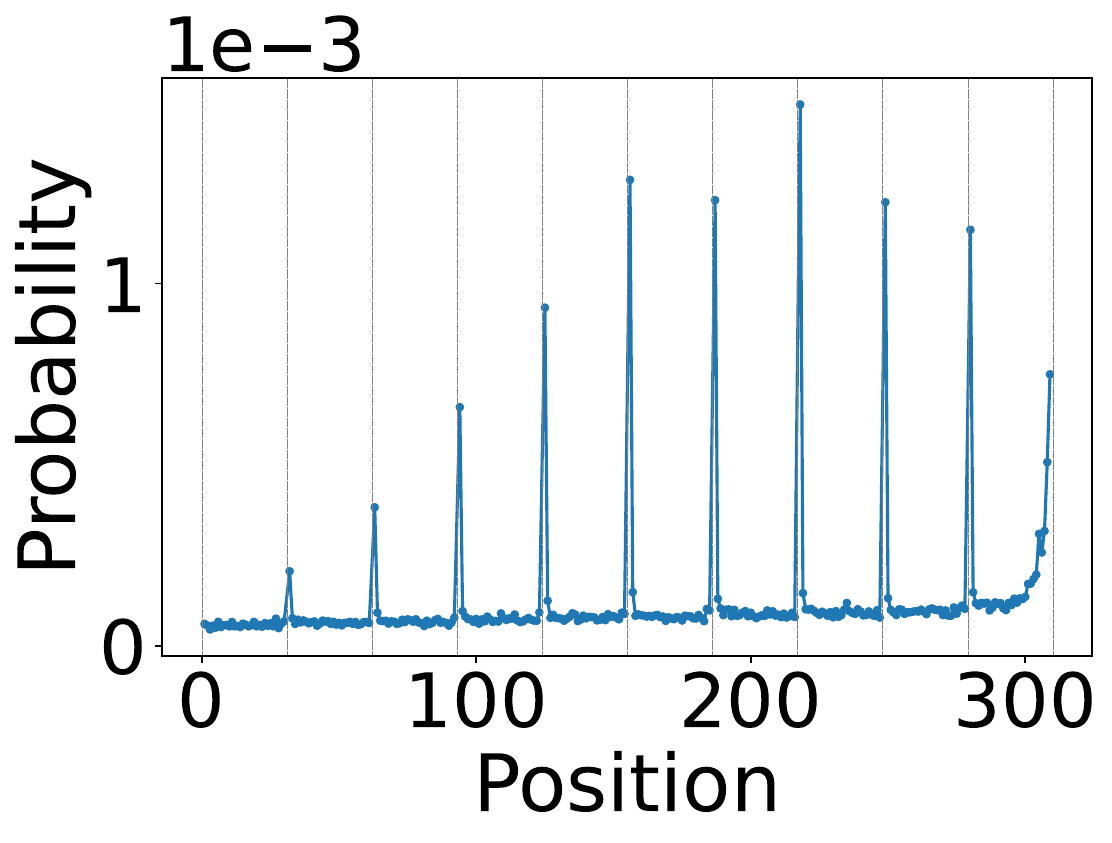} \\

\addlinespace[1pt]
\midrule
\addlinespace[1pt]

\FigGridVLabel{Mamba} &
\FigGridCell{Figures/prob_without_A/mamba-130m-hf_10_Repeats_10_Length_5000_Permutations.pdf} &
\FigGridCell{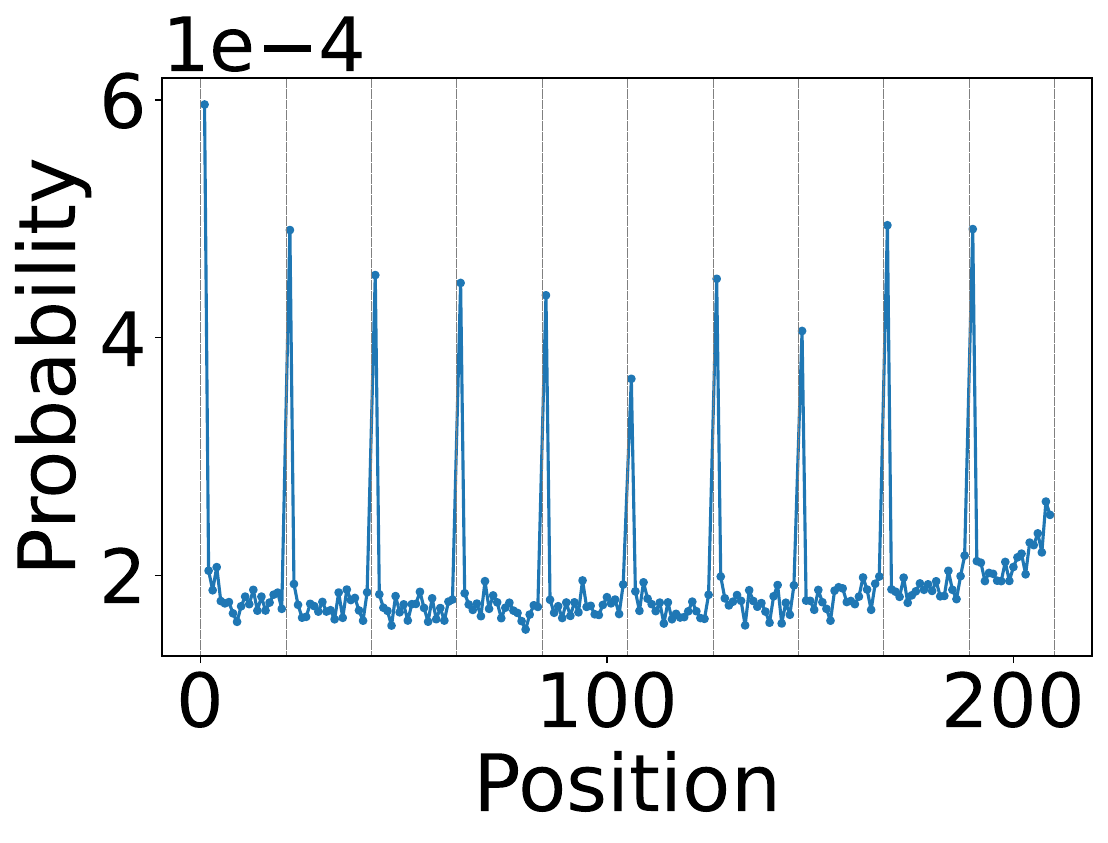} &
\FigGridCell{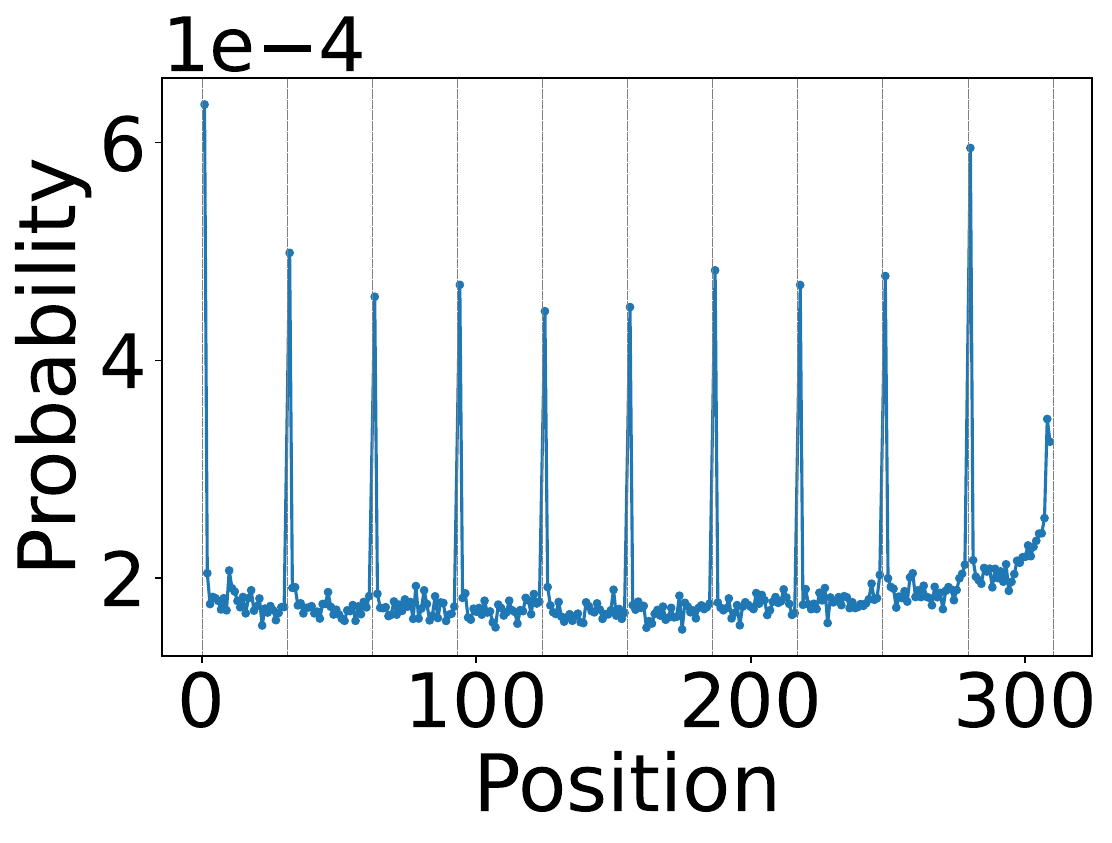} \\
\FigGridVLabel{Falcon-M} &
\FigGridCell{Figures/prob_without_A/Falcon3-Mamba-7B-Instruct_10_Repeats_10_Length_5000_Permutations.pdf} &
\FigGridCell{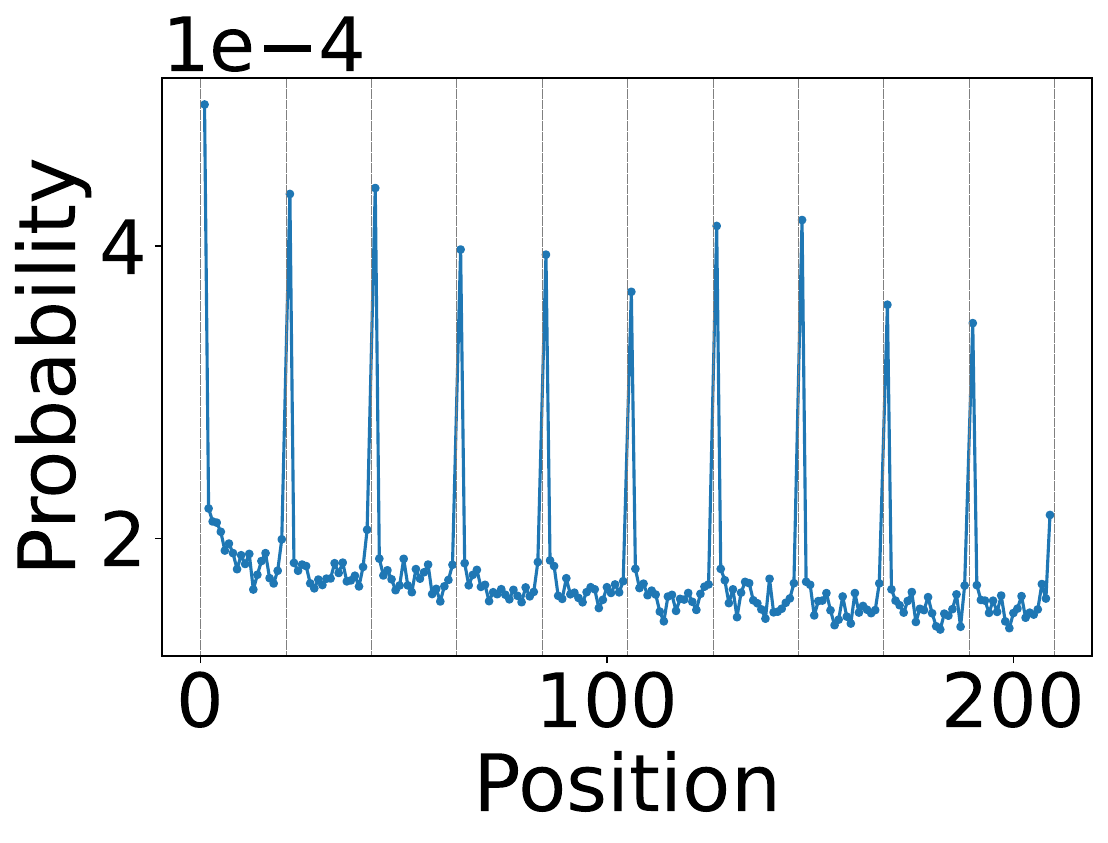} &
\FigGridCell{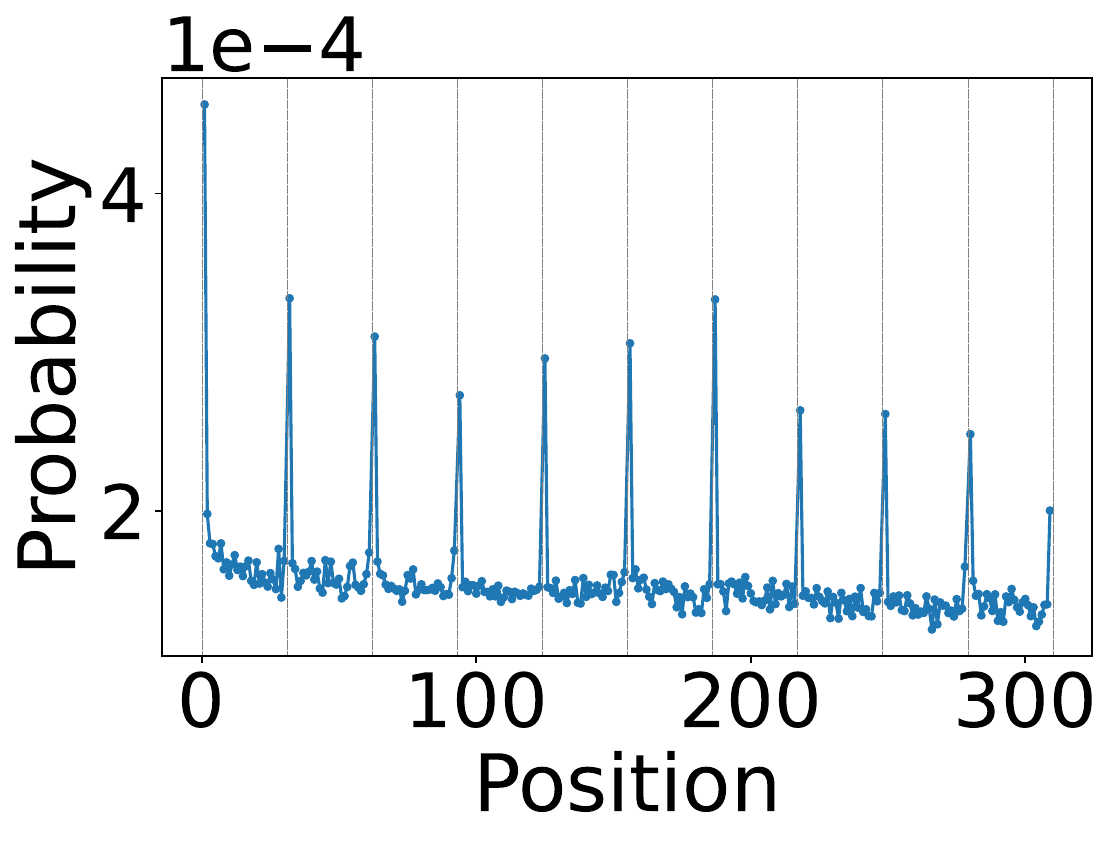} \\
\FigGridVLabel{R-Gemma} &
\FigGridCell{Figures/prob_without_A/recurrentgemma-9b-it_10_Repeats_10_Length_5000_Permutations.pdf} &
\FigGridCell{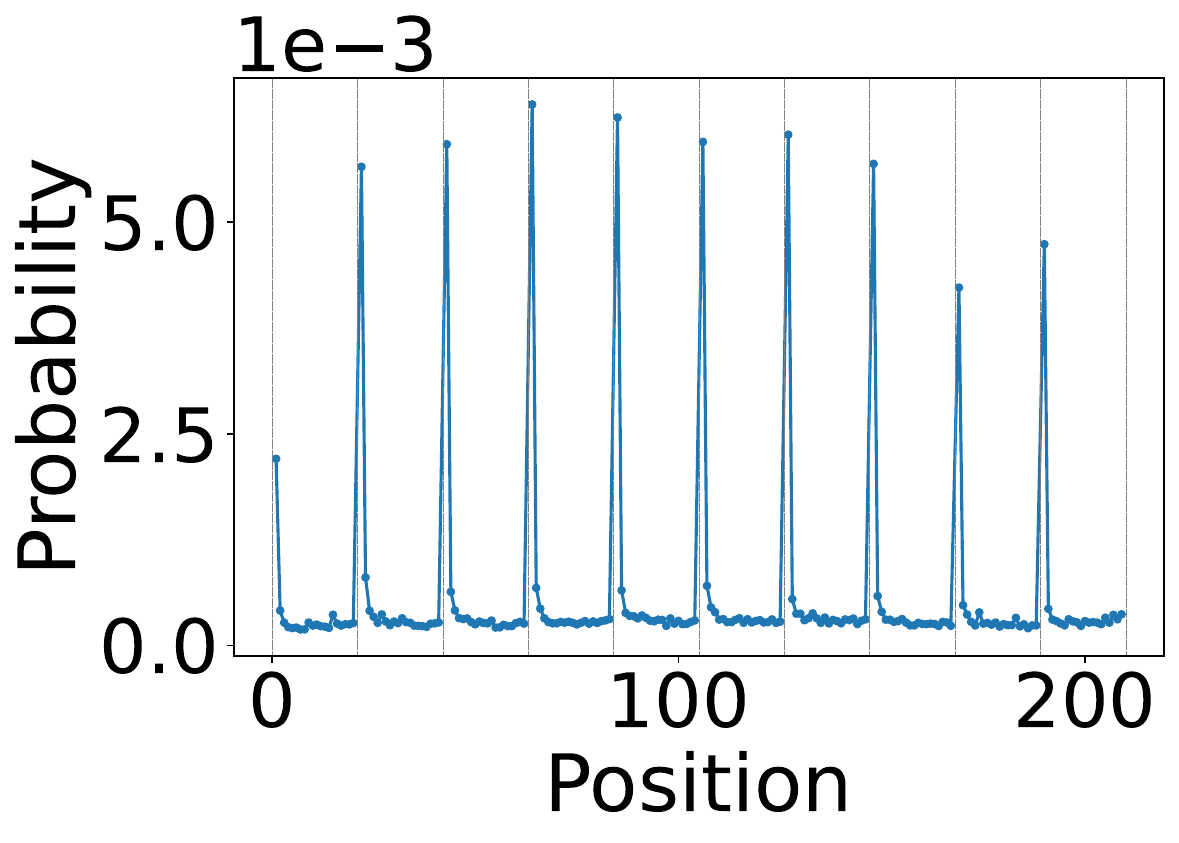} &
\FigGridCell{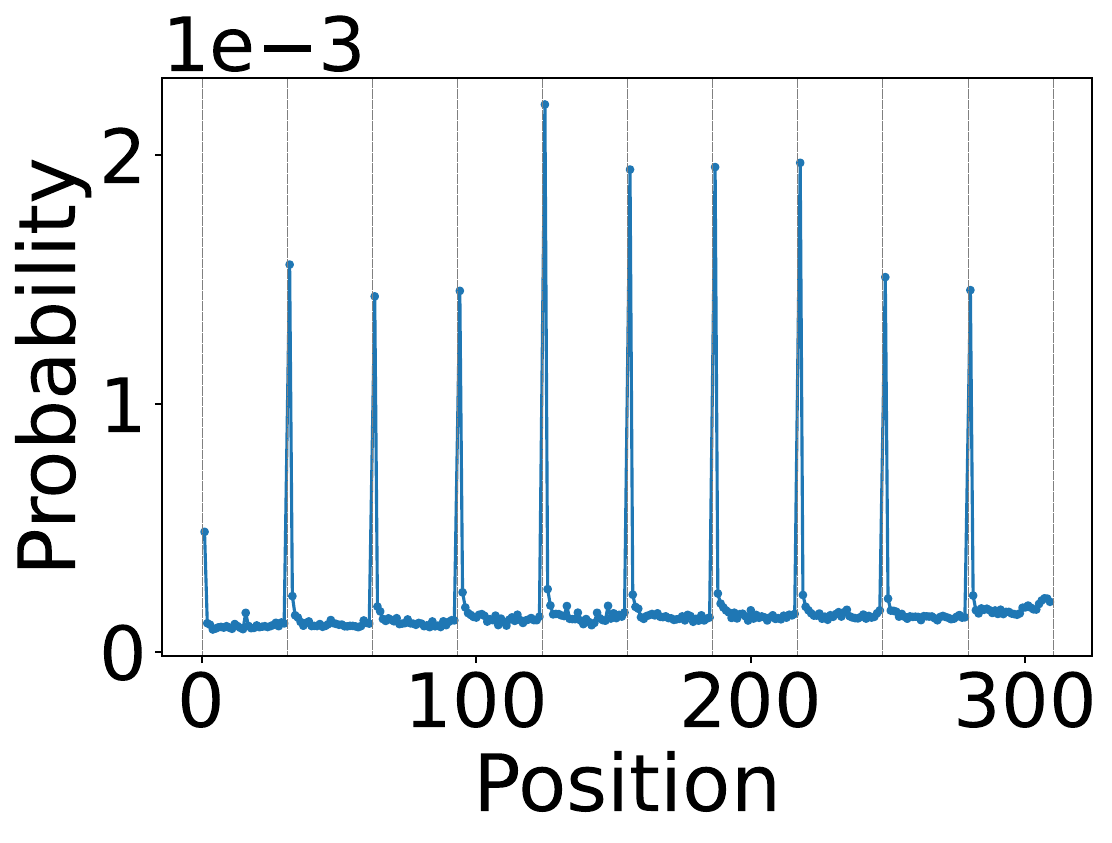} \\
\bottomrule
\end{tabular}
\end{adjustbox}
\endgroup
\caption{
Next-token probability vs. position, varying spacing between fixed tokens (columns, repeats=10). Shows \plusoneTok\ recall preference and positional biases.
}
\label{fig:single_target_all_spacings_all_tokens}
\end{figure*}

\newcommand{\FigGridVLabelSpanFive}[1]{%
  \begin{minipage}[c][\dimexpr5\FigGridCellH\relax][c]{\FigGridLabelW}%
    \centering
    \rotatebox[origin=c]{90}{\FigLabelFont\strut #1}%
  \end{minipage}%
}

\begin{figure*}[h!]
\centering
\begingroup
\setlength{\FigGridLabelW}{1.05cm}
\setlength{\FigGridSubLabelW}{0.55cm}
\setlength{\FigGridCellH}{1.85cm} 
\setlength{\FigGridLabelSep}{1pt}
\setlength{\FigGridSubLabelSep}{1pt}
\setlength{\FigGridColSep}{0.5pt}

\FigGridSetupTwo{5}
\renewcommand{\arraystretch}{1.0}

\begin{tabular}{@{}>{\centering\arraybackslash}m{\FigGridLabelW}@{\hspace{\FigGridLabelSep}}
                >{\centering\arraybackslash}m{\FigGridSubLabelW}@{\hspace{\FigGridSubLabelSep}}
                *{5}{>{\centering\arraybackslash}m{\FigGridPlotW}}@{}}
\toprule
& & \multicolumn{5}{c}{\FigHdr Ablations} \\
\cmidrule(lr){3-7}
& & \FigHdr 0 & \FigHdr 1 & \FigHdr 10 & \FigHdr 50 & \FigHdr 100 \\
\midrule

\multirow{5}{*}{\FigGridVLabelSpanFive{Induction}} &
\FigGridVSubLabel{P1} &
\FigGridCell{Figures/ep_prob_without_A_red/Llama-3.1-8B-Instruct_5_Repeats_200_Length_500_Permutations_0_ablations_induction_1_nth.pdf} &
\FigGridCell{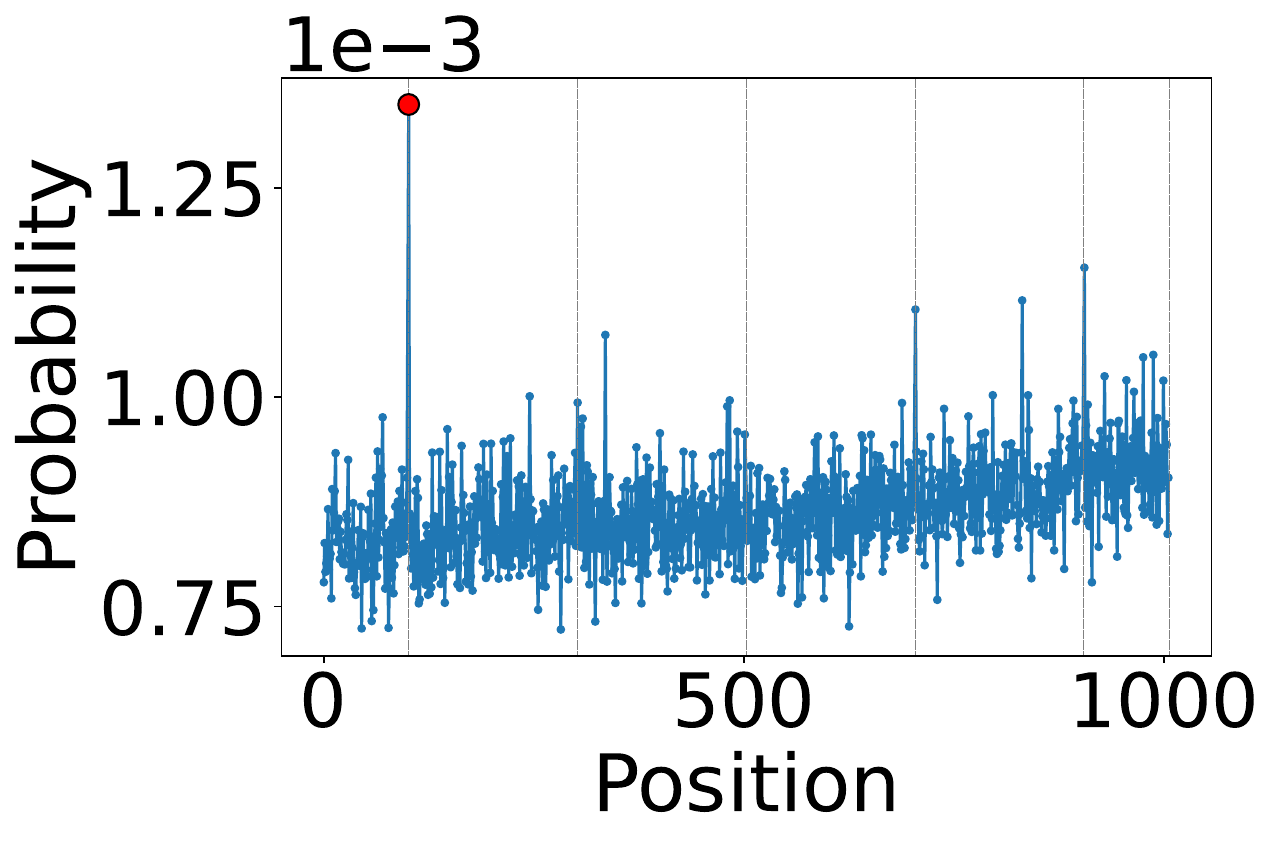} &
\FigGridCell{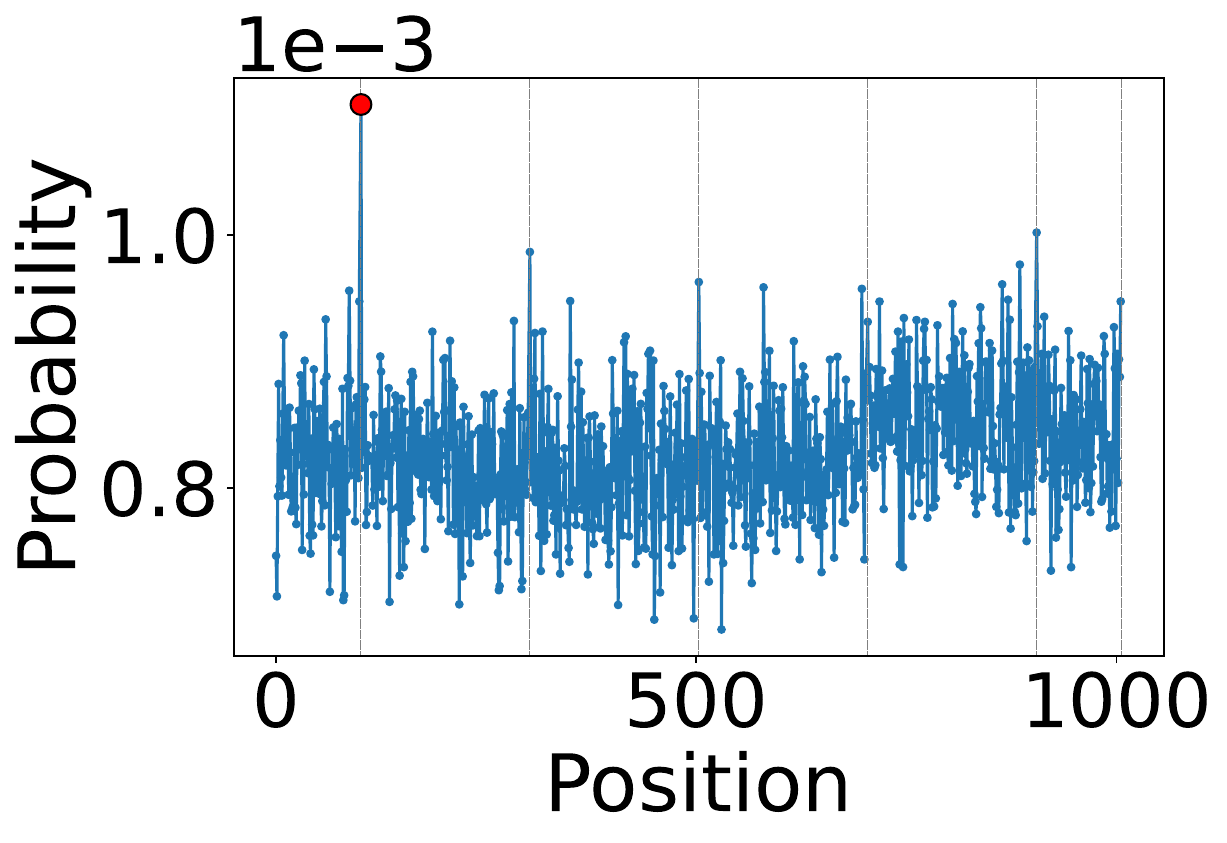} &
\FigGridCell{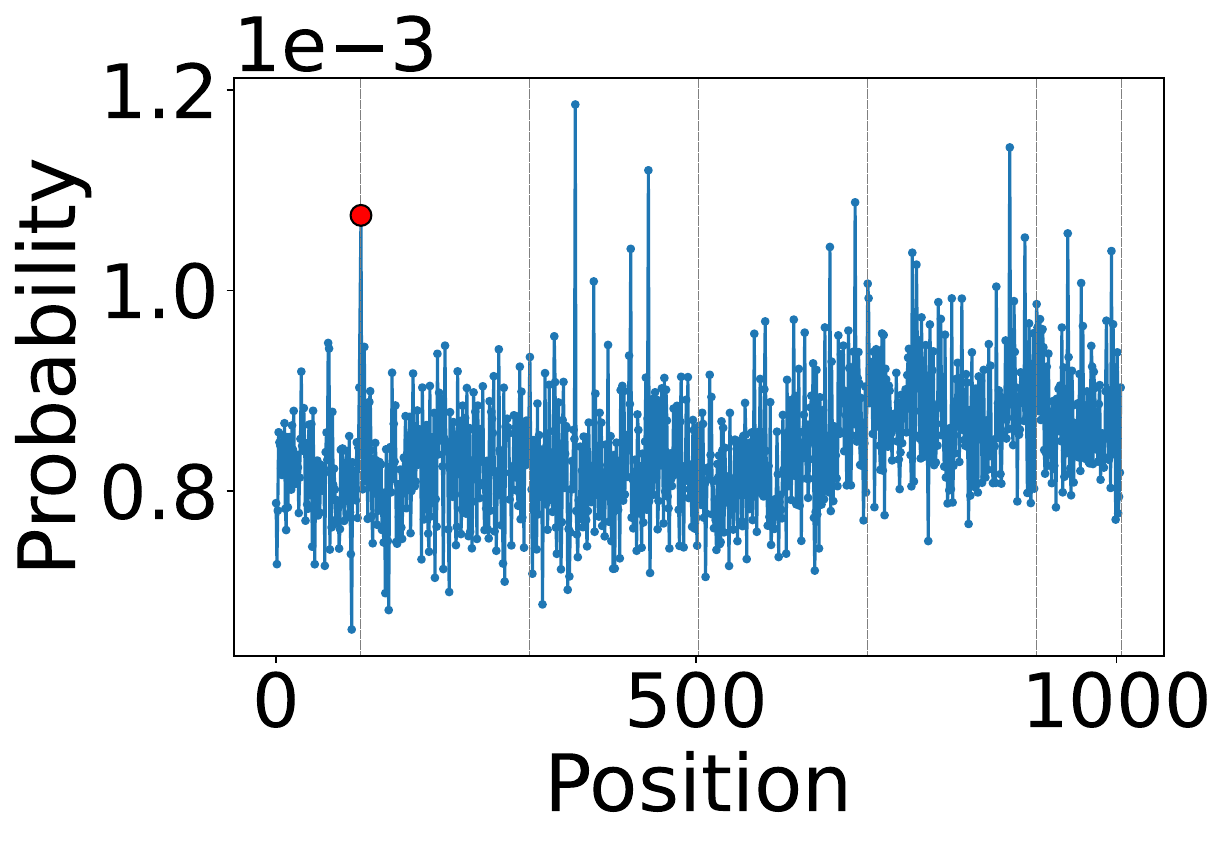} &
\FigGridCell{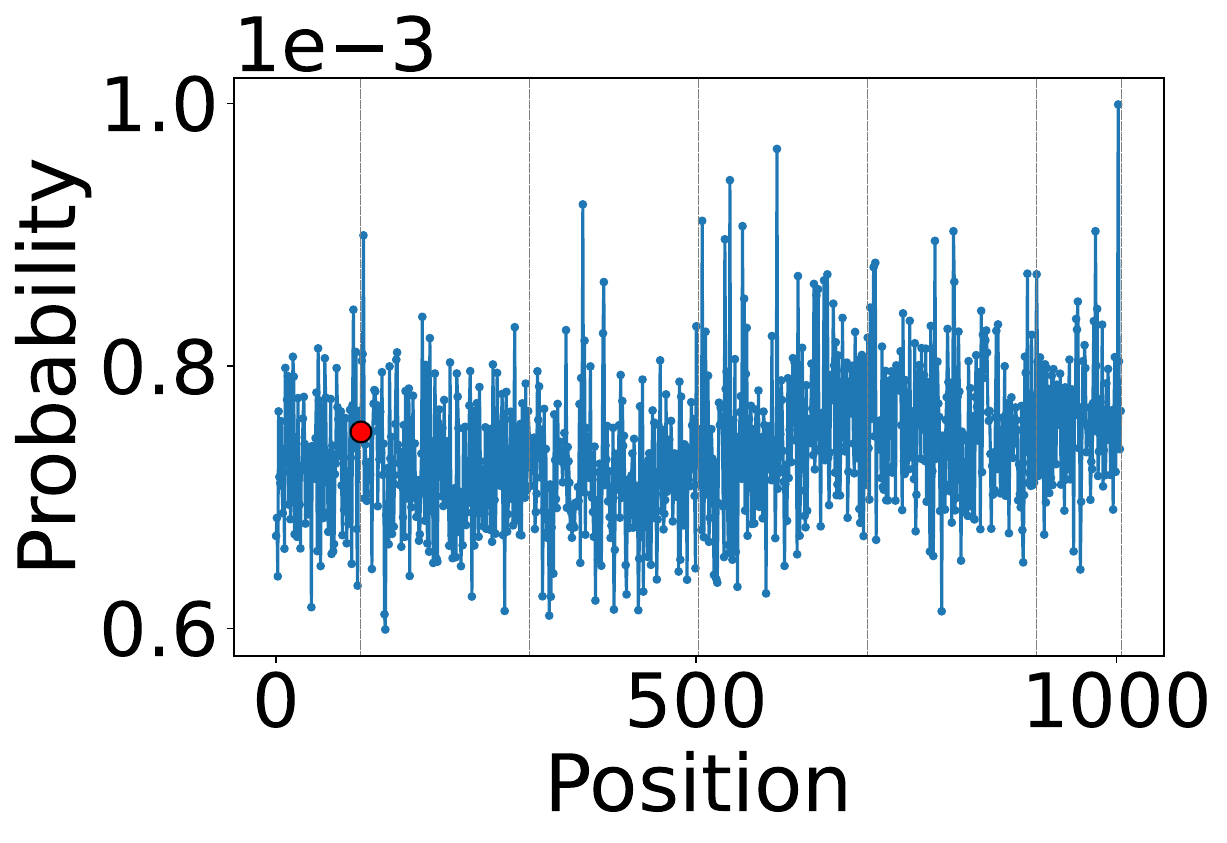} \\

& \FigGridVSubLabel{P2} &
\FigGridCell{Figures/ep_prob_without_A_red/Llama-3.1-8B-Instruct_5_Repeats_200_Length_500_Permutations_0_ablations_induction_2_nth.pdf} &
\FigGridCell{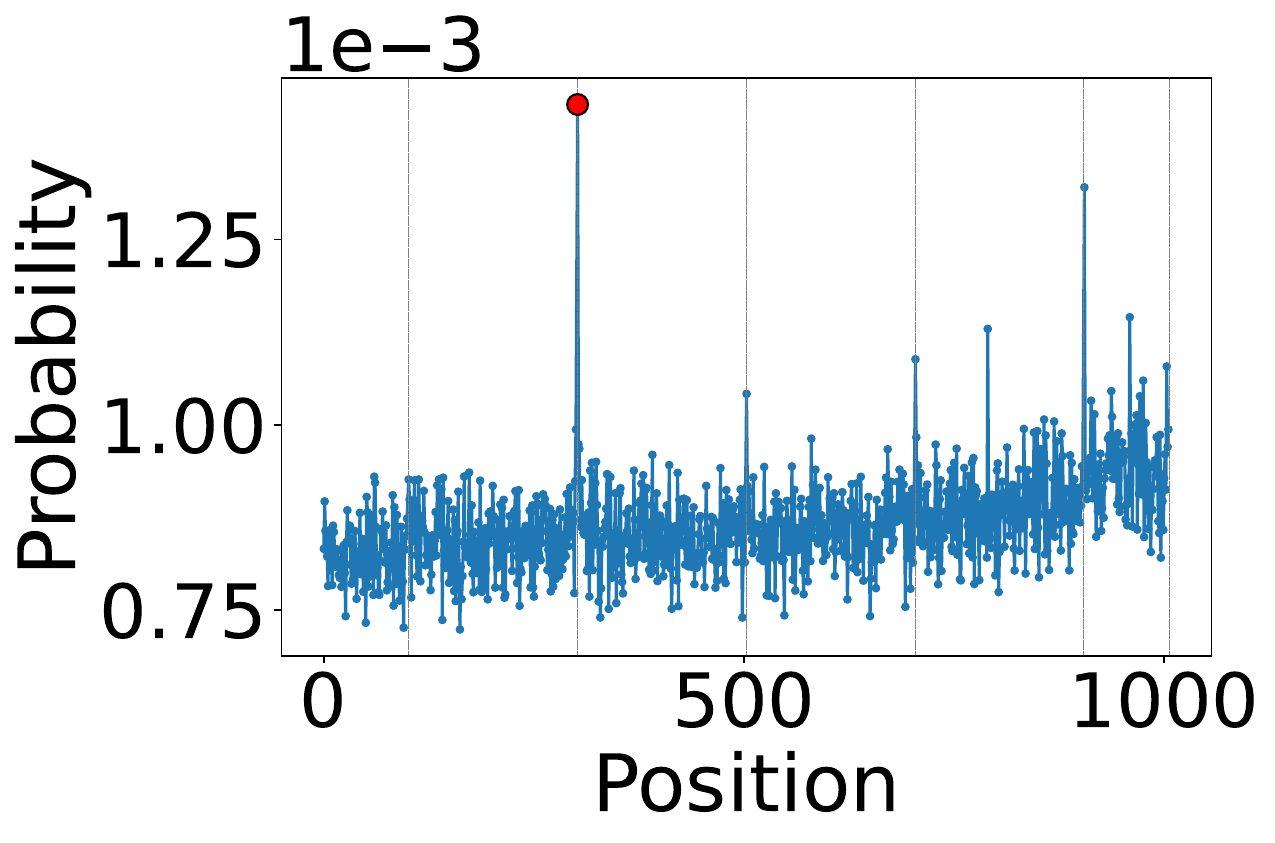} &
\FigGridCell{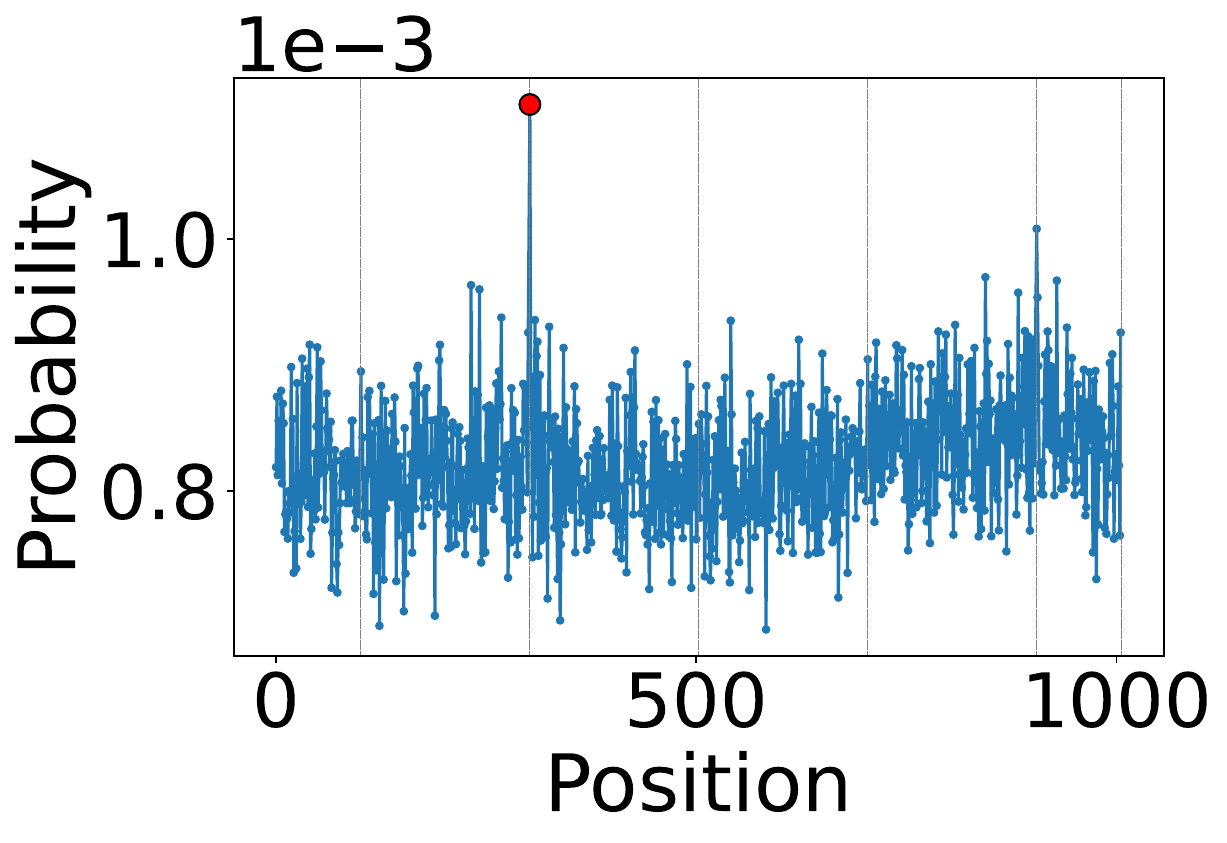} &
\FigGridCell{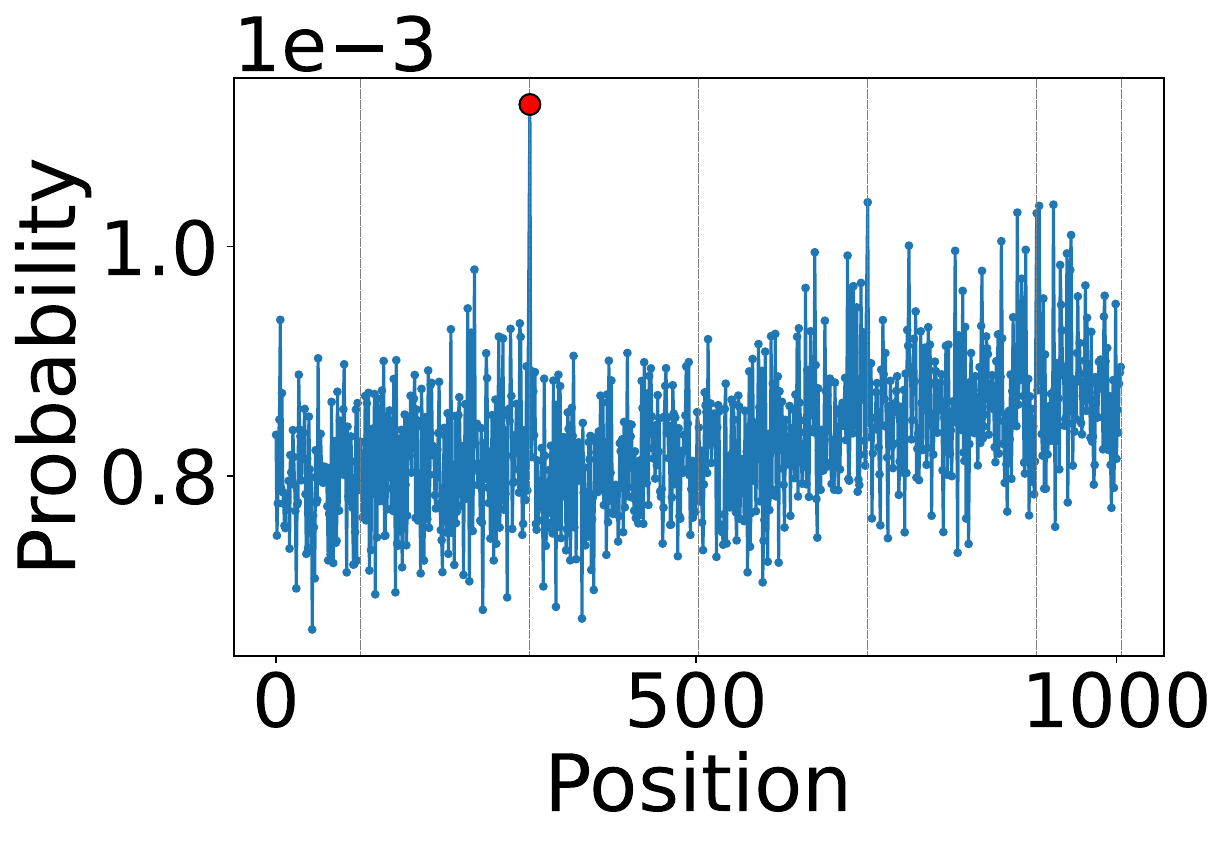} &
\FigGridCell{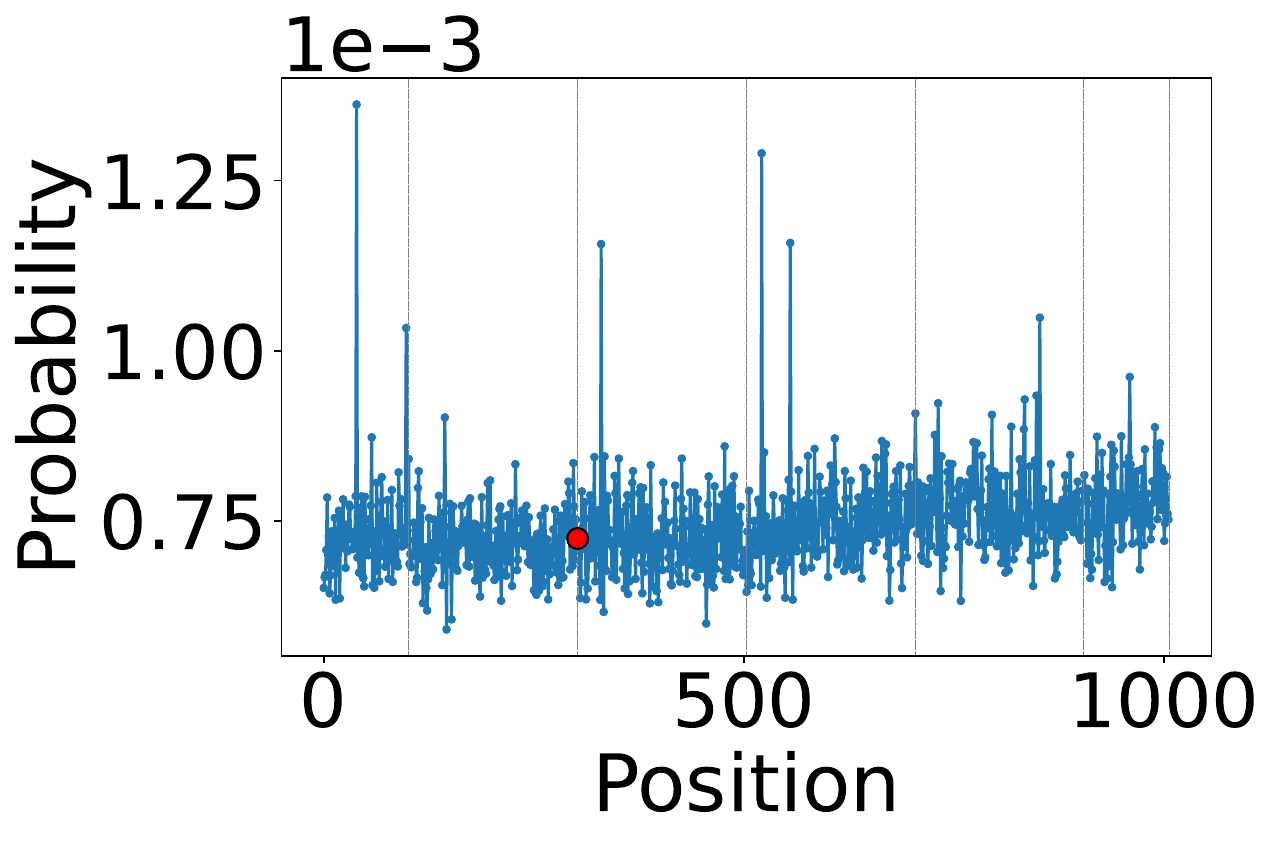} \\

& \FigGridVSubLabel{P3} &
\FigGridCell{Figures/ep_prob_without_A_red/Llama-3.1-8B-Instruct_5_Repeats_200_Length_500_Permutations_0_ablations_induction_3_nth.pdf} &
\FigGridCell{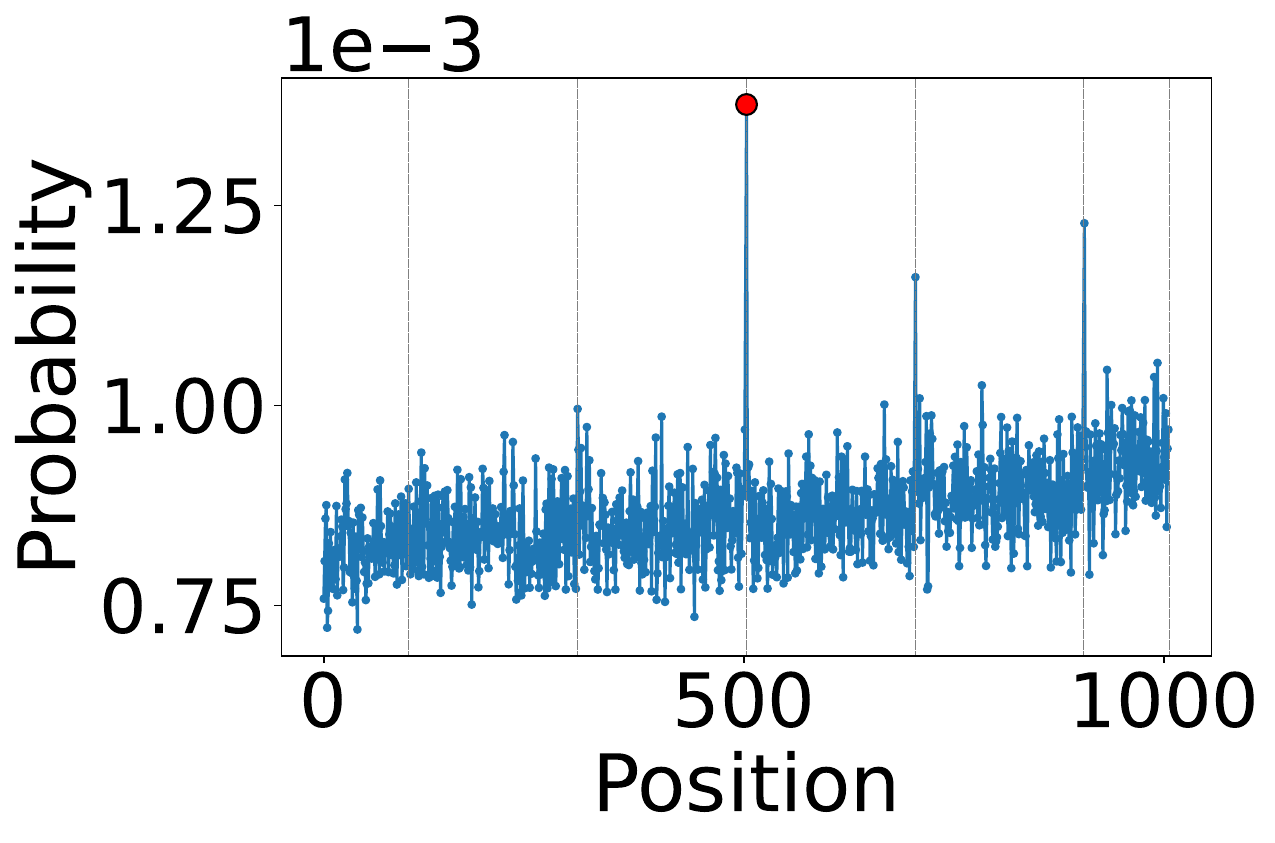} &
\FigGridCell{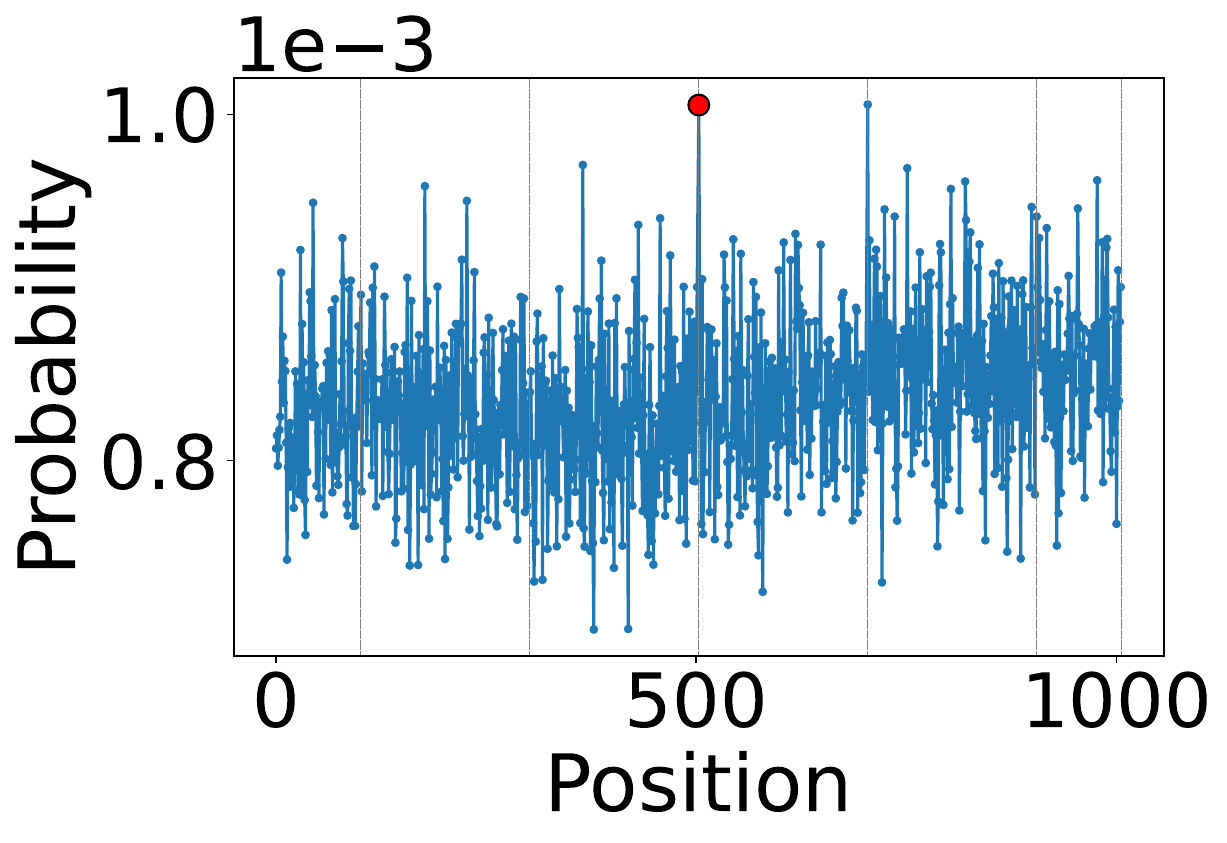} &
\FigGridCell{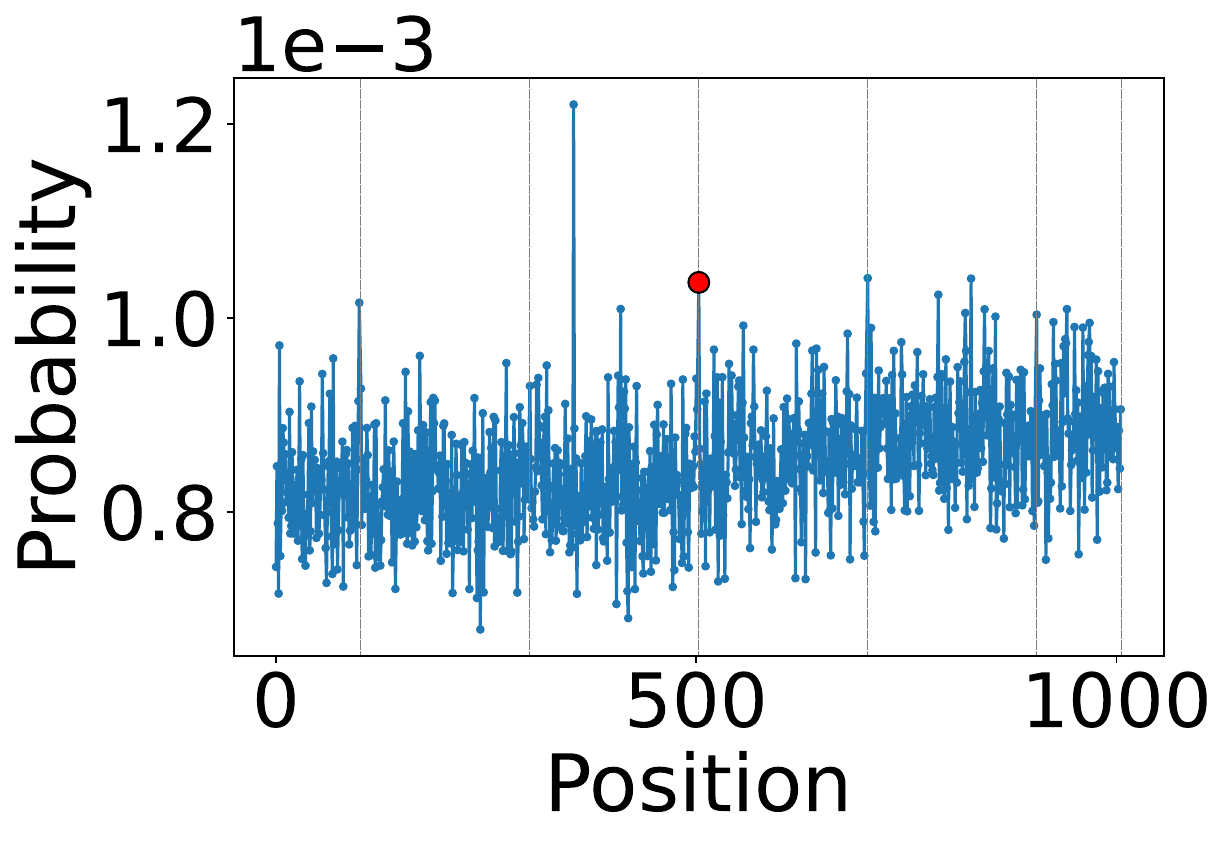} &
\FigGridCell{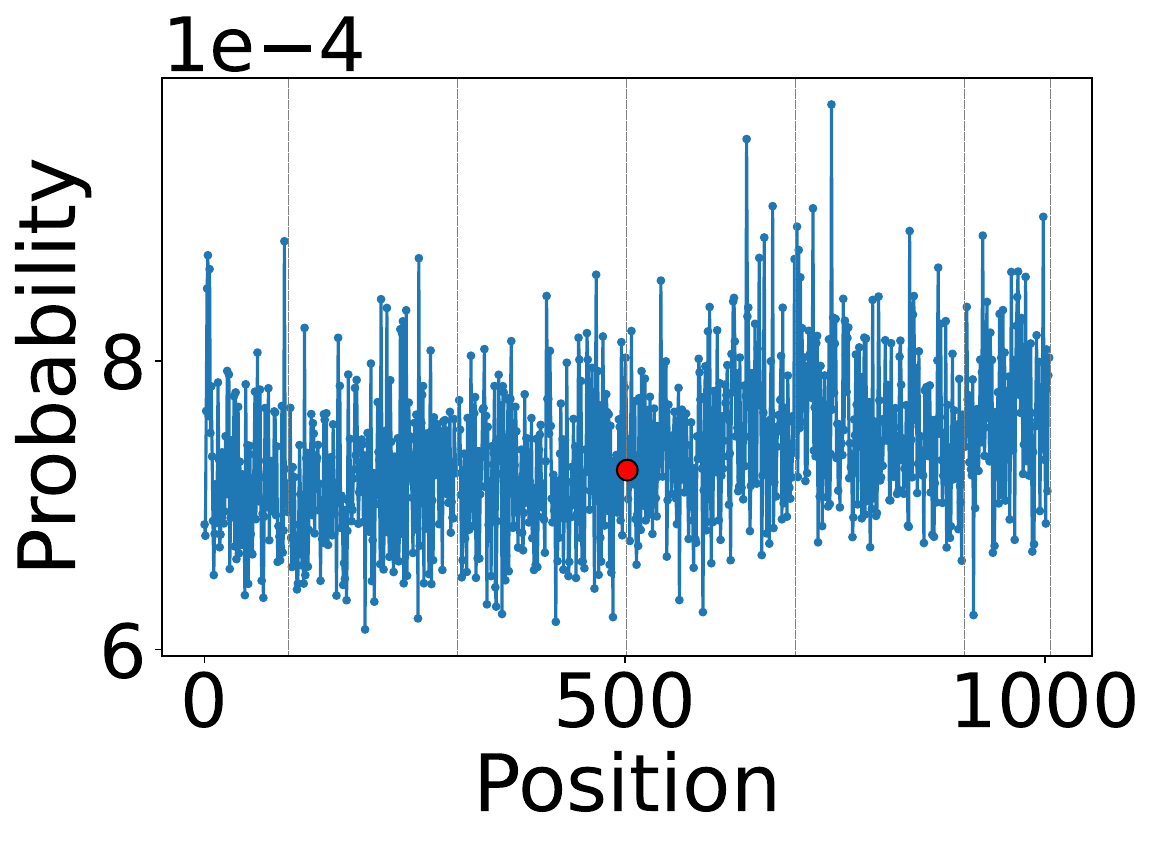} \\

& \FigGridVSubLabel{P4} &
\FigGridCell{Figures/ep_prob_without_A_red/Llama-3.1-8B-Instruct_5_Repeats_200_Length_500_Permutations_0_ablations_induction_4_nth.pdf} &
\FigGridCell{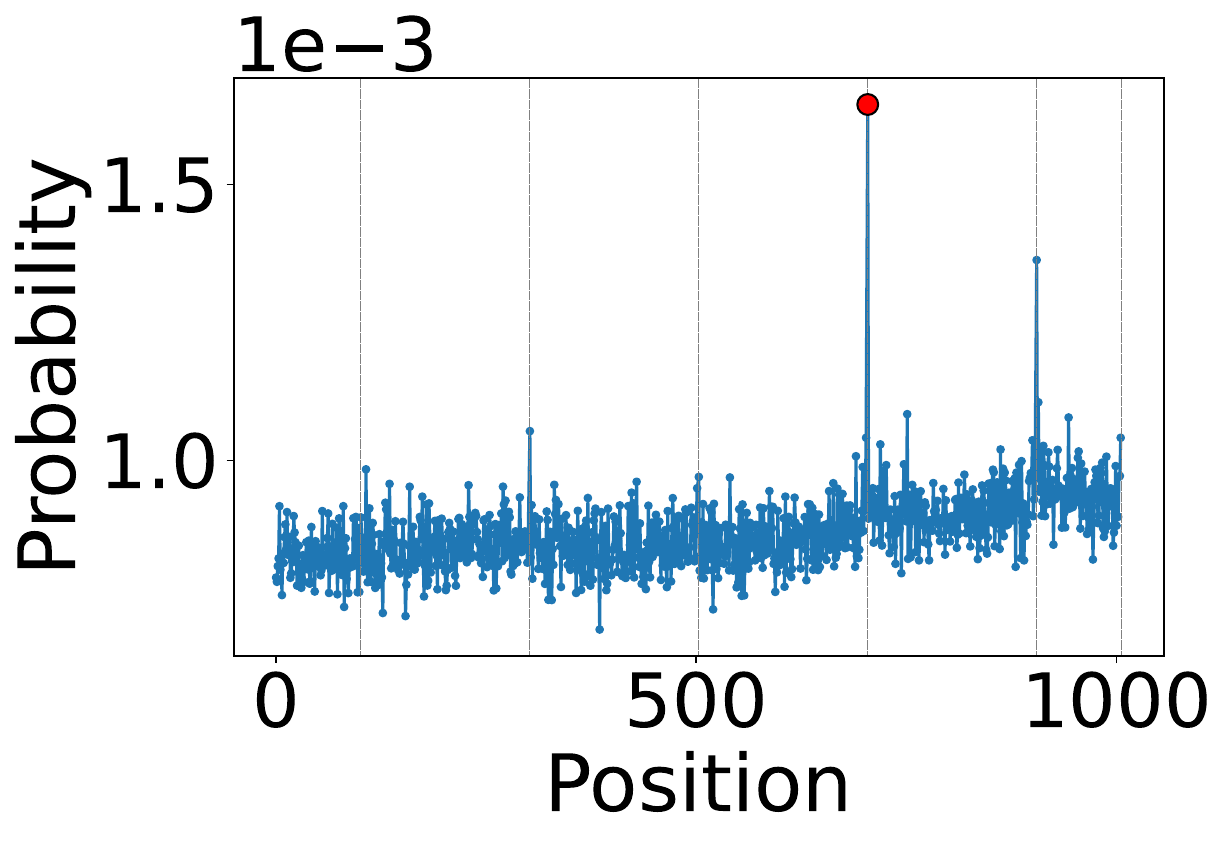} &
\FigGridCell{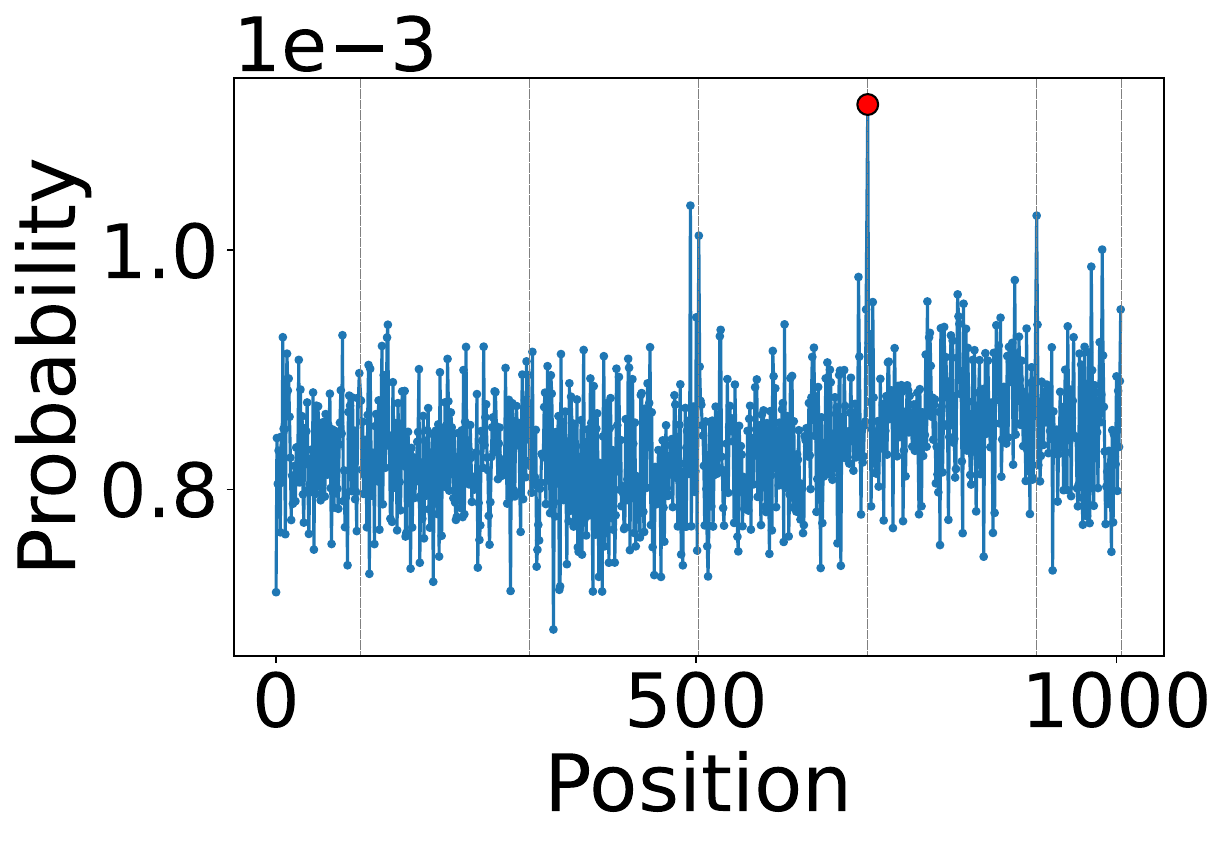} &
\FigGridCell{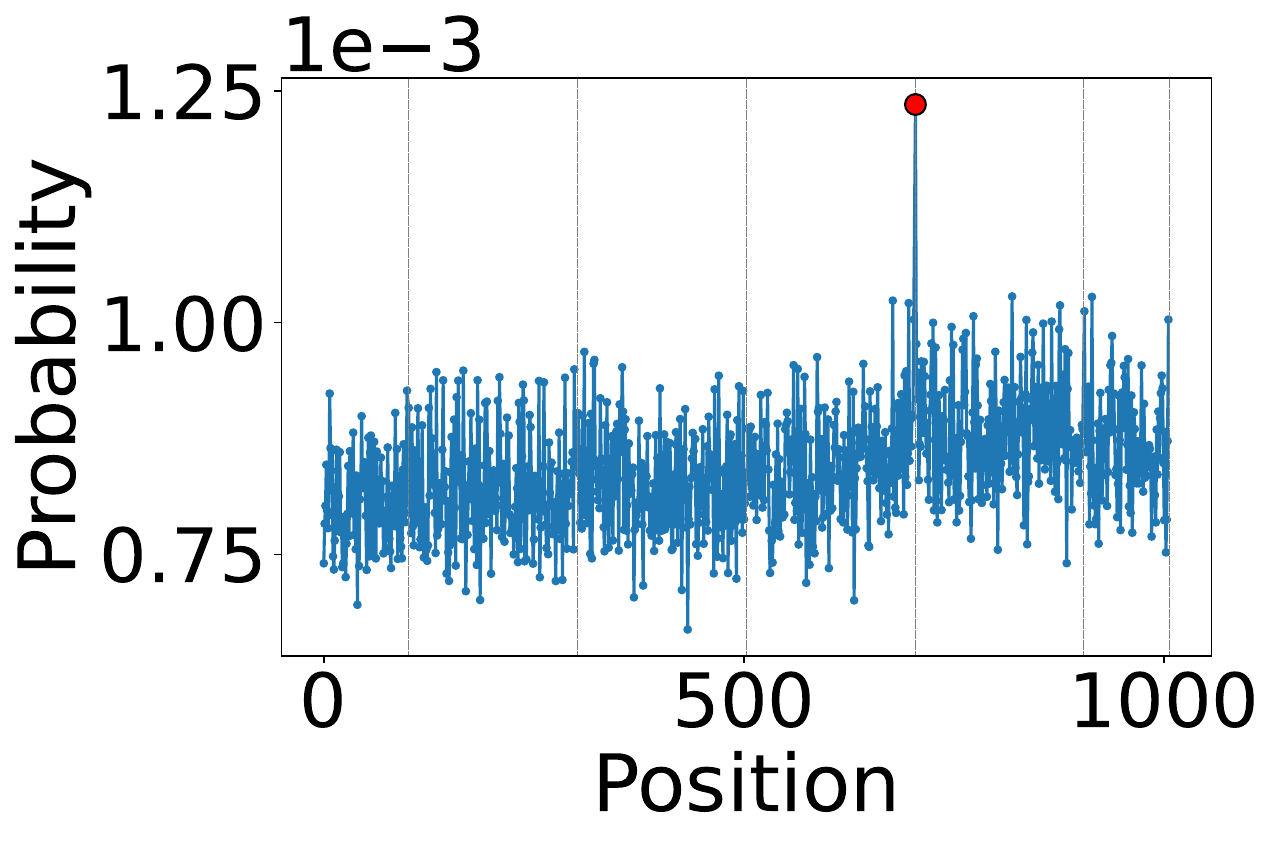} &
\FigGridCell{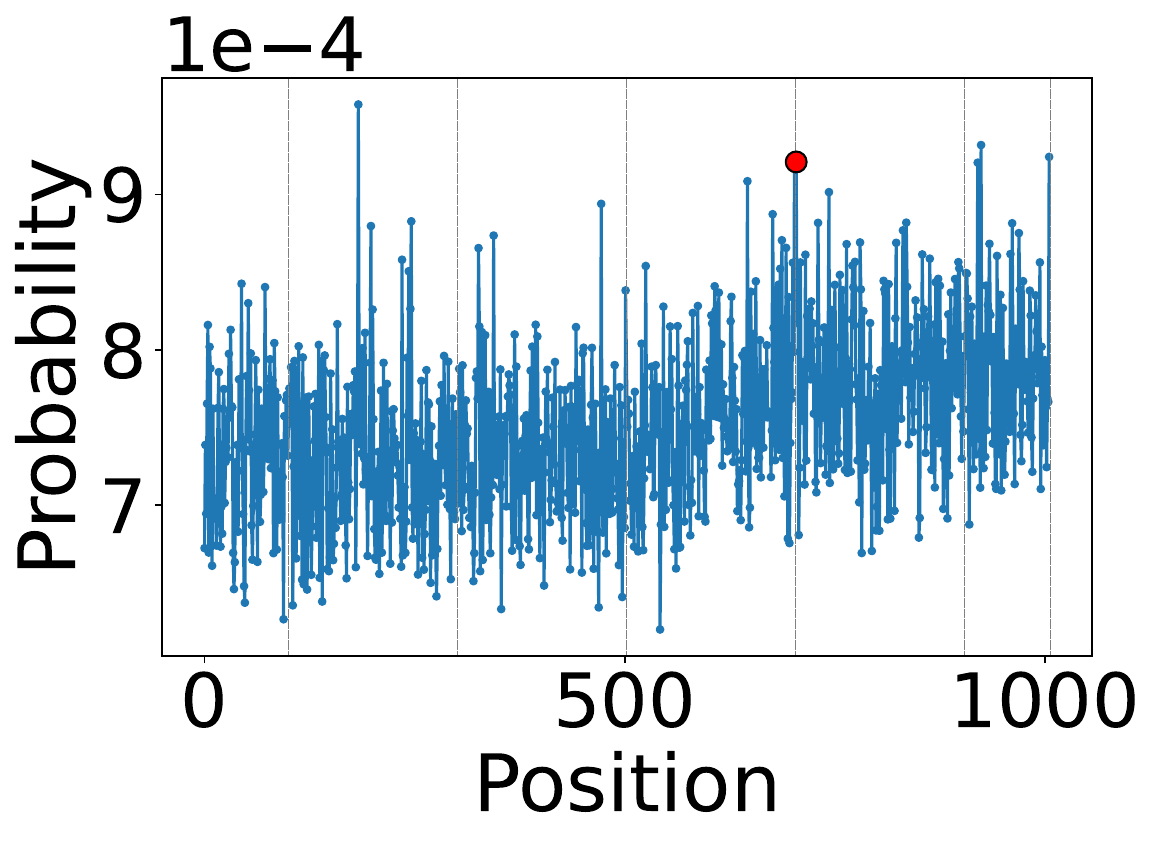} \\

& \FigGridVSubLabel{P5} &
\FigGridCell{Figures/ep_prob_without_A_red/Llama-3.1-8B-Instruct_5_Repeats_200_Length_500_Permutations_0_ablations_induction_5_nth.pdf} &
\FigGridCell{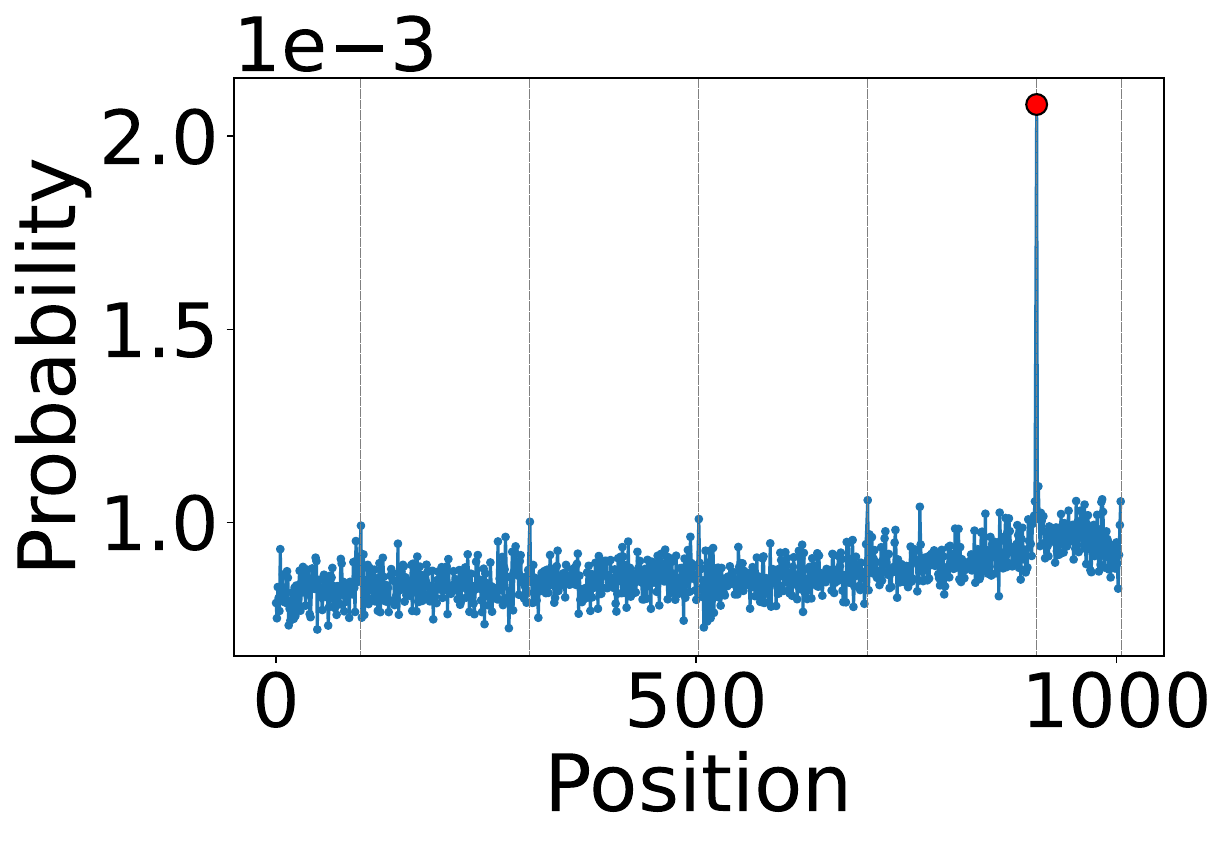} &
\FigGridCell{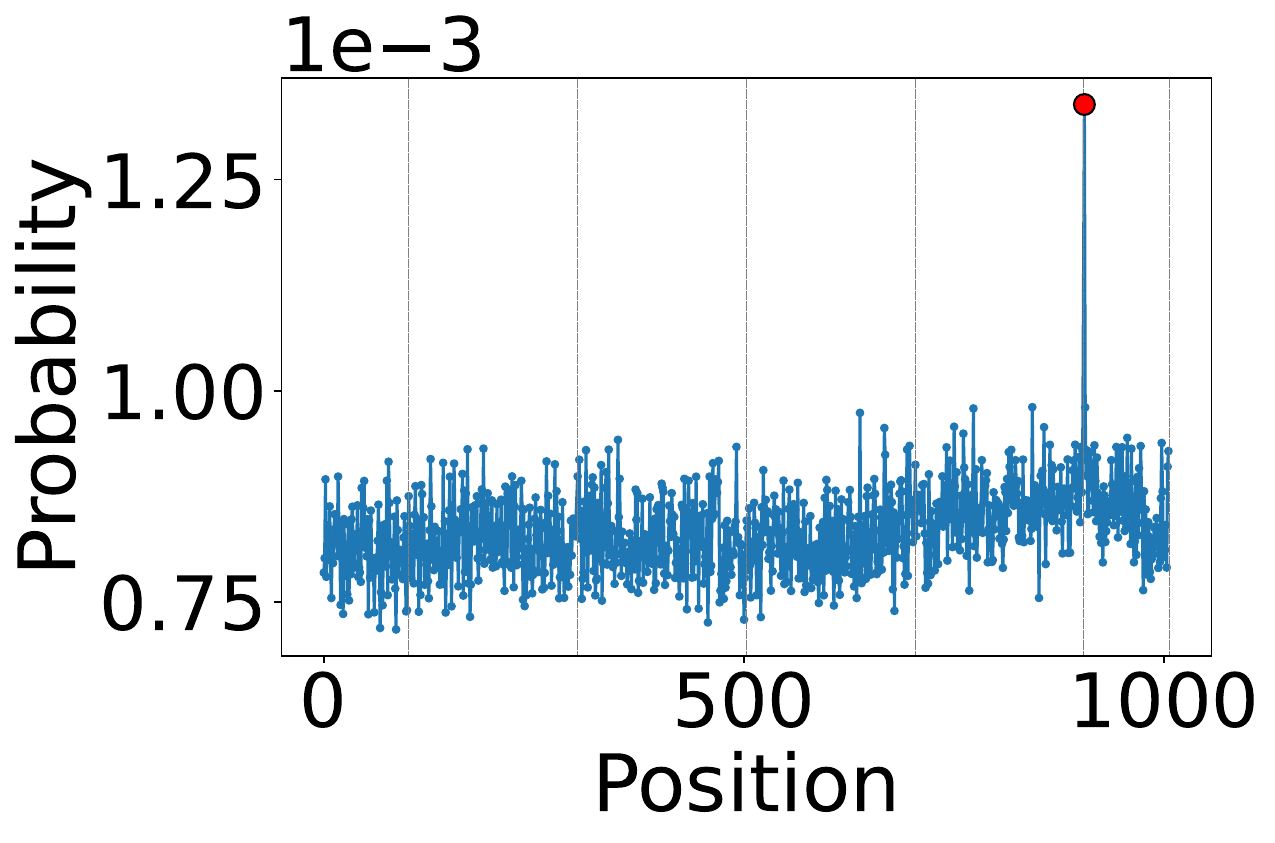} &
\FigGridCell{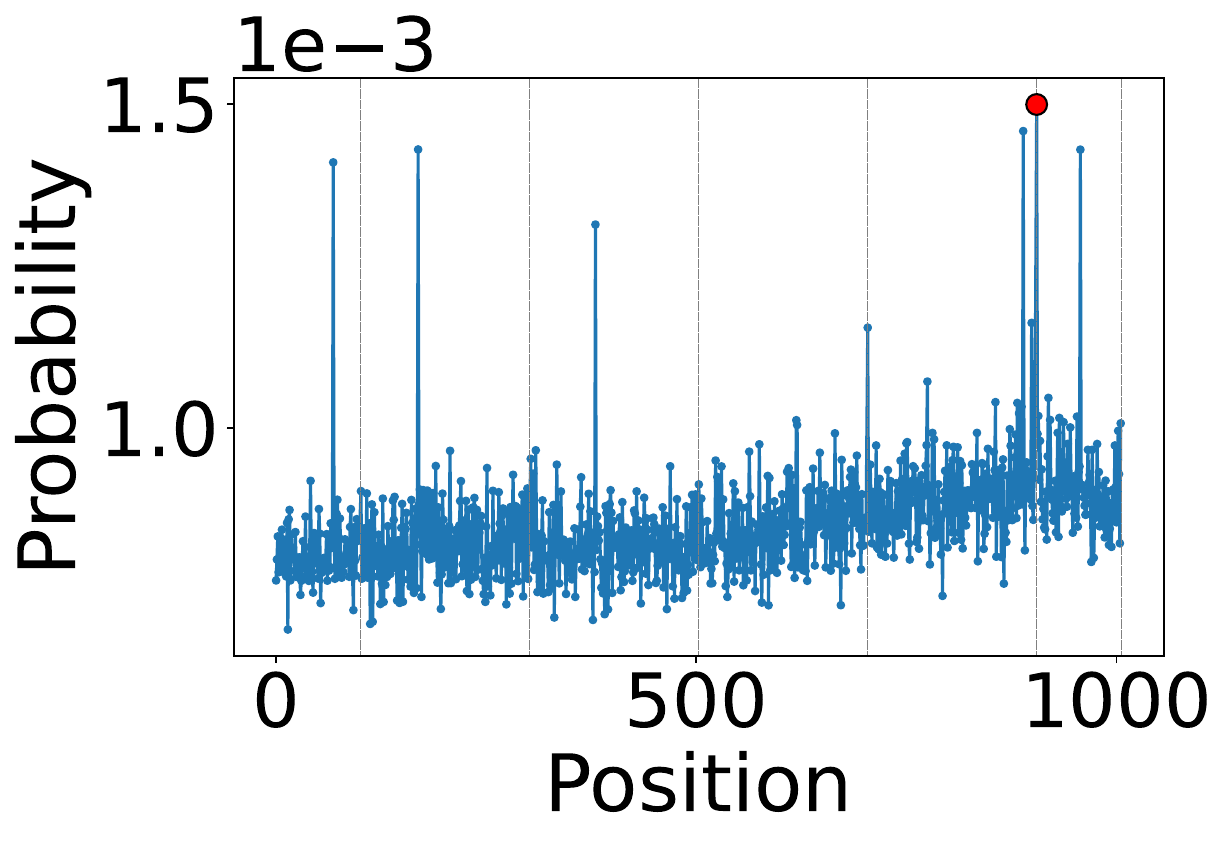} &
\FigGridCell{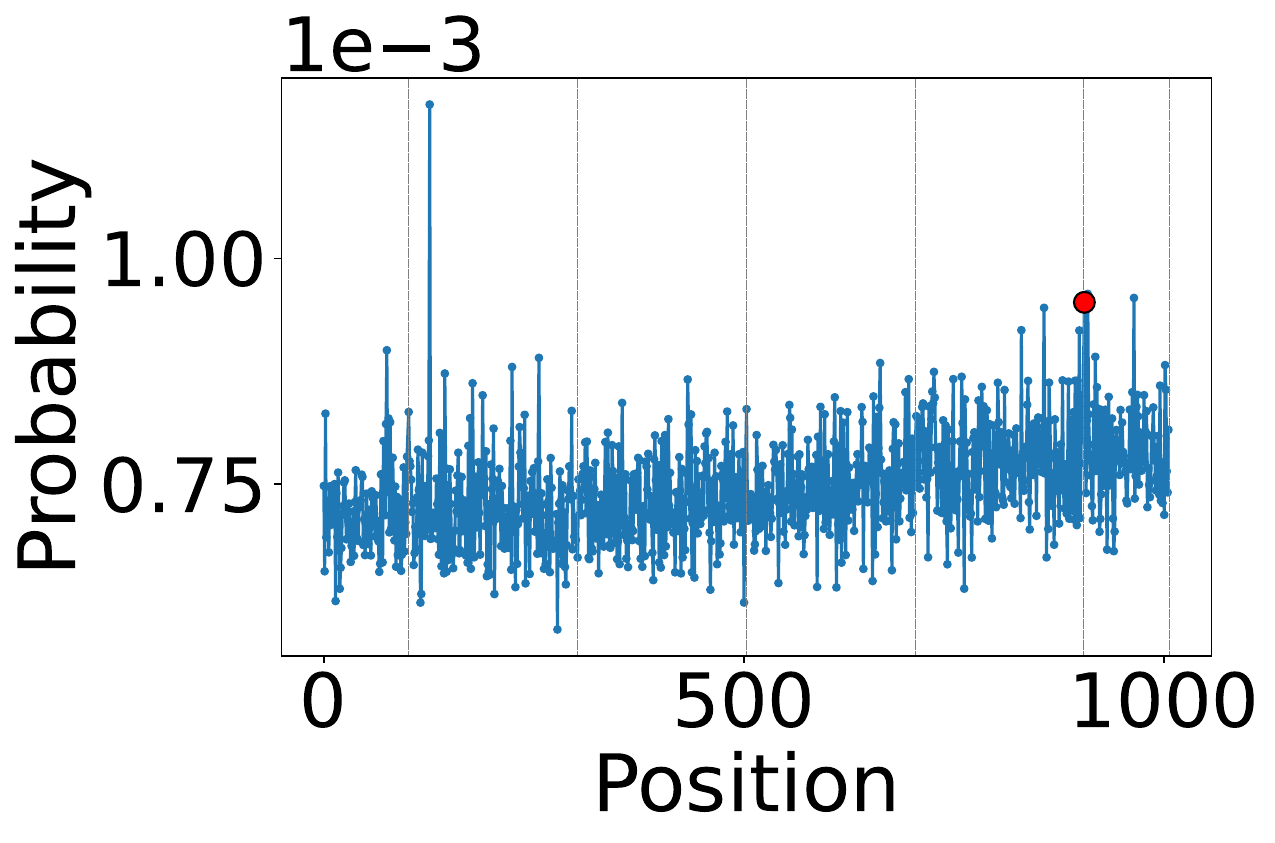} \\

\addlinespace[1pt]
\midrule
\addlinespace[1pt]

\multirow{5}{*}{\FigGridVLabelSpanFive{Random}} &
\FigGridVSubLabel{P1} &
\FigGridCell{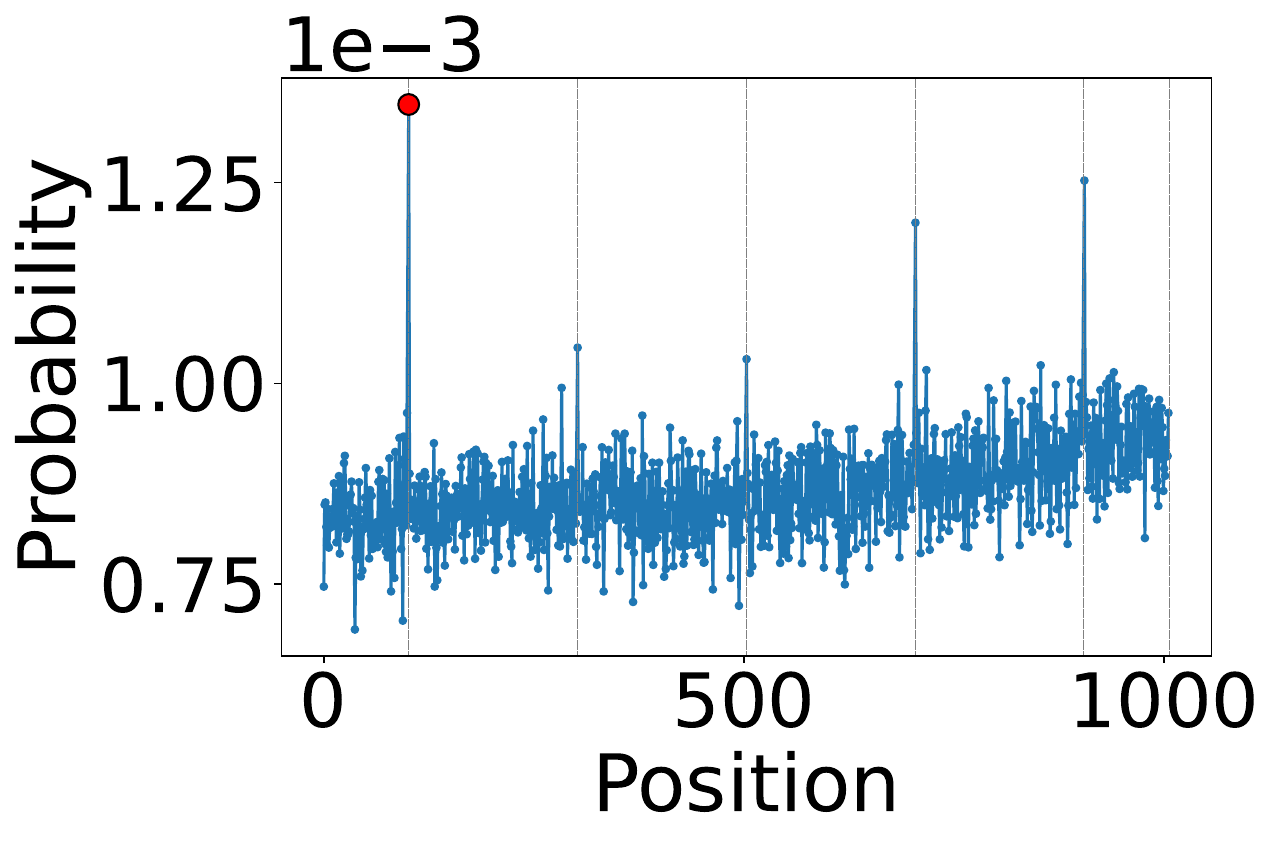} &
\FigGridCell{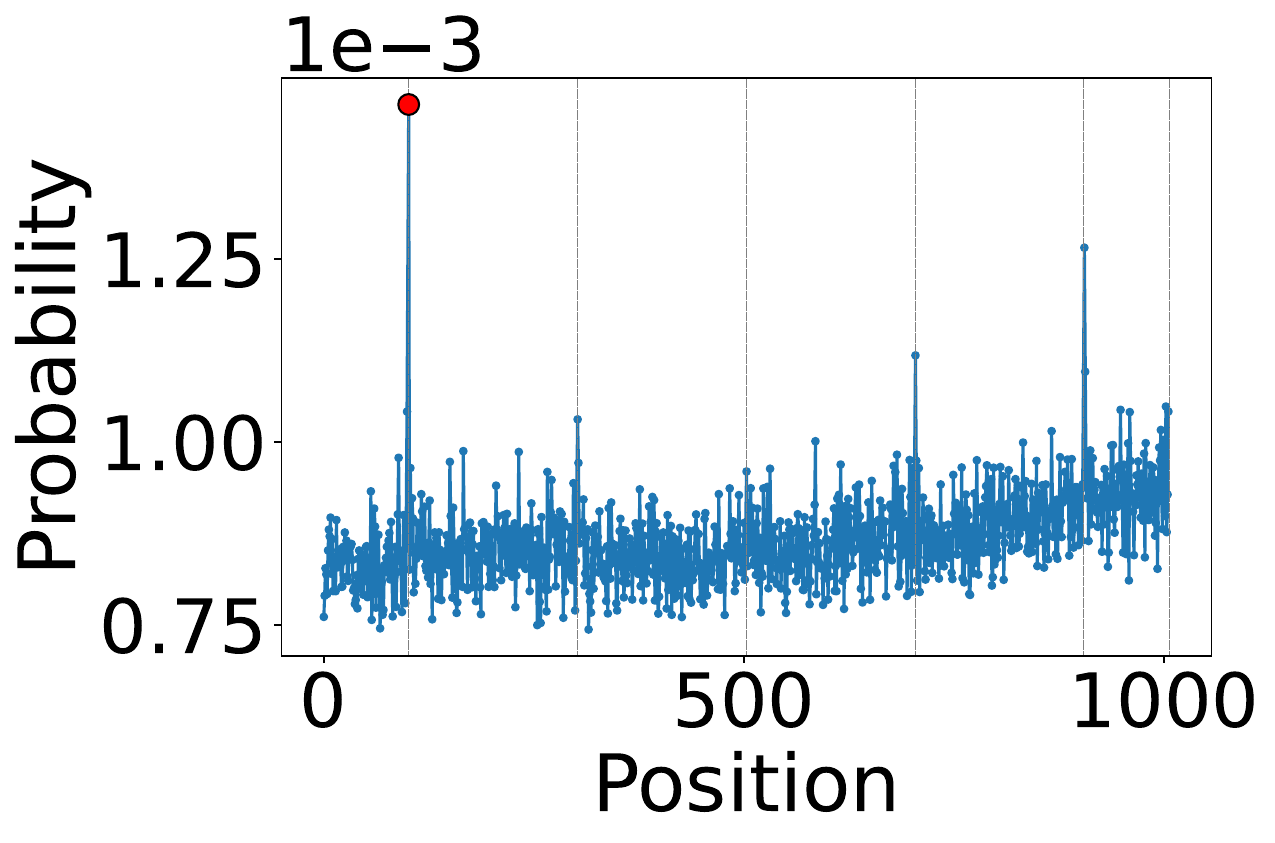} &
\FigGridCell{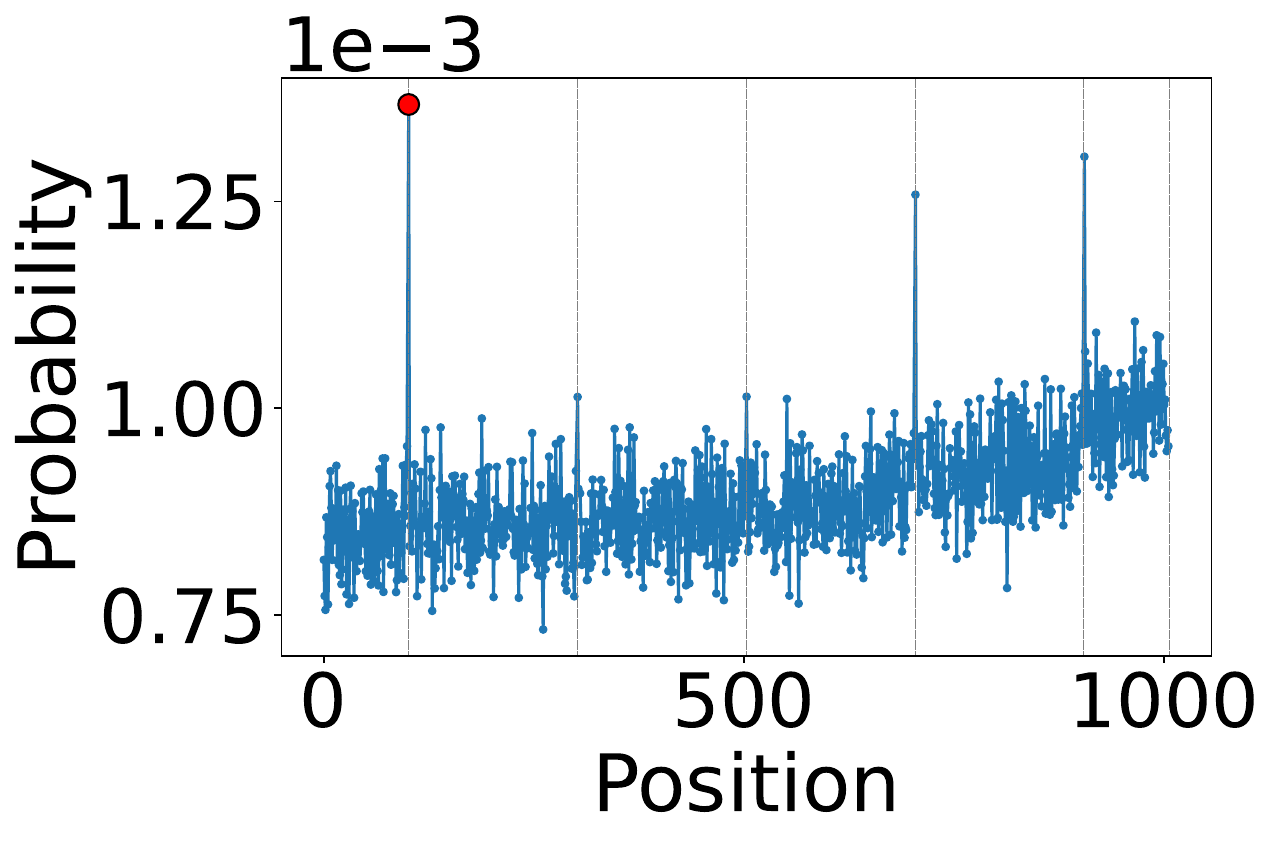} &
\FigGridCell{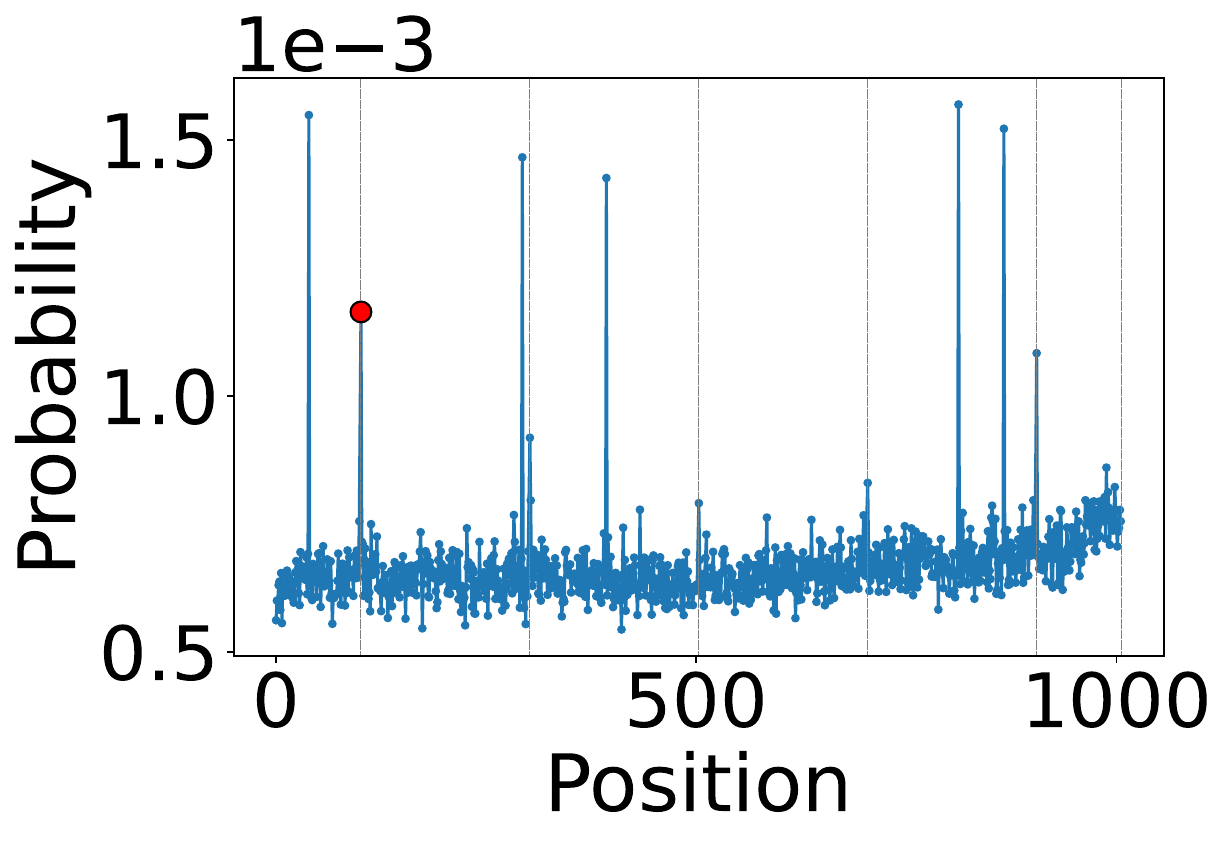} &
\FigGridCell{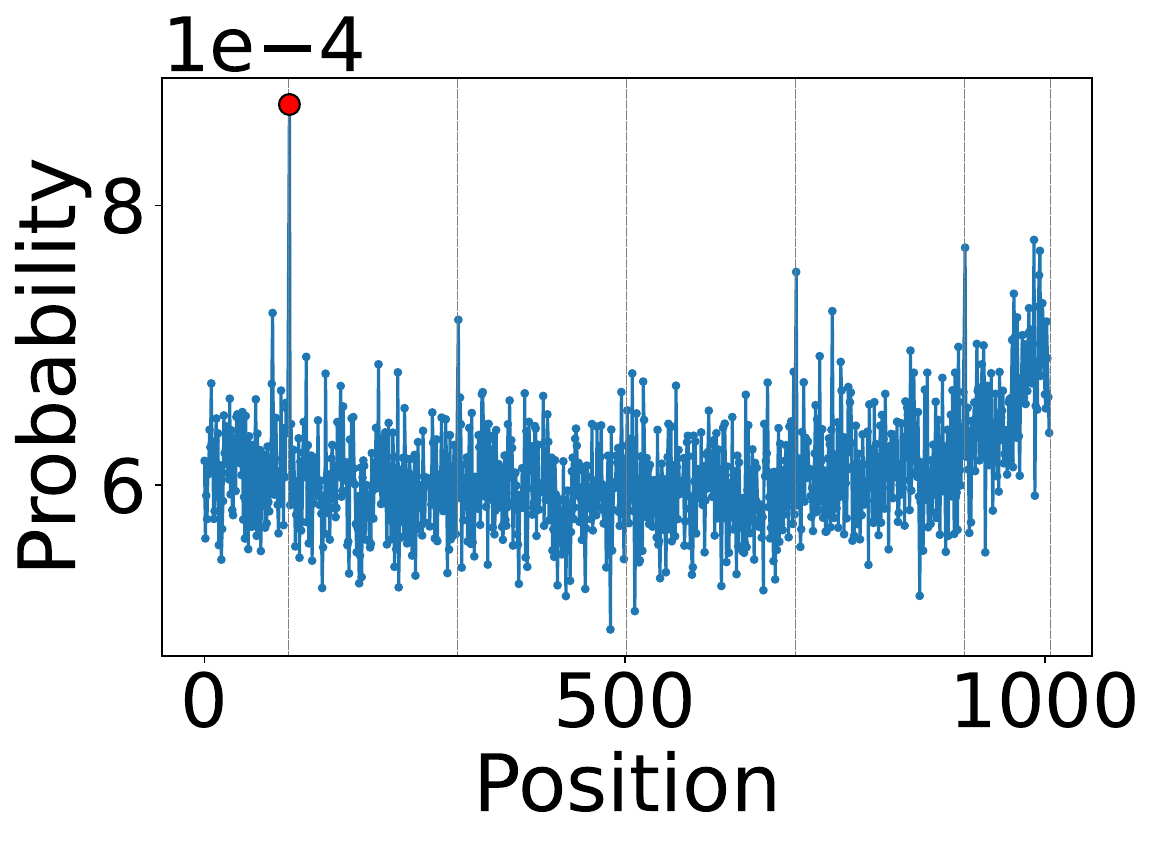} \\

& \FigGridVSubLabel{P2} &
\FigGridCell{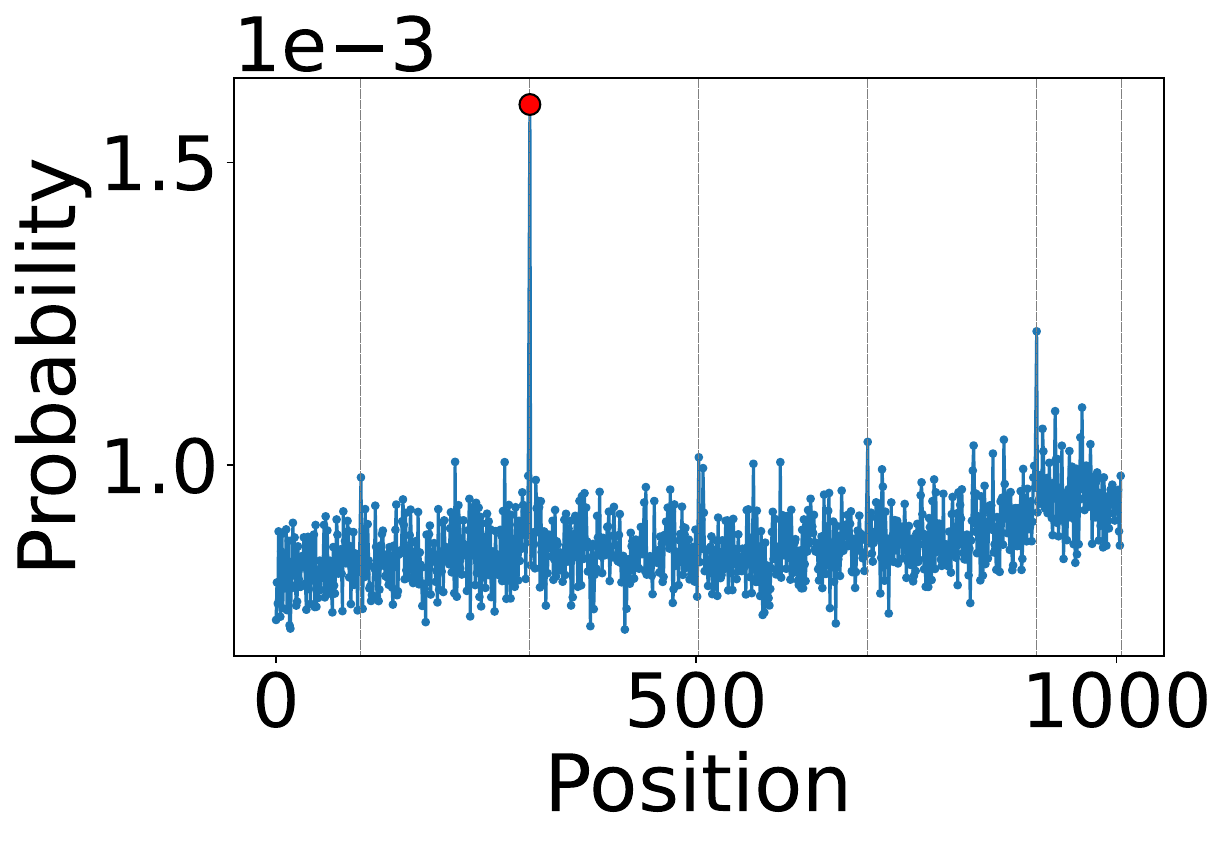} &
\FigGridCell{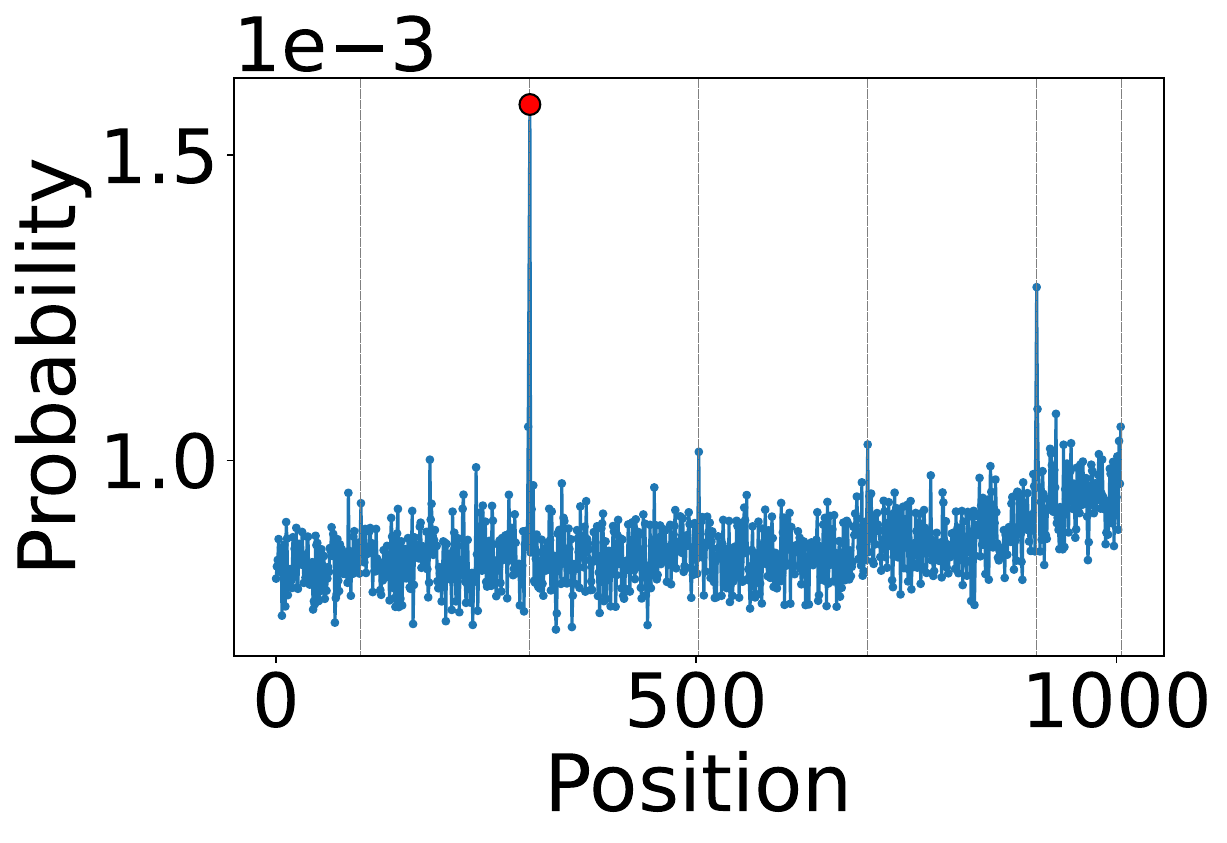} &
\FigGridCell{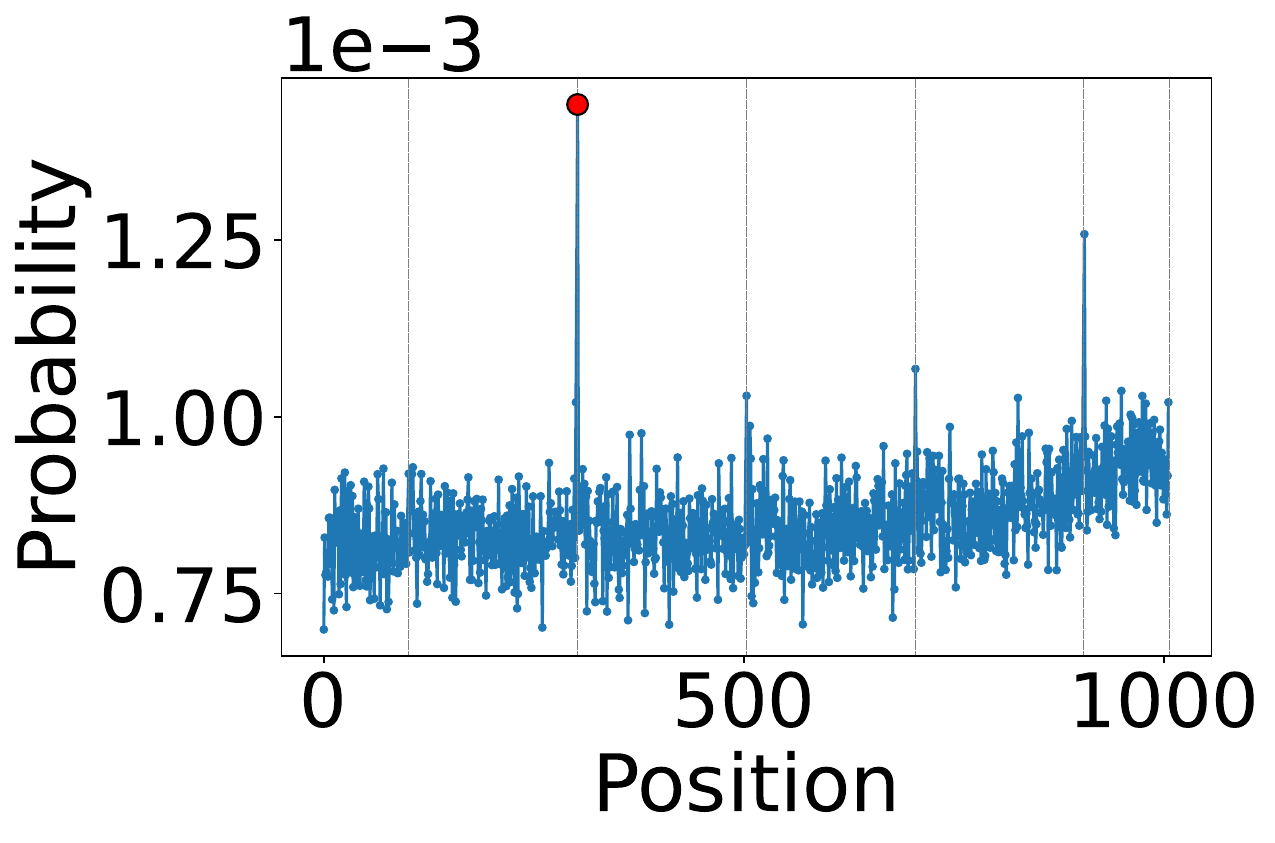} &
\FigGridCell{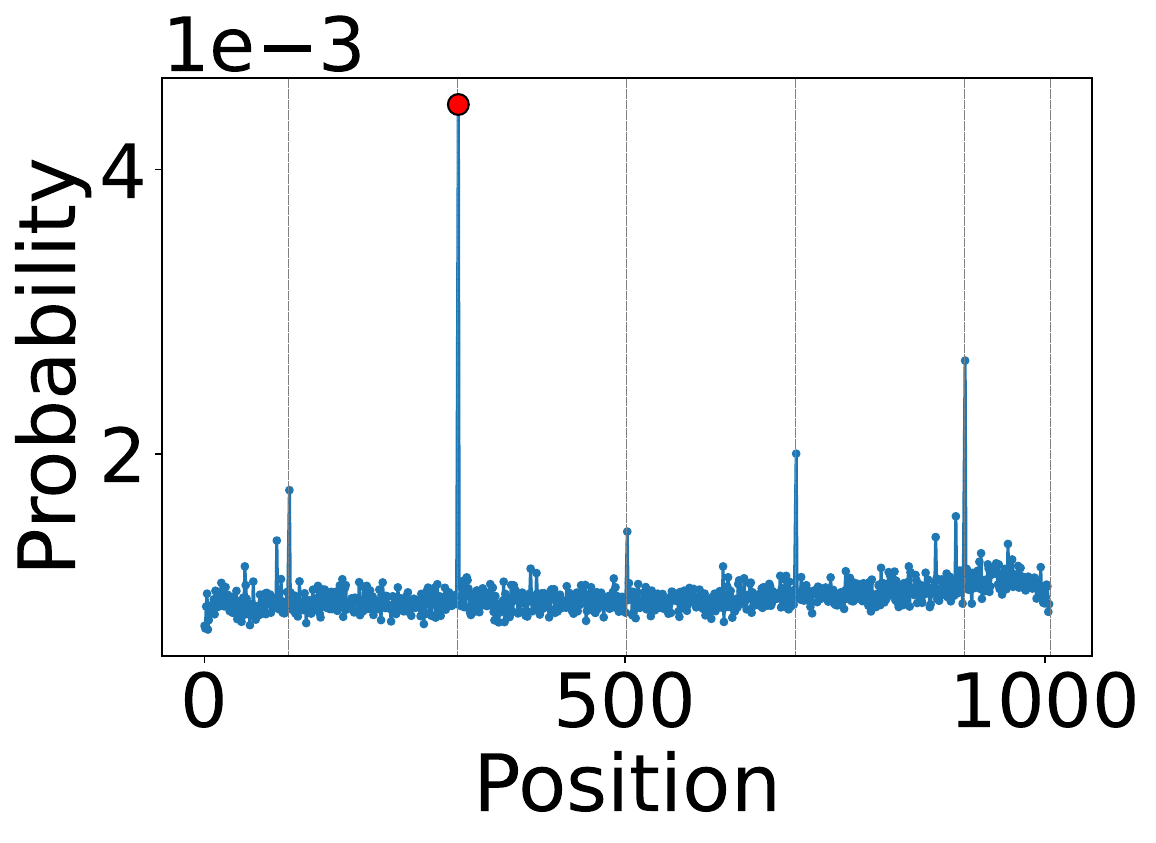} &
\FigGridCell{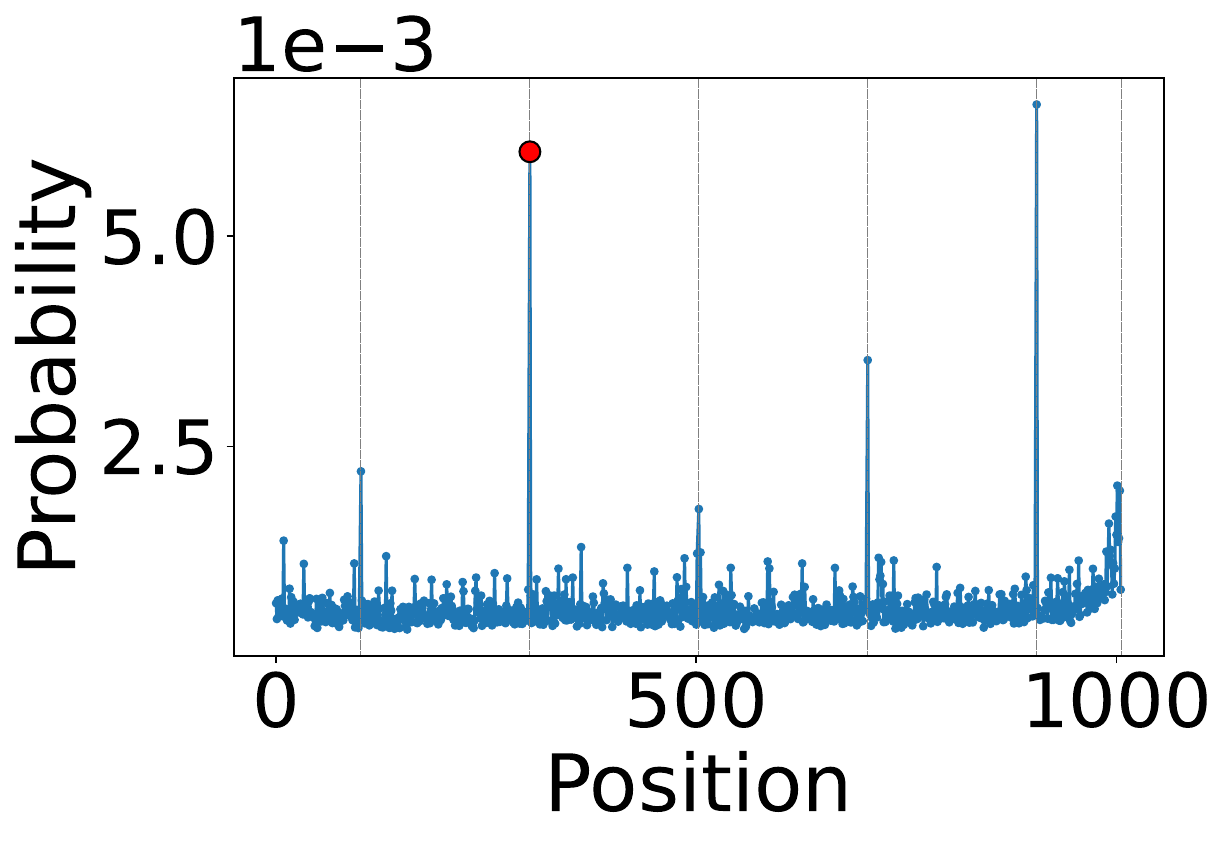} \\

& \FigGridVSubLabel{P3} &
\FigGridCell{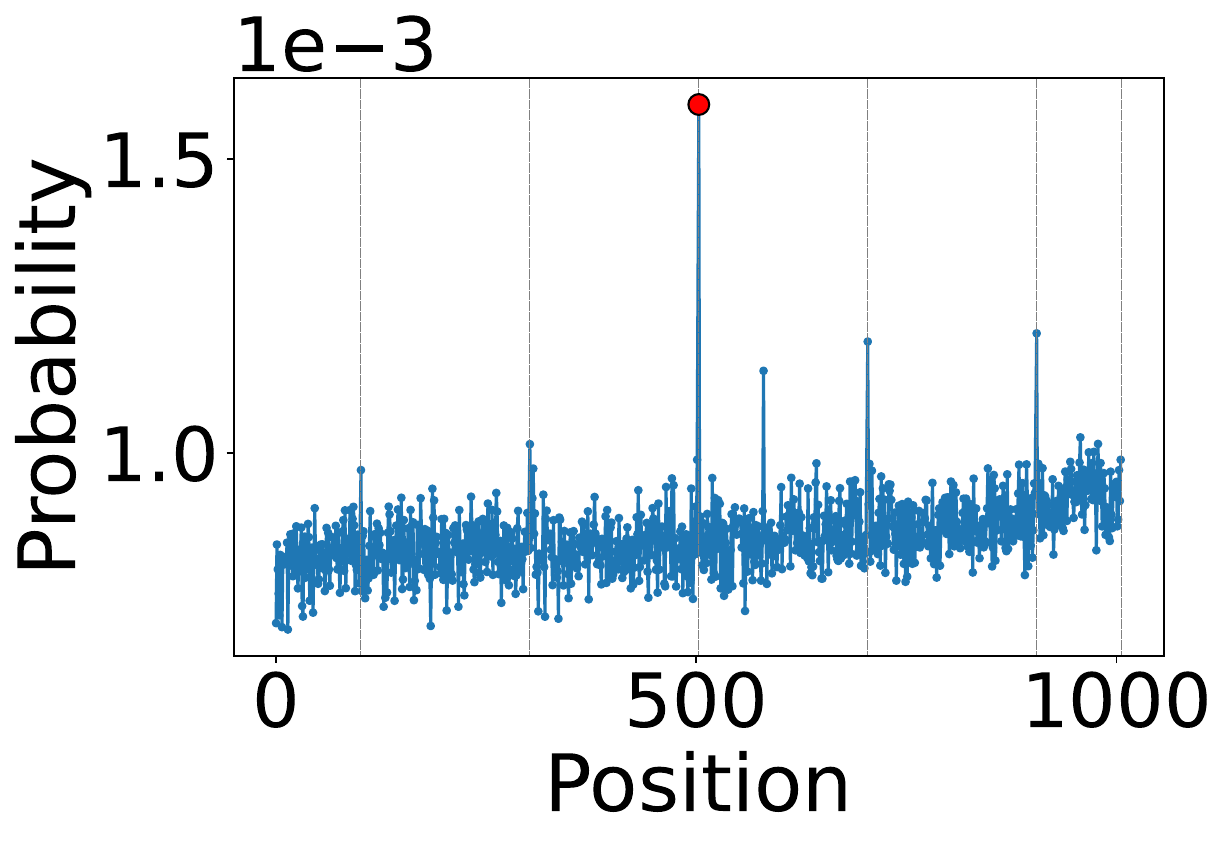} &
\FigGridCell{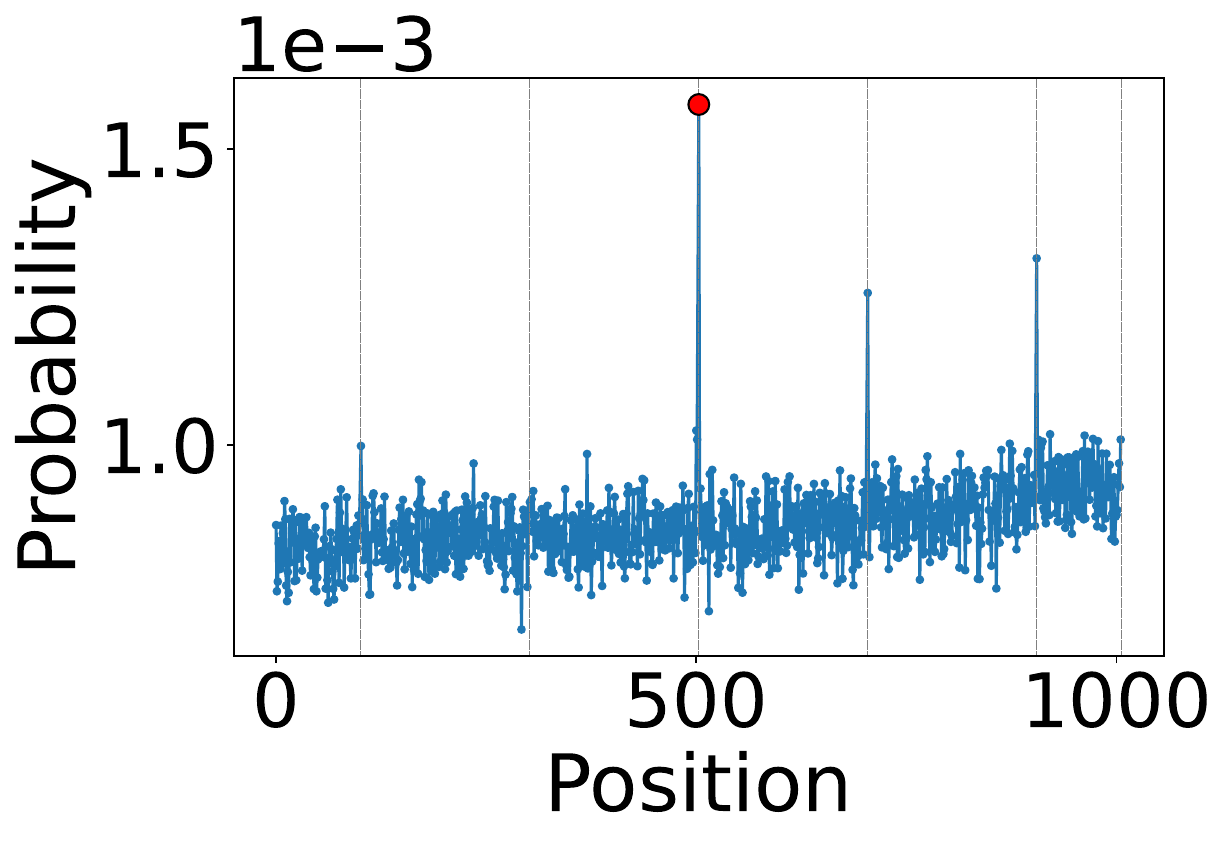} &
\FigGridCell{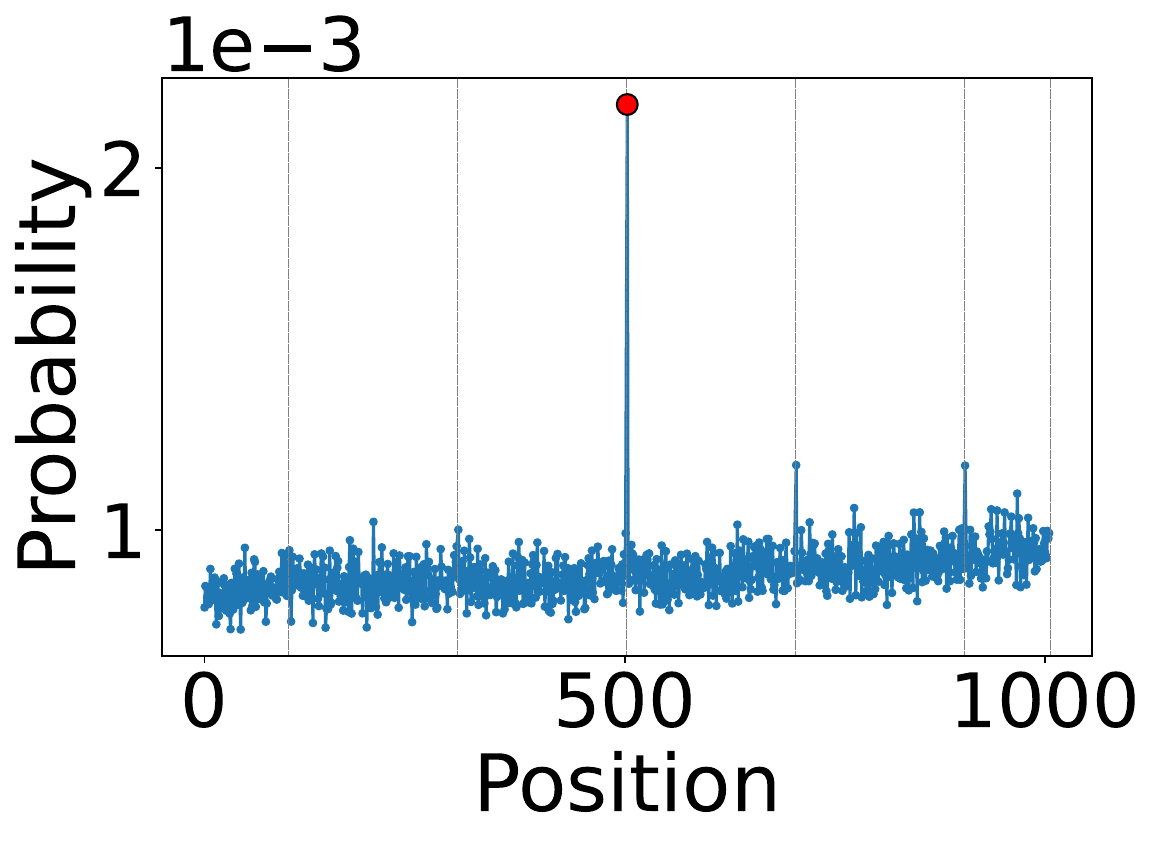} &
\FigGridCell{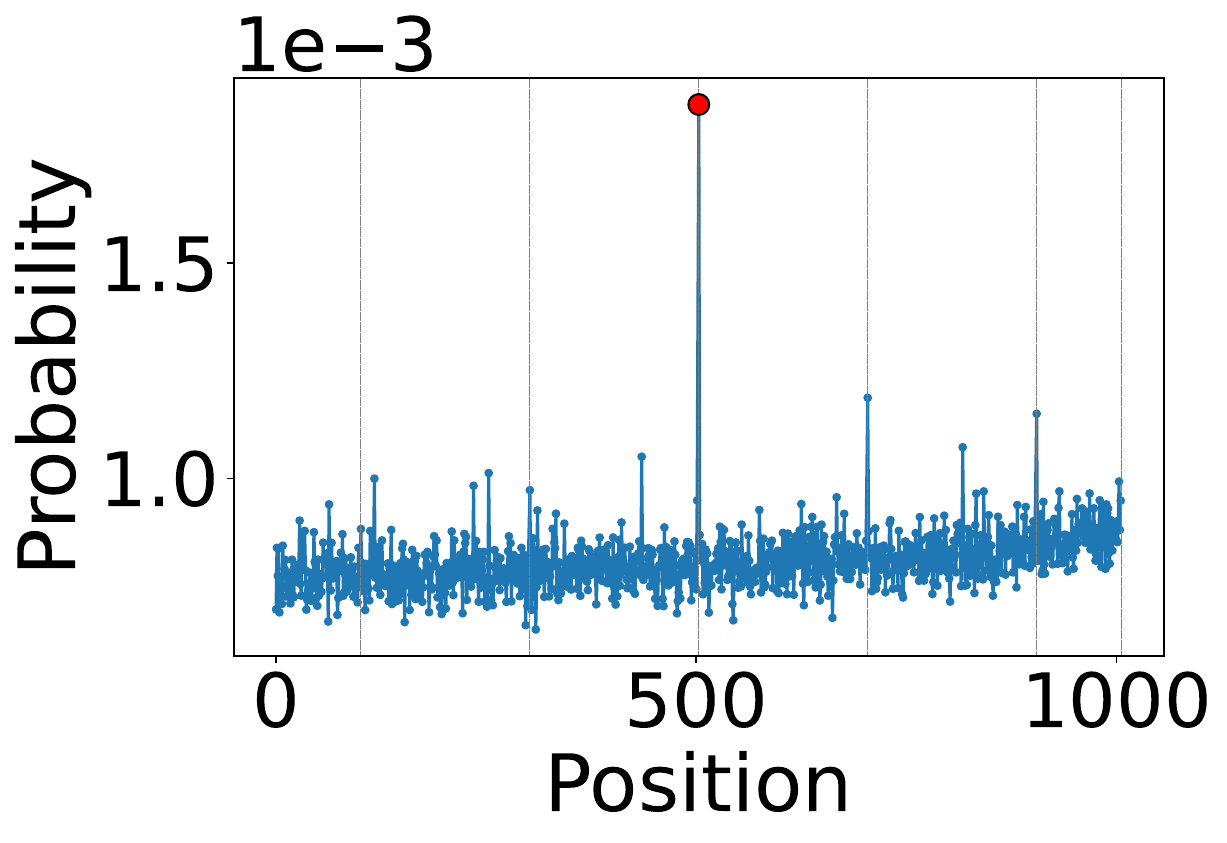} &
\FigGridCell{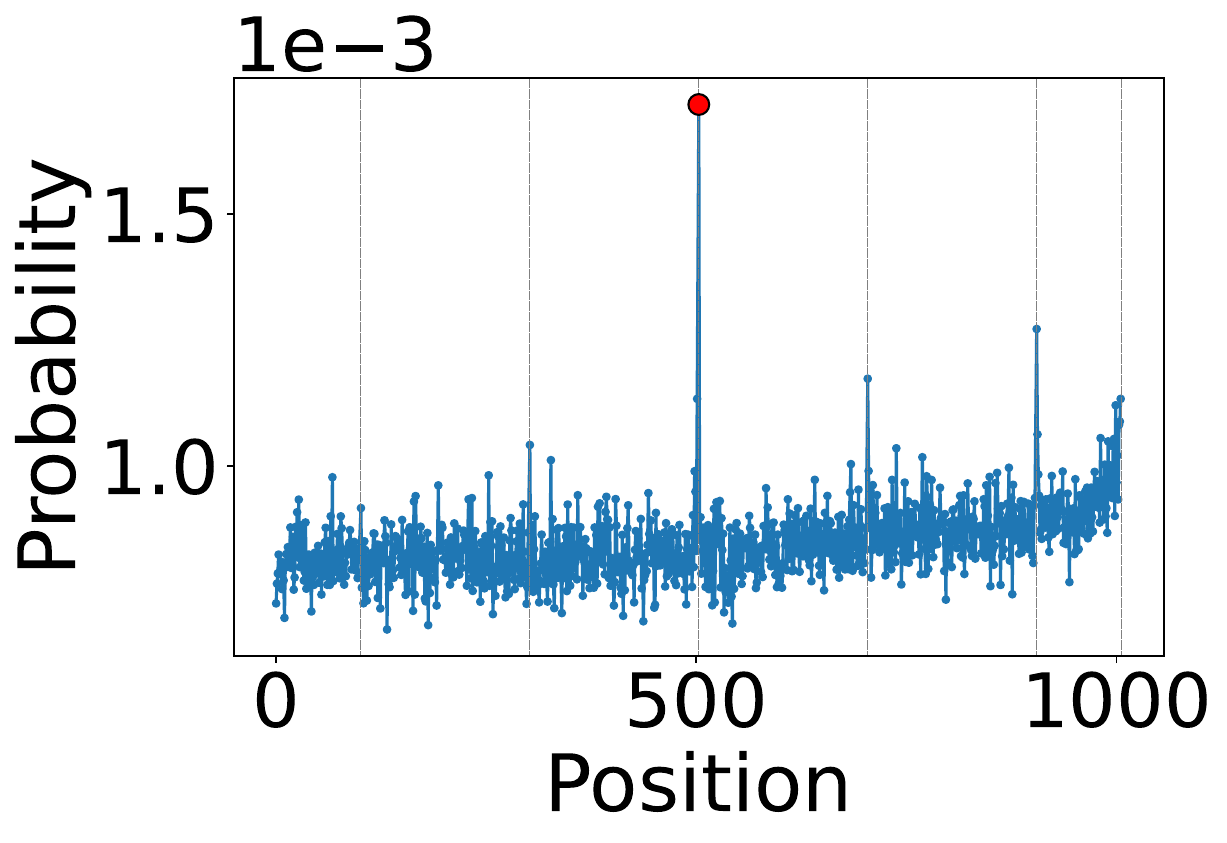} \\

& \FigGridVSubLabel{P4} &
\FigGridCell{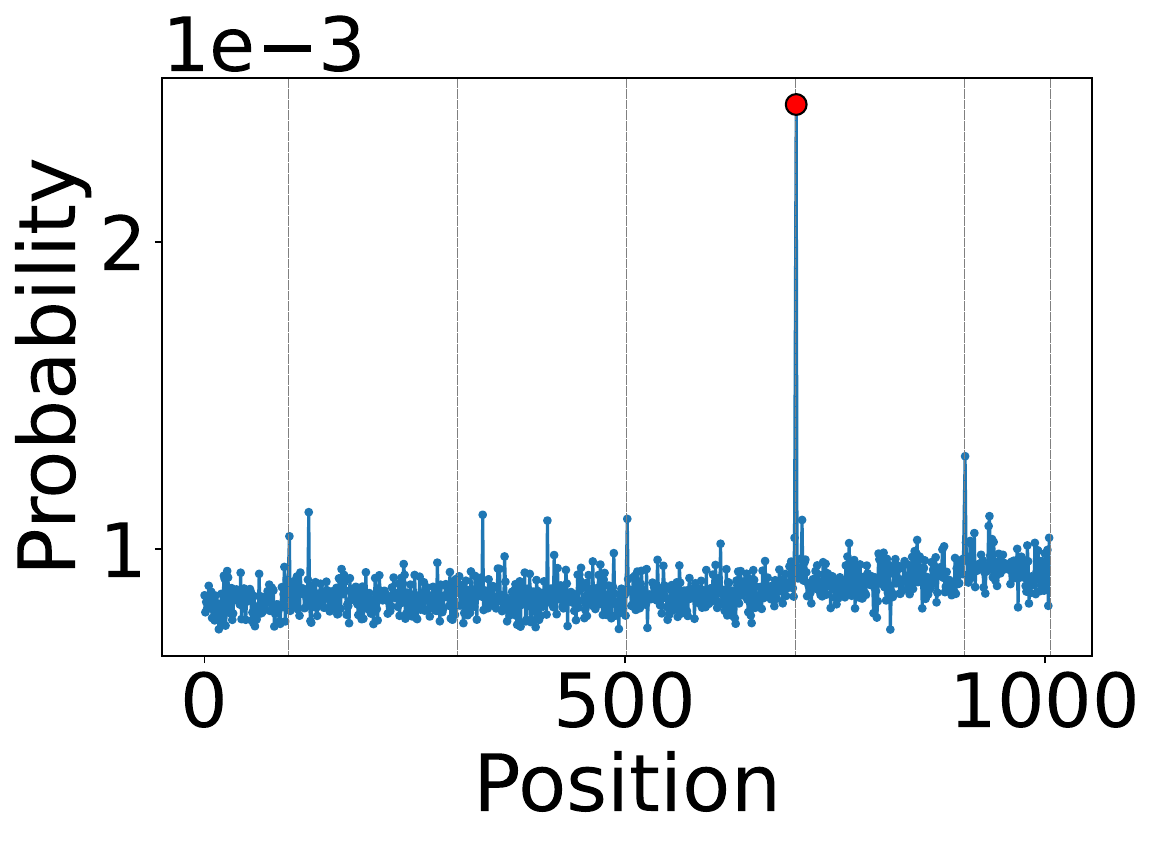} &
\FigGridCell{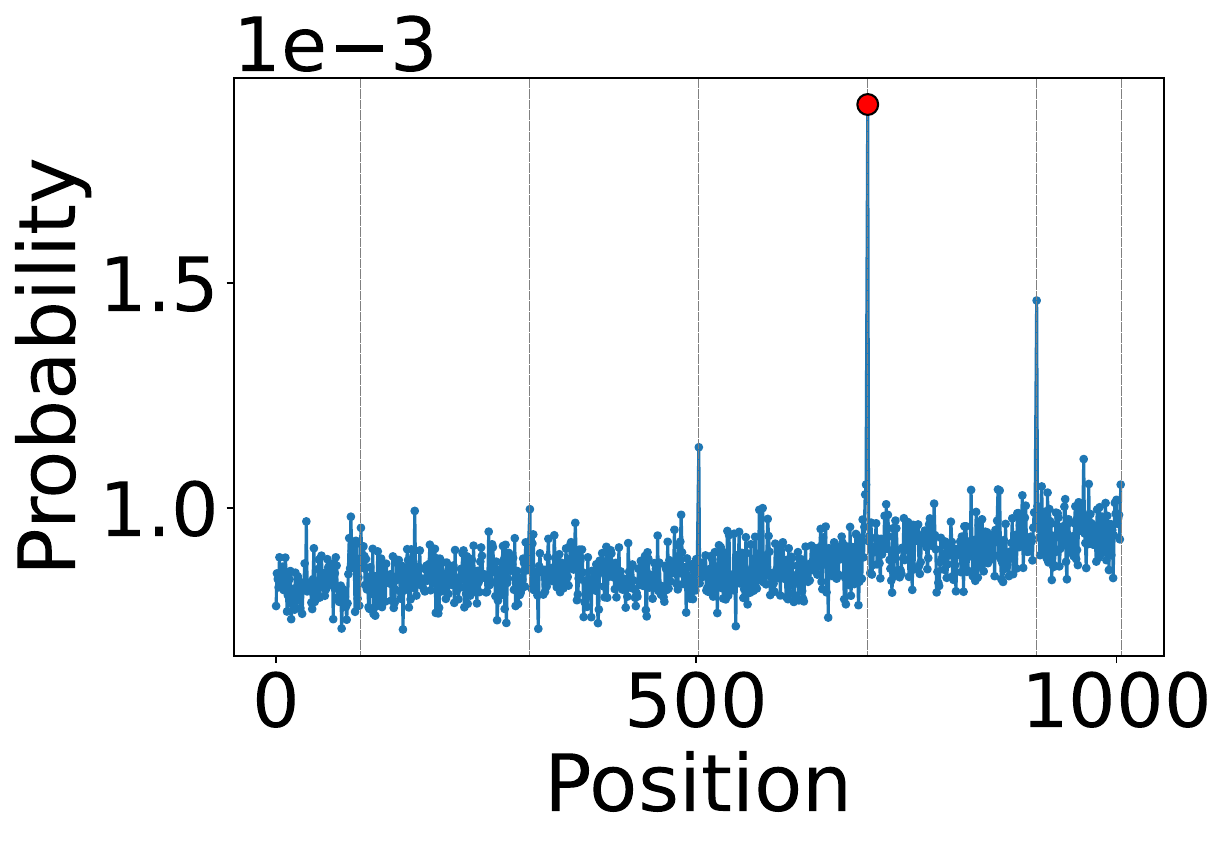} &
\FigGridCell{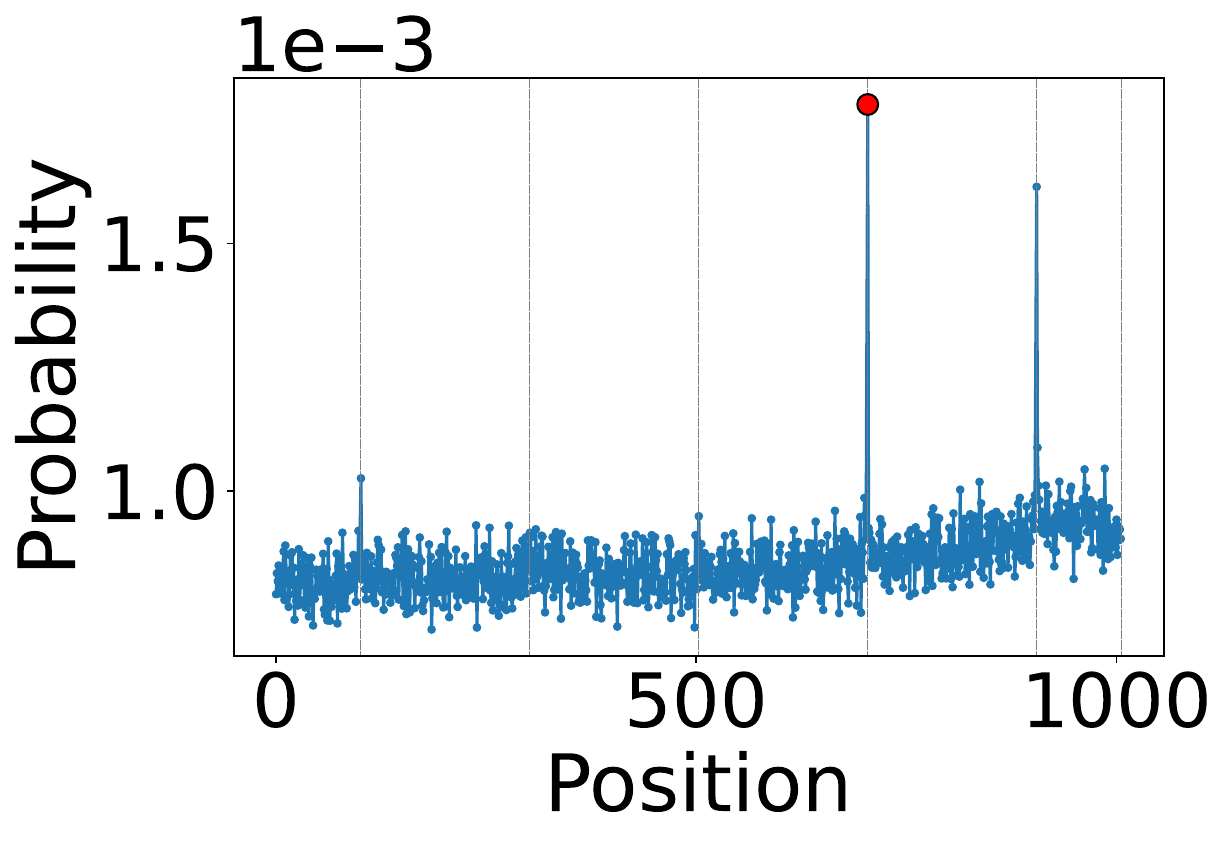} &
\FigGridCell{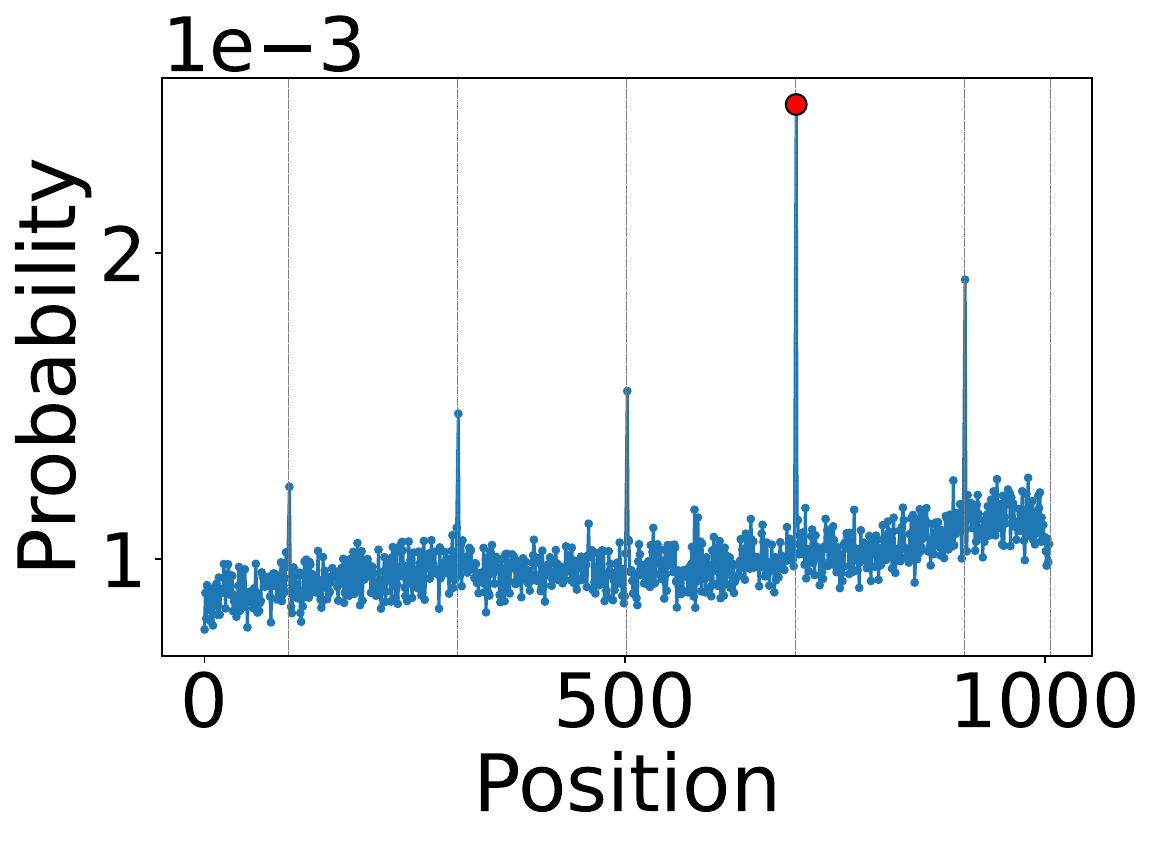} &
\FigGridCell{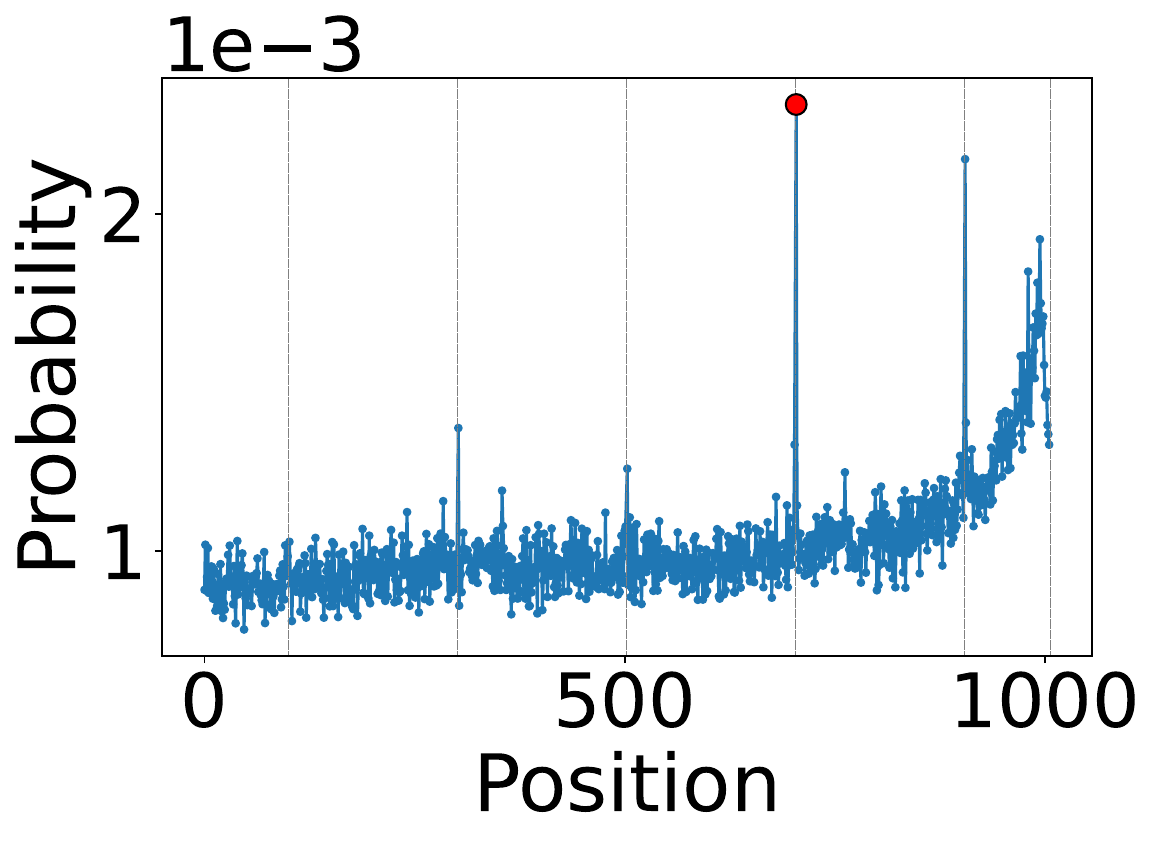} \\

& \FigGridVSubLabel{P5} &
\FigGridCell{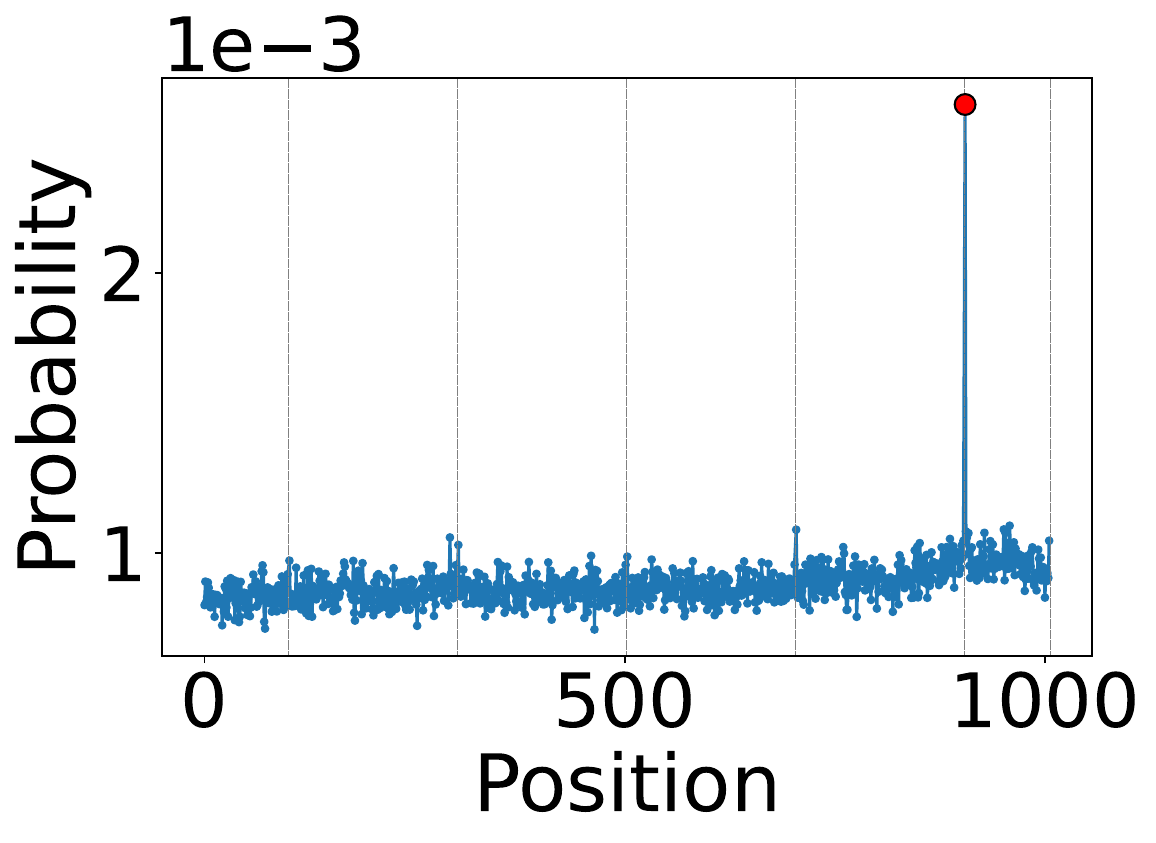} &
\FigGridCell{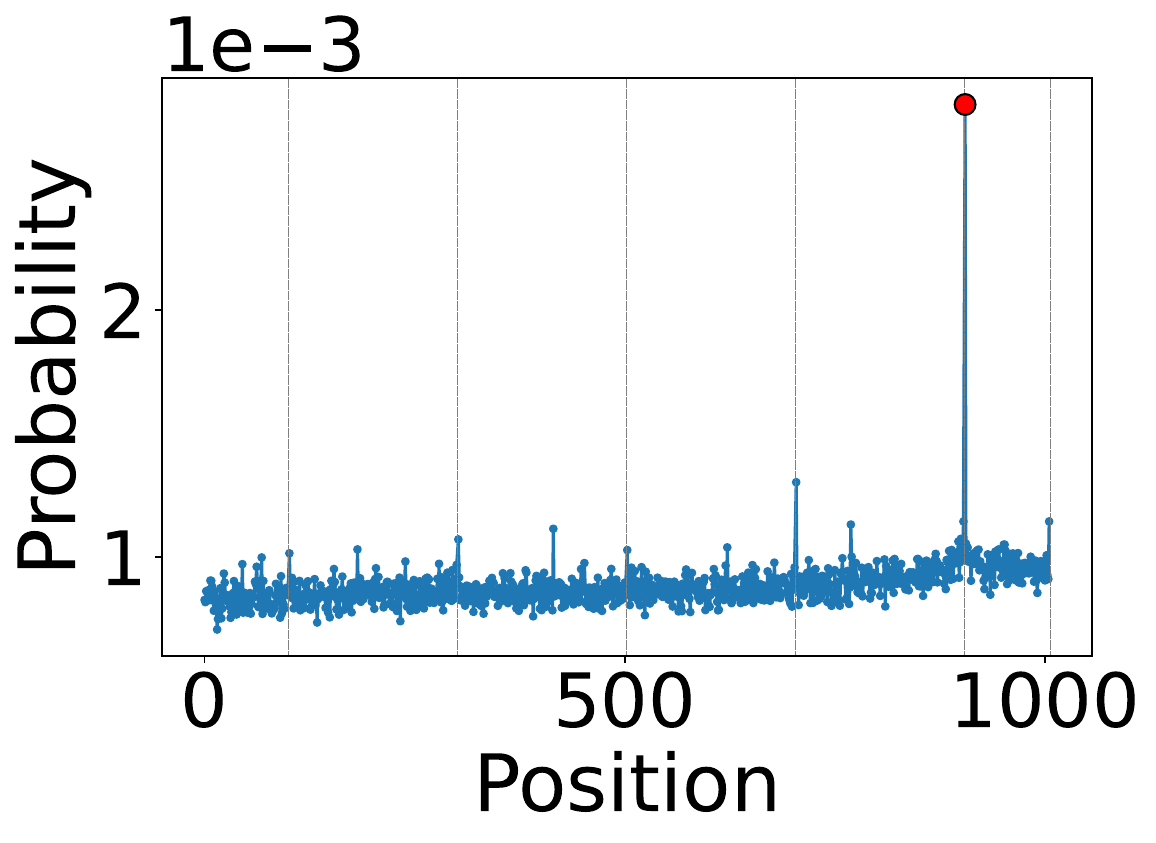} &
\FigGridCell{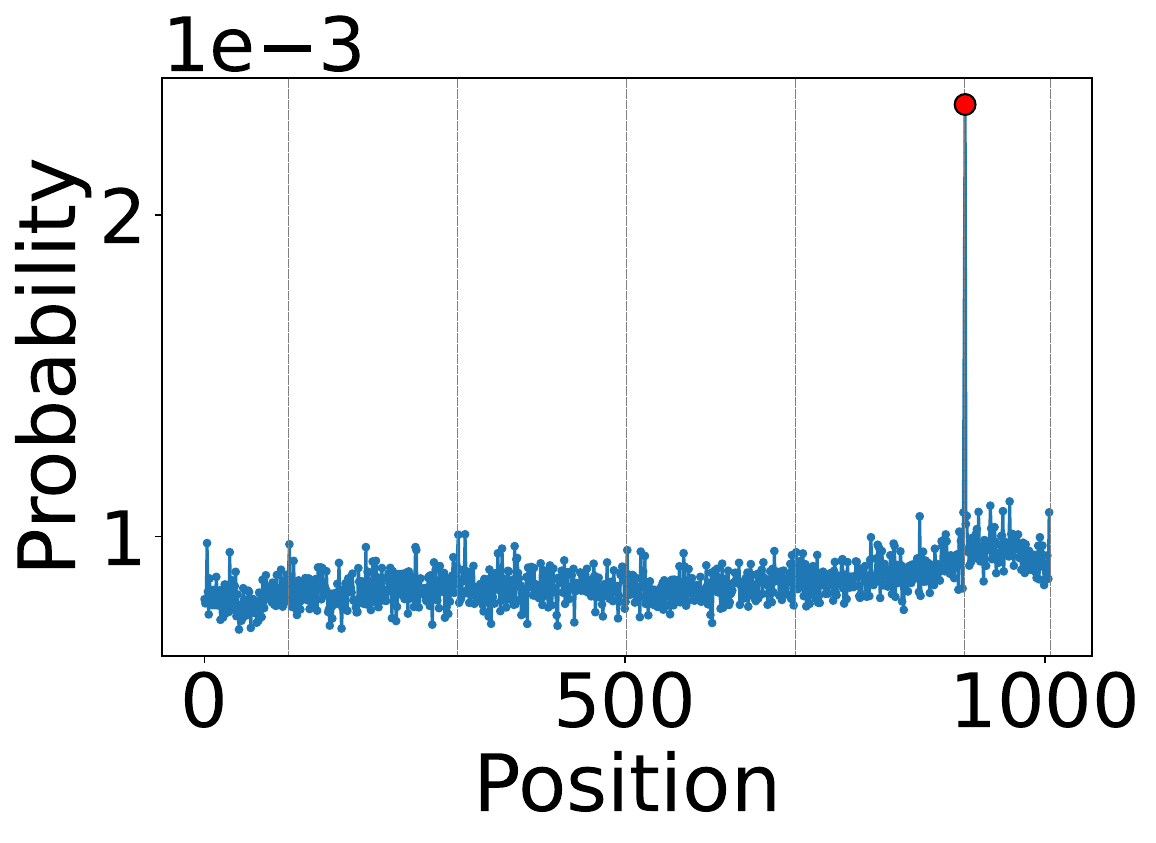} &
\FigGridCell{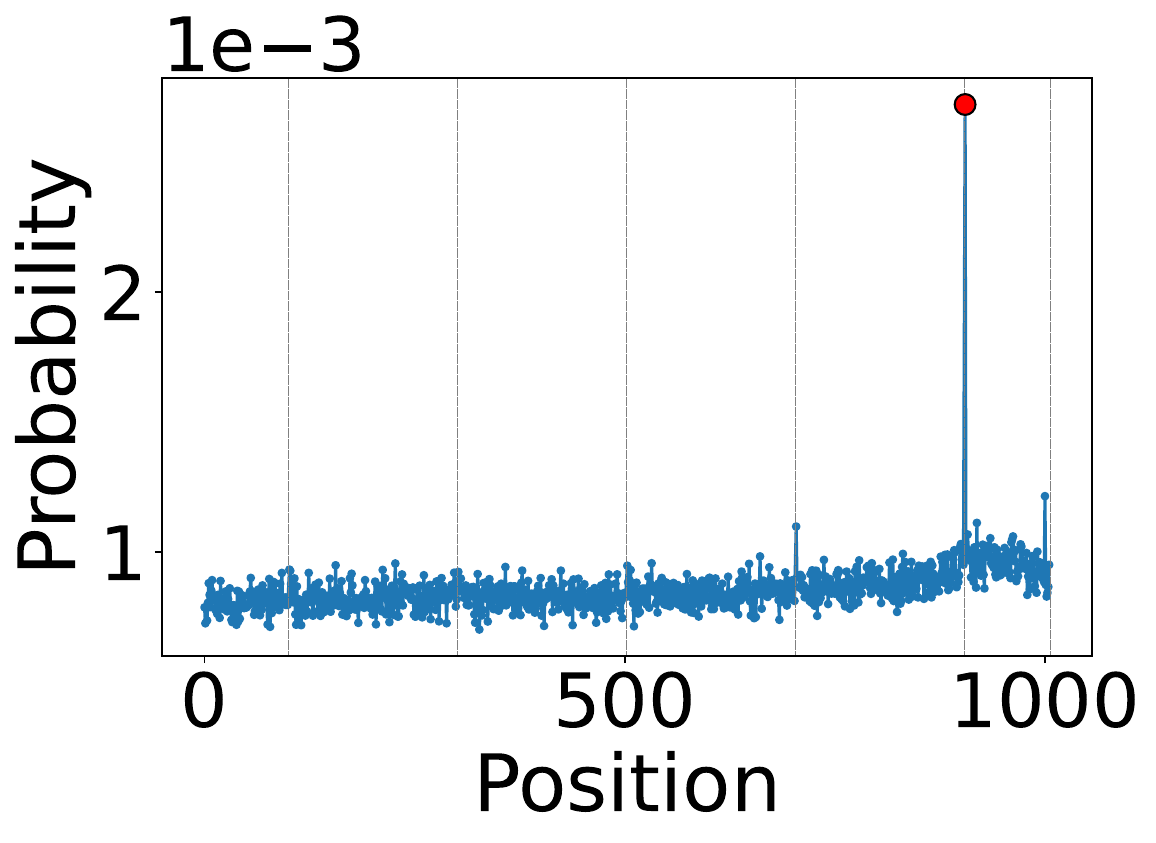} &
\FigGridCell{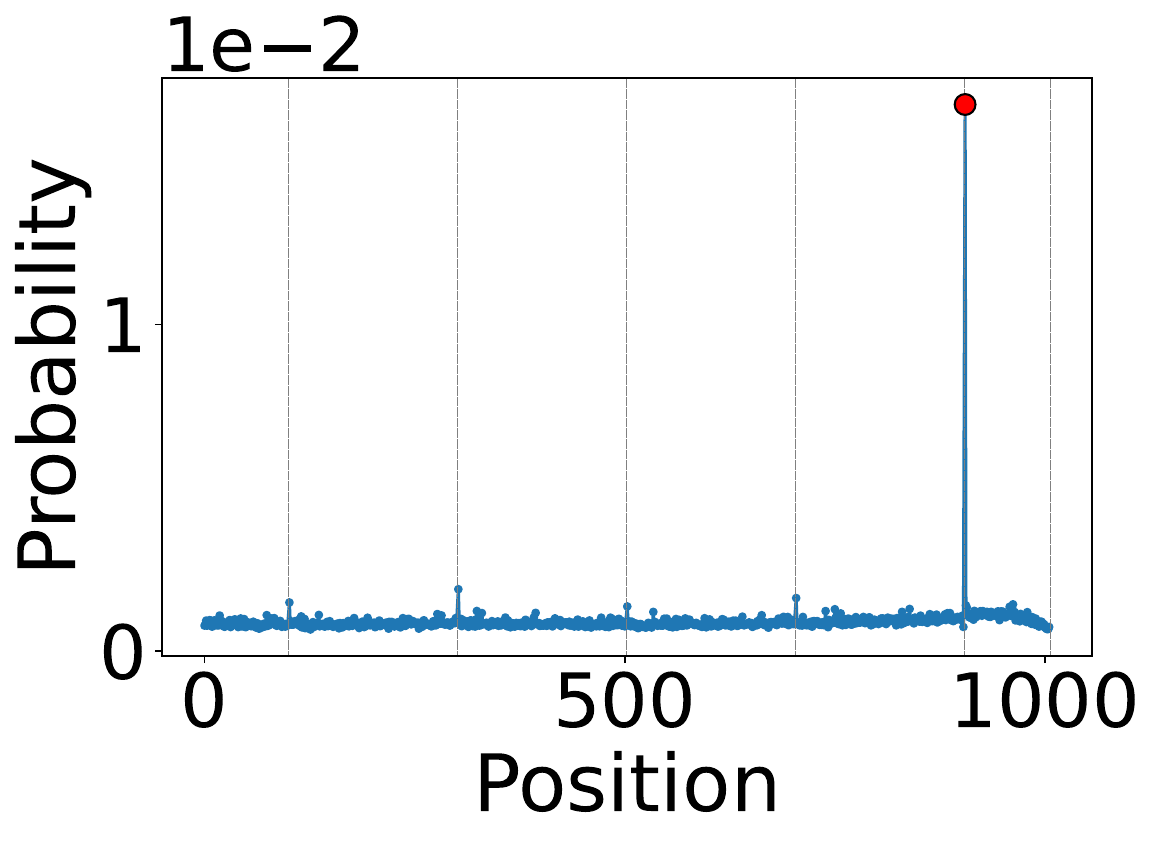} \\

\bottomrule
\end{tabular}
\endgroup
\caption{
Llama ablation effect (Exp. 2). Episodic retrieval probability after ablating Induction or Random heads (rows) probing different episode positions. Columns show number of ablated heads. Red dots mark the target token corresponding to the probed episode.
}
\label{fig:exp2_llama_ablation}
\end{figure*}

\begin{figure*}[h!]
\centering
\begingroup
\setlength{\FigGridLabelW}{1.05cm}
\setlength{\FigGridSubLabelW}{0.55cm}
\setlength{\FigGridCellH}{1.85cm} 
\setlength{\FigGridLabelSep}{1pt}
\setlength{\FigGridSubLabelSep}{1pt}
\setlength{\FigGridColSep}{0.5pt}

\FigGridSetupTwo{5}
\renewcommand{\arraystretch}{1.0}

\begin{tabular}{@{}>{\centering\arraybackslash}m{\FigGridLabelW}@{\hspace{\FigGridLabelSep}}
                >{\centering\arraybackslash}m{\FigGridSubLabelW}@{\hspace{\FigGridSubLabelSep}}
                *{5}{>{\centering\arraybackslash}m{\FigGridPlotW}}@{}}
\toprule
& & \multicolumn{5}{c}{\FigHdr Ablations} \\
\cmidrule(lr){3-7}
& & \FigHdr 0 & \FigHdr 1 & \FigHdr 10 & \FigHdr 50 & \FigHdr 100 \\
\midrule
\multirow{5}{*}{\FigGridVLabelSpanFive{Induction}} &
\FigGridVSubLabel{P1} &
\FigGridCell{Figures/ep_prob_without_A_red/Mistral-7B-Instruct-v0.1_5_Repeats_200_Length_500_Permutations_0_ablations_induction_1_nth.pdf} &
\FigGridCell{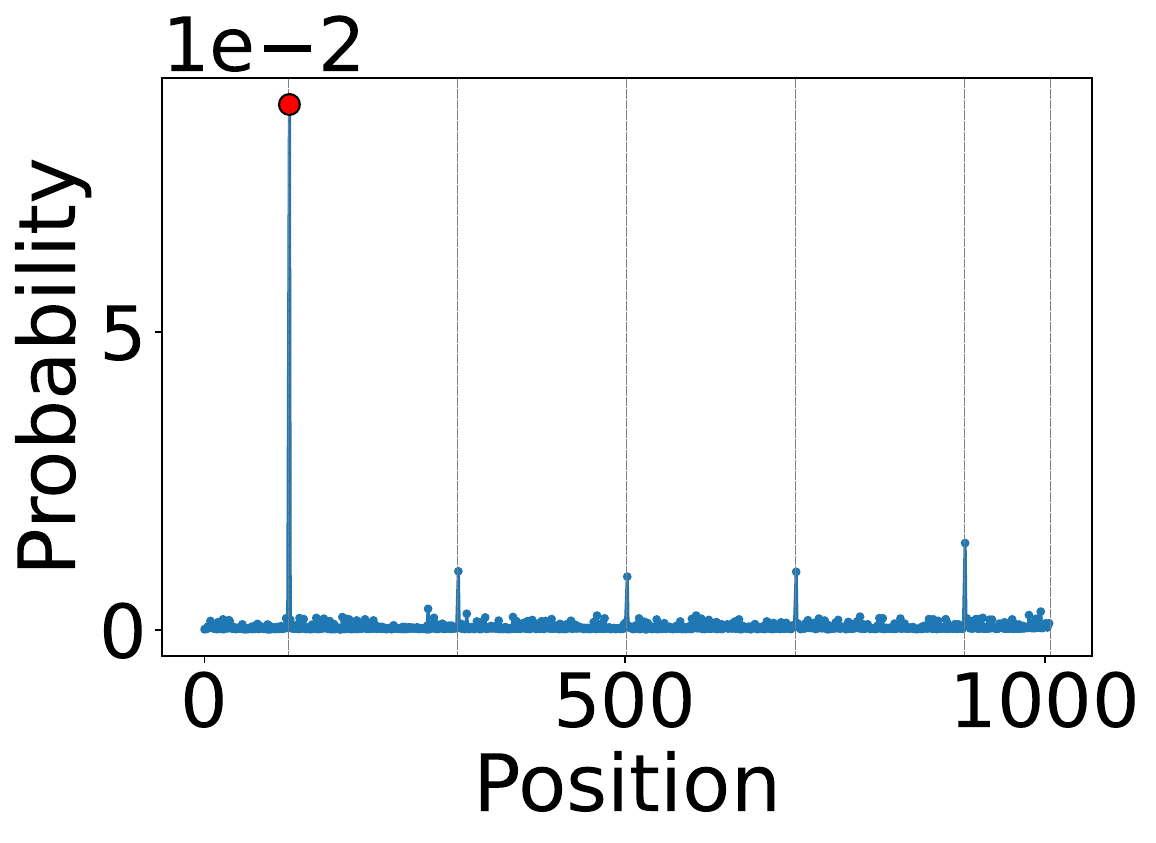} &
\FigGridCell{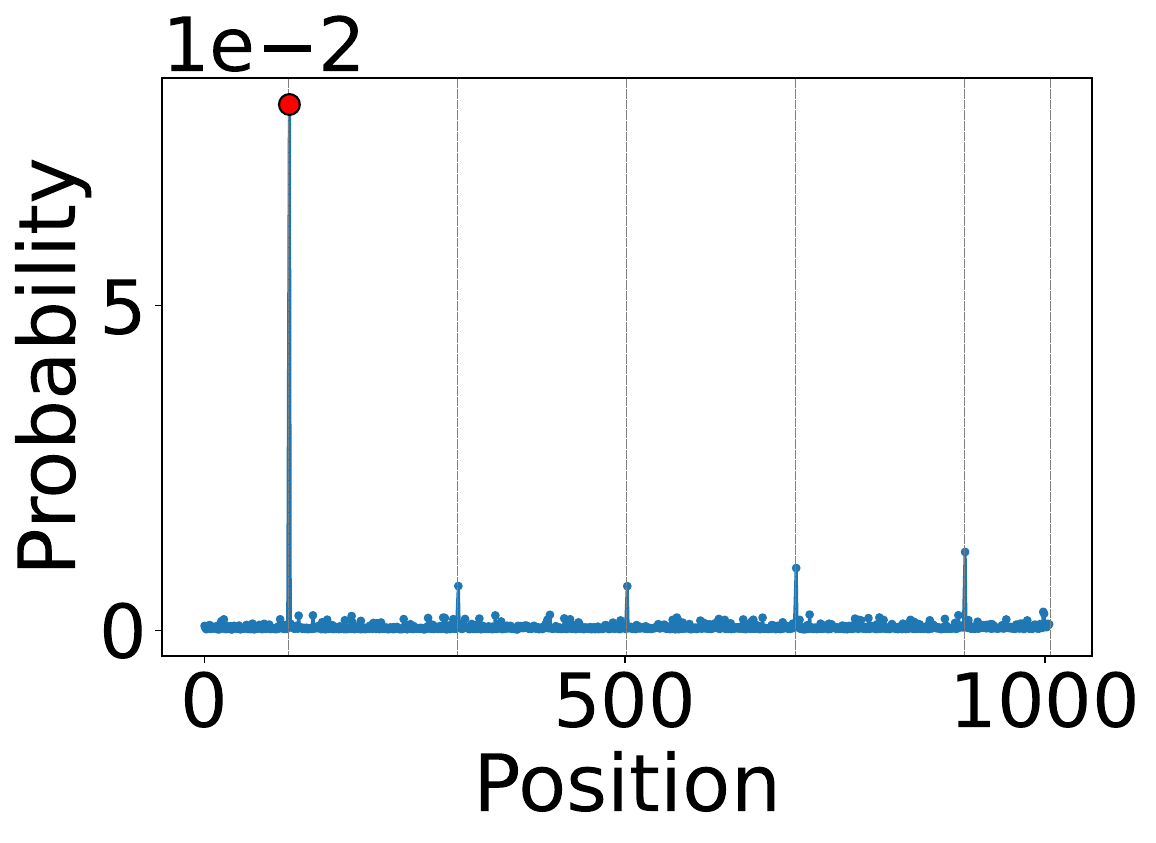} &
\FigGridCell{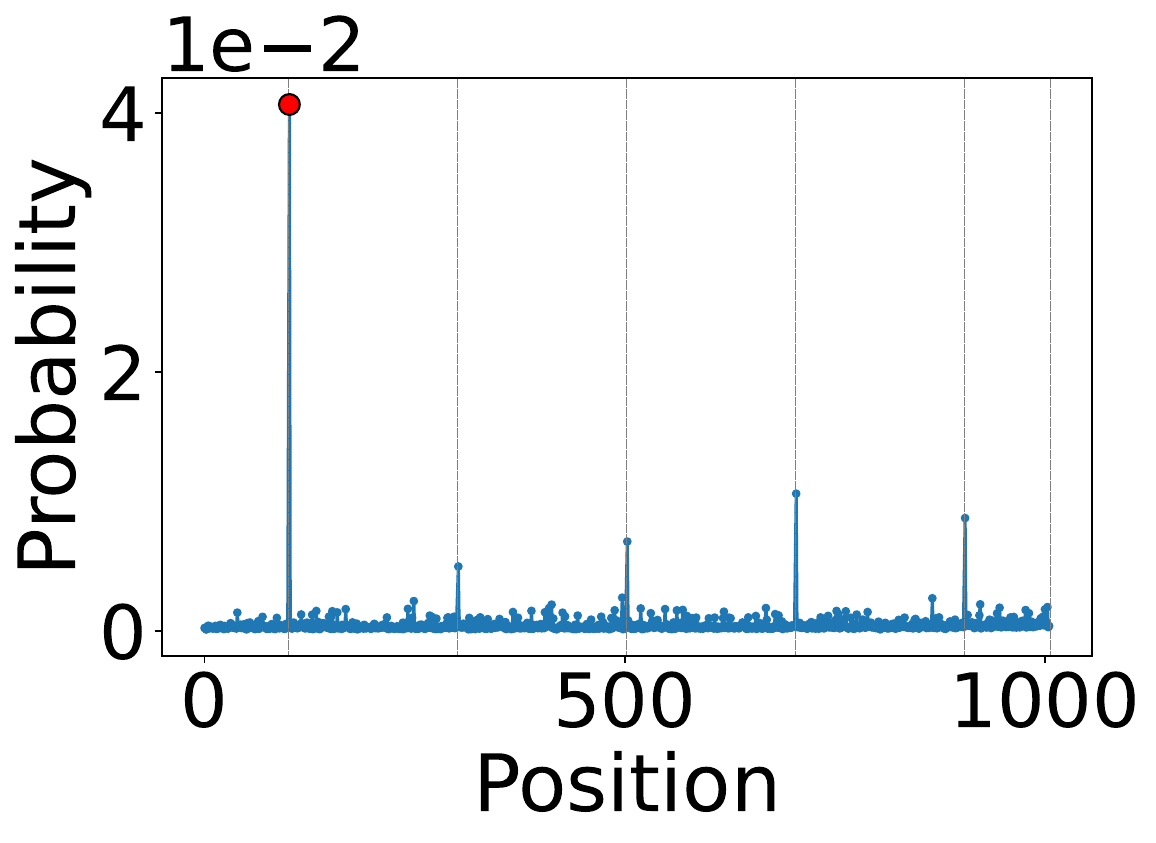} &
\FigGridCell{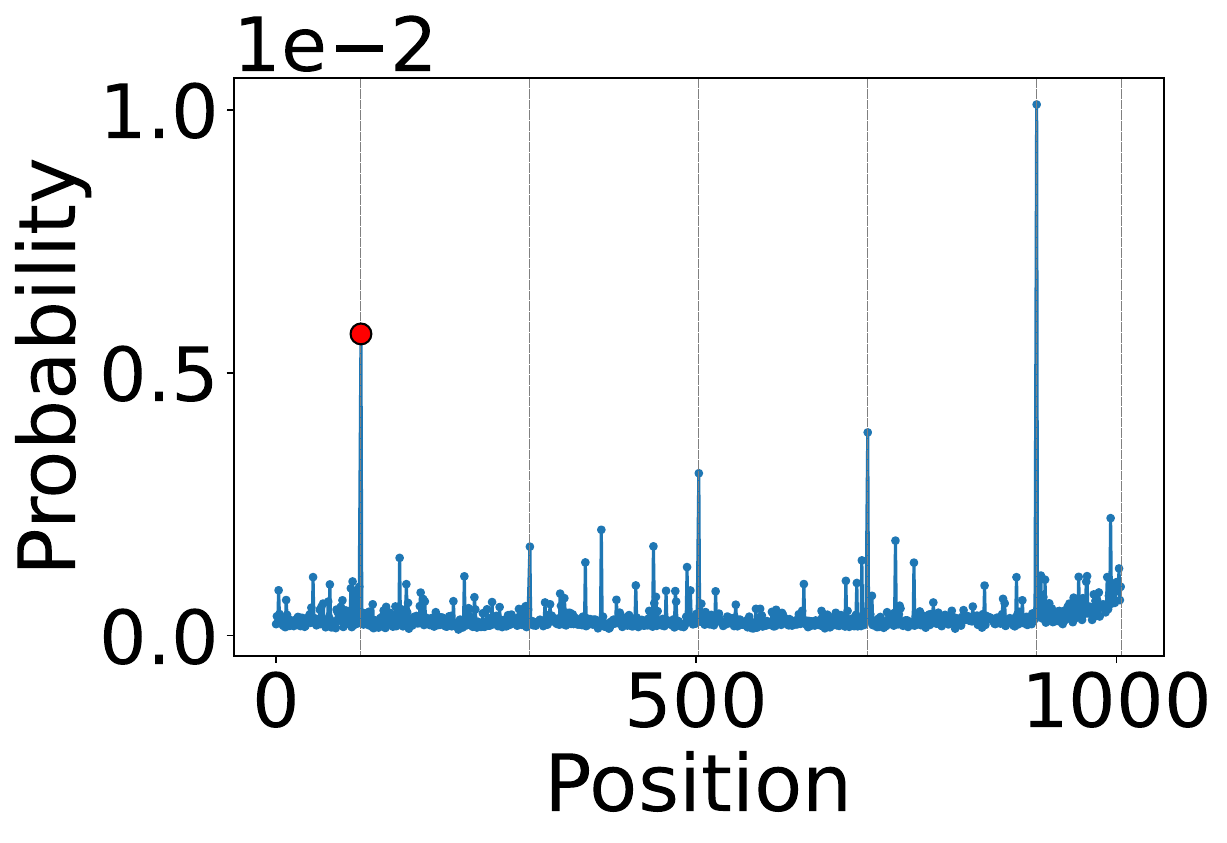} \\
& \FigGridVSubLabel{P2} &
\FigGridCell{Figures/ep_prob_without_A_red/Mistral-7B-Instruct-v0.1_5_Repeats_200_Length_500_Permutations_0_ablations_induction_2_nth.pdf} &
\FigGridCell{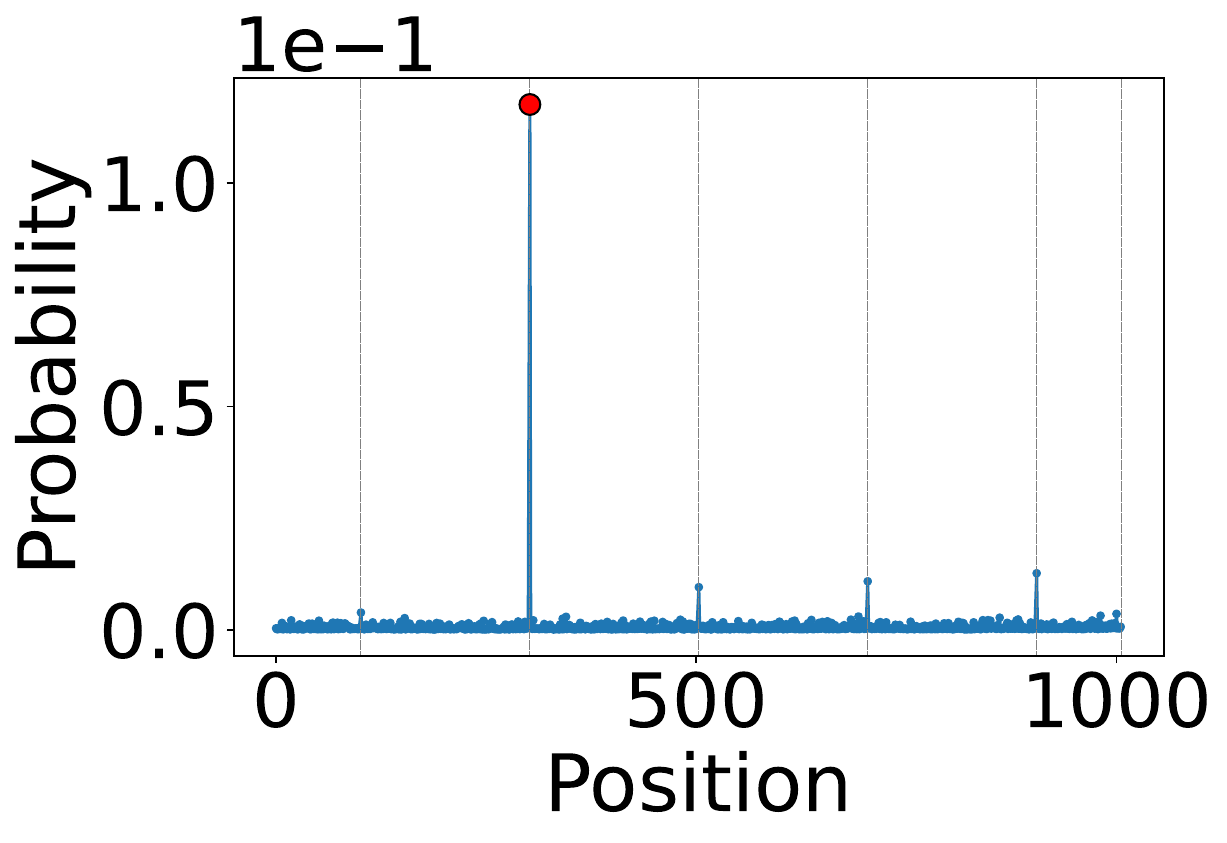} &
\FigGridCell{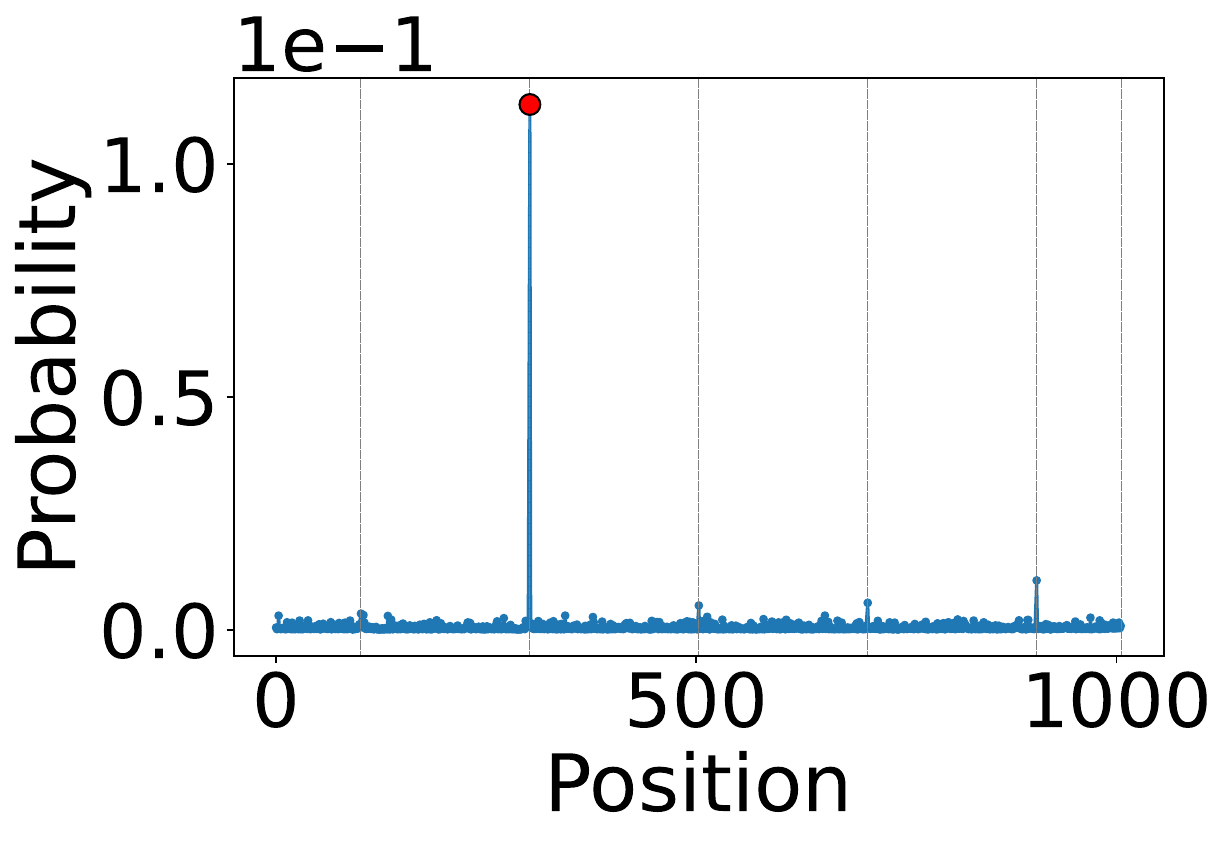} &
\FigGridCell{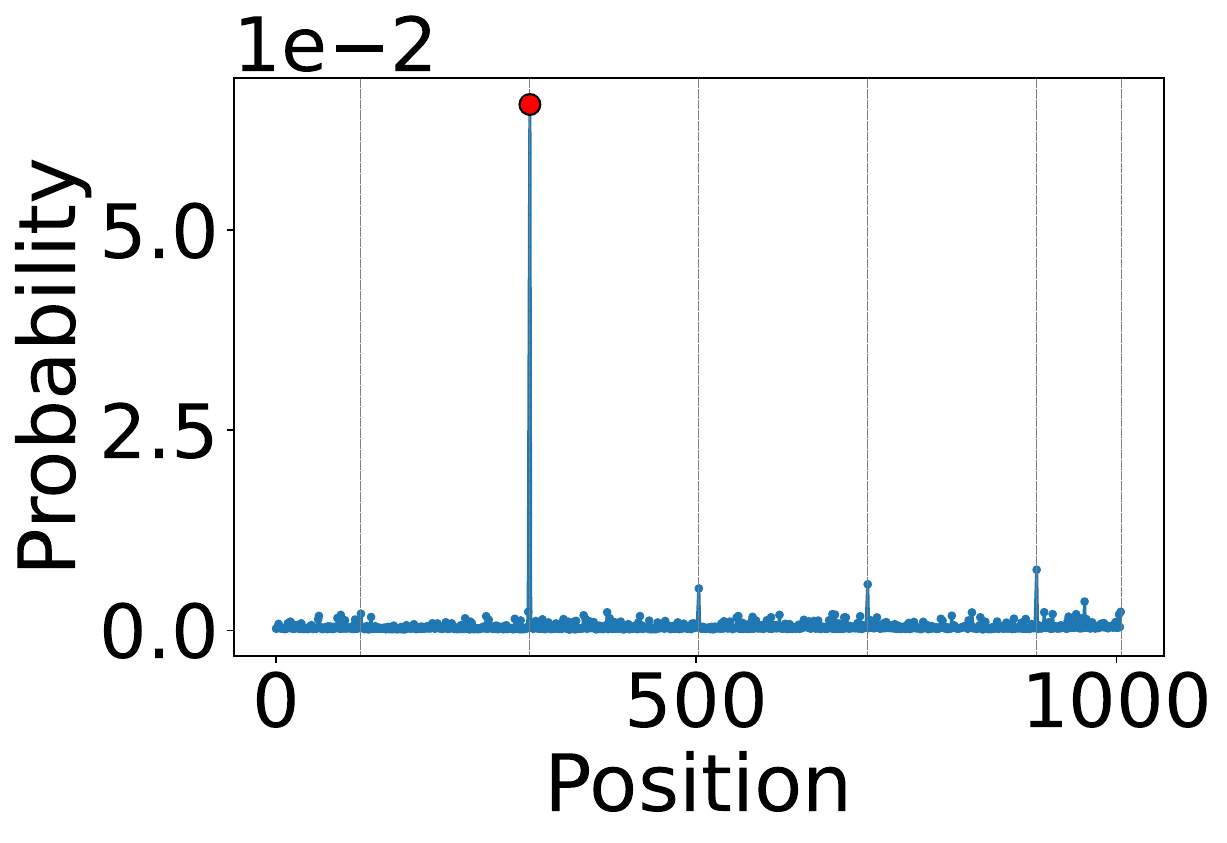} &
\FigGridCell{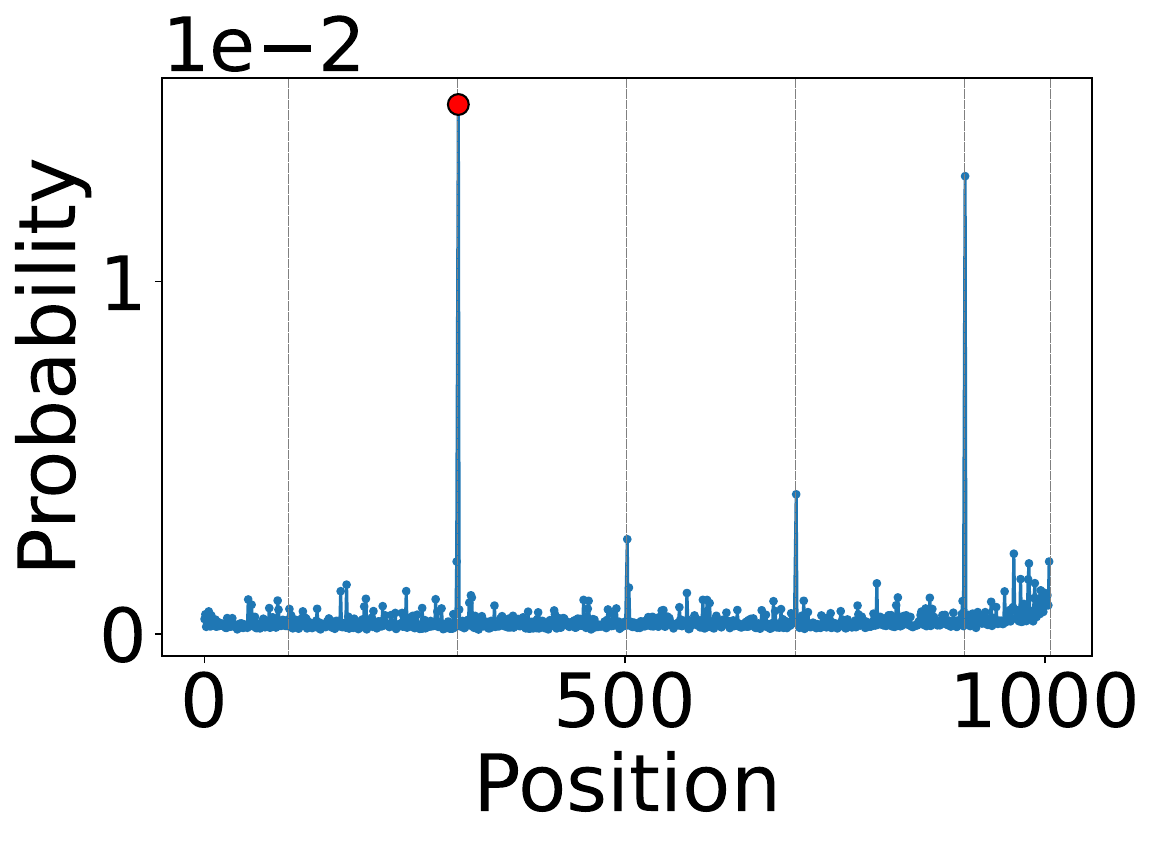} \\
& \FigGridVSubLabel{P3} &
\FigGridCell{Figures/ep_prob_without_A_red/Mistral-7B-Instruct-v0.1_5_Repeats_200_Length_500_Permutations_0_ablations_induction_3_nth.pdf} &
\FigGridCell{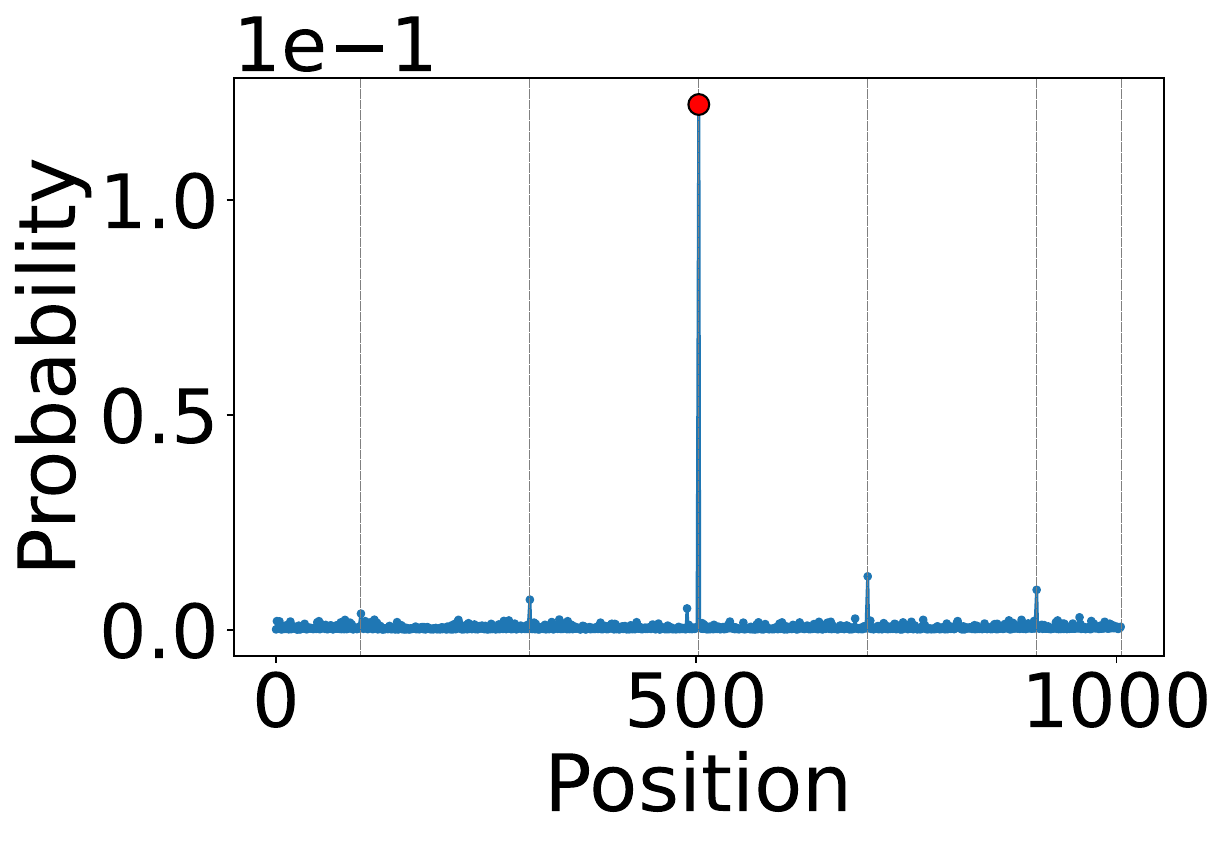} &
\FigGridCell{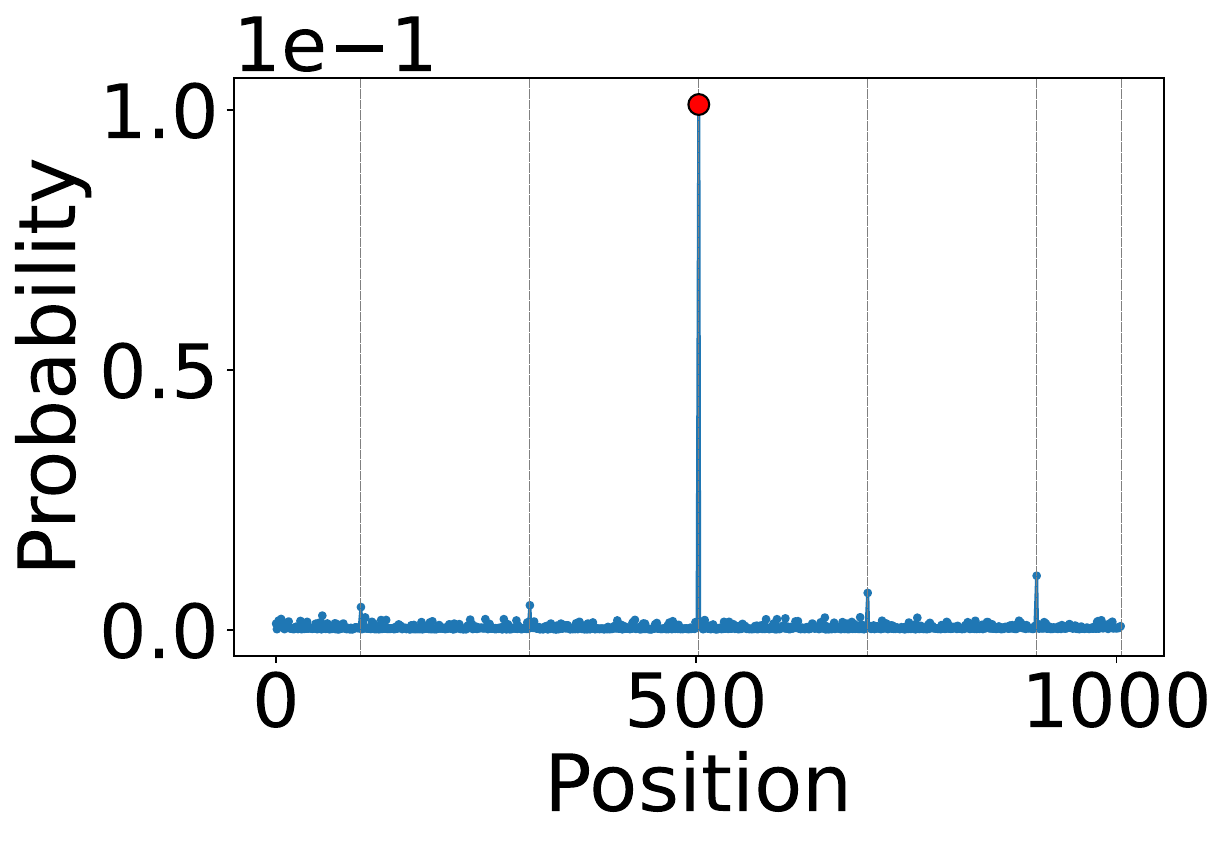} &
\FigGridCell{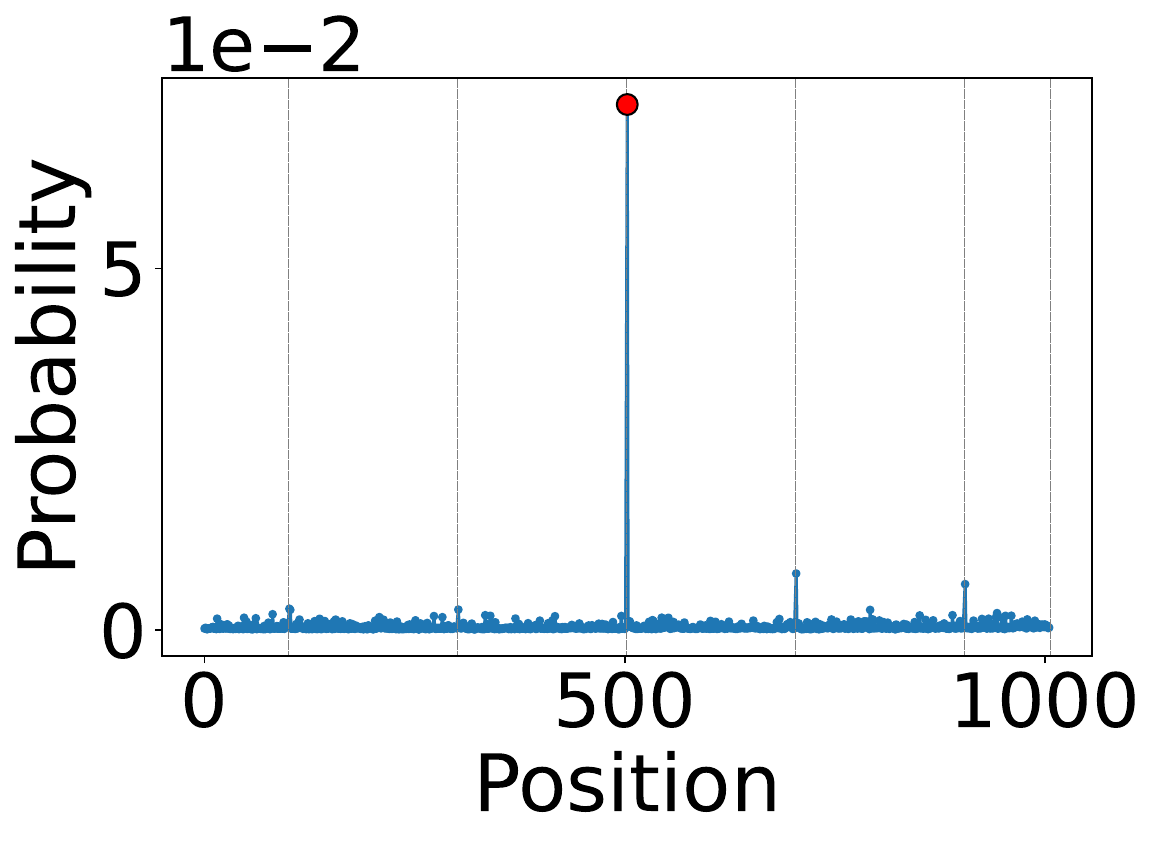} &
\FigGridCell{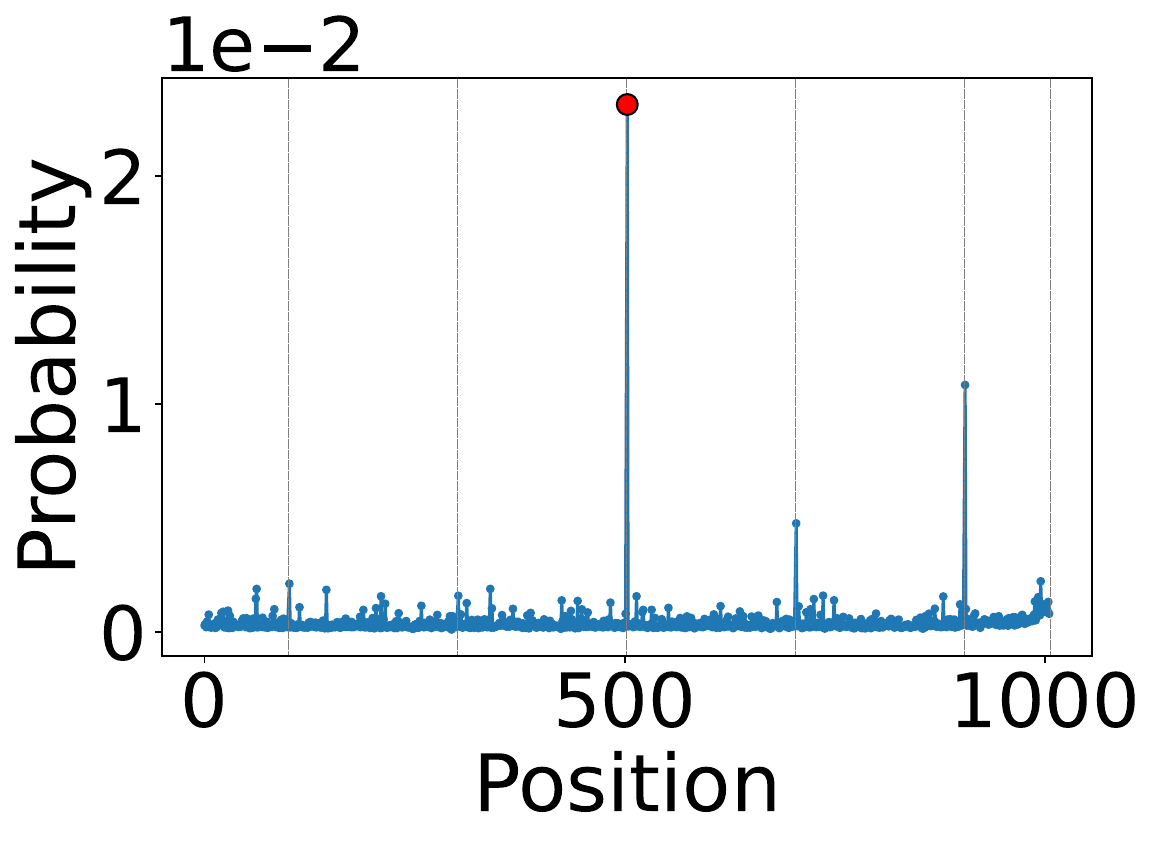} \\
& \FigGridVSubLabel{P4} &
\FigGridCell{Figures/ep_prob_without_A_red/Mistral-7B-Instruct-v0.1_5_Repeats_200_Length_500_Permutations_0_ablations_induction_4_nth.pdf} &
\FigGridCell{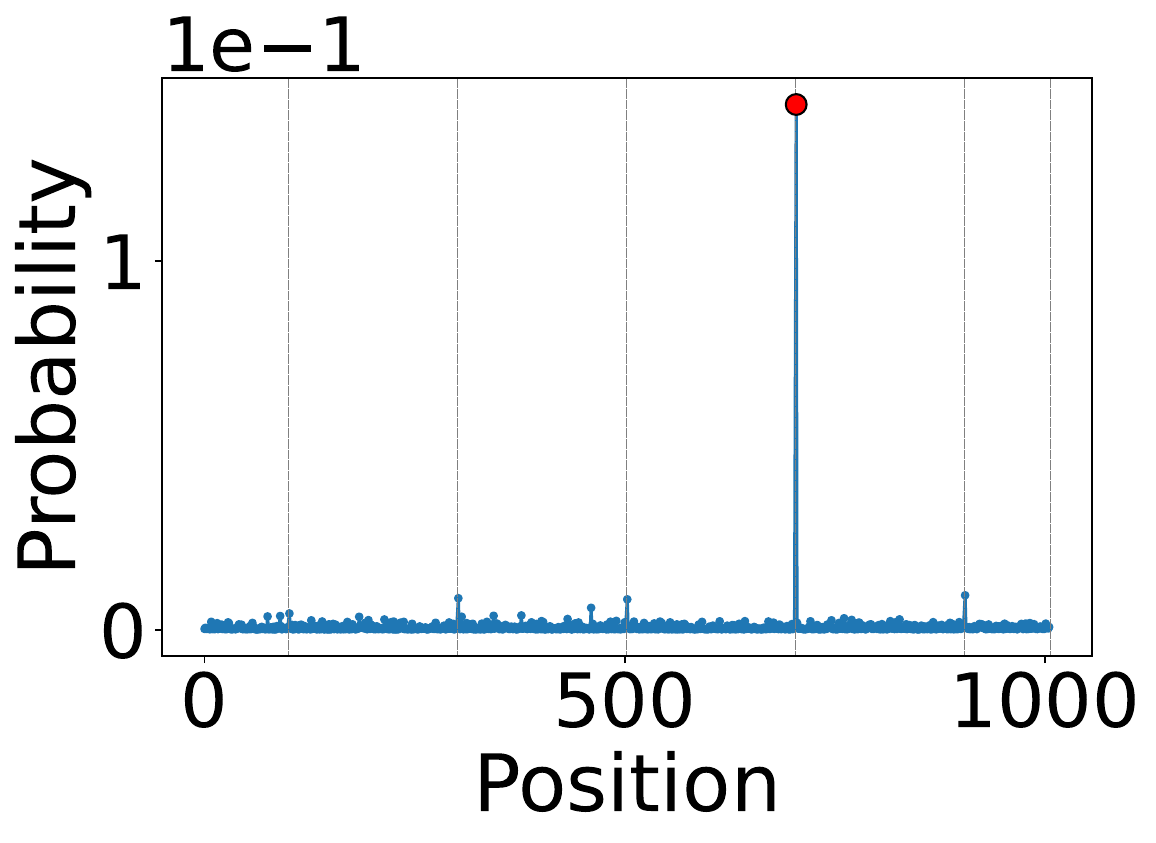} &
\FigGridCell{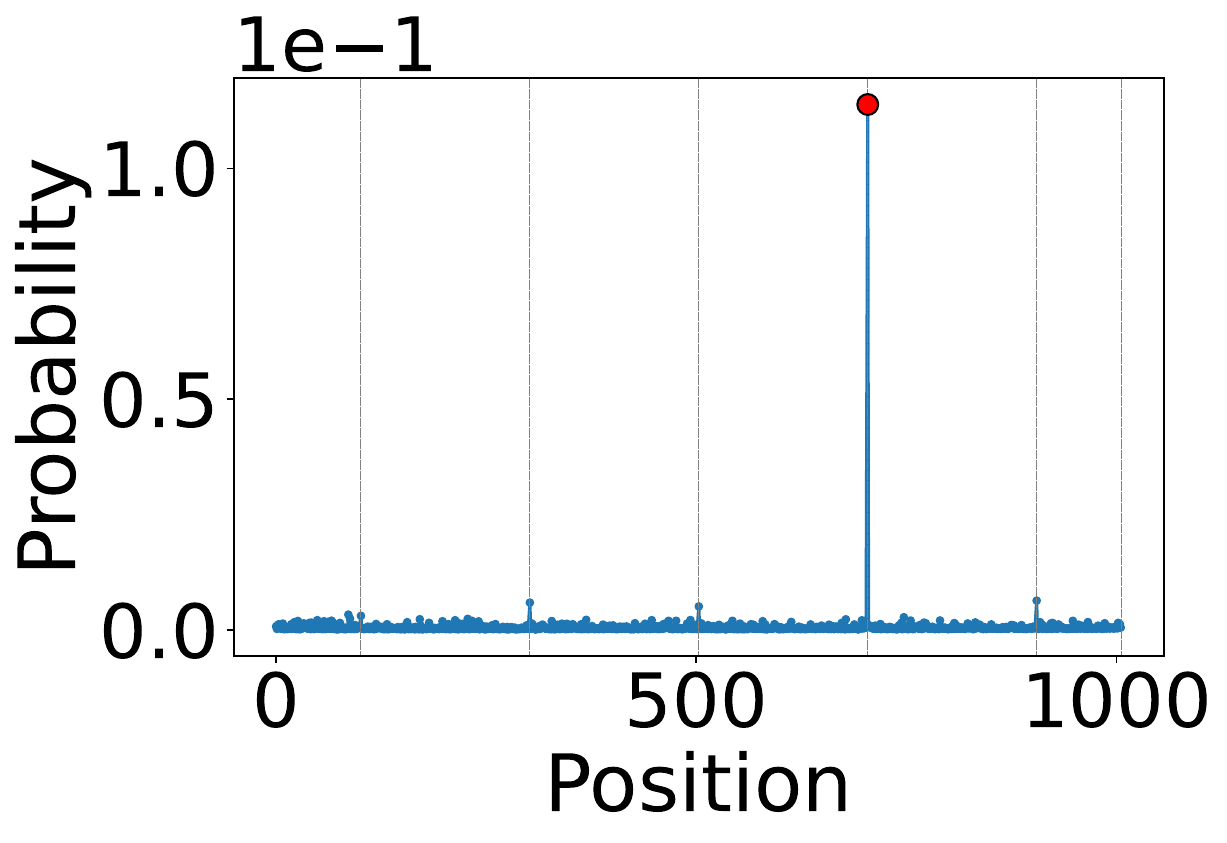} &
\FigGridCell{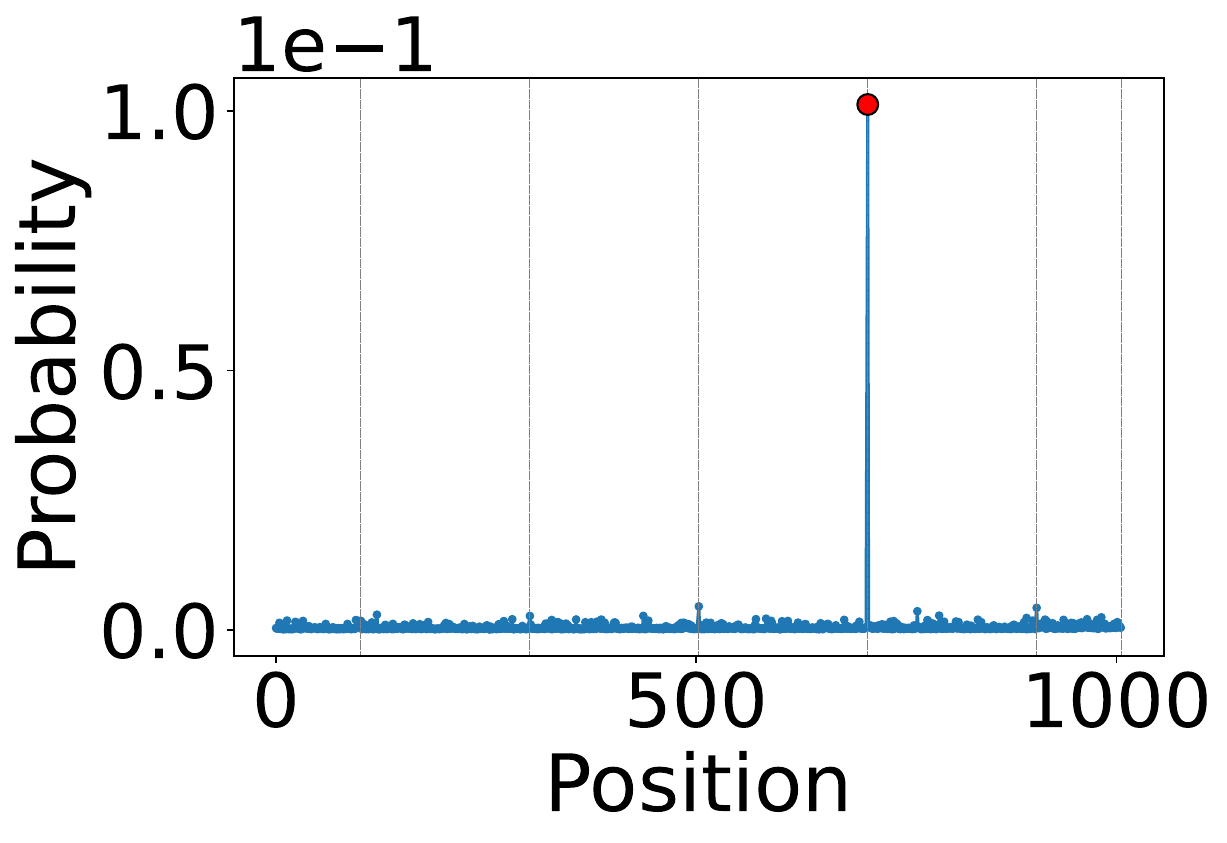} &
\FigGridCell{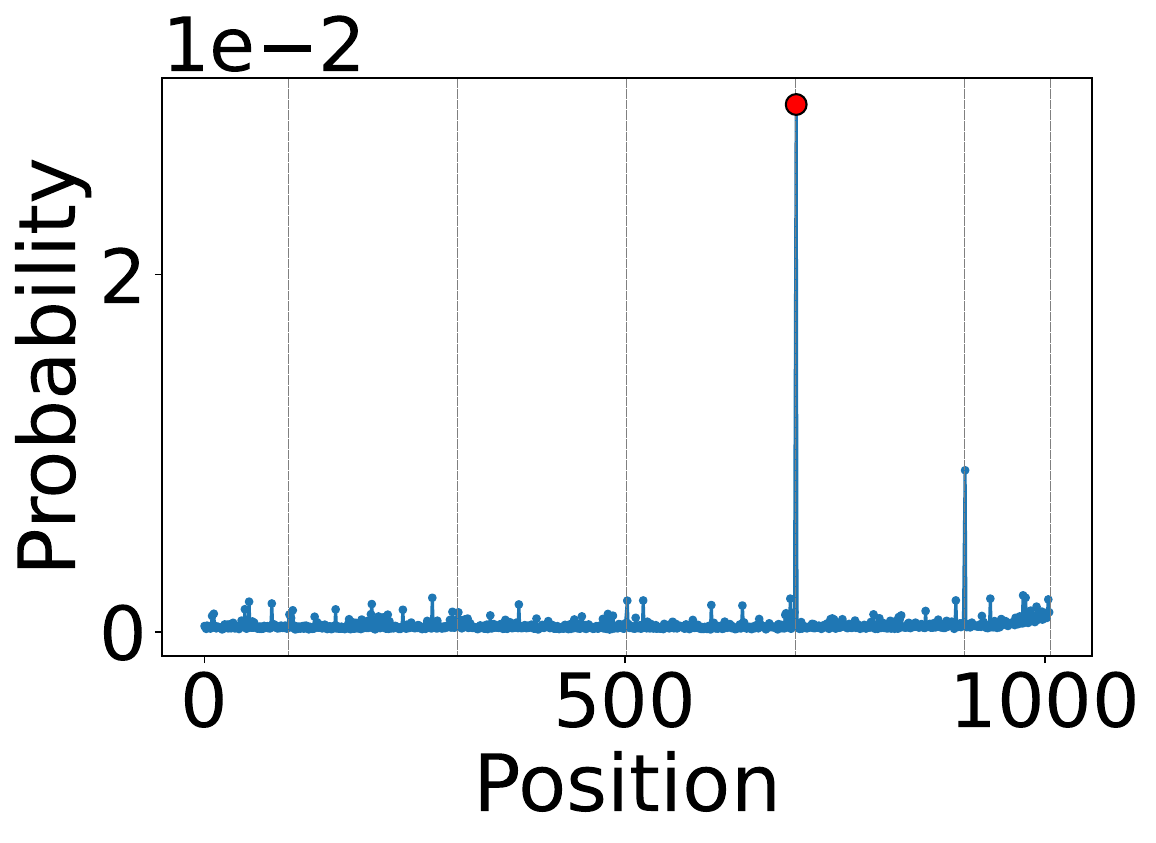} \\
& \FigGridVSubLabel{P5} &
\FigGridCell{Figures/ep_prob_without_A_red/Mistral-7B-Instruct-v0.1_5_Repeats_200_Length_500_Permutations_0_ablations_induction_5_nth.pdf} &
\FigGridCell{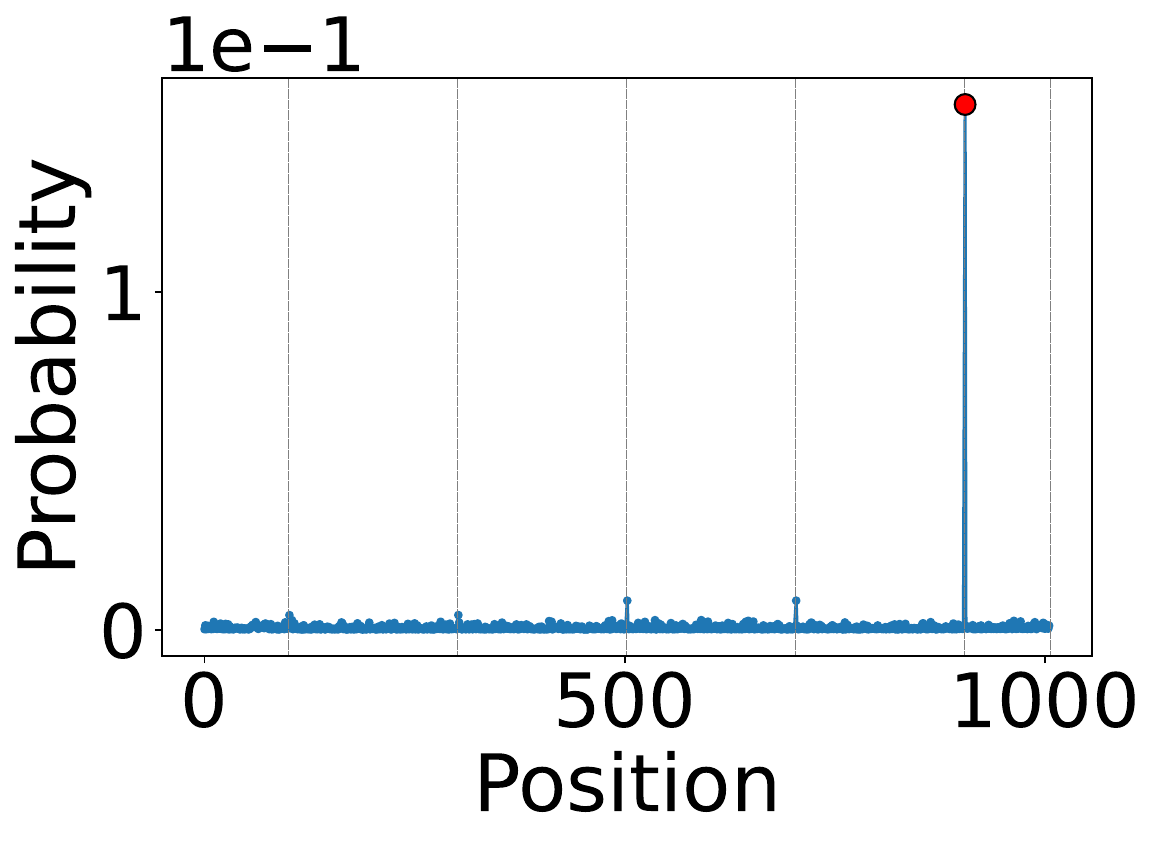} &
\FigGridCell{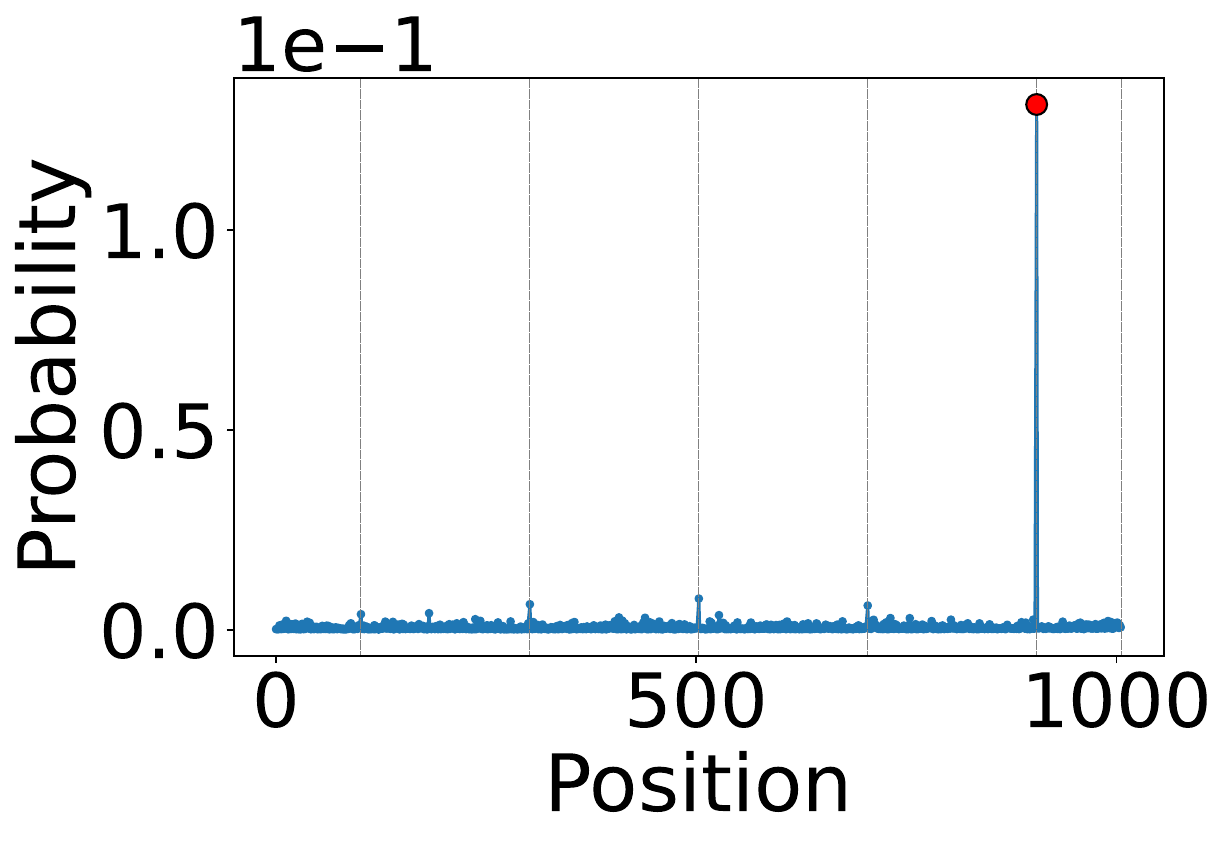} &
\FigGridCell{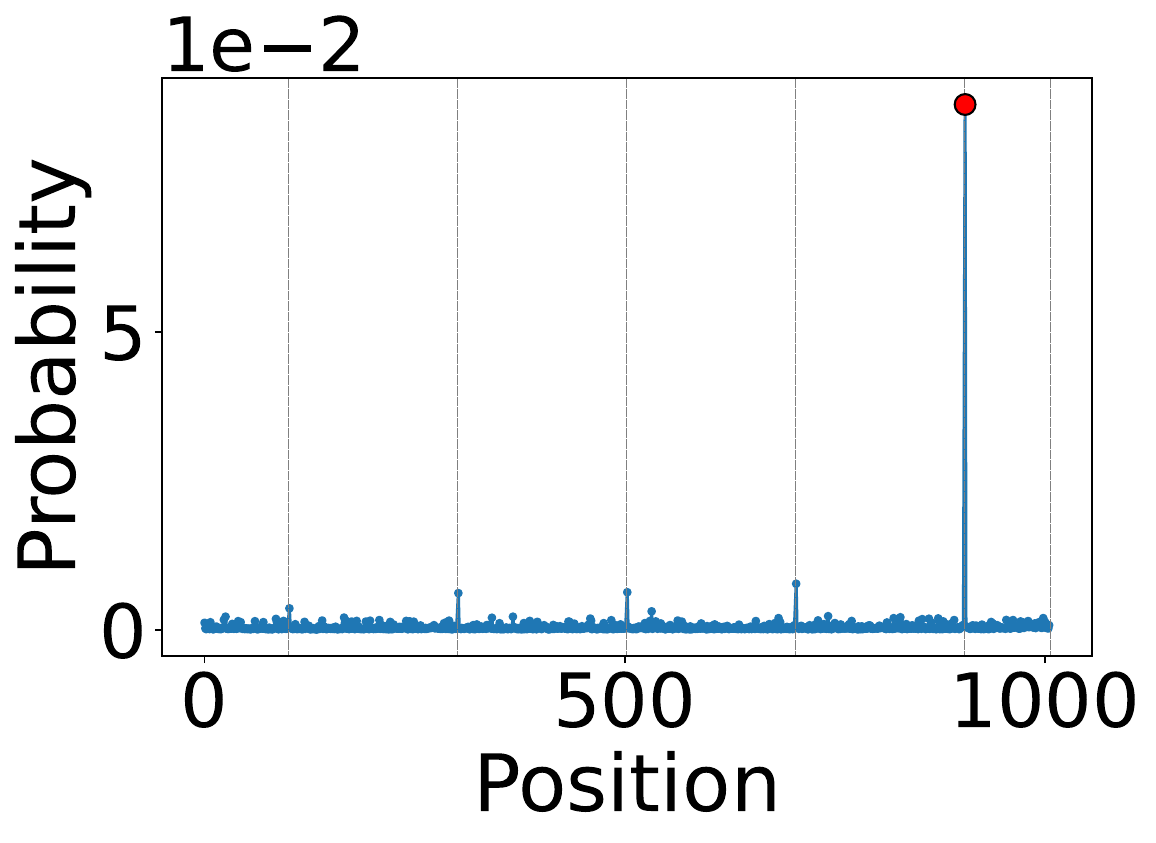} &
\FigGridCell{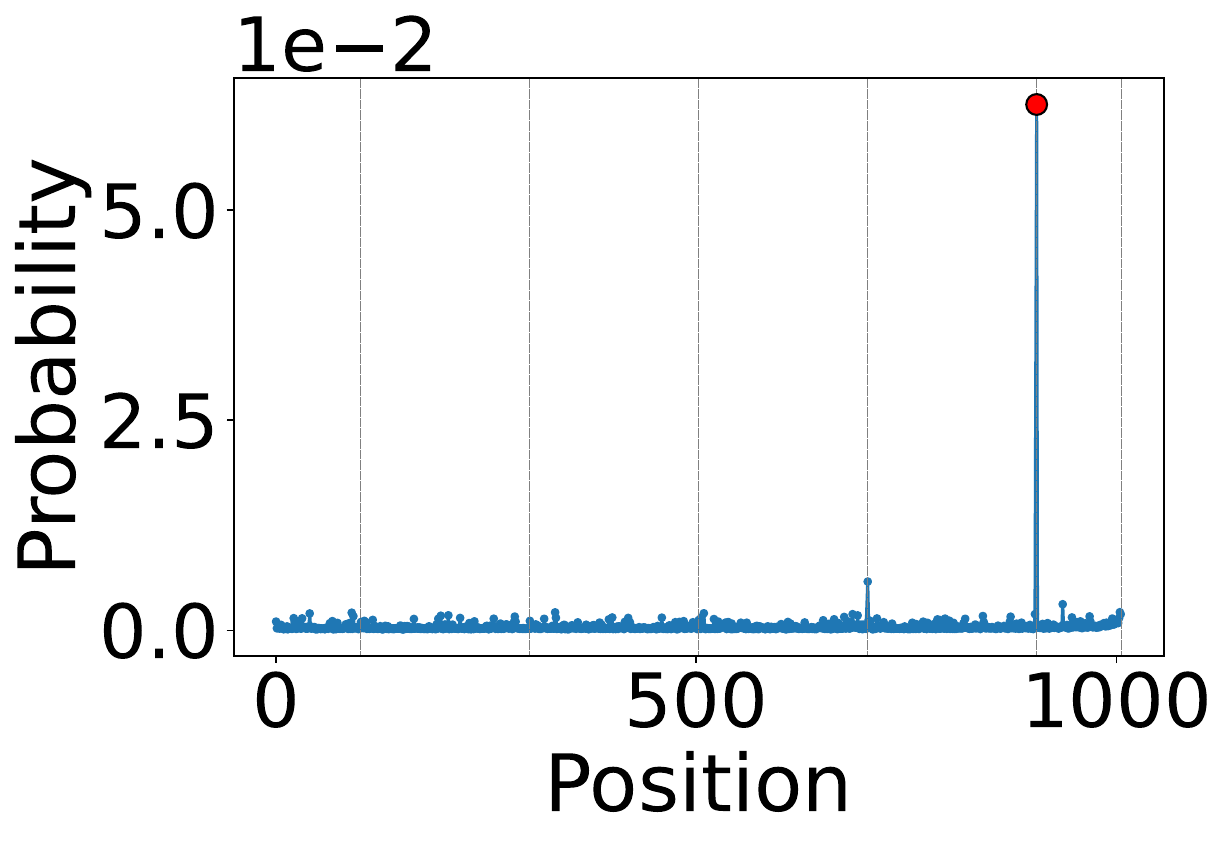} \\
\addlinespace[1pt]
\midrule
\addlinespace[1pt]
\multirow{5}{*}{\FigGridVLabelSpanFive{Random}} &
\FigGridVSubLabel{P1} &
\FigGridCell{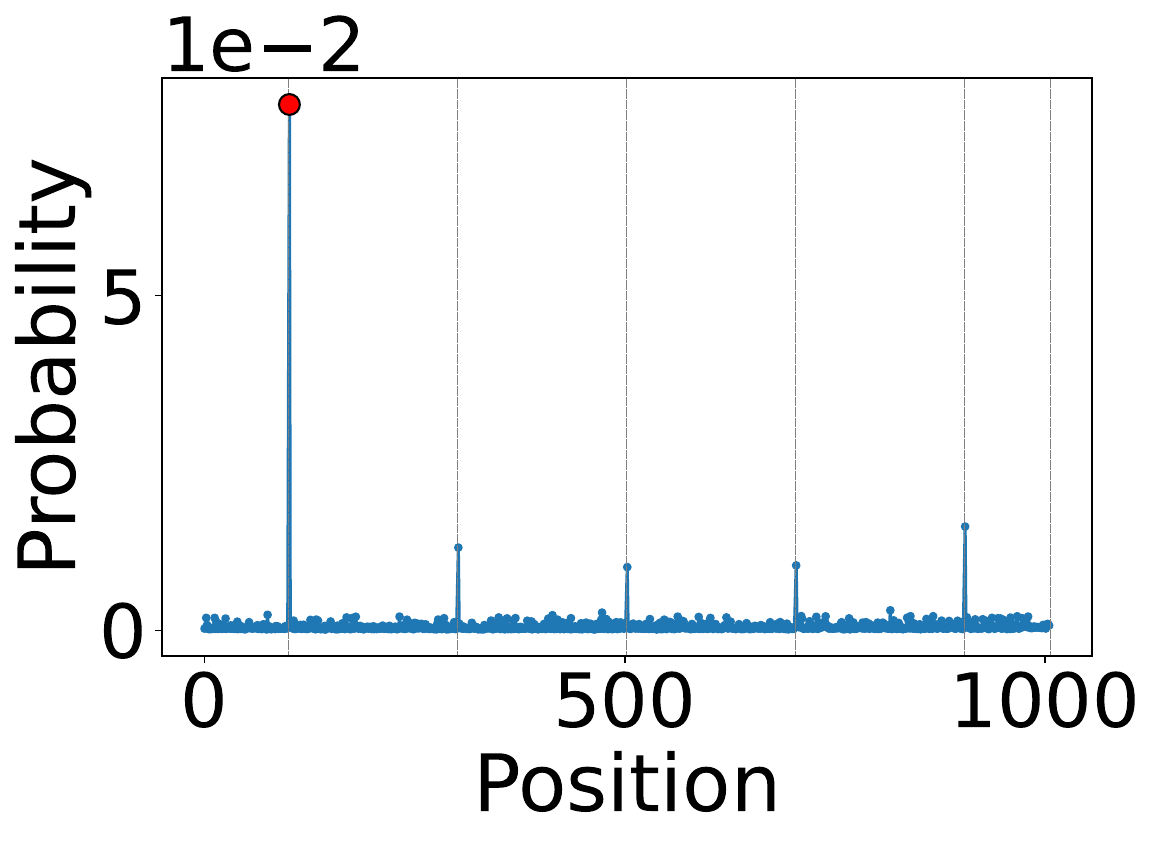} &
\FigGridCell{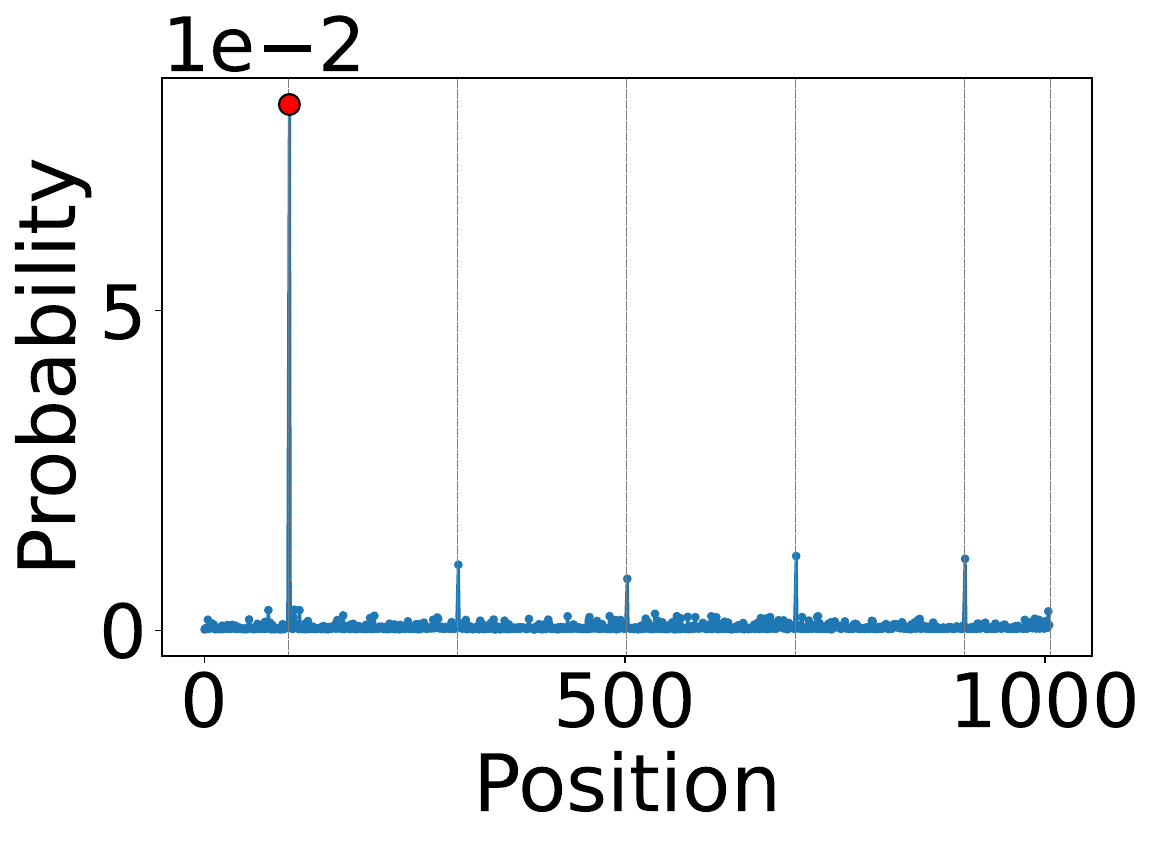} &
\FigGridCell{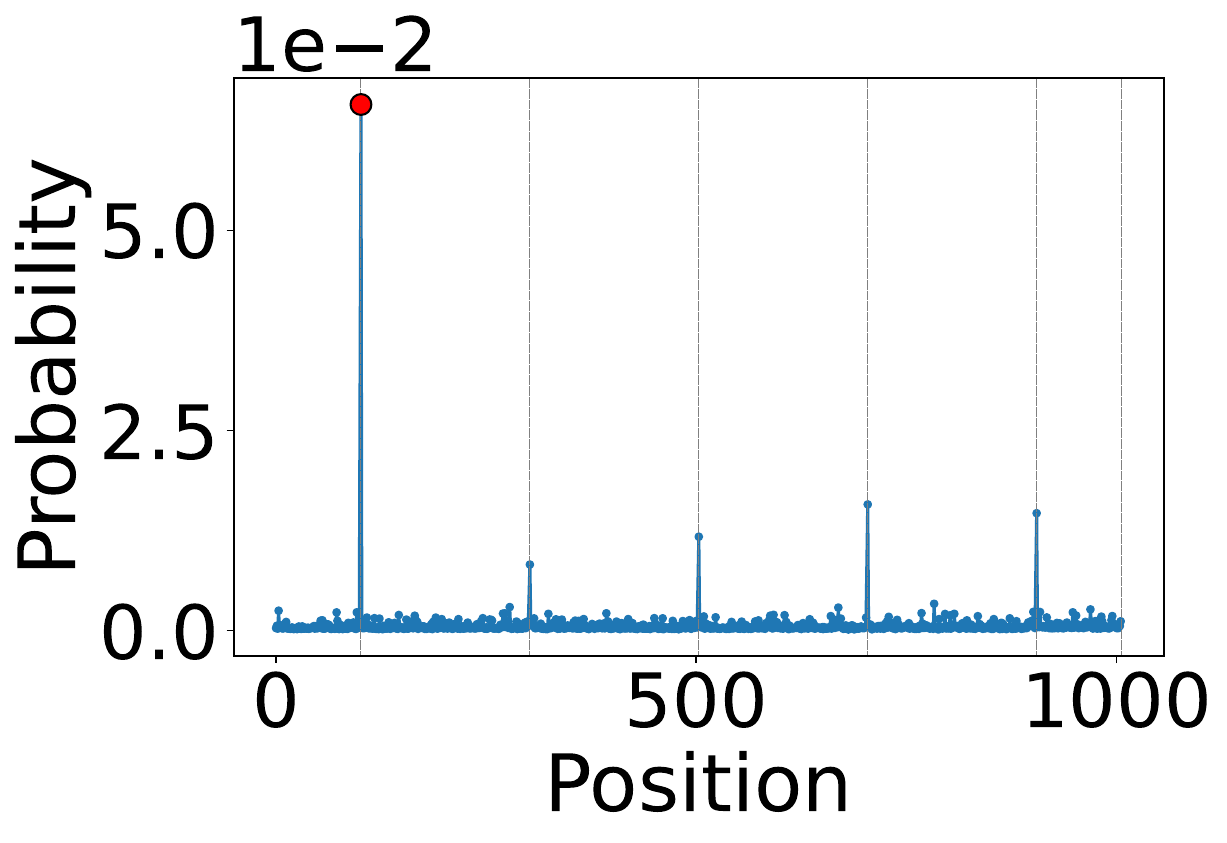} &
\FigGridCell{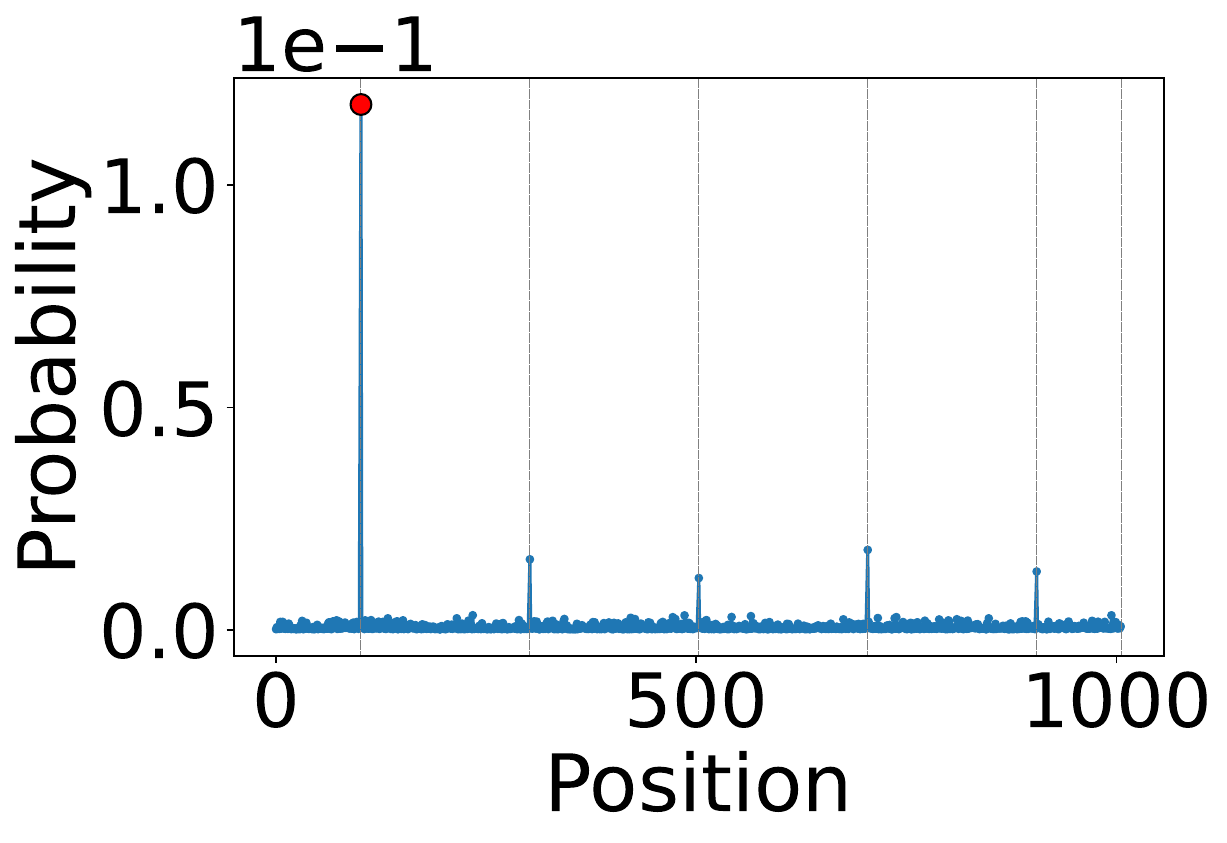} &
\FigGridCell{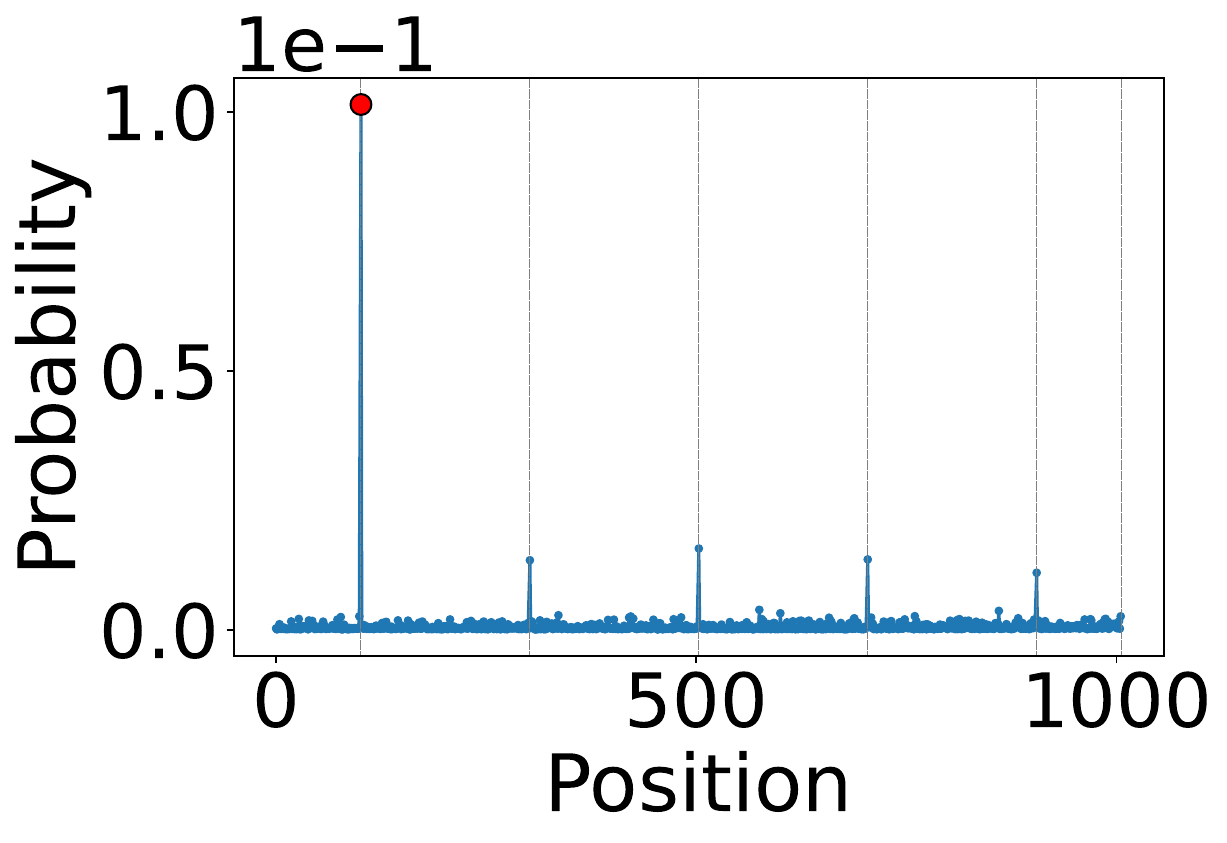} \\
& \FigGridVSubLabel{P2} &
\FigGridCell{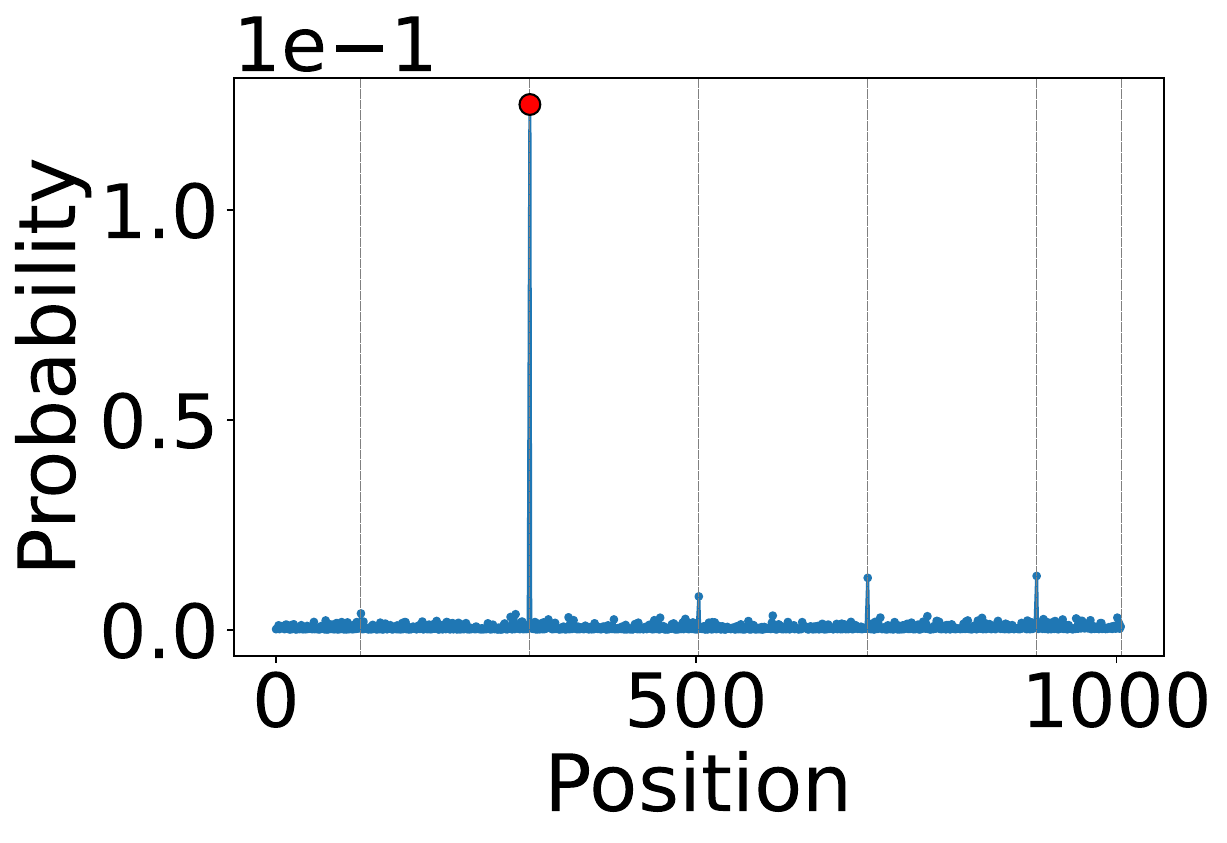} &
\FigGridCell{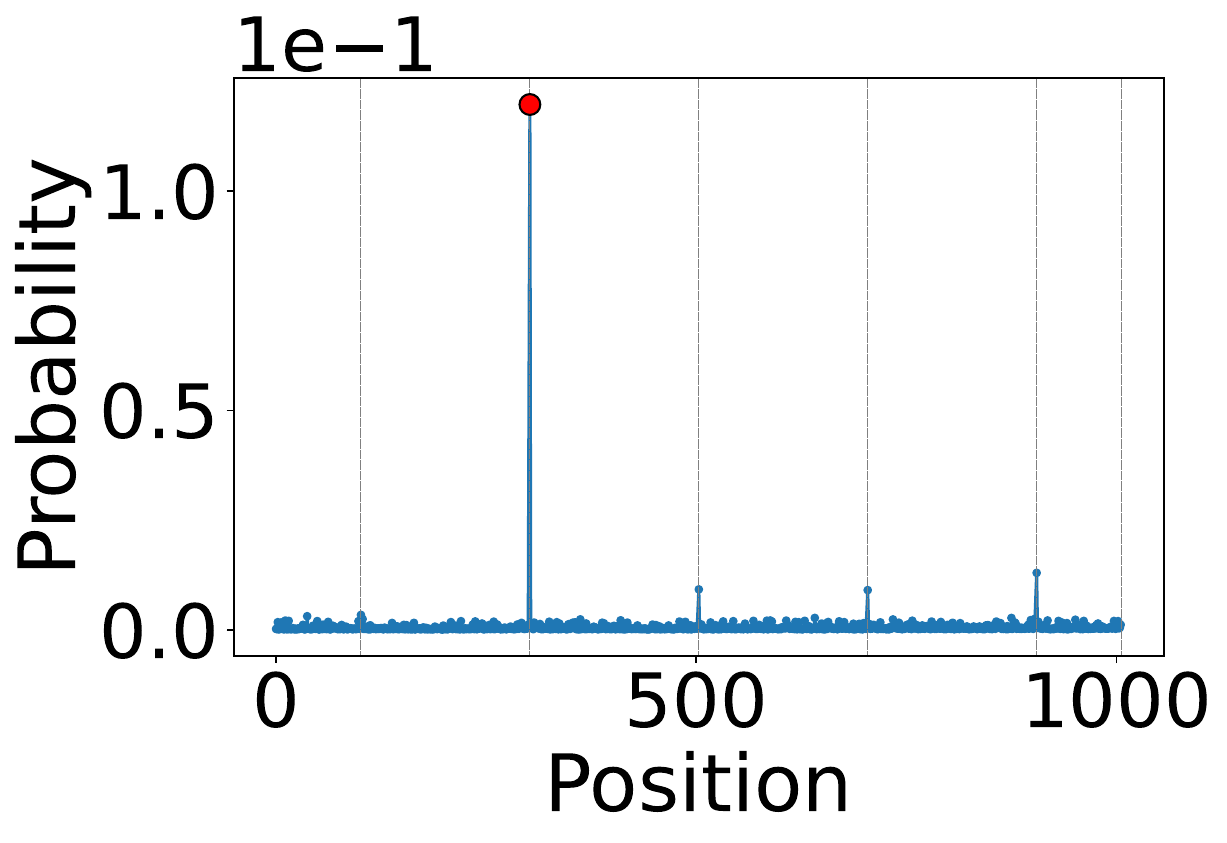} &
\FigGridCell{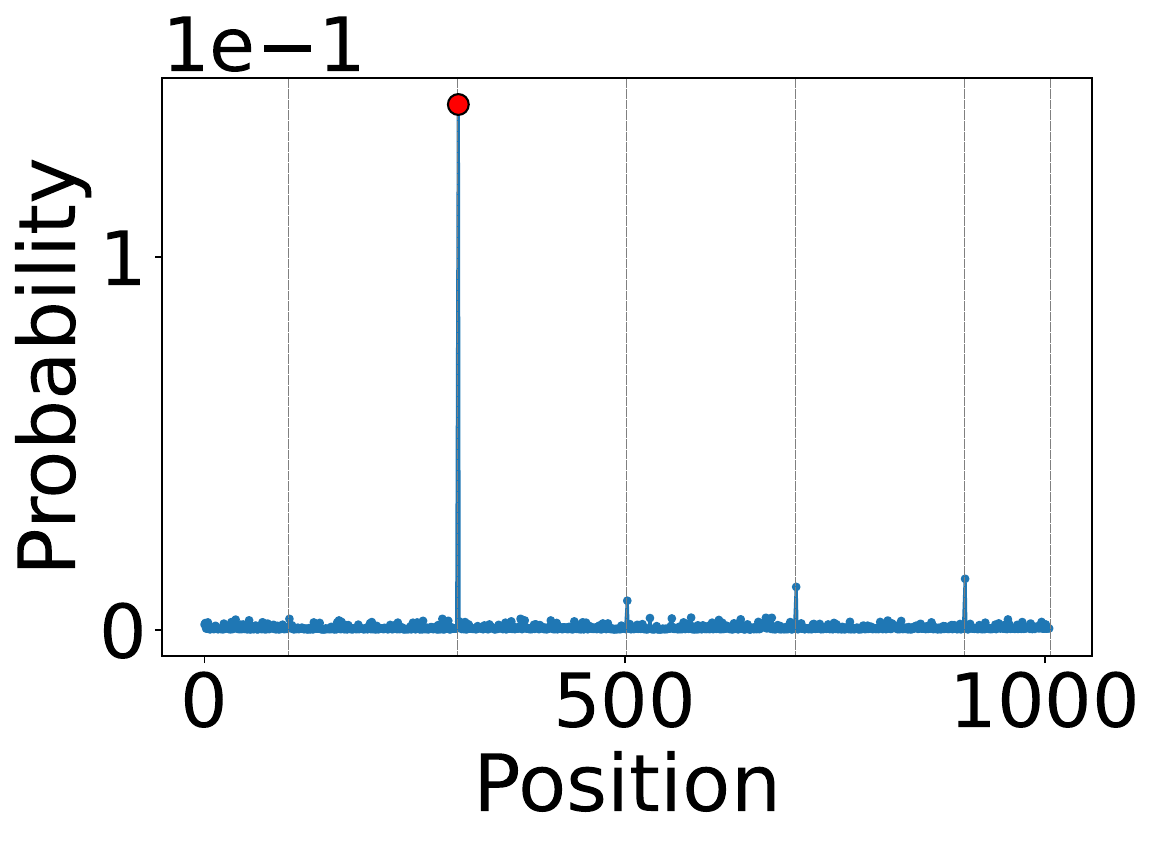} &
\FigGridCell{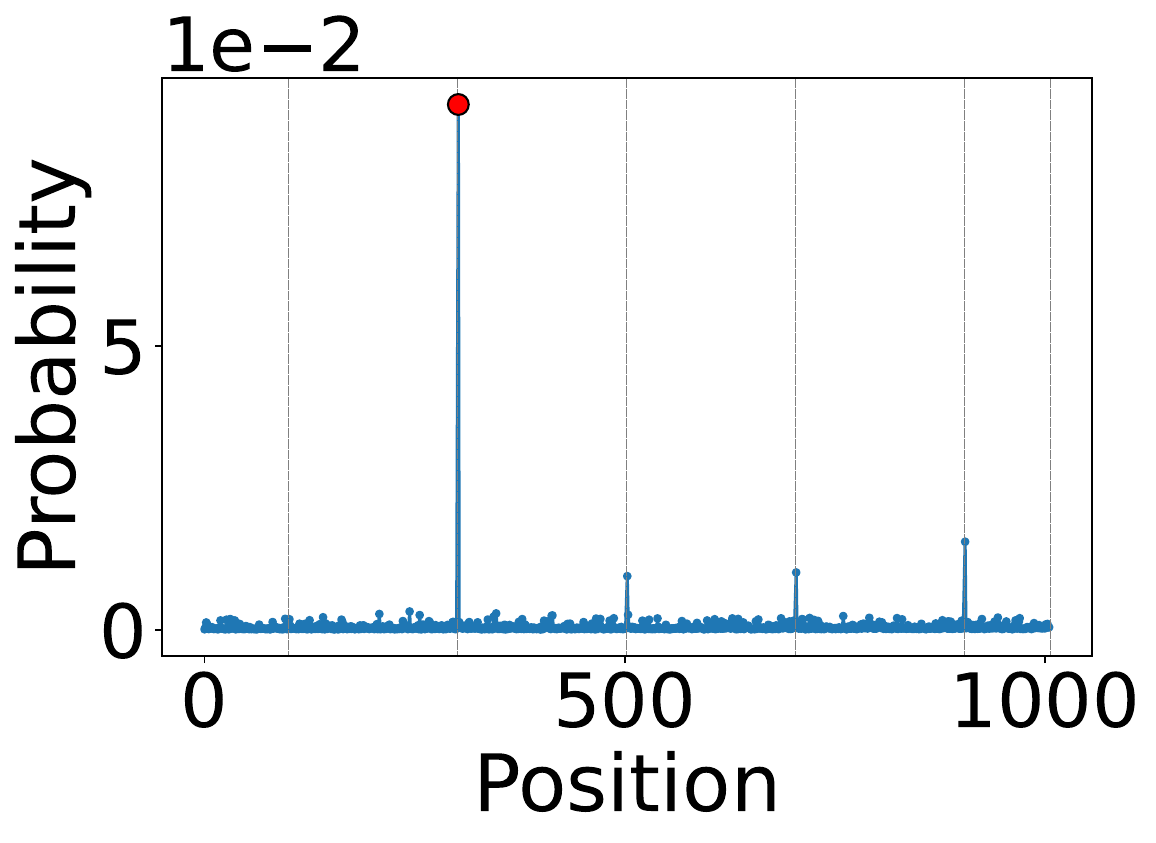} &
\FigGridCell{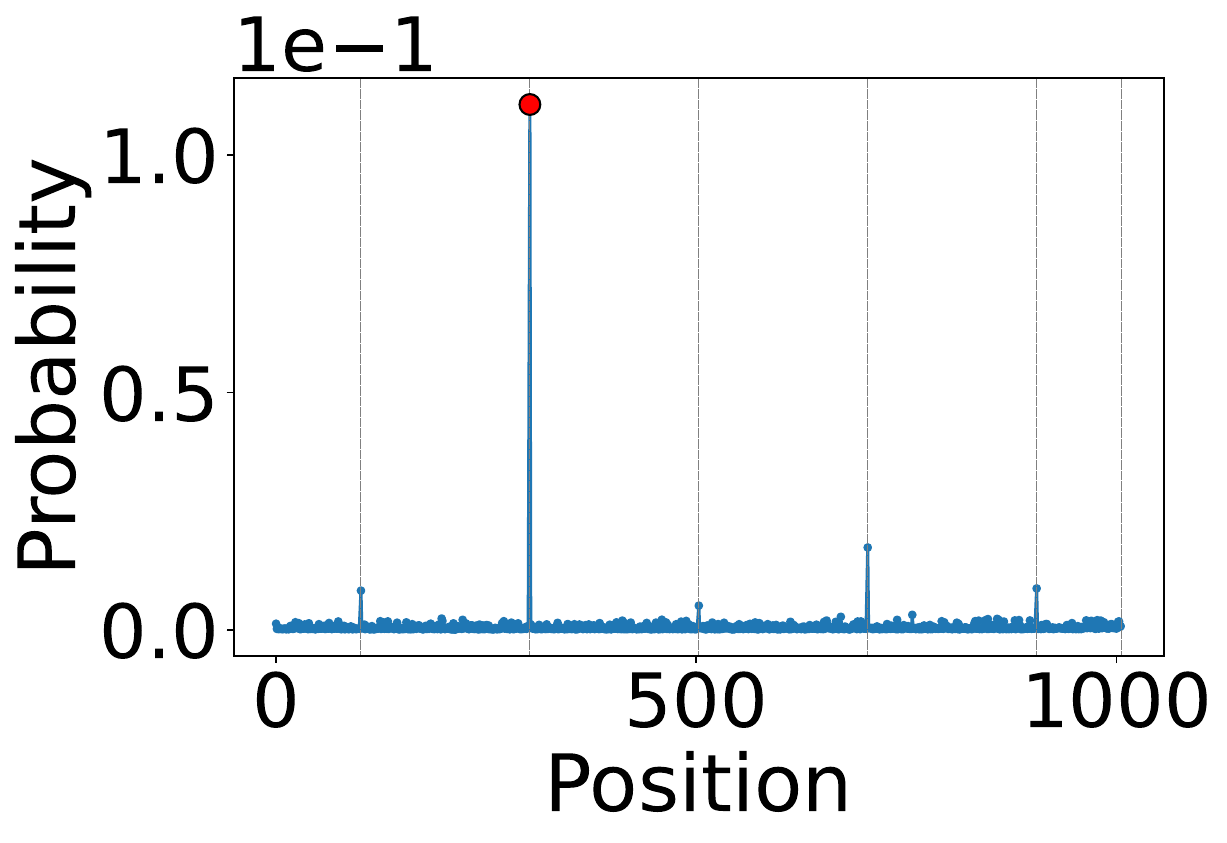} \\
& \FigGridVSubLabel{P3} &
\FigGridCell{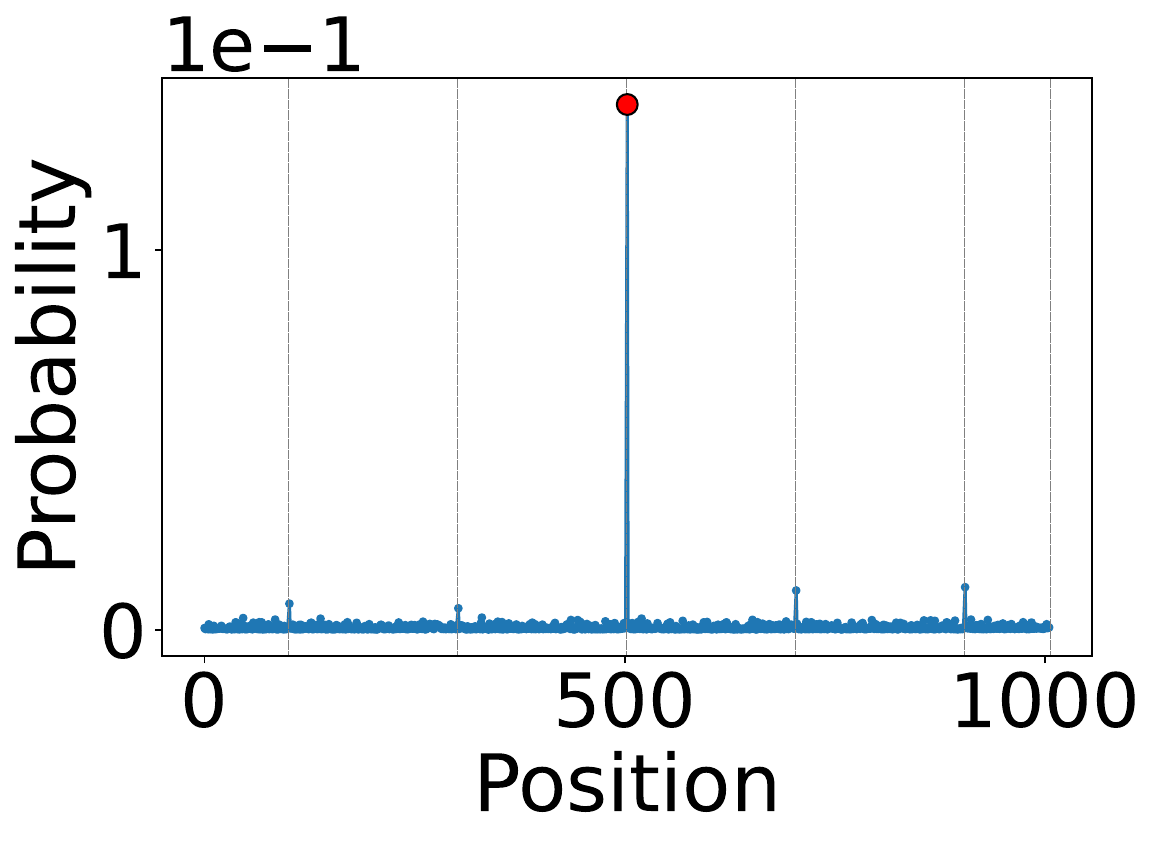} &
\FigGridCell{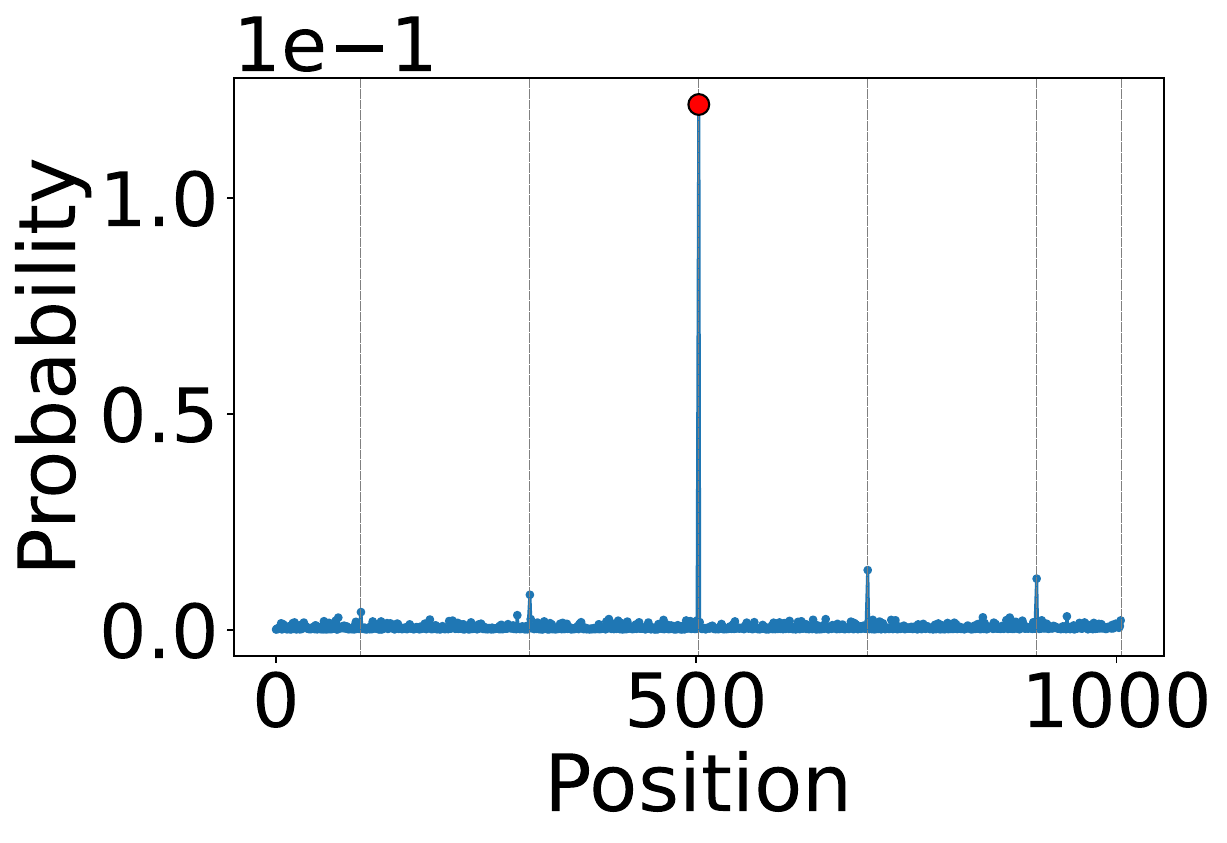} &
\FigGridCell{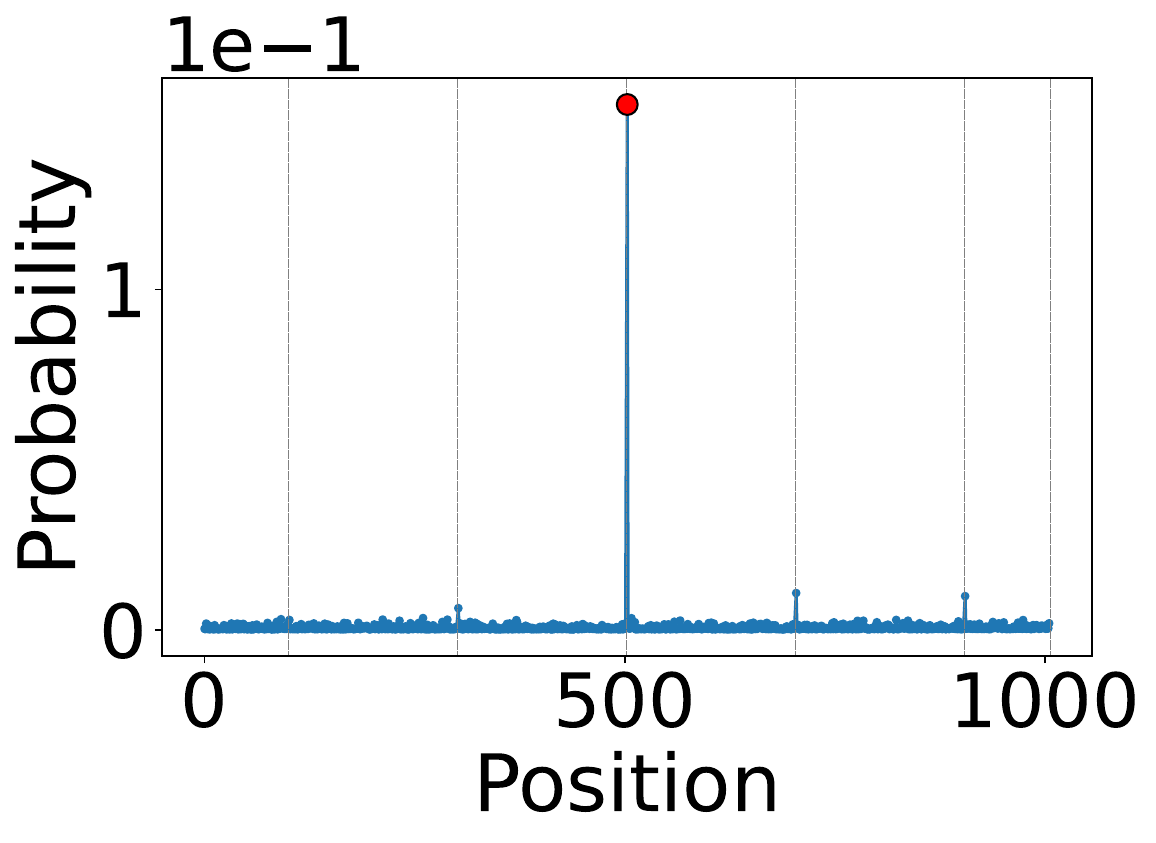} &
\FigGridCell{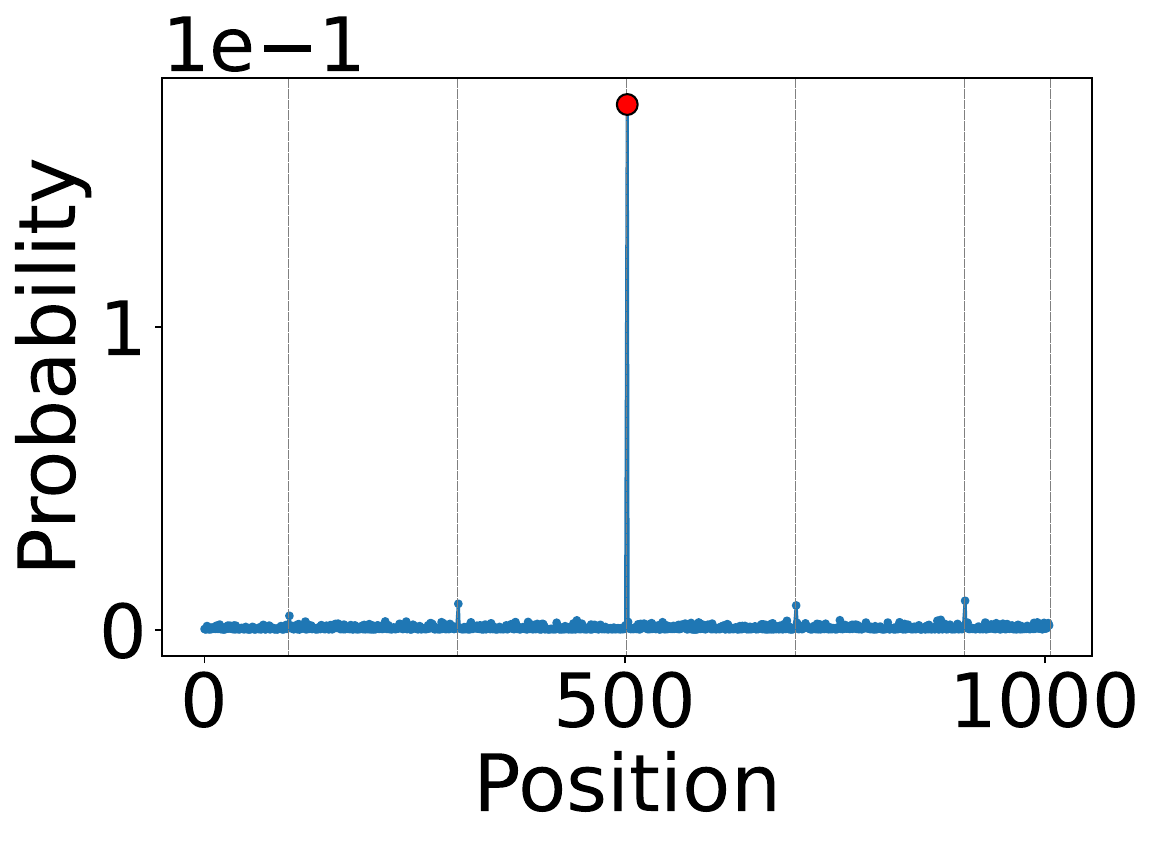} &
\FigGridCell{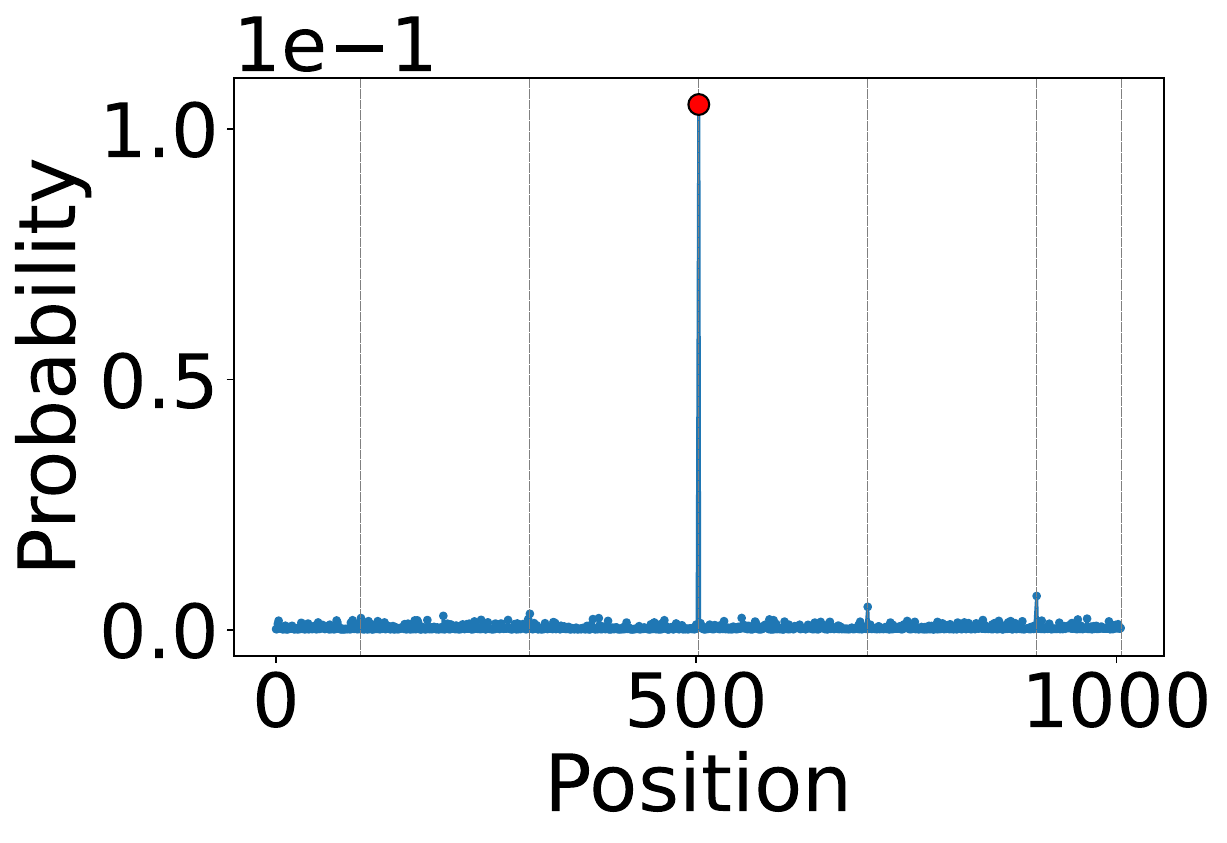} \\
& \FigGridVSubLabel{P4} &
\FigGridCell{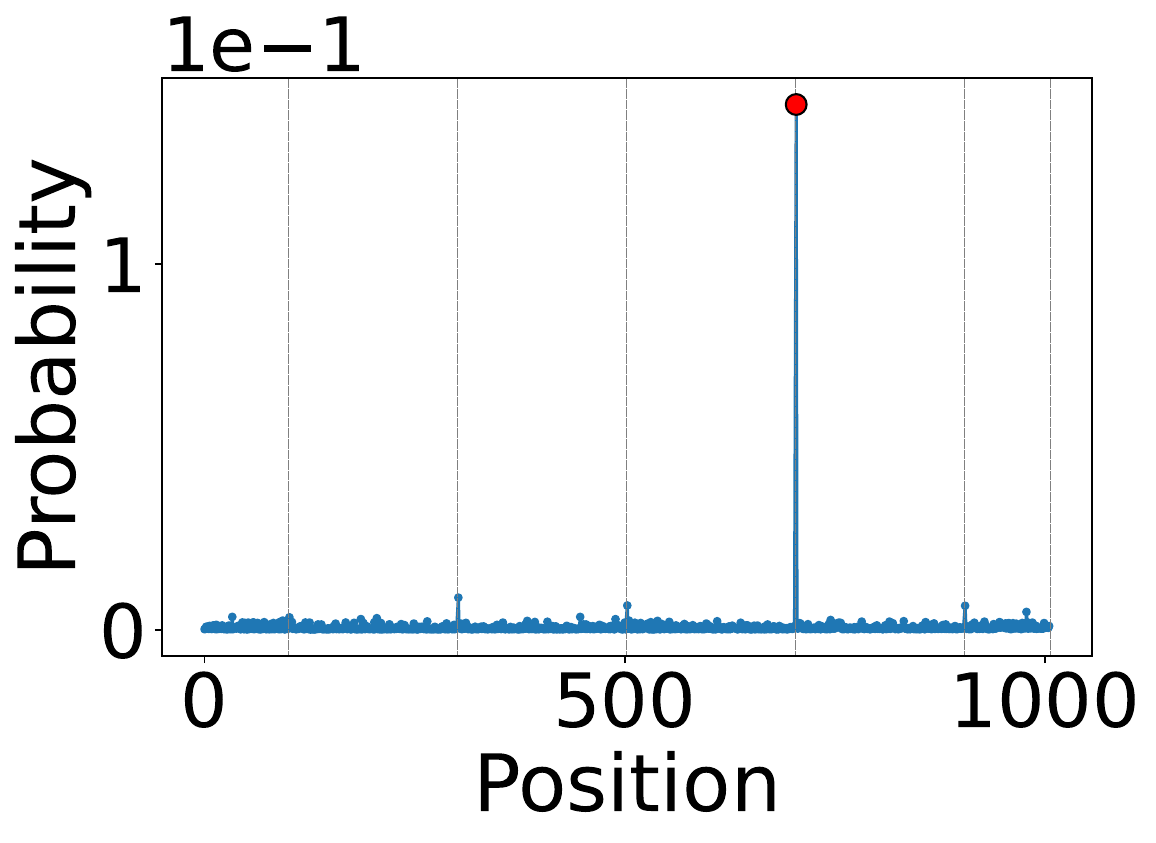} &
\FigGridCell{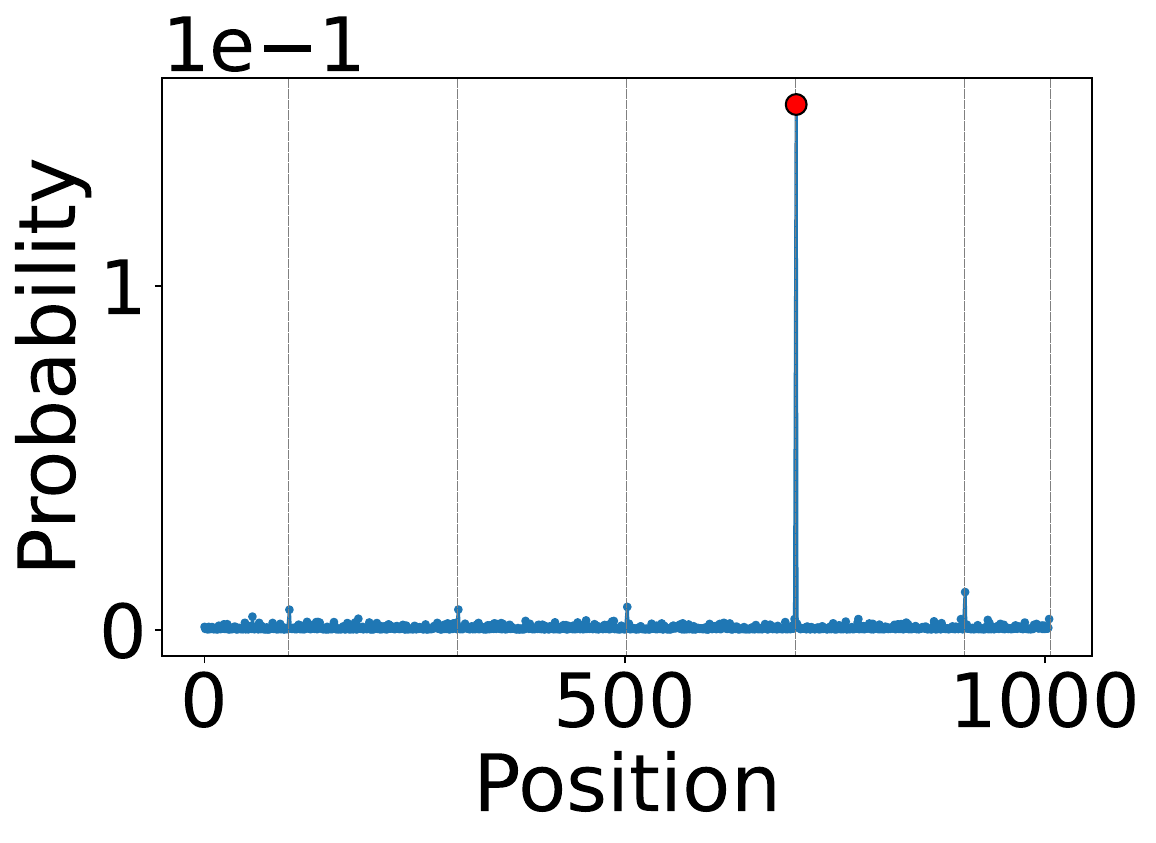} &
\FigGridCell{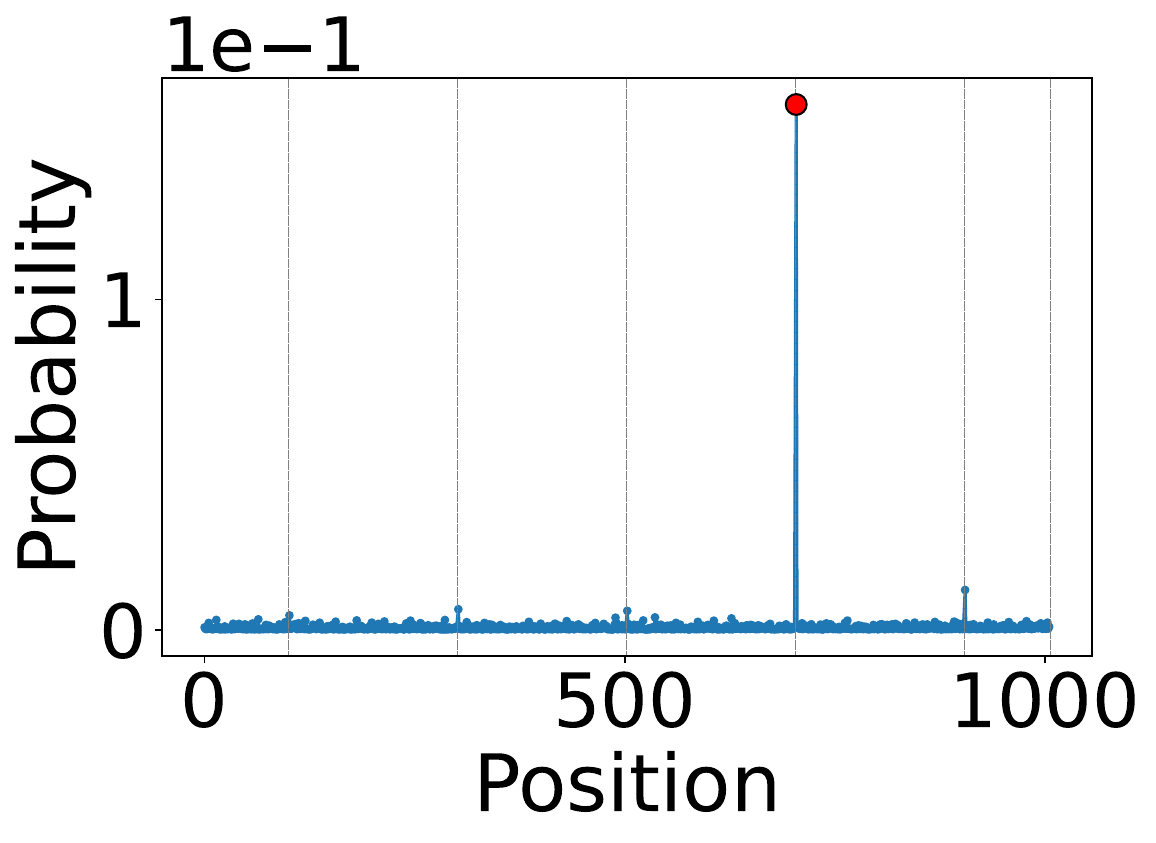} &
\FigGridCell{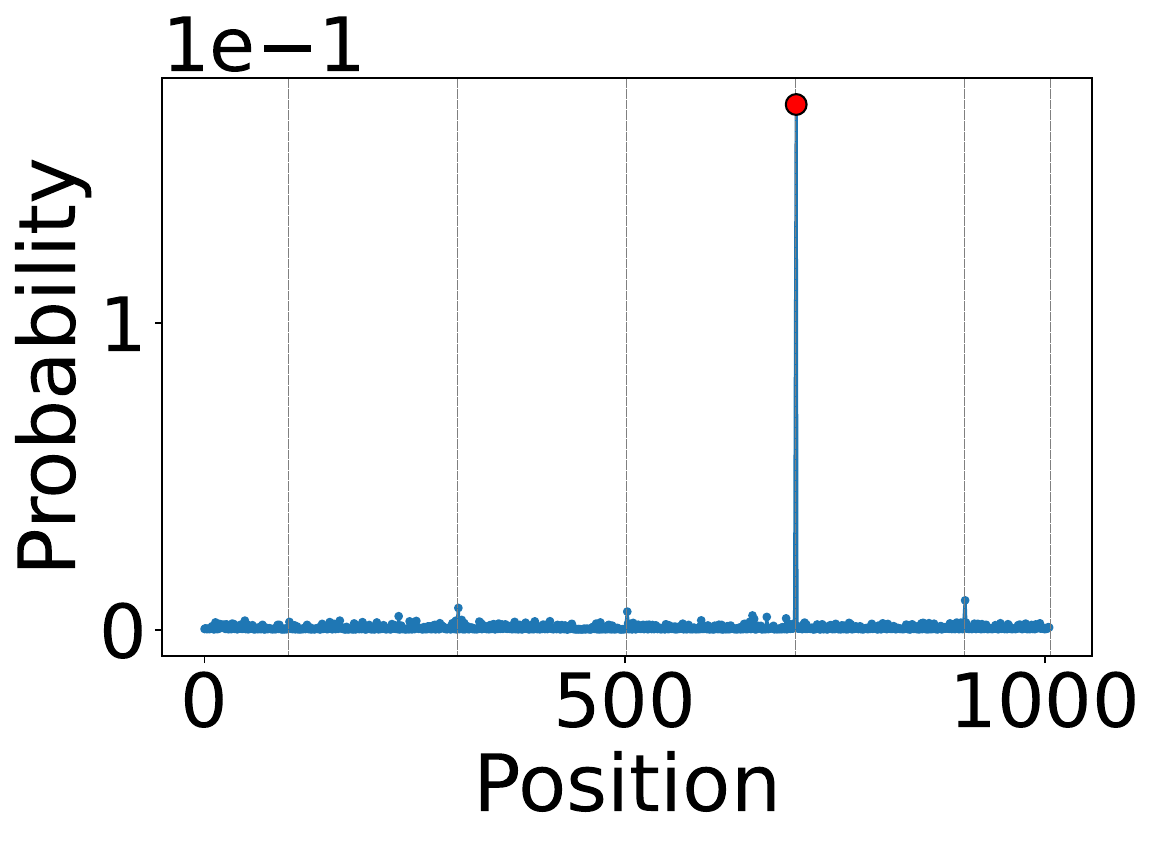} &
\FigGridCell{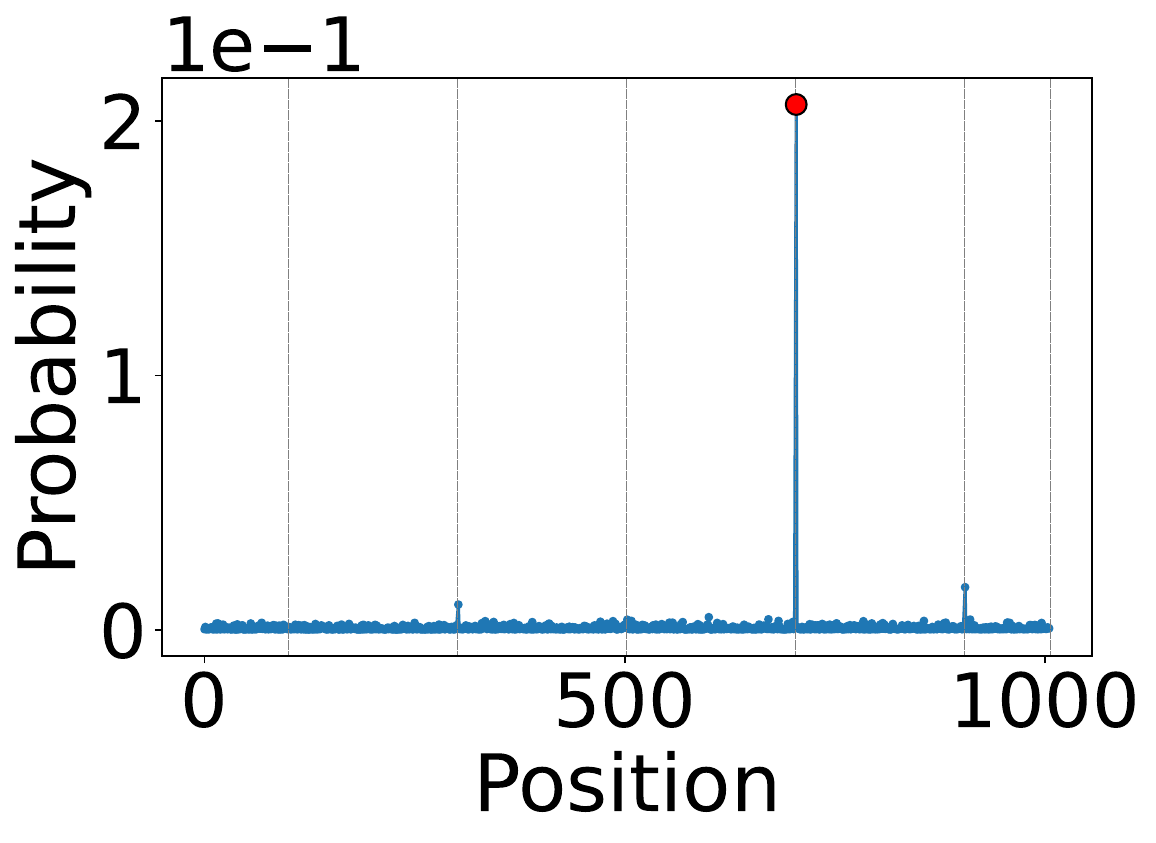} \\
& \FigGridVSubLabel{P5} &
\FigGridCell{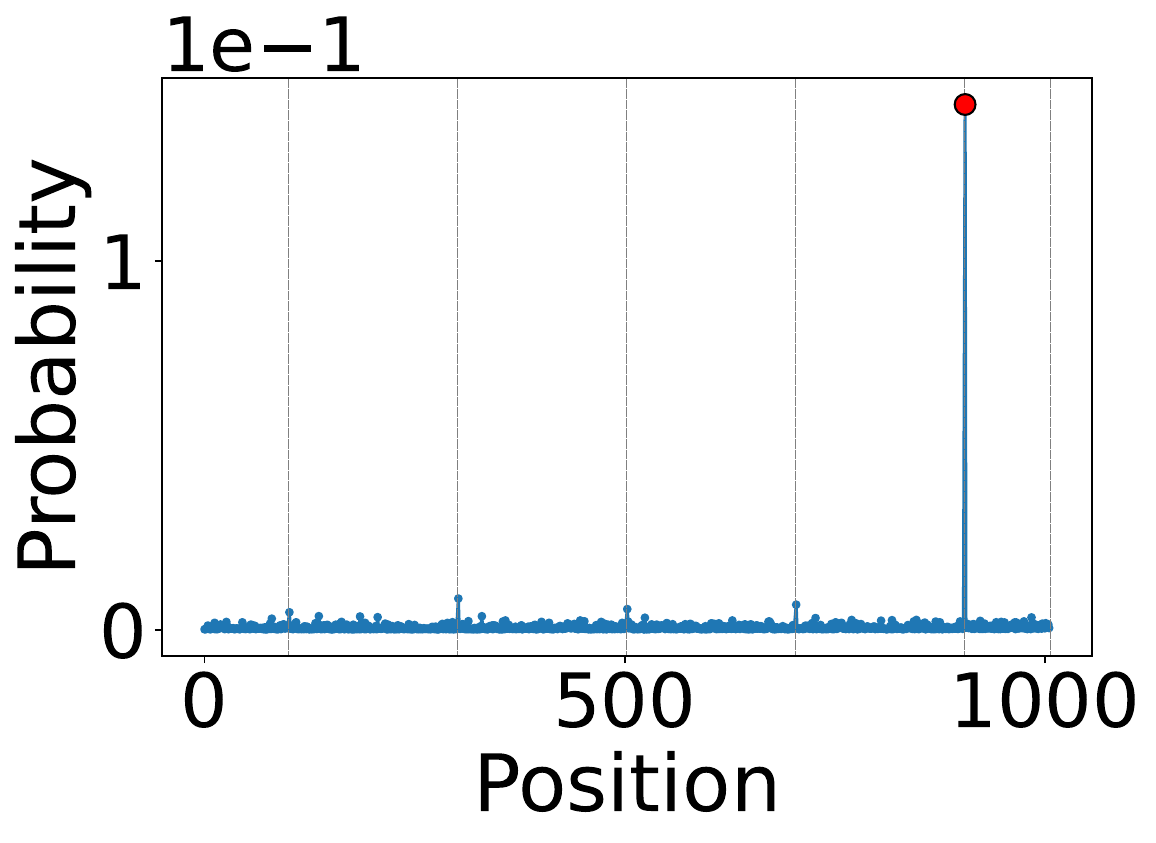} &
\FigGridCell{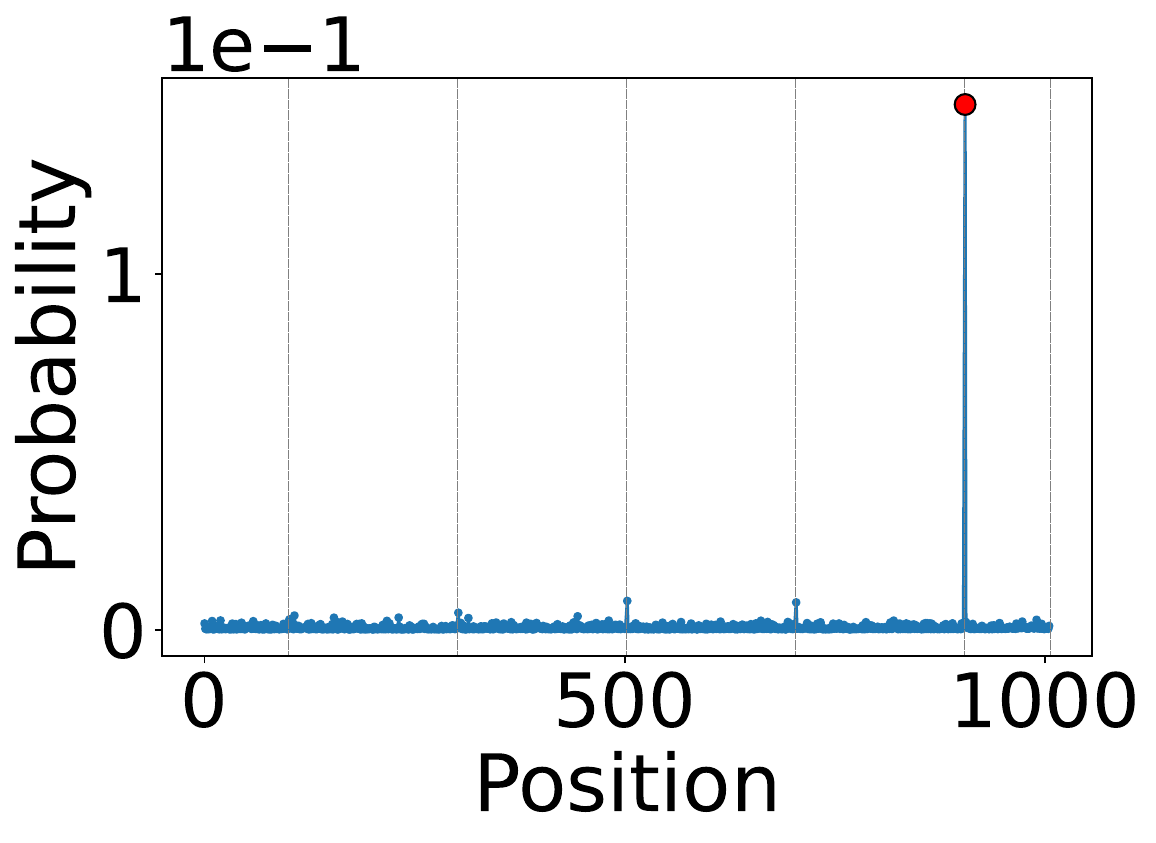} &
\FigGridCell{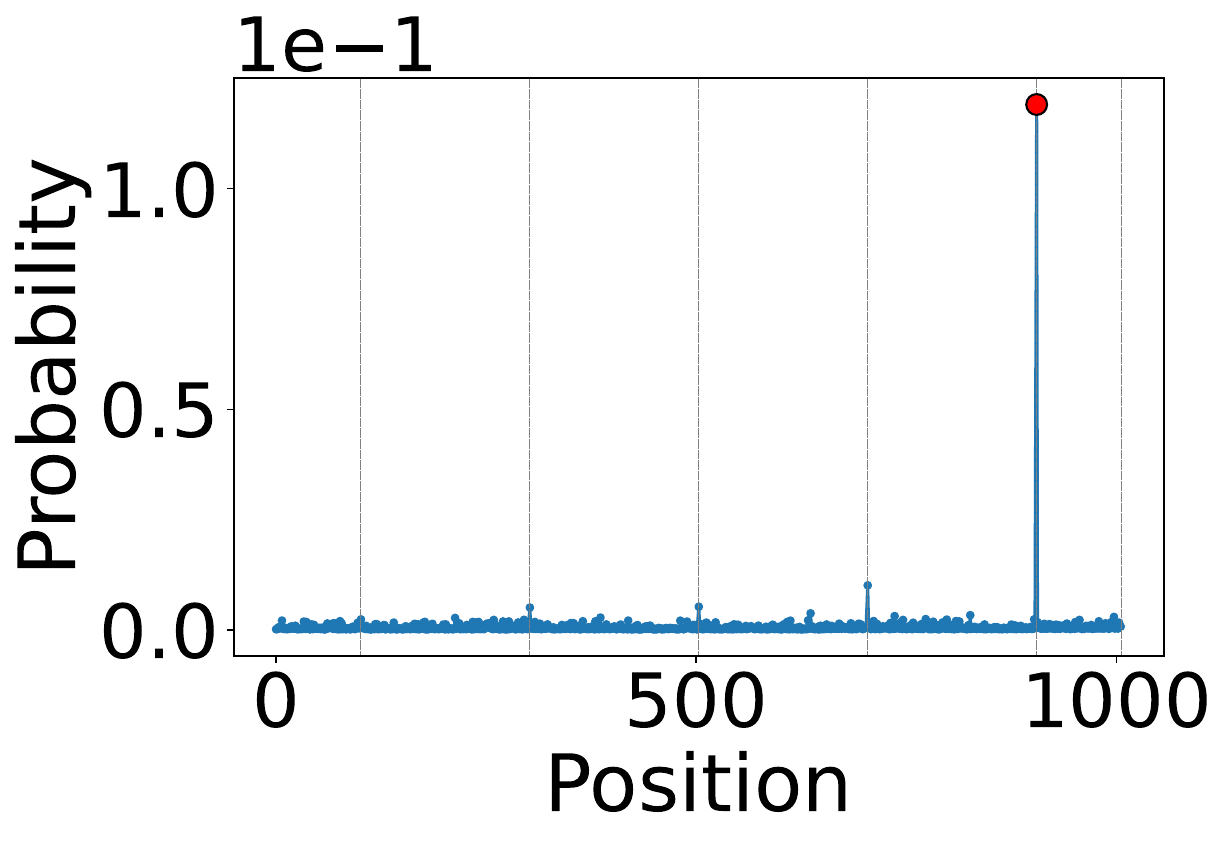} &
\FigGridCell{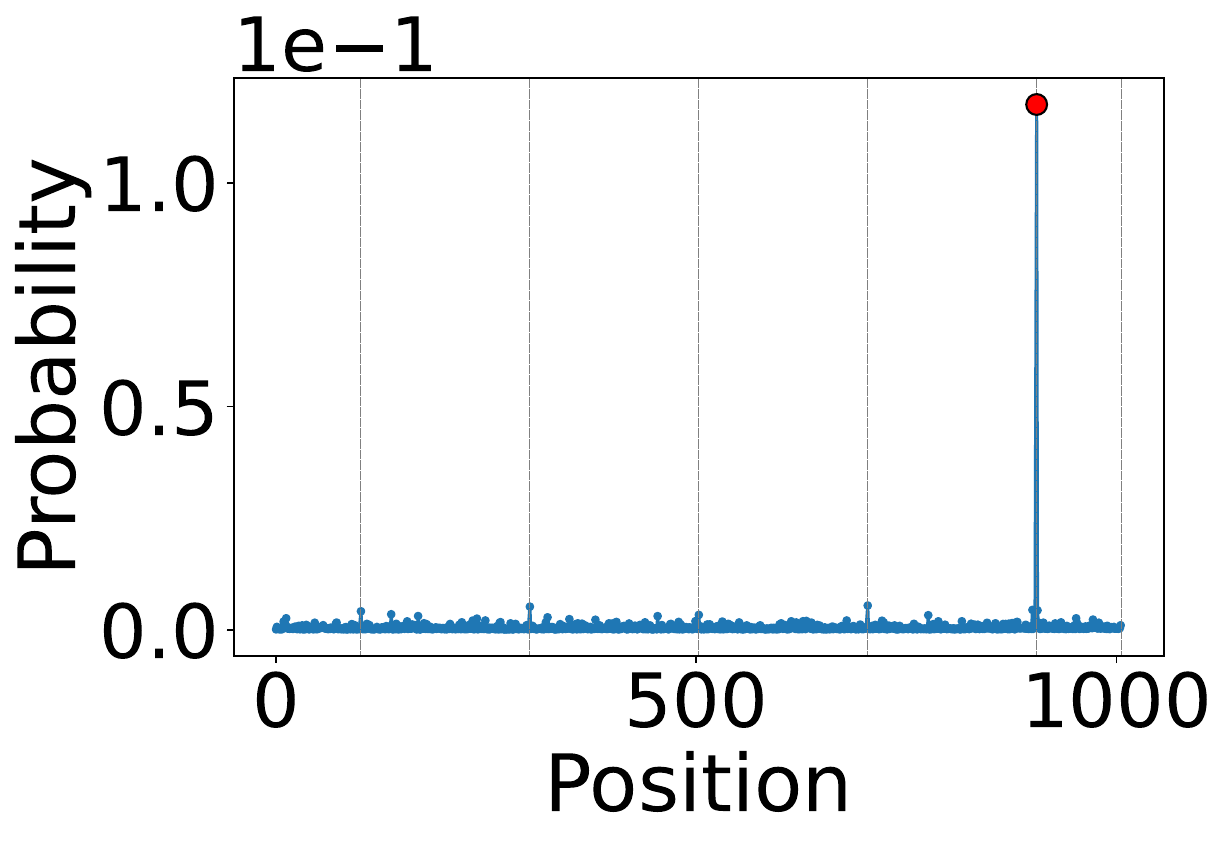} &
\FigGridCell{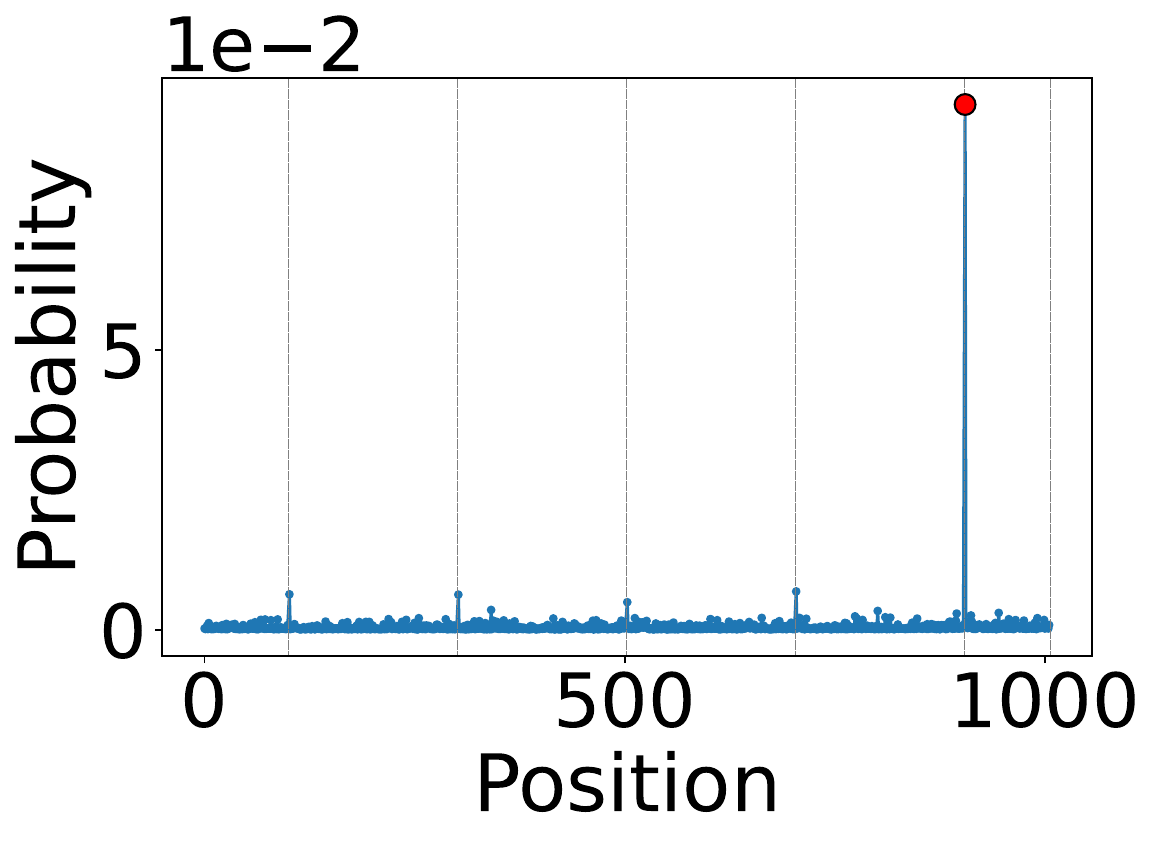} \\
\bottomrule
\end{tabular}
\endgroup
\caption{Mistral ablation effect (Exp. 2). Episodic retrieval probability after ablating Induction or Random heads. Columns show number of ablated heads.}
\label{fig:exp2_mistral_ablation}
\end{figure*}

\begin{figure*}[h!]
\centering
\begingroup
\setlength{\FigGridLabelW}{1.05cm}
\setlength{\FigGridSubLabelW}{0.55cm}
\setlength{\FigGridCellH}{1.85cm} 
\setlength{\FigGridLabelSep}{1pt}
\setlength{\FigGridSubLabelSep}{1pt}
\setlength{\FigGridColSep}{0.5pt}

\FigGridSetupTwo{5}
\renewcommand{\arraystretch}{1.0}

\begin{tabular}{@{}>{\centering\arraybackslash}m{\FigGridLabelW}@{\hspace{\FigGridLabelSep}}
                >{\centering\arraybackslash}m{\FigGridSubLabelW}@{\hspace{\FigGridSubLabelSep}}
                *{5}{>{\centering\arraybackslash}m{\FigGridPlotW}}@{}}
\toprule
& & \multicolumn{5}{c}{\FigHdr Ablations} \\
\cmidrule(lr){3-7}
& & \FigHdr 0 & \FigHdr 1 & \FigHdr 10 & \FigHdr 50 & \FigHdr 100 \\
\midrule

\multirow{5}{*}{\FigGridVLabelSpanFive{Induction}} &
\FigGridVSubLabel{P1} &
\FigGridCell{Figures/ep_prob_without_A_red/Qwen2.5-7B-Instruct_5_Repeats_200_Length_500_Permutations_0_ablations_induction_1_nth.pdf} &
\FigGridCell{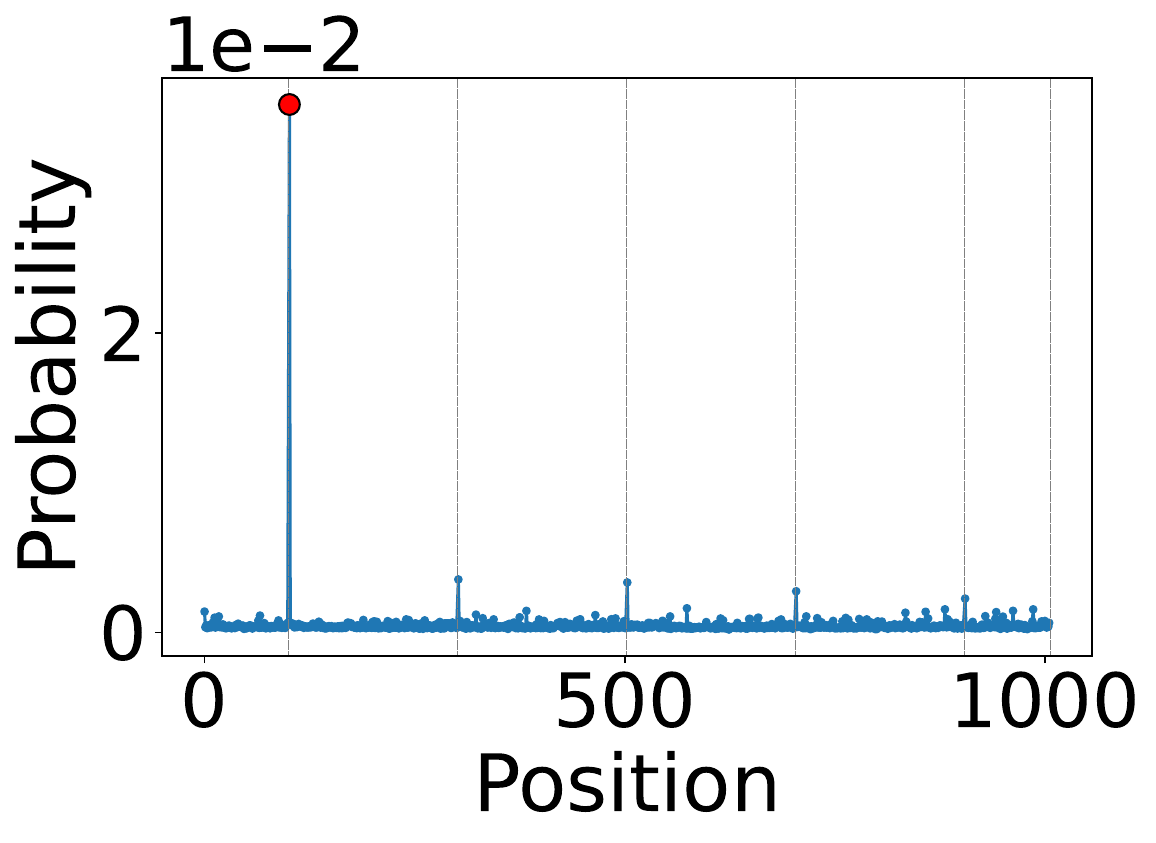} &
\FigGridCell{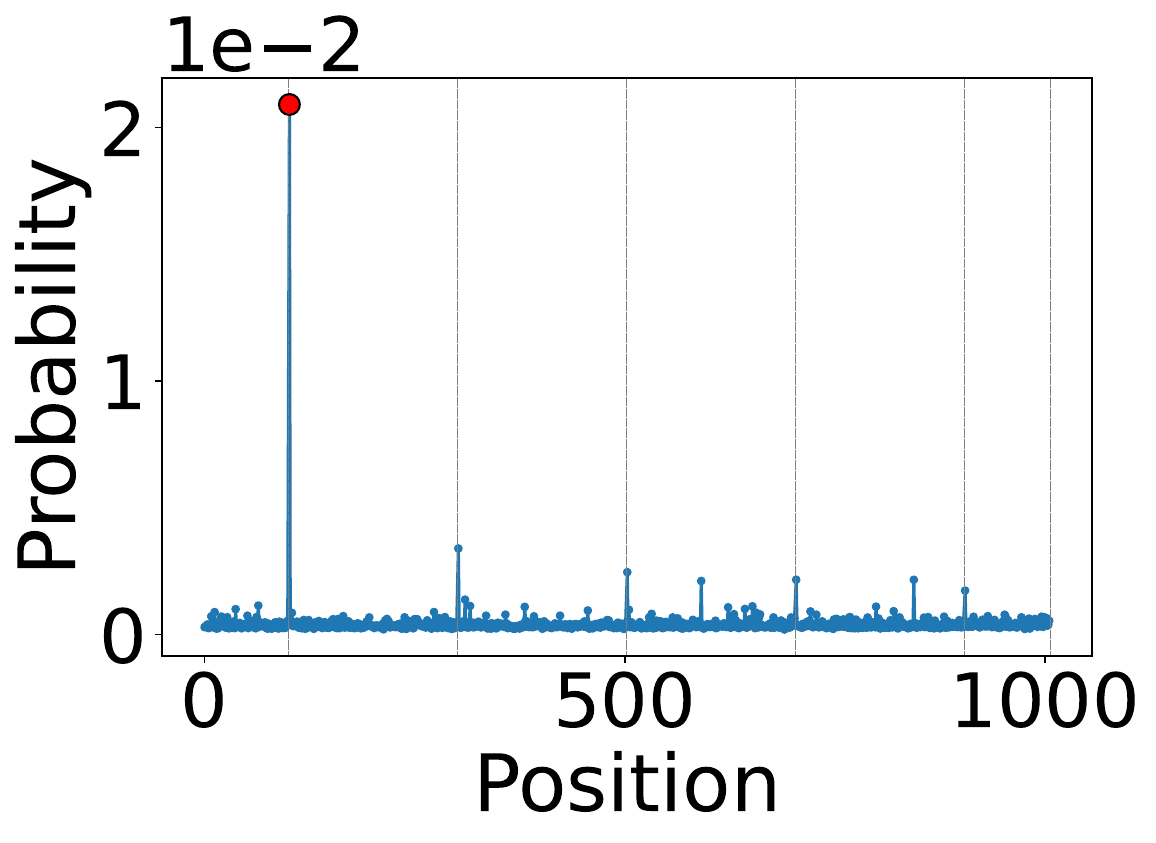} &
\FigGridCell{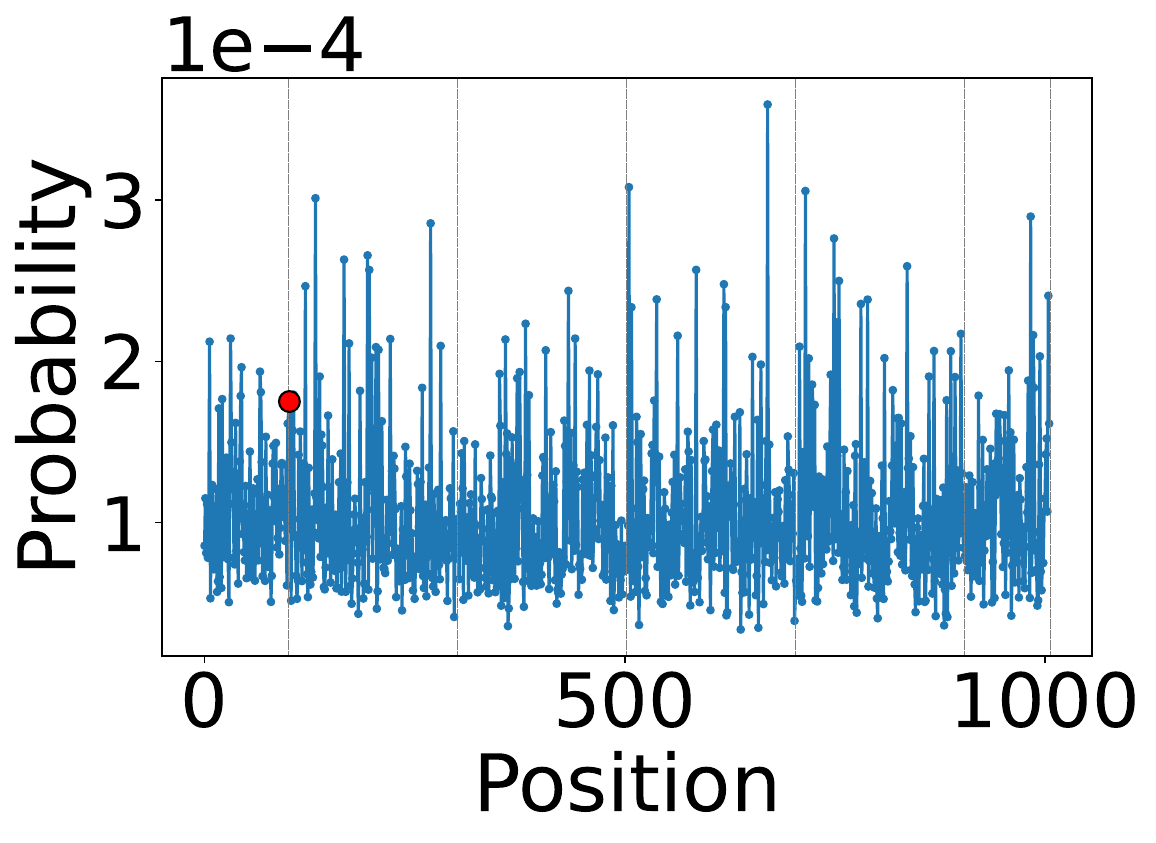} &
\FigGridCell{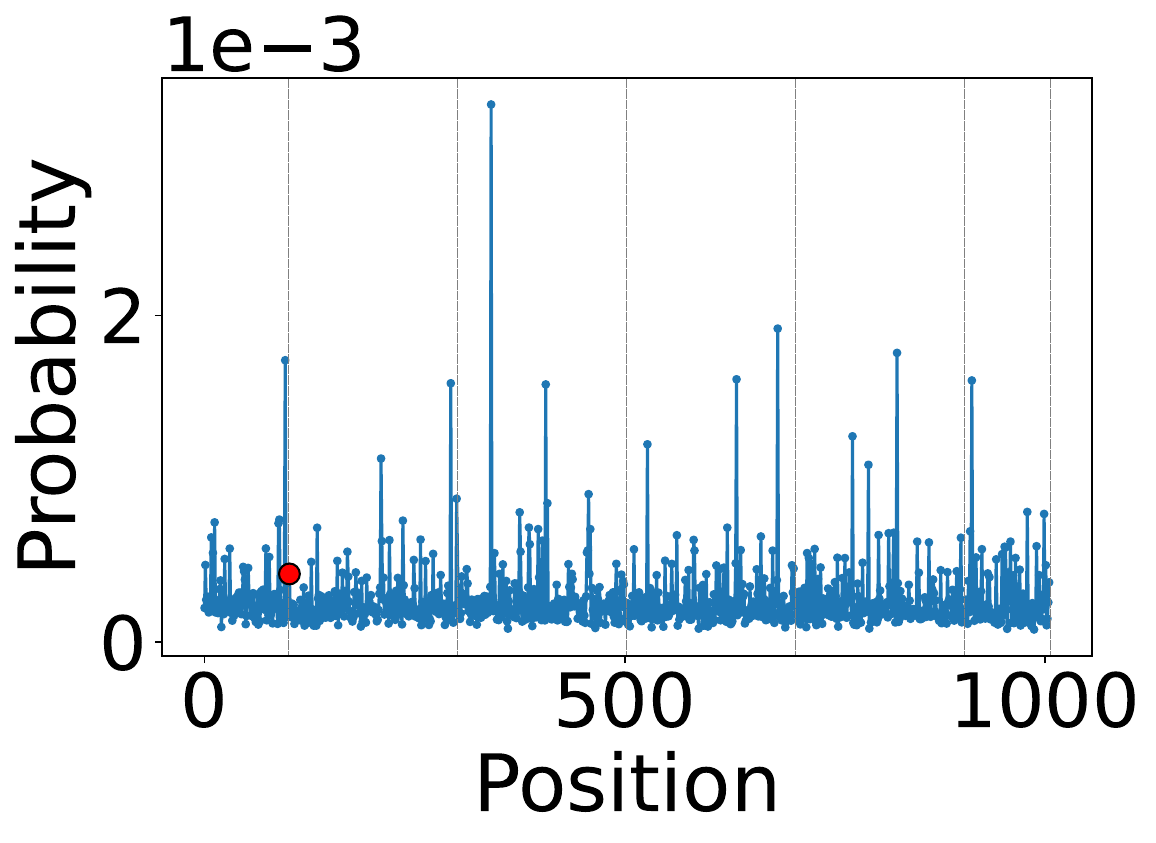} \\

& \FigGridVSubLabel{P2} &
\FigGridCell{Figures/ep_prob_without_A_red/Qwen2.5-7B-Instruct_5_Repeats_200_Length_500_Permutations_0_ablations_induction_2_nth.pdf} &
\FigGridCell{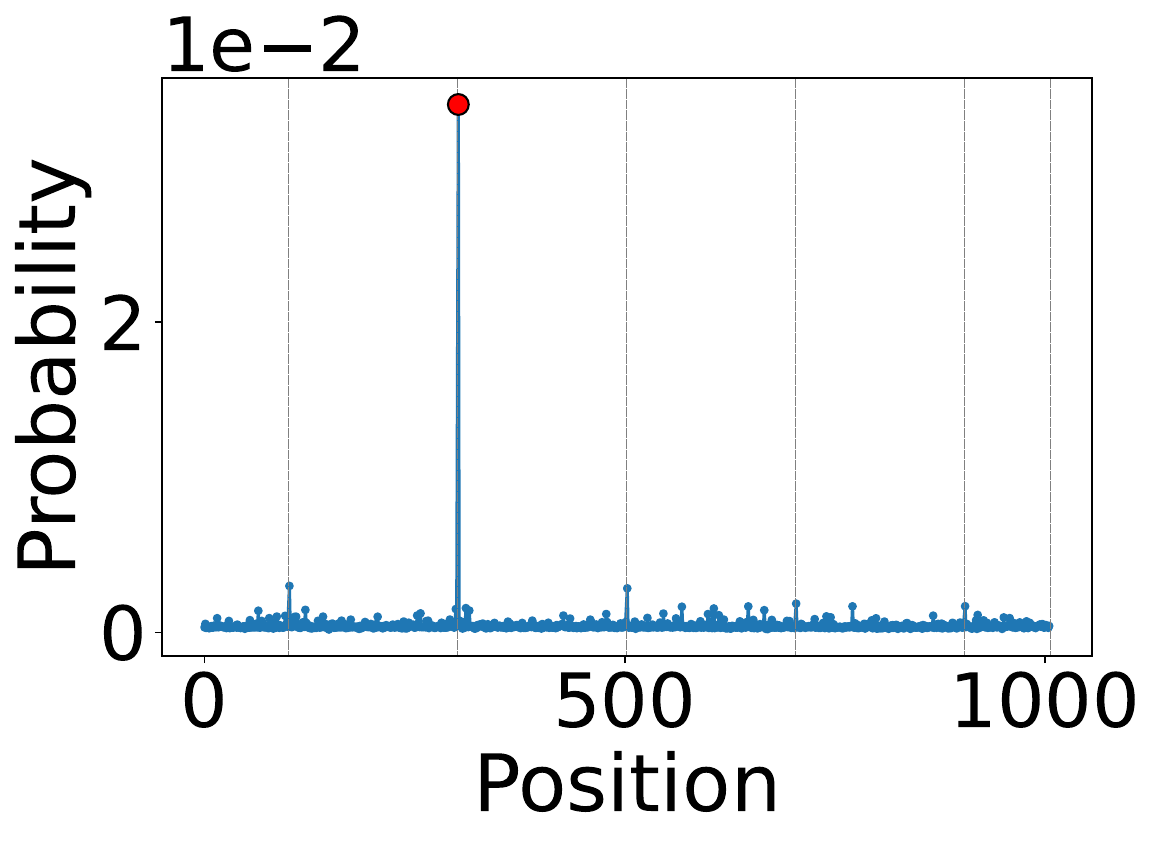} &
\FigGridCell{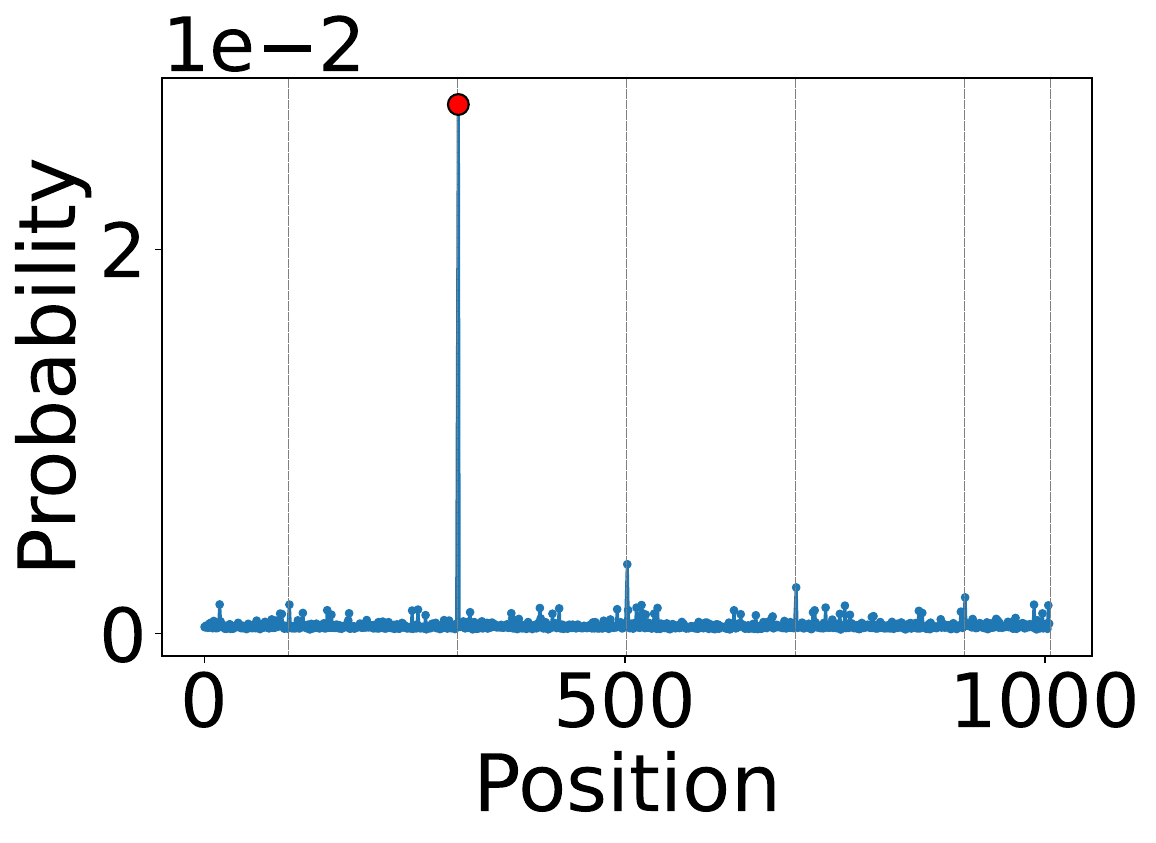} &
\FigGridCell{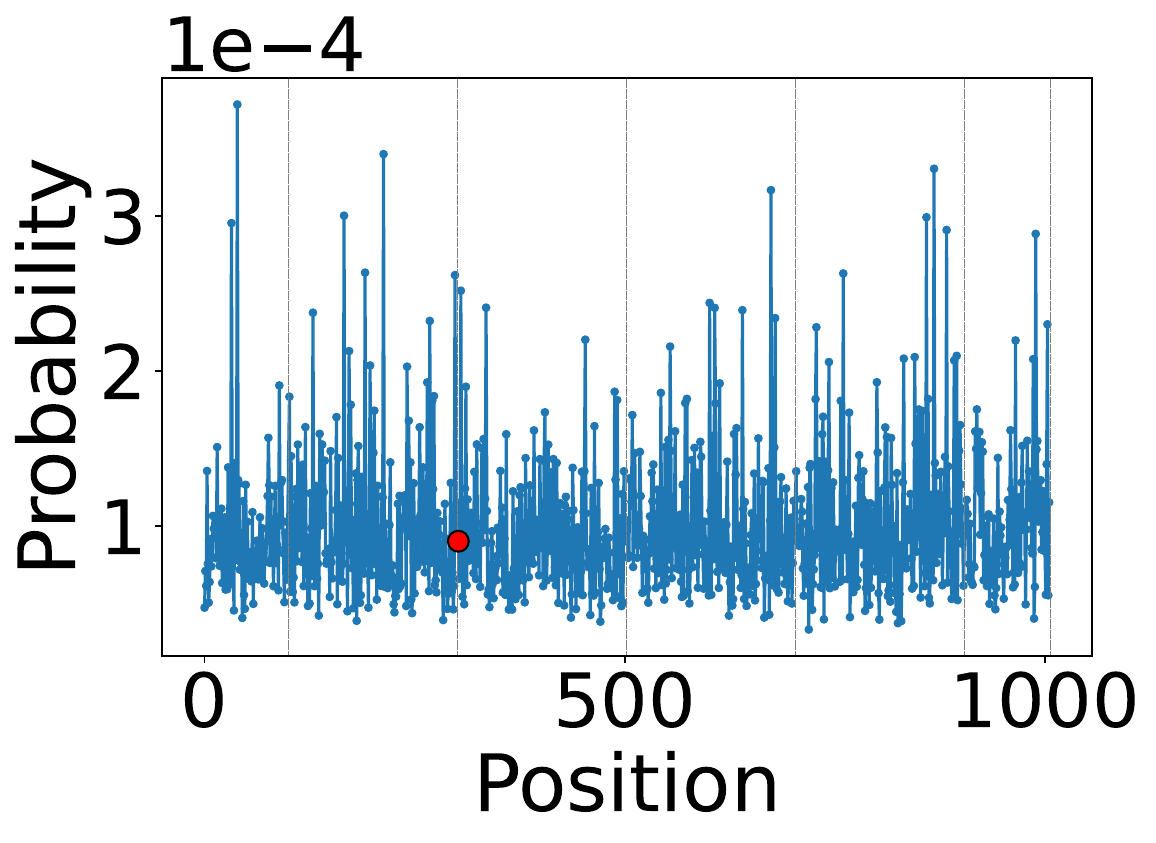} &
\FigGridCell{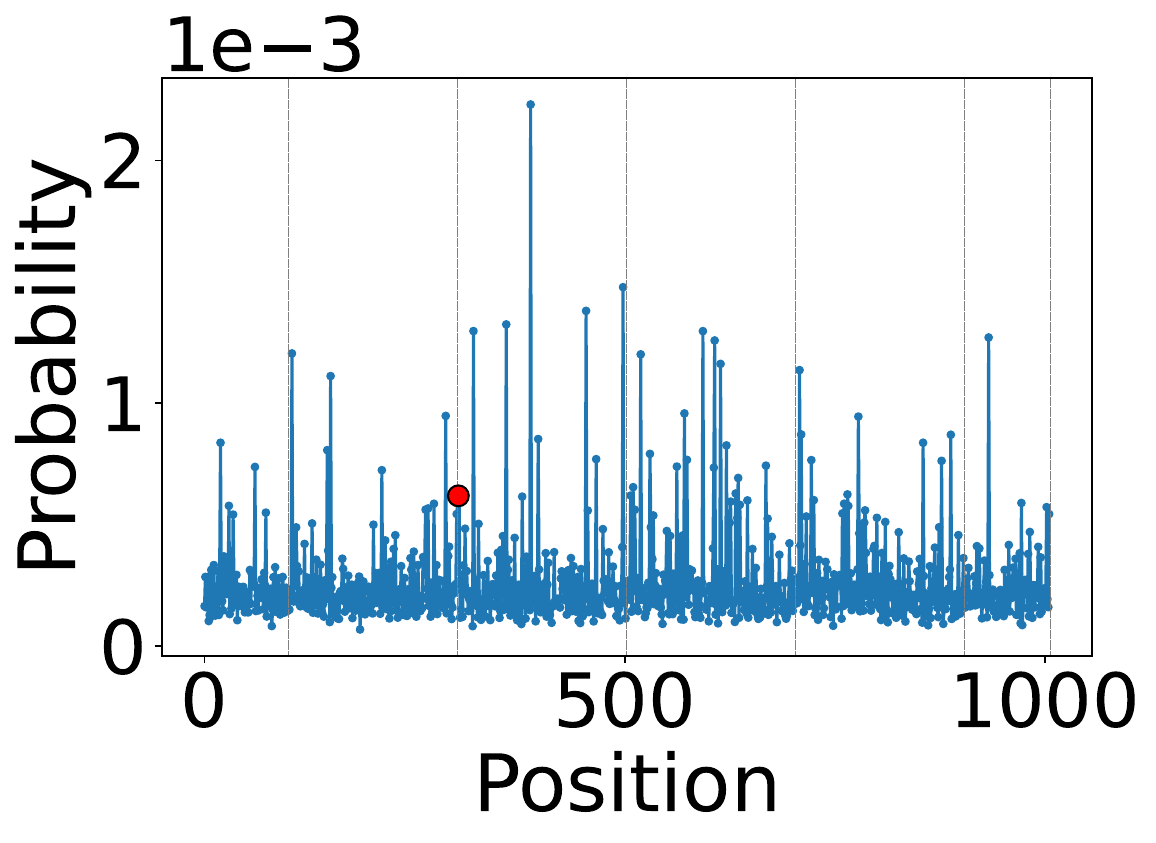} \\

& \FigGridVSubLabel{P3} &
\FigGridCell{Figures/ep_prob_without_A_red/Qwen2.5-7B-Instruct_5_Repeats_200_Length_500_Permutations_0_ablations_induction_3_nth.pdf} &
\FigGridCell{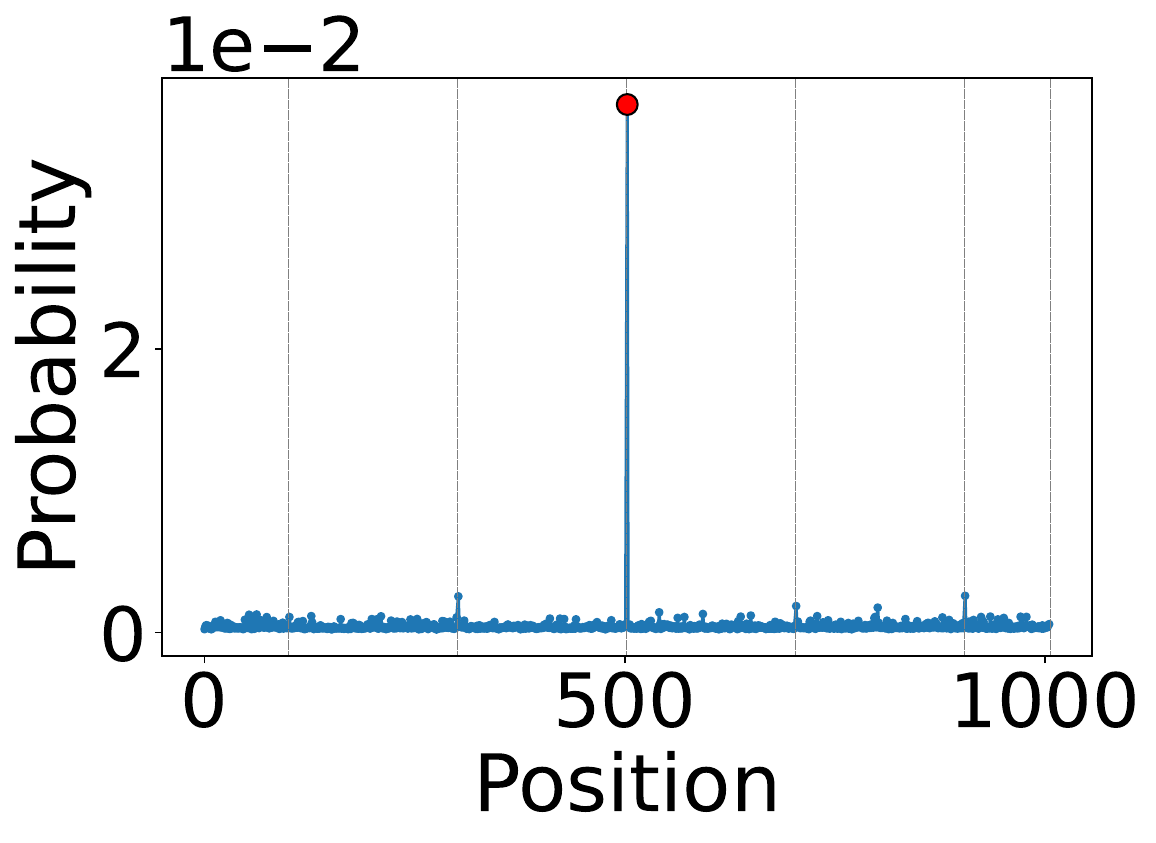} &
\FigGridCell{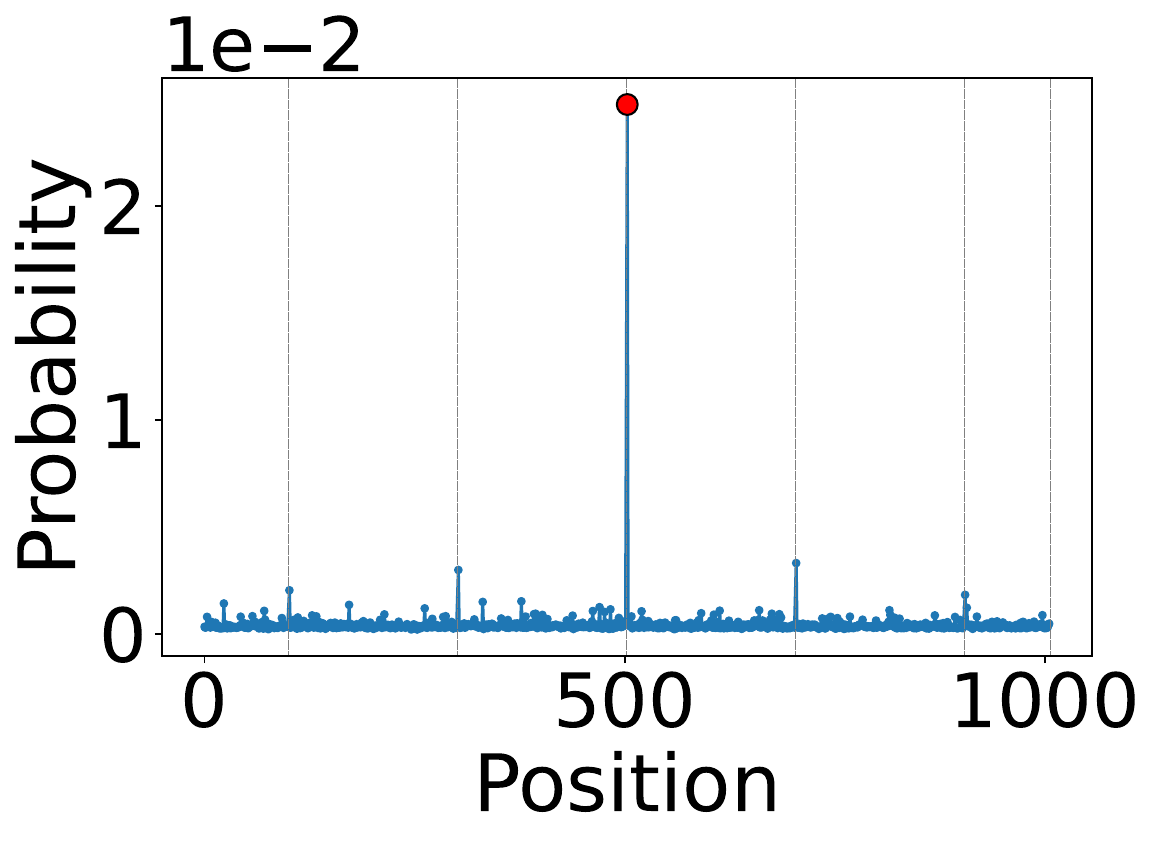} &
\FigGridCell{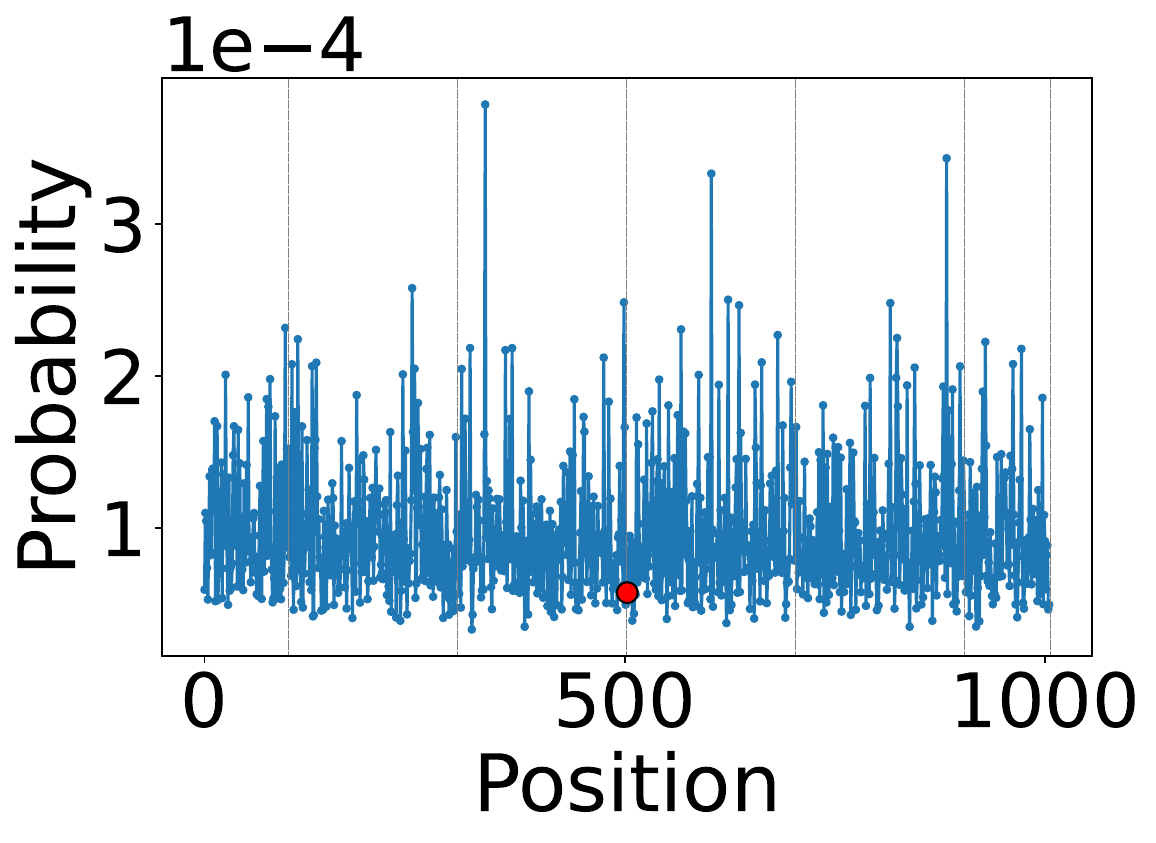} &
\FigGridCell{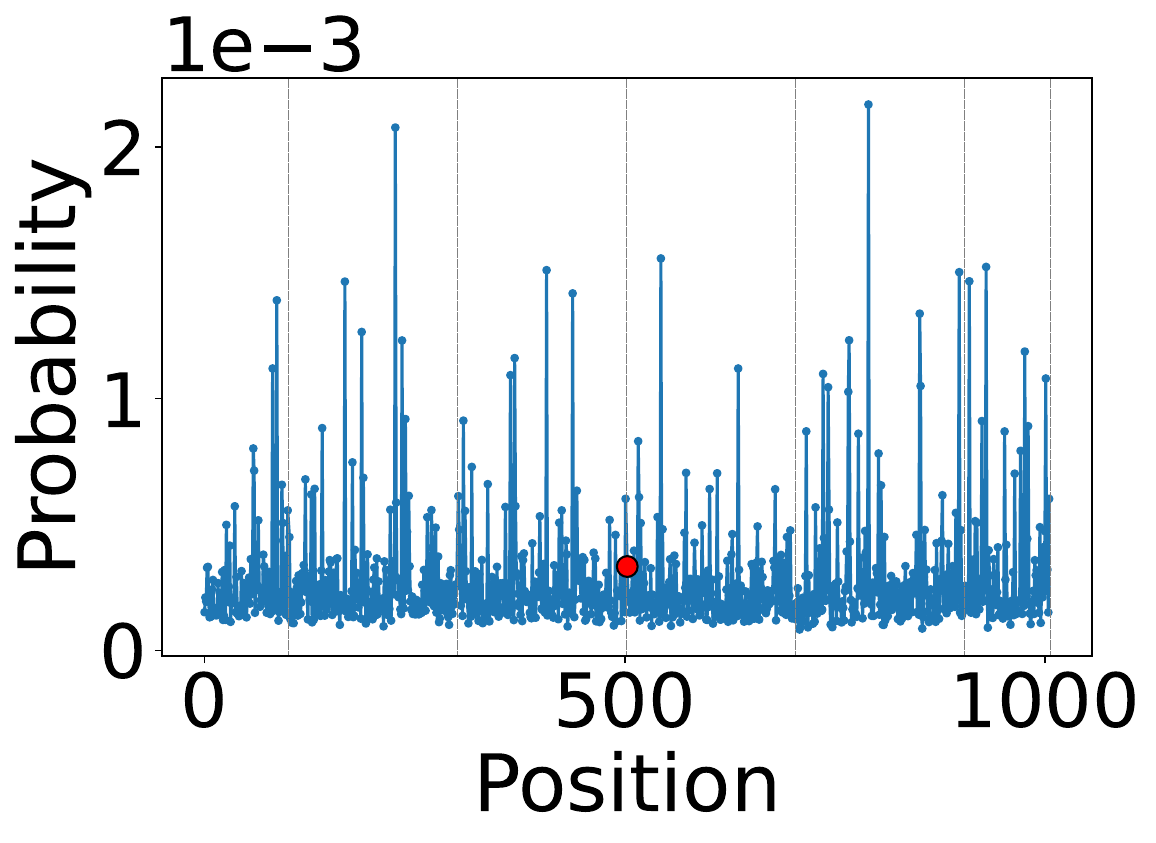} \\

& \FigGridVSubLabel{P4} &
\FigGridCell{Figures/ep_prob_without_A_red/Qwen2.5-7B-Instruct_5_Repeats_200_Length_500_Permutations_0_ablations_induction_4_nth.pdf} &
\FigGridCell{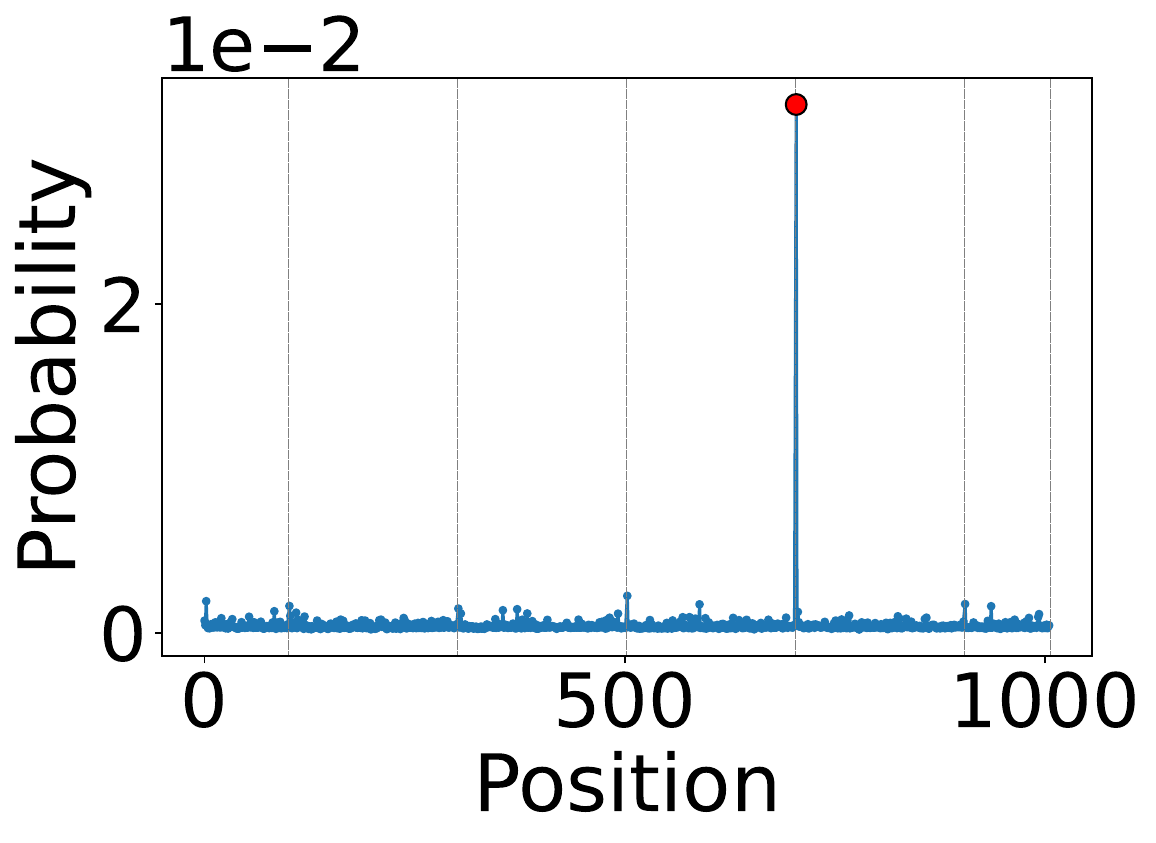} &
\FigGridCell{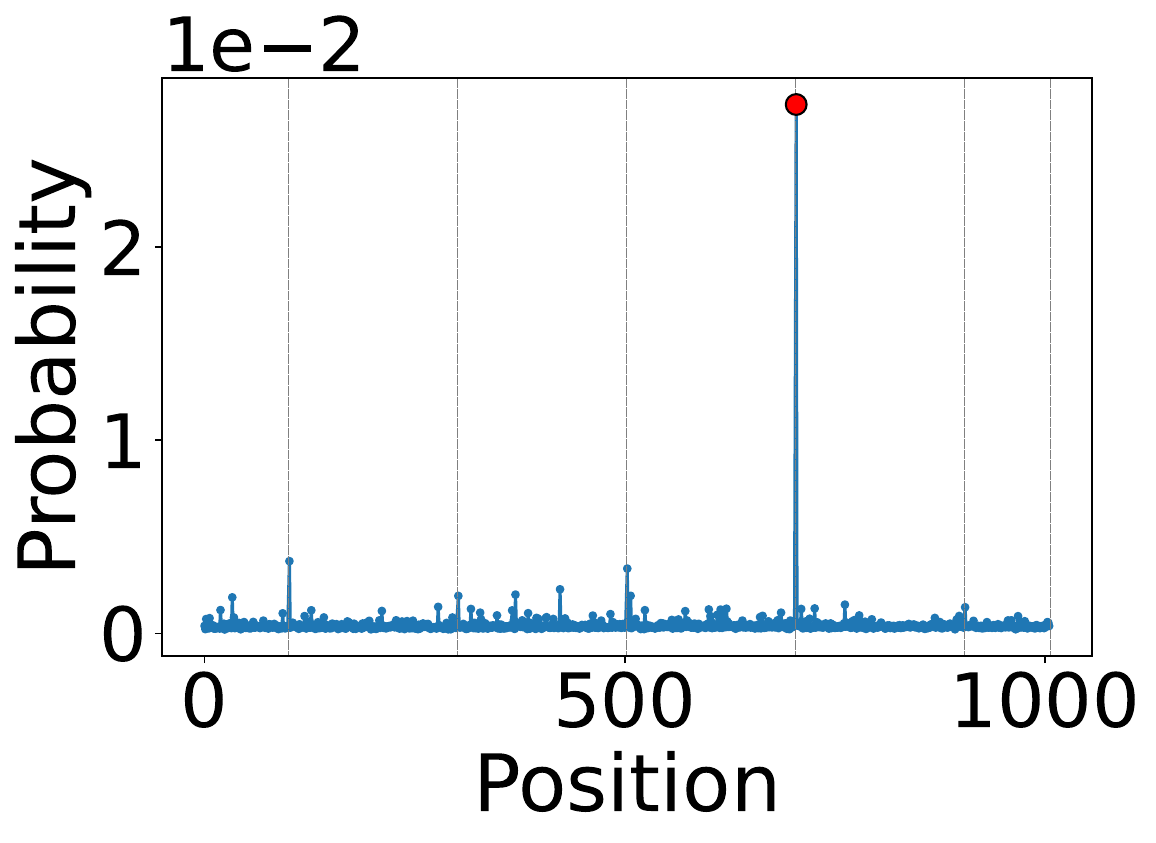} &
\FigGridCell{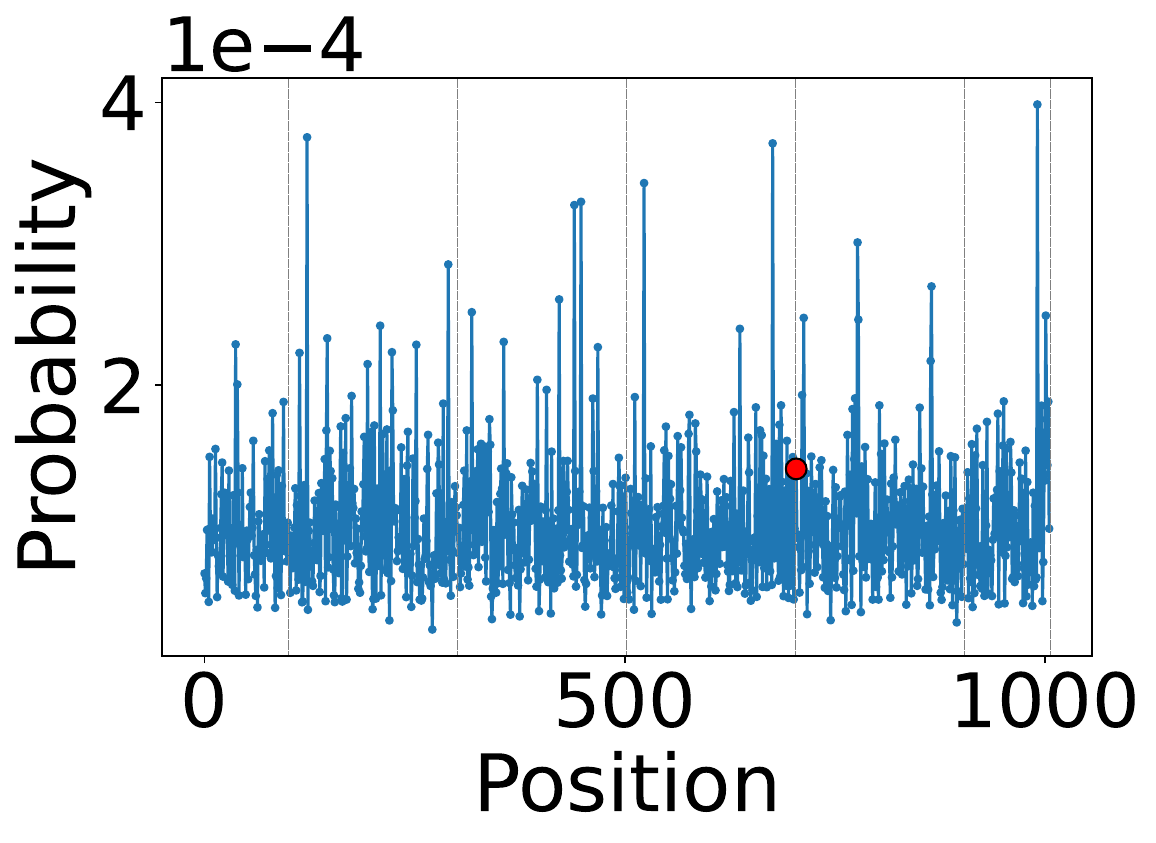} &
\FigGridCell{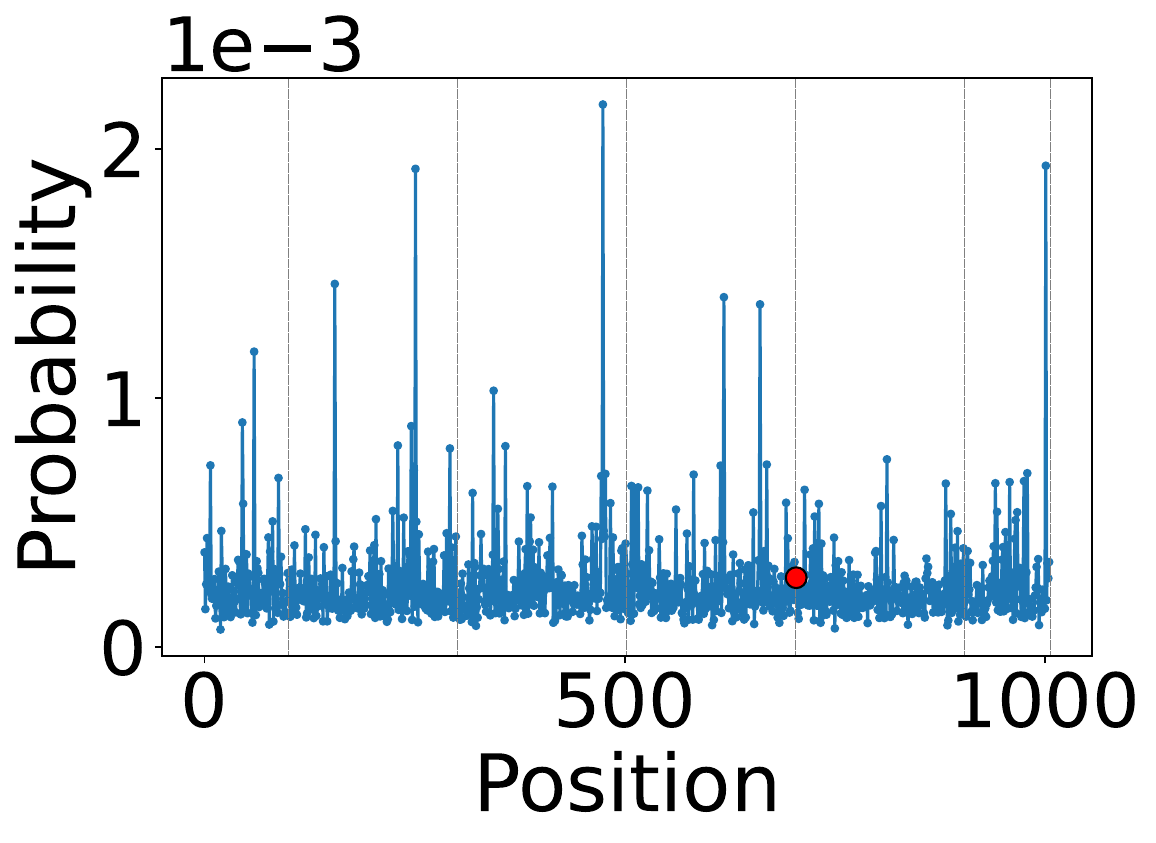} \\

& \FigGridVSubLabel{P5} &
\FigGridCell{Figures/ep_prob_without_A_red/Qwen2.5-7B-Instruct_5_Repeats_200_Length_500_Permutations_0_ablations_induction_5_nth.pdf} &
\FigGridCell{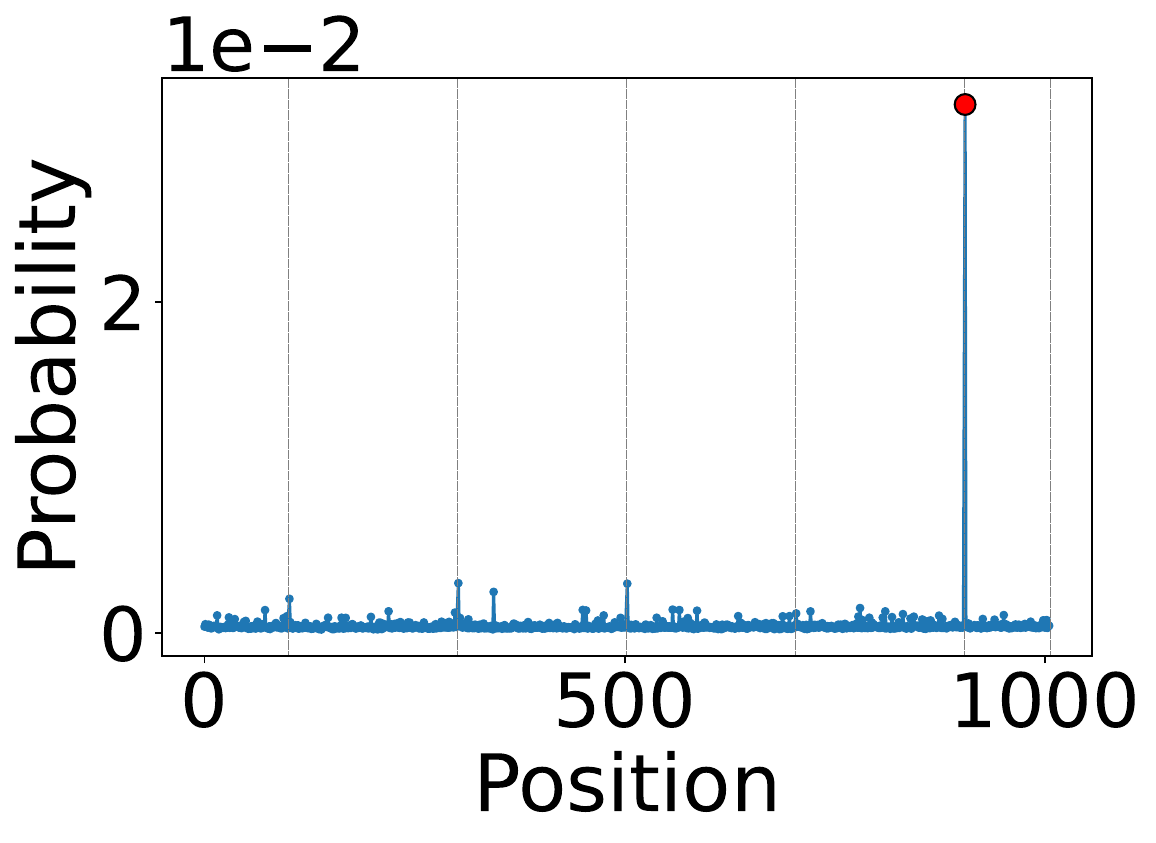} &
\FigGridCell{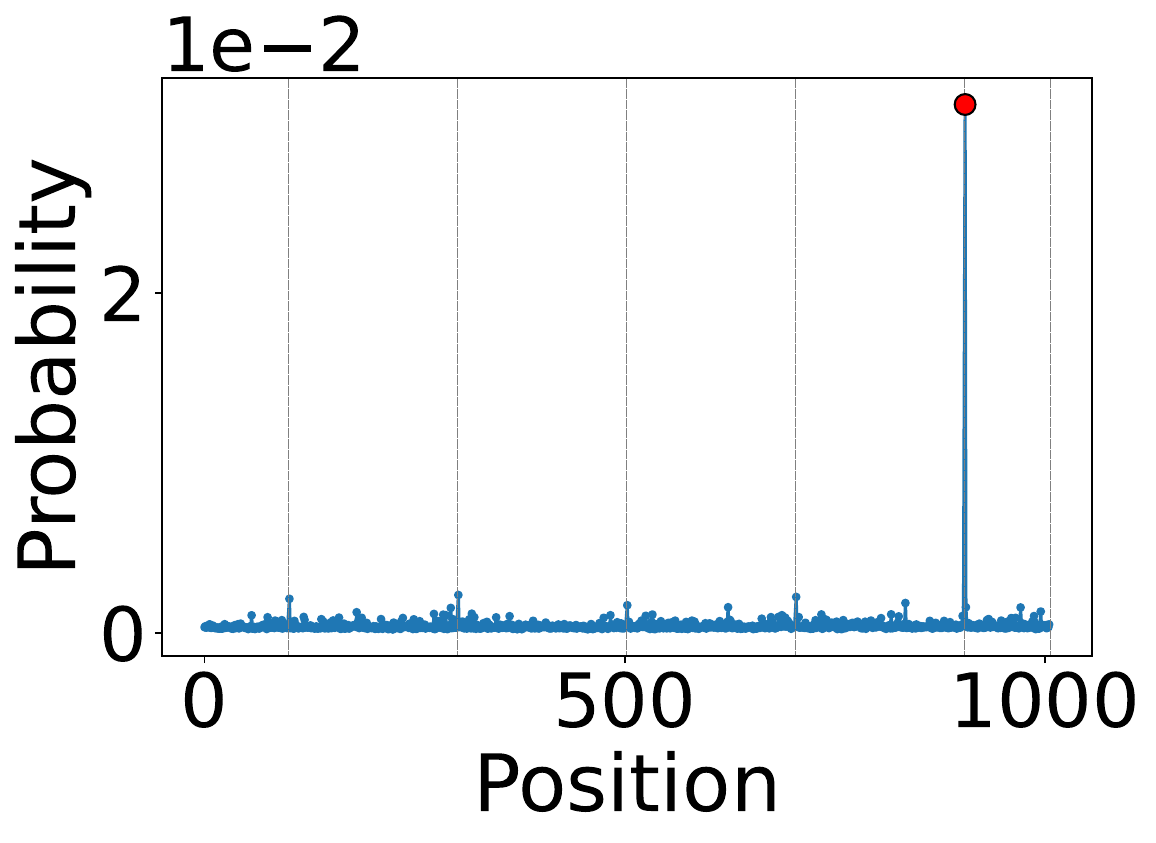} &
\FigGridCell{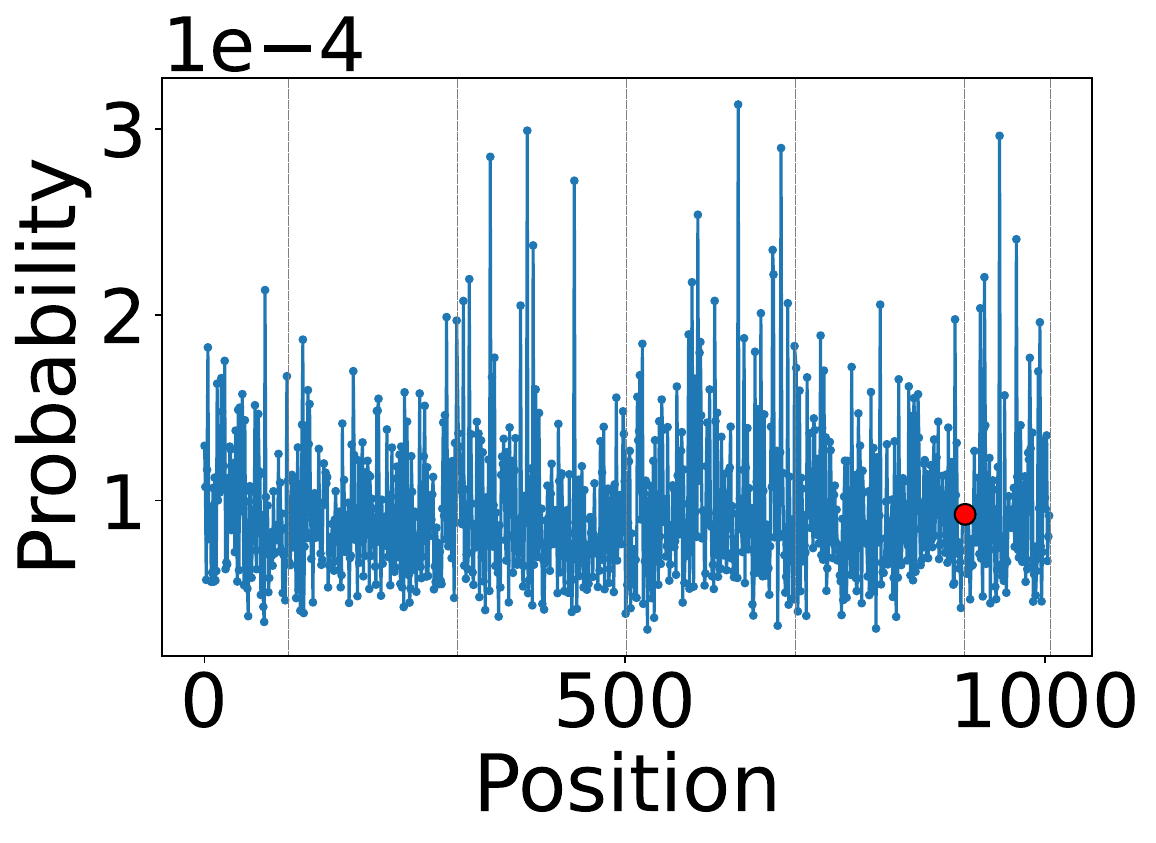} &
\FigGridCell{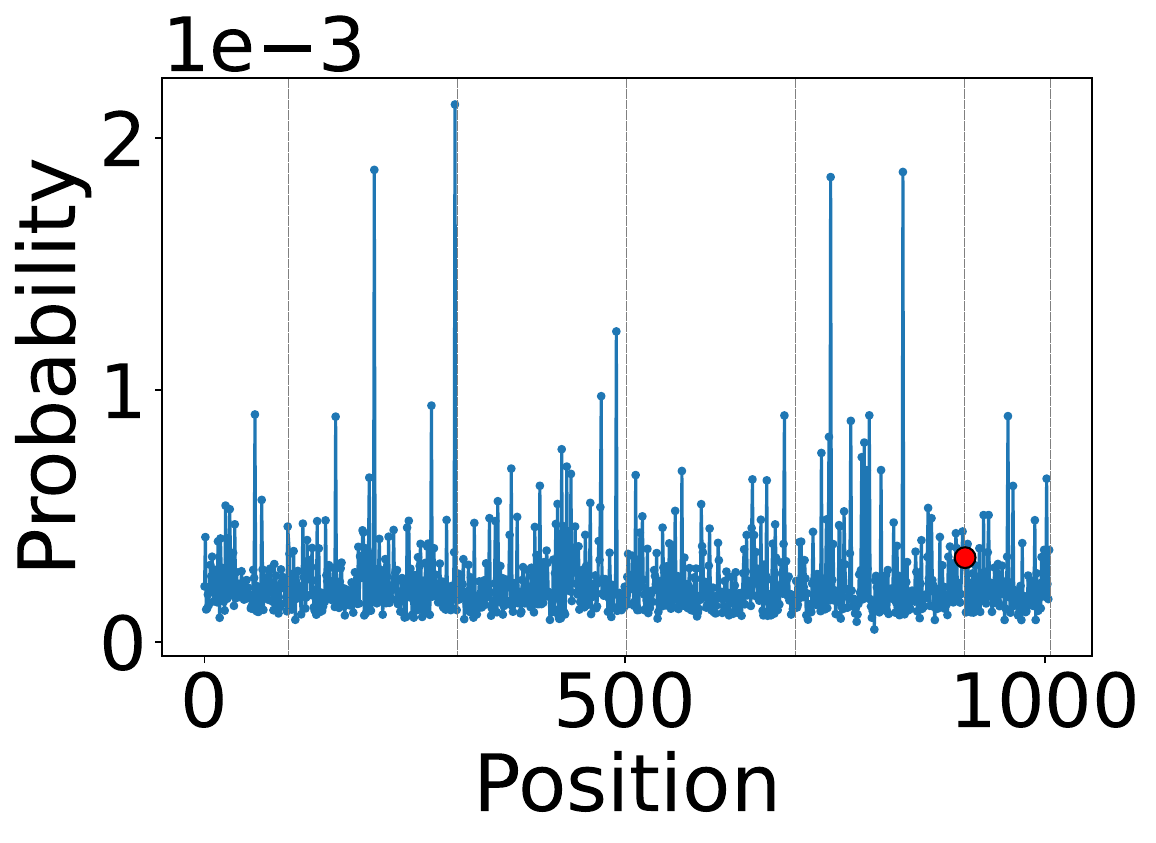} \\

\addlinespace[1pt]
\midrule
\addlinespace[1pt]

\multirow{5}{*}{\FigGridVLabelSpanFive{Random}} &
\FigGridVSubLabel{P1} &
\FigGridCell{Figures/ep_prob_without_A_red/Qwen2.5-7B-Instruct_5_Repeats_200_Length_500_Permutations_0_ablations_induction_1_nth.pdf} &
\FigGridCell{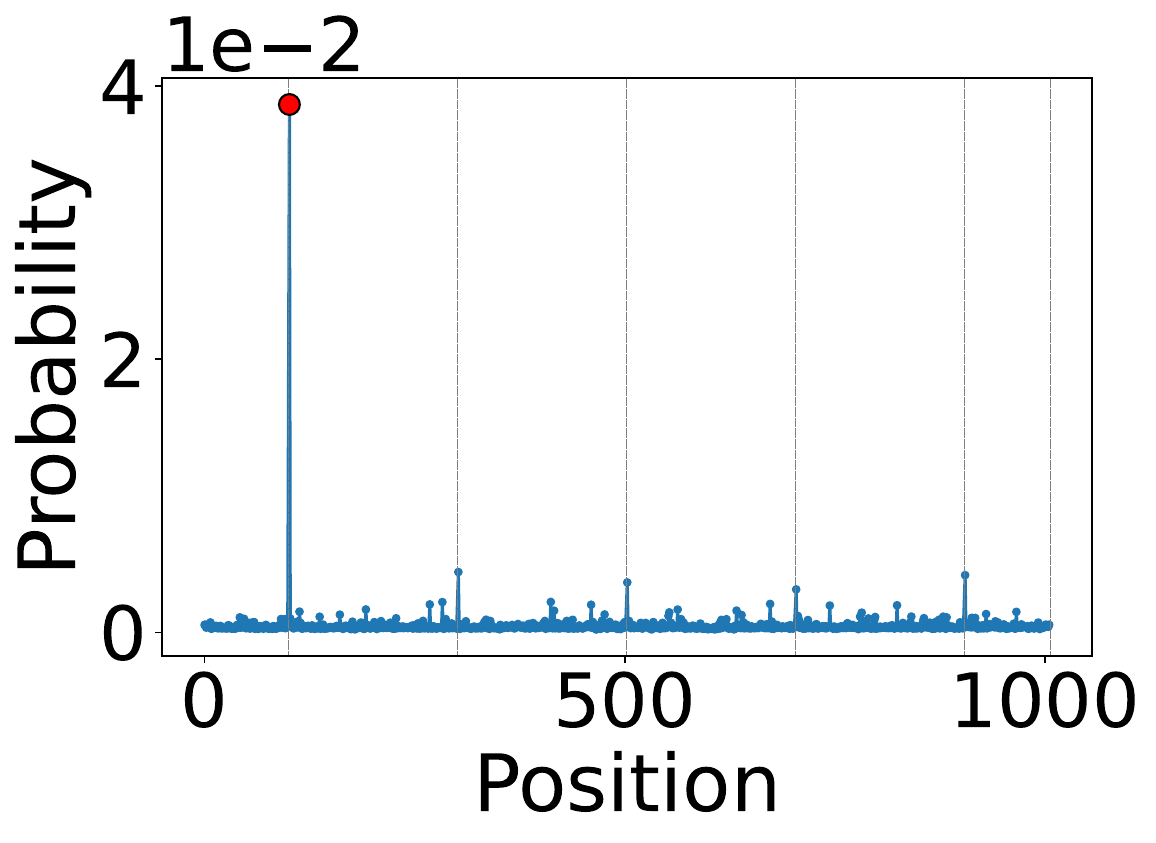} &
\FigGridCell{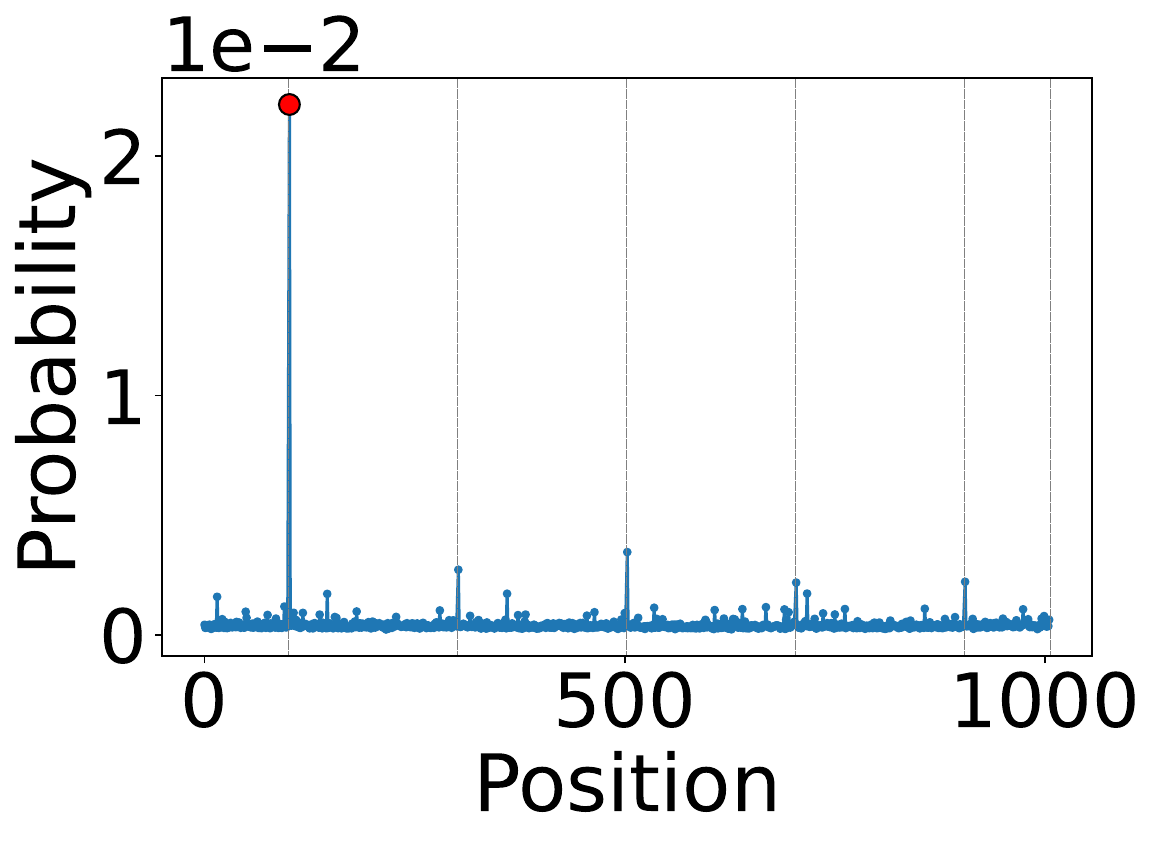} &
\FigGridCell{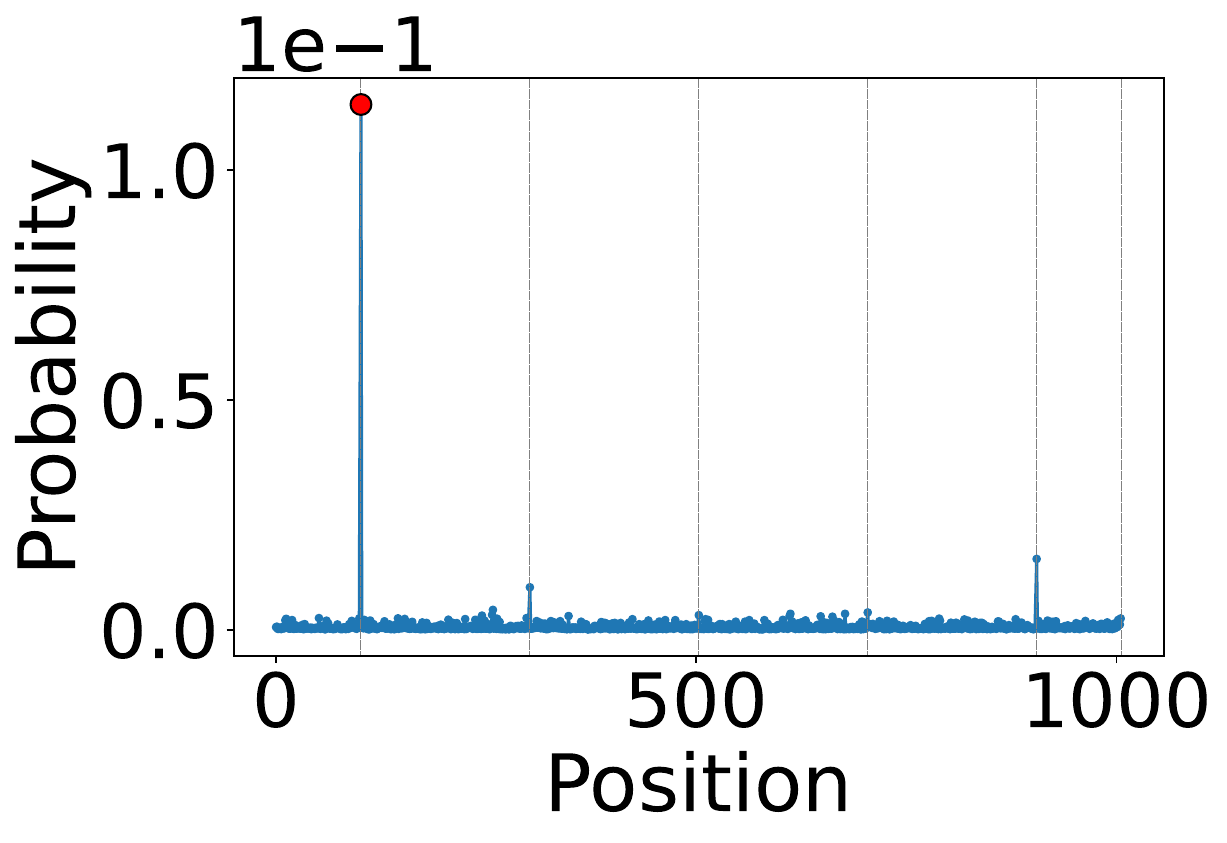} &
\FigGridCell{Figures/ep_prob_without_A_red/Qwen2.5-7B-Instruct_5_Repeats_200_Length_500_Permutations_100_ablations_random_1_nth.pdf} \\

& \FigGridVSubLabel{P2} &
\FigGridCell{Figures/ep_prob_without_A_red/Qwen2.5-7B-Instruct_5_Repeats_200_Length_500_Permutations_0_ablations_induction_2_nth.pdf} &
\FigGridCell{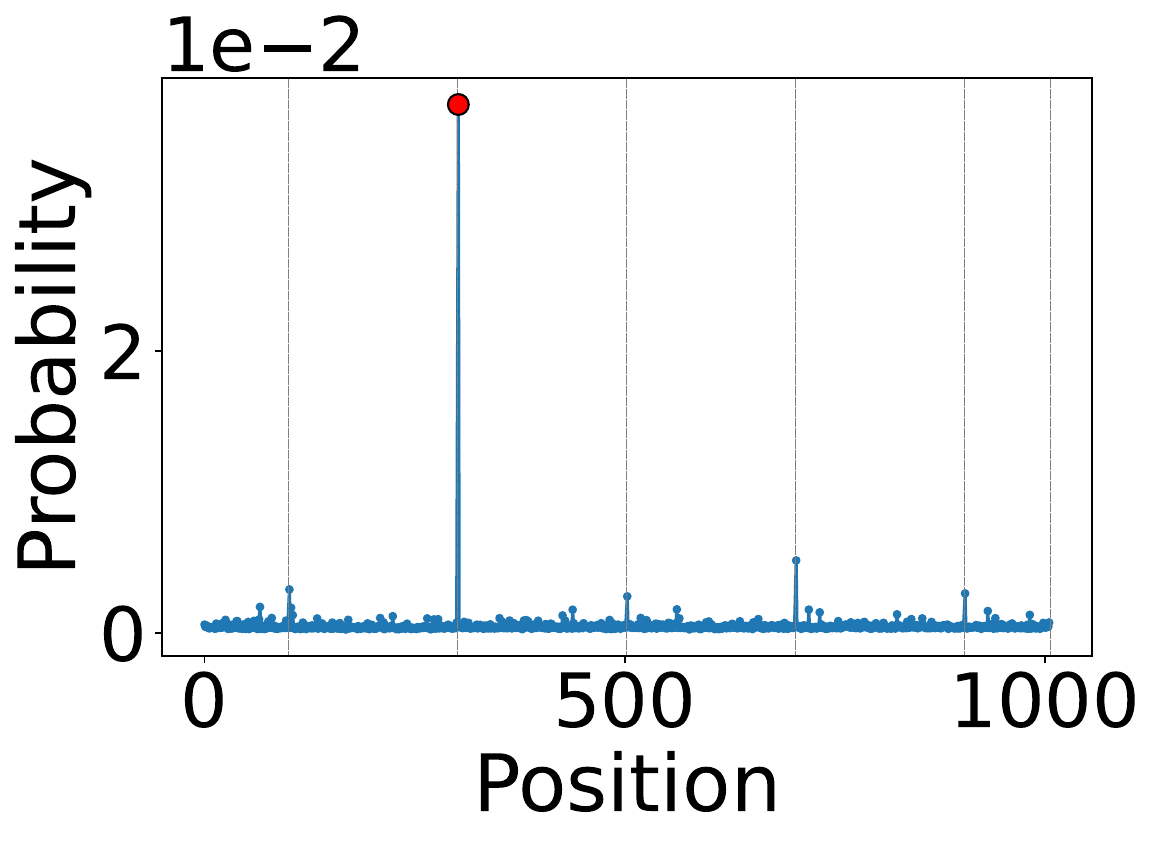} &
\FigGridCell{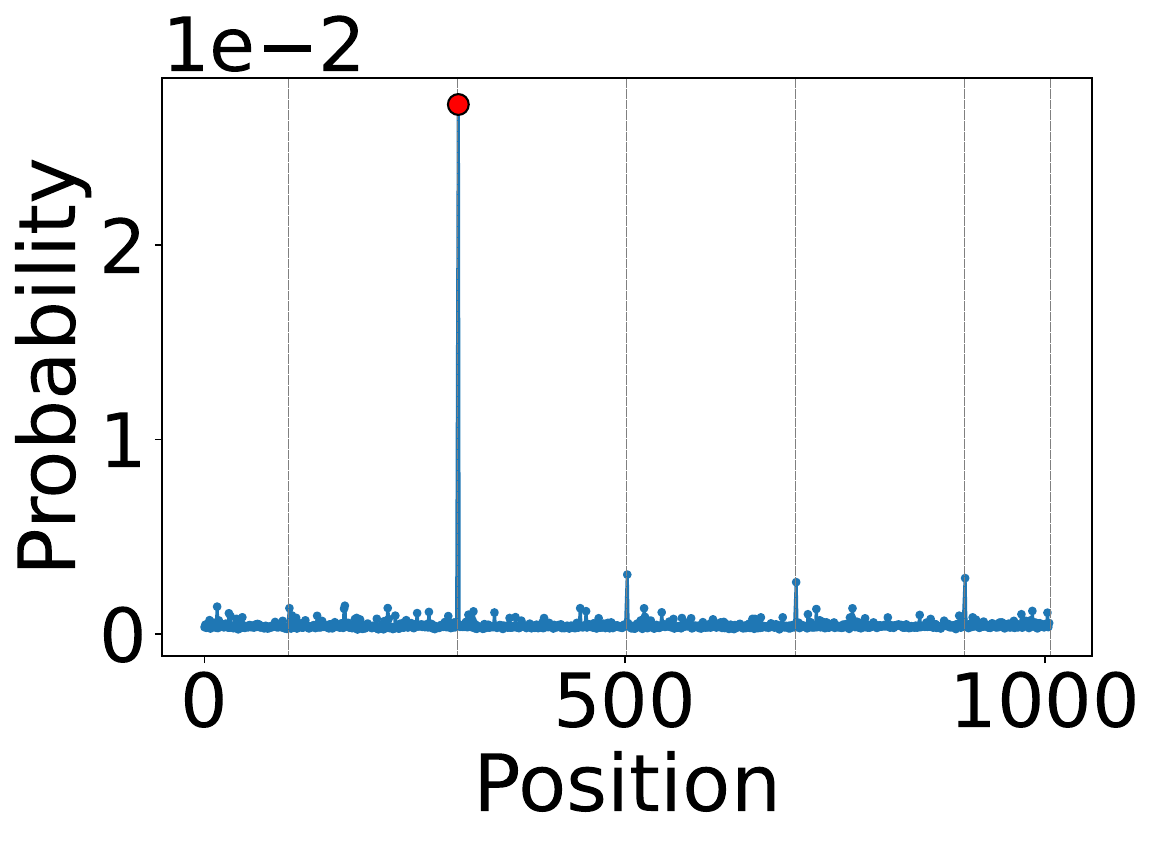} &
\FigGridCell{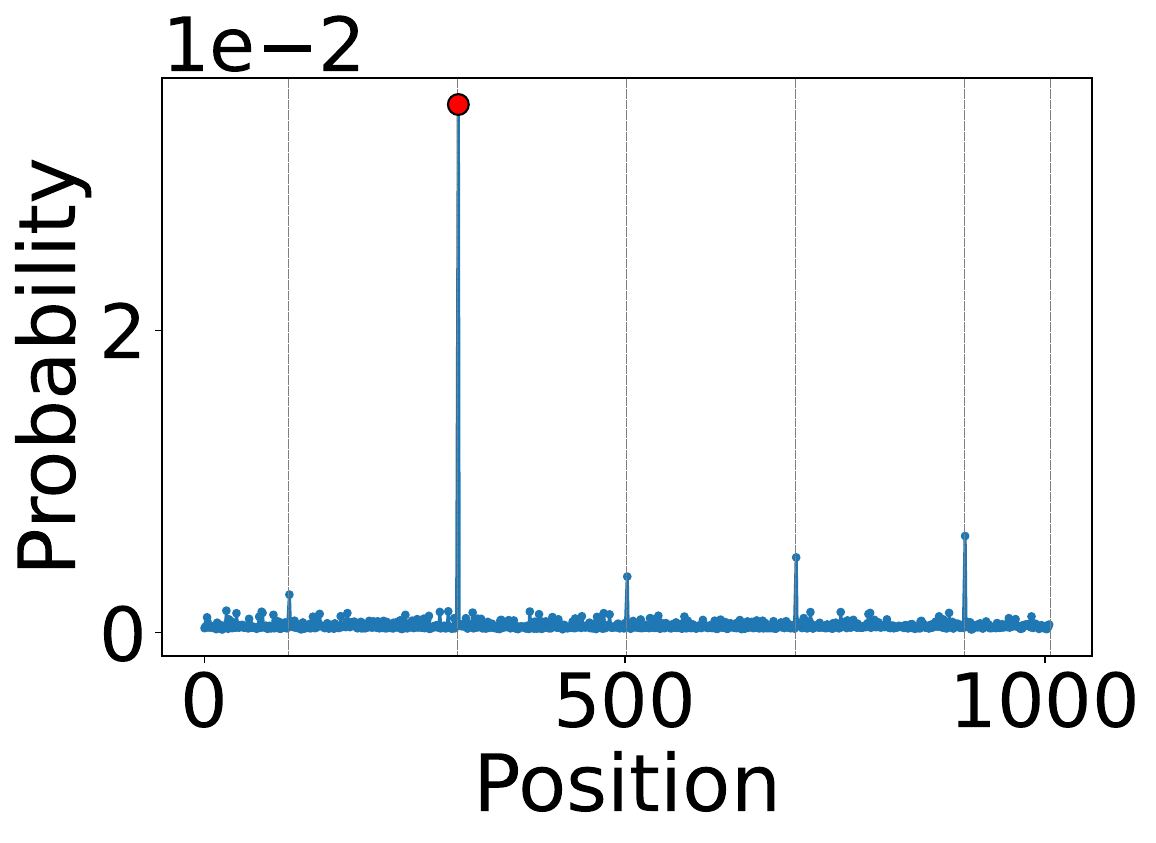} &
\FigGridCell{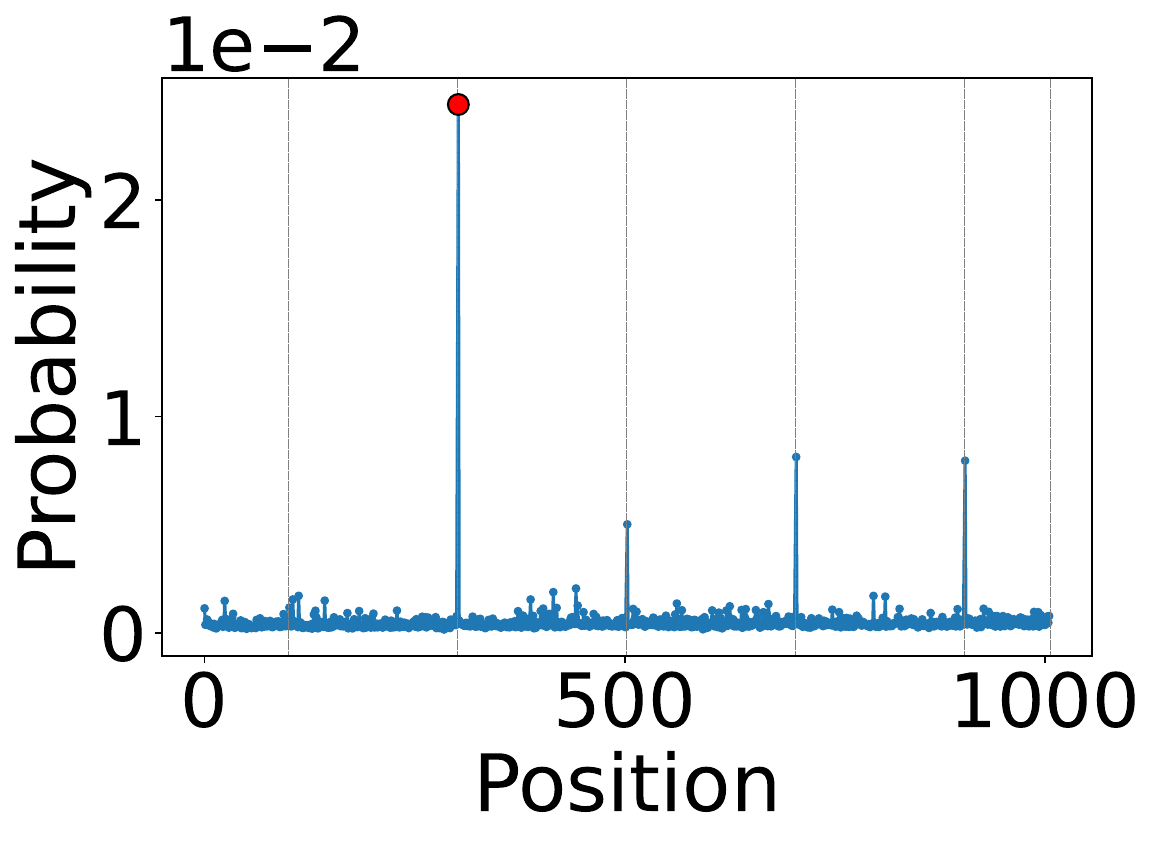} \\

& \FigGridVSubLabel{P3} &
\FigGridCell{Figures/ep_prob_without_A_red/Qwen2.5-7B-Instruct_5_Repeats_200_Length_500_Permutations_0_ablations_induction_3_nth.pdf} &
\FigGridCell{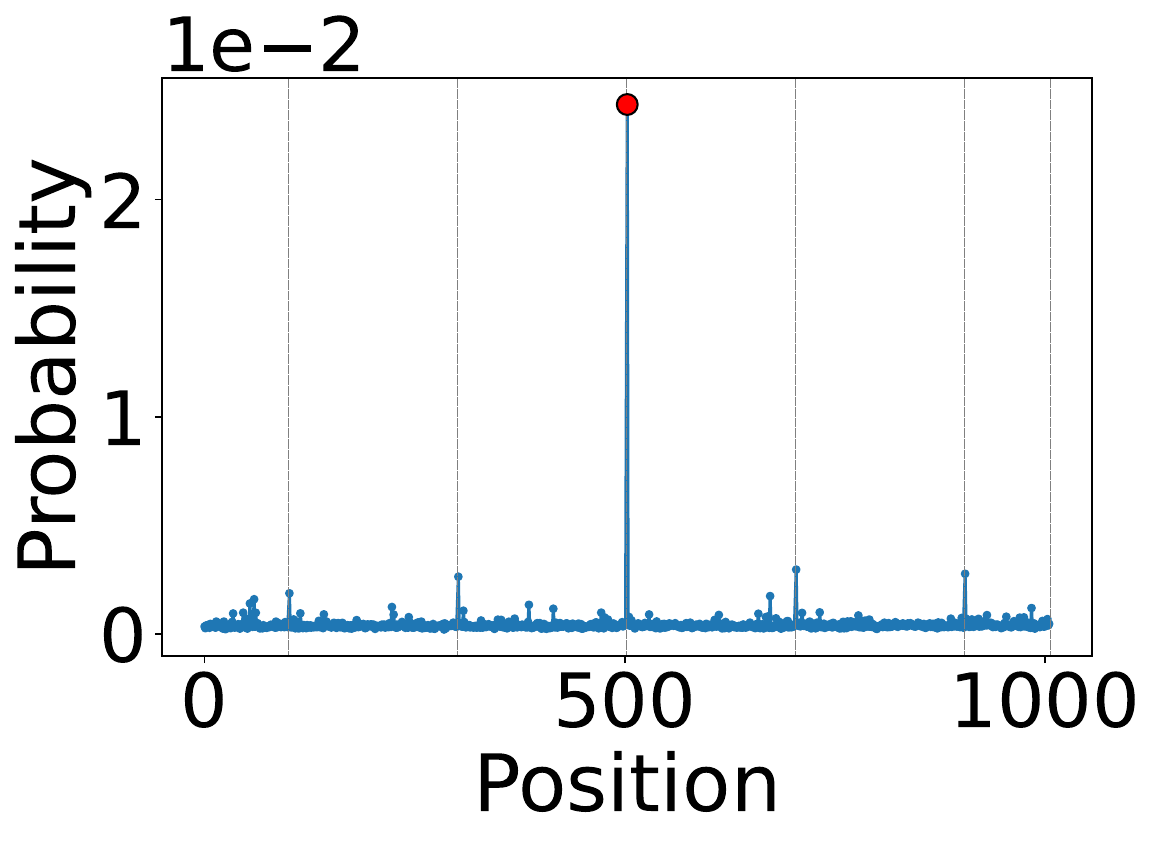} &
\FigGridCell{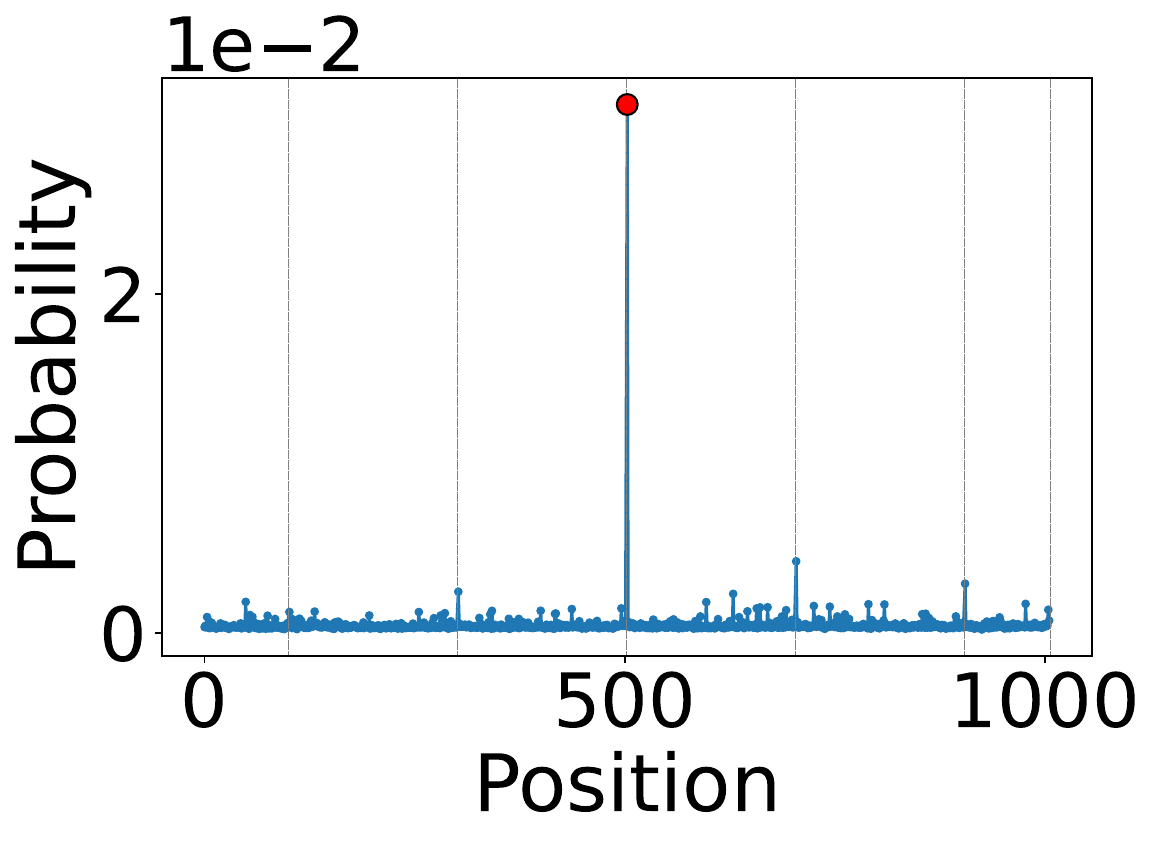} &
\FigGridCell{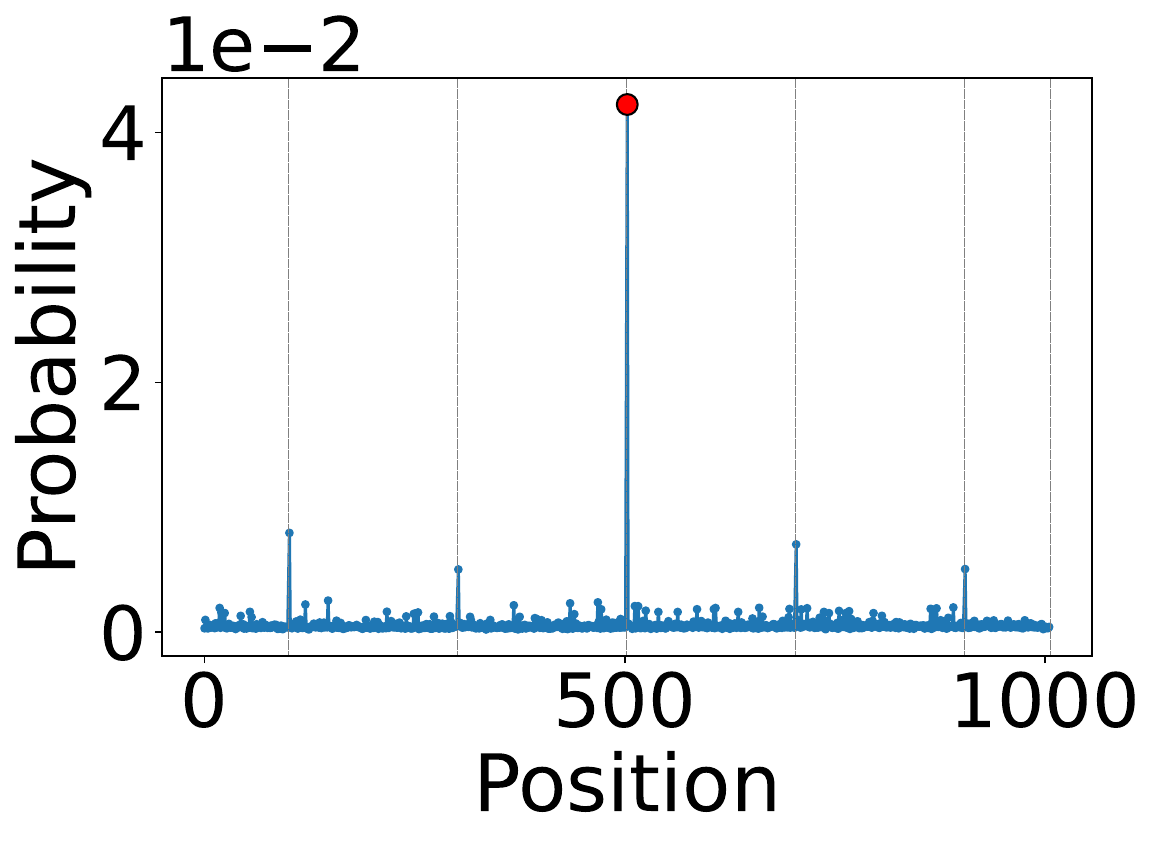} &
\FigGridCell{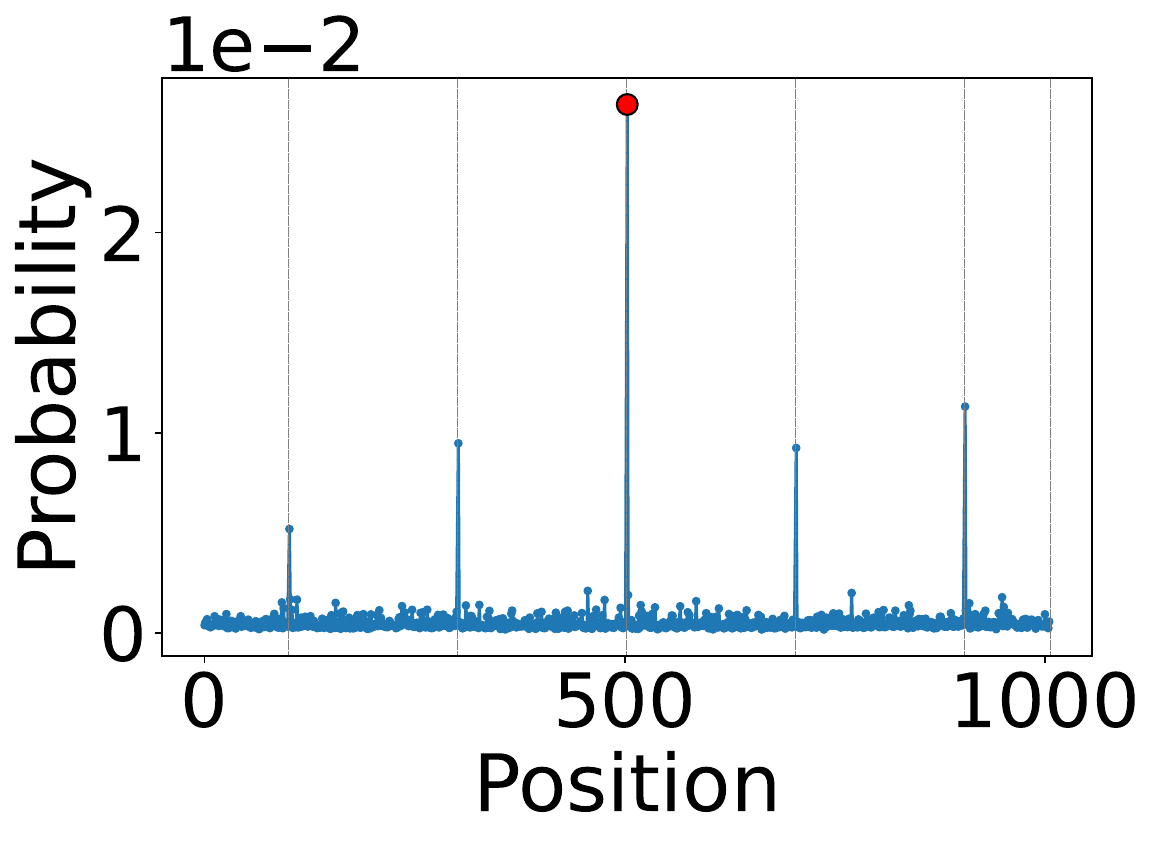} \\

& \FigGridVSubLabel{P4} &
\FigGridCell{Figures/ep_prob_without_A_red/Qwen2.5-7B-Instruct_5_Repeats_200_Length_500_Permutations_0_ablations_induction_4_nth.pdf} &
\FigGridCell{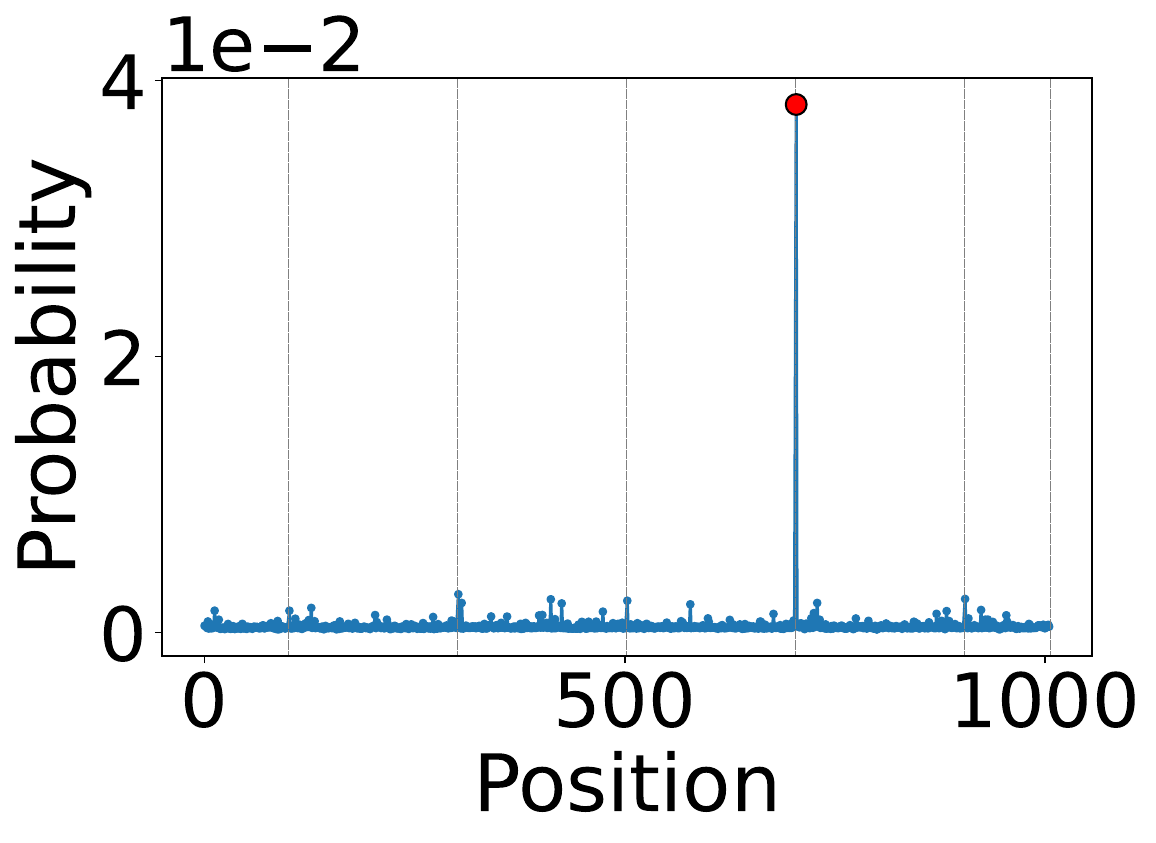} &
\FigGridCell{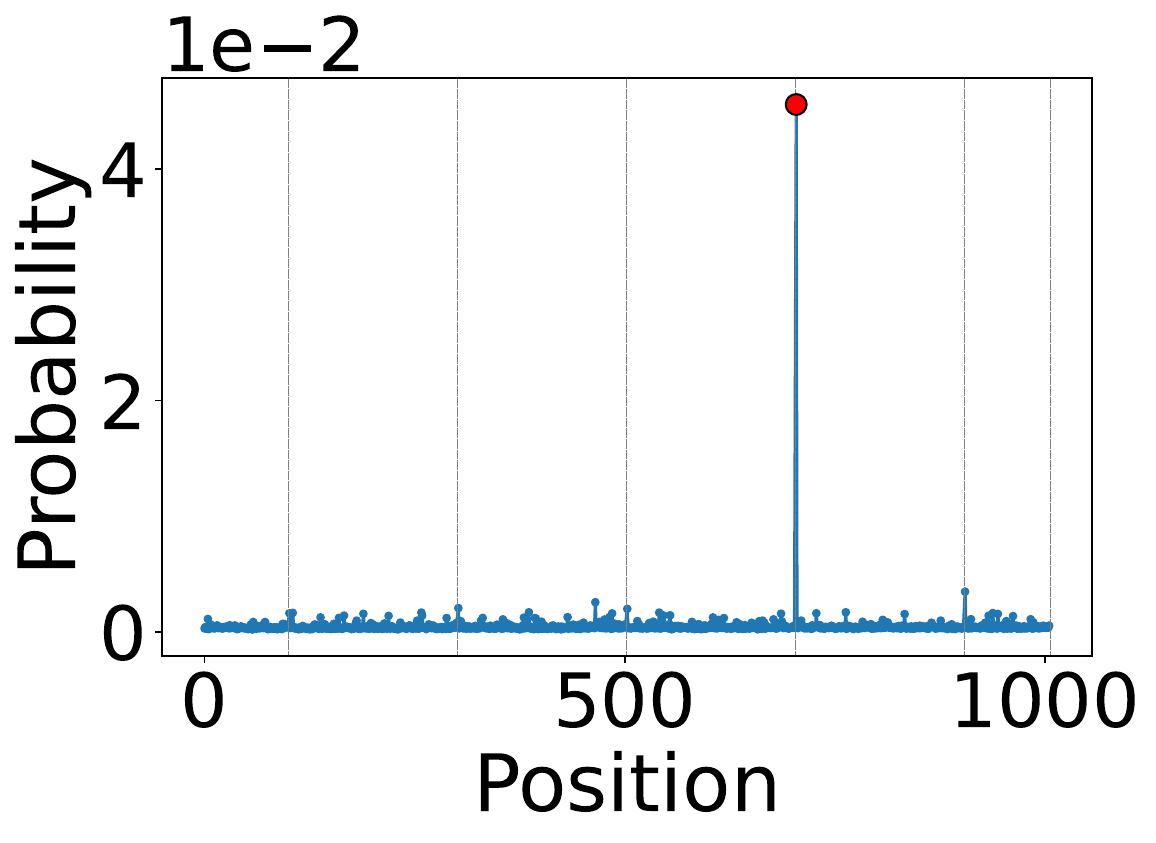} &
\FigGridCell{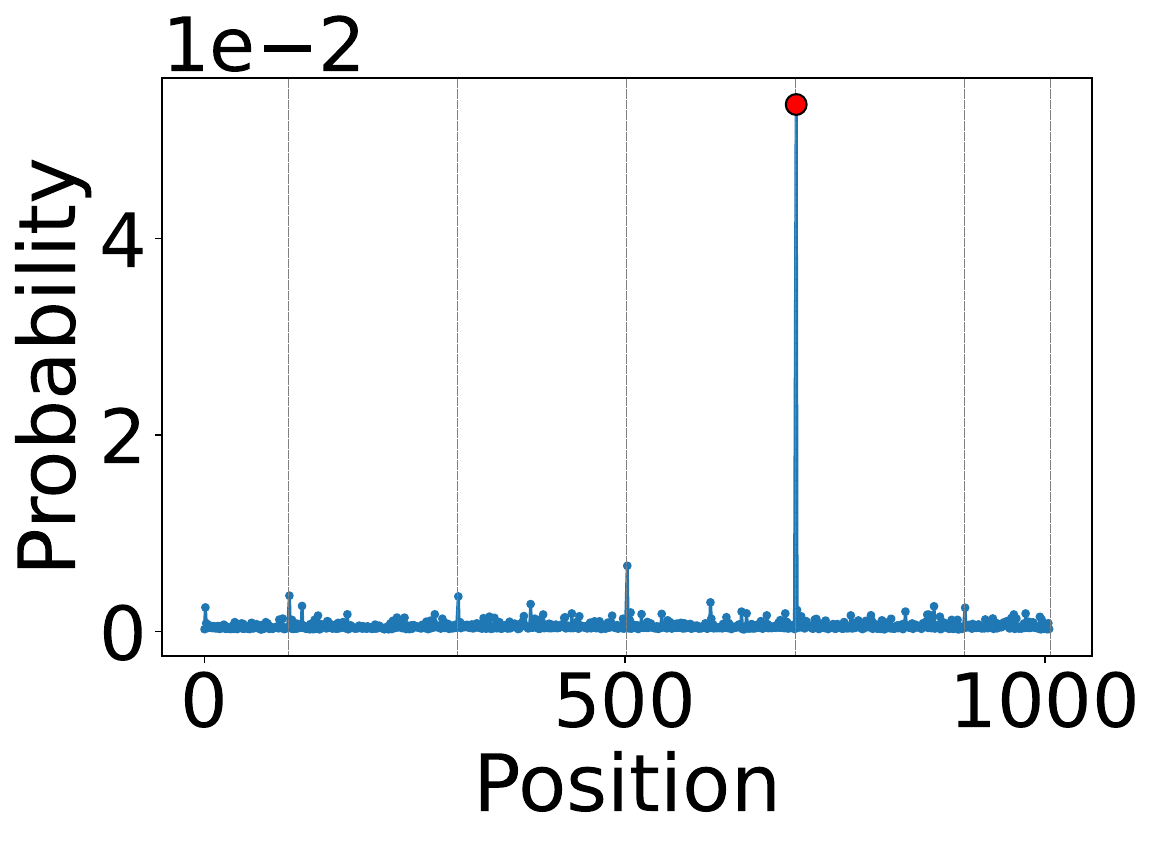} &
\FigGridCell{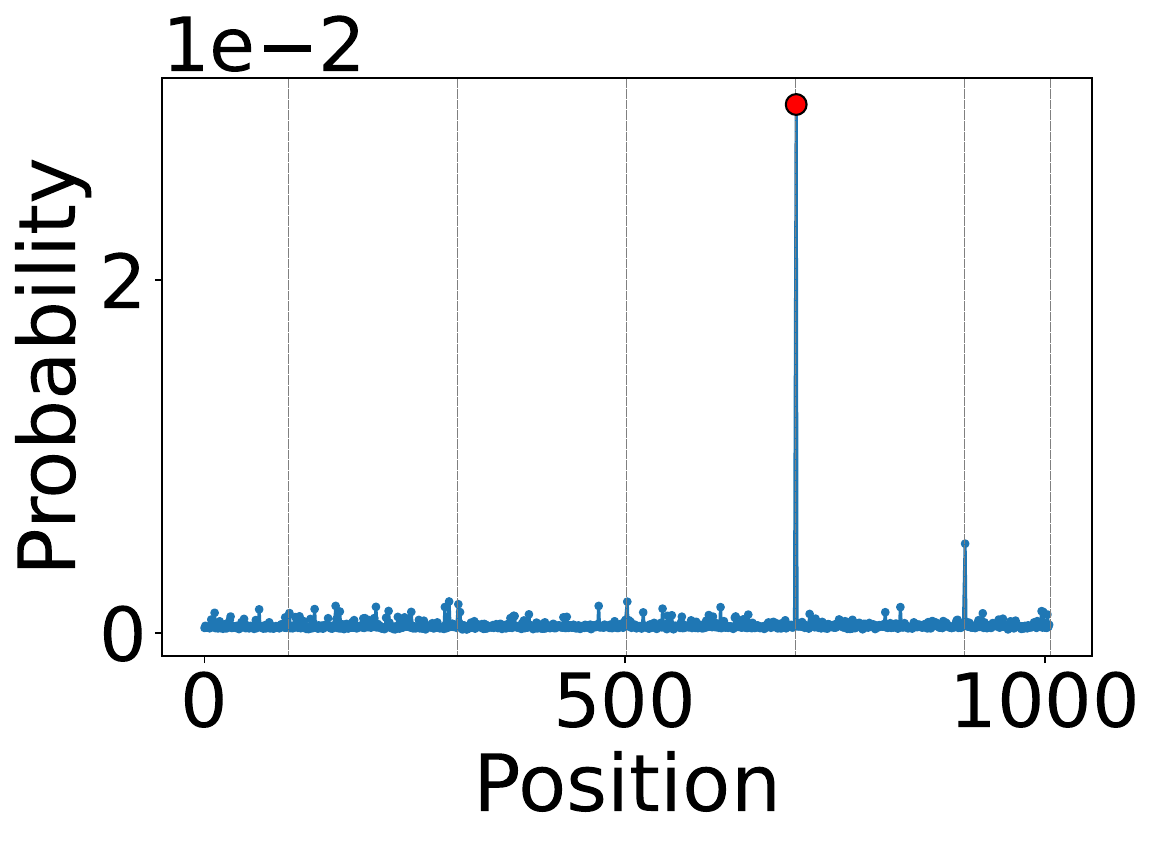} \\

& \FigGridVSubLabel{P5} &
\FigGridCell{Figures/ep_prob_without_A_red/Qwen2.5-7B-Instruct_5_Repeats_200_Length_500_Permutations_0_ablations_induction_5_nth.pdf} &
\FigGridCell{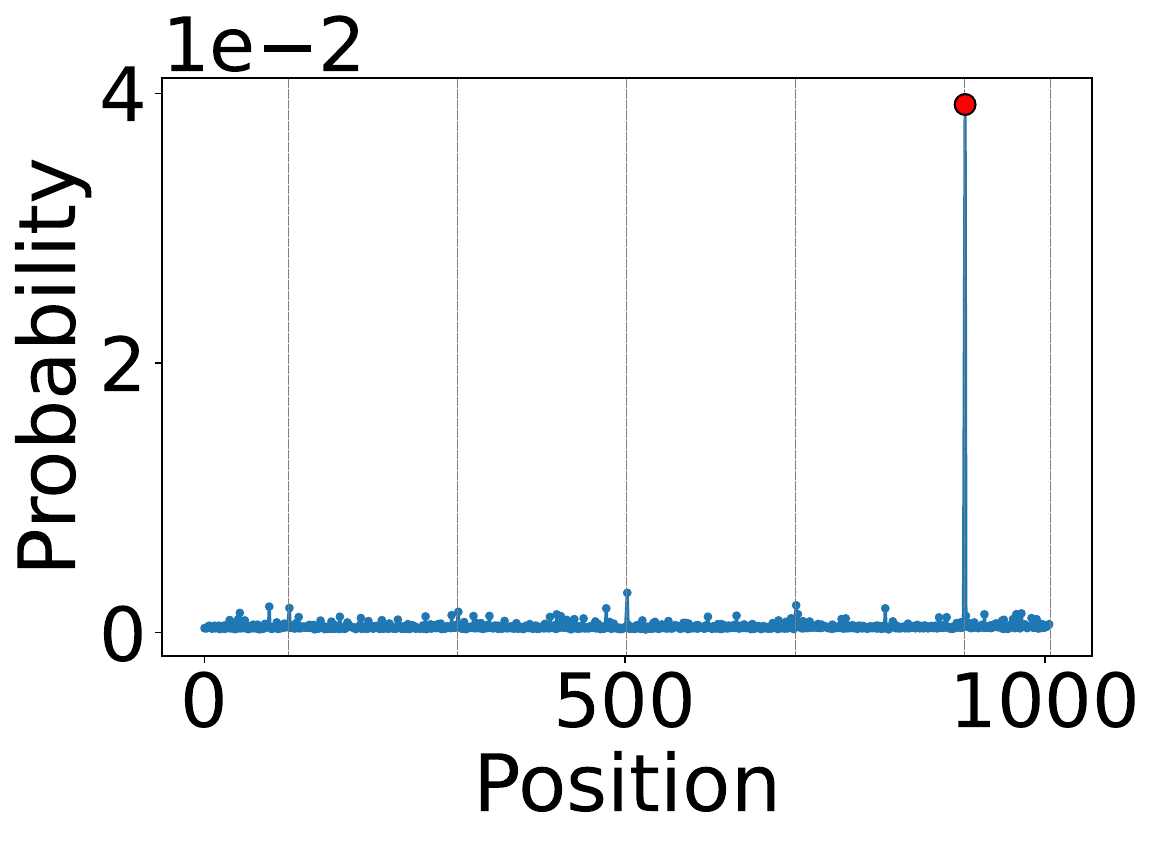} &
\FigGridCell{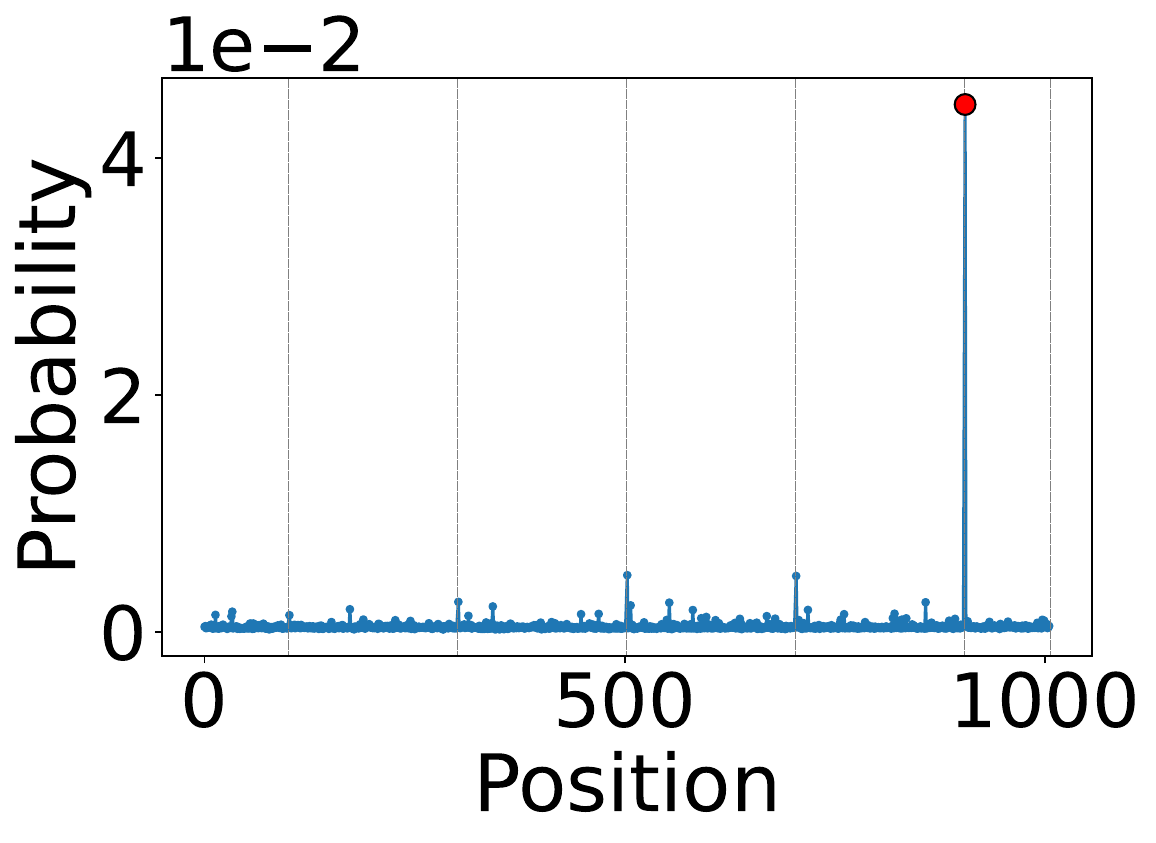} &
\FigGridCell{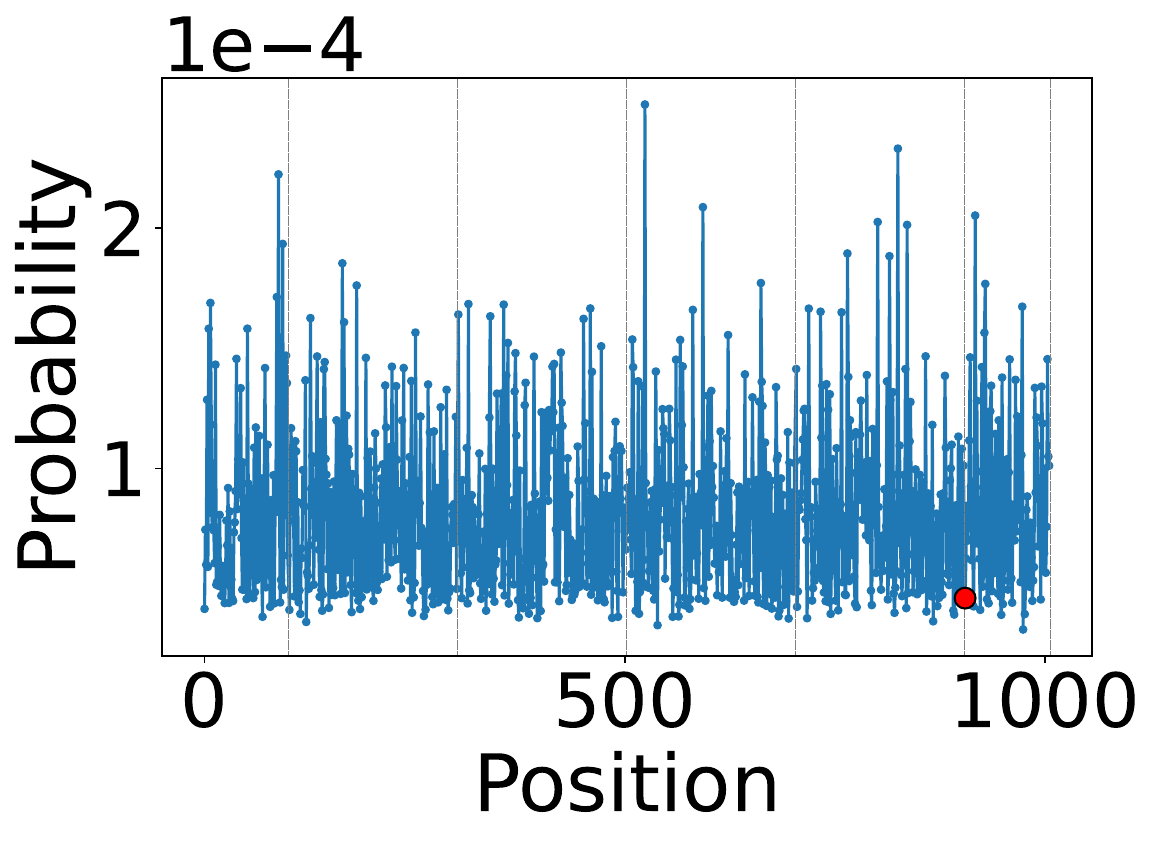} &
\FigGridCell{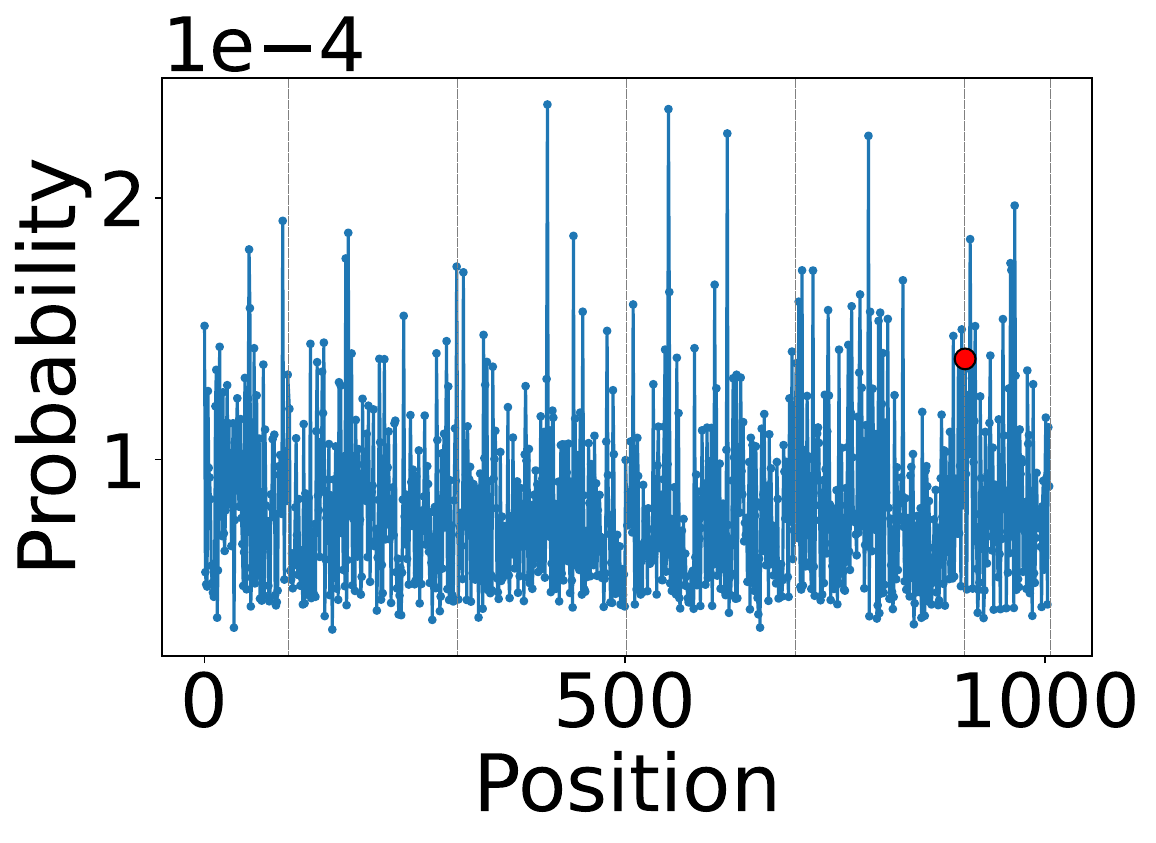} \\

\bottomrule
\end{tabular}
\endgroup
\caption{Qwen ablation effect (Exp. 2). Episodic retrieval probability after ablating Induction (Ind P1-P5) or Random (Rand P1-P5) heads. Columns show number of ablated heads.}
\label{fig:exp2_qwen_ablation}
\end{figure*}
\begin{figure*}[h!]
\centering
\begingroup
\setlength{\FigGridLabelW}{1.05cm}
\setlength{\FigGridSubLabelW}{0.55cm}
\setlength{\FigGridCellH}{1.85cm} 
\setlength{\FigGridLabelSep}{1pt}
\setlength{\FigGridSubLabelSep}{1pt}
\setlength{\FigGridColSep}{0.5pt}

\FigGridSetupTwo{5}
\renewcommand{\arraystretch}{1.0}

\begin{tabular}{@{}>{\centering\arraybackslash}m{\FigGridLabelW}@{\hspace{\FigGridLabelSep}}
                >{\centering\arraybackslash}m{\FigGridSubLabelW}@{\hspace{\FigGridSubLabelSep}}
                *{5}{>{\centering\arraybackslash}m{\FigGridPlotW}}@{}}
\toprule
& & \multicolumn{5}{c}{\FigHdr Ablations} \\
\cmidrule(lr){3-7}
& & \FigHdr 0 & \FigHdr 1 & \FigHdr 10 & \FigHdr 50 & \FigHdr 100 \\
\midrule
\multirow{5}{*}{\FigGridVLabelSpanFive{Induction}} &
\FigGridVSubLabel{P1} &
\FigGridCell{Figures/ep_prob_without_A_red/gemma-2-9b-it_5_Repeats_200_Length_500_Permutations_0_ablations_induction_1_nth.pdf} &
\FigGridCell{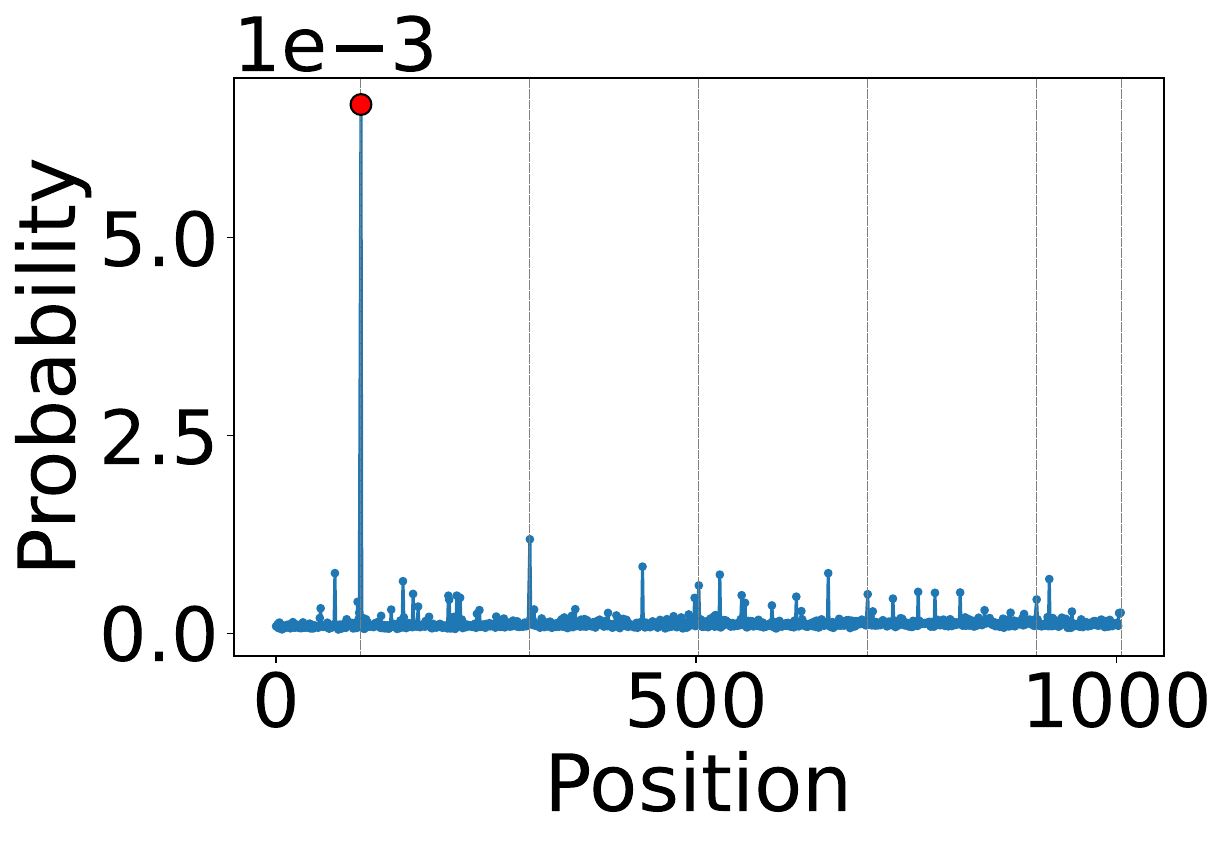} &
\FigGridCell{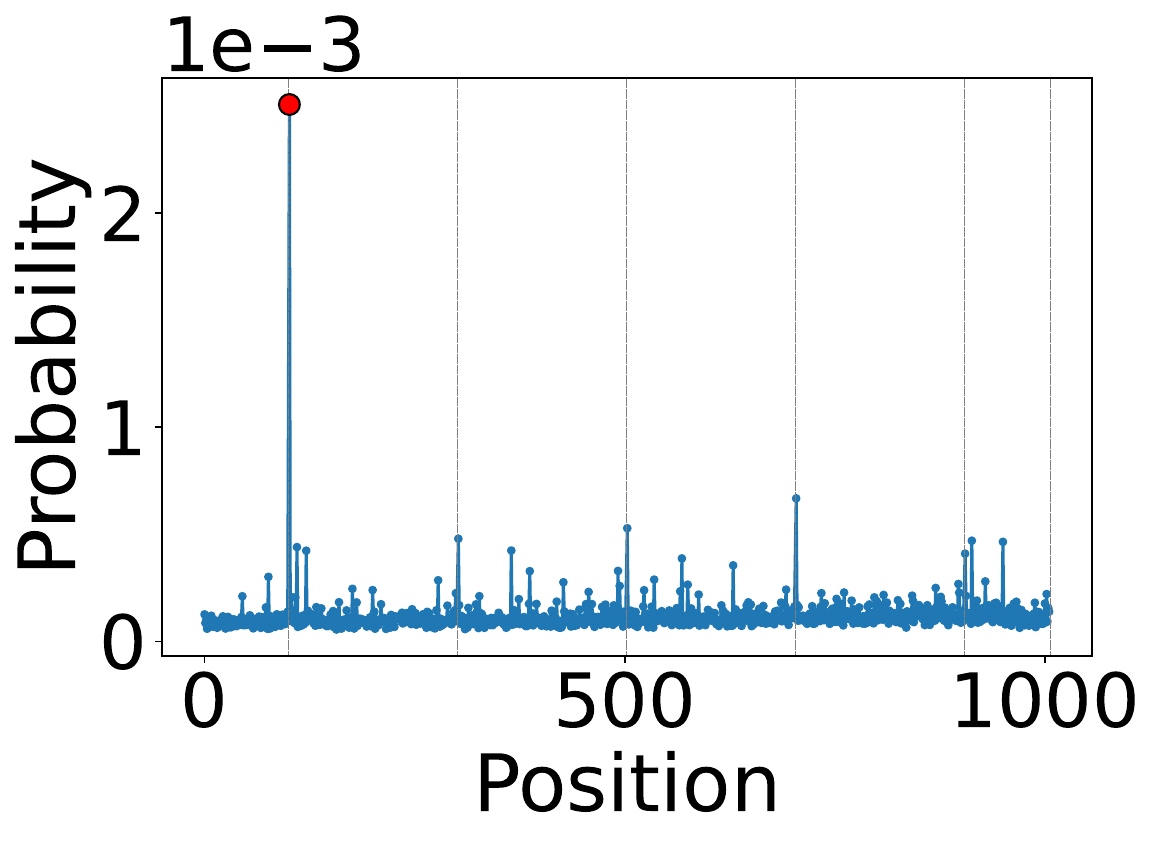} &
\FigGridCell{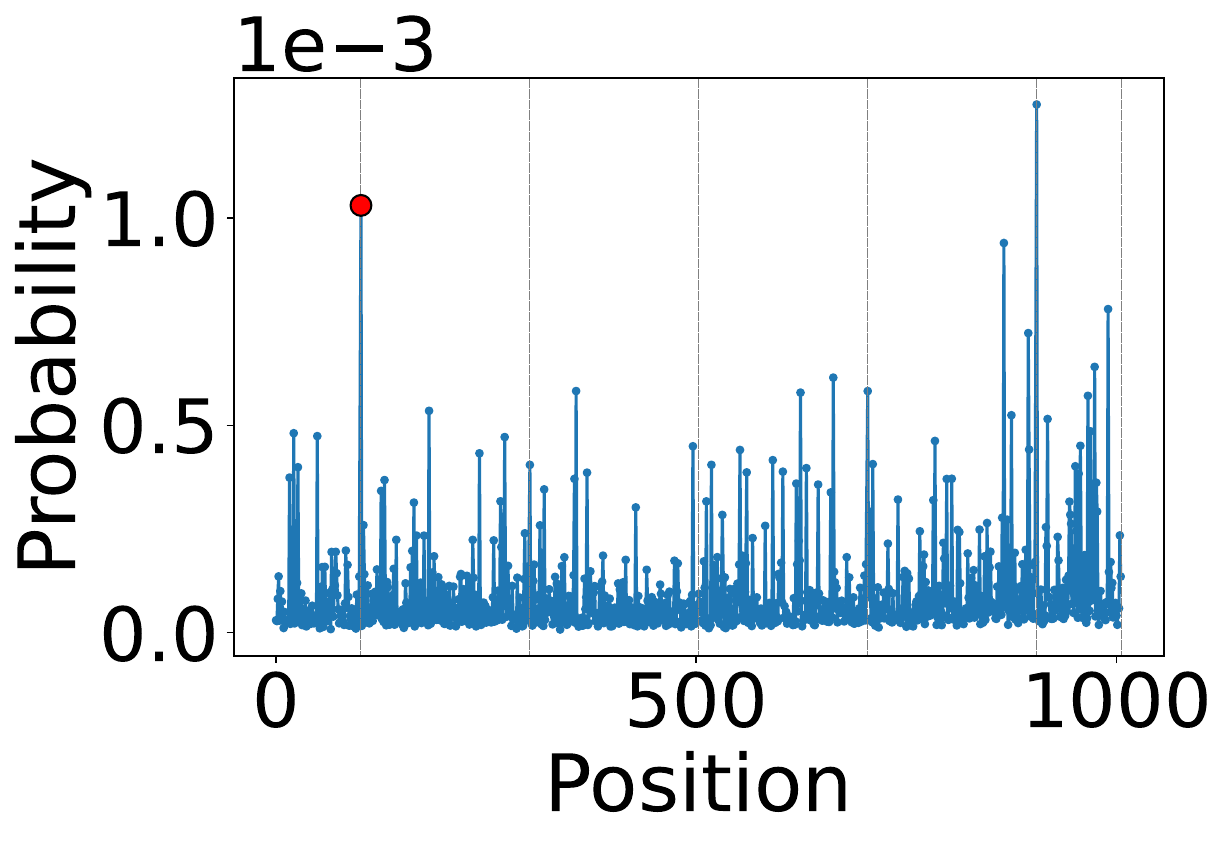} &
\FigGridCell{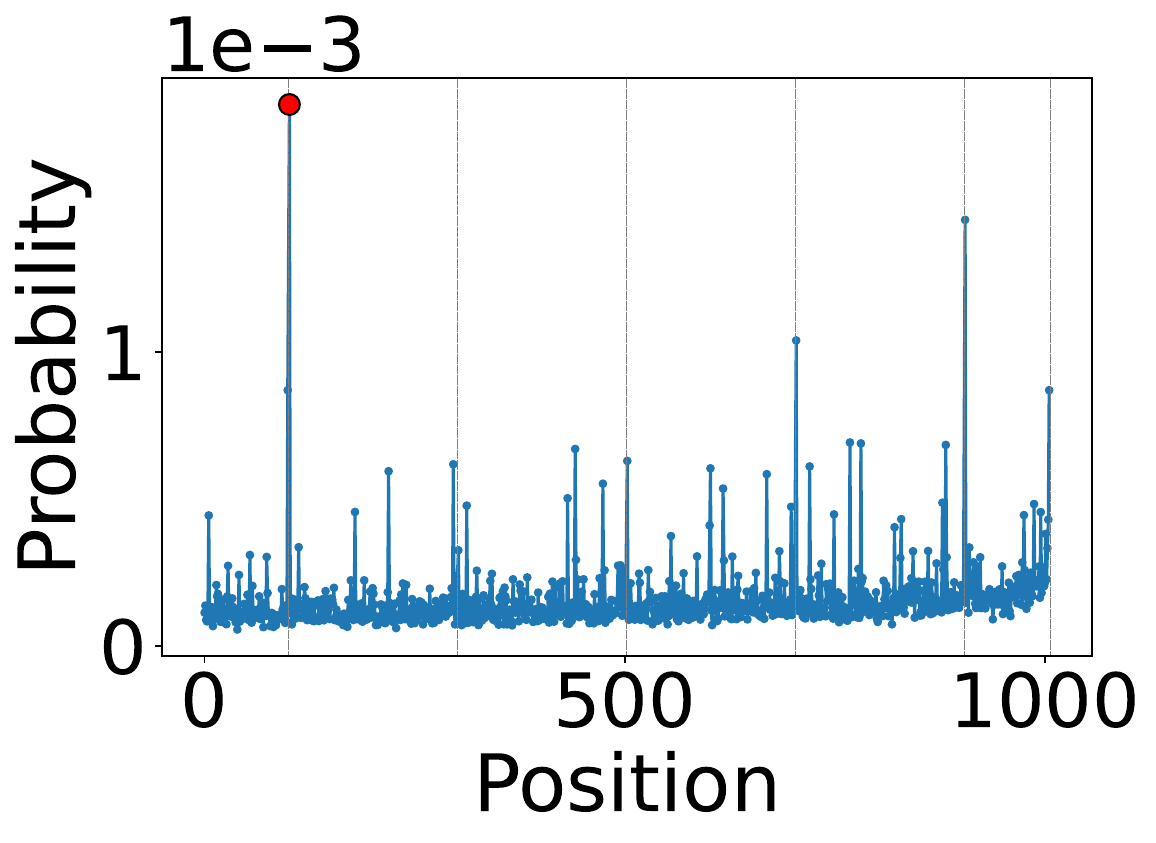} \\
& \FigGridVSubLabel{P2} &
\FigGridCell{Figures/ep_prob_without_A_red/gemma-2-9b-it_5_Repeats_200_Length_500_Permutations_0_ablations_induction_2_nth.pdf} &
\FigGridCell{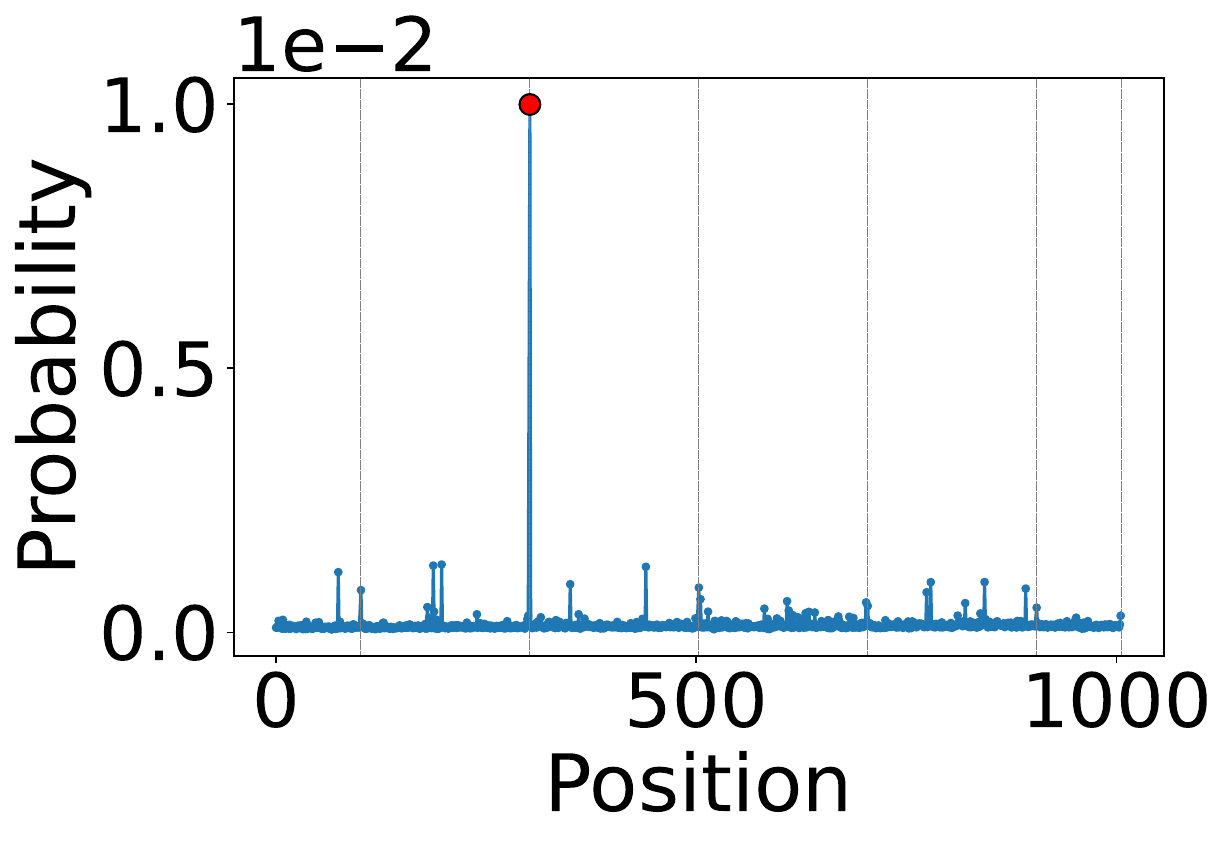} &
\FigGridCell{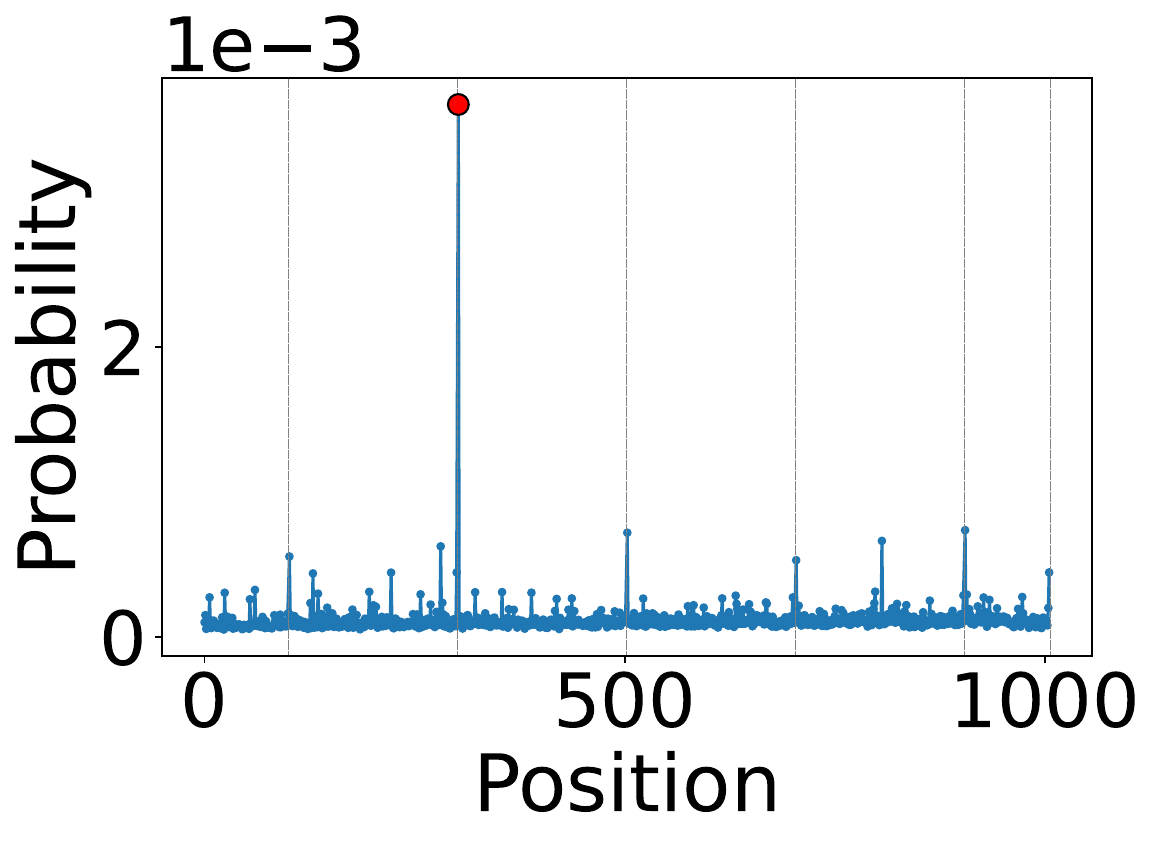} &
\FigGridCell{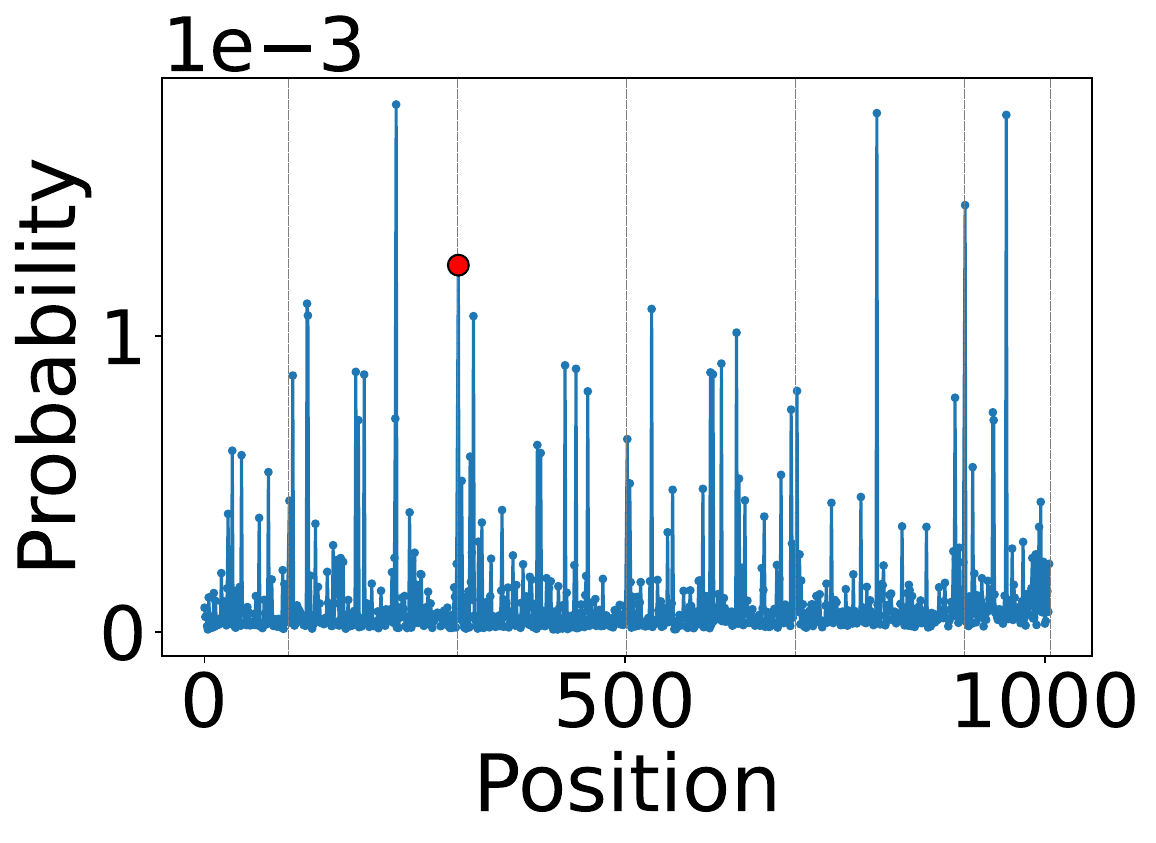} &
\FigGridCell{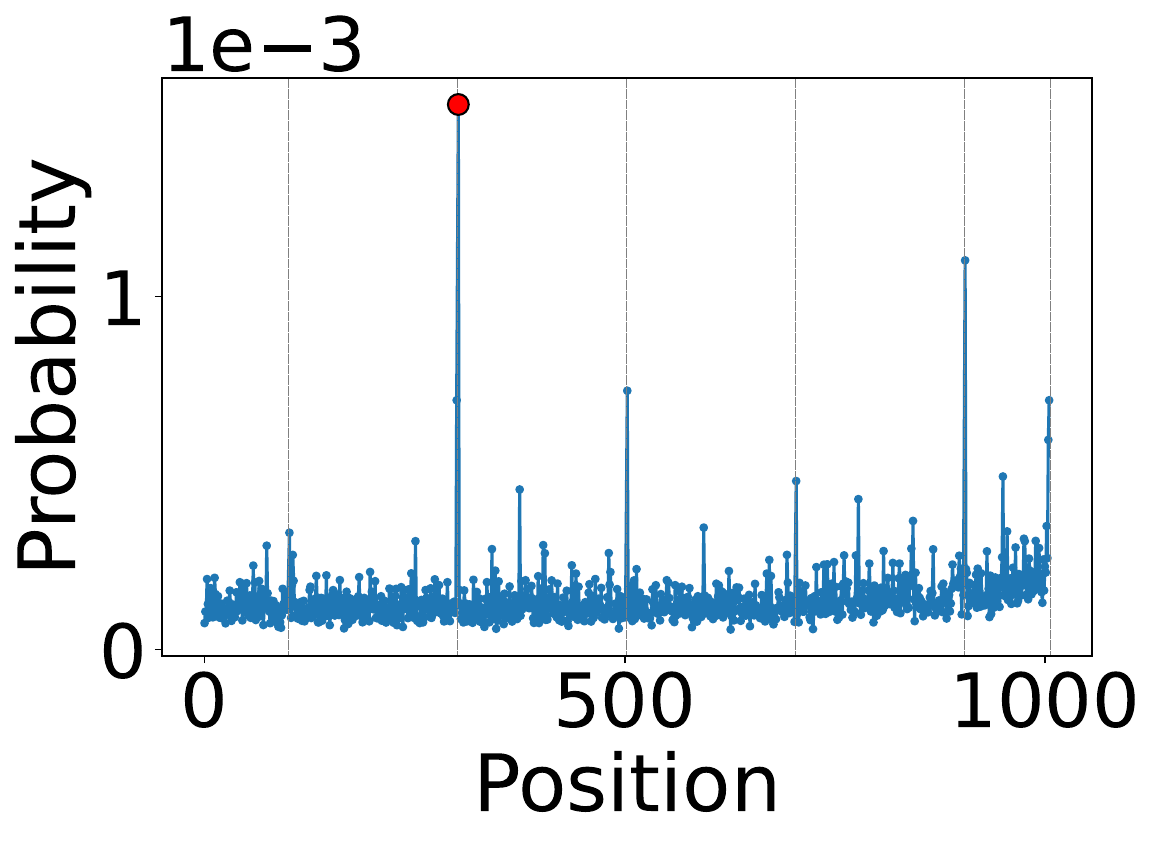} \\
& \FigGridVSubLabel{P3} &
\FigGridCell{Figures/ep_prob_without_A_red/gemma-2-9b-it_5_Repeats_200_Length_500_Permutations_0_ablations_induction_3_nth.pdf} &
\FigGridCell{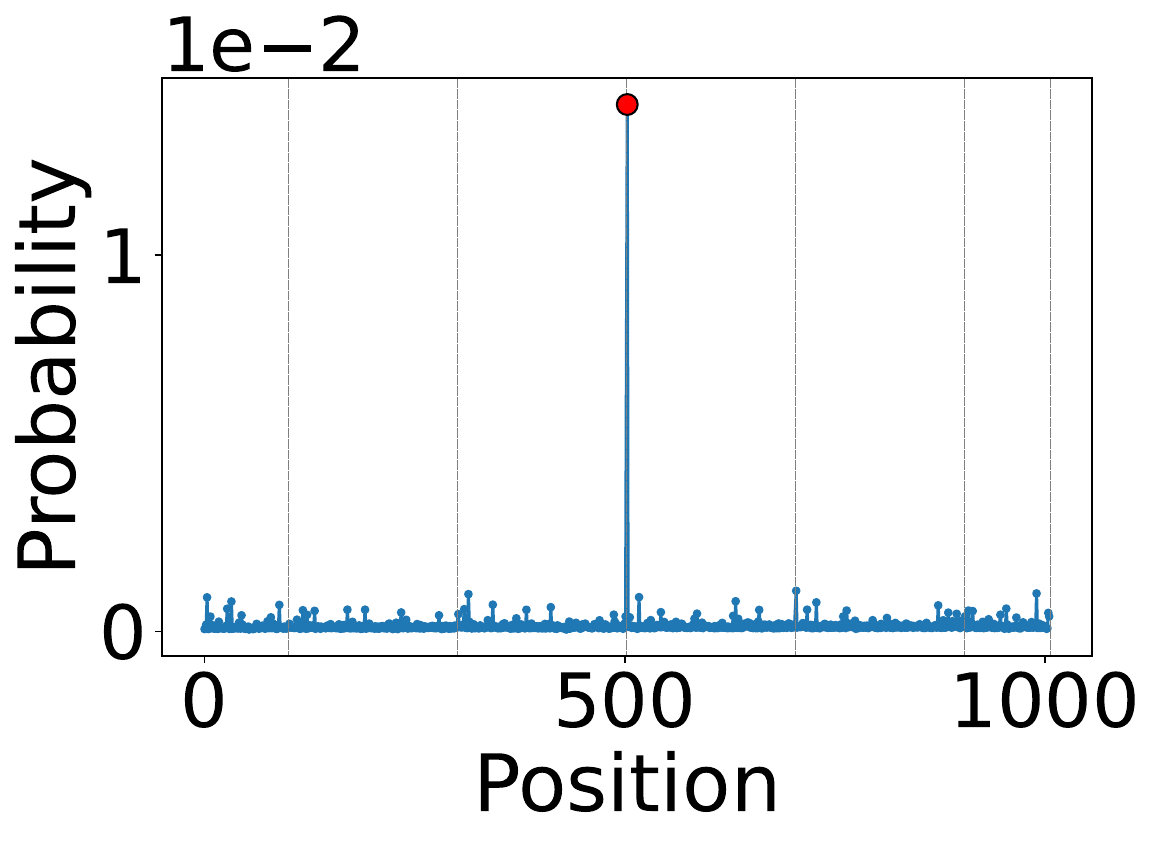} &
\FigGridCell{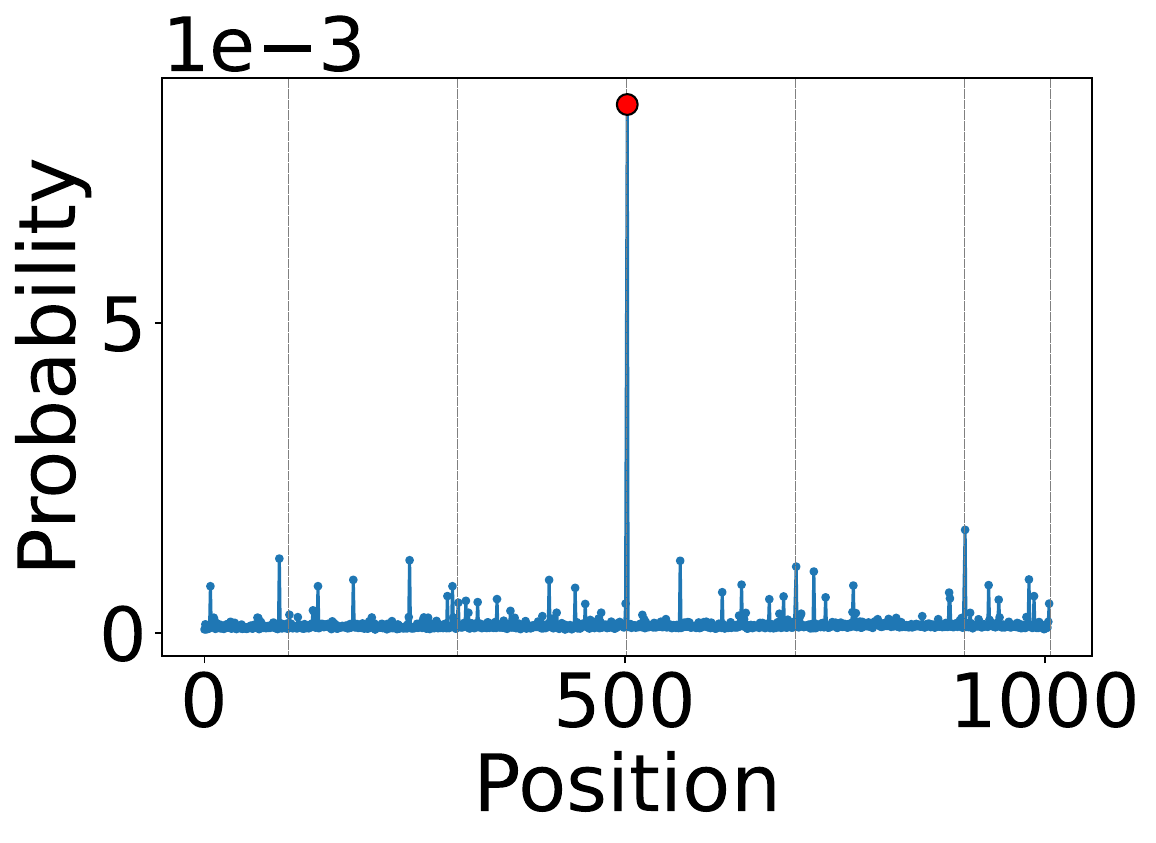} &
\FigGridCell{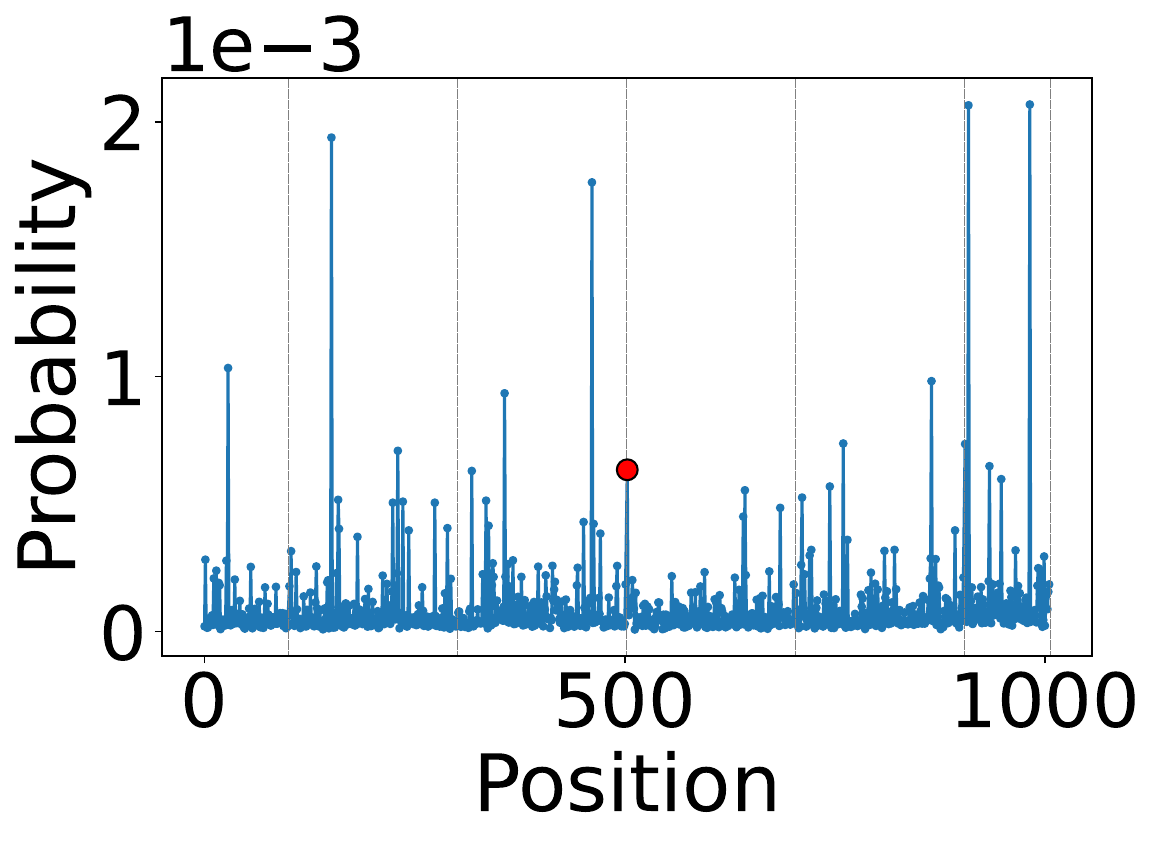} &
\FigGridCell{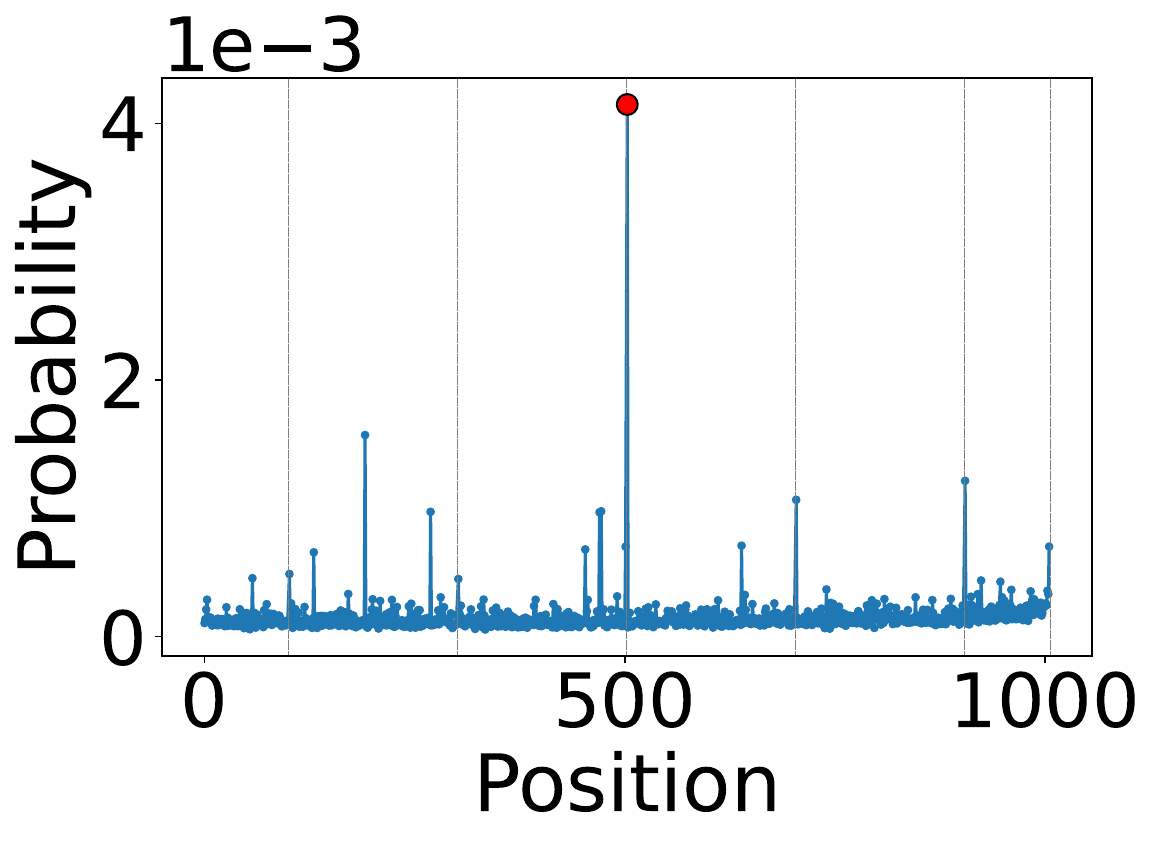} \\
& \FigGridVSubLabel{P4} &
\FigGridCell{Figures/ep_prob_without_A_red/gemma-2-9b-it_5_Repeats_200_Length_500_Permutations_0_ablations_induction_4_nth.pdf} &
\FigGridCell{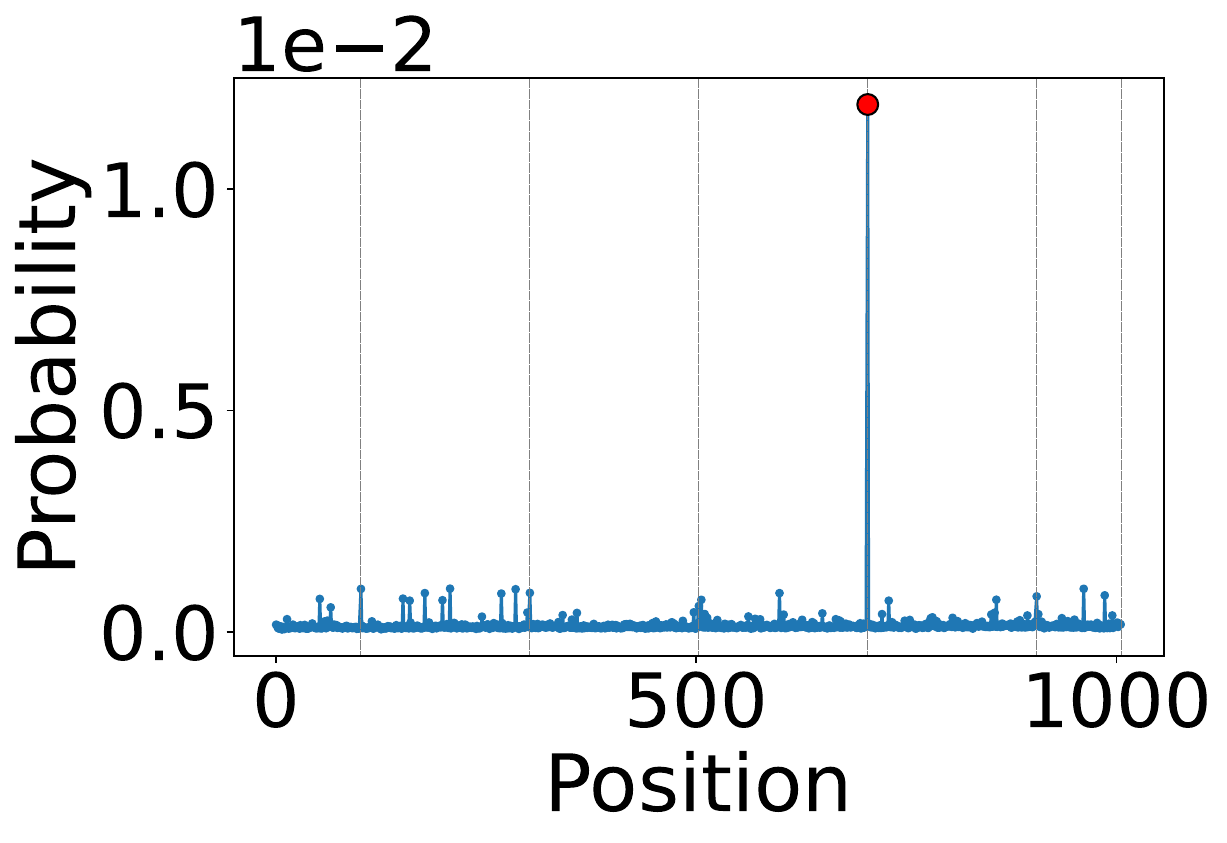} &
\FigGridCell{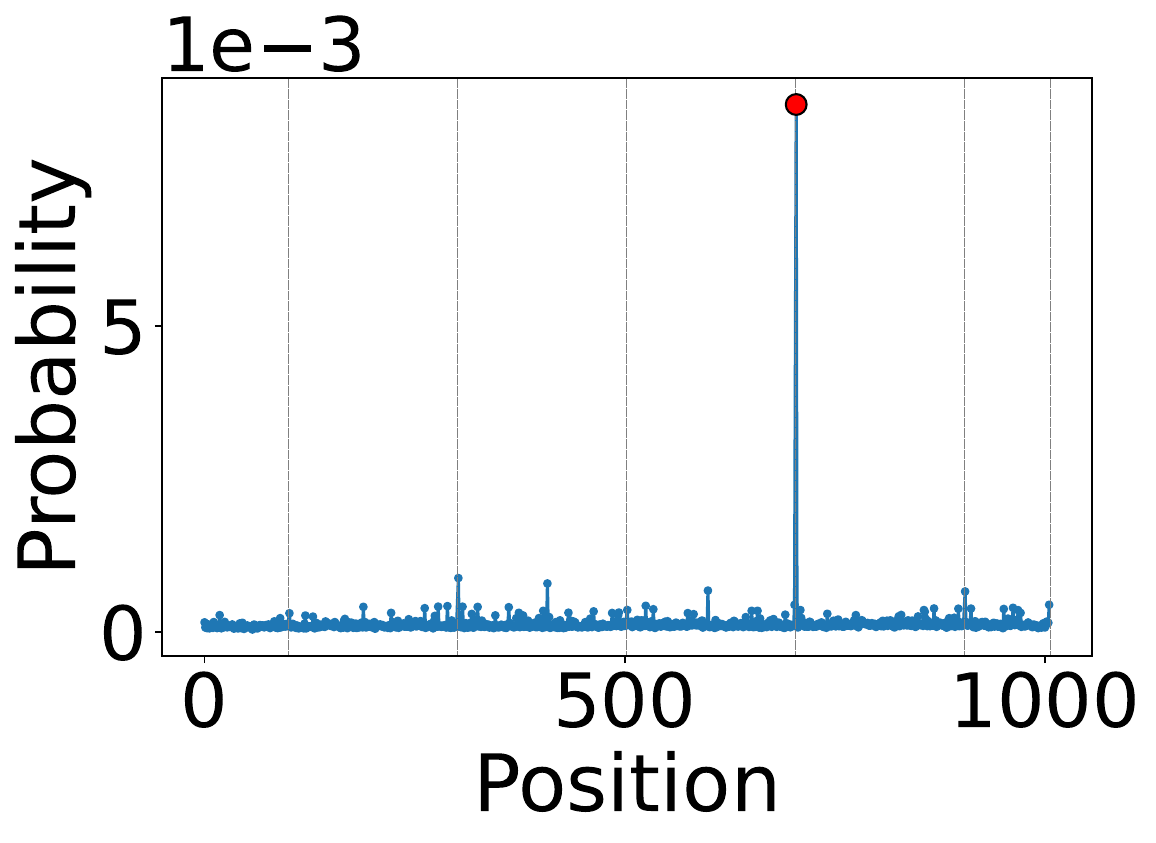} &
\FigGridCell{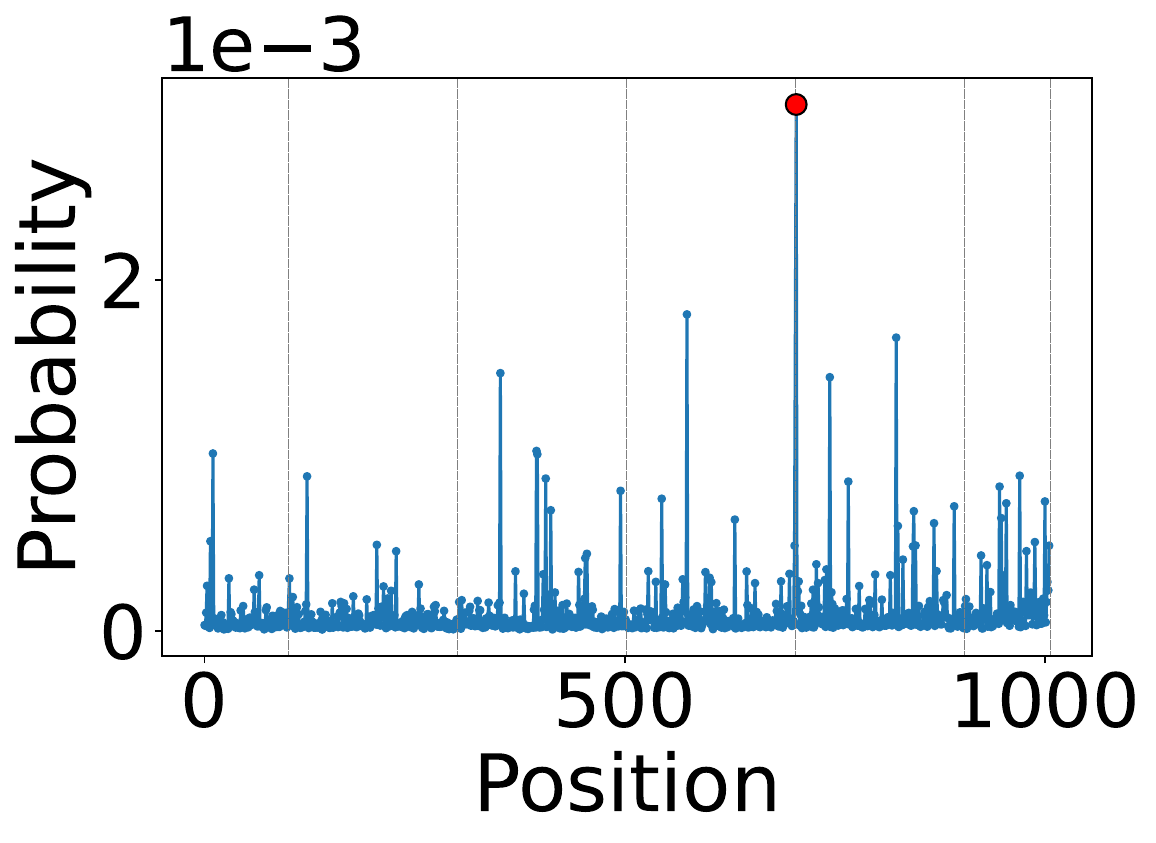} &
\FigGridCell{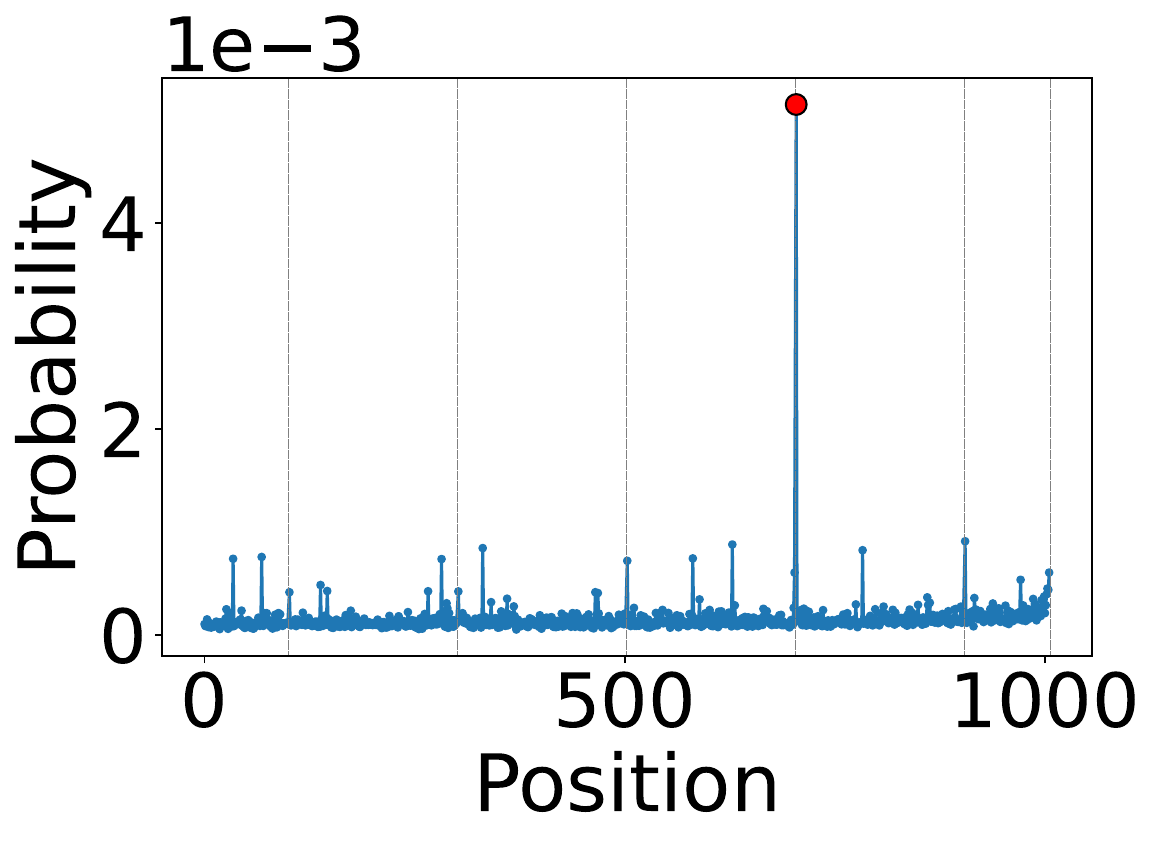} \\
& \FigGridVSubLabel{P5} &
\FigGridCell{Figures/ep_prob_without_A_red/gemma-2-9b-it_5_Repeats_200_Length_500_Permutations_0_ablations_induction_5_nth.pdf} &
\FigGridCell{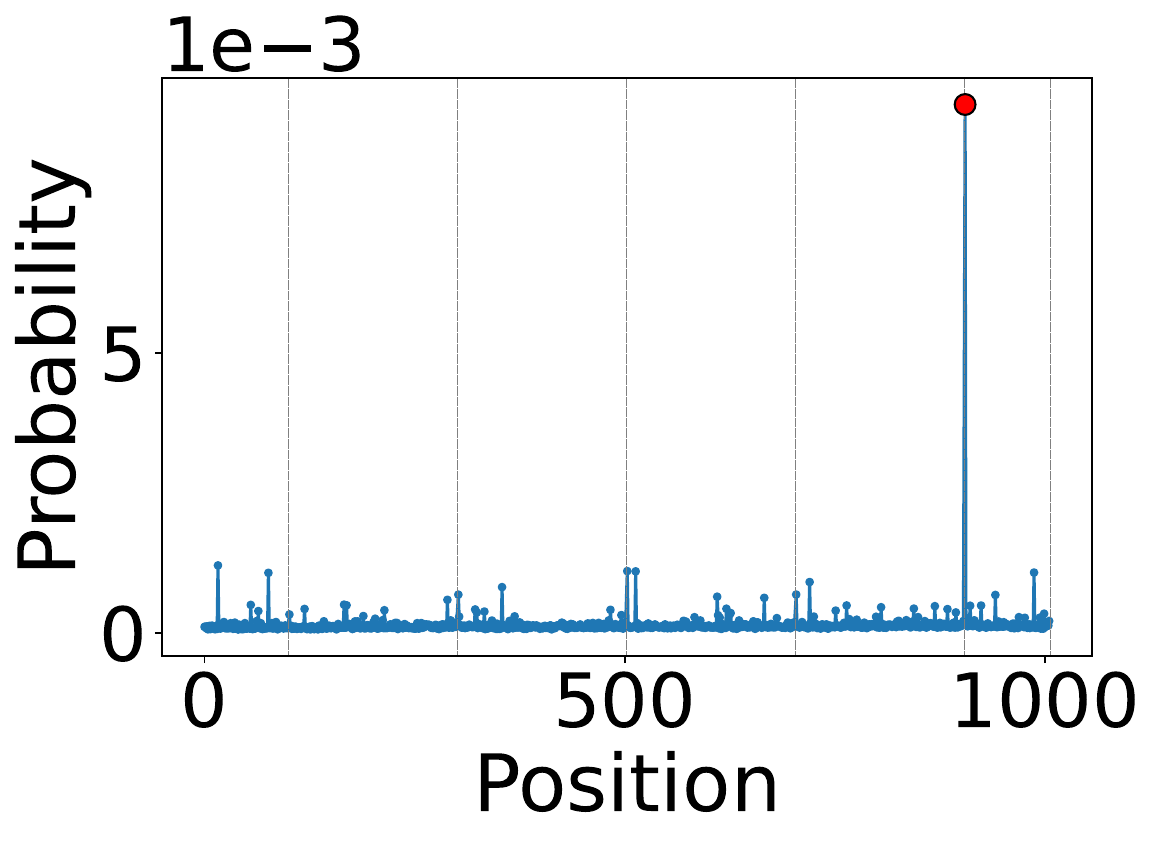} &
\FigGridCell{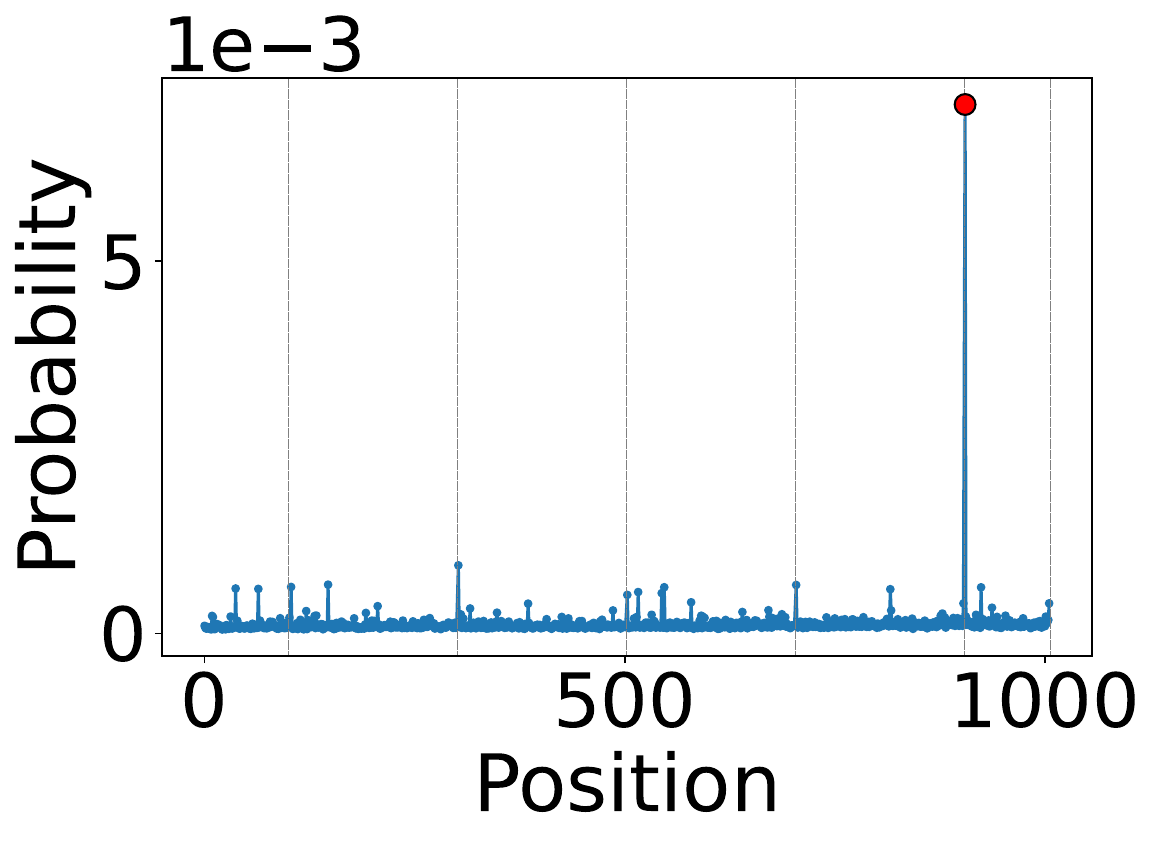} &
\FigGridCell{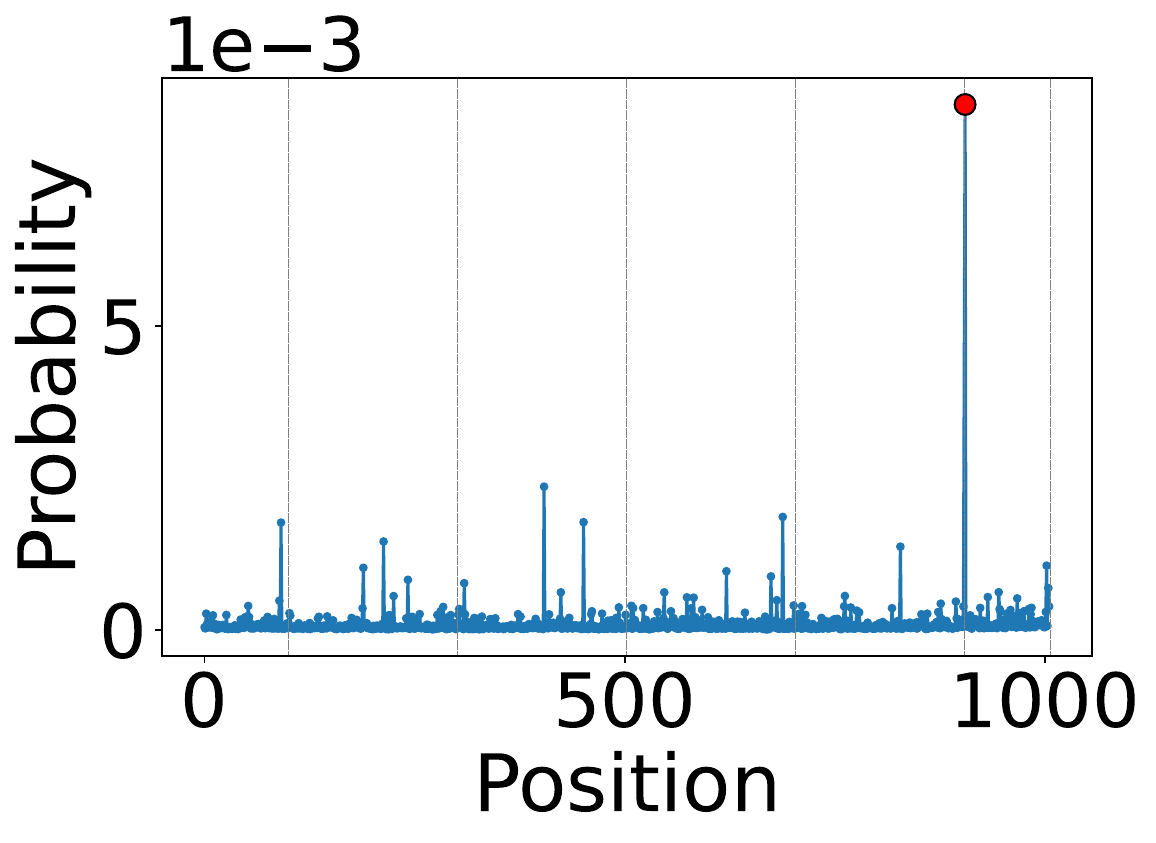} &
\FigGridCell{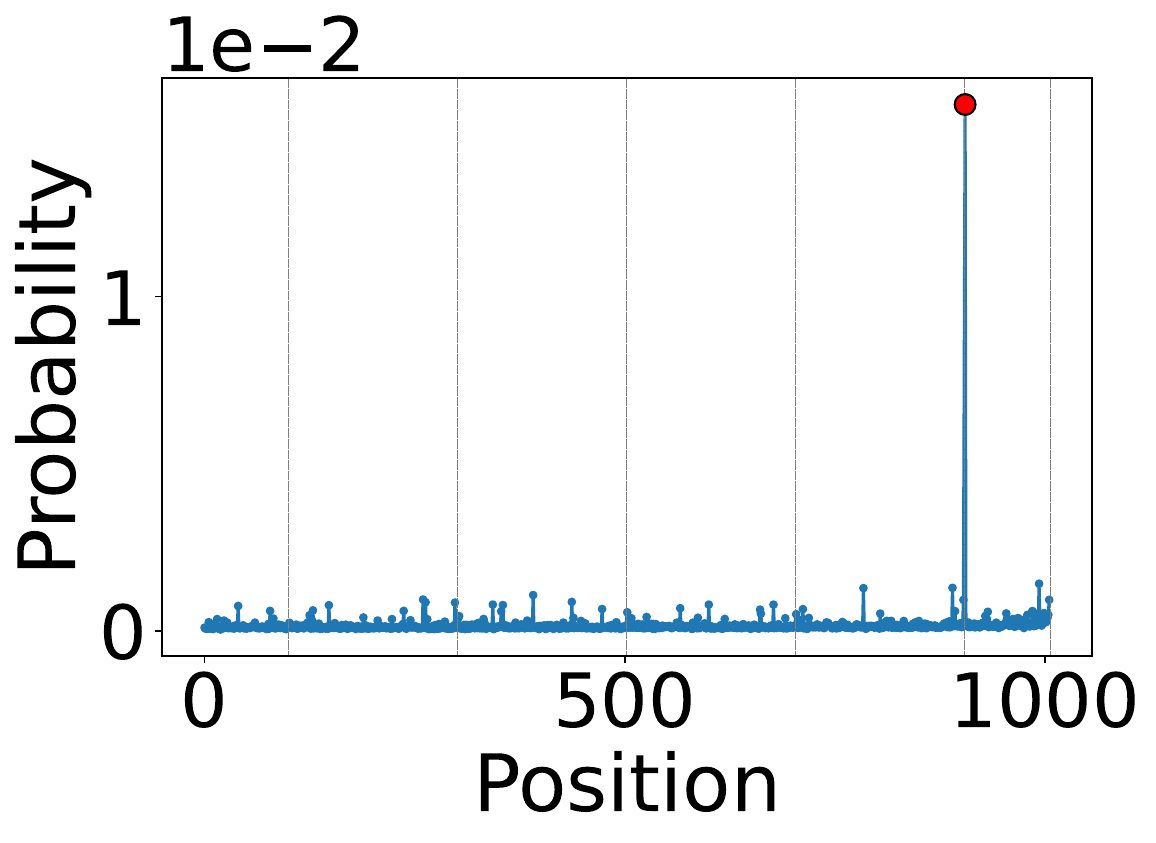} \\
\addlinespace[1pt]
\midrule
\addlinespace[1pt]
\multirow{5}{*}{\FigGridVLabelSpanFive{Random}} &
\FigGridVSubLabel{P1} &
\FigGridCell{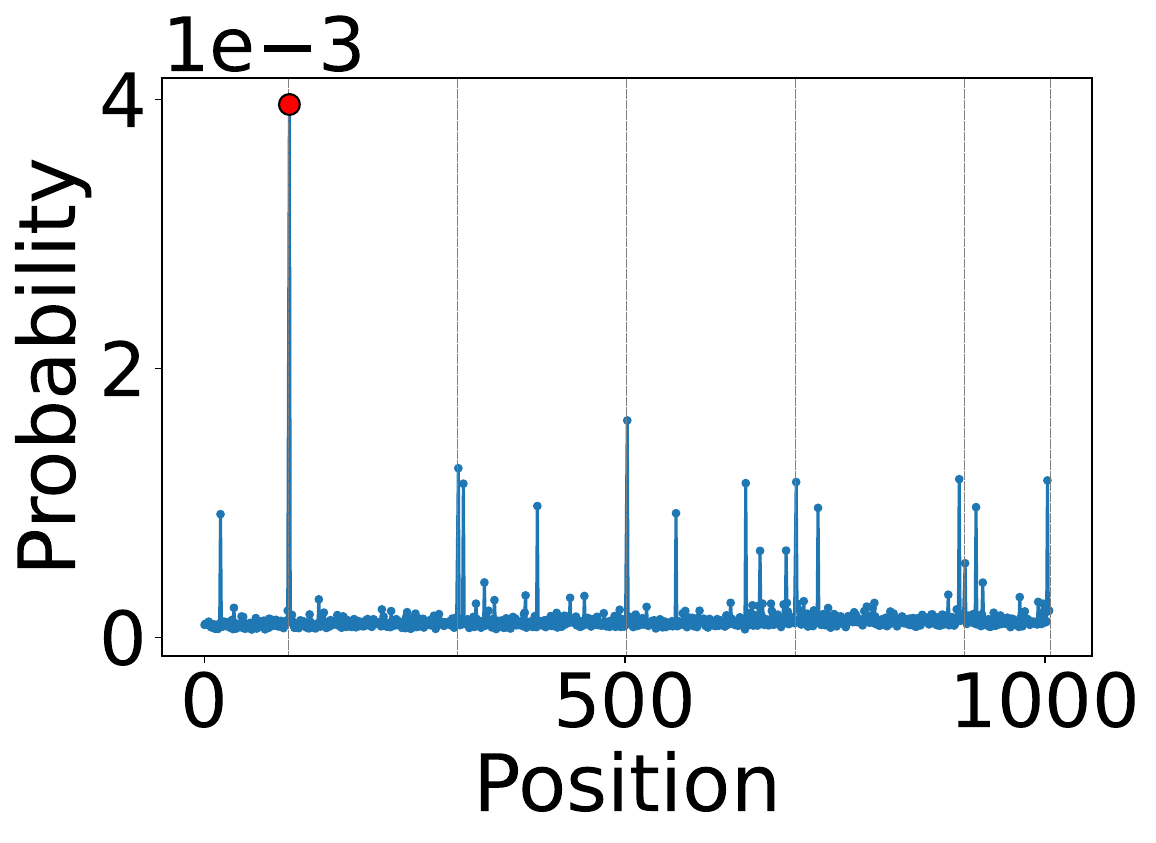} &
\FigGridCell{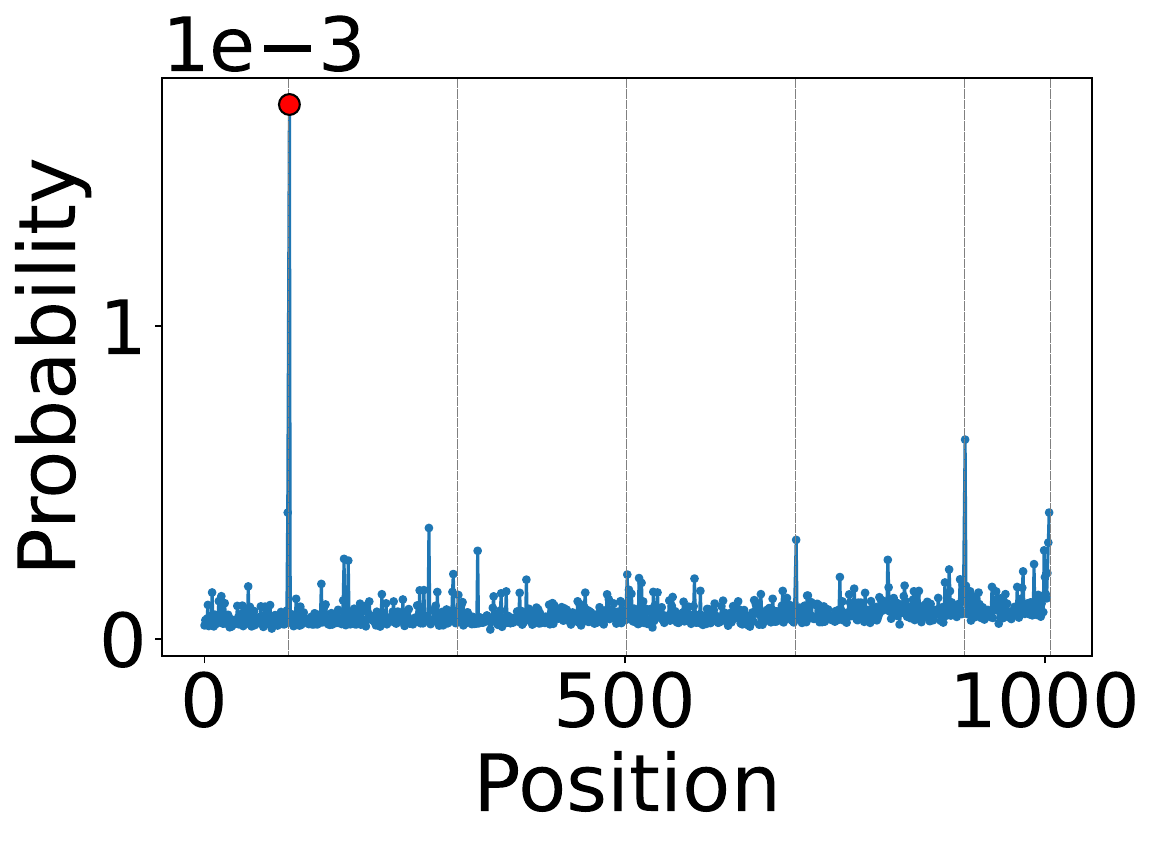} &
\FigGridCell{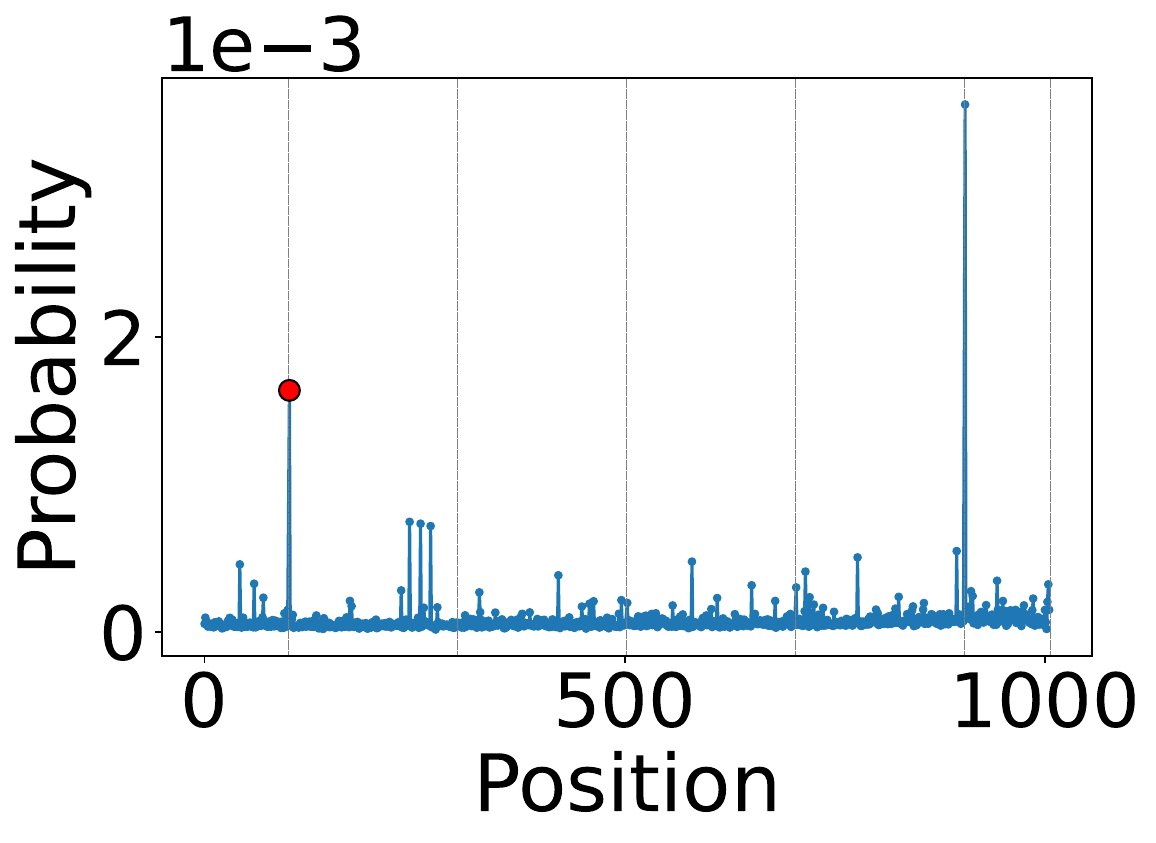} &
\FigGridCell{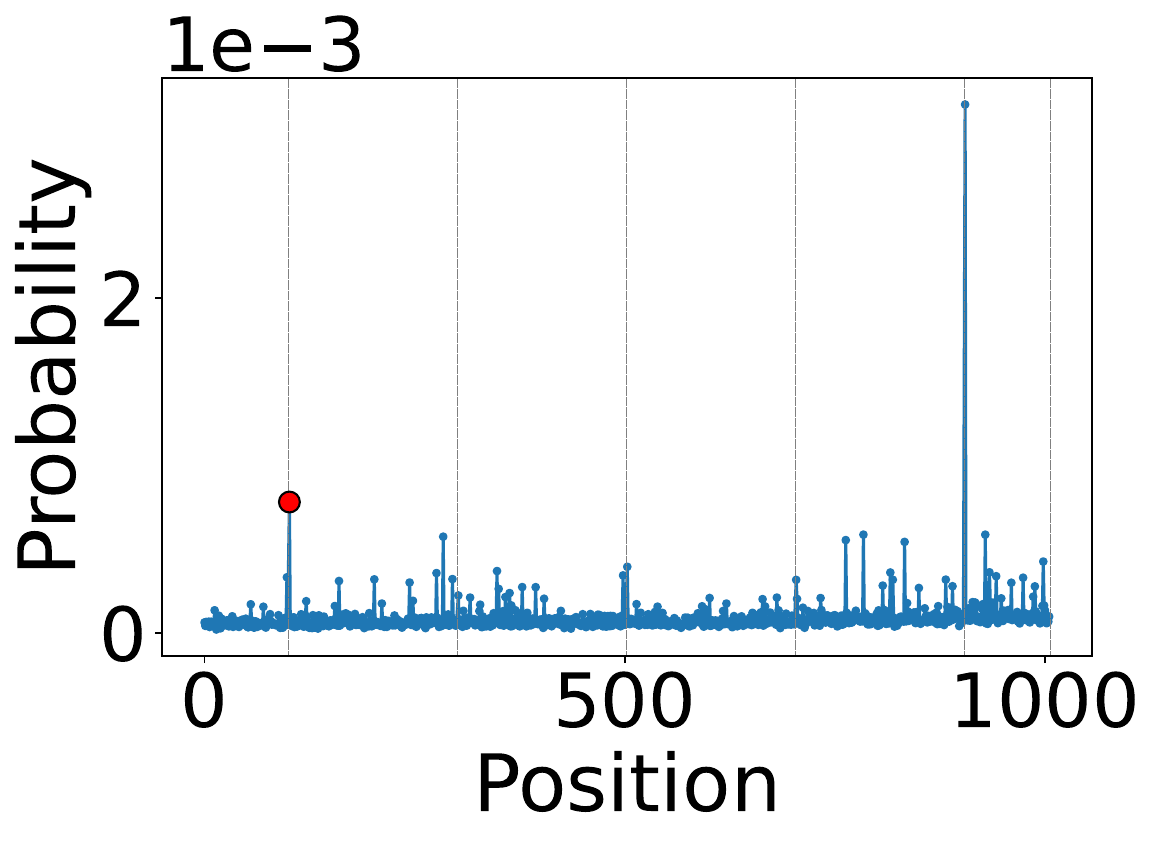} &
\FigGridCell{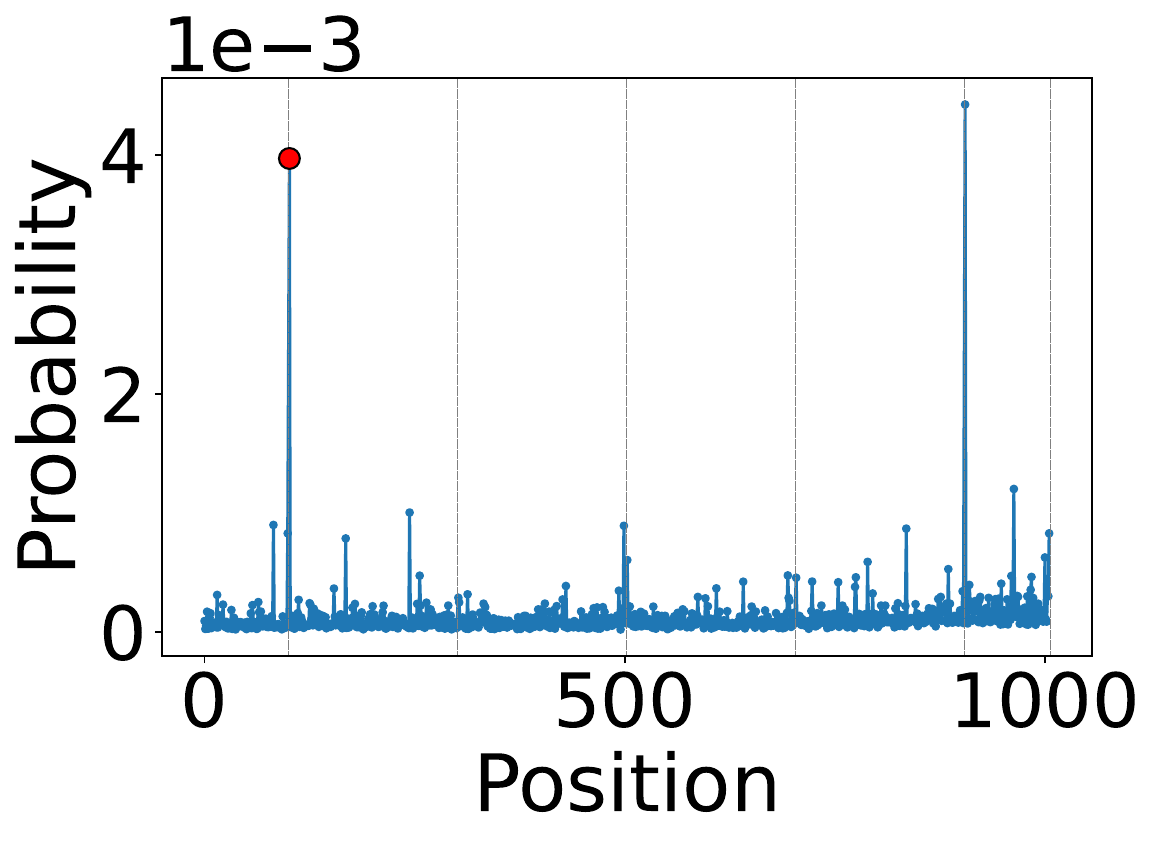} \\
& \FigGridVSubLabel{P2} &
\FigGridCell{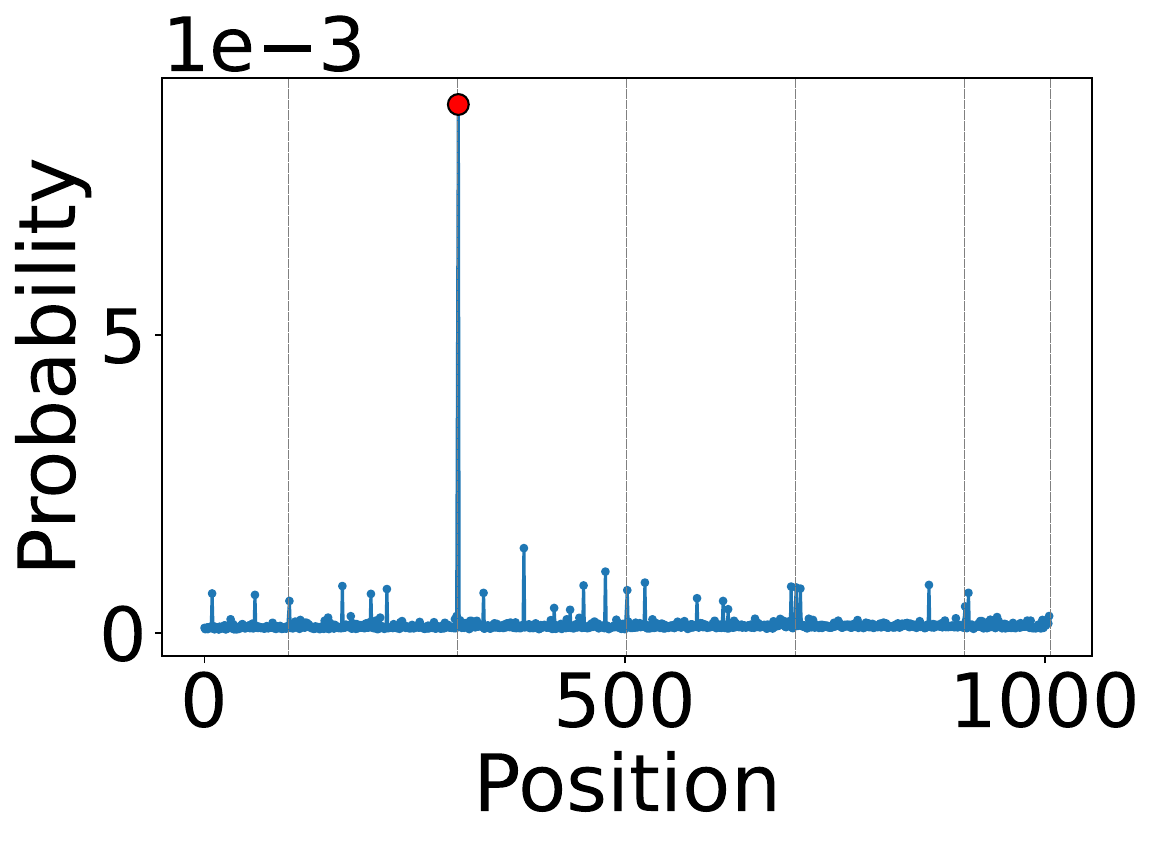} &
\FigGridCell{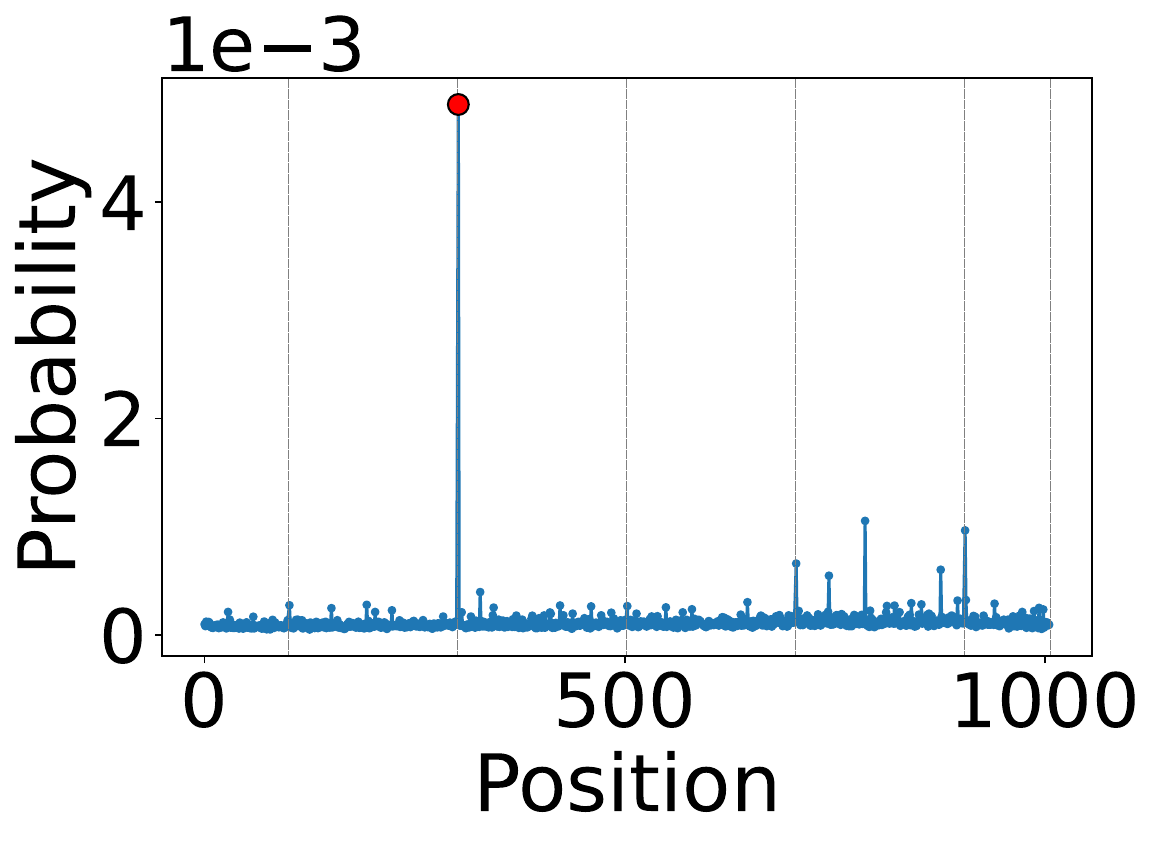} &
\FigGridCell{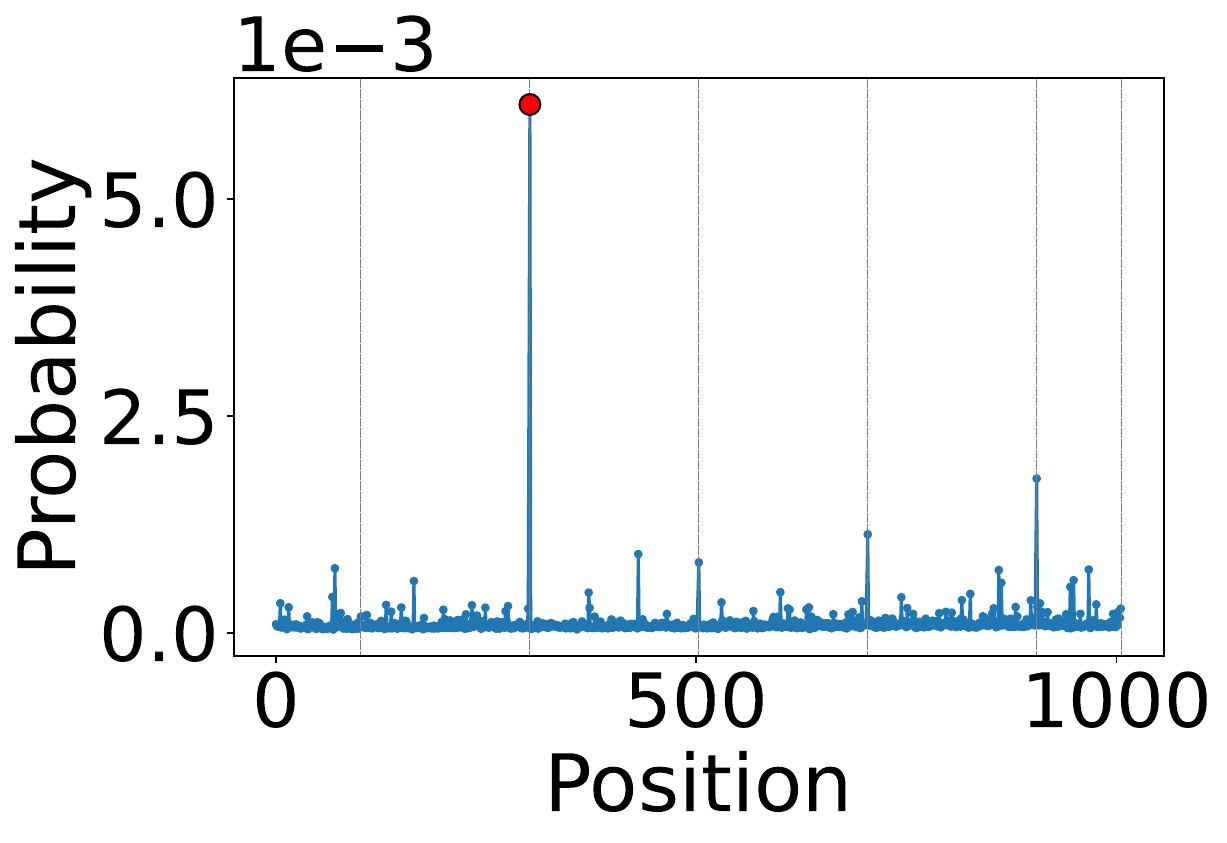} &
\FigGridCell{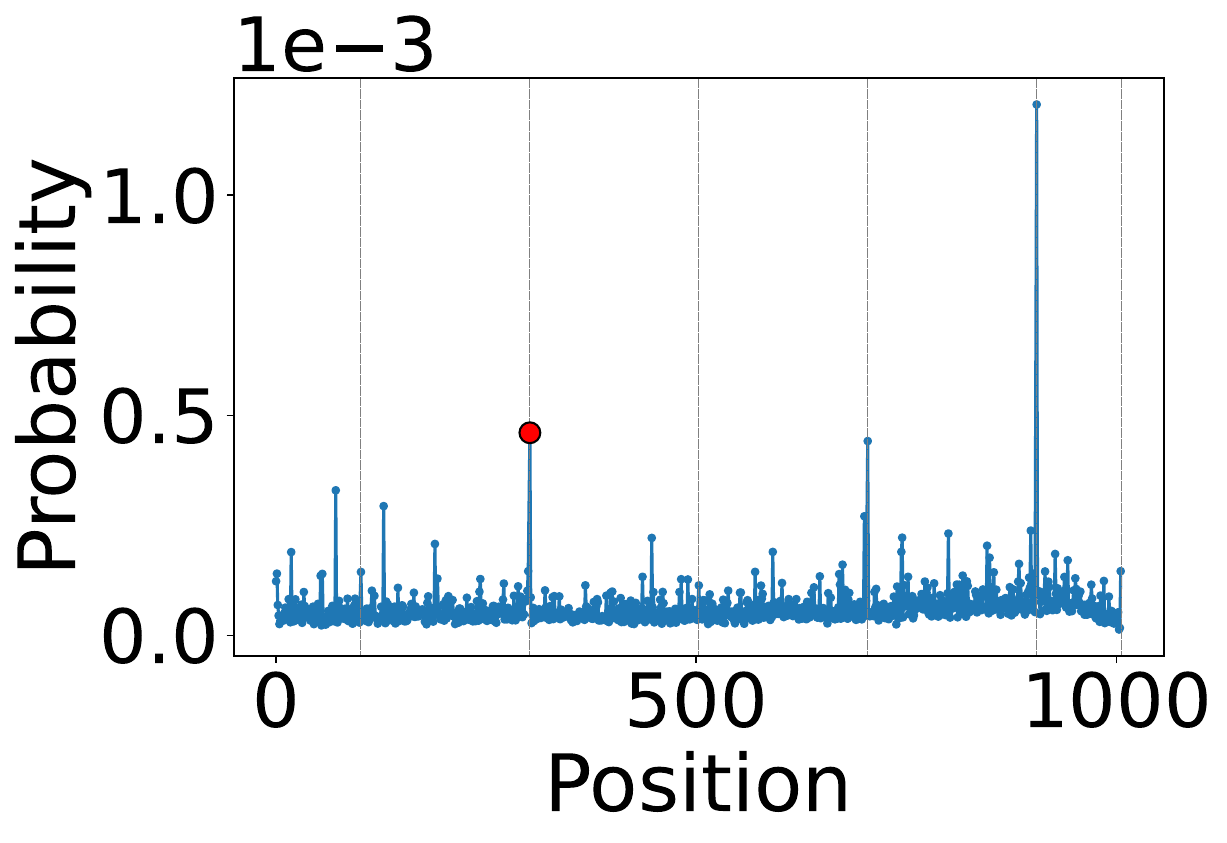} &
\FigGridCell{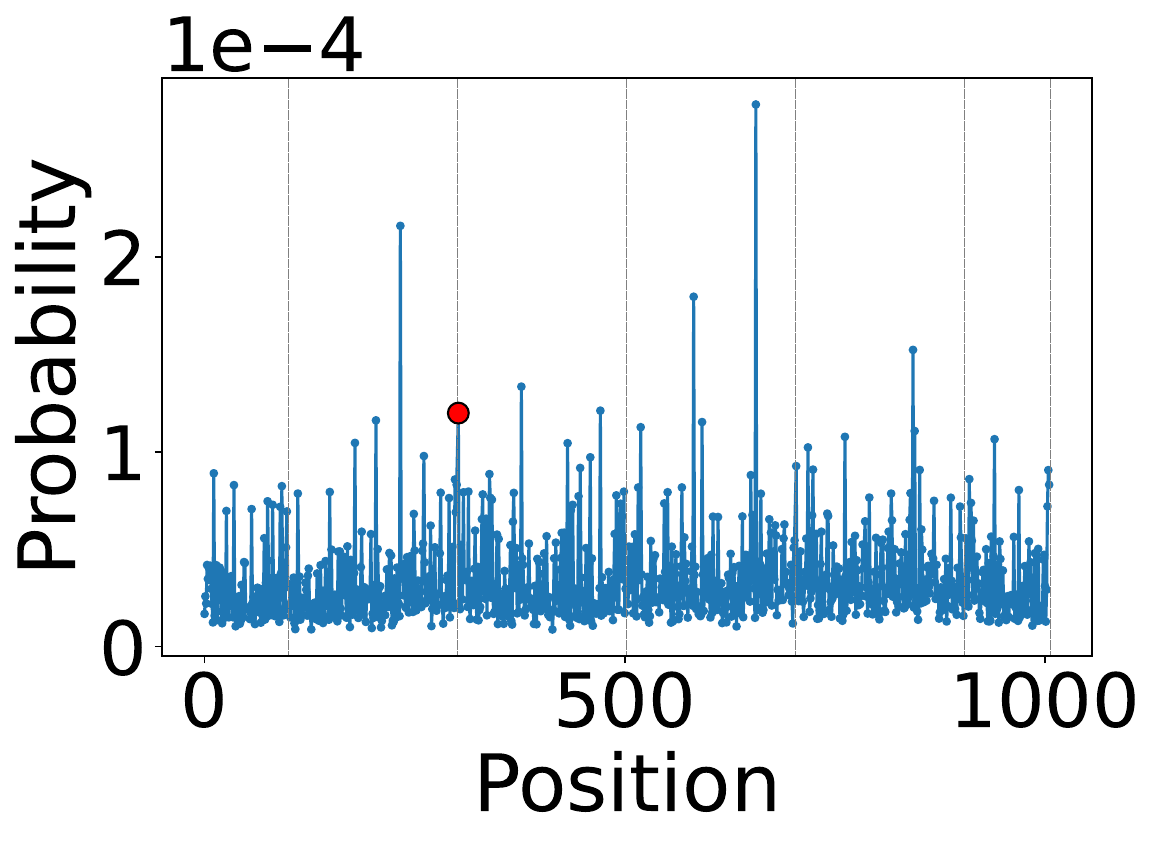} \\
& \FigGridVSubLabel{P3} &
\FigGridCell{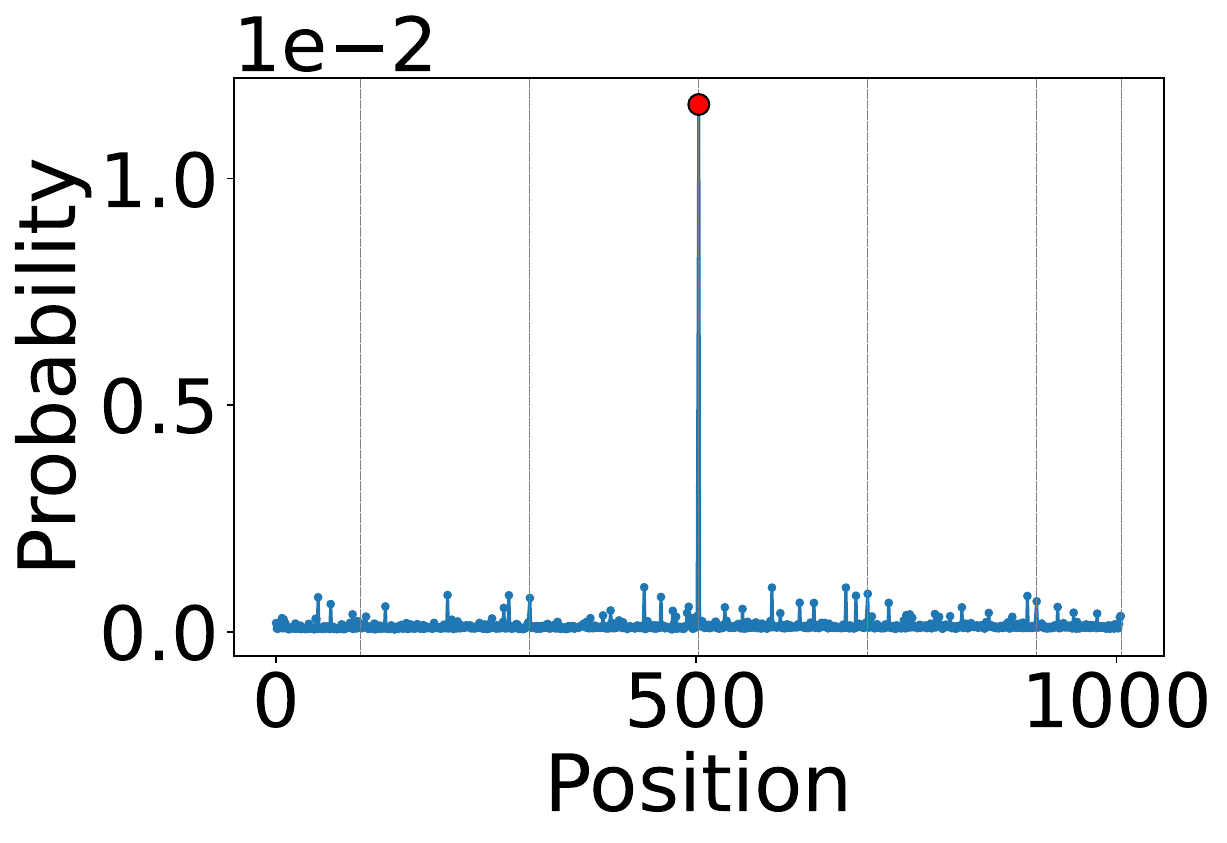} &
\FigGridCell{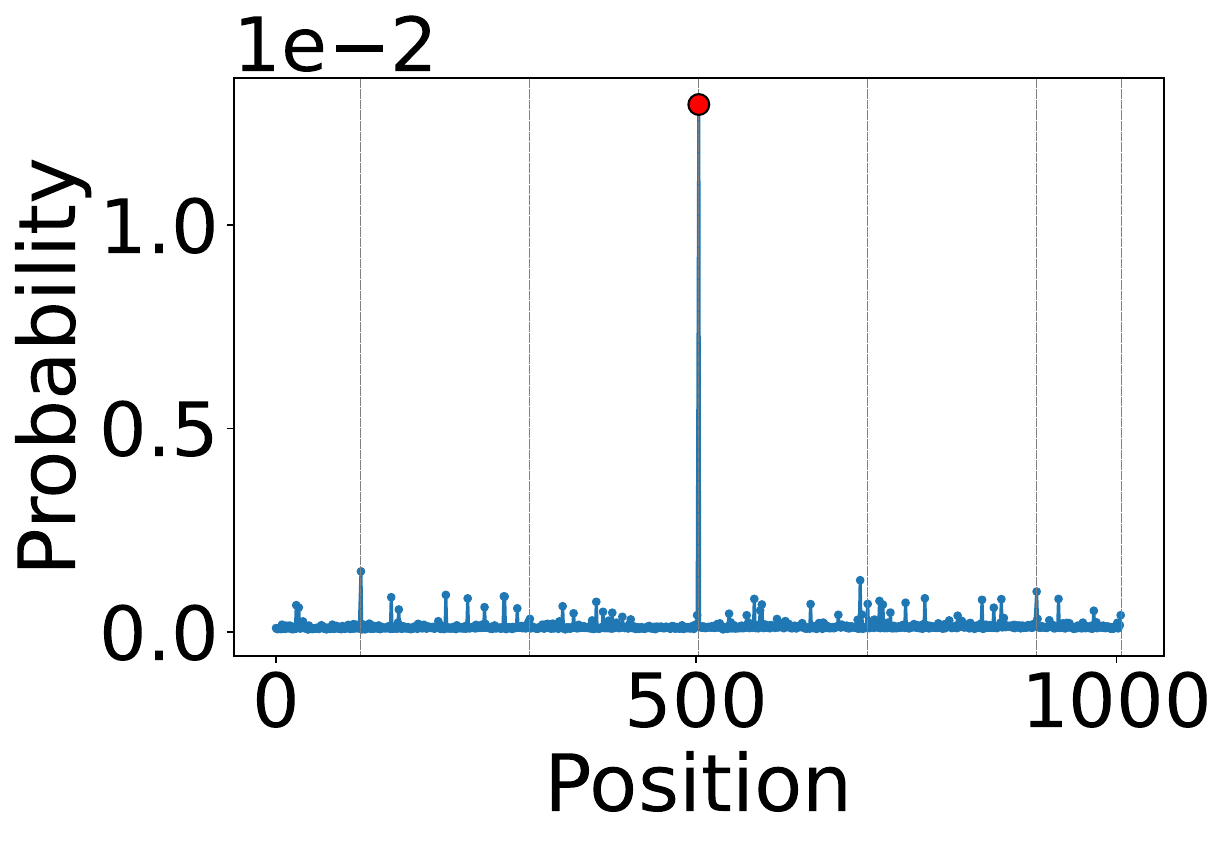} &
\FigGridCell{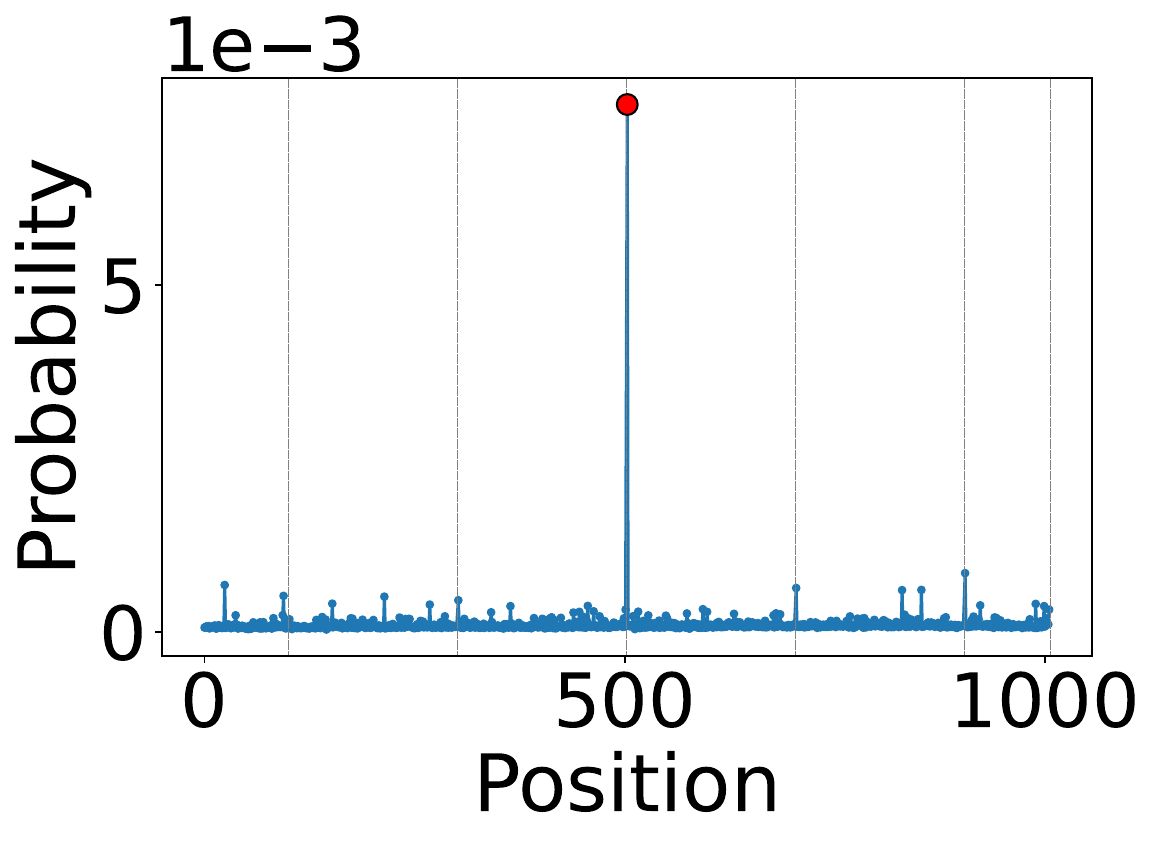} &
\FigGridCell{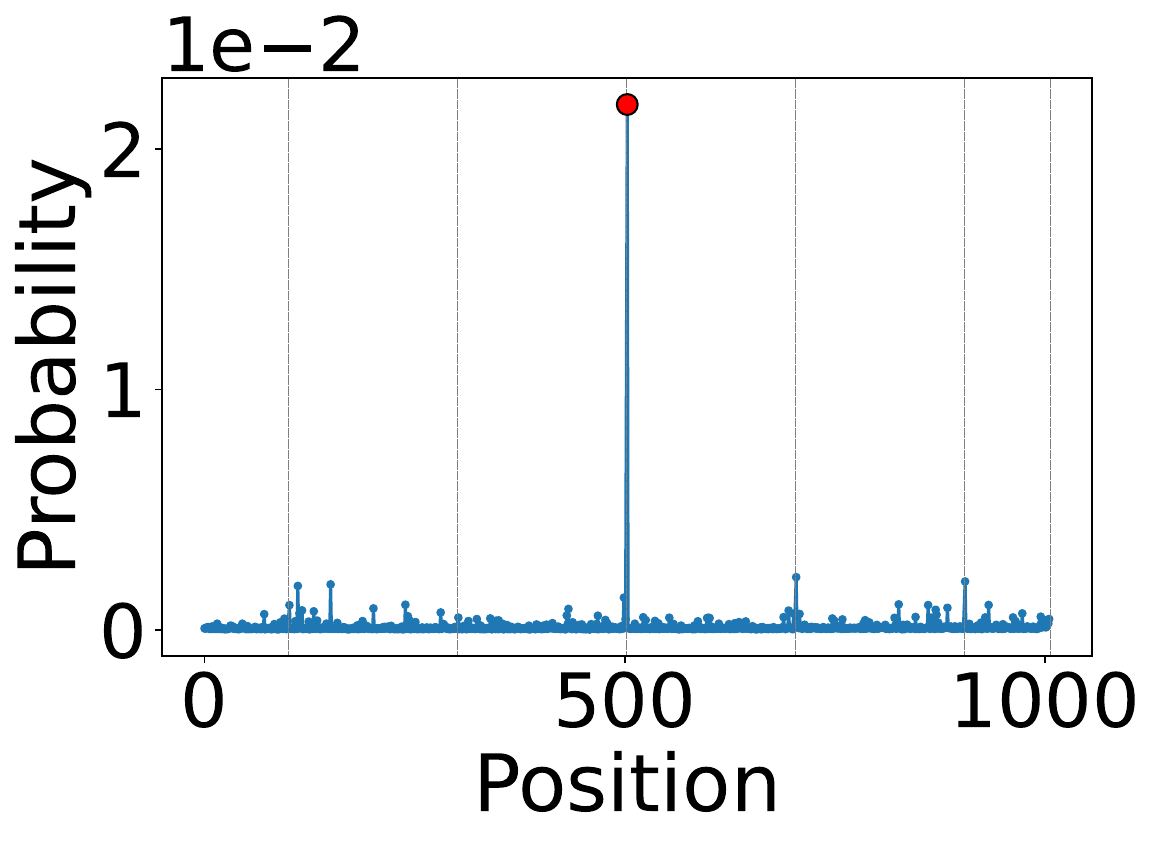} &
\FigGridCell{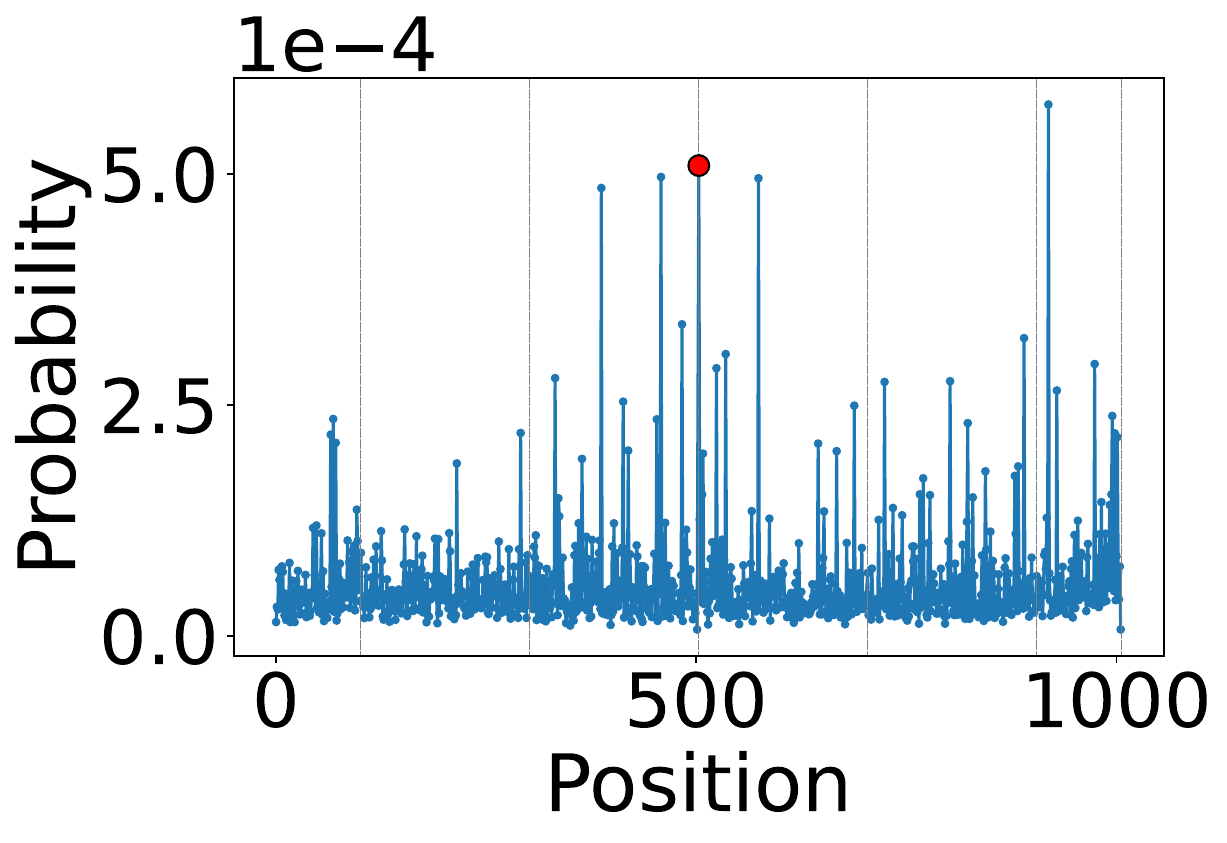} \\
& \FigGridVSubLabel{P4} &
\FigGridCell{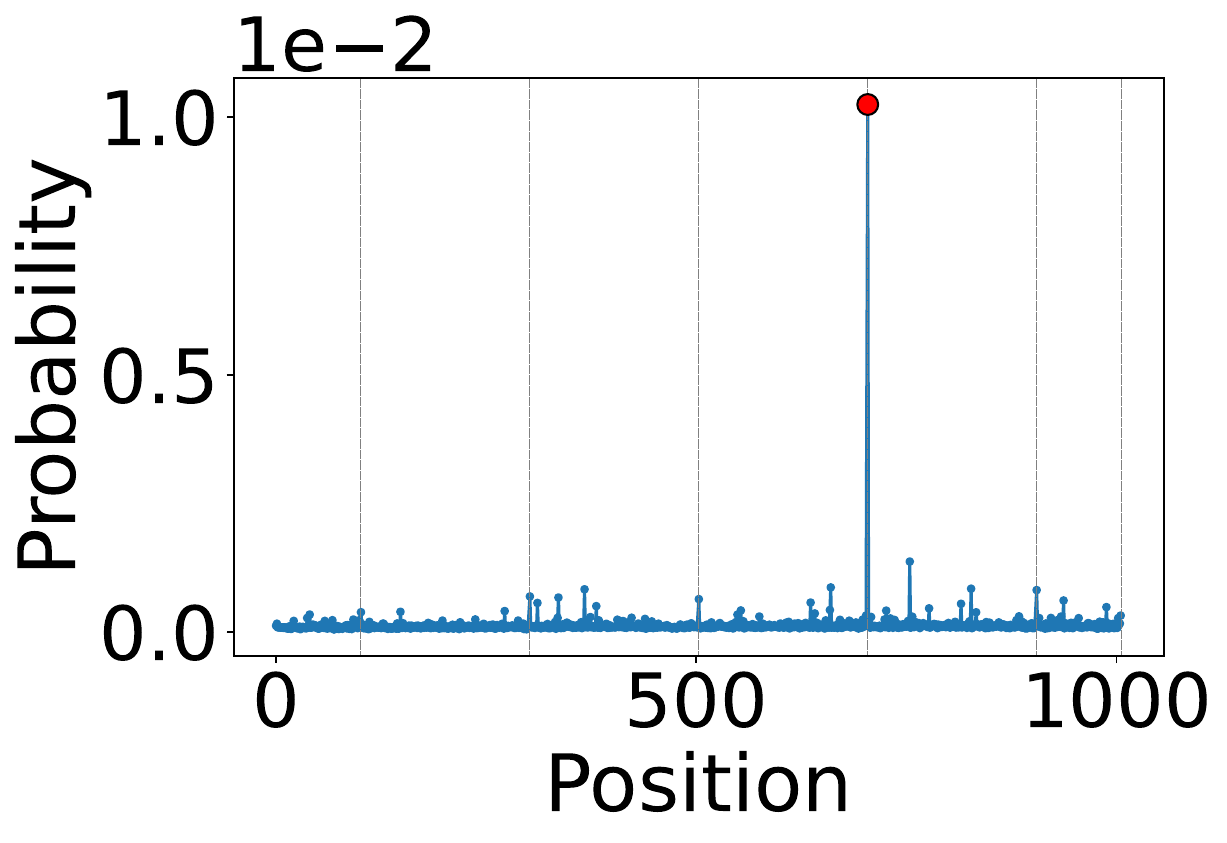} &
\FigGridCell{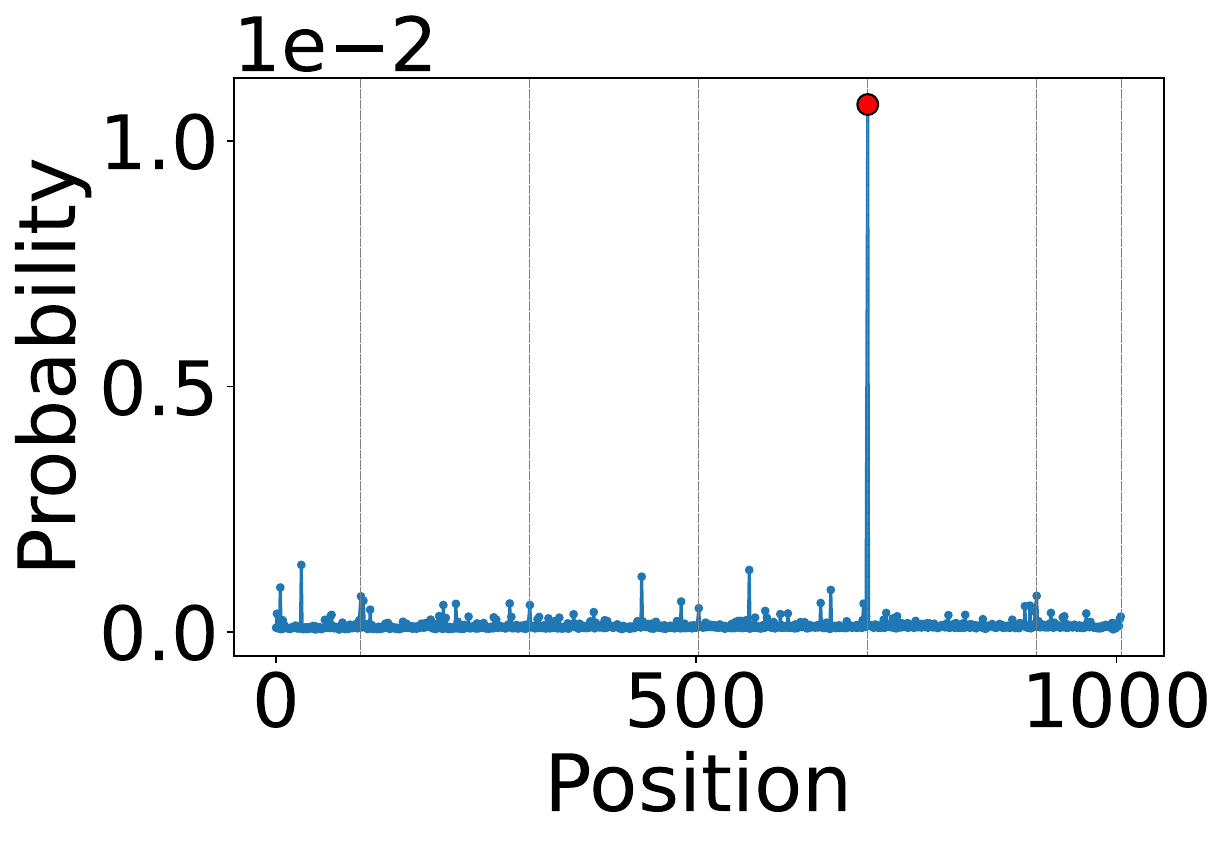} &
\FigGridCell{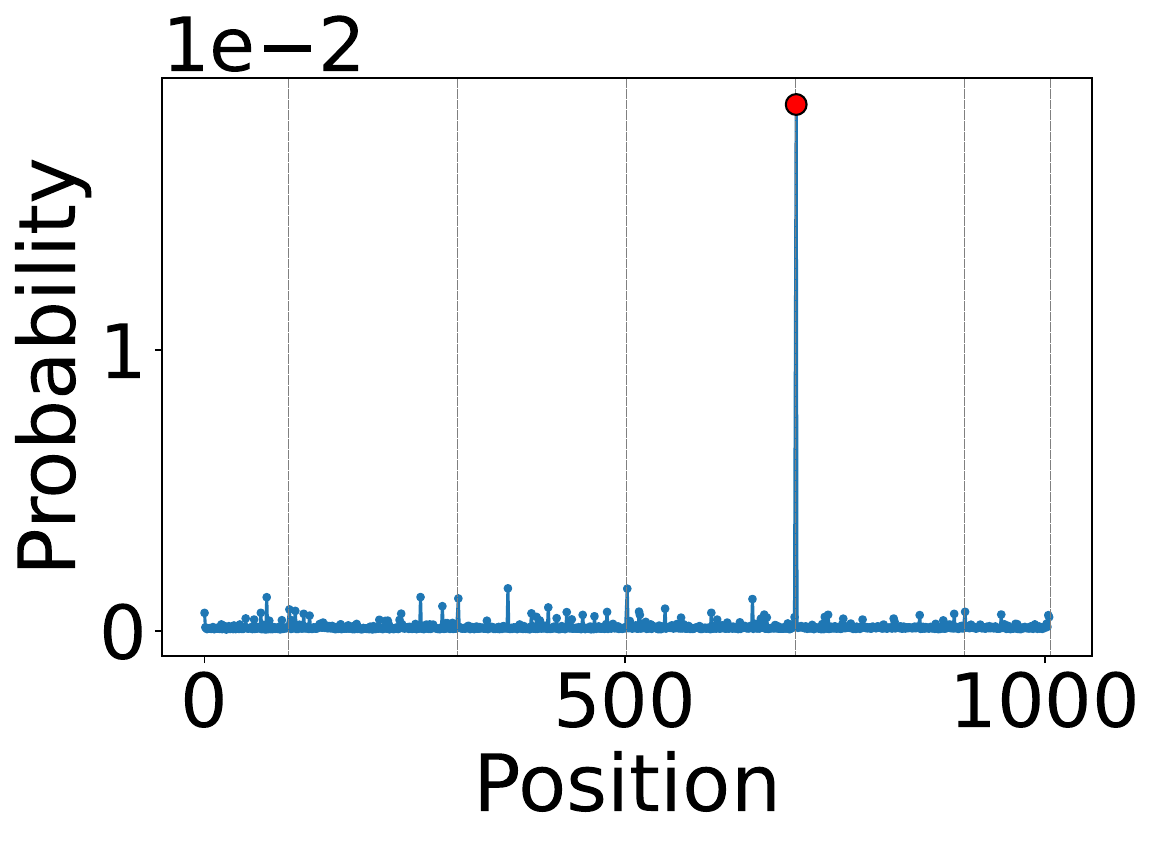} &
\FigGridCell{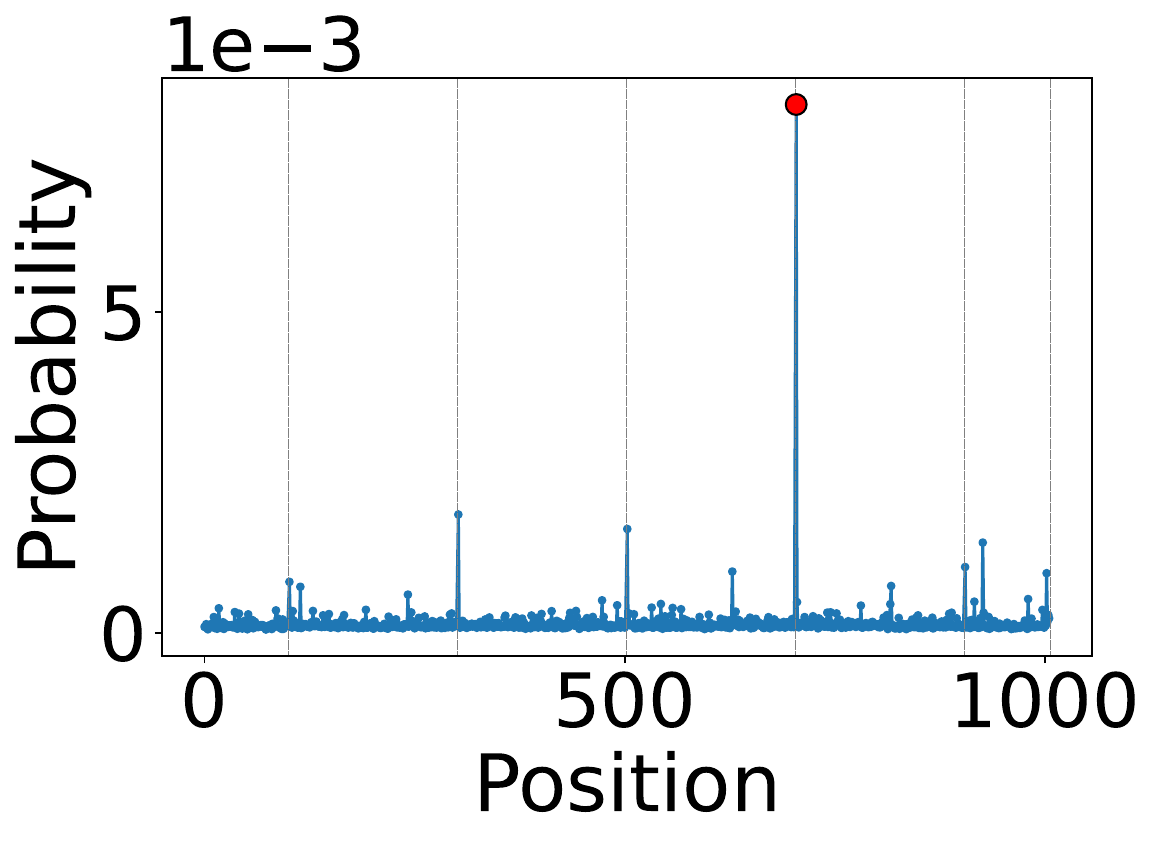} &
\FigGridCell{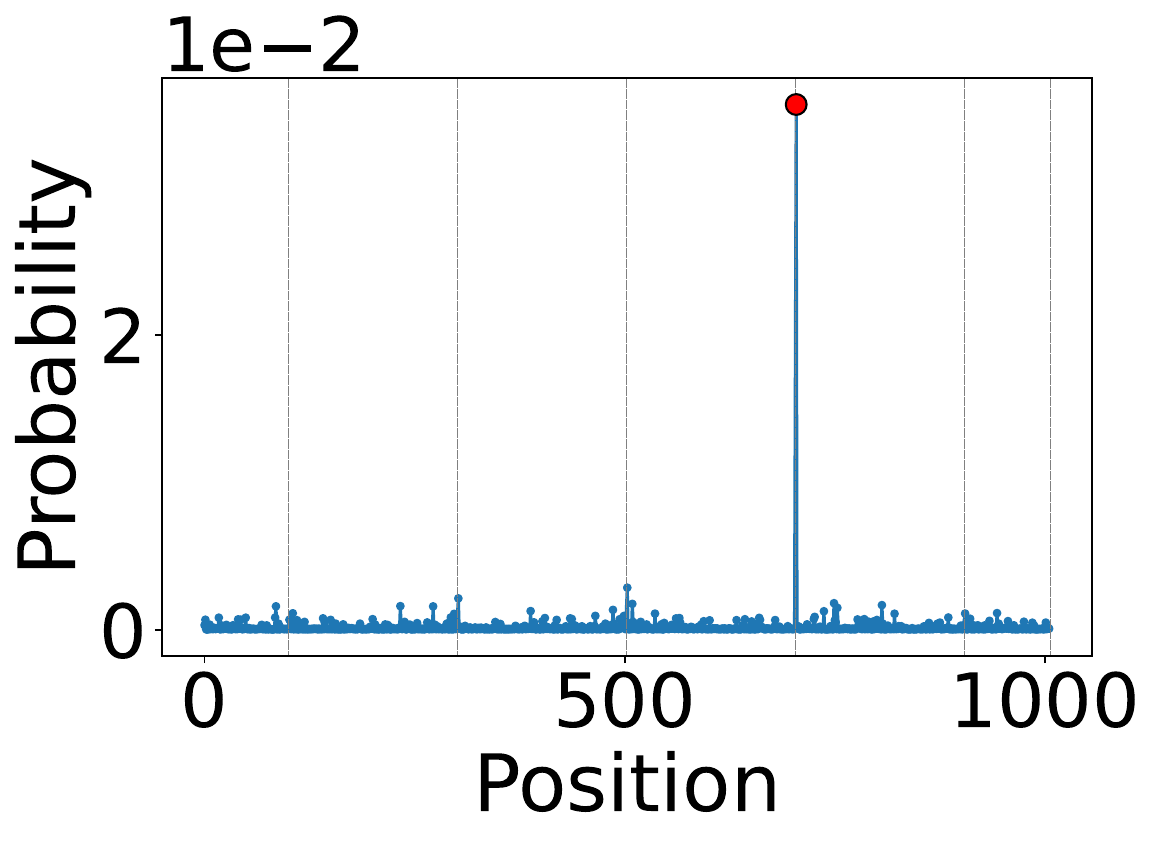} \\
& \FigGridVSubLabel{P5} &
\FigGridCell{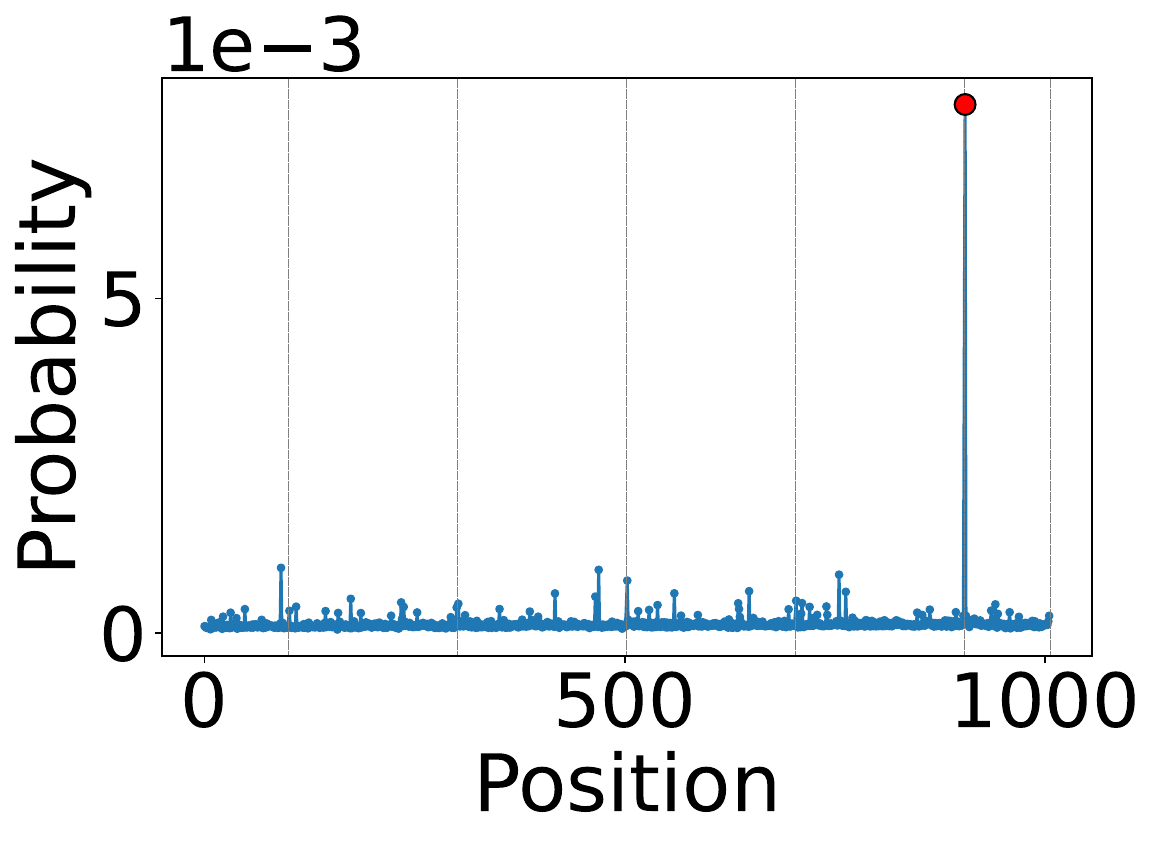} &
\FigGridCell{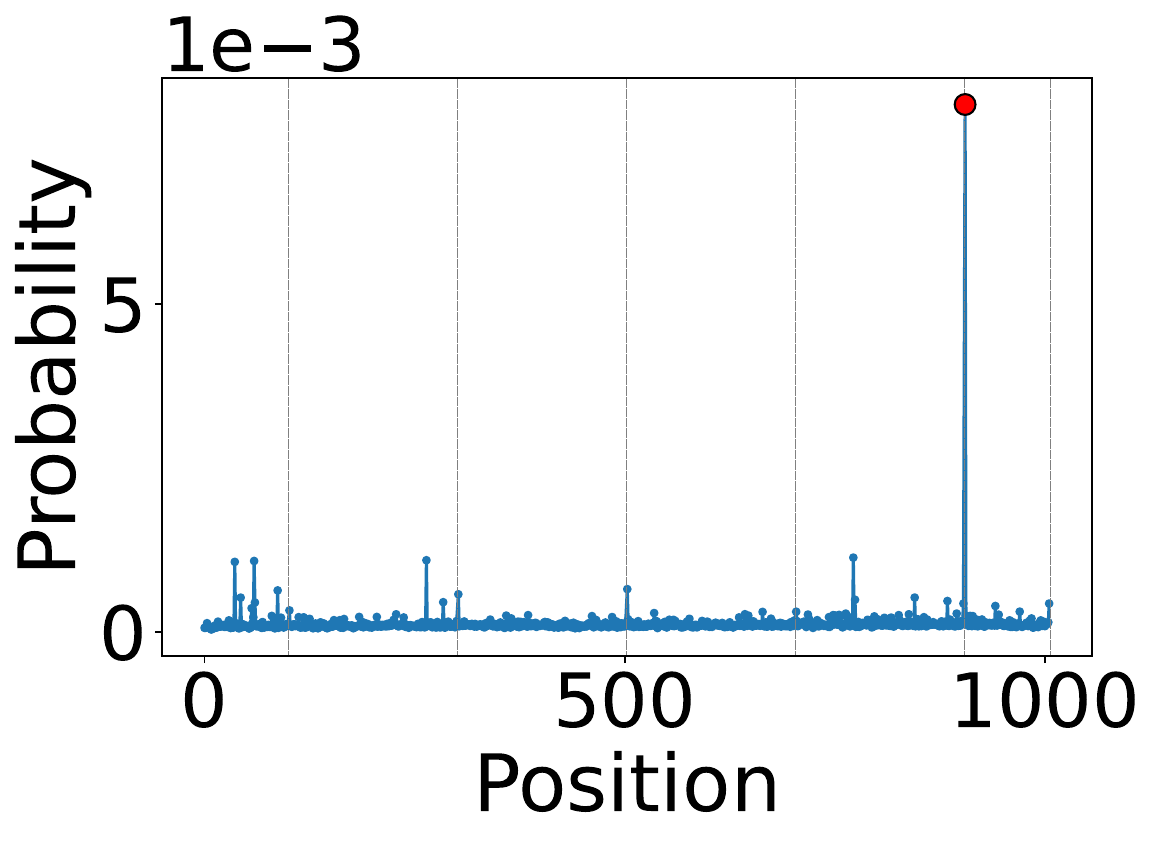} &
\FigGridCell{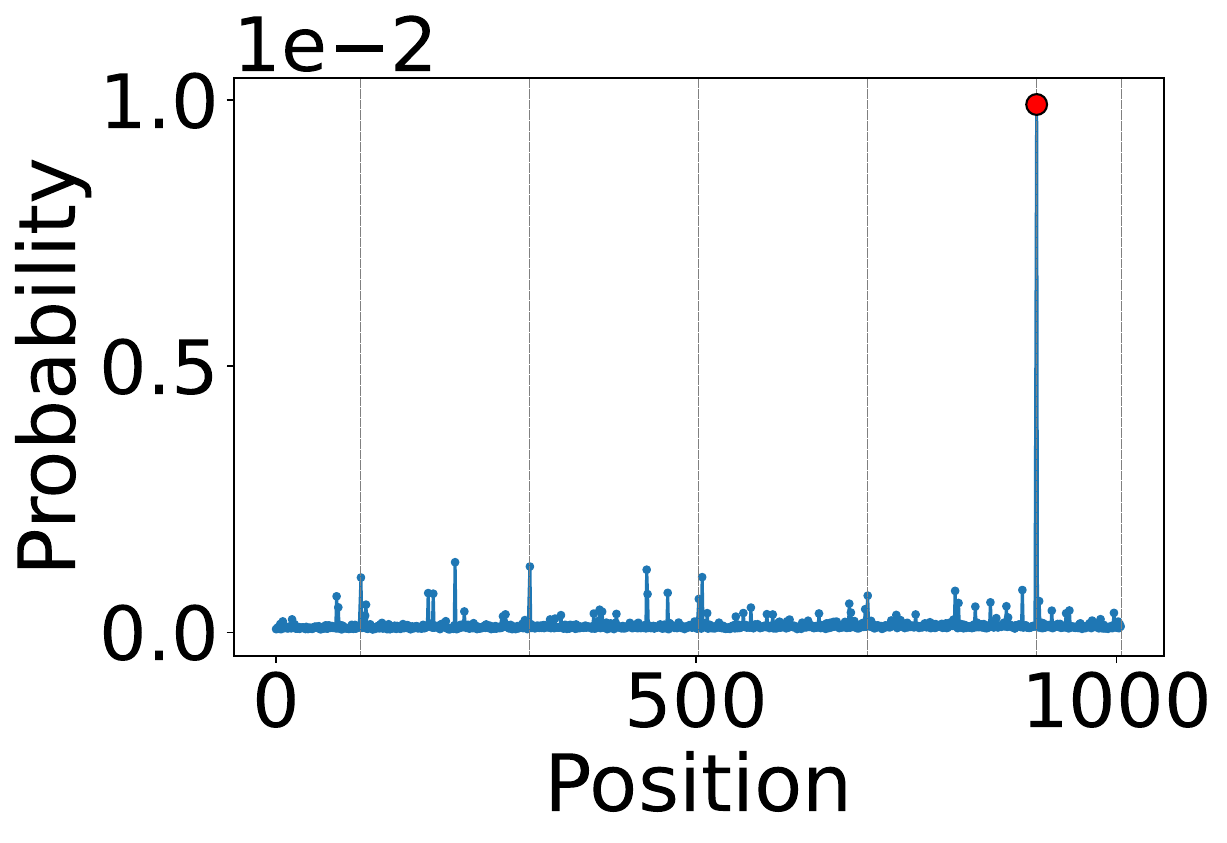} &
\FigGridCell{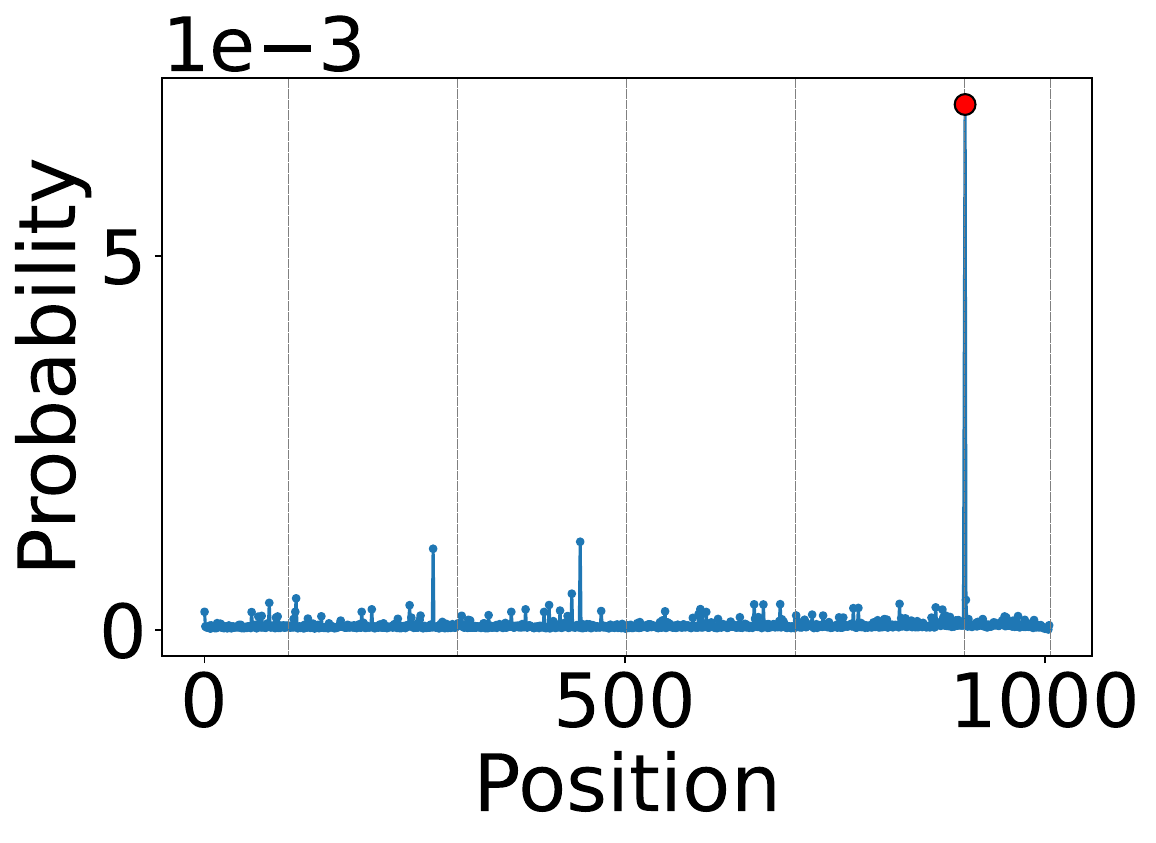} &
\FigGridCell{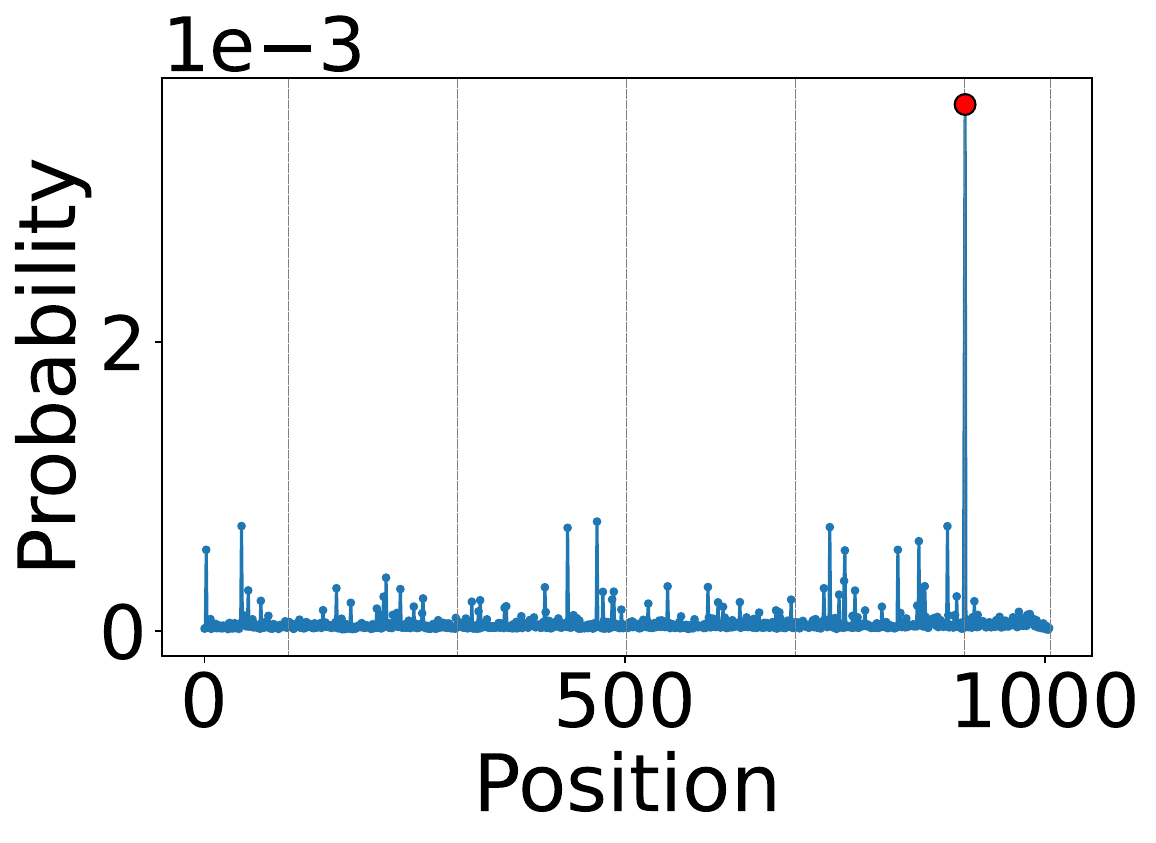} \\
\bottomrule
\end{tabular}
\endgroup
\caption{Gemma ablation effect (Exp. 2). Episodic retrieval probability after ablating Induction or Random heads. Columns show number of ablated heads.}
\label{fig:exp2_gemma_ablation}
\end{figure*}

\begin{figure*}[h!]
\centering
\begingroup
\setlength{\FigGridLabelW}{1.05cm}
\setlength{\FigGridSubLabelW}{0.55cm}
\setlength{\FigGridCellH}{1.85cm} 
\setlength{\FigGridLabelSep}{1pt}
\setlength{\FigGridSubLabelSep}{1pt}
\setlength{\FigGridColSep}{0.5pt}

\FigGridSetupTwo{4}
\renewcommand{\arraystretch}{1.0}

\begin{tabular}{@{}>{\centering\arraybackslash}m{\FigGridLabelW}@{\hspace{\FigGridLabelSep}}
                >{\centering\arraybackslash}m{\FigGridSubLabelW}@{\hspace{\FigGridSubLabelSep}}
                *{4}{>{\centering\arraybackslash}m{\FigGridPlotW}}@{}}
\toprule
& & \multicolumn{4}{c}{\FigHdr Ablations (\%)} \\
\cmidrule(lr){3-6}
& & \FigHdr 0 & \FigHdr 1 & \FigHdr 5 & \FigHdr 10 \\
\midrule
\multirow{5}{*}{\FigGridVLabelSpanFive{Induction}} &
\FigGridVSubLabel{P1} &
\FigGridCell{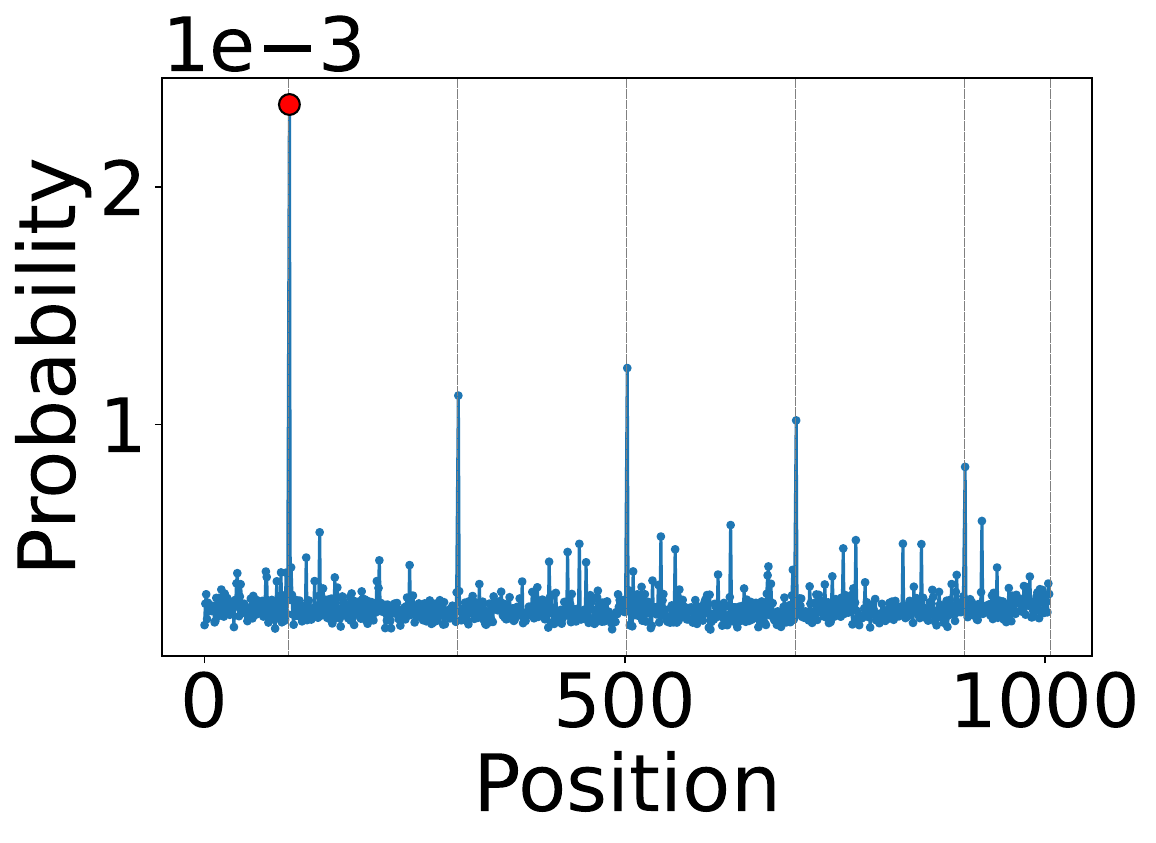} &
\FigGridCell{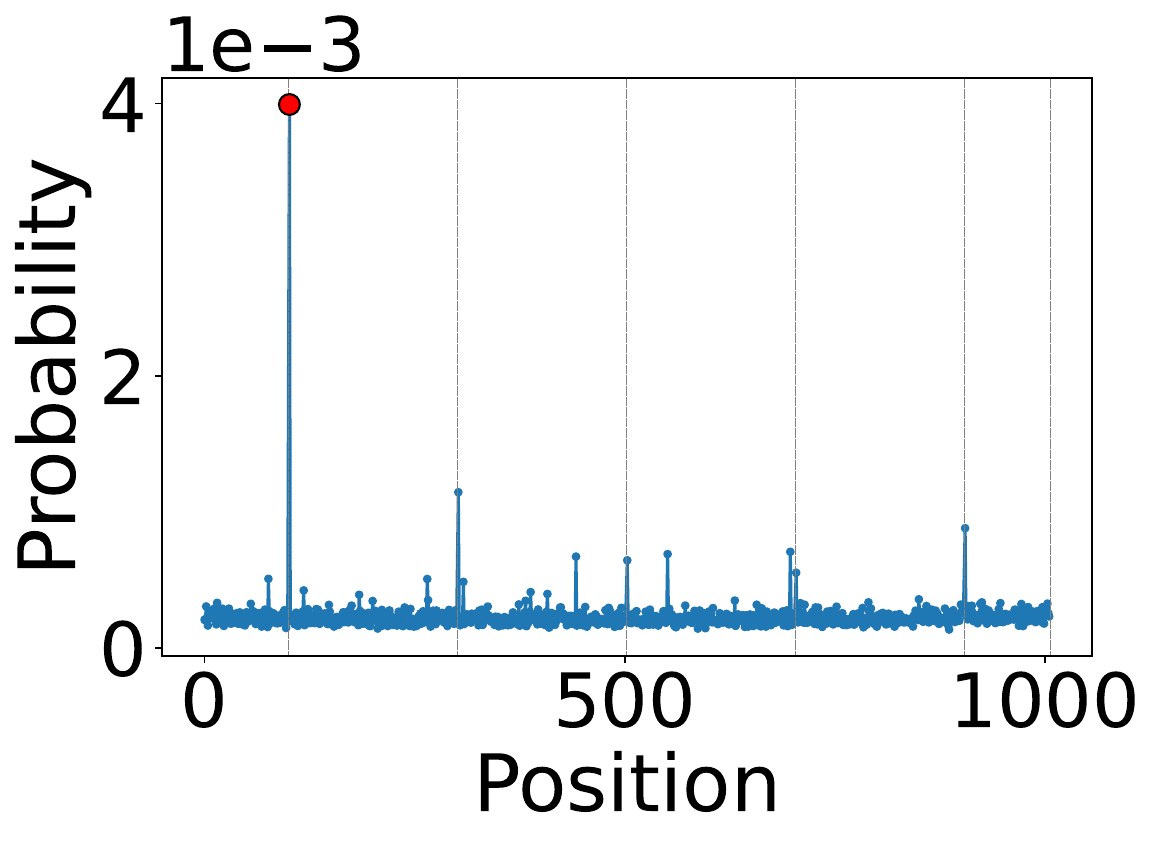} &
\FigGridCell{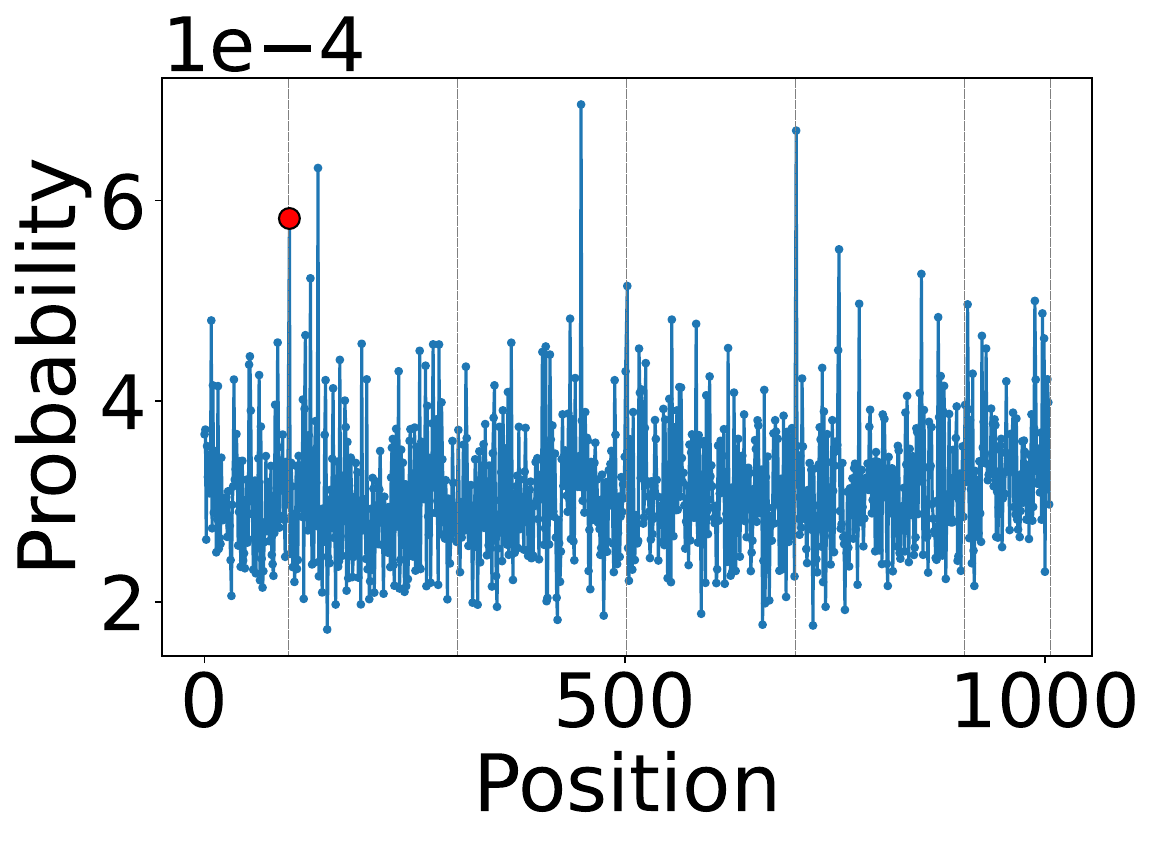} &
\FigGridCell{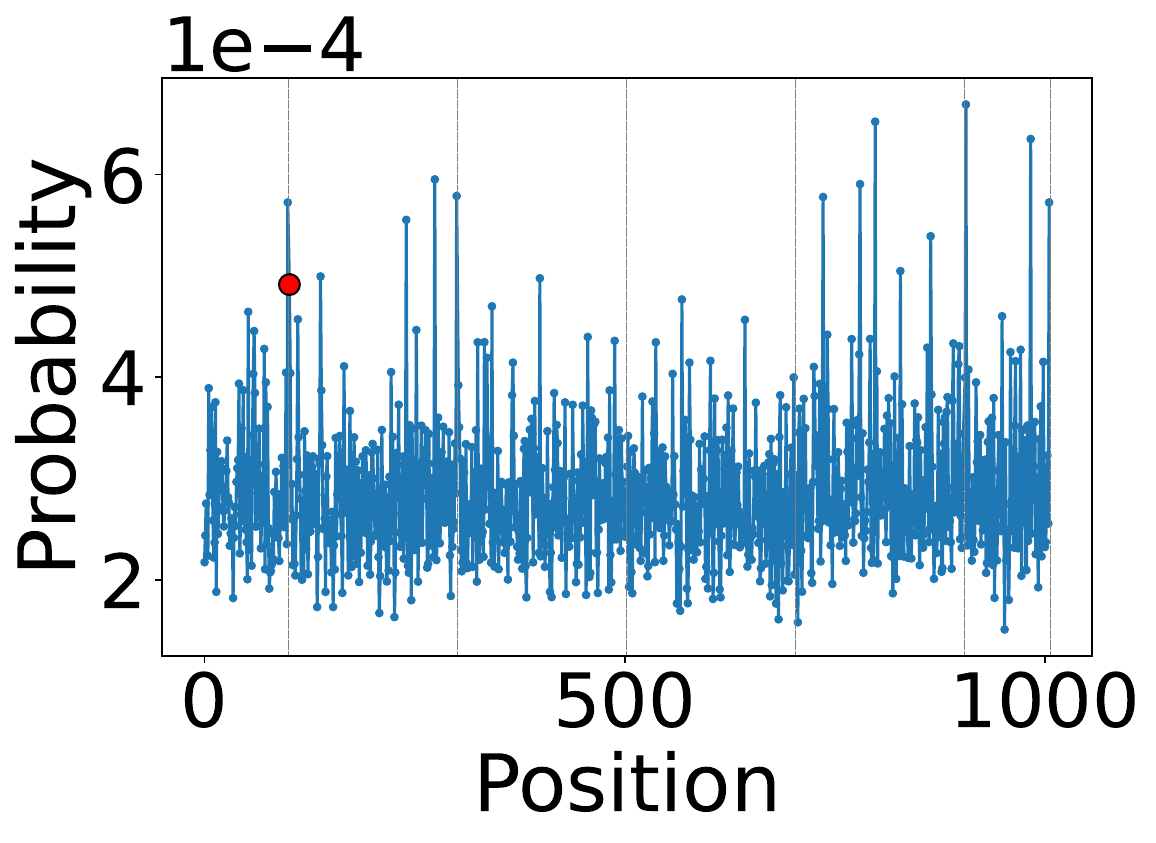} \\
& \FigGridVSubLabel{P2} &
\FigGridCell{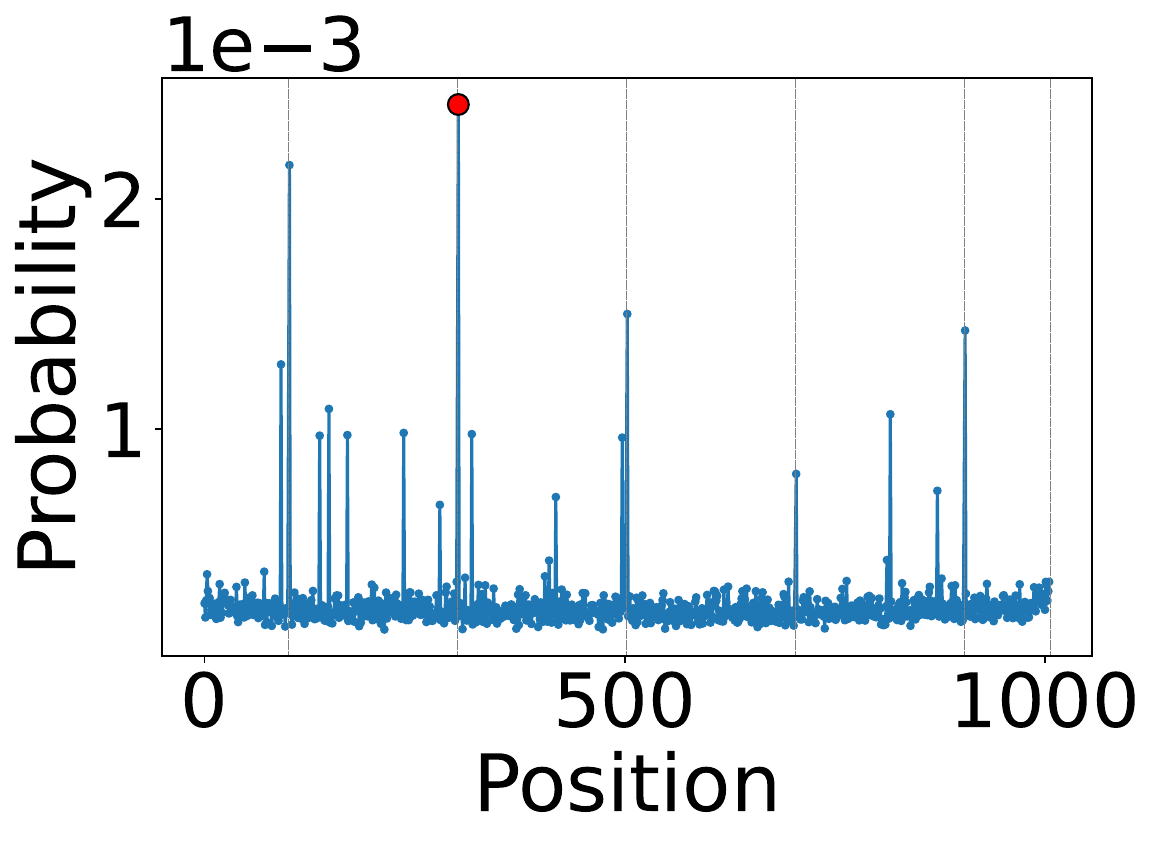} &
\FigGridCell{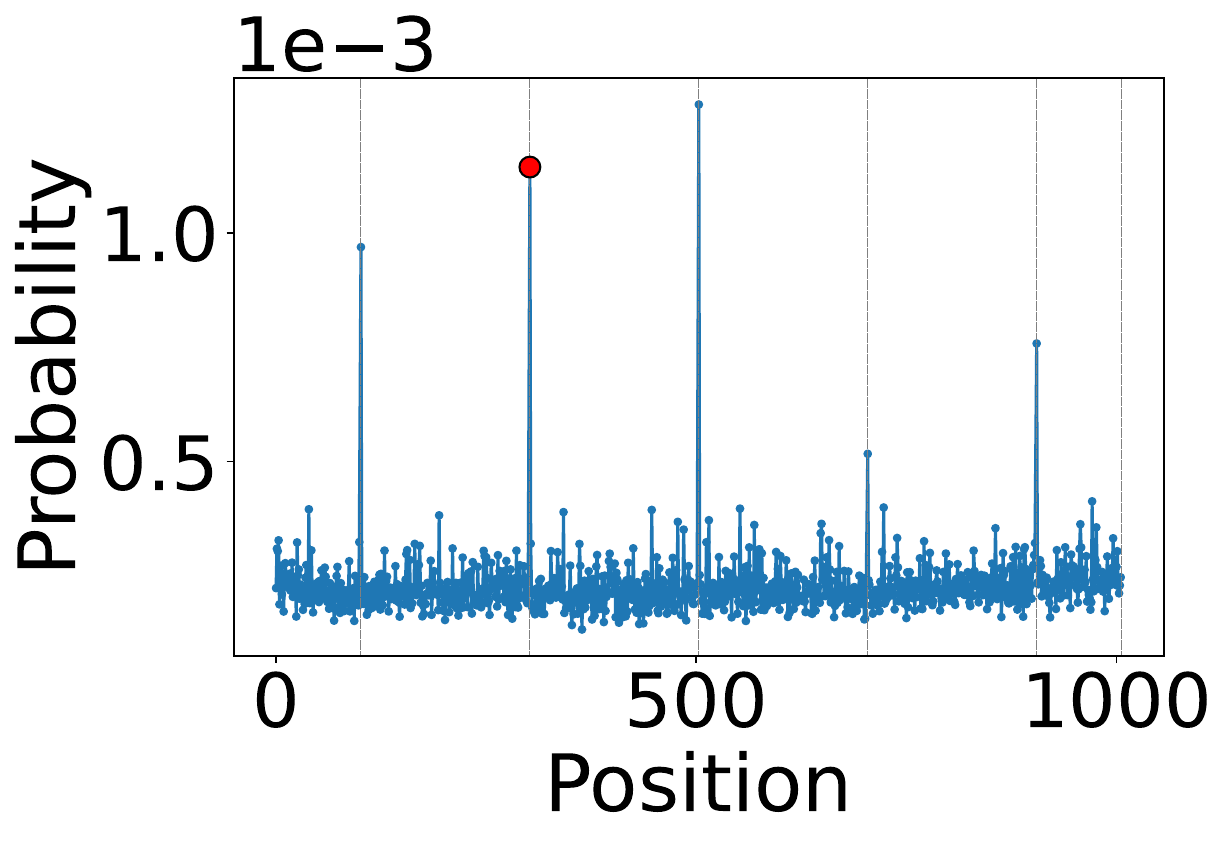} &
\FigGridCell{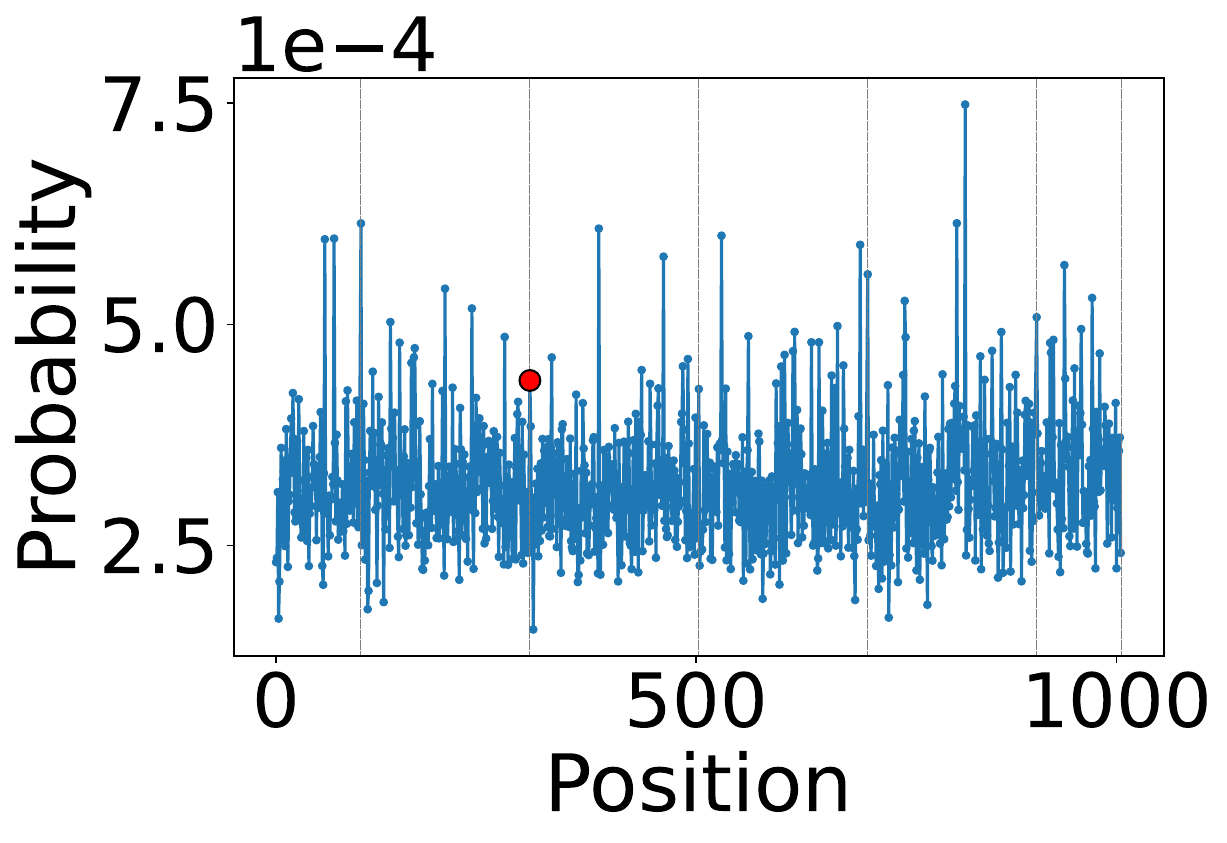} &
\FigGridCell{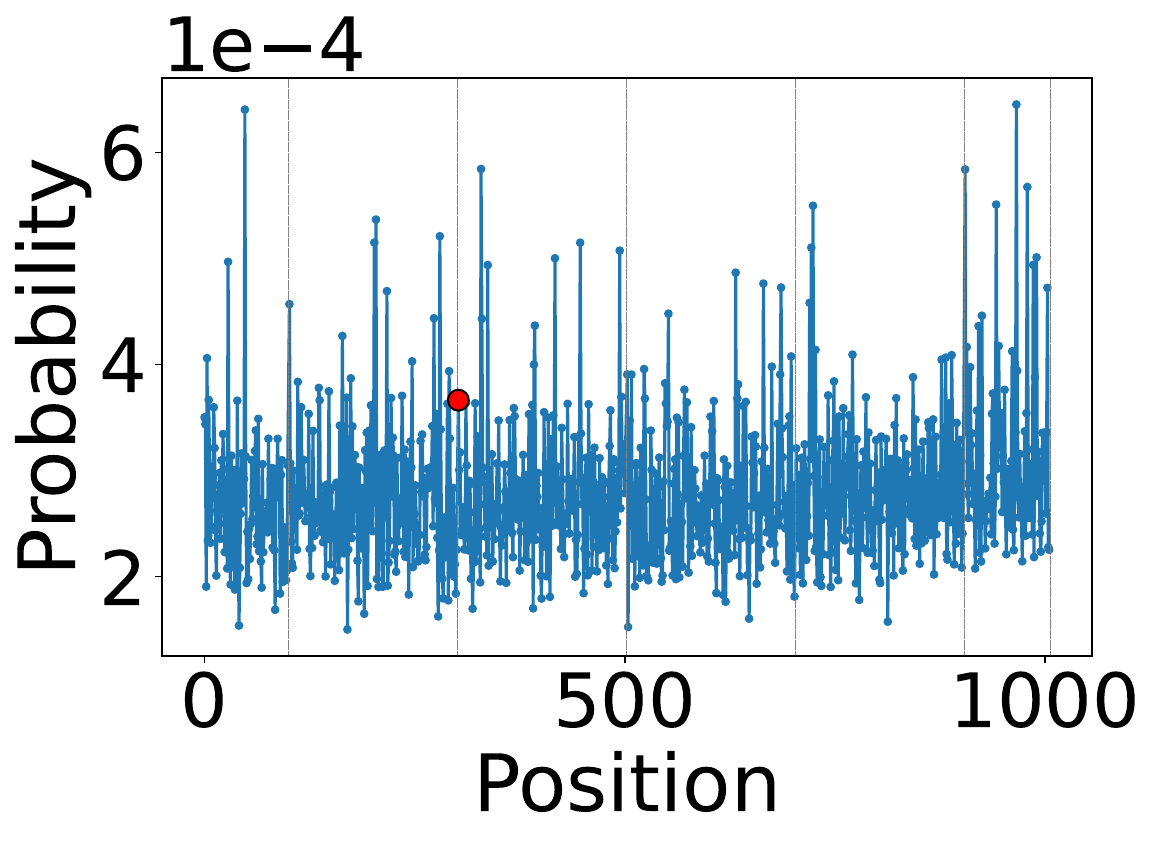} \\
& \FigGridVSubLabel{P3} &
\FigGridCell{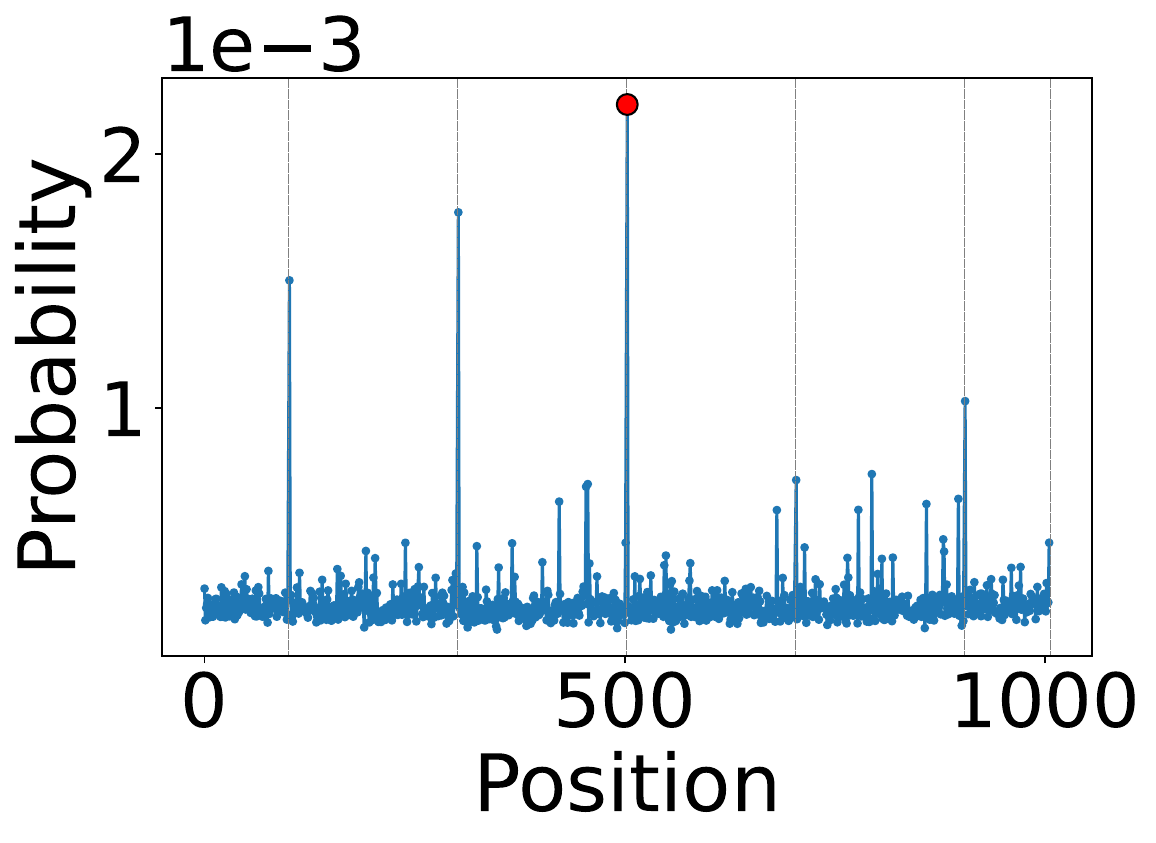} &
\FigGridCell{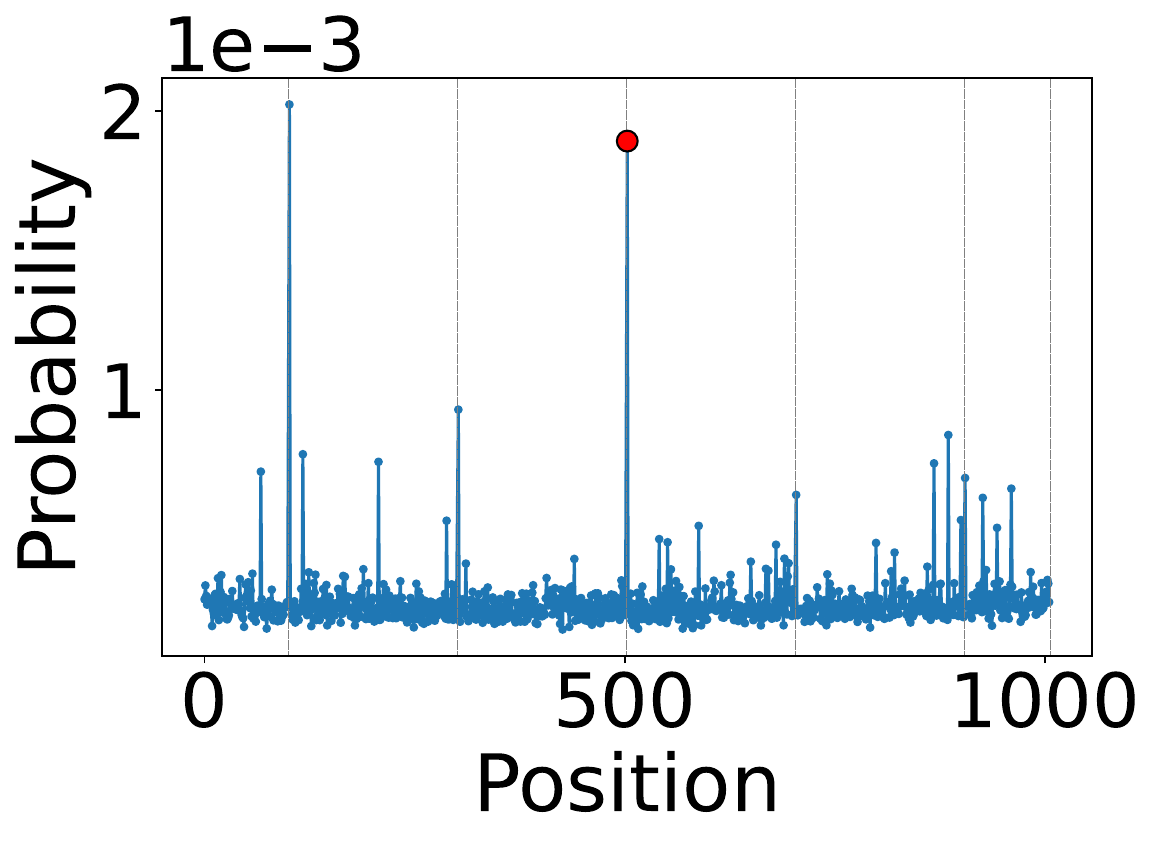} &
\FigGridCell{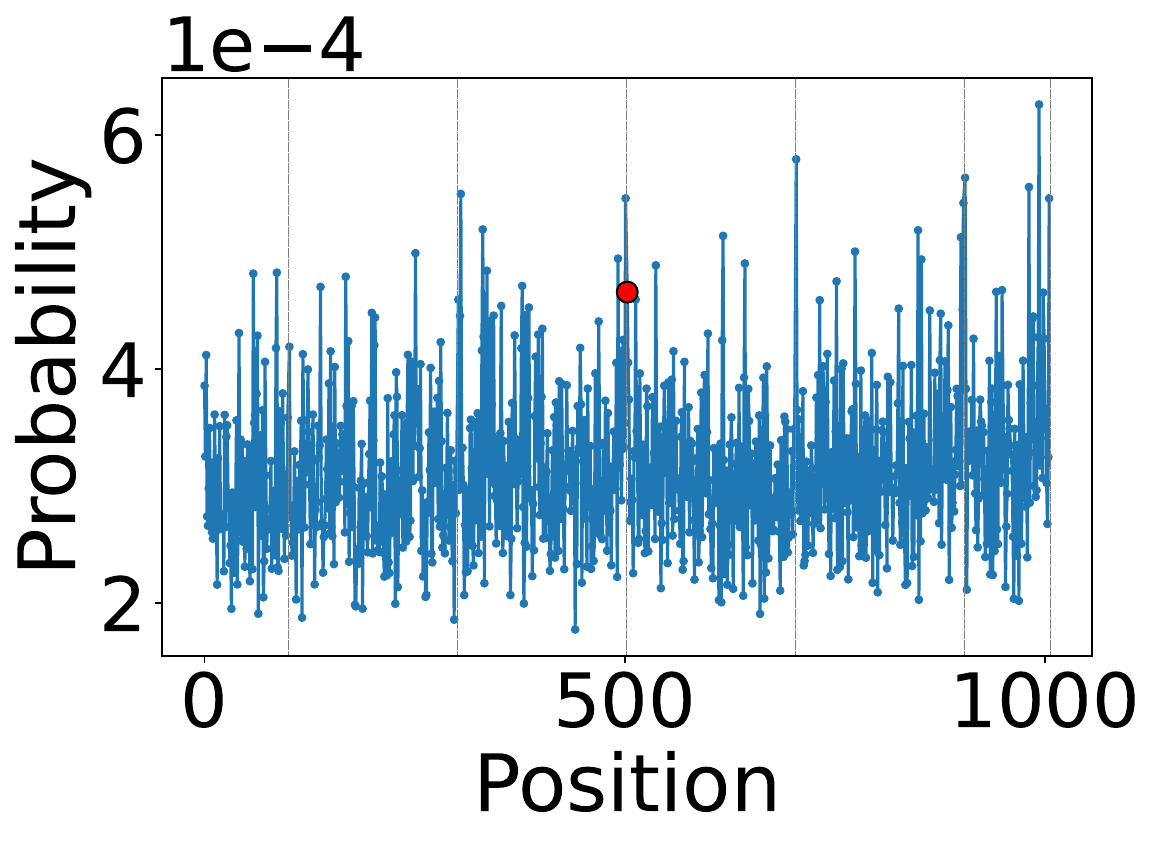} &
\FigGridCell{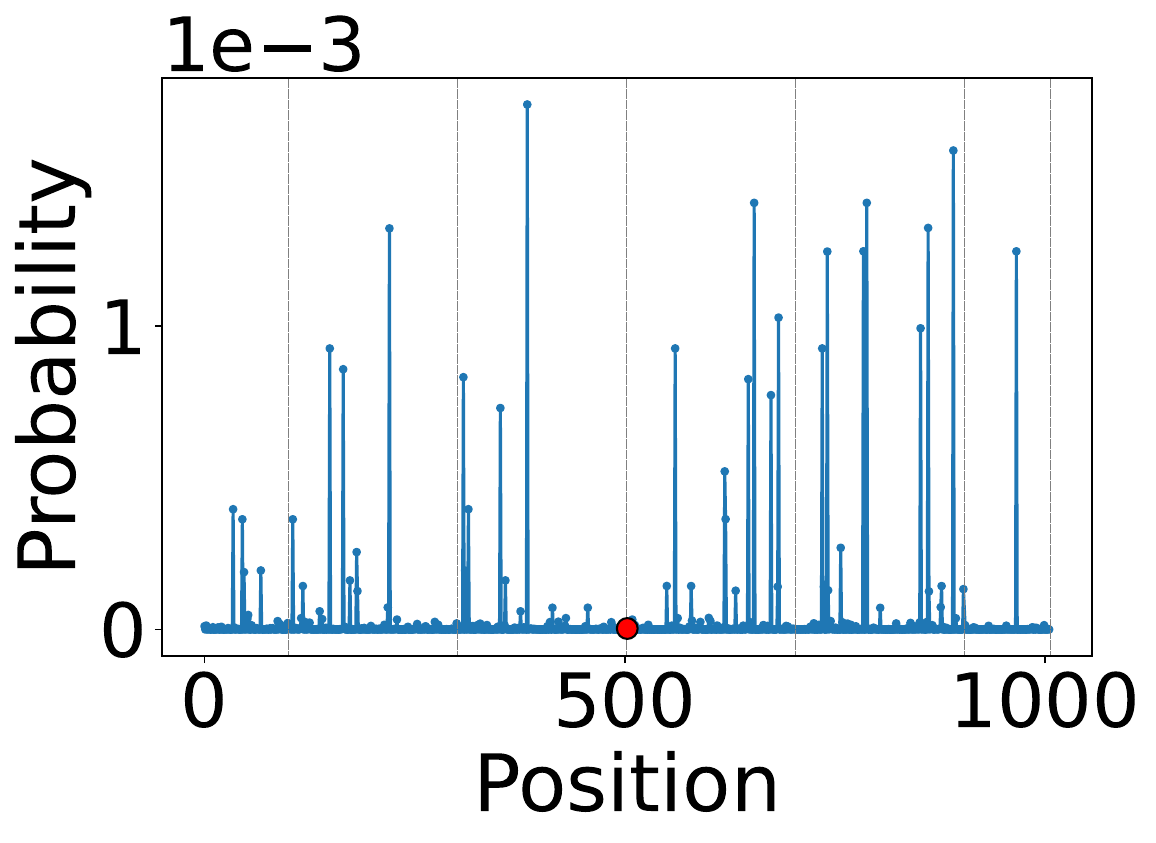} \\
& \FigGridVSubLabel{P4} &
\FigGridCell{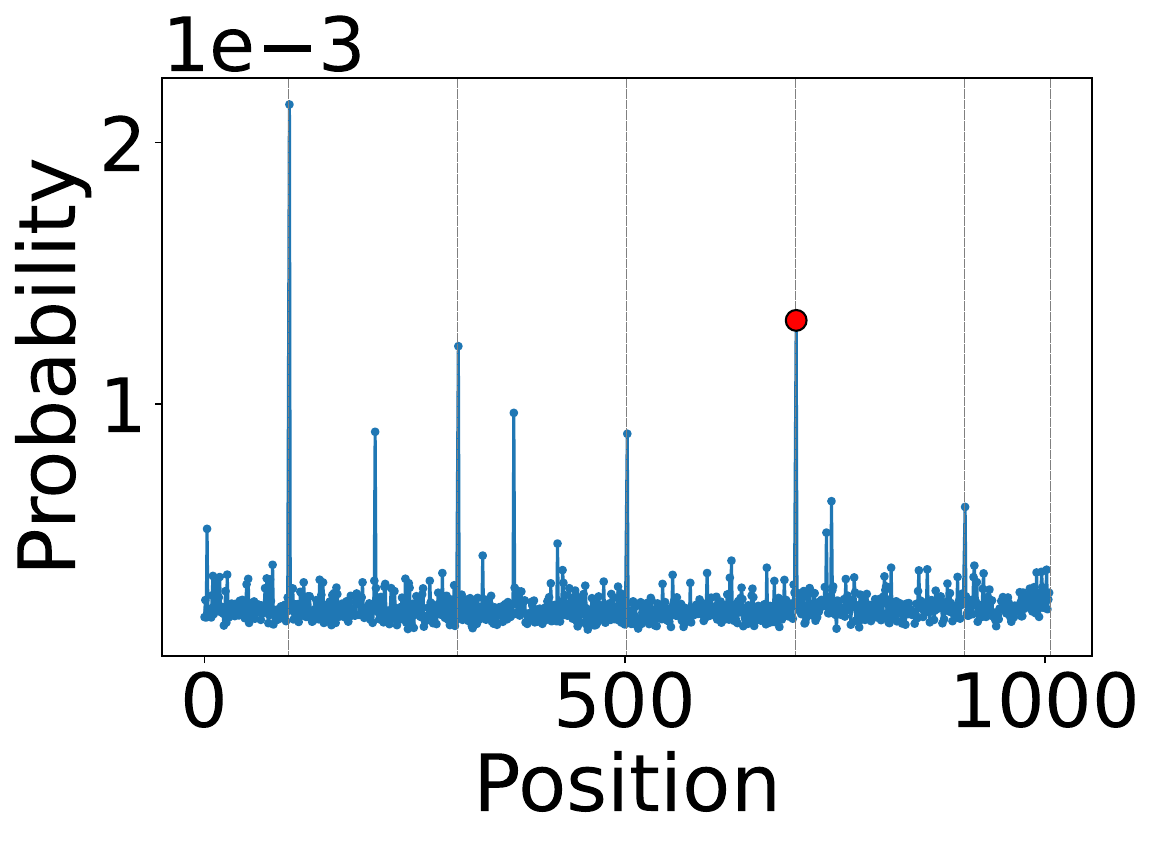} &
\FigGridCell{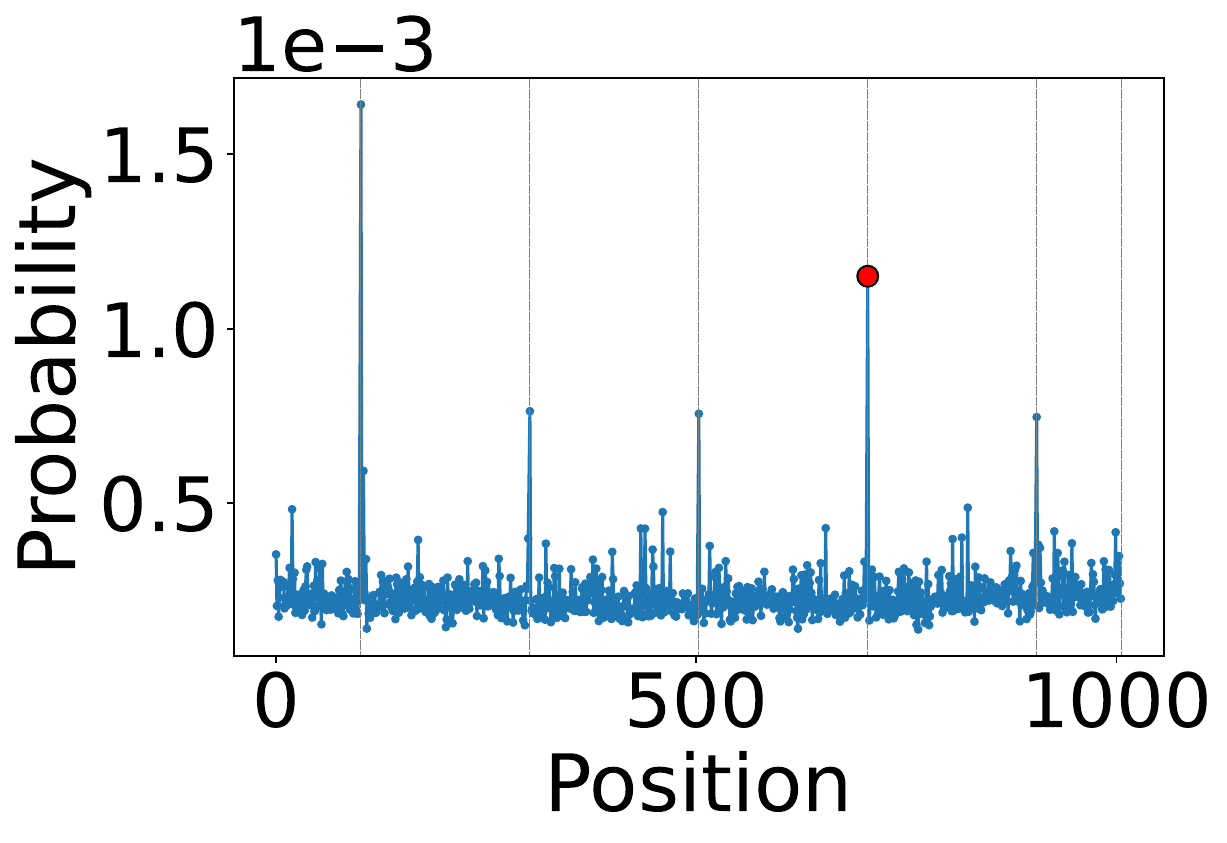} &
\FigGridCell{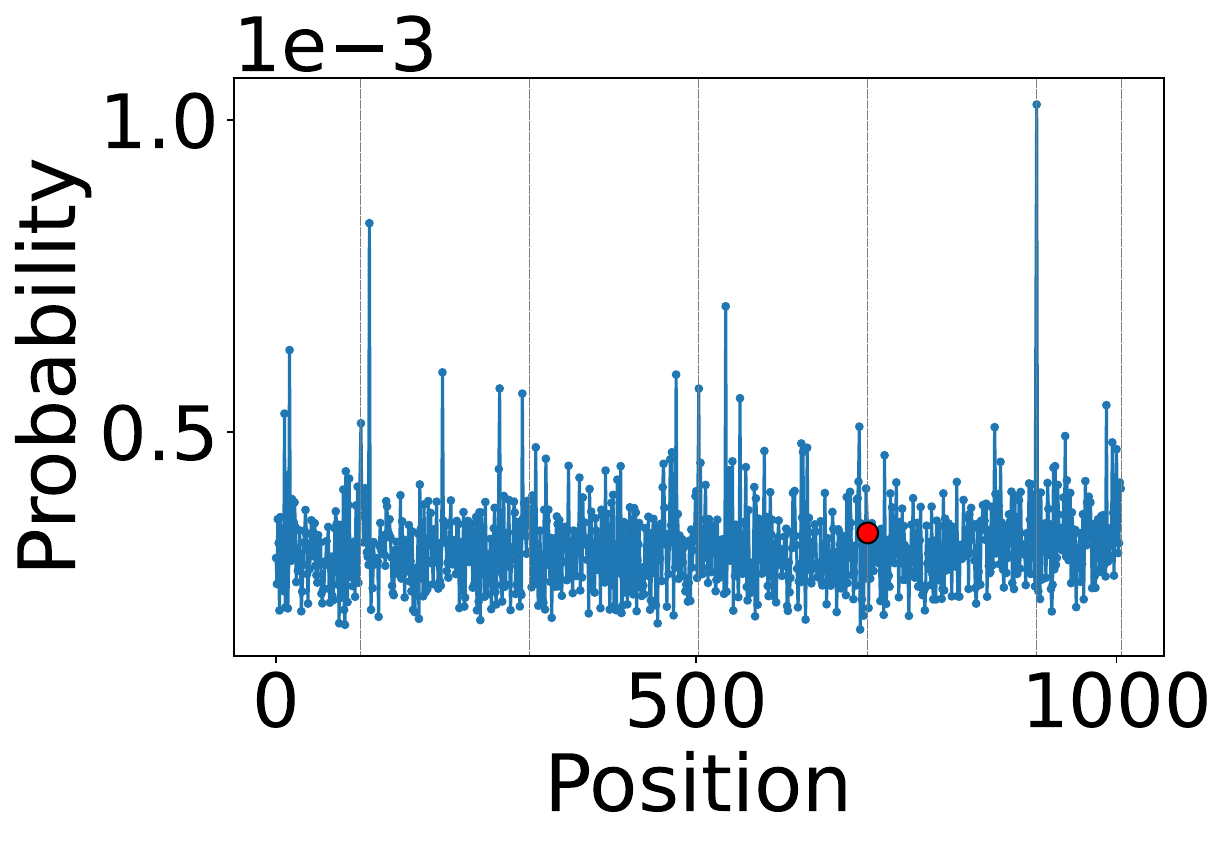} &
\FigGridCell{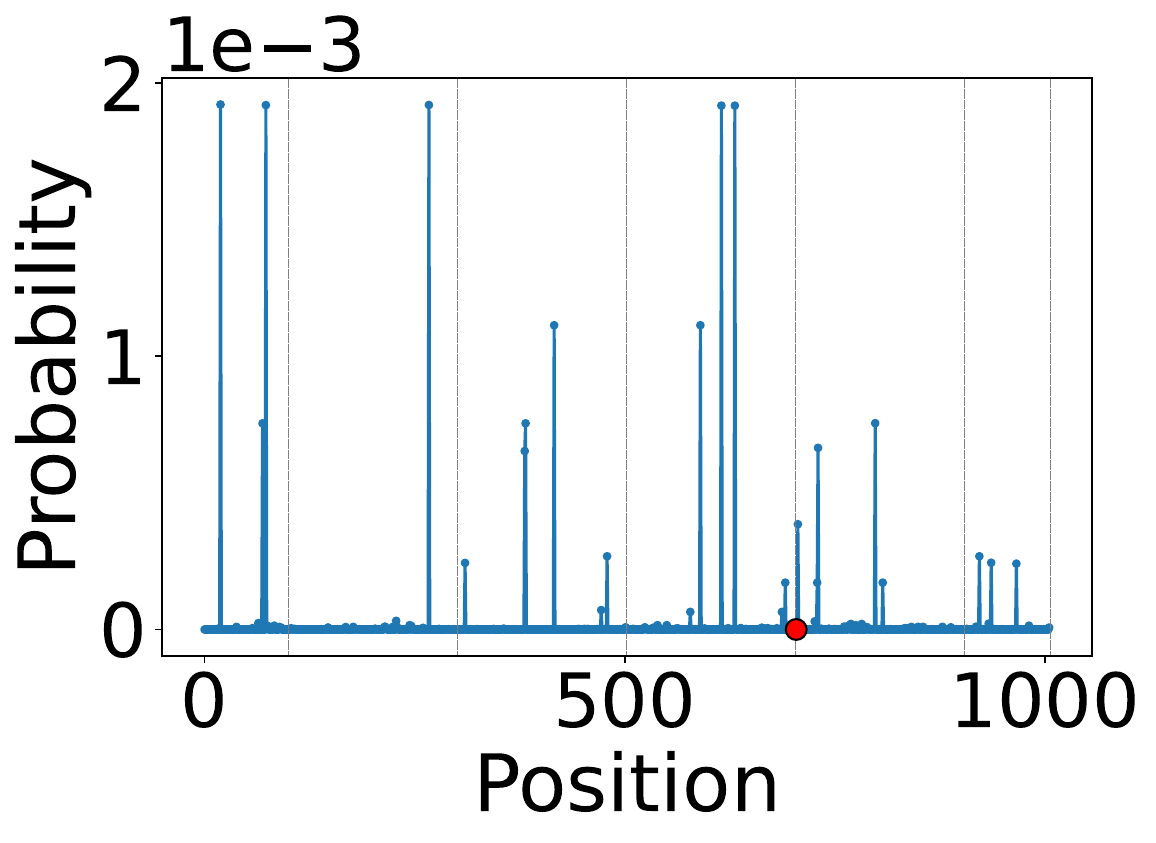} \\
& \FigGridVSubLabel{P5} &
\FigGridCell{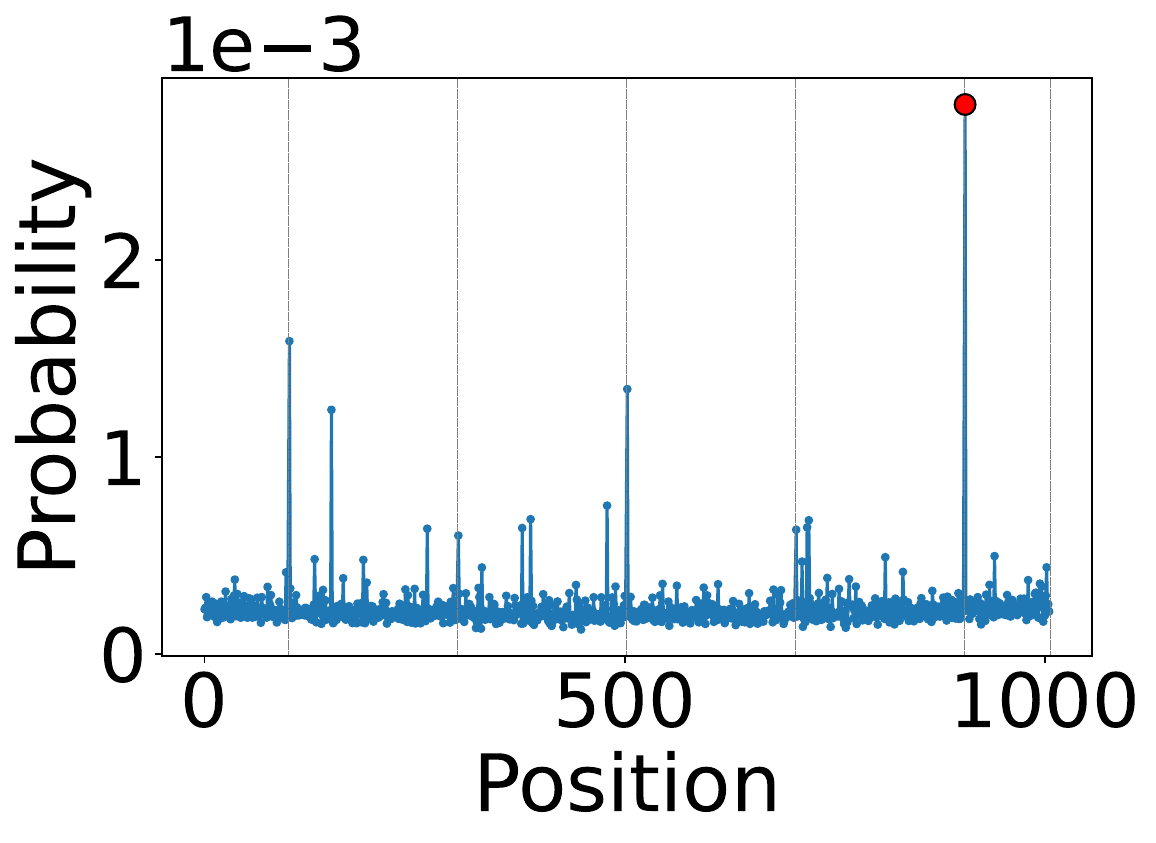} &
\FigGridCell{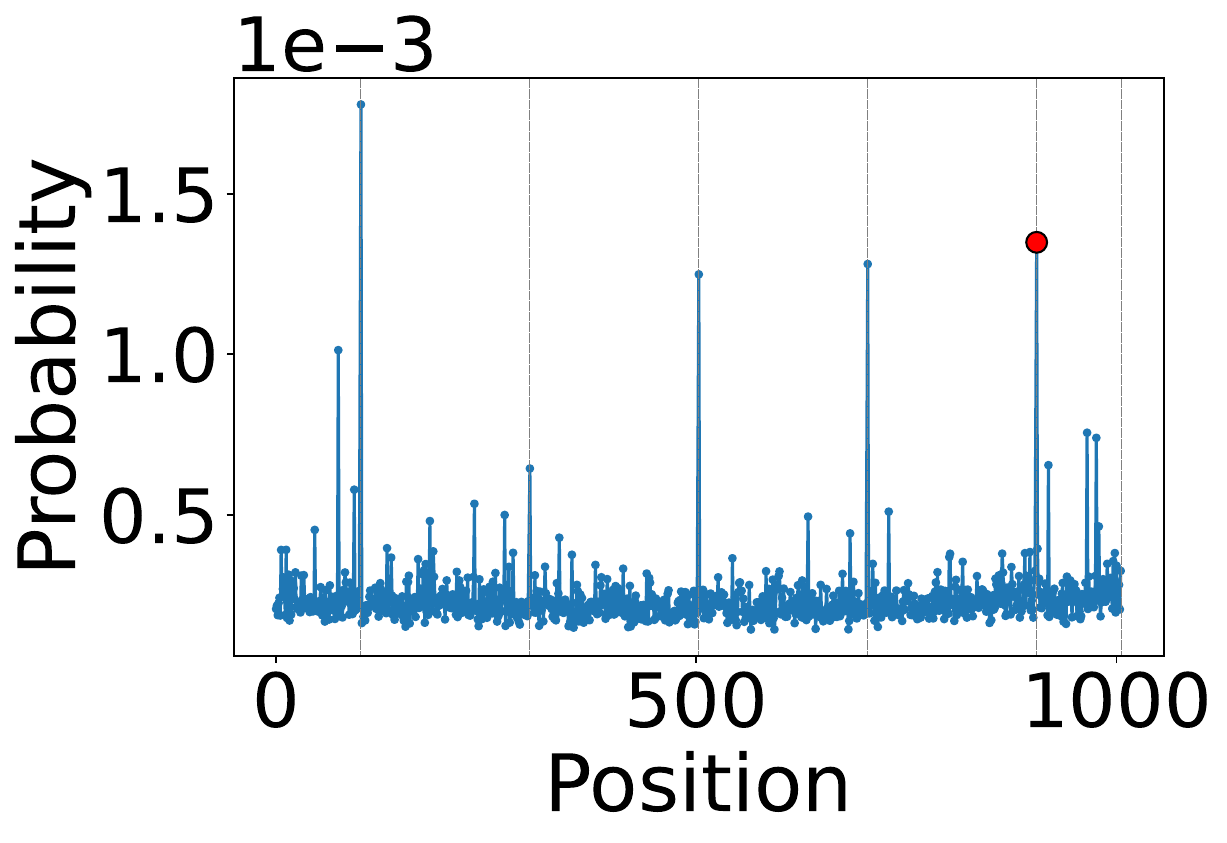} &
\FigGridCell{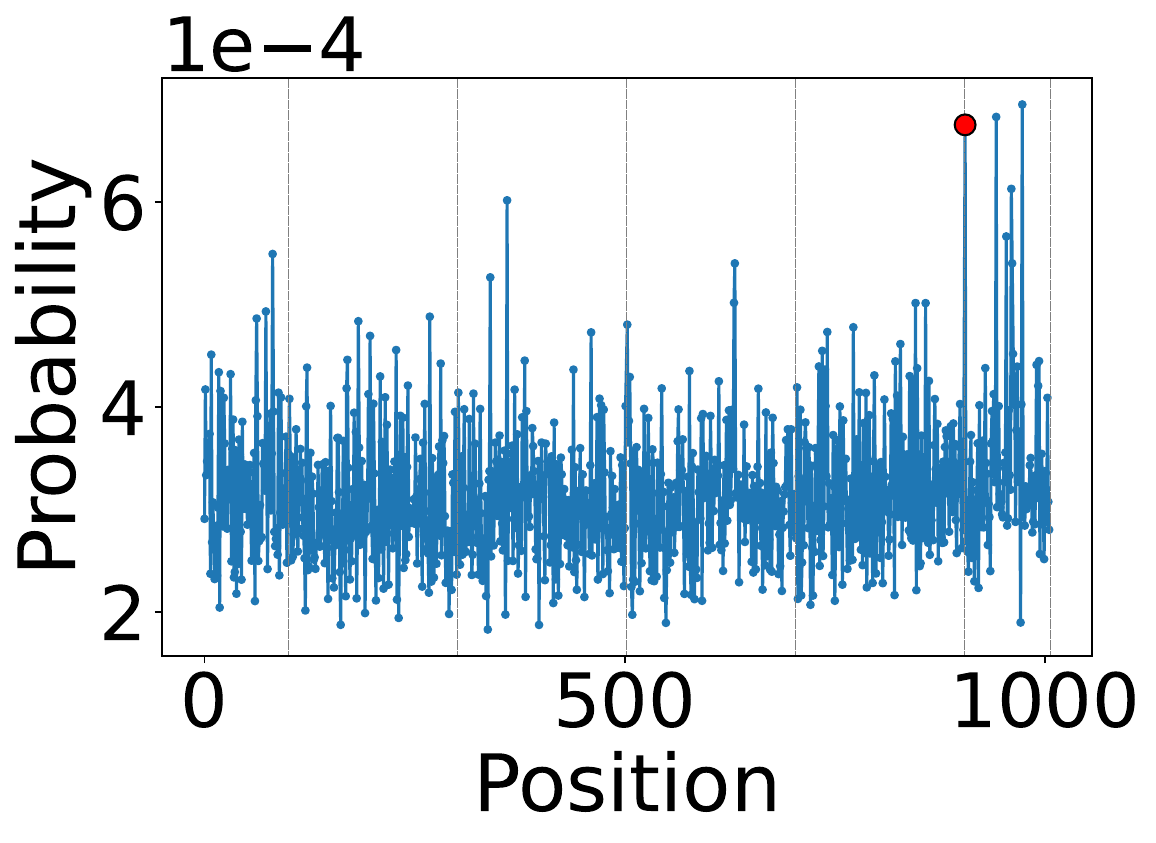} &
\FigGridCell{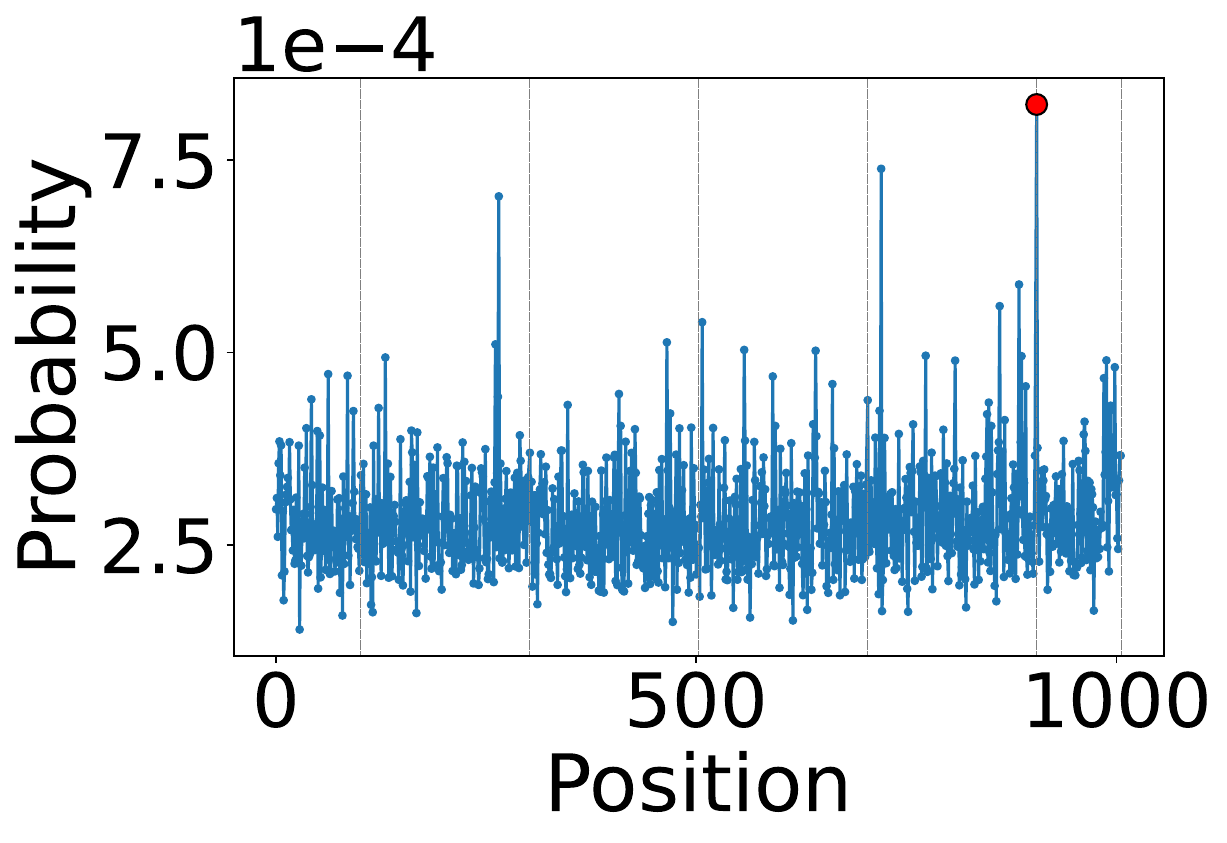} \\
\addlinespace[1pt]
\midrule
\addlinespace[1pt]
\multirow{5}{*}{\FigGridVLabelSpanFive{Random}} &
\FigGridVSubLabel{P1} &
\FigGridCell{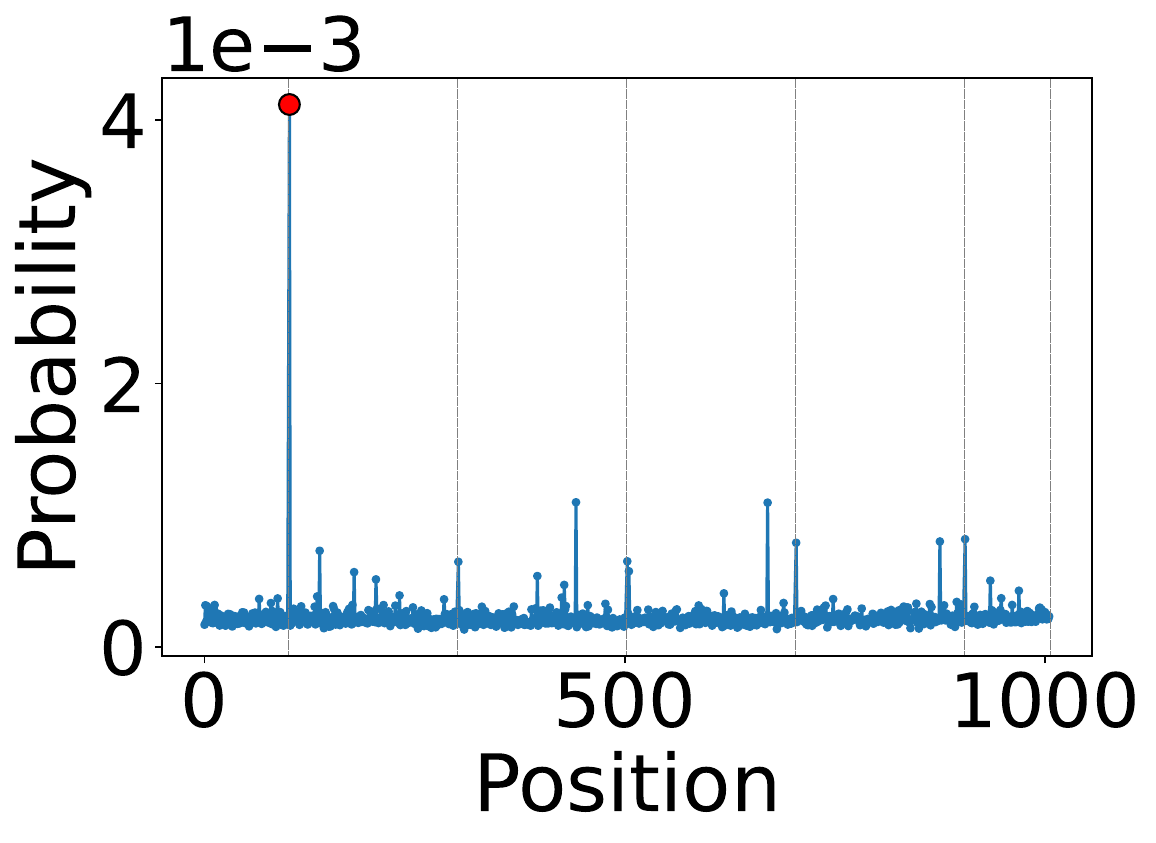} &
\FigGridCell{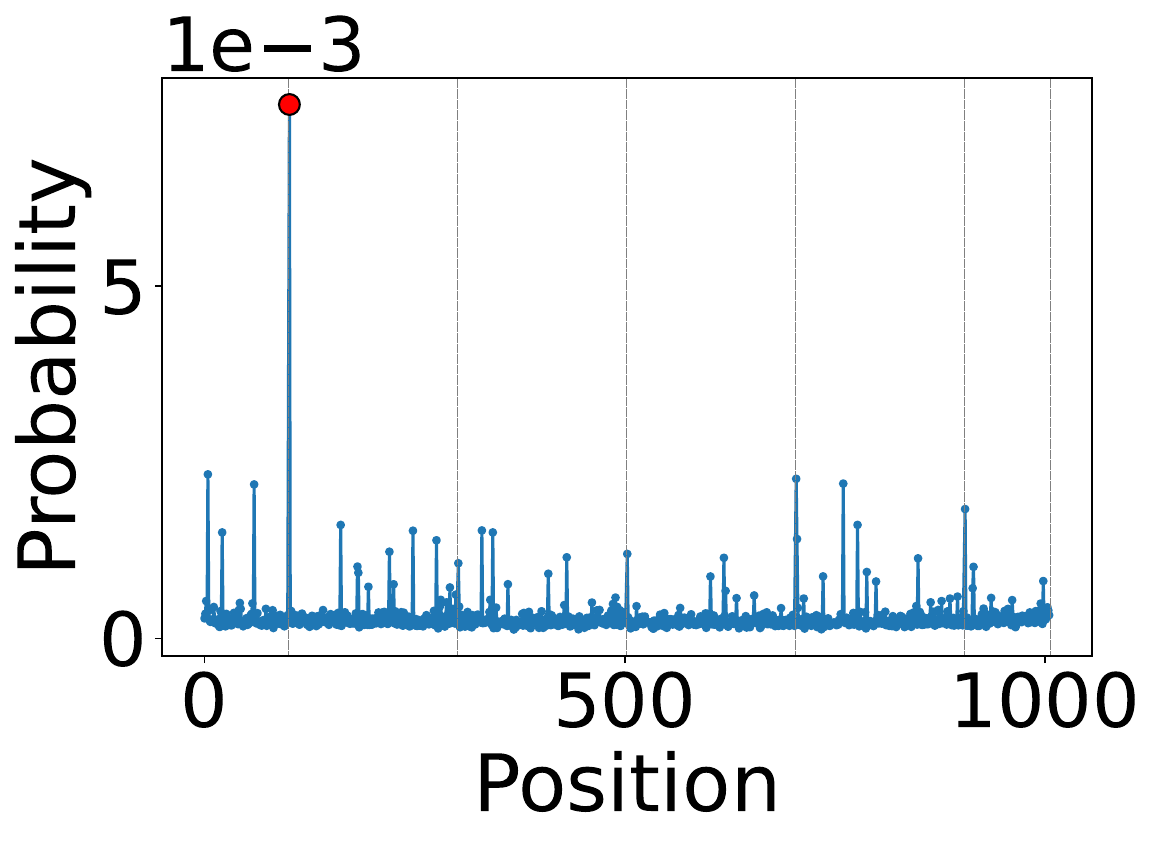} &
\FigGridCell{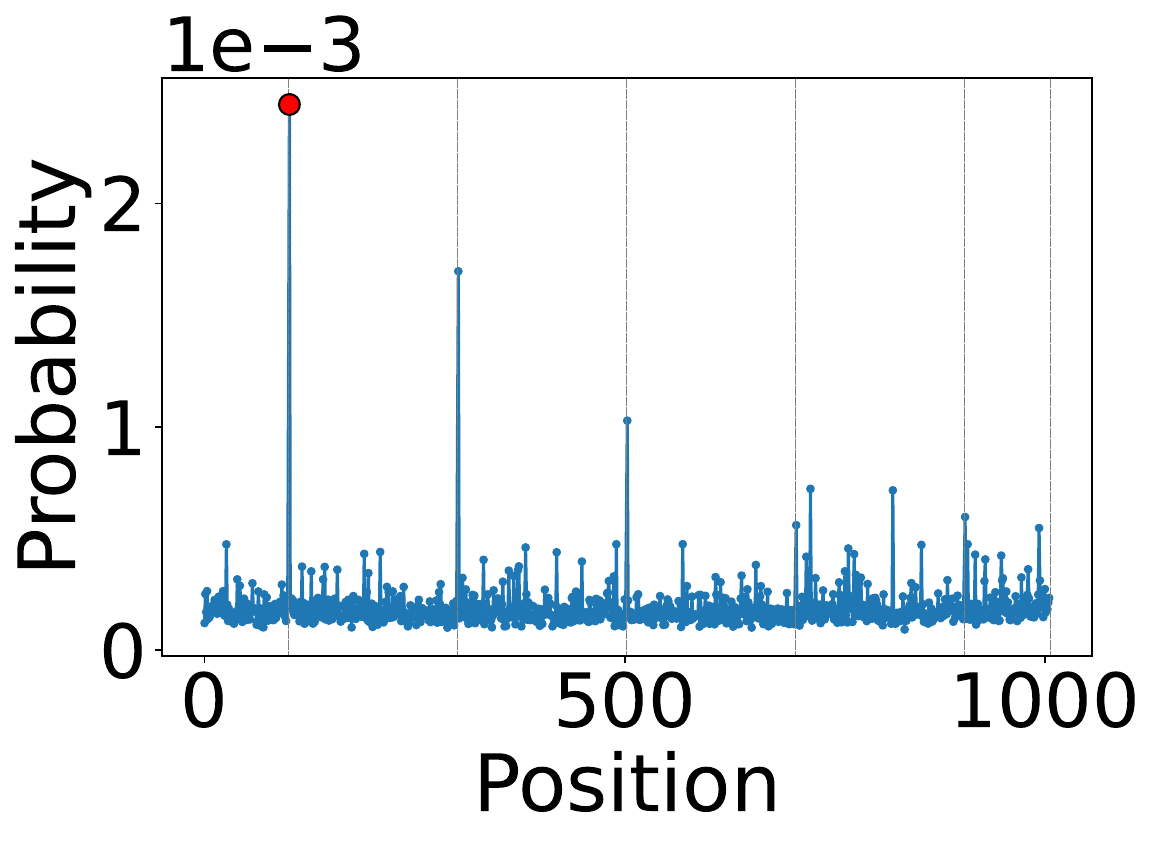} &
\FigGridCell{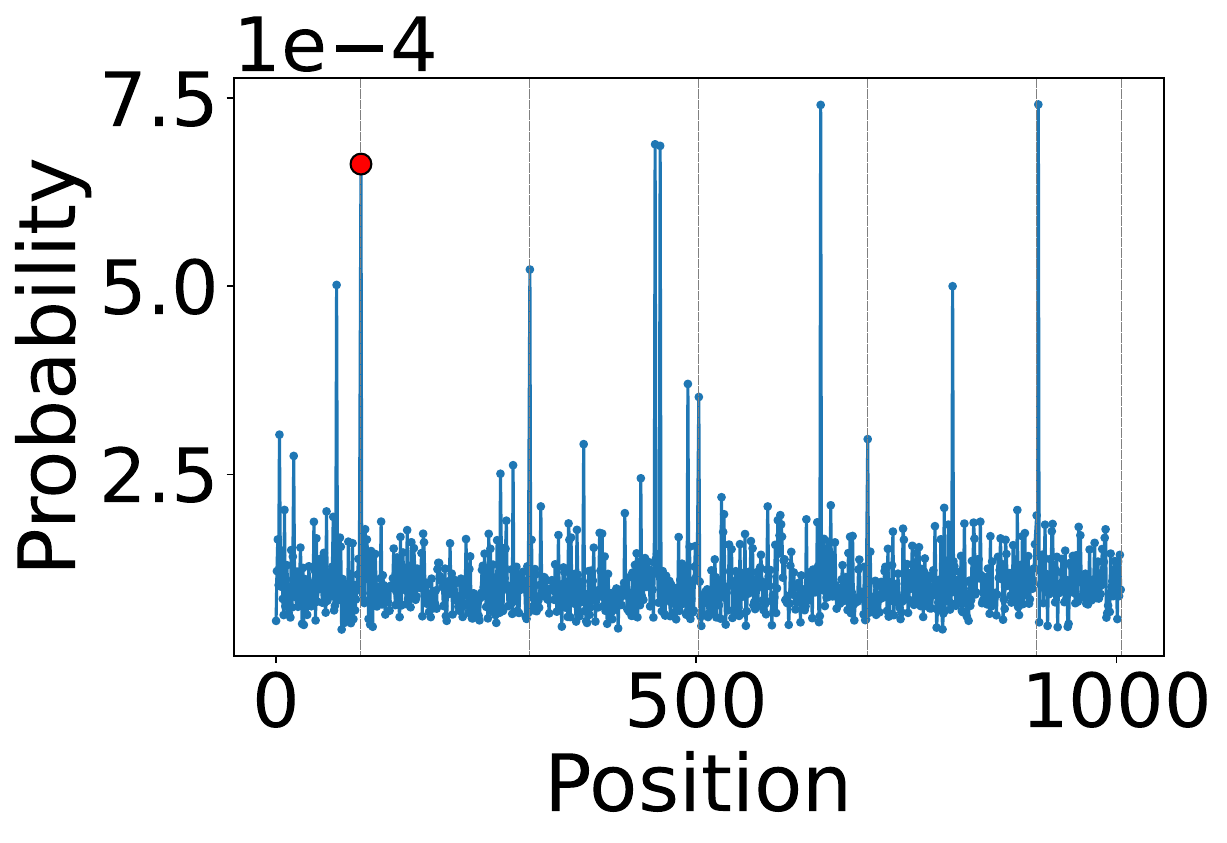} \\
& \FigGridVSubLabel{P2} &
\FigGridCell{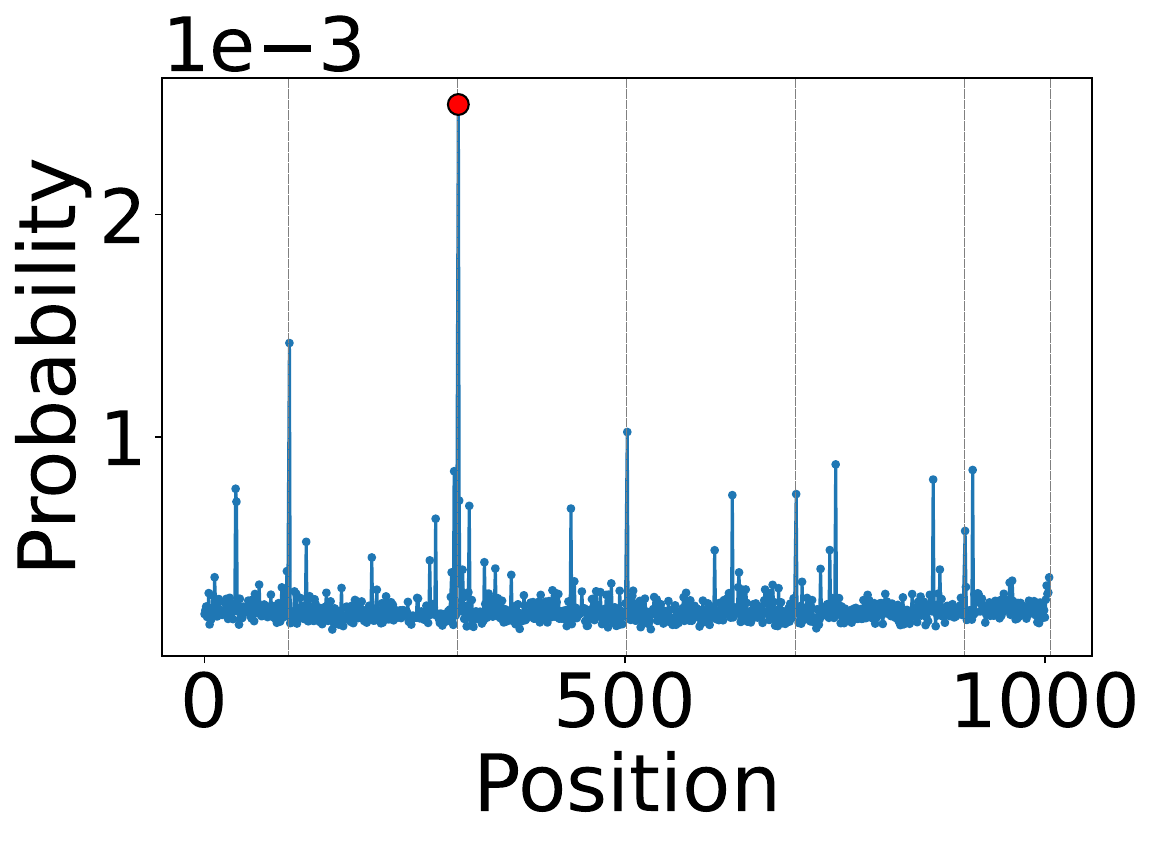} &
\FigGridCell{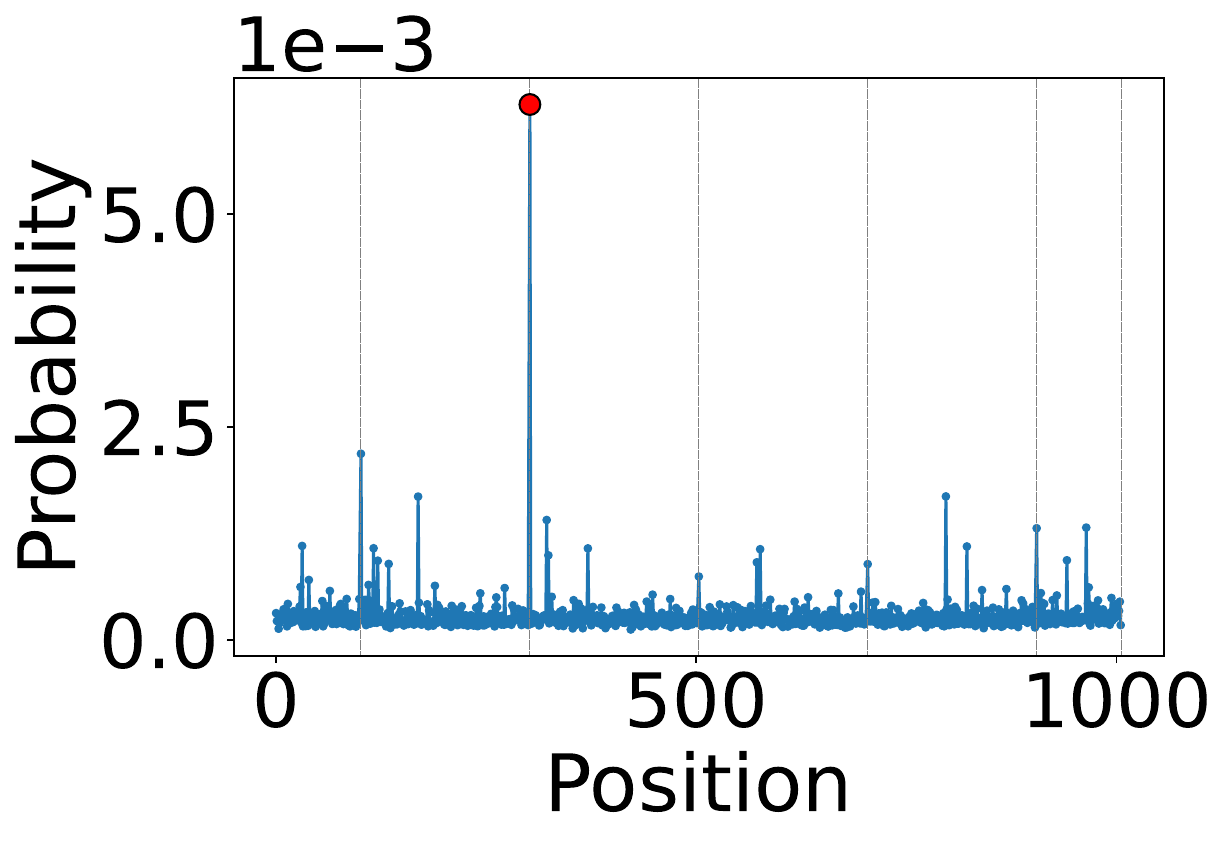} &
\FigGridCell{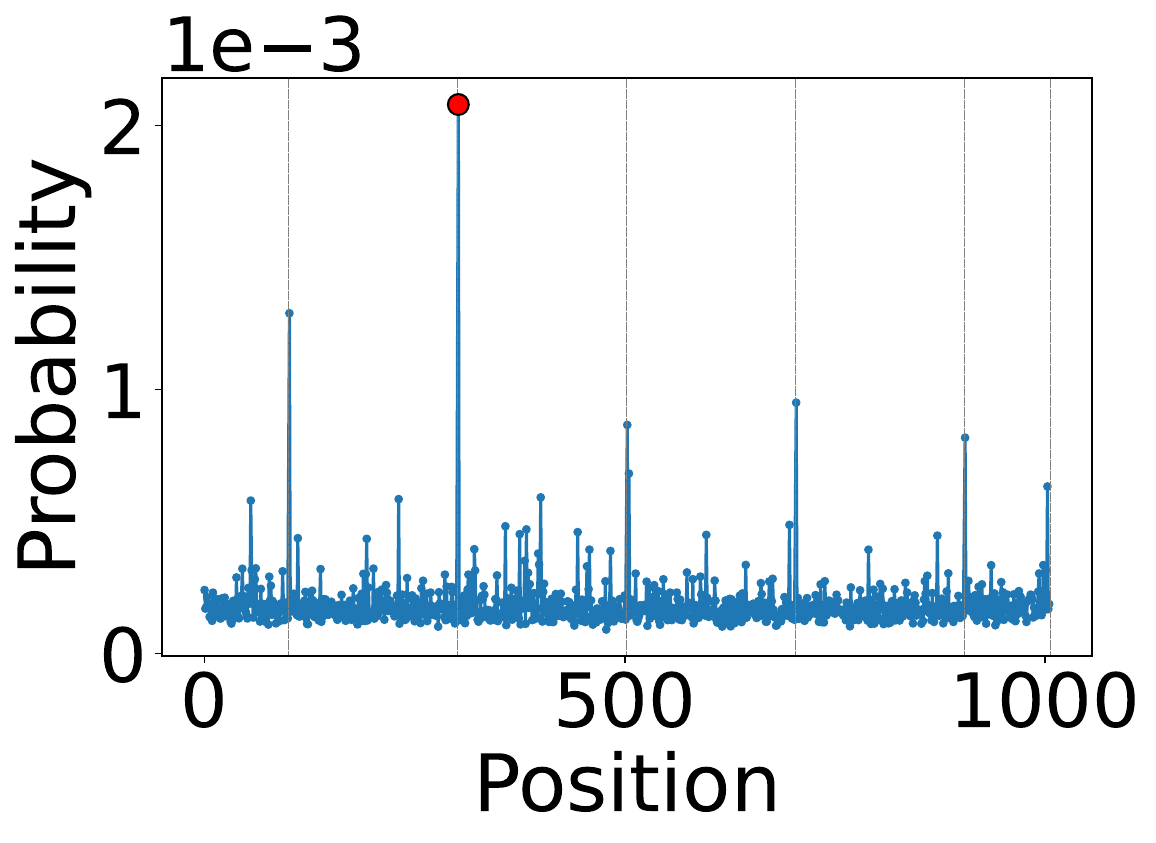} &
\FigGridCell{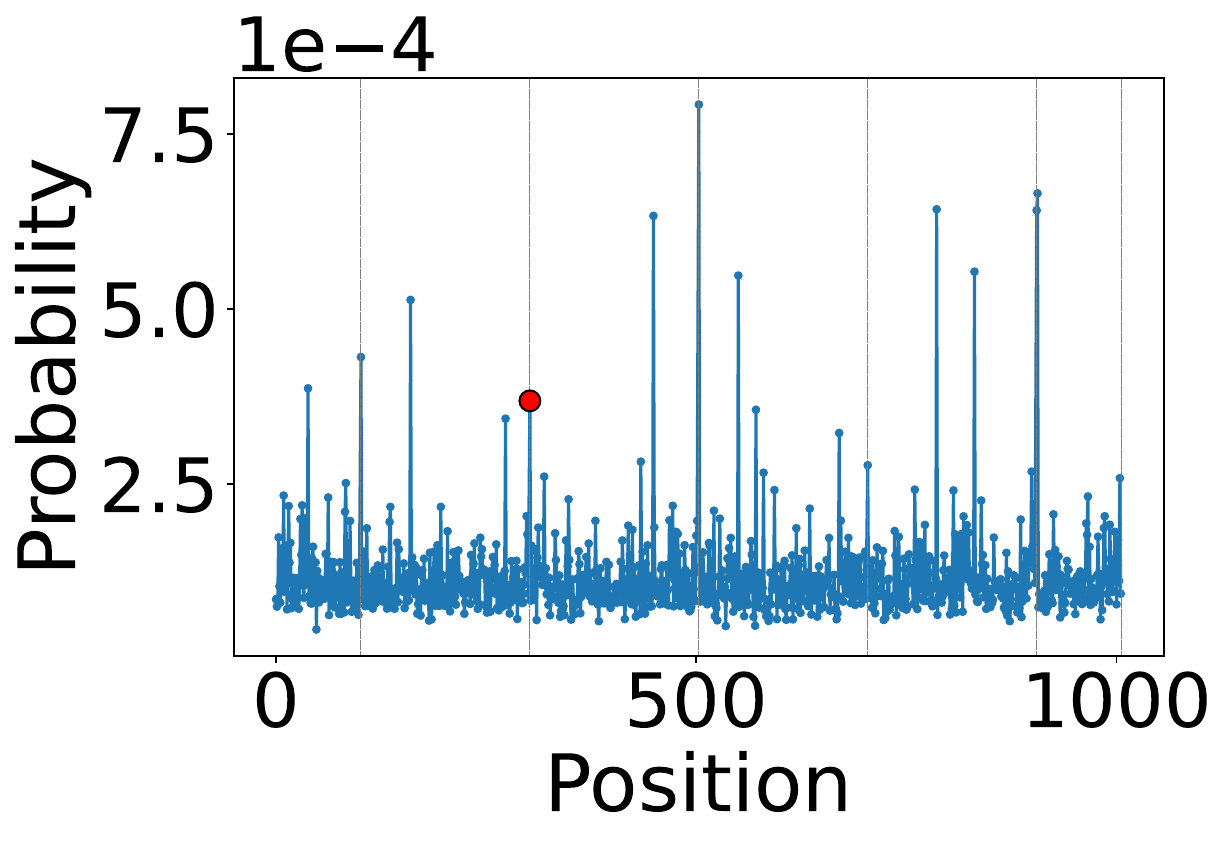} \\
& \FigGridVSubLabel{P3} &
\FigGridCell{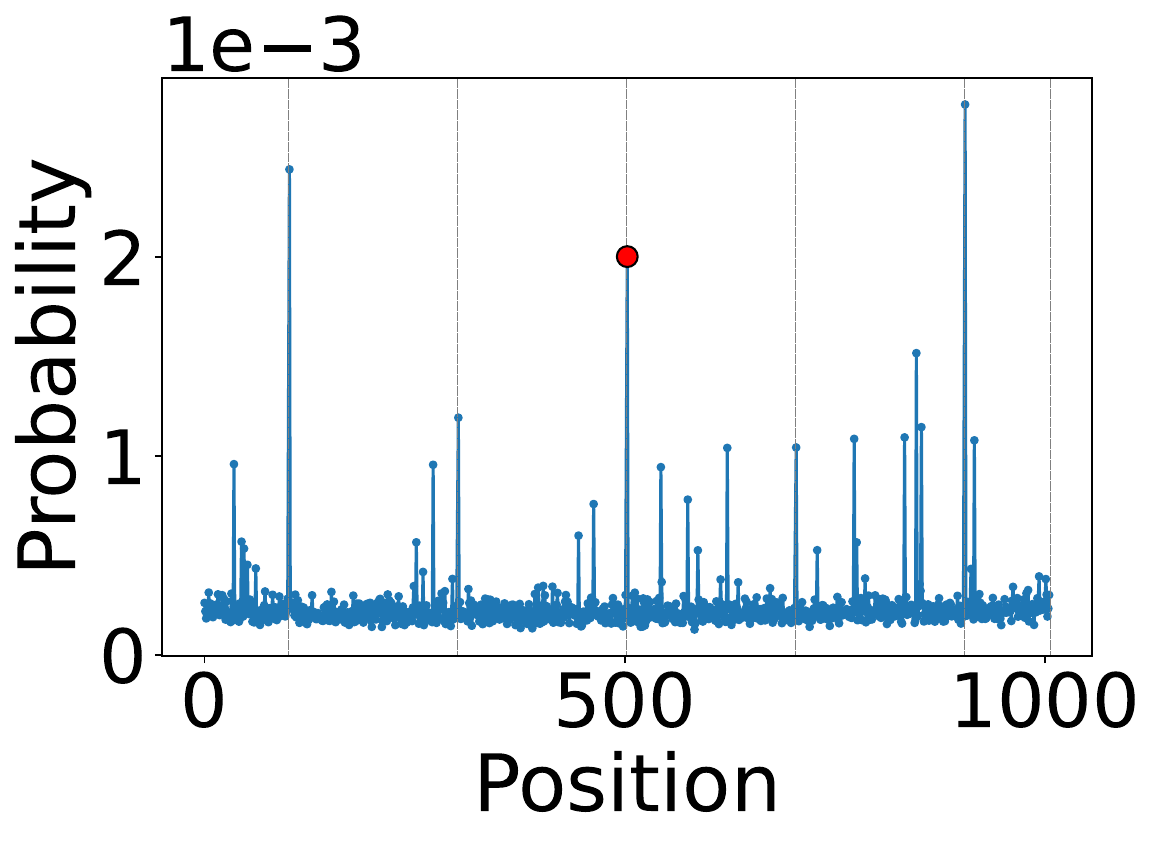} &
\FigGridCell{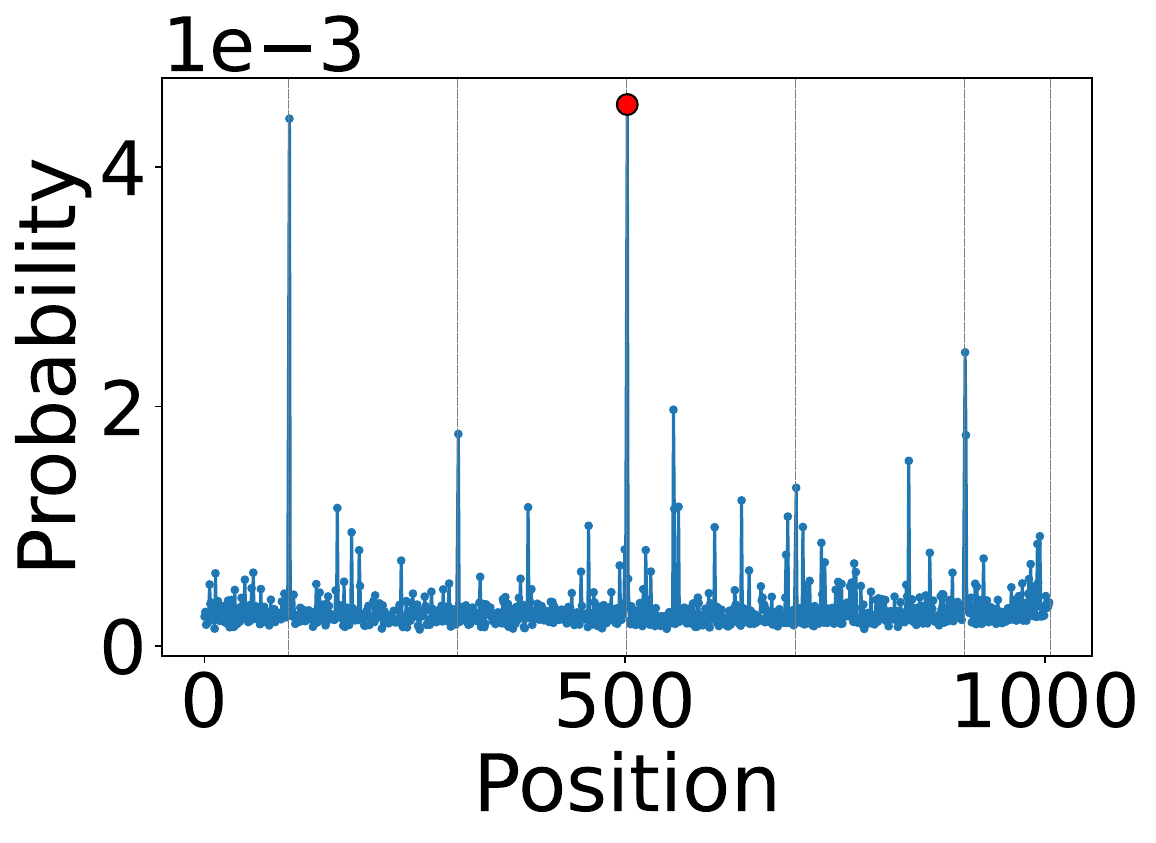} &
\FigGridCell{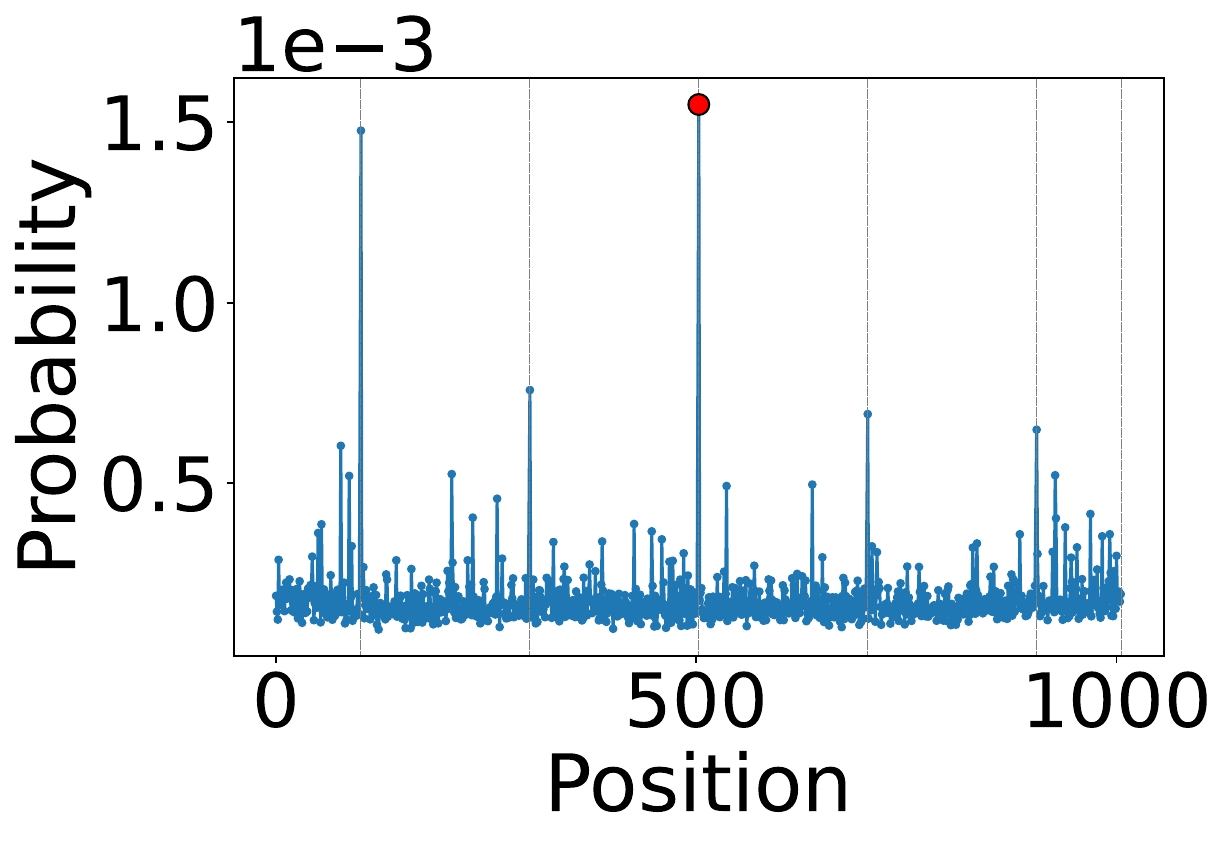} &
\FigGridCell{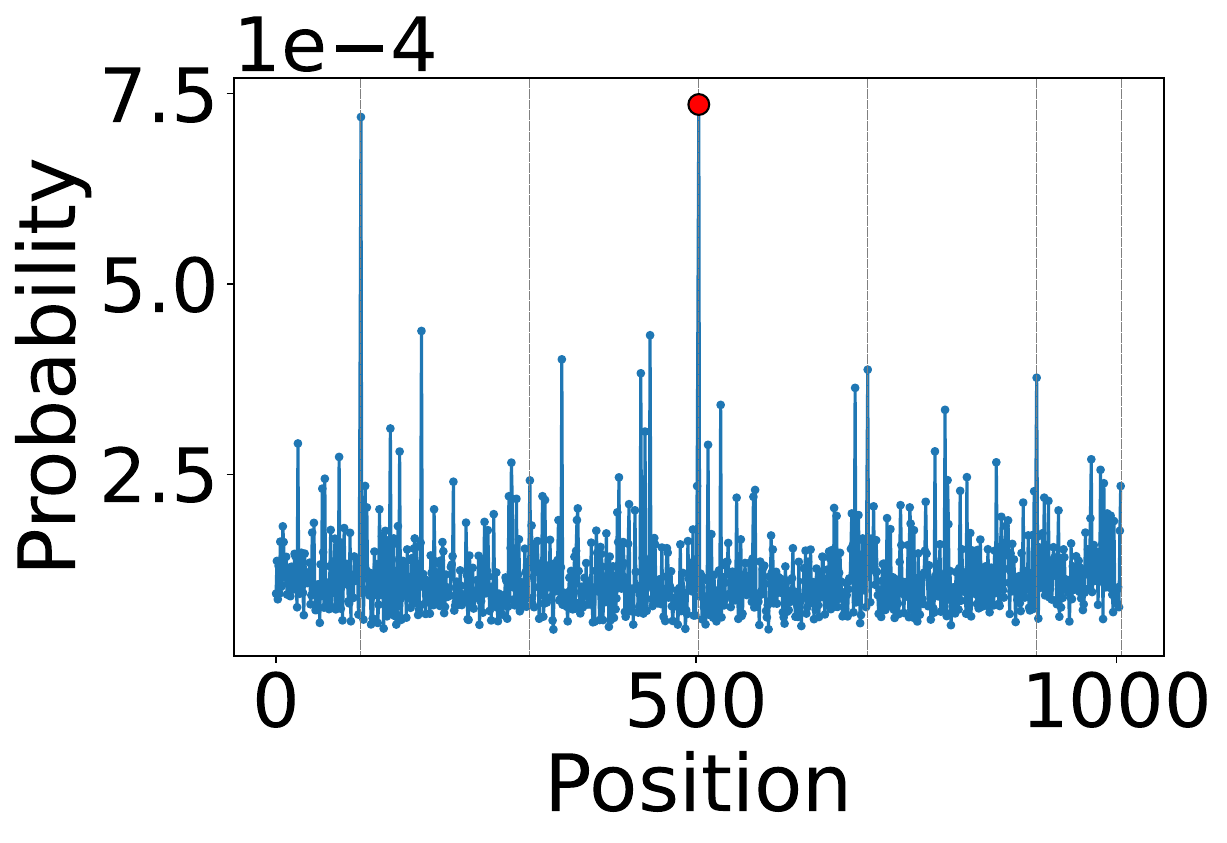} \\
& \FigGridVSubLabel{P4} &
\FigGridCell{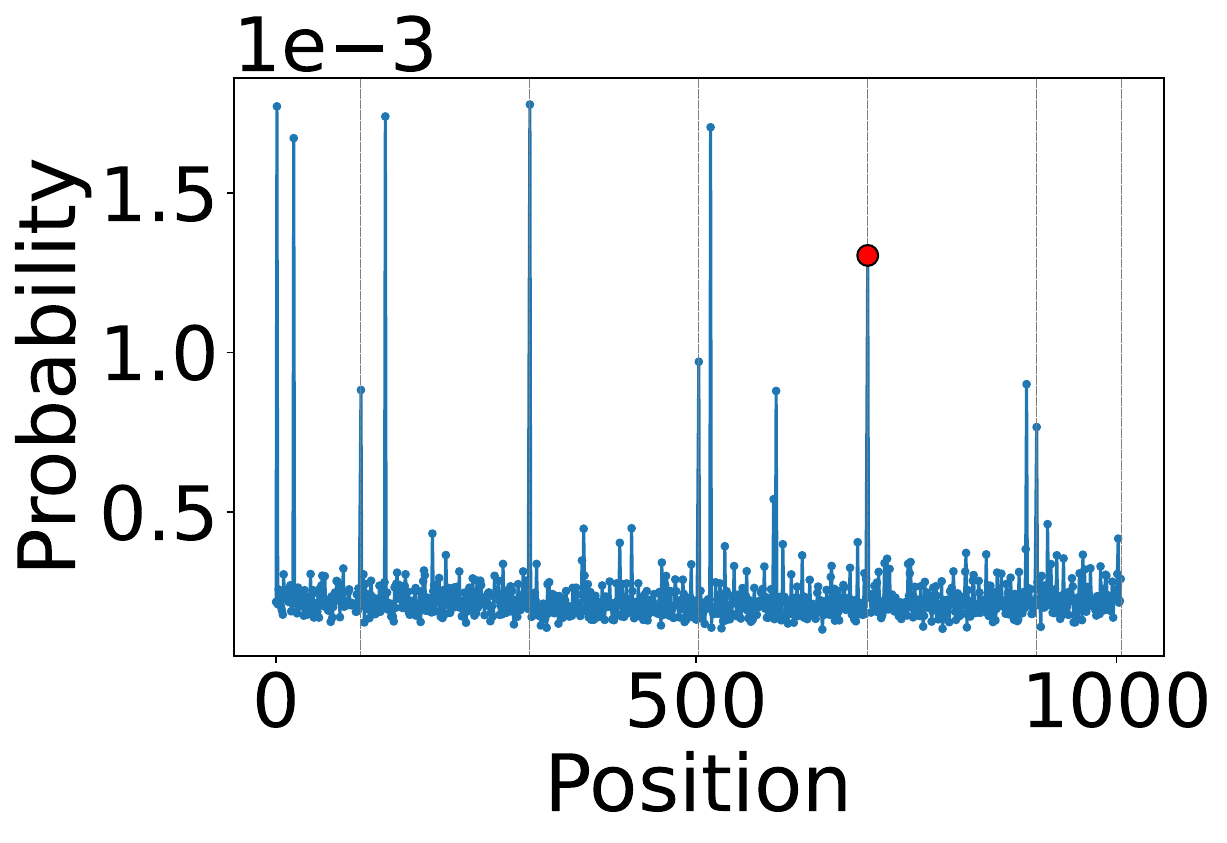} &
\FigGridCell{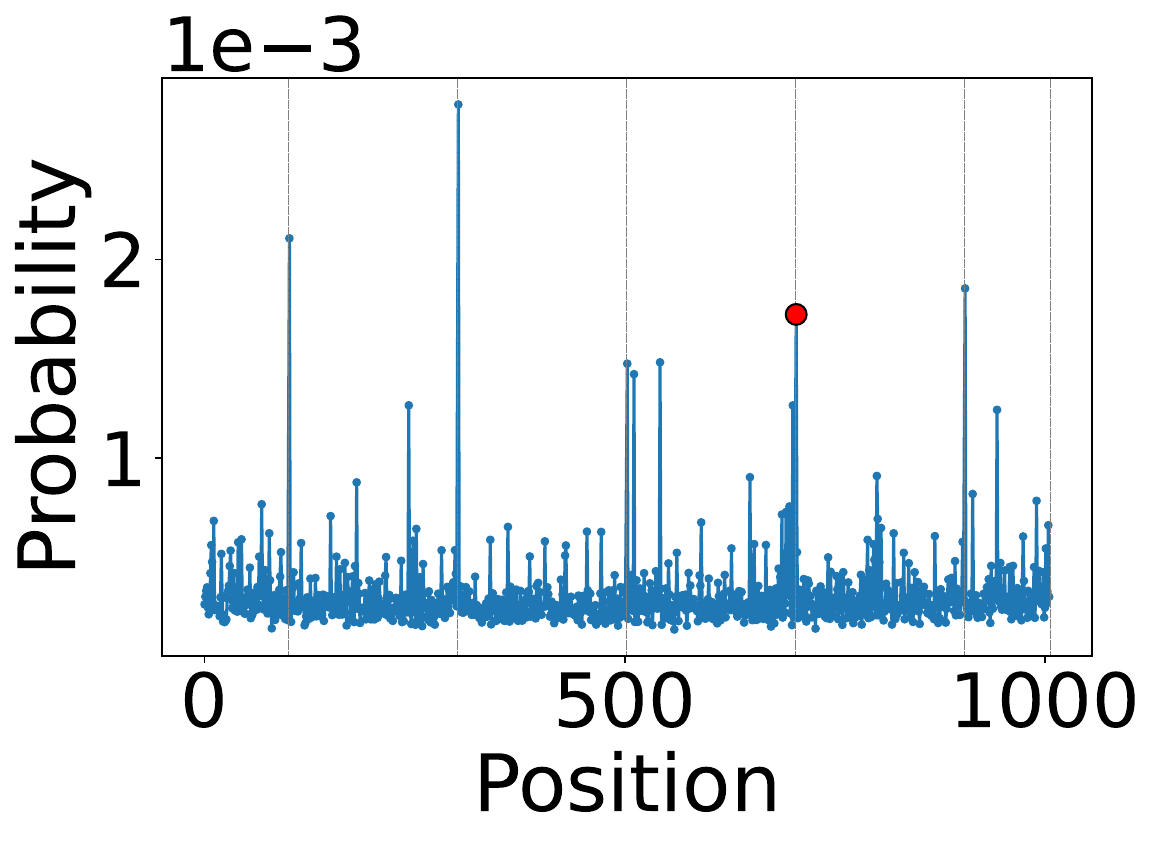} &
\FigGridCell{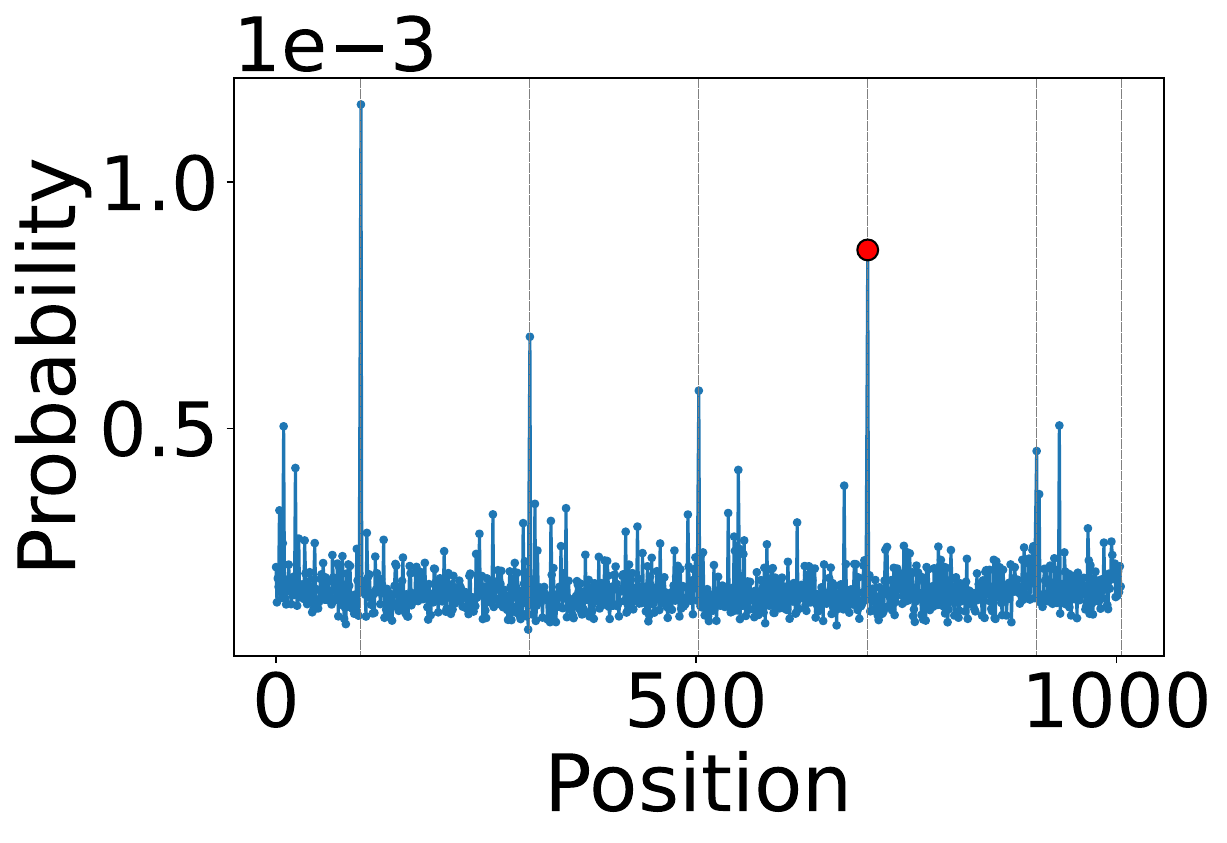} &
\FigGridCell{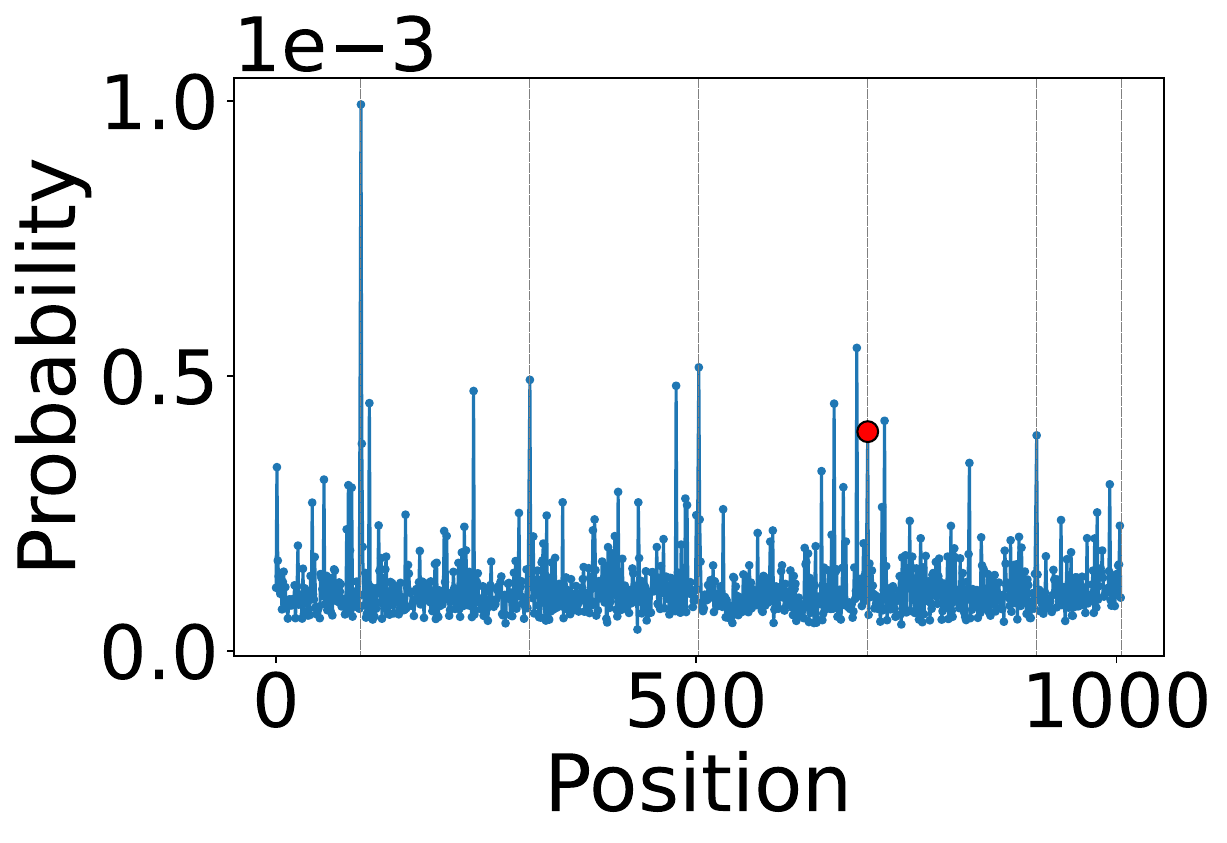} \\
& \FigGridVSubLabel{P5} &
\FigGridCell{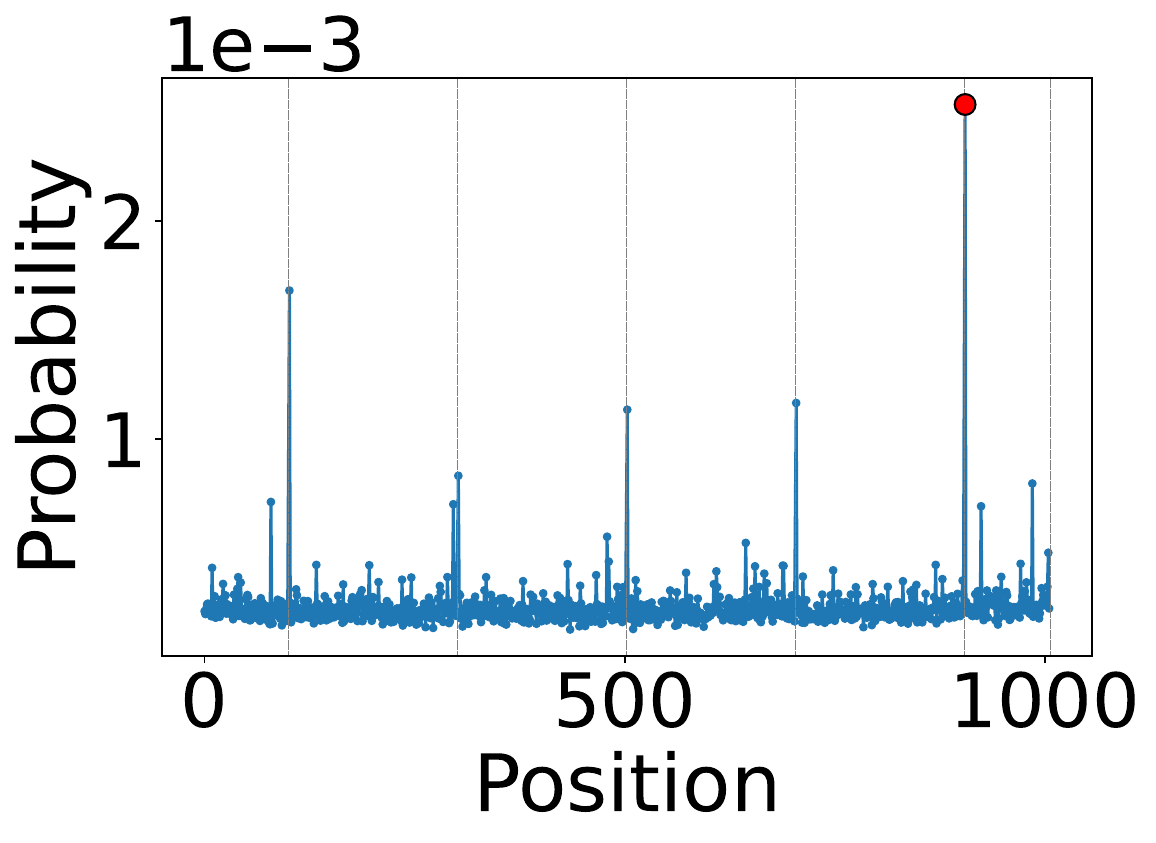} &
\FigGridCell{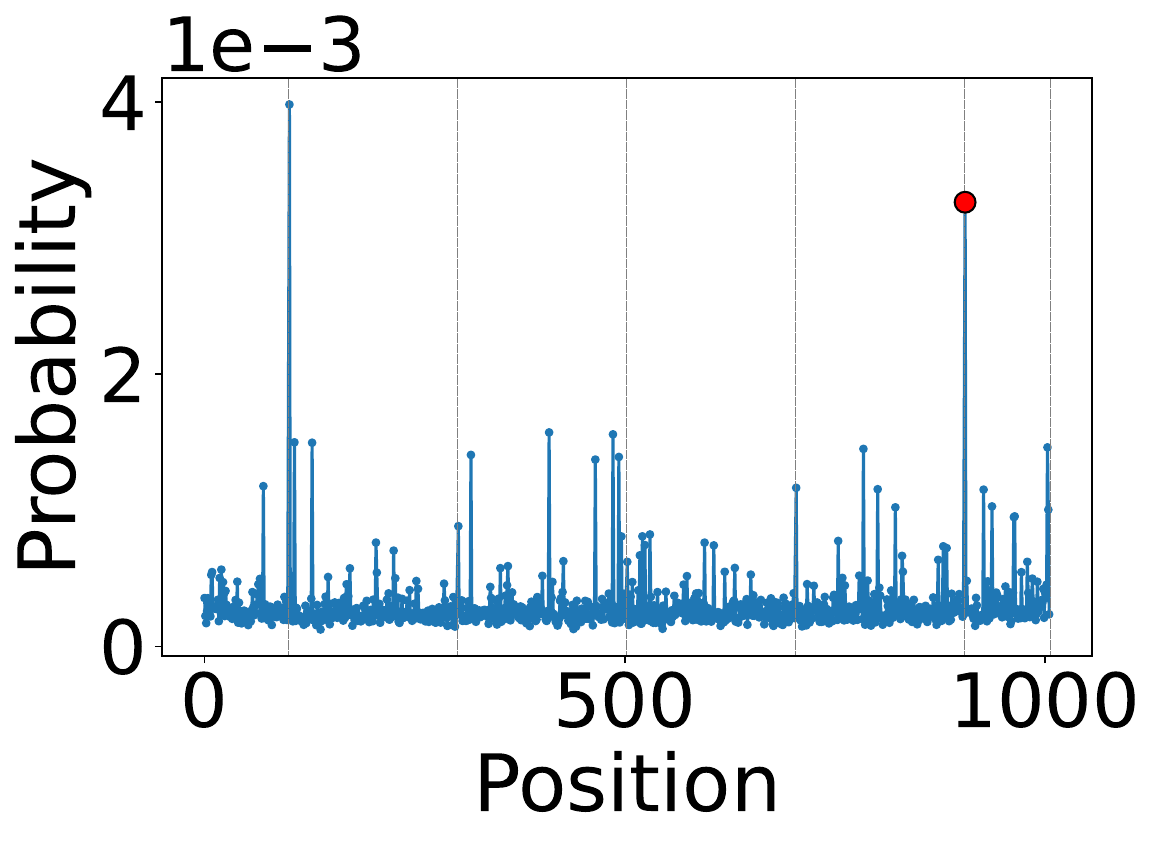} &
\FigGridCell{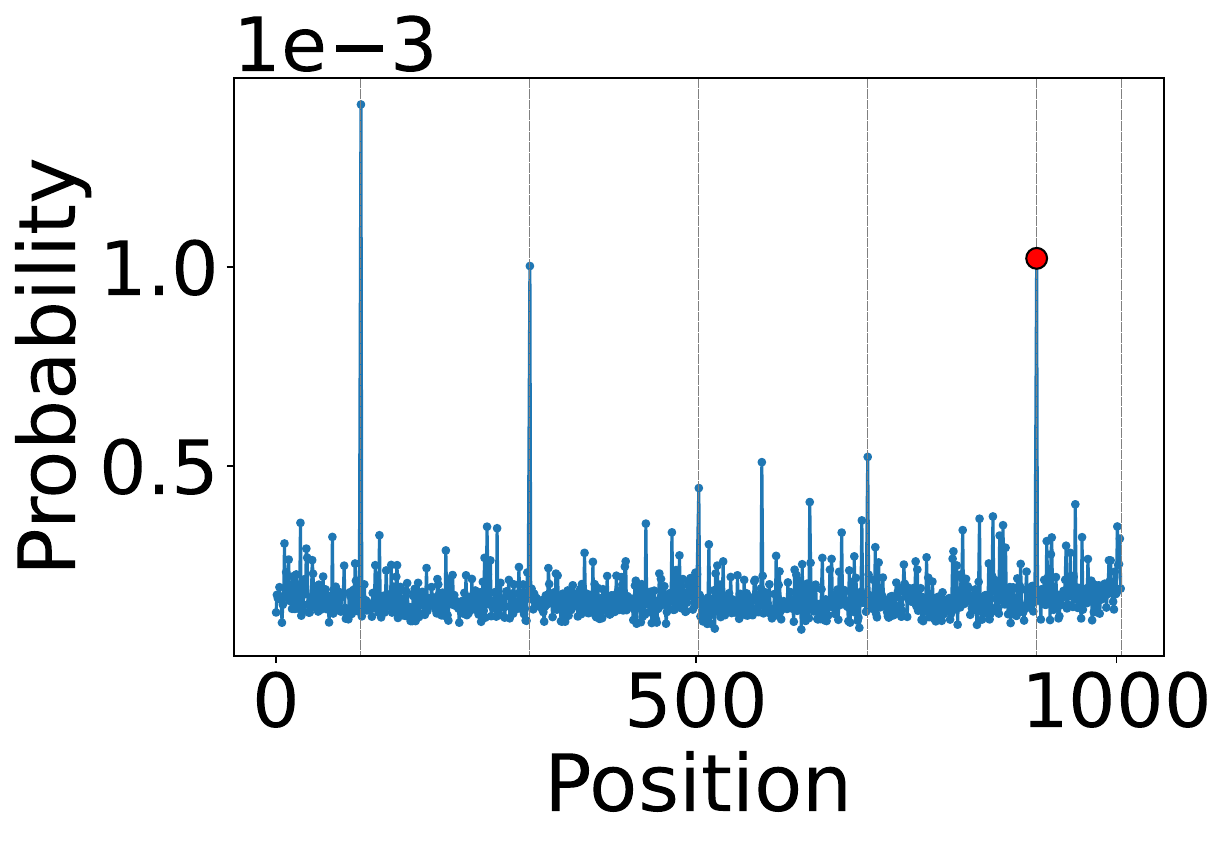} &
\FigGridCell{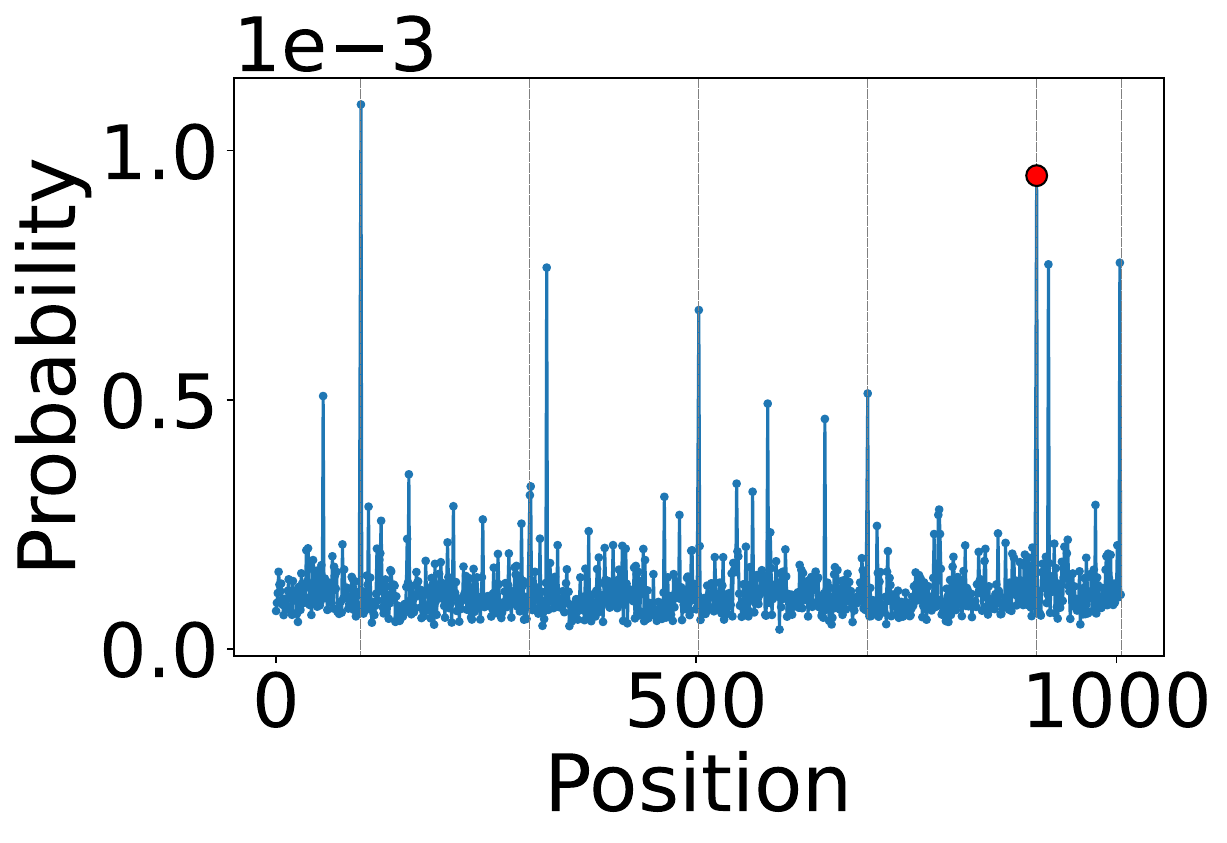} \\
\bottomrule
\end{tabular}
\endgroup
\caption{Mamba ablation effect (Exp. 2). Episodic retrieval probability after ablating Induction-related or Random channels. Columns show the percentage of ablated channels.}
\label{fig:exp2_mamba_ablation}
\end{figure*}

\begin{figure*}[h!]
\centering
\begingroup
\setlength{\FigGridLabelW}{1.05cm}
\setlength{\FigGridSubLabelW}{0.55cm}
\setlength{\FigGridCellH}{1.85cm} 
\setlength{\FigGridLabelSep}{1pt}
\setlength{\FigGridSubLabelSep}{1pt}
\setlength{\FigGridColSep}{0.5pt}

\FigGridSetupTwo{4}
\renewcommand{\arraystretch}{1.0}

\begin{tabular}{@{}>{\centering\arraybackslash}m{\FigGridLabelW}@{\hspace{\FigGridLabelSep}}
                >{\centering\arraybackslash}m{\FigGridSubLabelW}@{\hspace{\FigGridSubLabelSep}}
                *{4}{>{\centering\arraybackslash}m{\FigGridPlotW}}@{}}
\toprule
& & \multicolumn{4}{c}{\FigHdr Ablations (\%)} \\
\cmidrule(lr){3-6}
& & \FigHdr 0 & \FigHdr 1 & \FigHdr 5 & \FigHdr 10 \\
\midrule
\multirow{5}{*}{\FigGridVLabelSpanFive{Induction}} &
\FigGridVSubLabel{P1} &
\FigGridCell{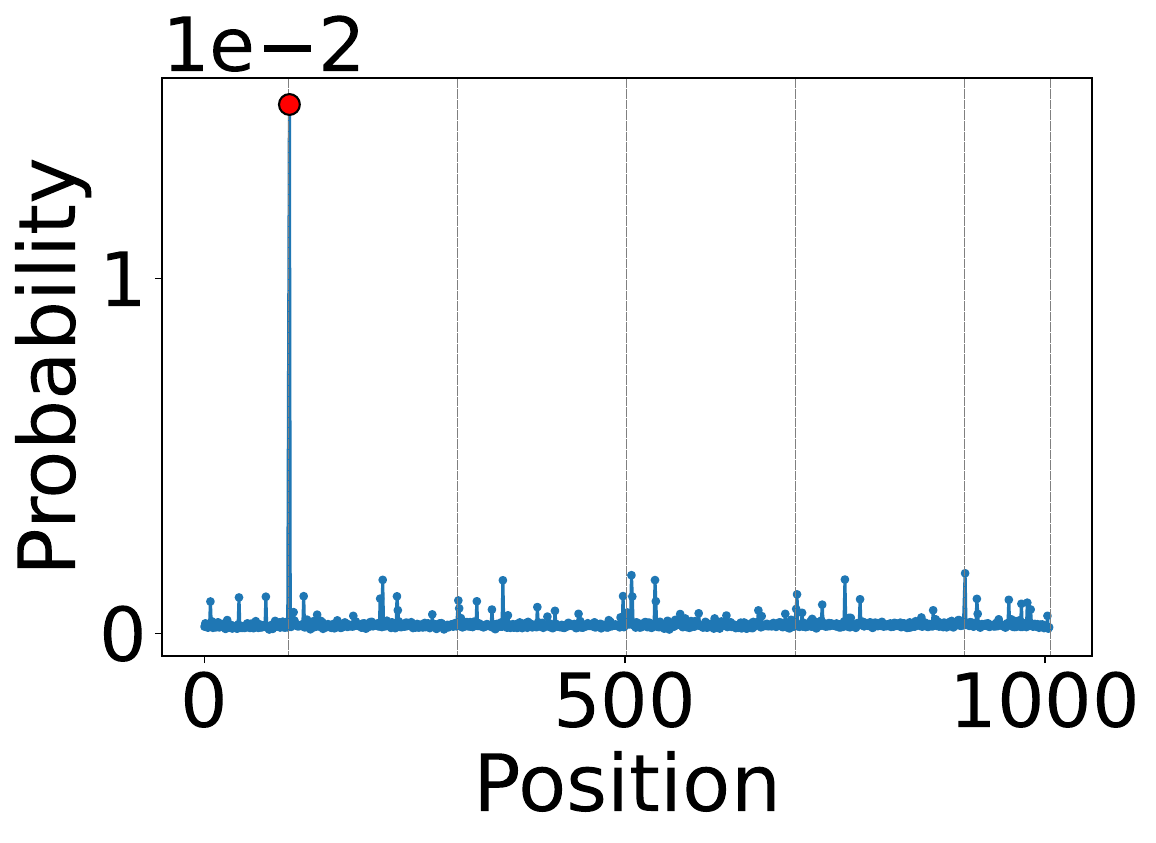} &
\FigGridCell{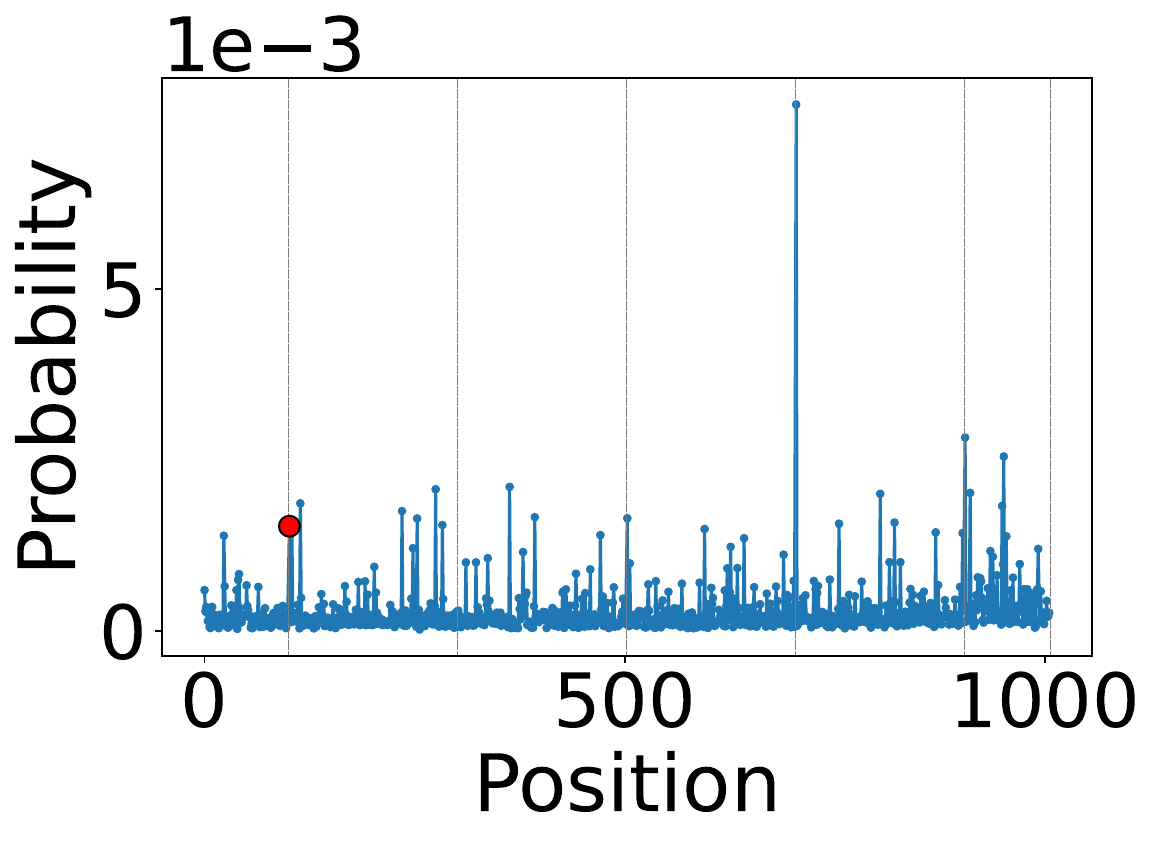} &
\FigGridCell{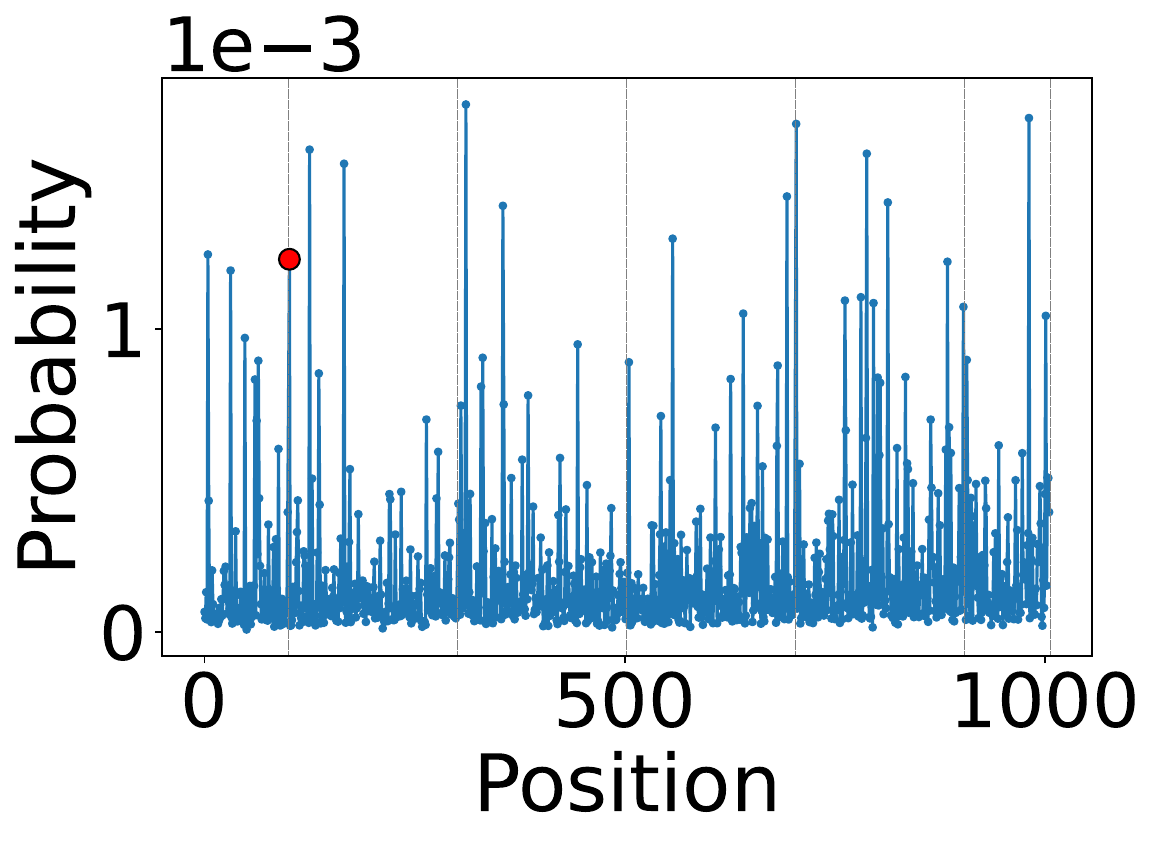} &
\FigGridCell{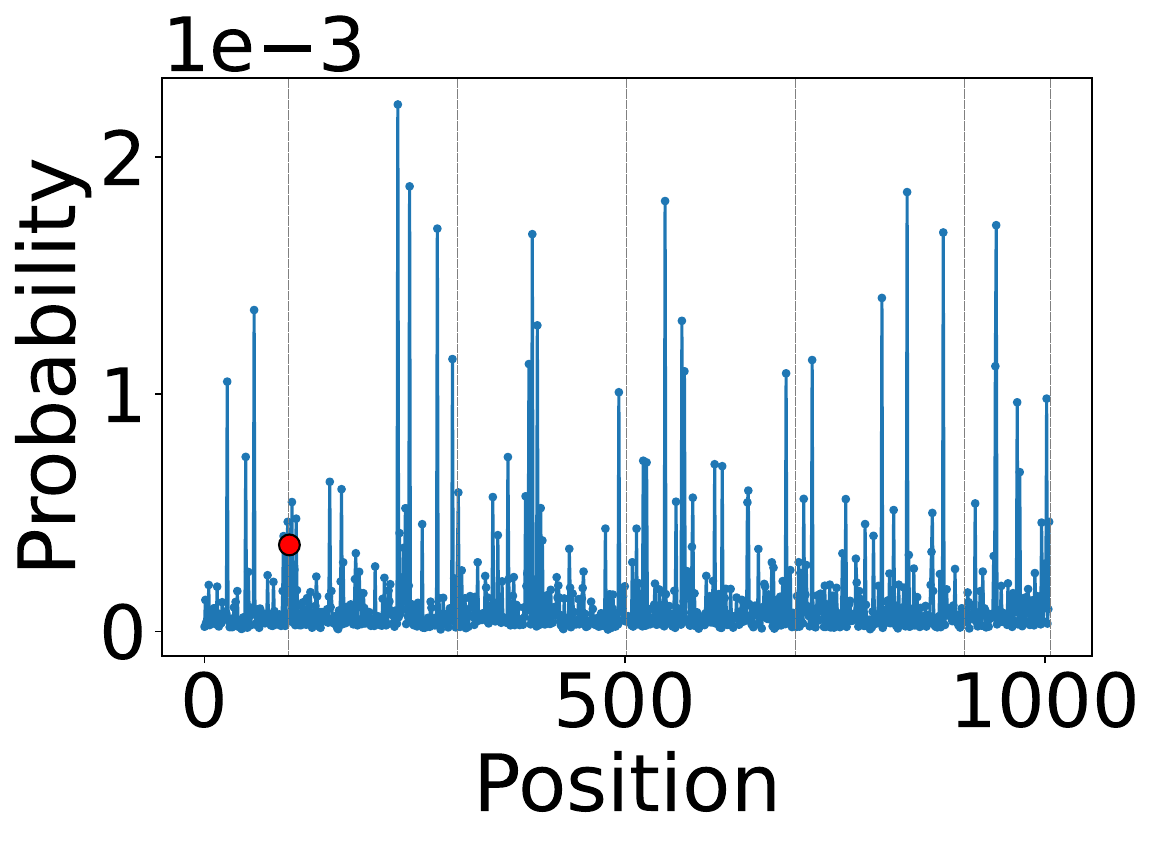} \\
& \FigGridVSubLabel{P2} &
\FigGridCell{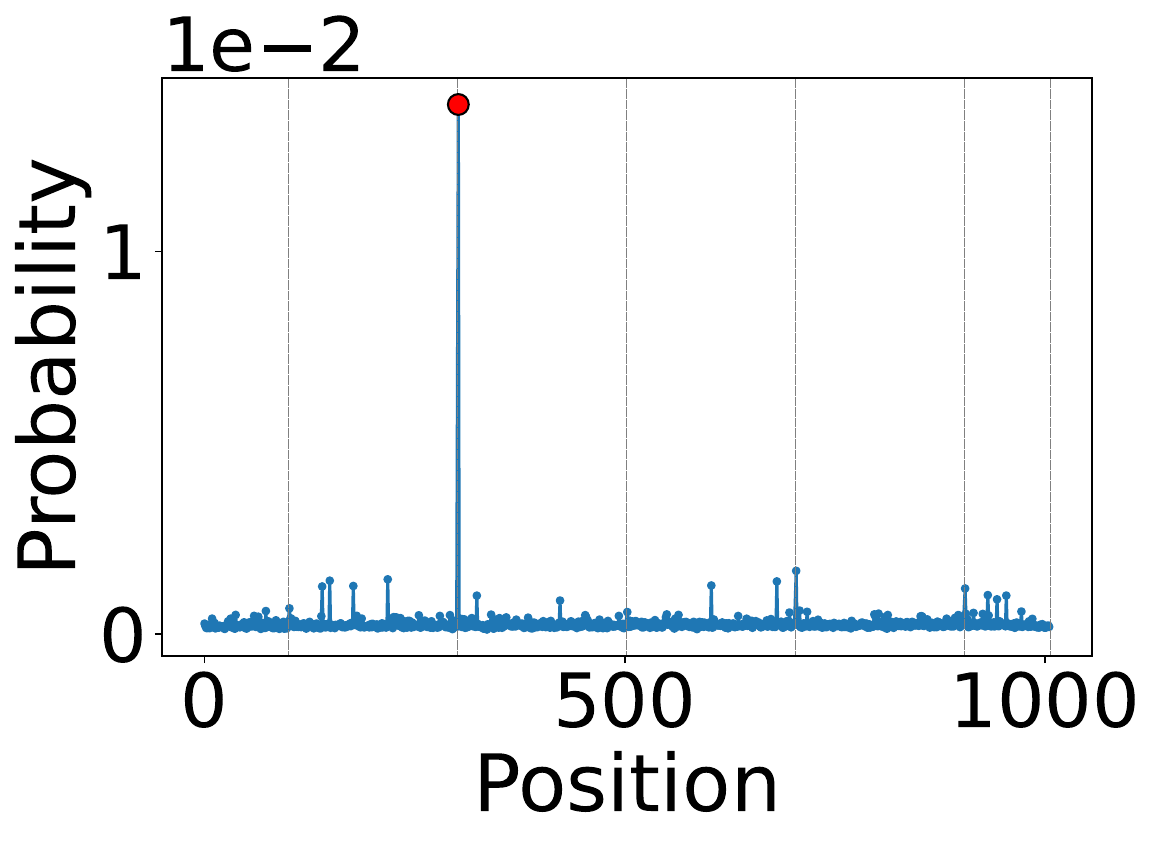} &
\FigGridCell{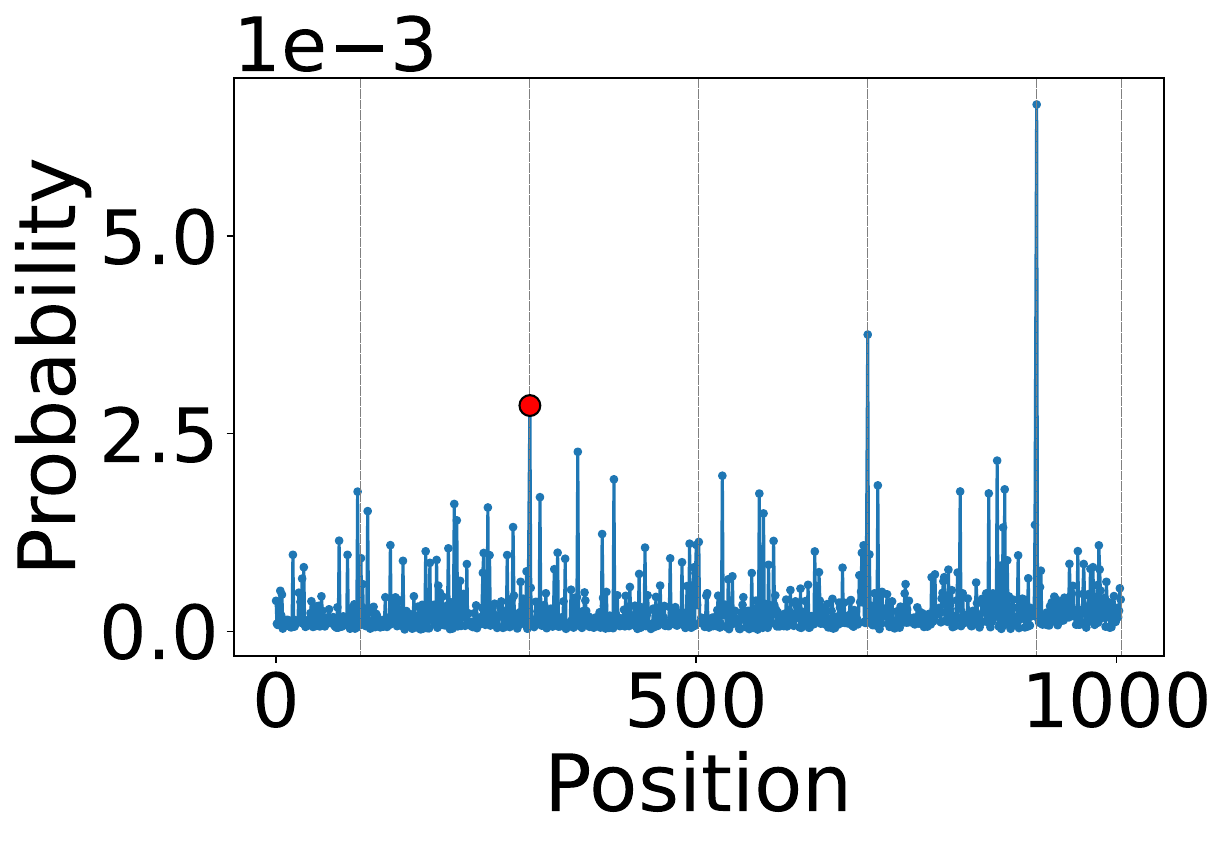} &
\FigGridCell{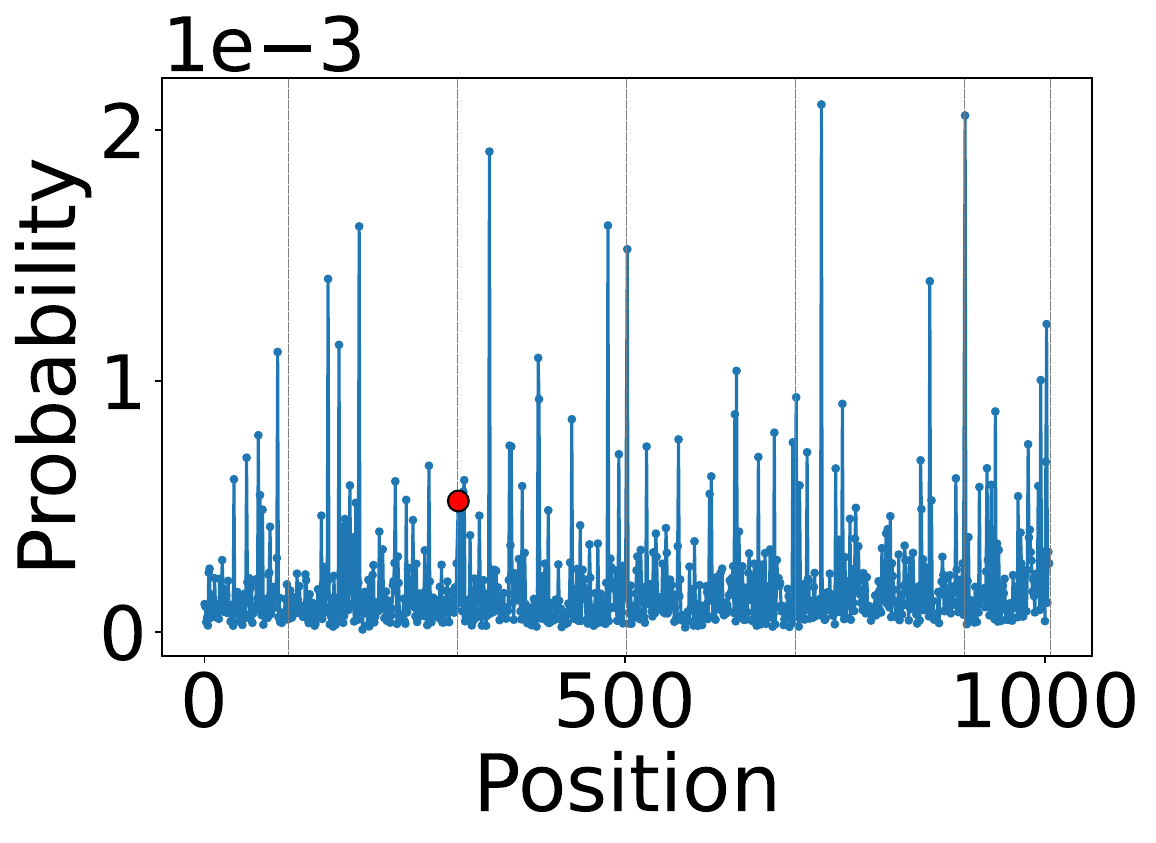} &
\FigGridCell{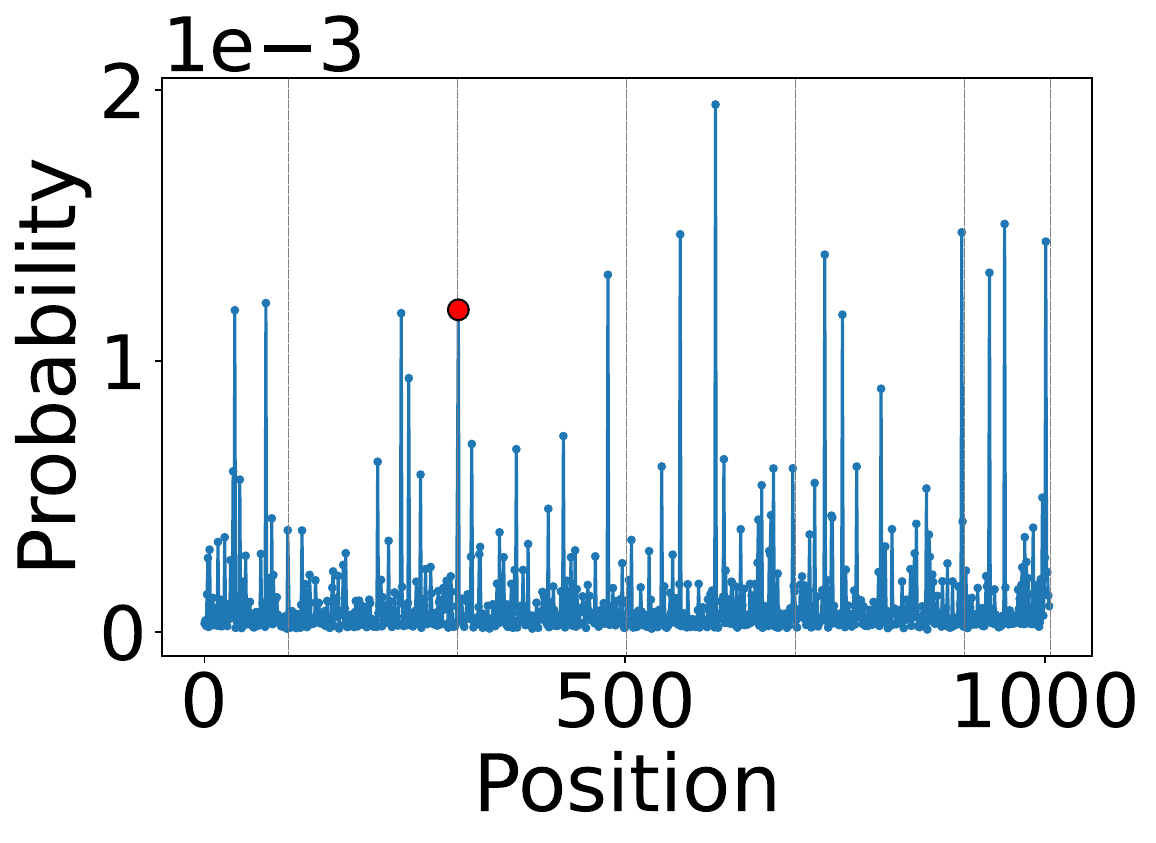} \\
& \FigGridVSubLabel{P3} &
\FigGridCell{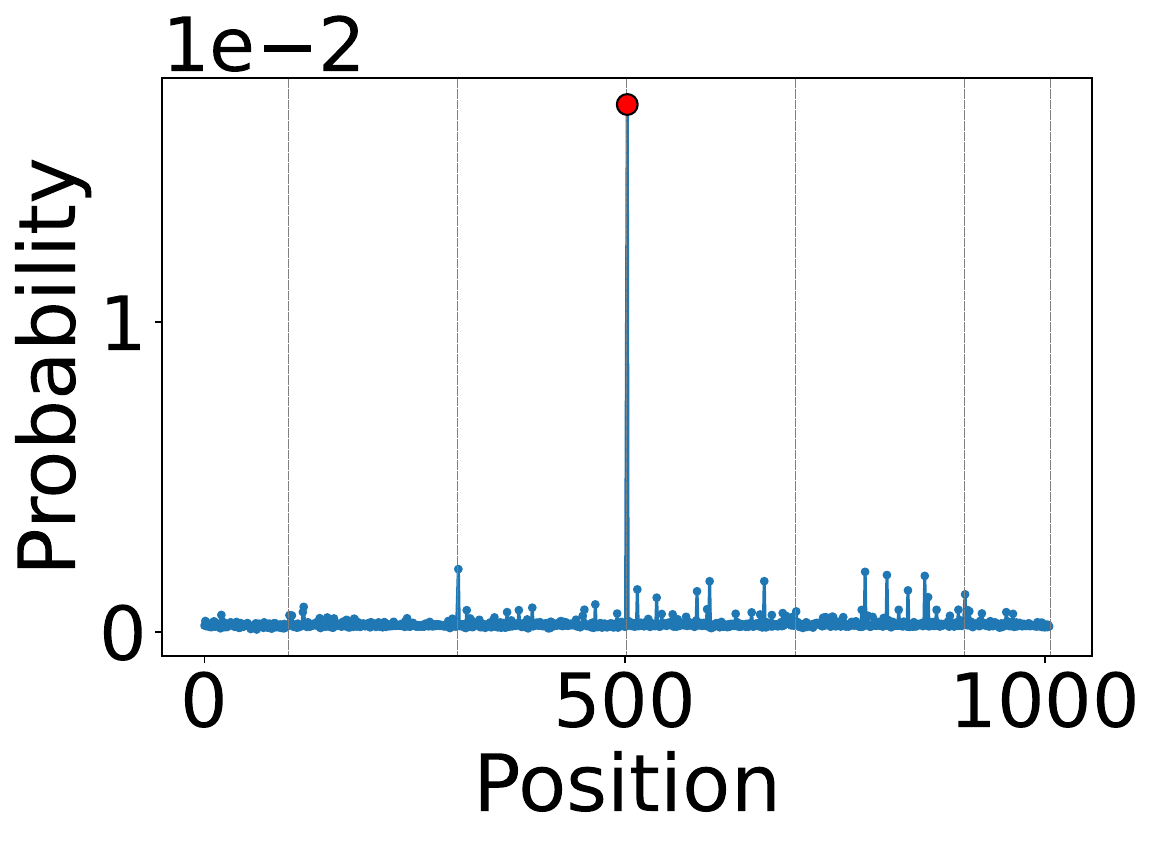} &
\FigGridCell{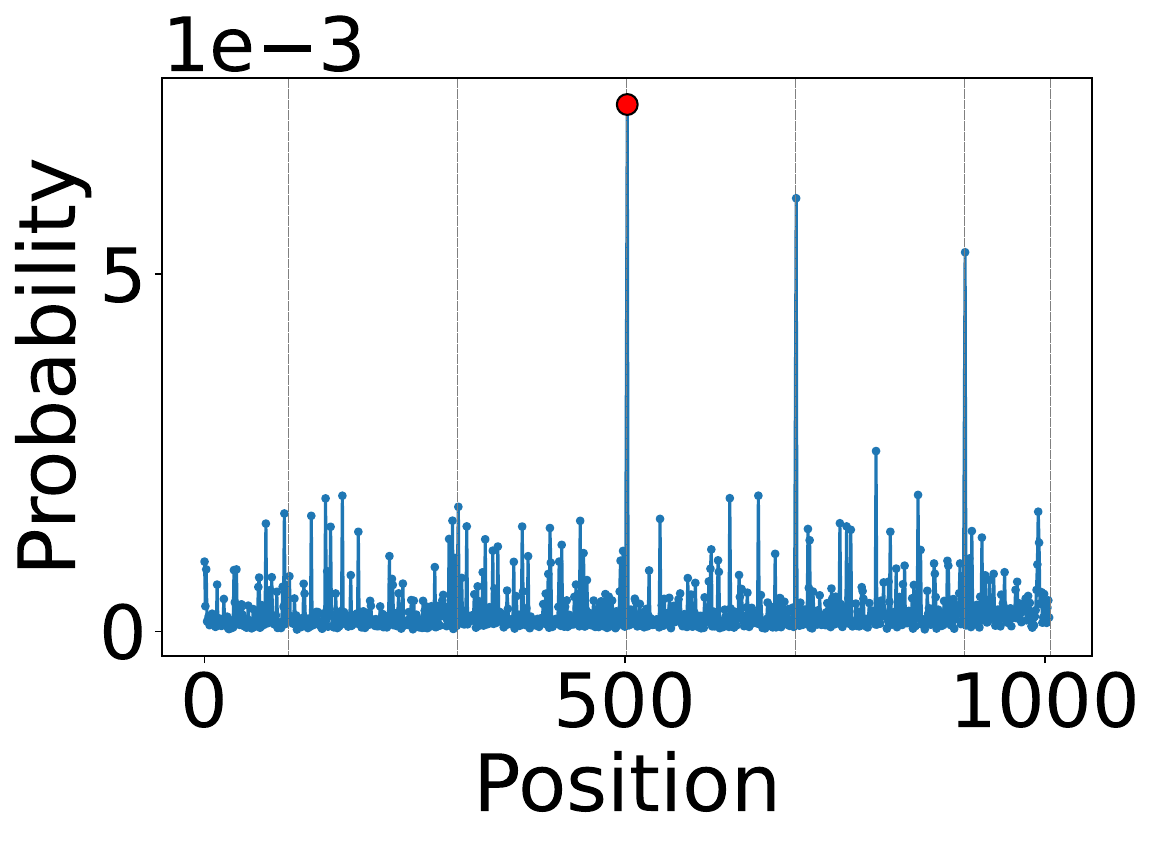} &
\FigGridCell{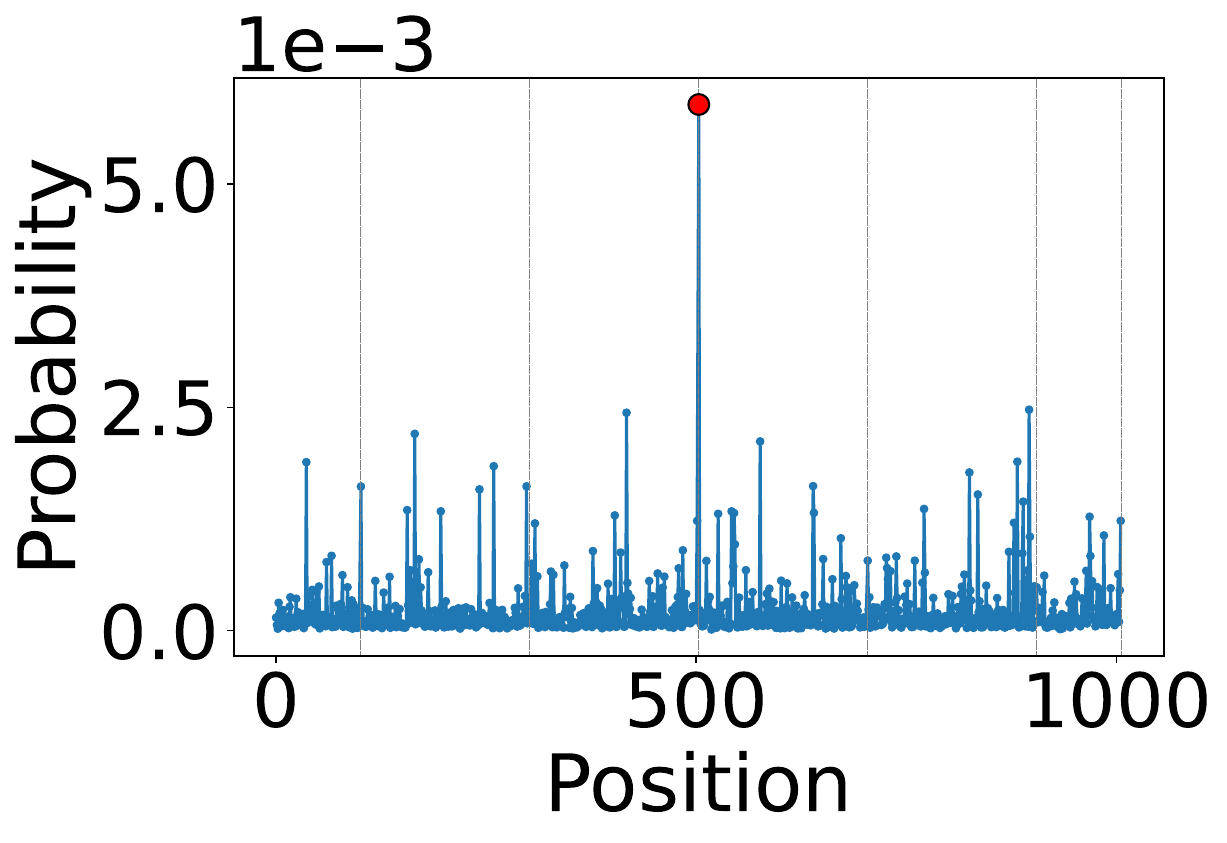} &
\FigGridCell{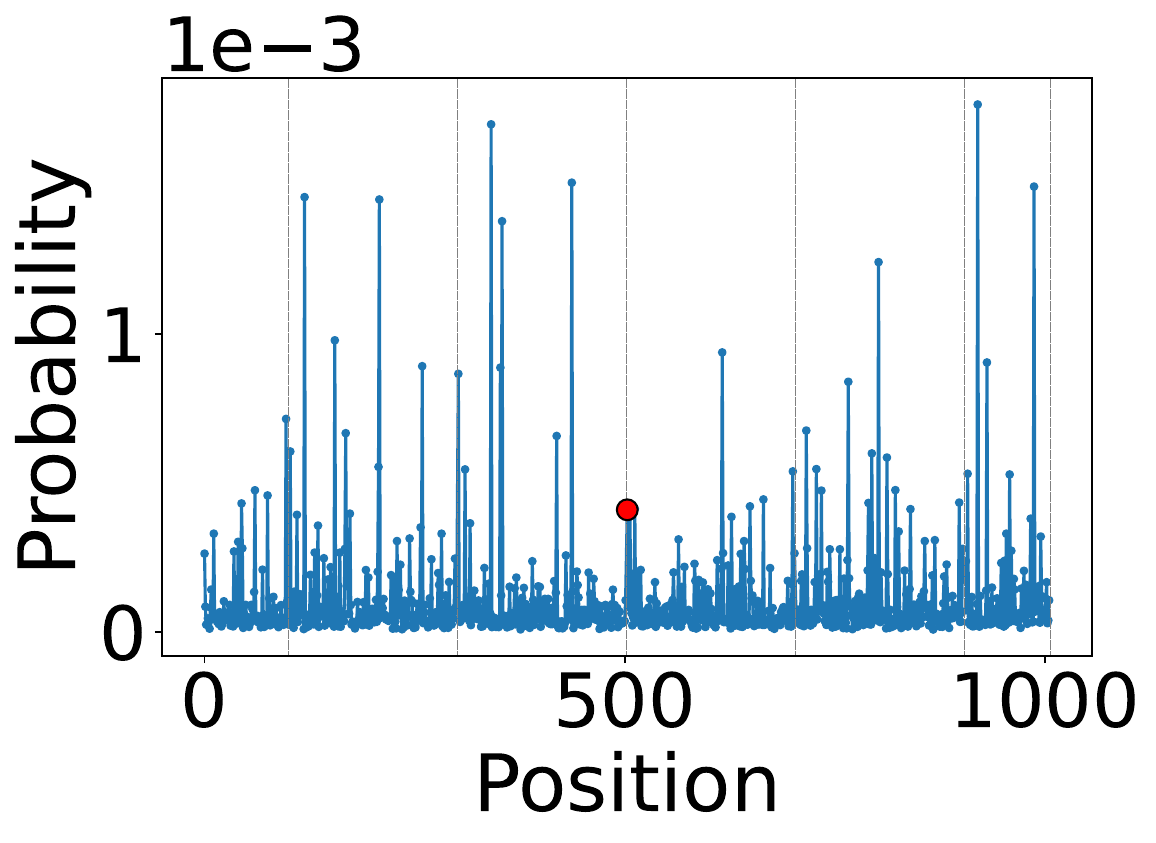} \\
& \FigGridVSubLabel{P4} &
\FigGridCell{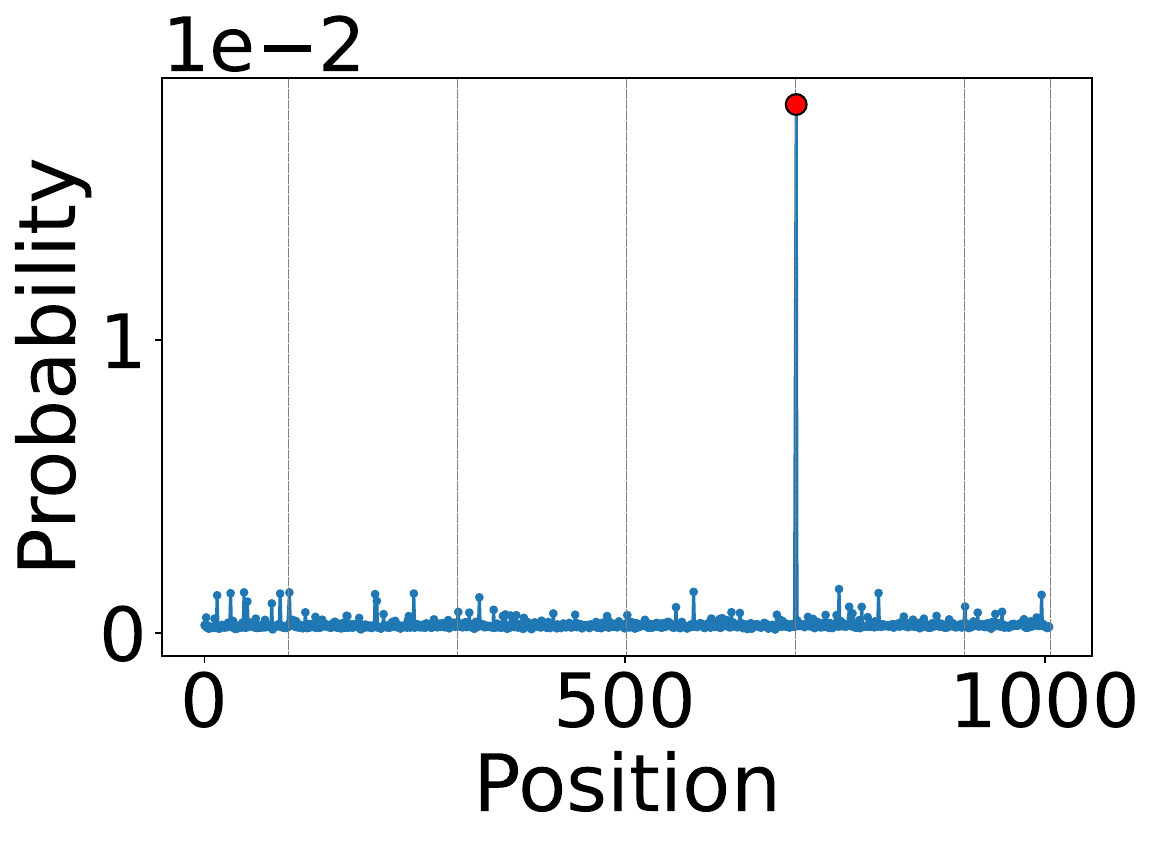} &
\FigGridCell{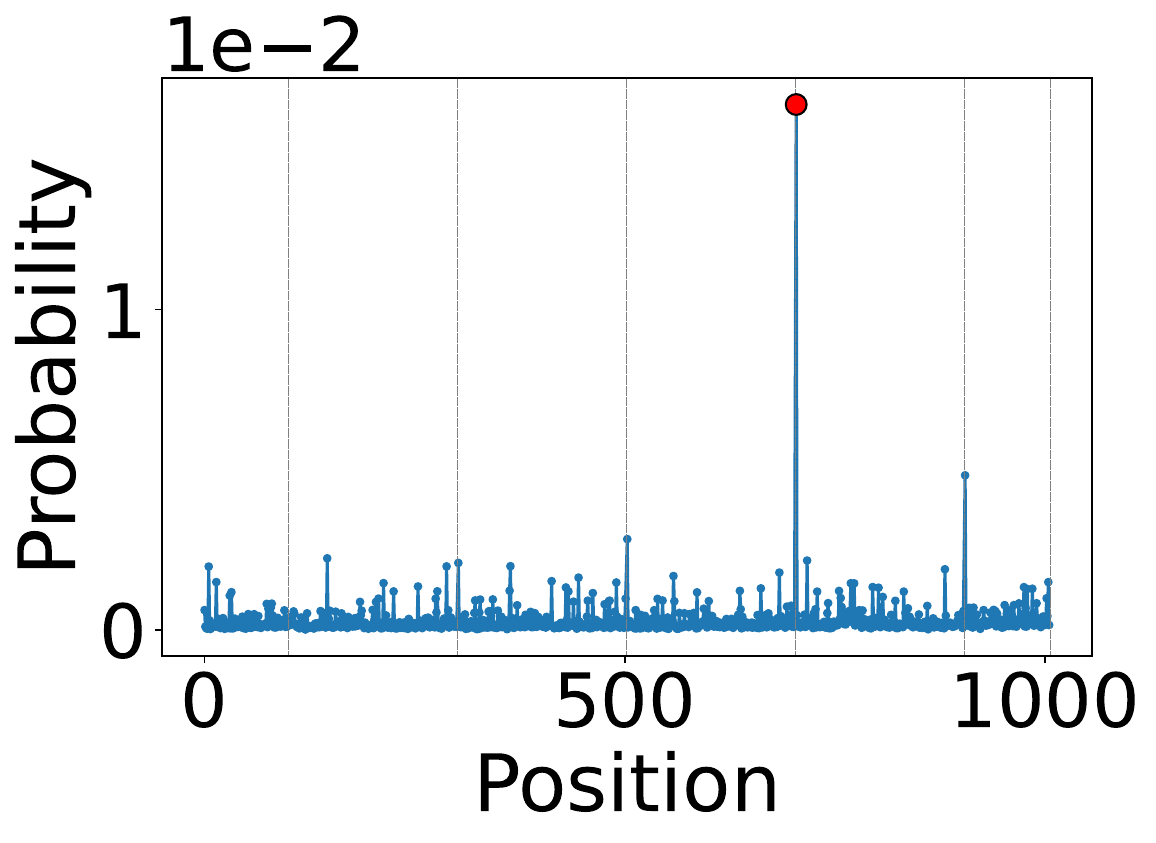} &
\FigGridCell{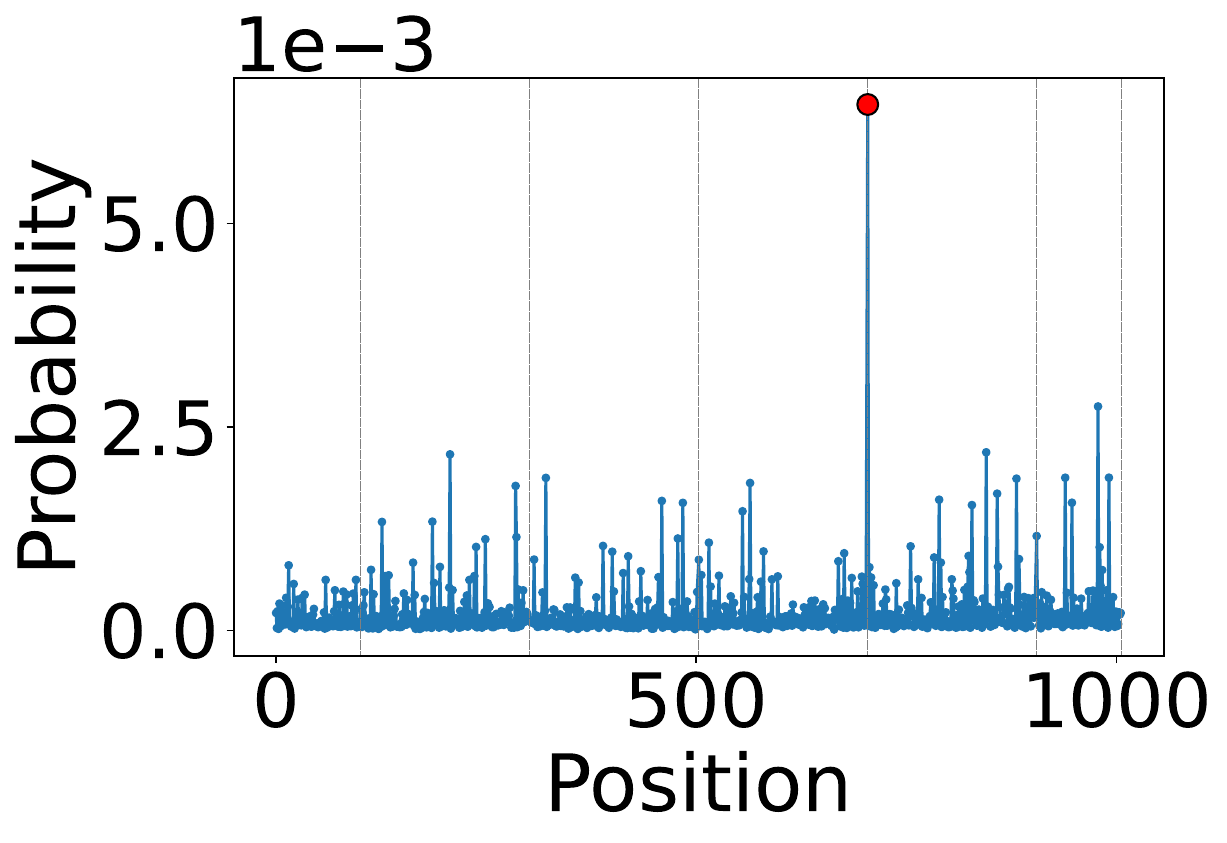} &
\FigGridCell{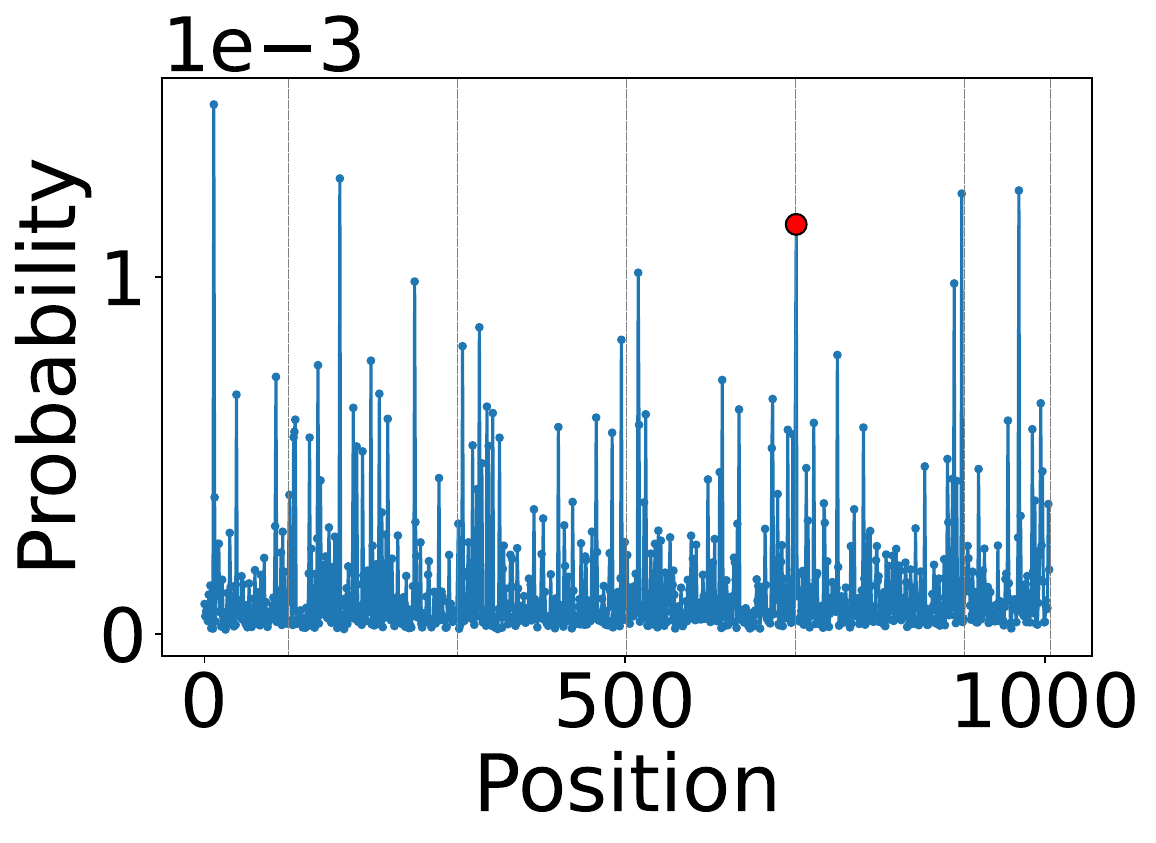} \\
& \FigGridVSubLabel{P5} &
\FigGridCell{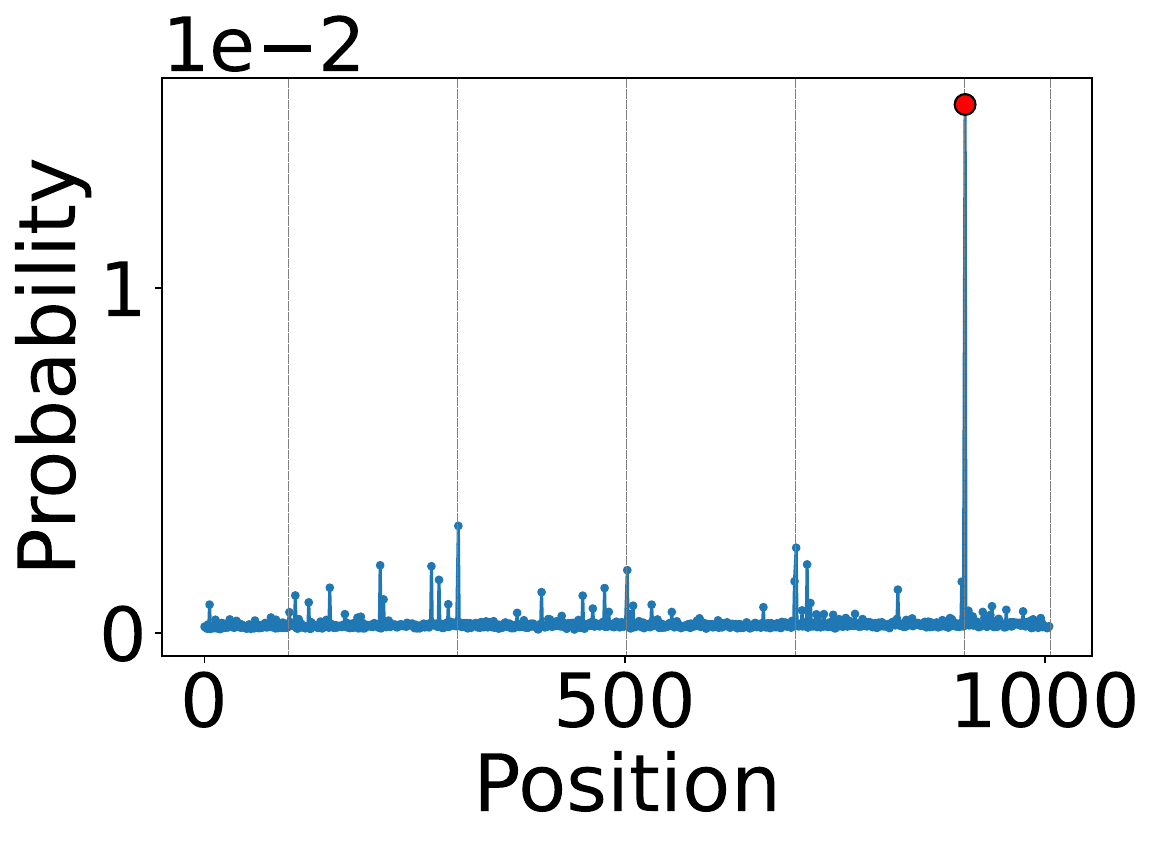} &
\FigGridCell{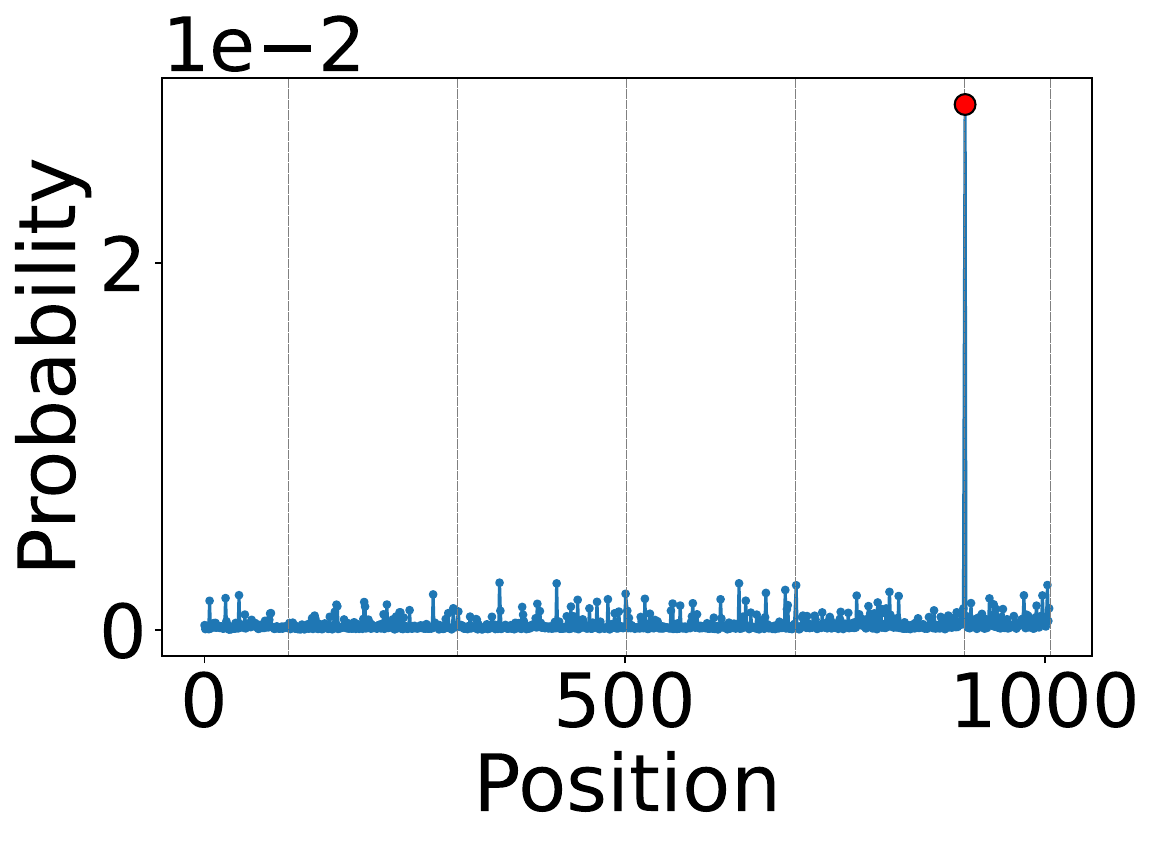} &
\FigGridCell{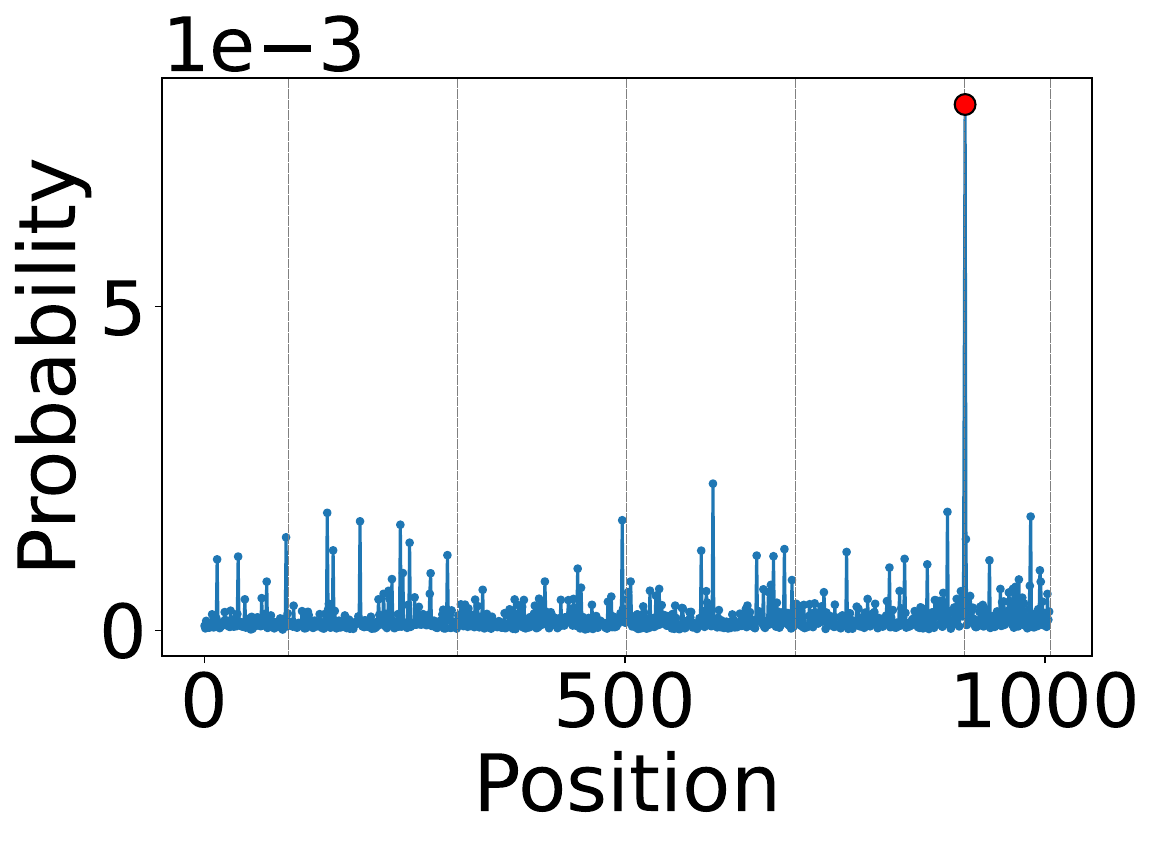} &
\FigGridCell{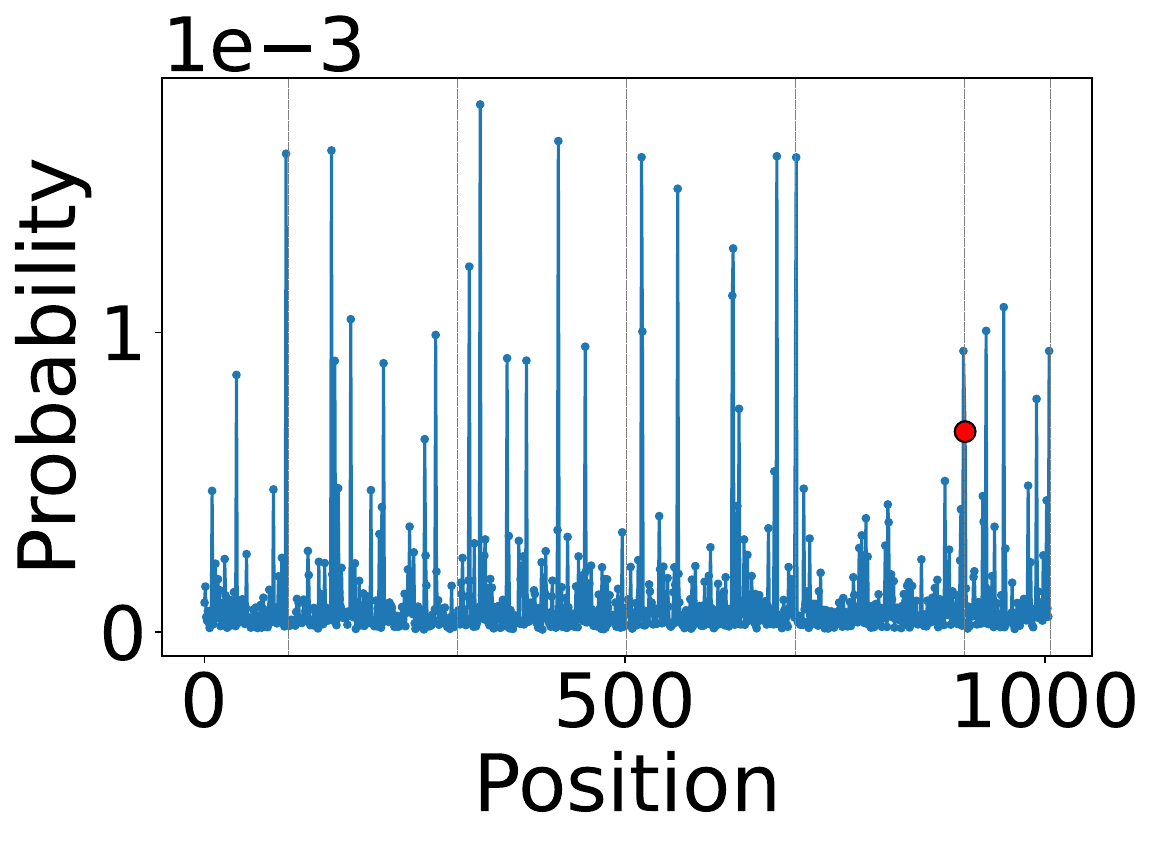} \\
\addlinespace[1pt]
\midrule
\addlinespace[1pt]
\multirow{5}{*}{\FigGridVLabelSpanFive{Random}} &
\FigGridVSubLabel{P1} &
\FigGridCell{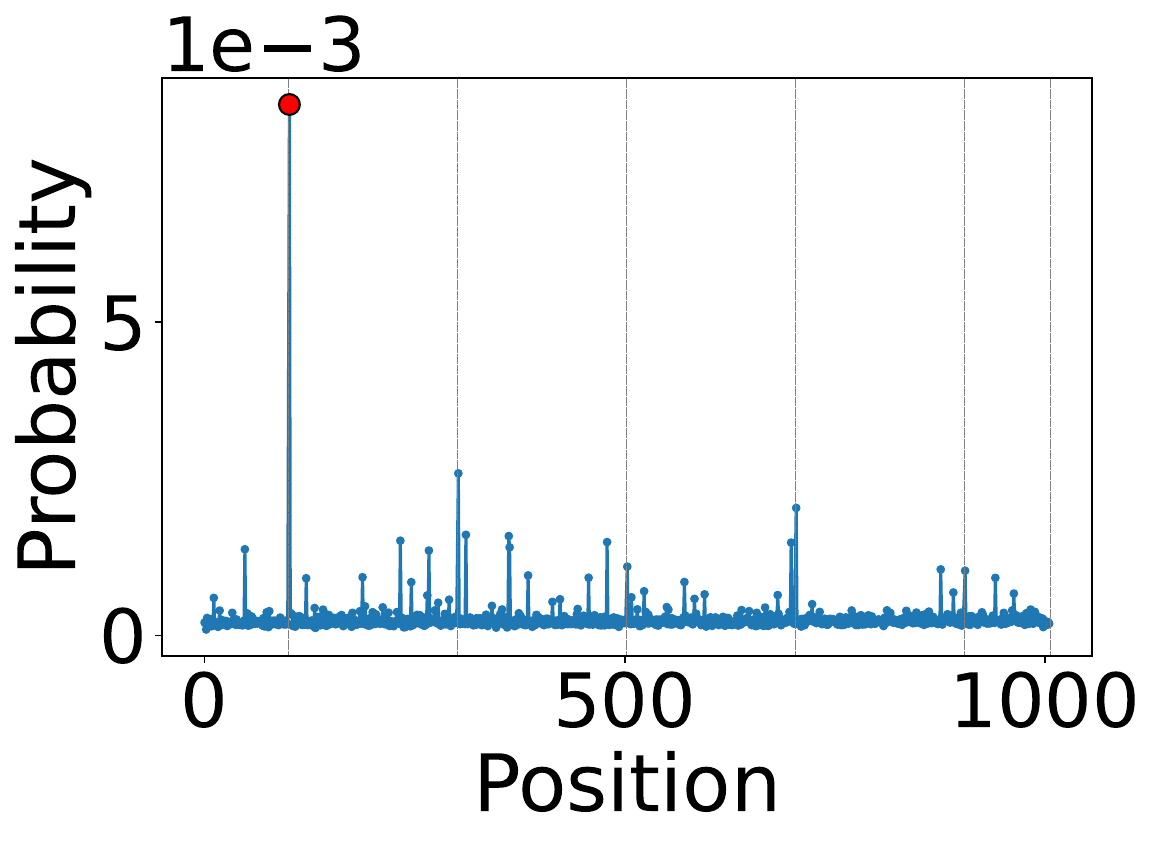} &
\FigGridCell{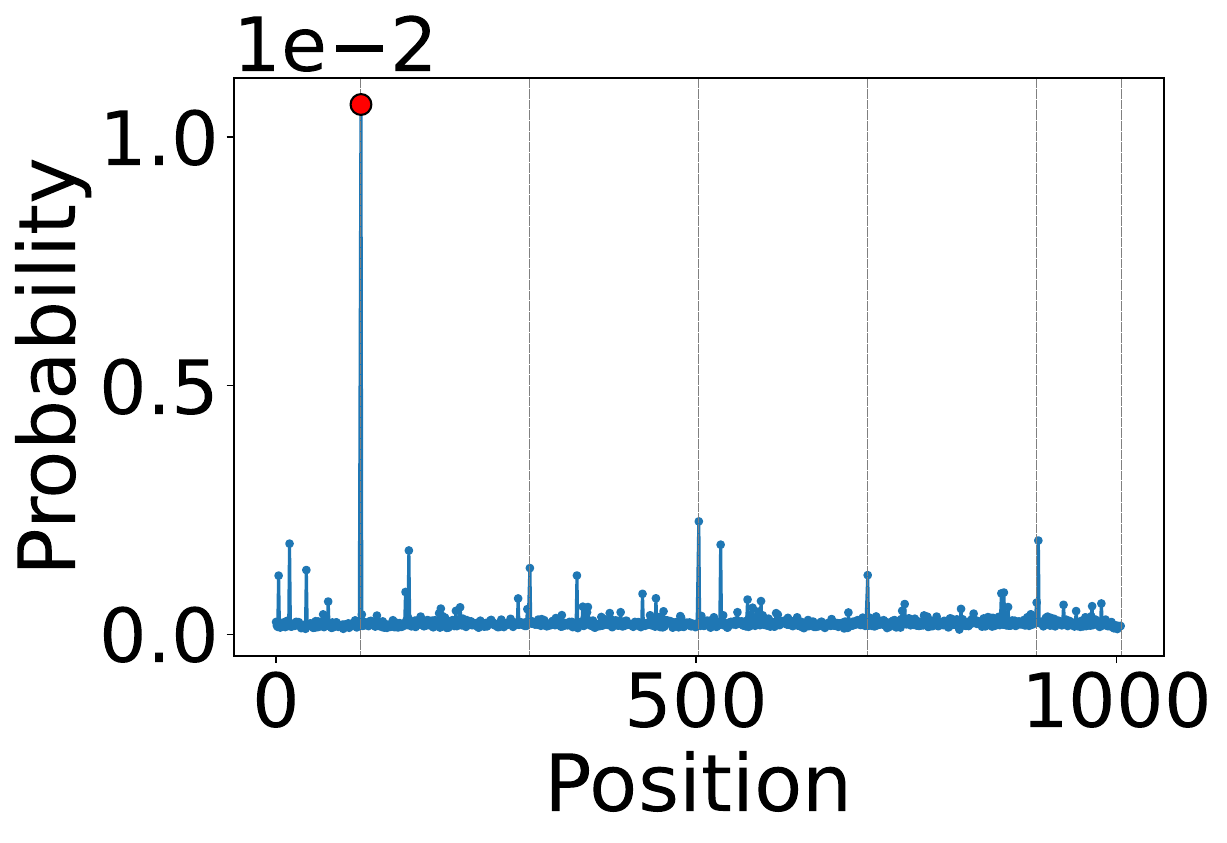} &
\FigGridCell{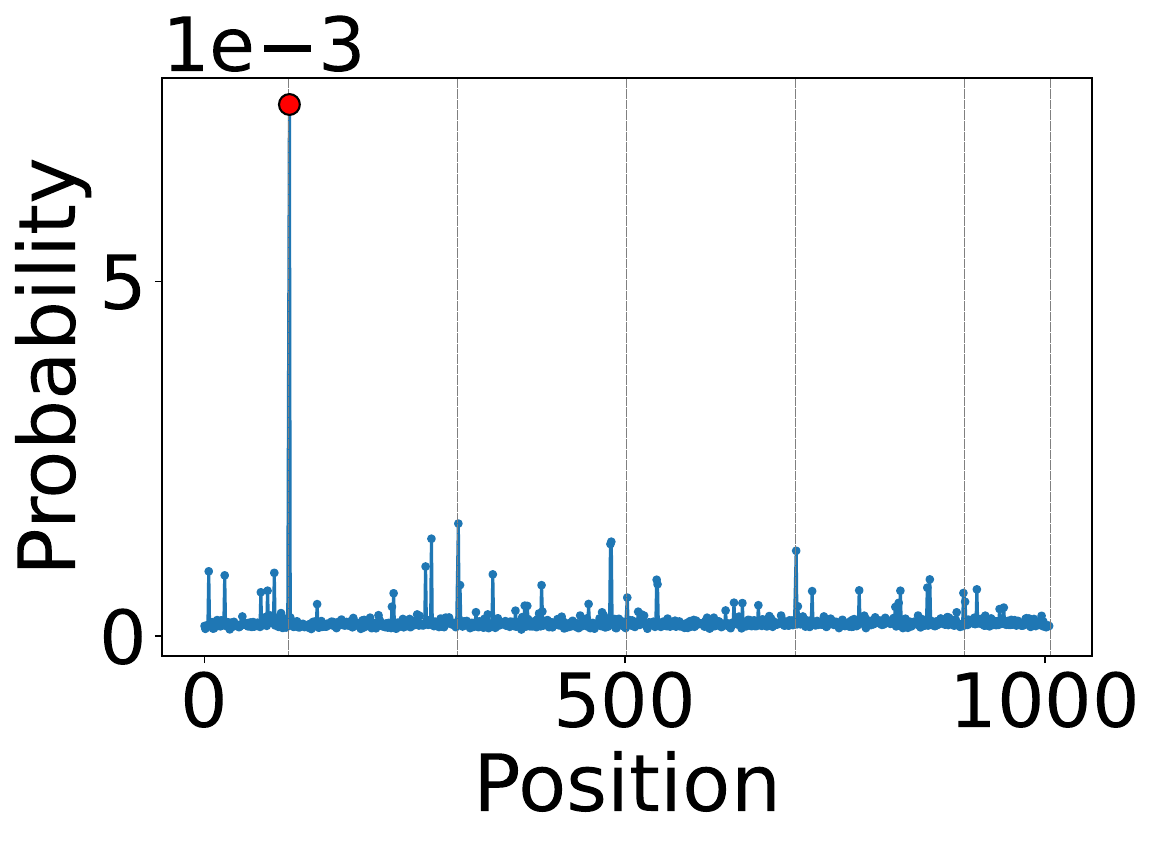} &
\FigGridCell{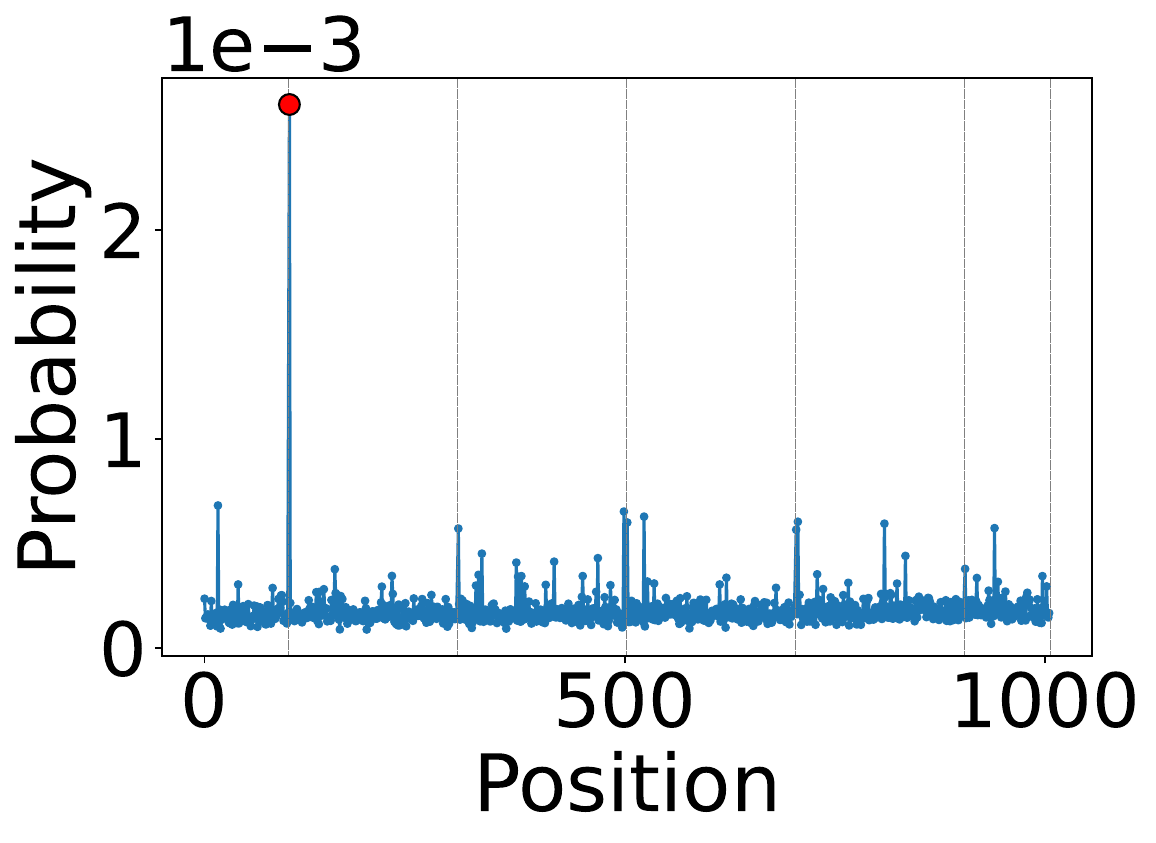} \\
& \FigGridVSubLabel{P2} &
\FigGridCell{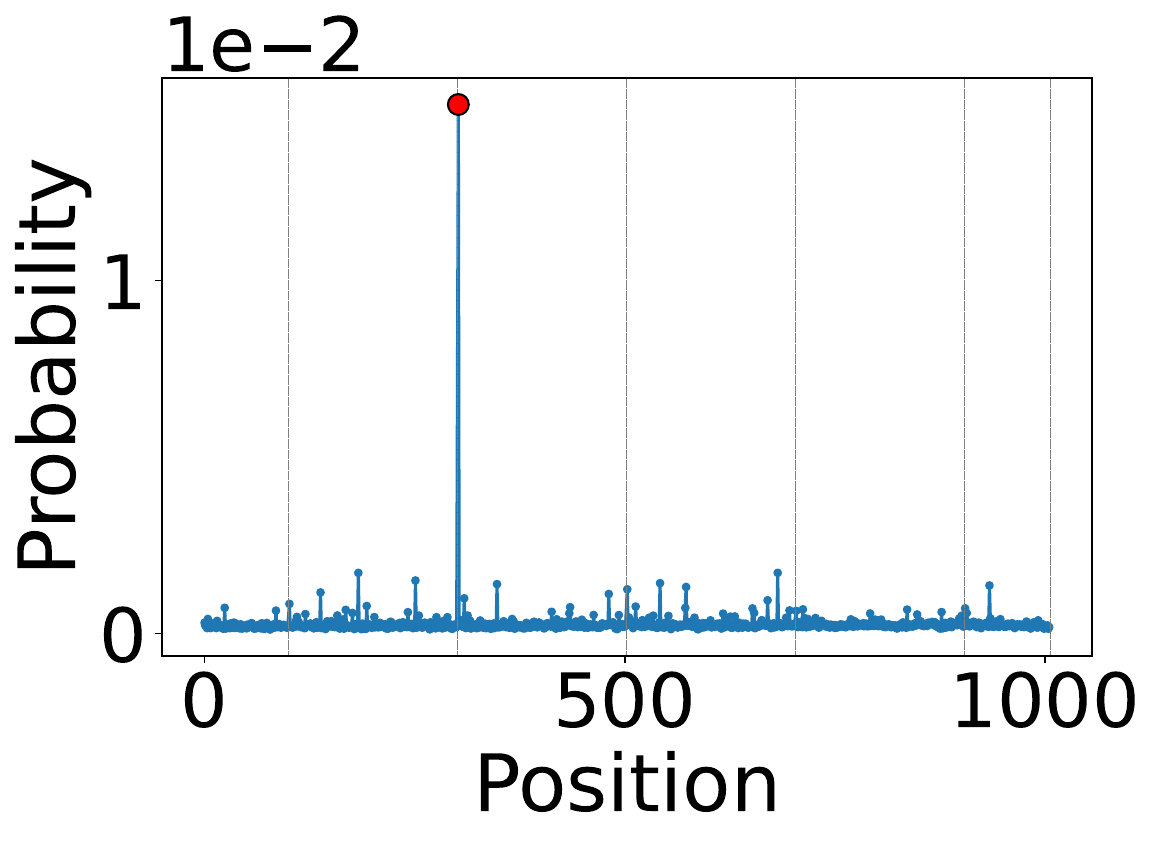} &
\FigGridCell{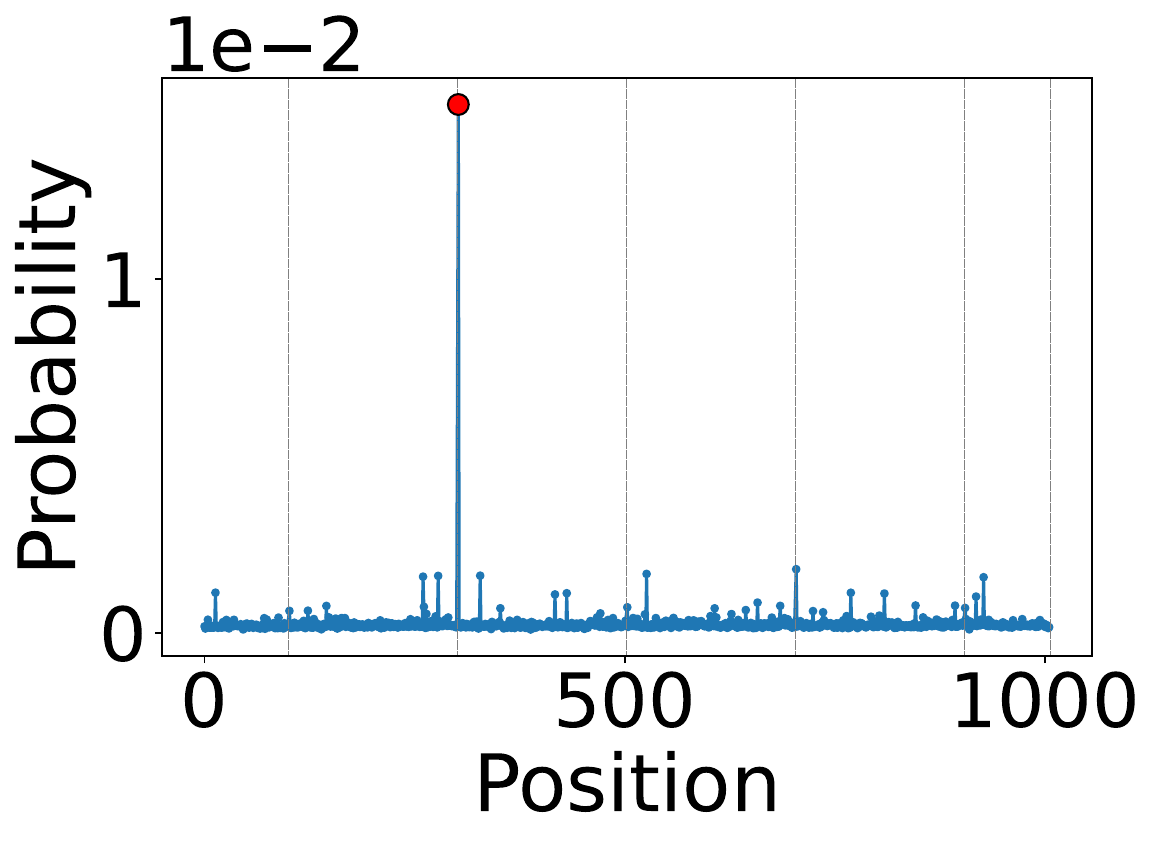} &
\FigGridCell{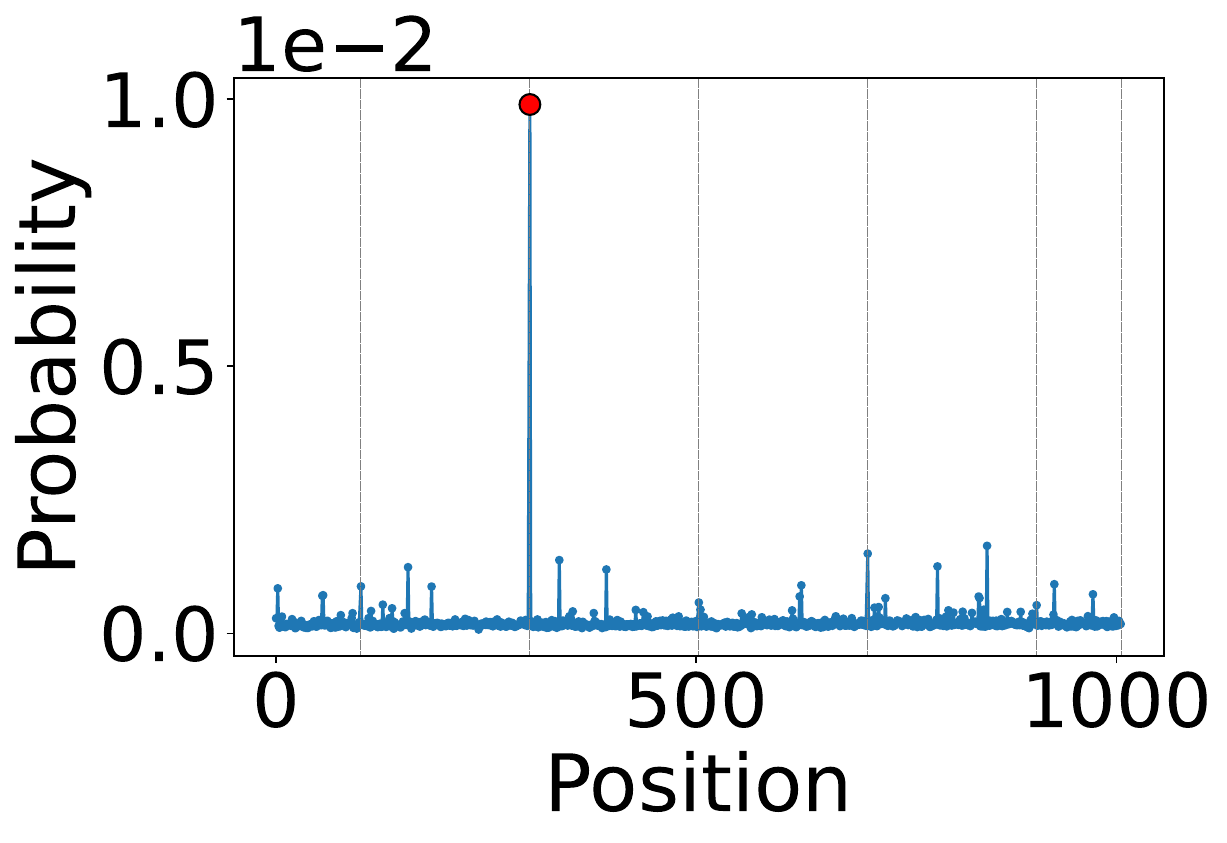} &
\FigGridCell{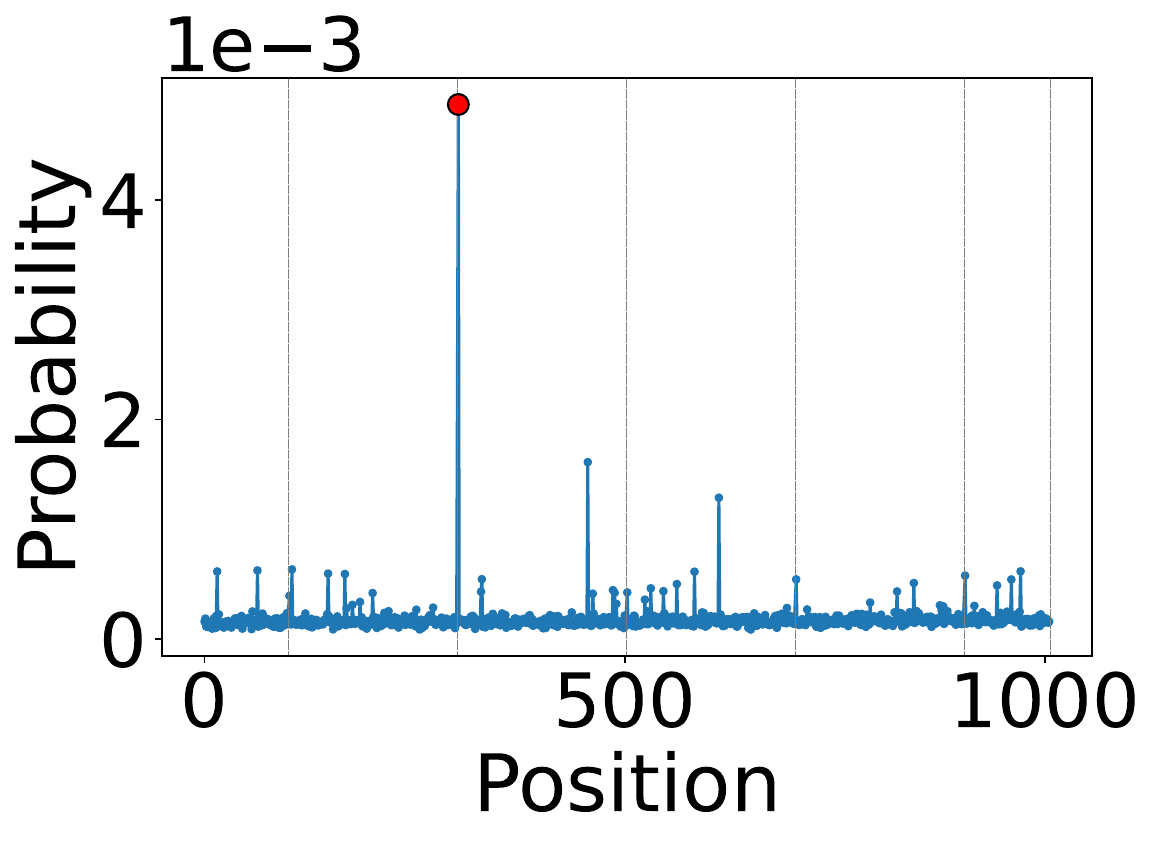} \\
& \FigGridVSubLabel{P3} &
\FigGridCell{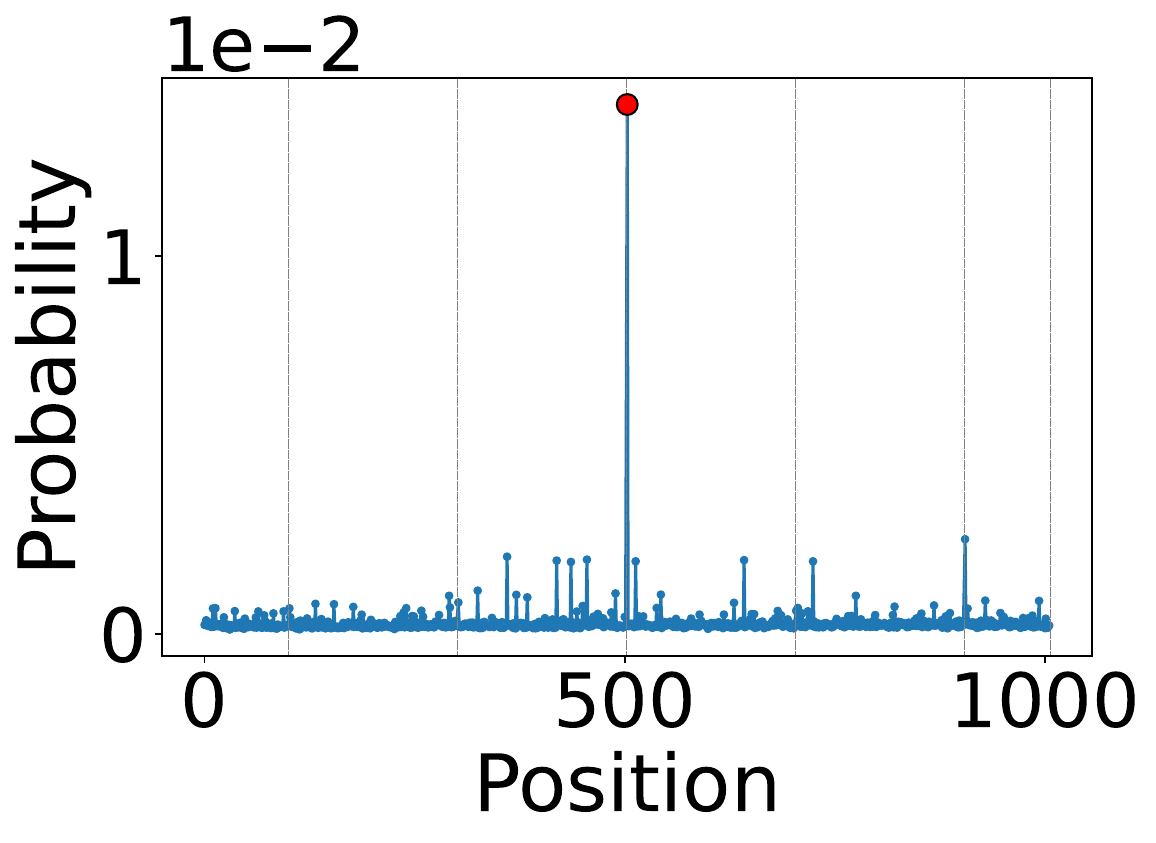} &
\FigGridCell{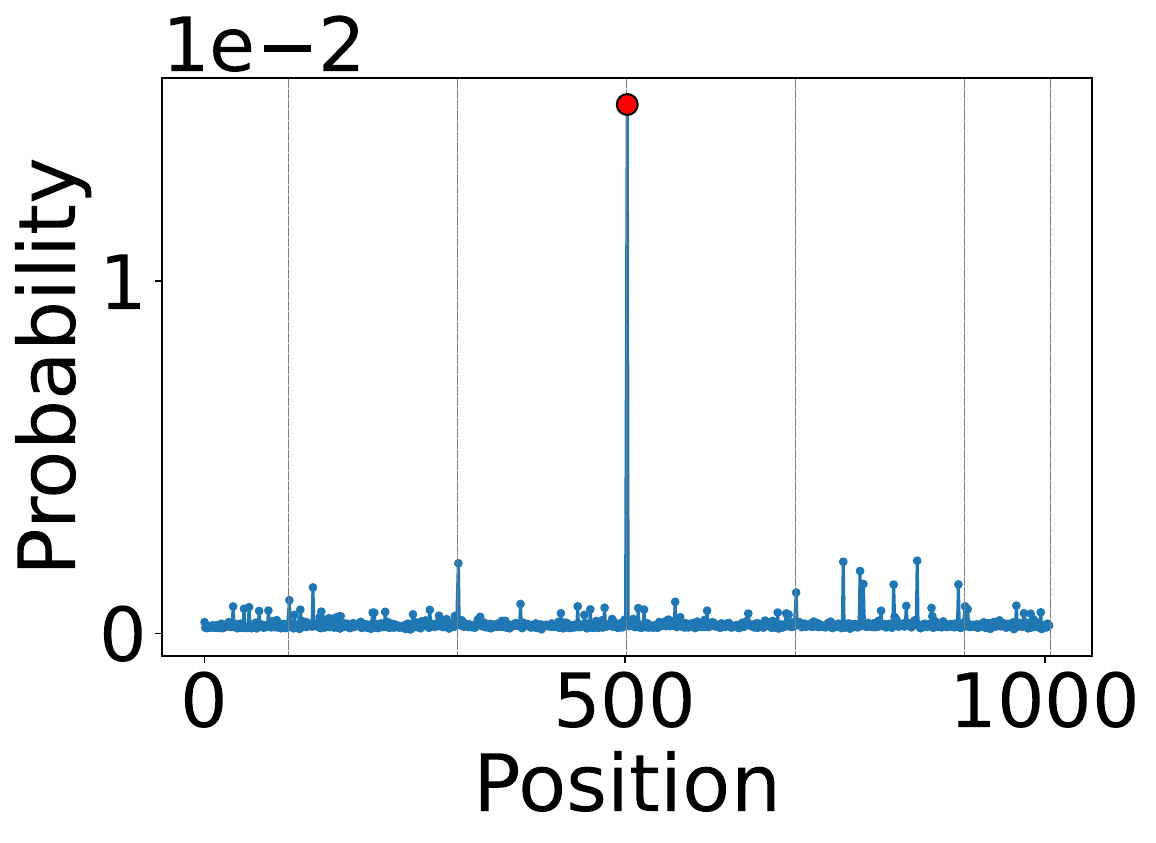} &
\FigGridCell{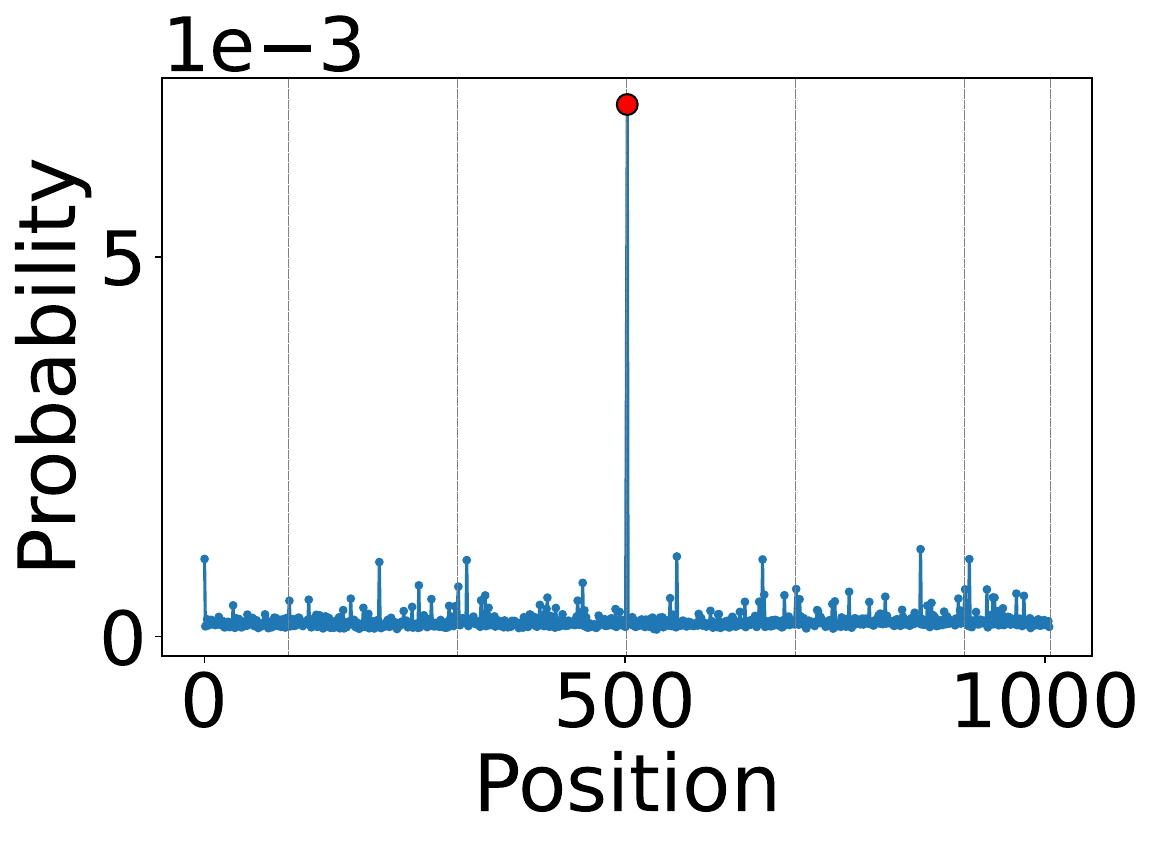} &
\FigGridCell{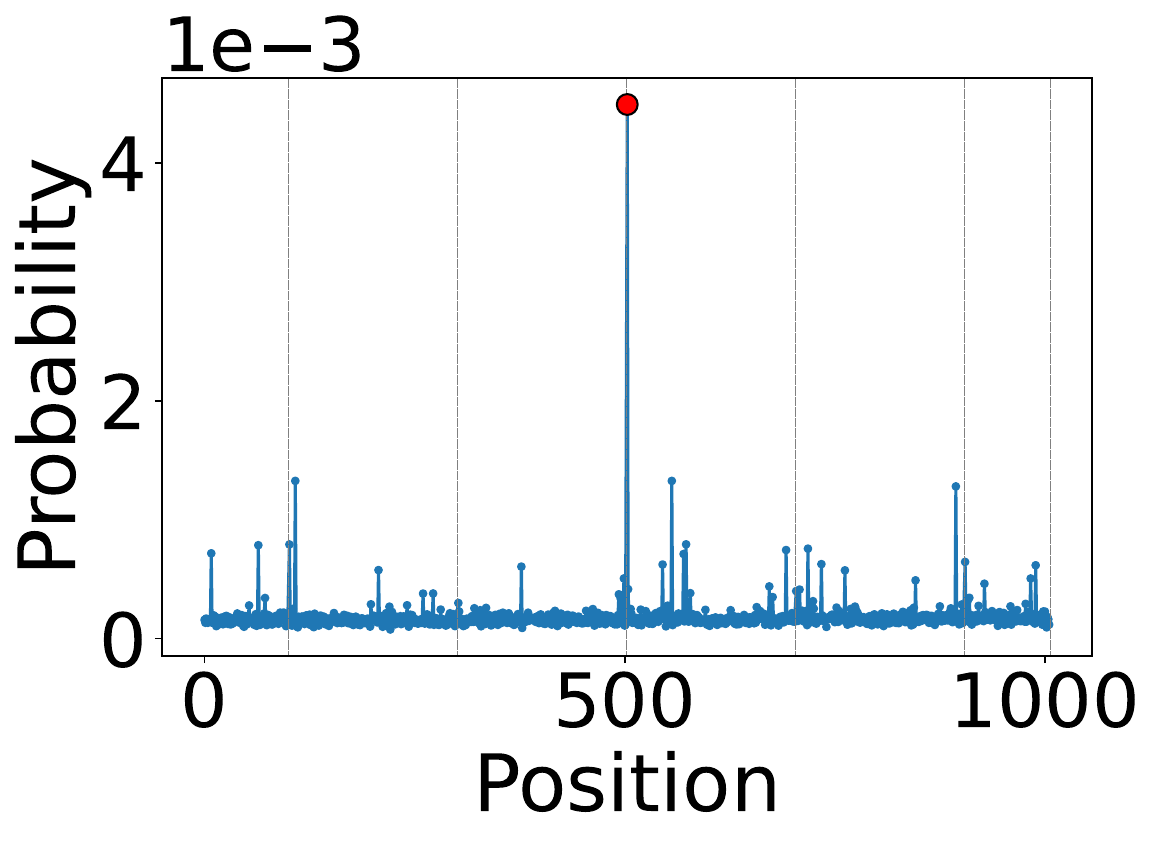} \\
& \FigGridVSubLabel{P4} &
\FigGridCell{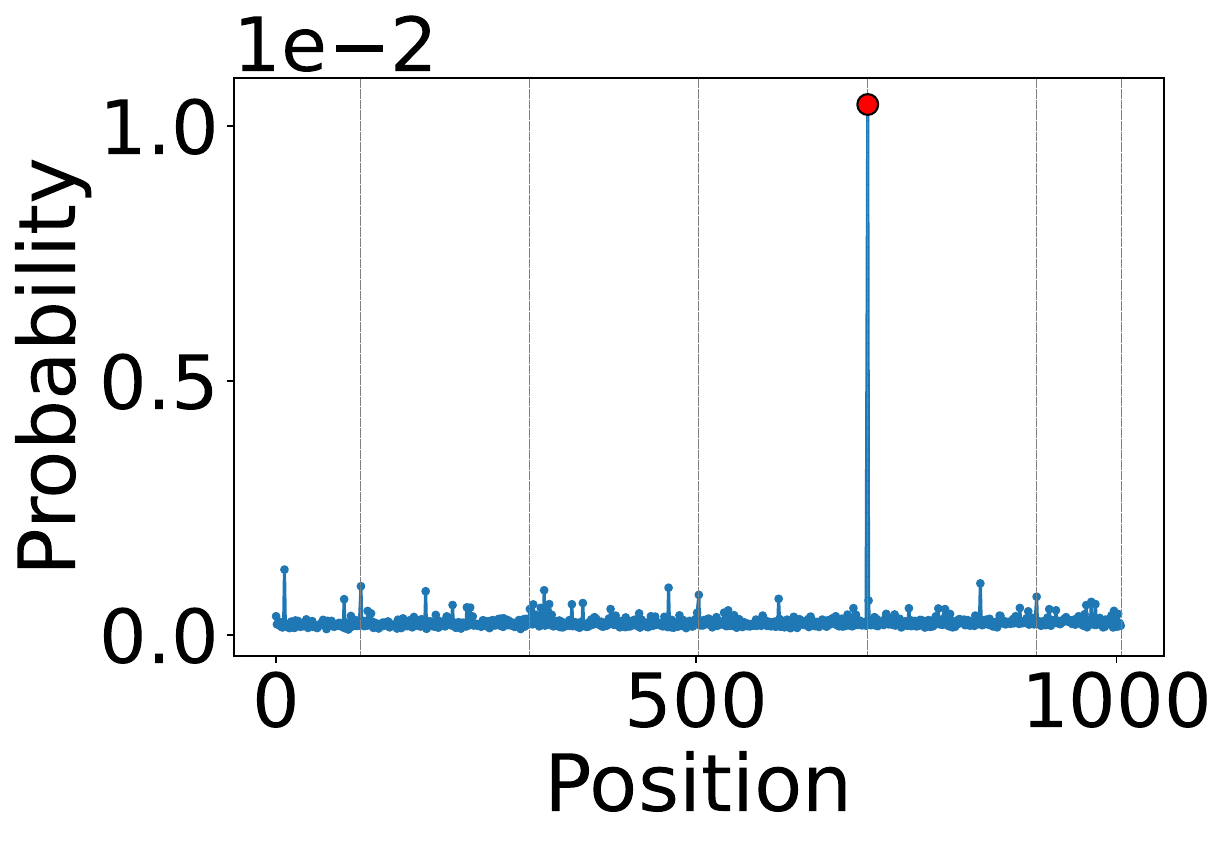} &
\FigGridCell{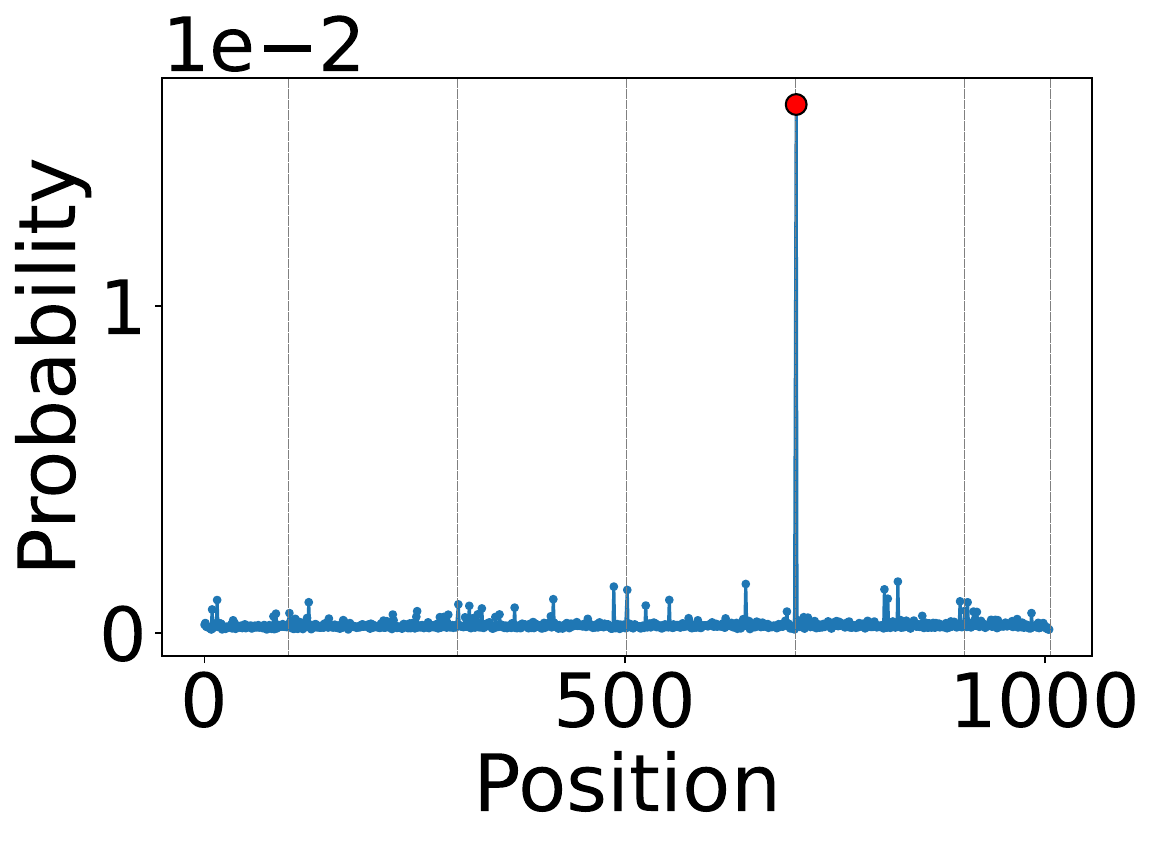} &
\FigGridCell{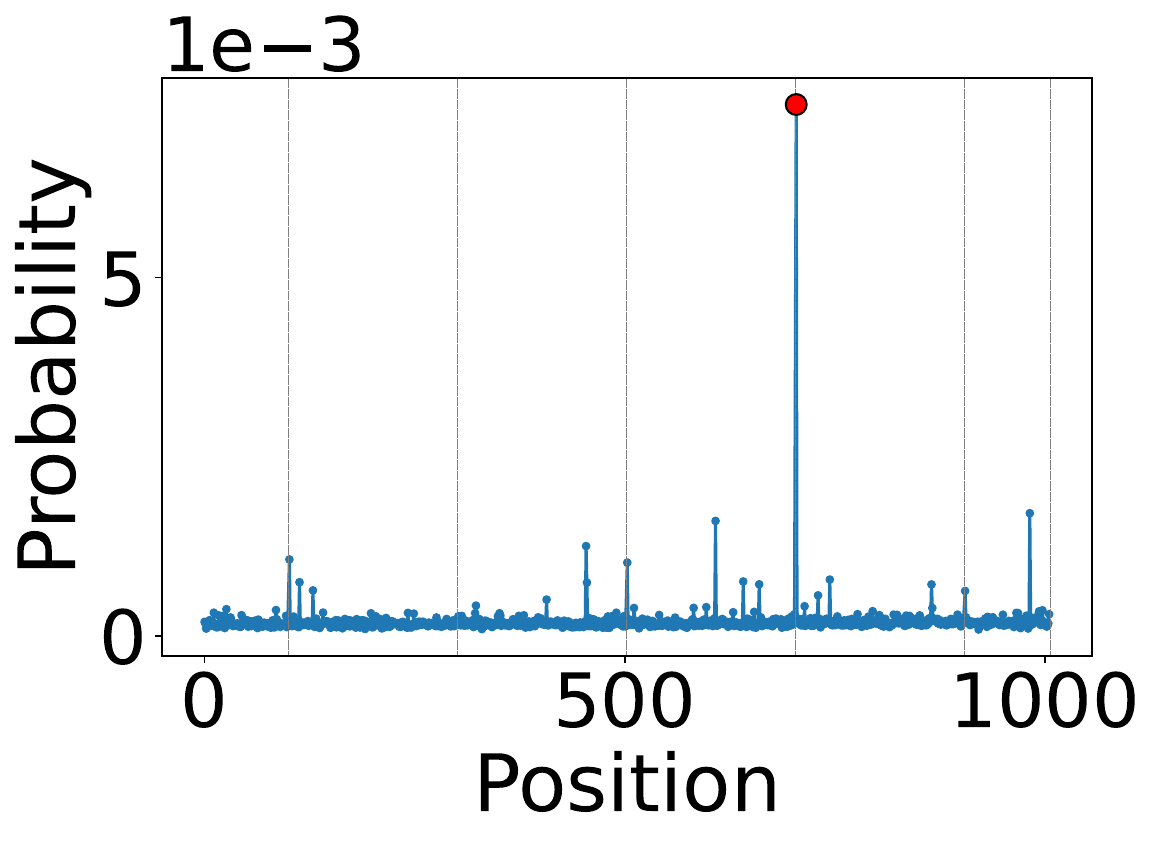} &
\FigGridCell{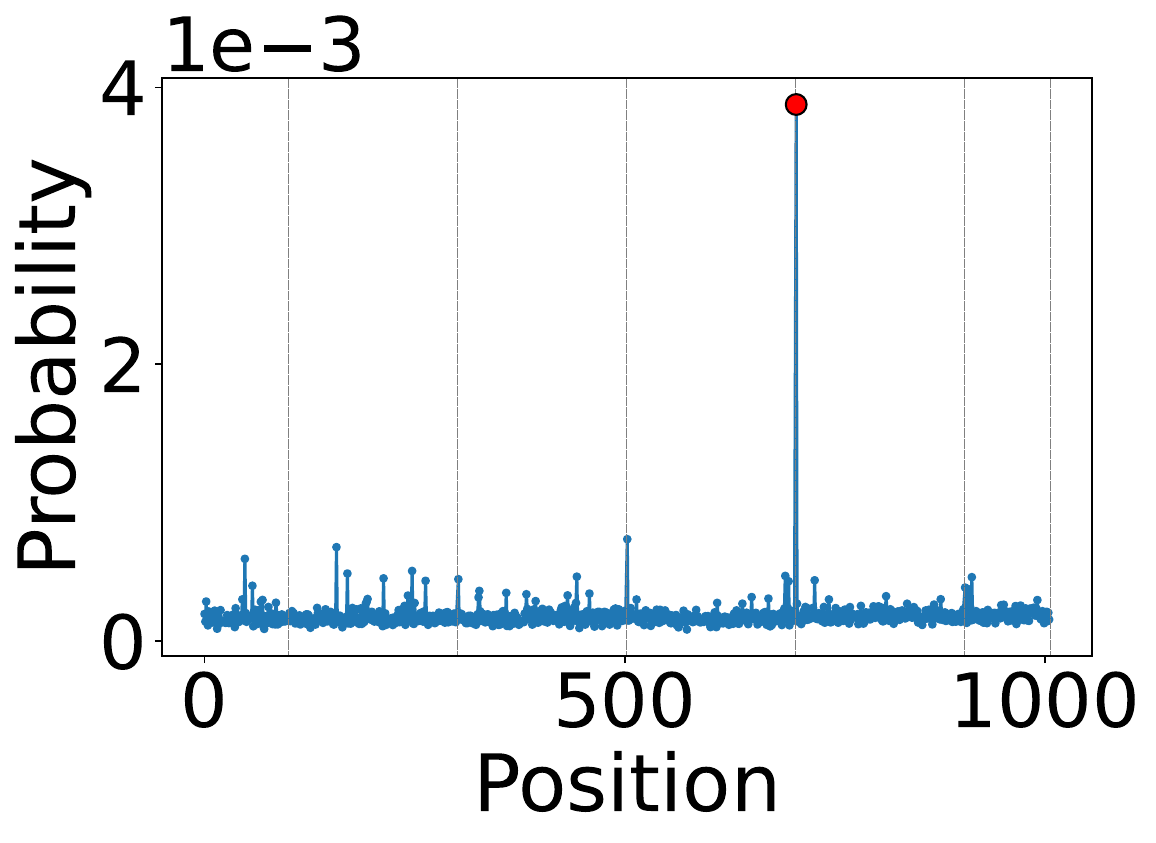} \\
& \FigGridVSubLabel{P5} &
\FigGridCell{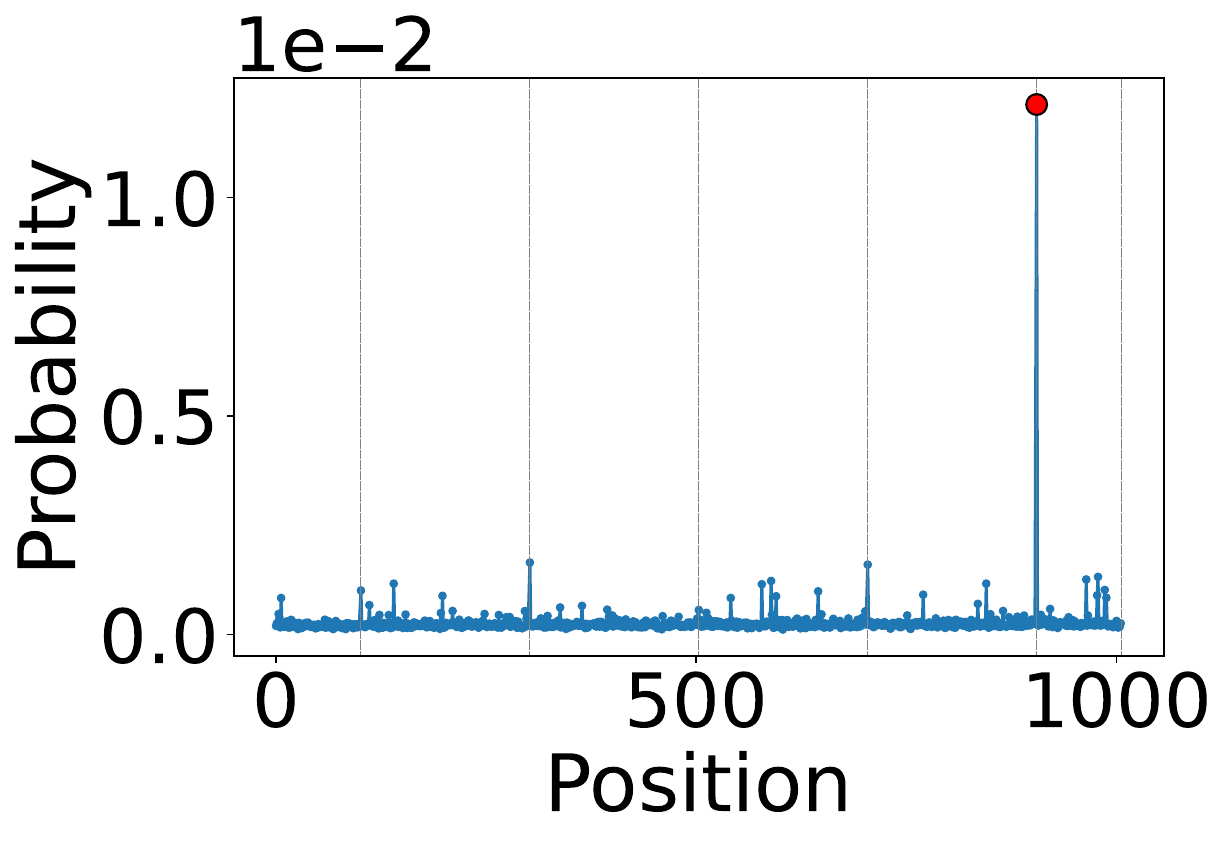} &
\FigGridCell{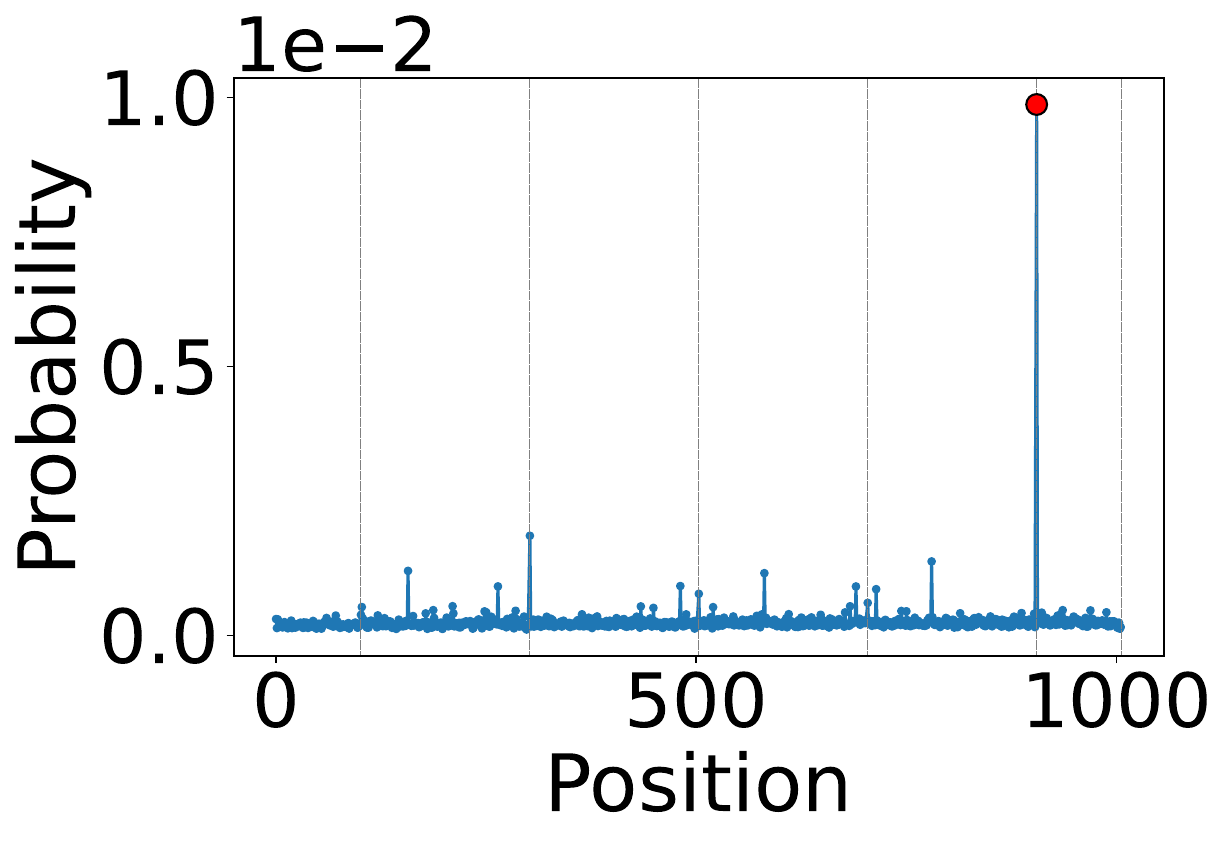} &
\FigGridCell{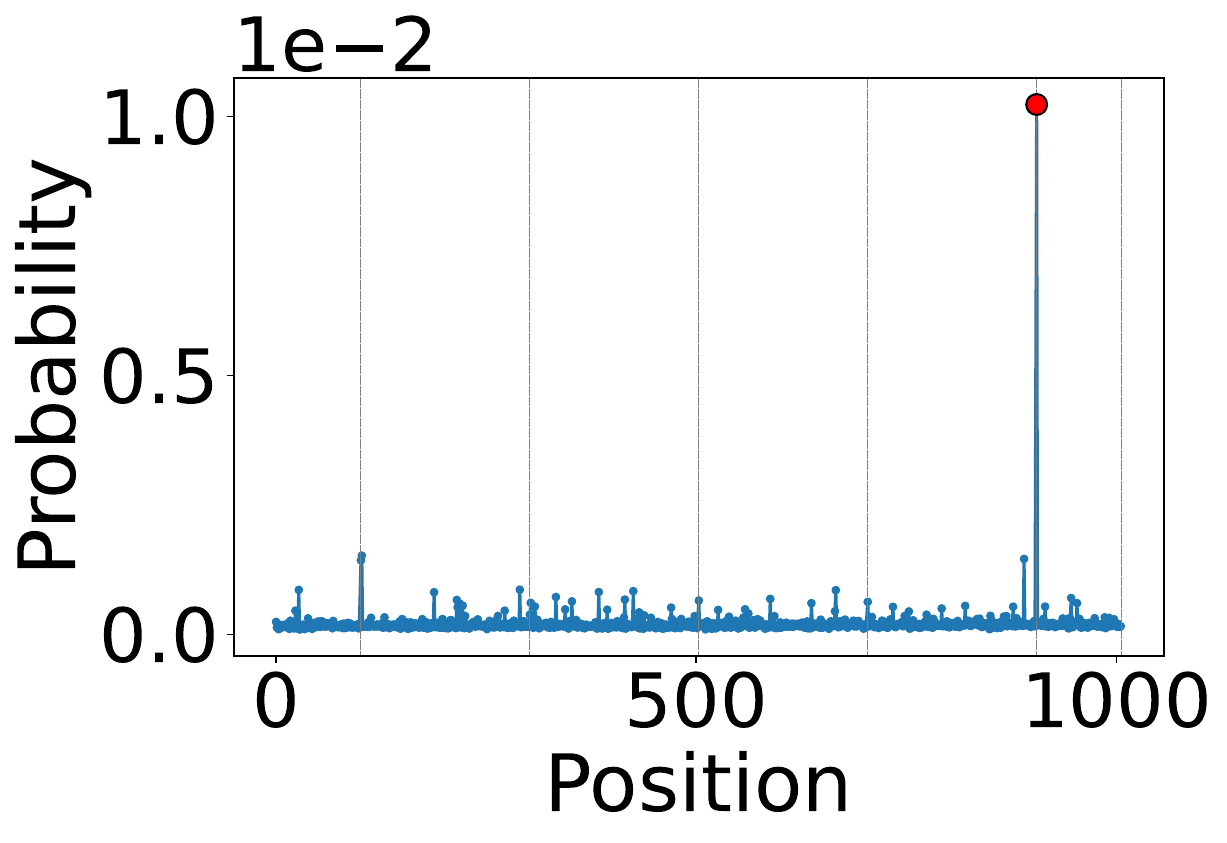} &
\FigGridCell{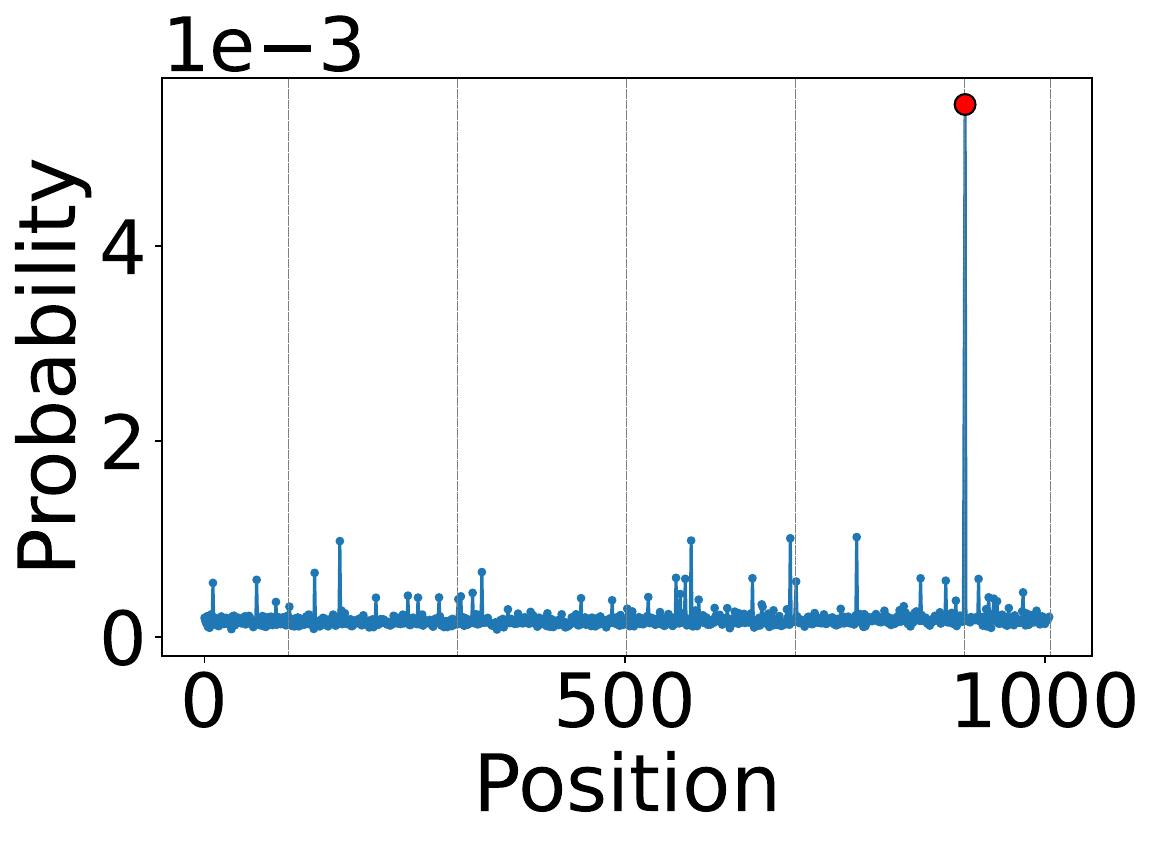} \\
\bottomrule
\end{tabular}
\endgroup
\caption{Recurrent-Gemma ablation effect (Exp. 2). Episodic retrieval probability after ablating Induction-related or Random channels. Columns show percentage of ablated channels.}
\label{fig:exp2_r-gemma_ablation}
\end{figure*}

\end{document}